%% file: main.tex
\documentclass[thnuscript,10pt,number,sort&compress]{elsarticle}

\include{preamble}

\begin{document}
	
	\input{IN_abstract}

	{\footnotesize
		\setlength{\parskip}{0.1em}
		\linespread{0.1}
		\tableofcontents
		}
	
	\section{Introduction}\label{sec:intro}
	\input{IN_introduction}

    \section{Neural PDEs and neural operators}\label{sec:intro:piml}
    \input{IN_preliminaries}
    
    \section{Modeling total uncertainty}\label{sec:uqt:pre}
    \input{IN_UQ_modeling}

    \section{Methods for uncertainty quantification}\label{sec:uqt}	
    \input{IN_UQ_methods}

    \section{Accuracy and uncertainty quality evaluation and improvement}\label{sec:eval}
    \input{IN_UQ_evaluation}

	\section{Comparative study}\label{sec:comp}
	
	\input{IN_comp_intro}
	
	\subsection{Discontinuous function approximation}\label{sec:comp:func}
	\input{IN_function_example}

	\subsection{Mixed deterministic PDE: Nonlinear time-dependent diffusion-reaction equation}\label{sec:comp:pinns}
	\input{IN_mixed_PINN}	
	
	\newpage
	
	\subsection{Mixed stochastic elliptic equation}\label{sec:comp:stochastic}
	\input{IN_stochastic}

	\subsection{Forward deterministic PDE: Nonlinear static diffusion-reaction equation}\label{sec:comp:pinns:forw}
	\input{IN_forward_PINN}

	\subsection{Operator learning and out-of-distribution data detection: 2D flow in heterogeneous porous media}\label{sec:comp:don}
	\input{IN_DON}
	
	\subsection{Performance versus cost}\label{sec:comp:accvscost}
	\input{IN_accVScost}

	\section{Summary}\label{sec:discussion}
	\input{IN_discussion}

    \phantomsection 
    \addcontentsline{toc}{section}{Acknowledgment} 
    \section*{Acknowledgments}
    
    This work was supported by OSD/AFOSR MURI grant FA9550-20-1-0358 and
    DOE PhILMs project no. DE-SC0019453. The authors would like to thank Dr. Khemraj Shukla of Brown University for reviewing parts of the paper and providing useful comments. 
    
    \phantomsection 
    \addcontentsline{toc}{section}{References} 
    \bibliographystyle{elsarticle-num-names}
    \bibliography{main}
	
	\numberwithin{equation}{section}
	\numberwithin{figure}{section}
	\numberwithin{table}{section}
	\addtocontents{toc}{\protect\setcounter{tocdepth}{1}}
	
	\begin{appendices}
	\newpage
	\clearpage
	\input{appendix/IN_app_modeling}
	\clearpage
	
	\newpage
	\input{appendix/IN_app_methods}
	\clearpage
	
	\newpage
	\input{appendix/IN_app_eval}
	\clearpage
	
	\newpage
	\input{appendix/IN_app_results}

	\end{appendices}

\end{document}

%% file: preamble.tex
\usepackage{csquotes,xpatch}

\usepackage{graphicx}
\usepackage{epstopdf} 
\usepackage{calrsfs}
\usepackage{physics}
\usepackage{mathtools}  
\usepackage{amsmath}
\usepackage{amssymb}
\usepackage{tabulary}
\usepackage{booktabs}

\usepackage{geometry}
\usepackage{blindtext}
\usepackage{ragged2e}
\usepackage{float}

\usepackage{epstopdf}
\usepackage{empheq} 

\usepackage{array}

\usepackage{graphics}
\graphicspath{ {figures/} }

\usepackage[labelfont=bf,justification=justified,singlelinecheck=false]{caption}
\usepackage{array}
\usepackage{longtable}
\usepackage[dvipsnames]{xcolor}

\usepackage{comment}

\usepackage[shortlabels]{enumitem}

\usepackage{wrapfig}

\usepackage{subcaption}
\usepackage{bbm}
\usepackage{tabularx}

\usepackage{indentfirst}

\usepackage{subcaption}

\usepackage[english]{babel}
\usepackage{blindtext}
\usepackage{amsmath,amsthm}
\usepackage{rotating}

\DeclareMathAlphabet{\pazocal}{OMS}{zplm}{m}{n}
\newcommand{\cH}{\pazocal{H}}
\newcommand{\cN}{\pazocal{N}}

\newcommand{\cD}{\pazocal{D}}

\newcommand{\cL}{\pazocal{L}}

\usepackage[ruled,vlined]{algorithm2e}
\SetKwComment{Comment}{$\triangleright$\ }{}

\usepackage{geometry}
\usepackage{marginnote}

\usepackage{dsfont}
\usepackage{diagbox}

\usepackage{multirow} 
\usepackage{multicol}

\usepackage[titletoc]{appendix}

\usepackage{pdflscape}
\usepackage{fancyhdr} 

\fancypagestyle{mylandscape}{
\fancyhf{} 
\fancyfoot{
\makebox[\textwidth][r]{
  \rlap{\hspace{.75cm}
    \smash{
      \raisebox{4.87in}{
        \rotatebox{90}{\thepage}}}}}}
}

\usepackage[hyperfootnotes=false]{hyperref}

\definecolor{mycolor}{RGB}{207,42,40}
\hypersetup{
	colorlinks
}

\AtBeginDocument{\hypersetup{citecolor=mycolor, linkcolor = mycolor}}
\usepackage{etoc}

\makeatletter
\def\ps@pprintTitle{%
  \let\@oddhead\@empty
  \let\@evenhead\@empty
  \let\@oddfoot\@empty
  \let\@evenfoot\@oddfoot
}
\makeatother

%% file: IN_abstract.tex
	\begin{frontmatter}
		
		\title{Uncertainty Quantification in Scientific Machine Learning: \\
		Methods, Metrics, and Comparisons}
		
		\author[brown]{Apostolos~F~Psaros \corref{equal}}
		\author[brown]{Xuhui~Meng \corref{equal}}
		\author[brown]{Zongren~Zou}
		\author[ling]{Ling~Guo}
		\author[brown,PNNL]{George~Em~Karniadakis\corref{email}}
		\address[brown]{Division of Applied Mathematics, Brown University, Providence, RI 02906, USA}
		\address[ling]{Department of Mathematics, Shanghai Normal University, Shanghai, China}
        \address[PNNL]{Pacific Northwest National Laboratory, Richland, WA 99354, USA}
		
		\cortext[equal]{The first two authors contributed equally to this work.}
		\cortext[email]{Corresponding Author}\ead{george_karniadakis@brown.edu}
		
		\begin{abstract}
			Neural networks (NNs) are currently changing the computational paradigm on how to combine data with mathematical laws in physics and engineering in a profound way, tackling challenging inverse and ill-posed problems not solvable with traditional methods. However, quantifying errors and uncertainties in NN-based inference is more complicated than in traditional methods.
			This is because in addition to aleatoric uncertainty associated with noisy data, there is also uncertainty due to limited data, but also due to NN hyperparameters, overparametrization, optimization and sampling errors as well as model misspecification. Although there are some recent works on uncertainty quantification (UQ) in NNs, there is no systematic investigation of suitable methods towards quantifying the {\em total uncertainty} effectively and efficiently even for function approximation, and  there is even less work on solving partial differential equations and learning operator mappings between infinite-dimensional function spaces using NNs.
        	In this work, we present a comprehensive framework that includes uncertainty modeling, new and existing solution methods, as well as evaluation metrics and post-hoc improvement approaches. To demonstrate the applicability and reliability of our framework, we present an extensive comparative study in which various methods are tested on prototype problems, including problems with mixed input-output data, and stochastic problems in high dimensions. In the Appendix, we include a comprehensive description of all the UQ methods employed, which we will make available as open-source library of all codes included in this framework. 					
		\end{abstract}
		
		\begin{keyword}
			scientific machine learning, stochastic partial differential equations, uncertainty quantification, physics-informed neural networks, neural operator learning, Bayesian framework
		\end{keyword}
		
	\end{frontmatter}


%% file: IN_introduction.tex
\subsection{Motivation and scope of the paper}

Accurate modeling and prediction of the dynamic response of physical systems remains an open scientific problem, despite many successful research efforts in the past decades.  
Persistent challenges pertain, indicatively, to solving ill-posed problems, e.g., with unknown parameters and boundary conditions, to assimilating noisy and/or gappy data, and to reducing computational cost, especially for high-dimensional systems. 
In this regard, physics-informed and more generally scientific machine learning (SciML) are emerging interdisciplinary areas of research, which can offer effective new tools for addressing some of the aforementioned challenges; see \cite{karniadakis2021physicsinformed} for a comprehensive review as well as \cite{alber2019integrating,willard2021integrating}.

In general, machine learning, and especially deep learning which uses neural networks (NNs), can deal with massive amounts of data and make accurate predictions fast; see, e.g., \cite{goodfellow2016deep, higham2019deep}.
SciML includes a variety of techniques aiming at seamlessly combining observational data with available physical models. They can be divided into three main categories: (1) the ones that seek to emulate physical systems by utilizing large data amounts, such as the deep operator network (DeepONet) developed by \citet{lu2021learning} (see also \cite{goswami2021physicsinformed,wang2021learning} for different versions of DeepONet and \cite{lanthaler2021error} for theoretical results, as well as \cite{li2021fourier} for a different approach termed Fourier neural operator); (2) the ones that encode physics into the NN architecture, such as the architectures developed by \citet{darbon2021neural}; and (3) the ones that add physical soft constraints in the NN optimization process, such as the physics-informed neural network (PINN) developed by \citet{raissi2019physicsinformed} (see also \cite{lagaris1998artificial,sirignano2018dgm,pang2019fpinns,jagtap2020extended,meng2020ppinn,kharazmi2021hpvpinns} for different versions of PINN and \cite{shin2020convergence,mishra2021estimates} for theoretical results). 
These techniques flourish, in principle, where the applicability of conventional solvers diminishes.
Specifically, they can address mixed problems when only partial observational data and information regarding the underlying physical system are available (e.g., \cite{raissi2019physicsinformed}); it is straightforward to incorporate noisy and multi-fidelity data (e.g., \cite{meng2020composite,meng2021multifidelity}); they can provide physically consistent predictions even for extrapolatory tasks (e.g., \cite{yang2021bpinns}); they do not require computationally expensive mesh generation (e.g., \cite{raissi2019physicsinformed}); and open-source software is available and continuously extended (e.g., \cite{lu2021deepxde,hennigh2021nvidia}). 

\begin{figure}[!ht]
	\centering
	\includegraphics[width=.35\linewidth]{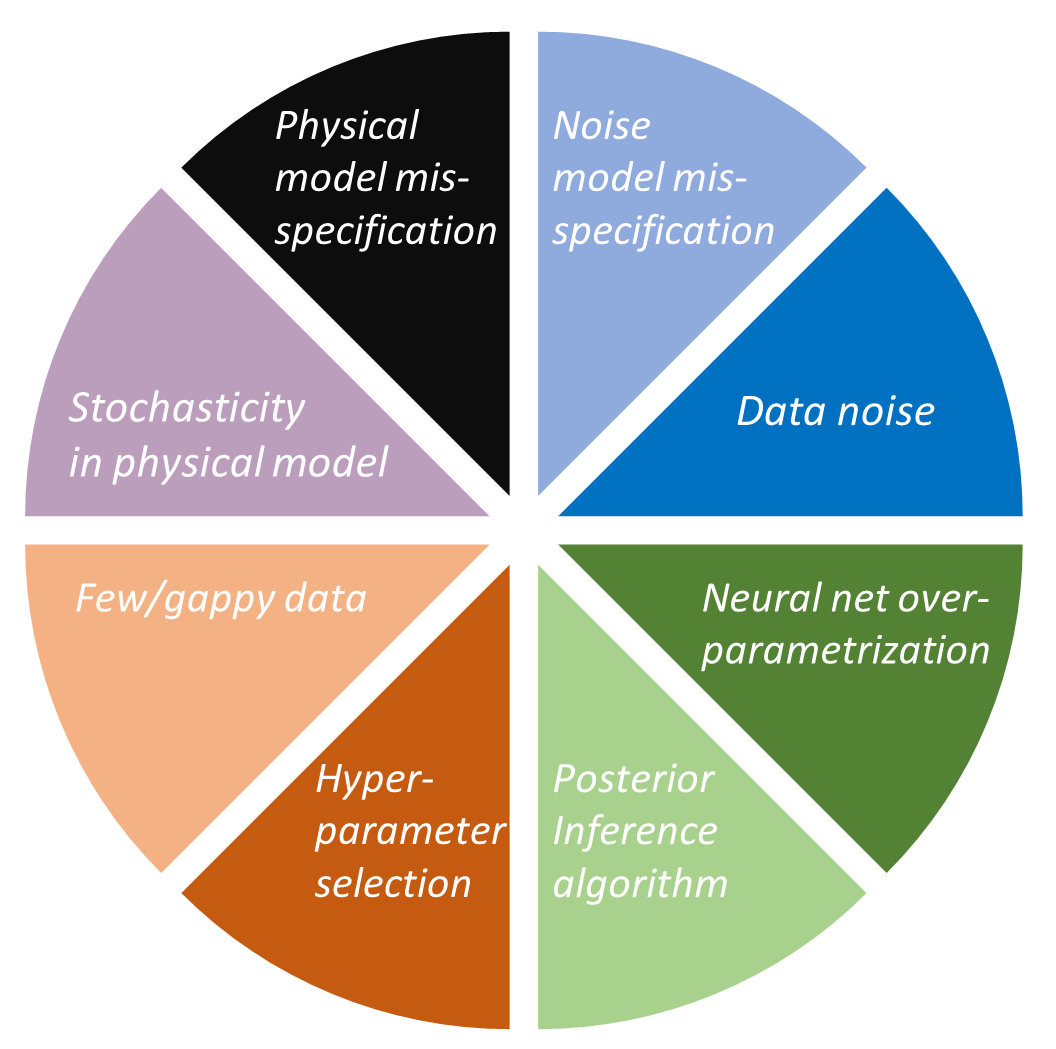}
	\caption{A qualitative breakdown of {\em total} uncertainty describing the contributions from data (noisy, gappy); physical models (misspecification, stochasticity); neural networks (architecture, hyperparameters, overparametrization); and posterior inference. 
	Aleatoric uncertainty is due to noisy data and cannot be reduced.
	Epistemic uncertainty is due to noisy and limited data, as well as neural network overparametrization. 
	Note that all uncertainties in this figure are either modeled, addressed, or discussed in this work, except for model error due to physical model misspecification.	 	
	}
	\label{fig:intro:pie}
\end{figure}

Quantifying prediction uncertainty associated with, for example, noisy and limited data as well as NN overparametrization, is paramount for deep learning to be reliably used in critical applications involving physical and biological systems.
The most successful family of UQ methods so far in deep learning has been based on the Bayesian framework \cite{neal1993probabilistic,mackay1995bayesian,neal1995bayesian,barber1998ensemble,lampinen2001bayesian,gelman2015bayesian,graves2011practical,neal2012mcmc,hoffman2013stochastic,ma2015complete,gal2016uncertainty,wilson2020case,wilson2020bayesian}. 
Alternative methods are based, indicatively, on ensembles of NN optimization iterates or independently trained NNs \cite{lakshminarayanan2017simple,huang2017snapshot,mandt2017stochastic,garipov2018loss,khan2018fast,pearce2018highquality,maddox2019simple,fort2020deep,he2020bayesian,franzese2020isotropic,kessler2020practical,rahaman2021uncertainty}, as well as on the evidential framework \cite{malinin2018predictive,sensoy2018evidential,amini2020deep,charpentier2020posterior,kopetzki2020evaluating,malinin2020regression,charpentier2021natural,meinert2021multivariate,ulmer2021survey}. 
Although Bayesian methods and ensembles are thoroughly discussed in this paper, the interested reader is also directed to the recent review studies in \cite{polson2017deep,kabir2018neural,caldeira2020deeply,goan2020bayesian,jospin2020handson,loquercio2020general,stahl2020evaluation,wang2020survey,gawlikowski2021survey,abdar2021review,hullermeier2021aleatoric,nado2021uncertainty,zhou2021survey} for more information.    
Clearly, in the context of SciML, which may involve differential equations with unknown or uncertain terms and parameters, UQ becomes an even more demanding task; see Fig.~\ref{fig:intro:all:problems} for a sketch of problems of interest and Fig.~\ref{fig:intro:pie} for a summary of considered uncertainty sources.
Despite some recent progress towards combining UQ with either data-driven or physics-informed optimization objectives \cite{tripathy2018deep,grigo2019physicsaware,yang2019conditional,olivier2021bayesian,patel2021ganbased,xu2021solving,oleary2021stochastic,yeo2021variational,yang2019highlyscalable,atkinson2020bayesian,geneva2020multifidelity,huang2020learning,karumuri2020simulatorfree,yang2020physicsinformed,yang2020bayesian,bajaj2021robust,daw2021pidgan,fuhg2021interval,gao2021wasserstein,lin2021accelerated,meng2021multifidelity,molnar2021flow,wang2021efficient,yang2021bpinns,tsilifis2021inverse,guo2021normalizing,hall2021ginns}, UQ methods are still scarcely utilized within SciML. 
One potential reason is that UQ methods are under-utilized even within the broader deep learning community; it is thus a developing field that is not universally trusted and understood yet.
In addition, physical considerations, although of utmost importance in many problems, make the SciML techniques more complicated and computationally expensive than their standard deep learning counterparts, and further reinforce the hesitation of the machine learning community towards UQ. 

Motivated by the aforementioned open issues, in this paper we present a unified and comprehensive framework of UQ in SciML with emphasis placed on techniques founded on the PINN and DeepONet underlying frameworks and their extensions.
The framework can be combined with approaches other than PINN and DeepONet, such as the Fourier neural operator developed in \cite{li2021fourier}.
Specifically, (1) we consider various classes of problems and summarize pertinent solution methods; (2) provide a detailed deep learning-based uncertainty modeling procedure; (3) review and integrate into SciML the most widely used UQ methods; (4) review and integrate uncertainty evaluation metrics as well as post-training improvement approaches; and (5) present an extensive comparative study, in which various uncertainty modeling and quantification methods are tested on problems of interest in computational science and engineering. 
Overall, with the present paper we intend to lay the groundwork for UQ to be routinely utilized in SciML, with the expectation that more focused studies will be undertaken in the future by other researchers. 

\subsection{Problem formulation}\label{sec:intro:problem:form}

Consider a nonlinear partial differential equation (PDE) describing the dynamics of a physical system as follows:
\begin{subequations}\label{eq:intro:piml:pinn:pde}
	\begin{align}
		\pazocal{F}_{\lambda, \xi}[u(x; \xi)] &= f(x; \xi) \text{, } x \in  \Omega\text{, } \xi \in \Xi, 
		\label{eq:intro:piml:pinn:pde:a}\\
		\pazocal{B}_{\lambda, \xi}[u(x; \xi)] & = b(x; \xi) \text{, } x \in \Gamma, 
		\label{eq:intro:piml:pinn:pde:b}
	\end{align}
\end{subequations}
where $x$ is the $D_x$-dimensional space-time coordinate, $\Omega$ is a bounded domain with boundary $\Gamma$, $\xi$ is a random event in a probability space $\Xi$, while $f(x; \xi)$ and $u(x; \xi)$ represent the $D_u$-dimensional source term and sought solution evaluated at $(x, \xi)$, respectively. 
Further, $\pazocal{F}_{\lambda, \xi}$ is a general differential operator; $\pazocal{B}_{\lambda, \xi}$ is a boundary/initial condition operator acting on the domain boundary $\Gamma$; and $b$ is a boundary/initial condition term.
Furthermore, $\lambda(x; \xi)$, with $x \in \Omega_{\lambda} \subset \Omega$, denotes the problem parameters.

In this regard, we consider four problem scenarios, as summarized in Table~\ref{tab:problem:form} and Fig.~\ref{fig:intro:all:problems}, and delineated below.
We refer to problems where the operators $\pazocal{F}_{\lambda, \xi}$ and $\pazocal{B}_{\lambda, \xi}$ are known, i.e., the differential operators that define them are specified, as \textit{neural PDEs}, because we use neural networks (NNs) throughout the present work to solve them.
We refer to problems involving unknown operators $\pazocal{F}_{\lambda, \xi}$ and $\pazocal{B}_{\lambda, \xi}$ as \textit{neural operators}, because we use NNs to learn them. 

The first scenario pertains to a ``forward'' deterministic PDE problem, where $\pazocal{F}_{\lambda, \xi}$ and $\pazocal{B}_{\lambda, \xi}$ are deterministic and known. 
That is, $\xi$ is fixed and dropped for simplicity.
The problem parameters $\lambda(x)$ are also known and $u(x)$ is unknown.
Given noisy data of the functions $f$ and $b$ sampled at random finite locations in $\Omega$ and $\Gamma$, respectively, the objective of this problem is to obtain the solution $u(x)$ at every $x \in \Omega$.
The dataset in this case is expressed as $\cD = \{\cD_f, \cD_b\}$, where $\cD_f = \{x_i, f_i\}_{i=1}^{N_f}$, $\cD_b = \{x_i, b_i\}_{i=1}^{N_b}$, $f_i = f(x_i)$, and $b_i = b(x_i)$.
Clearly, although we use the same symbol $x_i$ in $\cD_f$ and $\cD_b$, the data locations in the space domains of $f$ and $b$ are in general different, and this holds for all the datasets in this paper.

\input{IN_table_problem}

The second scenario pertains to a ``mixed'' deterministic problem, where $\pazocal{F}_{\lambda, \xi}$, $\pazocal{B}_{\lambda, \xi}$ are deterministic and known, while $\lambda(x), u(x)$ are partially unknown. 
Given noisy data of $u$ and $\lambda$ sampled at random finite locations in $\Omega$ and $\Omega_{\lambda}$, respectively, as well as noisy data of $f$ and $b$, the objective of this problem is to obtain the solution $u(x)$ at every $x \in \Omega$ and the problem parameters $\lambda(x)$ at every $x \in \Omega_{\lambda}$.
The dataset in this case is expressed as $\cD = \{\cD_f, \cD_b, \cD_u, \cD_{\lambda}\}$, where $\cD_u = \{x_i, u_i\}_{i=1}^{N_u}$, $\cD_{\lambda} = \{x_i, \lambda_i\}_{i=1}^{N_{\lambda}}$, $u_i = u(x_i)$, $\lambda_i = \lambda(x_i)$, and $\cD_f, \cD_b$ are given as above.

The third scenario pertains to a mixed stochastic PDE (SPDE) problem, where $\pazocal{F}_{\lambda, \xi}$, $\pazocal{B}_{\lambda, \xi}$ are stochastic and known, while $\lambda(x; \xi), u(x; \xi)$ are partially unknown. 
The available data consists of $N$ noisy paired realizations of $f, b, u$, and $\lambda$, each one associated with a different $\xi \in \Xi$.
The realizations are sampled at random finite locations in the corresponding $x$ domains.
Specifically, the dataset in this case is expressed as $\cD = \{\cD_f, \cD_b, \cD_u, \cD_{\lambda}\}$, where $\cD_f = \{F_i\}_{i=1}^{N}$, $\cD_b = \{B_i\}_{i=1}^{N}$, $\cD_u = \{U_i\}_{i=1}^{N}$, and $\cD_{\lambda} = \{\Lambda_i\}_{i=1}^{N}$.
The corresponding sub-datasets for each value of $i$ are given as $F_i = \{x_j^{(i)}, f_j^{(i)}\}_{j=1}^{N_f}$, $B_i = \{x_j^{(i)}, b_j^{(i)}\}_{j=1}^{N_b}$, $U_i = \{x_j^{(i)}, u_j^{(i)}\}_{j=1}^{N_u}$, and $\Lambda_i = \{x_j^{(i)}, \lambda_j^{(i)}\}_{j=1}^{N_{\lambda}}$.
The objective of this problem is to obtain the statistics, distribution and/or moments, of $u(x)$ at every $x \in \Omega$ and of $\lambda(x)$ at every $x \in \Omega_{\lambda}$.

The fourth scenario pertains to an operator learning problem, where $\pazocal{F}_{\lambda, \xi}$, $\pazocal{B}_{\lambda, \xi}$ are stochastic and unknown, while $\lambda(x; \xi), u(x; \xi)$ are partially unknown. 
The available data consists of paired realizations of $f, b, u$, and $\lambda$, each one associated with a different $\xi \in \Xi$.
The realizations are considered noisy in general for $u$, and clean for $f, b$, and $\lambda$; see also discussion in Section~\ref{sec:uqt:pre:don}.
Further, the objective of this problem is to learn the operator mapping defined by Eq.~\eqref{eq:intro:piml:pinn:pde}.
We refer to this procedure as \textit{pre-training} and the required dataset is the same as in the mixed stochastic problem, except with clean data for $f, b$, and $\lambda$.
For employing the learned operator during \textit{inference} for an unseen $\xi' \in \Xi$, we are given noisy and limited data of $f, b, u$, and $\lambda$, and the objective is to obtain $f, b, u$, and $\lambda$ at every $x$ in the corresponding domains.  
The inference dataset in this case is expressed as $\cD' = \{\cD_f', \cD_b', \cD_u', \cD_{\lambda}'\}$, where $\cD_f' = \{x_i, f_i\}_{i=1}^{N_f'}$, $\cD_b' = \{x_i, b_i\}_{i=1}^{N_b'}$, $\cD_u' = \{x_i, u_i\}_{i=1}^{N_u'}$, and $\cD_{\lambda}' = \{x_i, \lambda_i\}_{i=1}^{N_{\lambda}'}$.
Finally, note that a special case of inference data is where $f, b$ and $\lambda$ data are available at the complete set of locations used in pre-training, and the only sought outcome is the solution $u$.
See also Table~\ref{tab:comp:don:problems} in the comparative study for an overview of problem cases specific to operator learning.

\subsection{Novel contributions of this work}

In addition to the review and the comparative study we perform in this work, we also present several original contributions summarized below:

\begin{enumerate}
    \item We test and integrate into physics-informed neural networks, neural operators, and SPDE solvers various methods for posterior inference, prior learning, data noise modeling, as well as for post-training calibration.
    \item We demonstrate how to solve  function approximation problems with heteroscedastic noise given historical data by using functional priors.
    \item We propose a new technique for solving the deterministic forward PDE problem by combining Gaussian process regression with generative adversarial networks and compare with existing methods.
    \item We solve mixed PDE problems with heteroscedastic noise in the source term, the problem parameters, and the solution data.
    \item We solve mixed SPDE problems given noisy stochastic realizations, and we propose a new NN architecture for quantifying uncertainty using polynomial chaos.
    \item We demonstrate how to deal with noisy and incomplete inference data given a pre-trained neural operator on clean data.
    \item We propose methods to detect out-of-distribution data in neural operators, which is critical for risk-related applications.
    \item We present
    a formulation towards a unified UQ framework that addresses diverse problems in scientific machine learning, by seamlessly combining physics with new and historical data, which could be contaminated with various types of noise.
\end{enumerate}

\subsection{Organization of the paper}

\begin{figure}[!ht]
	\centering
	\includegraphics[width=.9\linewidth]{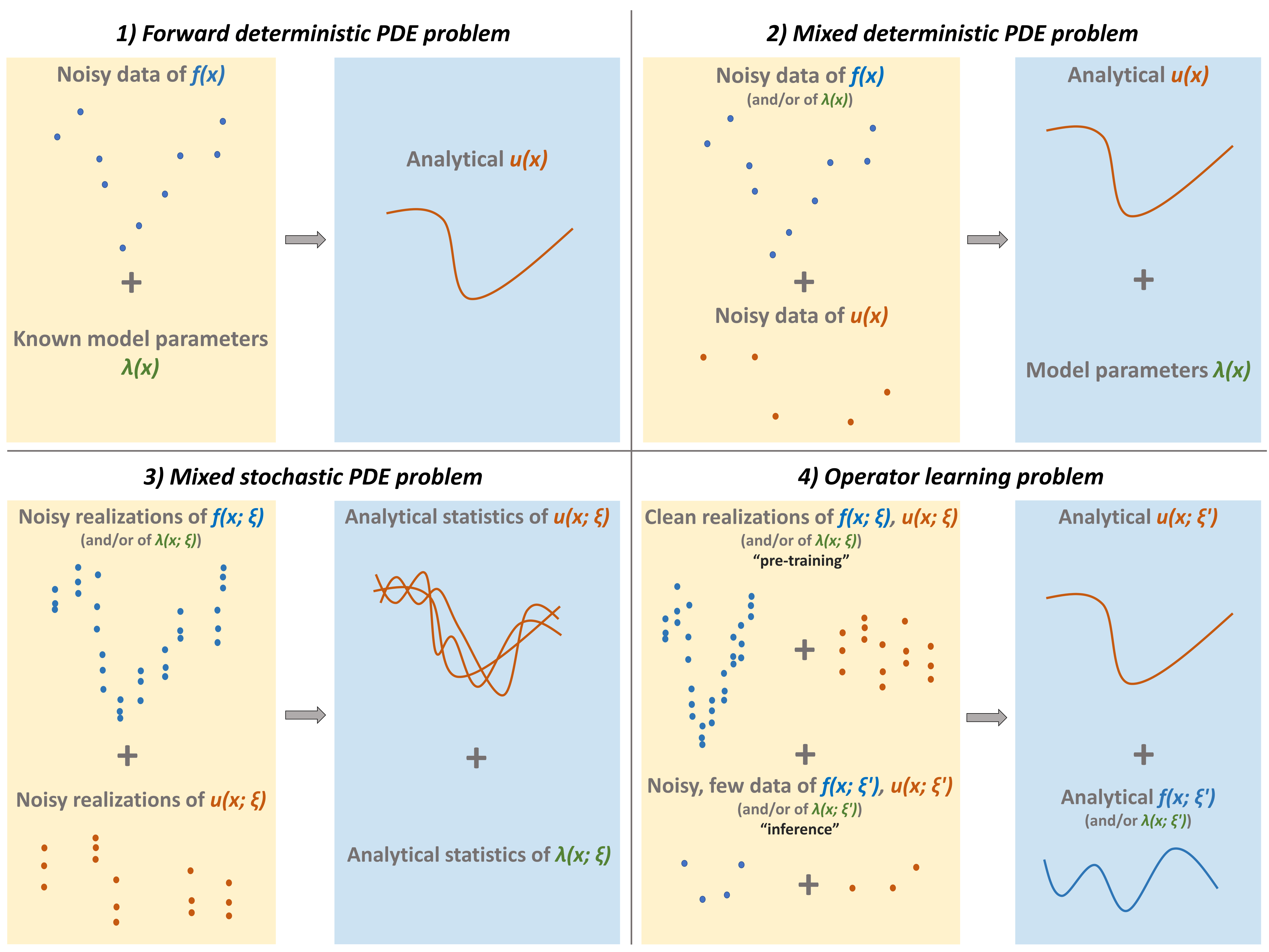}
	\caption{Inputs (orange panels) - Outputs (blue panels) of the problems described in Table~\ref{tab:problem:form} corresponding to equation $\pazocal{F}_{\lambda, \xi}[u(x; \xi)] = f(x; \xi) \text{, } x \in  \Omega\text{, } \xi \in \Xi \ \text{ and } \ 
		\pazocal{B}_{\lambda, \xi}[u(x; \xi)] = b(x; \xi) \text{, } x \in \Gamma$.
	Problems 1-3 relate to \textit{neural (S)PDEs}, whereas problem 4 relates to \textit{neural operators}.
	Problem 1 relates to a standard ``forward'' formulation.
	Problems 2-4 to ``mixed'' formulations, where both solution and parametric partially known fields, e.g., $\lambda(x)$, are sought.
	We refer to outputs as ``analytical'' (or ``analytical statistics'') because they are expressed via a neural network for the corresponding entire continuous space-time domain.
	We refer to functions sampled at random finite locations containing \textit{aleatoric uncertainty} as ``noisy data''.
	Similarly, we refer to stochastic realizations sampled at random finite locations containing aleatoric uncertainty as ``noisy realizations''.
	Operator learning consists of two phases, namely, pre-training and inference, as explained in Section~\ref{sec:intro:problem:form}; see also Table~\ref{tab:comp:don:problems}.
	Lastly, $b$ is either known or treated similarly to $f$.
	}
	\label{fig:intro:all:problems}
\end{figure}

We organize the paper as follows.
In Section~\ref{sec:intro:piml} we briefly present the PINN and DeepONet methods, which are the building blocks of all the SciML techniques we discuss in this paper. 
In Section~\ref{sec:uqt:pre} and the corresponding Appendix~\ref{app:modeling}, we develop an uncertainty modeling framework that includes data noise modeling, review of standard NN training, predictive distributions based on posterior samples, priors and hyperpriors, addressing model misspecification, as well as uncertainty modeling in PINN and DeepONet methods. 
In Section~\ref{sec:uqt} and the corresponding Appendix~\ref{app:methods}, we present two families of posterior inference methods, namely Bayesian methods and ensembles, we present functional priors for reducing the required amount of data and physics knowledge, as well as techniques for solving stochastic differential equations. 
In Section~\ref{sec:uqt:uni}, we develop a unified model that represents all the problems of Table~\ref{tab:problem:form} and Fig.~\ref{fig:intro:all:problems} and includes all solution methods in this paper as special cases. 
In Section~\ref{sec:eval} and the corresponding Appendix~\ref{app:eval}, we propose a set of quality evaluation metrics for the discussed methods and we review and integrate three post-hoc quality improvement approaches into SciML. 
In Section~\ref{sec:comp} and the corresponding Appendix~\ref{app:comp}, we perform a comparative study including a function approximation problem as well as various problem cases of Table~\ref{tab:problem:form} and Fig.~\ref{fig:intro:all:problems}. 
Lastly, in Section~\ref{sec:discussion}, we discuss the findings of this work and future research directions. 

%% file: IN_table_problem.tex
\begin{table}[ht]
	\centering
	\footnotesize
	\begin{tabular}{r|l||r|l}
		\toprule	
		\multicolumn{4}{c}{\textbf{Considered problem scenarios corresponding to the following equation:}} \\
		\multicolumn{4}{c}{$\pazocal{F}_{\lambda, \xi}[u(x; \xi)] = f(x; \xi) \text{, } x \in  \Omega\text{, } \xi \in \Xi \ \text{ and } \ 
		\pazocal{B}_{\lambda, \xi}[u(x; \xi)] = b(x; \xi) \text{, } x \in \Gamma$} \\
		\midrule
		\multicolumn{2}{c||}{\textit{\textbf{1) Forward deterministic PDE problem}}}&\multicolumn{2}{c}{\textit{\textbf{2) Mixed deterministic PDE problem}}}\\
		\midrule
		$\pazocal{F}_{\lambda, \xi}, \pazocal{B}_{\lambda, \xi}$&deterministic (fixed $\xi$) and known&$\pazocal{F}_{\lambda, \xi}, \pazocal{B}_{\lambda, \xi}$&deterministic (fixed $\xi$) and known\\
		{\color{RoyalBlue} $f(x), b(x)$}& noisy data &{\color{RoyalBlue} $f(x), b(x)$}& noisy data \\
		{\color{Orange} $u(x)$}& unknown  &{\color{Orange} $u(x)$}& partially unknown (noisy data) \\
		{\color{ForestGreen} $\lambda(x)$}& known &{\color{ForestGreen} $\lambda(x)$}& partially unknown (noisy data) \\
		Dataset& $\cD = \{\cD_f, \cD_b\}$, where &Dataset& $\cD = \{\cD_f, \cD_b, \cD_u, \cD_{\lambda}\}$, where \\
		& $\cD_f = \{x_i, f_i\}_{i=1}^{N_f}$, $\cD_b = \{x_i, b_i\}_{i=1}^{N_b}$ &&  $\cD_f = \{x_i, f_i\}_{i=1}^{N_f}$, $\cD_b = \{x_i, b_i\}_{i=1}^{N_b}$ \\
		& &&  $\cD_u = \{x_i, u_i\}_{i=1}^{N_u}$, $\cD_{\lambda} = \{x_i, \lambda_i\}_{i=1}^{N_{\lambda}}$\\
		Objective&$u(x)$, $\forall x \in \Omega$&Objective&$u(x), \lambda(x)$, $\forall x \in \Omega, \Omega_{\lambda}$\\
		\midrule
		\midrule
    	\multicolumn{2}{c||}{\textit{\textbf{3) Mixed stochastic PDE problem}}}&\multicolumn{2}{c}{\textit{\textbf{4) Operator learning problem}}}\\
		\midrule
		$\pazocal{F}_{\lambda, \xi}, \pazocal{B}_{\lambda, \xi}$&stochastic ($\xi \in \Xi$) and known&$\pazocal{F}_{\lambda, \xi}, \pazocal{B}_{\lambda, \xi}$&stochastic ($\xi \in \Xi$) and unknown\\
		{\color{RoyalBlue} $f(x; \xi), b(x; \xi)$}& noisy realizations &{\color{RoyalBlue} $f(x; \xi), b(x; \xi)$}& partially unknown (clean realizations) \\
		{\color{Orange} $u(x; \xi)$}& partially unknown (noisy realizations)  &{\color{Orange}$u(x; \xi)$}& partially unknown (noisy realizations) \\
		{\color{ForestGreen}$\lambda(x; \xi)$}& partially unknown (noisy realizations) &{\color{ForestGreen}$\lambda(x; \xi)$}& partially unknown (clean realizations) \\
		Dataset& $\cD = \{\cD_f, \cD_b, \cD_u, \cD_{\lambda}\}$, where &Dataset& $\cD = \{\cD_f, \cD_b, \cD_u, \cD_{\lambda}\}$, where \\
		& $\cD_f = \{F_i\}_{i=1}^{N}$, $\cD_b = \{B_i\}_{i=1}^{N}$ &
		(pre-training)&  $\cD_f = \{F_i\}_{i=1}^{N}$, $\cD_b = \{B_i\}_{i=1}^{N}$ \\
		& $\cD_u = \{U_i\}_{i=1}^{N}$, $\cD_{\lambda} = \{\Lambda_i\}_{i=1}^{N}$ &&  $\cD_u = \{U_i\}_{i=1}^{N}$, $\cD_{\lambda} = \{\Lambda_i\}_{i=1}^{N}$ \\
		&$F_i = \{x_j^{(i)}, f_j^{(i)}\}_{j=1}^{N_f}$, $B_i = \{x_j^{(i)}, b_j^{(i)}\}_{j=1}^{N_b}$ && $F_i = \{x_j^{(i)}, f_j^{(i)}\}_{j=1}^{N_f}$, $B_i = \{x_j^{(i)}, b_j^{(i)}\}_{j=1}^{N_b}$\\
		&$U_i = \{x_j^{(i)}, u_j^{(i)}\}_{j=1}^{N_u}$, $\Lambda_i = \{x_j^{(i)}, \lambda_j^{(i)}\}_{j=1}^{N_{\lambda}}$ && $U_i = \{x_j^{(i)}, u_j^{(i)}\}_{j=1}^{N_u}$, $\Lambda_i = \{x_j^{(i)}, \lambda_j^{(i)}\}_{j=1}^{N_{\lambda}}$\\
		Objective&statistics of $u(x), \lambda(x)$, $\forall x \in \Omega, \Omega_{\lambda}$&Objective&$f(x; \xi'), b(x; \xi'), u(x; \xi'), \lambda(x; \xi')$,\\
		&&& $\forall x \in \Omega, \Gamma, \Omega, \Omega_{\lambda}$, given inference data   \\
		&&Inference data& $\cD' = \{\cD_f', \cD_b', \cD_u', \cD_{\lambda}'\}$, where \\
		& &&  $\cD_f' = \{x_i, f_i\}_{i=1}^{N_f'}$, $\cD_b' = \{x_i, b_i\}_{i=1}^{N_b'}$ \\
		& &&  $\cD_u' = \{x_i, u_i\}_{i=1}^{N_u'}$, $\cD_{\lambda}' = \{x_i, \lambda_i\}_{i=1}^{N_{\lambda}'}$\\
		\bottomrule
	\end{tabular}
	\caption{
	Problem scenarios considered in this paper.
	Problems 1-3 relate to \textit{neural (S)PDEs}, whereas problem 4 relates to \textit{neural operators}.
	Problem 1 relates to a standard ``forward'' formulation.
	Problems 2-4 to ``mixed'' formulations, where both solution and parametric partially known fields, e.g., $\lambda(x)$, are sought.
	We refer to functions sampled at random finite locations containing \textit{aleatoric uncertainty} as ``noisy data''.
	Similarly, we refer to stochastic realizations sampled at random finite locations containing aleatoric uncertainty as ``noisy realizations''.
	Operator learning consists of two phases, namely, pre-training and inference, as explained in Section~\ref{sec:intro:problem:form}; see also Table~\ref{tab:comp:don:problems}.
	}
	\label{tab:problem:form}
\end{table}

%% file: IN_preliminaries.tex
In this paper, we obtain PDE solutions and operator mapping approximations that are endowed with uncertainty.
In this regard, we solve the problems of Table~\ref{tab:problem:form} and Fig.~\ref{fig:intro:all:problems} by combining available data with physical knowledge.
In this section, we summarize the vanilla PINN and DeepONet methods for solving PDEs and learning operators, respectively.
Because both methods utilize NNs, we refer to the former as neural PDEs and to the latter as neural operators.
These are the building blocks for several extensions that we discuss in Section~\ref{sec:uqt}, such as functional priors for reducing the required amount of data and physics knowledge, as well as methods for solving SPDEs. A glossary of various terms we use in this section is provided in Table~\ref{tab:glossary}.

\subsection{Solving forward and mixed PDE problems: Overview of PINN method}
\label{sec:intro:piml:pinn}

\begin{figure}[!ht]
	\centering
	\includegraphics[width=.6\linewidth]{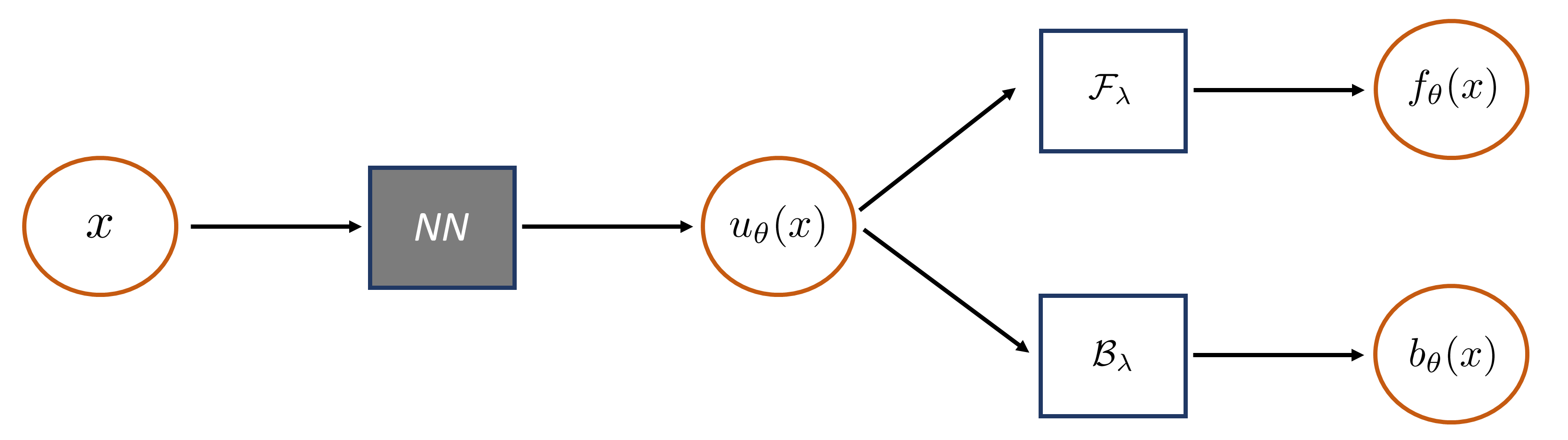}
	\caption{The PINN architecture.
		PINNs consist of three NN approximators $u_{\theta}$, $f_{\theta}$, and $b_{\theta}$ that are connected via both the operators $\pazocal{F}_{\lambda}$ and $\pazocal{B}_{\lambda}$ of Eq.~\eqref{eq:intro:piml:pinn:pde} and parameter sharing, i.e., $f_{\theta} = \pazocal{F}_{\lambda}[u_{\theta}]$ and $b_{\theta} = \pazocal{B}_{\lambda}[u_{\theta}]$, where $\theta$ are the shared parameters.
		The NN approximators $u_{\theta}$, $f_{\theta}$, and $b_{\theta}$ are trained simultaneously because of parameter sharing to fit the complete dataset.
		The model depicted in this figure can be construed as a special case of Fig.~\ref{fig:intro:all:techniques}.
	}
	\label{fig:intro:piml:pinn:arch}
\end{figure}

The PINN method, developed by \citet{raissi2019physicsinformed} (first published in arxiv in 2017), addresses the forward deterministic PDE problem by constructing a NN approximator, parametrized by $\theta$, for modeling the solution $u(x)$.
The approximator $u_{\theta}(x)$ is substituted into Eq.~\eqref{eq:intro:piml:pinn:pde} via authomatic differentiation for producing $f_{\theta} = \pazocal{F}_{\lambda}[u_{\theta}]$ and $b_{\theta} = \pazocal{B}_{\lambda}[u_{\theta}]$. 
PINN modeling for forward problems is illustrated in Fig.~\ref{fig:intro:piml:pinn:arch}.
Further, the NN parameters $\theta$ are optimized for fitting the dataset $\cD = \{\cD_f, \cD_b\}$ using the mean squared error (MSE) loss.
For solving the mixed deterministic PDE problem, an additional NN can be constructed for modeling $\lambda(x)$.
For notation simplicity, the parameters of both NNs are denoted by $\theta$. 
Finally, the optimization problem for obtaining $\theta$ in the mixed problem scenario with the dataset $\cD = \{\cD_f, \cD_b, \cD_u, \cD_{\lambda}\}$ is cast as
\begin{multline}\label{eq:intro:piml:pinn:loss:1}
	\hat{\theta} = \underset{\theta}{\mathrm{argmin}}~\pazocal{L}(\theta) \text{, where } 
	\pazocal{L}(\theta) = \frac{w_f}{N_f}\sum_{i=1}^{N_f}
		||f_{\theta}(x_i) - f_i||_2^2 +
	 \frac{w_b}{N_b}\sum_{i=1}^{N_b}
		||b_{\theta}(x_i)-b_i||_2^2 \\
	+  \frac{w_u}{N_u}\sum_{i=1}^{N_u}
		||u_{\theta}(x_i) - u_i||_2^2
	+ \frac{w_{\lambda}}{N_{\lambda}}\sum_{i=1}^{N_{\lambda}}
		||\lambda_{\theta}(x_i) - \lambda_i||_2^2.
\end{multline}
In this equation, $\{w_f, w_b, w_u, w_{\lambda}\}$ are objective function weights for balancing the various terms in Eq.~\eqref{eq:intro:piml:pinn:loss:1}.
The interested reader is directed to \cite{wang2020understanding,mcclenny2020selfadaptive,psaros2021metalearning} for online and offline adaptive weight techniques.
Following determination of $\hat{\theta}$, $u_{\hat{\theta}}$ and $\lambda_{\hat{\theta}}$ can be evaluated at any $x$ in $\Omega$, $\Omega_{\lambda}$, respectively.

\subsection{Learning operator mappings: Overview of DeepONet method}\label{sec:intro:piml:don}

\begin{figure}[!ht]
	\centering
	\includegraphics[width=.6\linewidth]{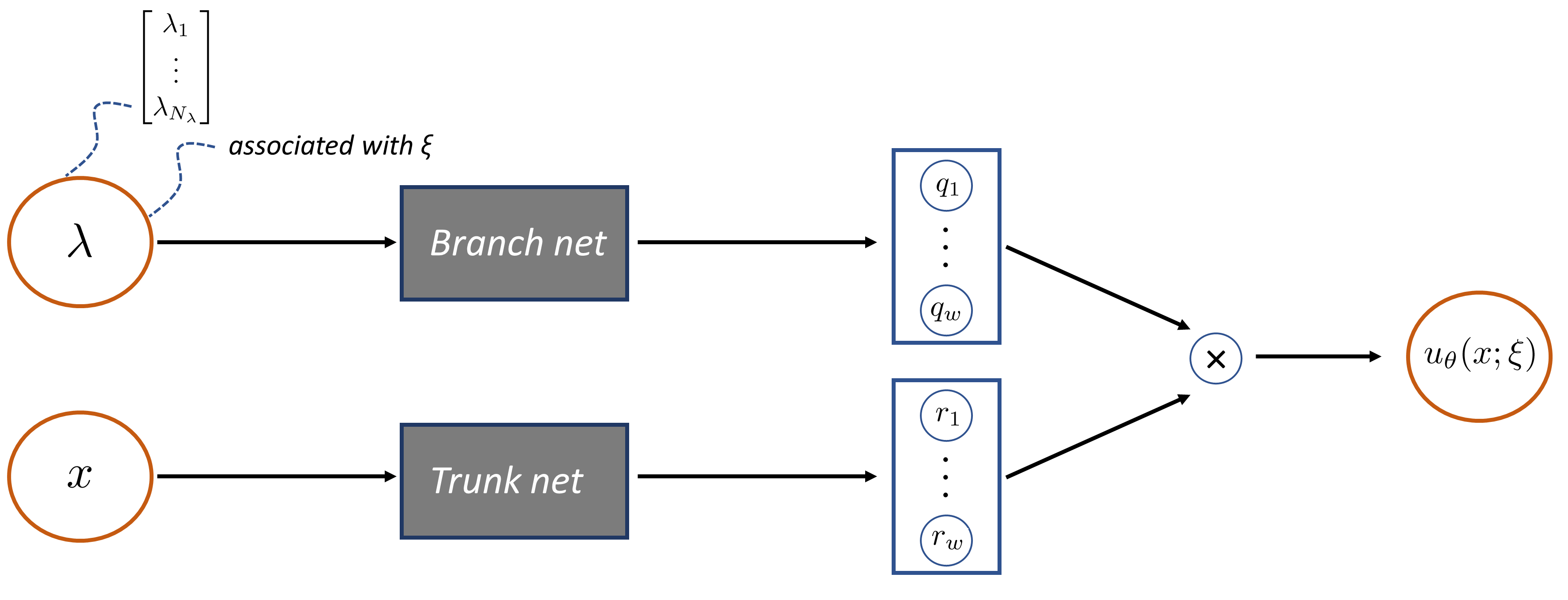}
	\caption{The DeepONet architecture corresponding to output $u_{\theta}(x; \xi)$, where $\xi$ is associated with an input function $\lambda(x; \xi)$.
	It consists of two sub-networks, one for the input $x$ where the solution $u$ is evaluated, called trunk net, and one for the input $\lambda$ (a vector of size $N_{\lambda}$), called branch net.
	The vectors $[q_1, \dots, q_w]$ and $[r_1, \dots, r_w]$ denote the outputs of the branch and trunk nets, respectively, and ``$\times$'' denotes their inner product.
	The model depicted in this figure can be construed as a special case of Fig.~\ref{fig:intro:all:techniques}.
	}
	\label{fig:intro:piml:don:arch}
\end{figure}

The DeepONet method, developed by \citet{lu2021learning} (first published in arxiv in 2019), addresses the operator learning problem by constructing a NN approximator, parametrized by $\theta$, for modeling the solution $u(x; \xi)$.
In the following and without loss of generality, we consider $f$ and $b$ in Eq.~\eqref{eq:intro:piml:pinn:pde} as known. 
Further, each value of $\xi \in \Xi$ in Eq.~\eqref{eq:intro:piml:pinn:pde} corresponds to a different $\lambda(x; \xi)$ function, and thus a different solution $u(x; \xi)$. 
The NN $u_{\theta}(x; \xi)$ takes as input the $\lambda_j^{(i)}$ values from the sub-dataset $\Lambda_i = \{x_j^{(i)}, \lambda_j^{(i)}\}_{j=1}^{N_{\lambda}}$.
The locations $\{x_j^{(i)}\}_{j=1}^{N_{\lambda}}$ are considered the same for all $i$ values, without loss of generality.
Inspired by the universal approximation theorem for operators, the DeepONet architecture consists of two sub-networks, one for the input $x$, called trunk net, and one for the input $\lambda$, called branch net.
Considering a one-dimensional output $u_{\theta}(x; \xi)$, i.e., $D_u=1$, each of the two networks produces a $w$-dimensional output.
The two outputs are merged by computing their inner product for producing the final prediction $u_{\theta}(x; \xi)$; see Fig.~\ref{fig:intro:piml:don:arch}.
For learning the operator mapping from $\lambda$ to $u$ with the DeepONet method, the minimization problem with the MSE loss
\begin{equation}\label{eq:intro:piml:don:min}
	\hat{\theta} =  \underset{\theta}{\mathrm{argmin}}~\pazocal{L}(\theta) \text{, where } 
	\pazocal{L}(\theta) = \frac{1}{NN_u}\sum_{i=1}^{N}\sum_{j=1}^{N_u}
	||u_{\theta}(x_j^{(i)}; \xi_i) - u_j^{(i)}||_2^2    
\end{equation}	
is solved. 
We refer to this procedure as pre-training or just training, depending on whether the inference data is limited and noisy, and requires further processing, or clean and can be directly used for inference.
Specifically, during inference for a new value $\xi' \in \Xi$, the vanilla DeepONet considers a clean dataset $\cD_{\lambda}' = \{x_i, \lambda_i\}_{i=1}^{N_{\lambda}'}$. 
If the location set $\{x_i\}_{i=1}^{N_{\lambda}'}$ is not the same as the locations $\{x_j^{(i)}\}_{j=1}^{N_{\lambda}}$ used in pre-training, interpolation must be performed because the locations of the input $\lambda$ to DeepONet are fixed; see Table~\ref{tab:comp:don:problems} for an overview of different cases. 
The corresponding solution $u(x; \xi')$ can be obtained for each required $x \in \Omega$ by performing a forward DeepONet pass.
Finally, note that in addition to supervised learning using data as described above, DeepONet can alternatively be trained by employing only the PINN loss during training.
As shown in \cite{goswami2021physicsinformed}, the most effective training is the hybrid one, where both physics and data are used in the training loss; see also \cite{wang2021learning}.  

%% file: IN_UQ_modeling.tex
In this section, we first show how to model and propagate total uncertainty in function approximation problems.
Total uncertainty, which can be construed as the sum of the uncertainties depicted in Fig.~\ref{fig:intro:pie}, is quantified in this section using both a distribution, e.g., for obtaining confidence intervals, and first- as well as second-order statistics.
Further, we integrate these concepts into SciML.
A glossary of various terms we use in this section is provided in Table~\ref{tab:glossary}.
The interested reader is also directed to Appendix~\ref{app:modeling} for supplementary material to this section.

\subsection{Uncertainty in function approximation}\label{sec:uqt:pre:bma}

Given a set of paired noisy observations $\cD = \{x_i,u_i\}_{i=1}^{N}$, our goal is to construct a distribution $p(u|x, \cD, \cH)$ for predicting the value of $u$ at any new location $x$. 
Although for brevity $\cH$ is henceforth dropped, unless considered necessary, we emphasize that all probabilistic distributions discussed in this paper encapsulate a set of user assumptions and preferences (e.g., NN architecture, likelihood function, etc.) that we collectively denote by $\cH$.
For each $x$, it is commonly assumed that $u(x)$ produced by the data-generating process contains both a deterministic part $u_c(x)$ as well as some additive noise $\epsilon_u$, i.e., $u(x) = u_c(x) + \epsilon_u$.
This noise is typically called \textit{data or aleatoric uncertainty}. 
For modeling this process, we assume a \textit{likelihood function} denoted by $p(u|x, \theta)$, with parameters $\theta=[\theta_1,\dots,\theta_K]^T$ to be inferred from the available data.
Further, for defining $p(u|x, \theta)$ we construct a model $u_{\theta}(x)$ that captures the deterministic part of $u(x)$ at some location $x$ and assume a model for the noise.
For example, the factorized Gaussian likelihood function, given as
\begin{equation}\label{eq:uqt:pre:bma:Glike}
	p(u|x, \theta) = \cN(u|u_{\theta}(x), diag(\Sigma_{u}^2)) = \prod_{d = 1}^{D_u}\frac{1}{\sqrt{2\pi \sigma_{u}^2}}\exp(-\frac{(u_d-u_{\theta}(x)_d)^2}{2\sigma_{u}^2}), 
\end{equation}
with the subscript $d$ denoting each of the $D_u$ dimensions of $u$, is often used for multi-dimensional function approximation problems.
In Eq.~\eqref{eq:uqt:pre:bma:Glike}, the output vector $u_{\theta}(x)$ is the mean of the Gaussian distribution assumed for $u$ at location $x$, $diag(\Sigma_{u}^2)$ is a diagonal covariance matrix with $\Sigma_{u}^2 = [\sigma_{u}^2,\dots,\sigma_{u}^2]$ in the diagonal, and $\sigma_{u}^2$ may either be known or assumed, or inferred from data.
Furthermore, see Section~\ref{app:modeling:postemp} for addressing model misspecification cases where, for example, the employed likelihood function is incorrect.

The value of $u$ at a location $x$ given the data $\cD$ is a random variable denoted by $(u|x, \cD)$.
To obtain the density of $(u|x, \cD)$ we integrate out the model parameters $\theta$, i.e., 
\begin{equation}\label{eq:uqt:pre:bma:new:1}
	p(u|x, \cD) = \int p(u|x, \theta)p(\theta|\cD) d\theta.
\end{equation}
This predictive probability density function (PDF) is commonly referred to as Bayesian model average (BMA).
In Eq.~\eqref{eq:uqt:pre:bma:new:1}, $p(\theta|\cD)$ denotes the \textit{posterior density}, which is a distribution over the ``plausible'' parameter values based on the data.
The uncertainty regarding the NN parameters $\theta$ is typically called \textit{epistemic uncertainty}.
Using Bayes' rule, the posterior $p(\theta|\cD)$ can be obtained via
\begin{equation}\label{eq:uqt:pre:bma:post}
	p(\theta|\cD) =\frac{p(\cD|\theta)p(\theta)}{p(\cD)}.
\end{equation}
In Eq.~\eqref{eq:uqt:pre:bma:post}, $p(\cD|\theta)$ is the likelihood of the data, i.e., $p(\cD|\theta) = \prod_{i=1}^{N}p(u_i|x_i, \theta)$ for independent and identically distributed (i.i.d.) data; $p(\theta)$ is the prior probability of the parameters $\theta$ as defined by the model $\cH$; and $p(\cD)$ is called \textit{marginal likelihood} or \textit{evidence} because it represents the probability that out of all possible datasets modeled by $\cH$, it ``happens'' that we observe $\cD$.
The evidence $p(\cD)$ is given as 
\begin{equation}\label{eq:uqt:pre:bma:marglike}
	p(\cD) =\int p(\cD|\theta)p(\theta) d\theta, 
\end{equation}
i.e., given random samples drawn from the prior $p(\theta)$ and then used in conjuction with the likelihood function, $p(\cD)$ represents the probability that the dataset $\cD$ is produced.

\begin{figure}[!ht]
	\centering
	\includegraphics[width=.8\linewidth]{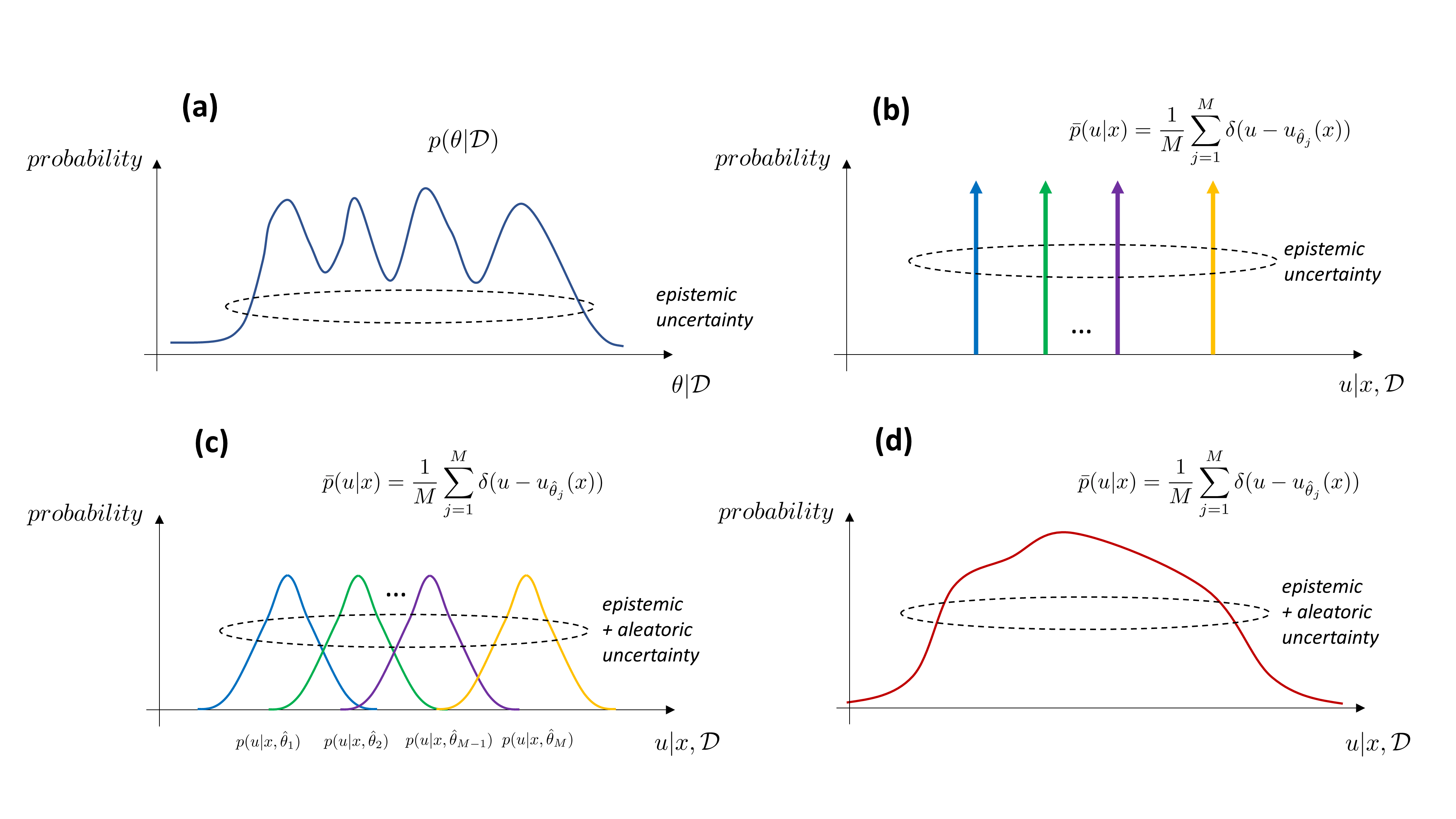}
	\caption{Total uncertainty represented by the predictive distribution $\bar{p}(u|x)$:
		\textbf{(a)} Epistemic uncertainty is taken into account via the posterior $p(\theta|\cD)$, \textbf{(b)} $p(\theta|\cD)$ is approximated via representative samples $\{\hat{\theta}_j\}_{j=1}^M$, \textbf{(c)} aleatoric uncertainty is added to each sample from $p(\theta|\cD)$ via the distributions $\{p(u|x, \hat{\theta}_j)\}_{j=1}^M$, \textbf{(d)} $\bar{p}(u|x)$ approximates total uncertainty by adding up the distributions $\{p(u|x, \hat{\theta}_j)\}_{j=1}^M$.
		If each $p(u|x, \hat{\theta}_j)$ is a Gaussian distribution, $\bar{p}(u|x)$ corresponds to a Gaussian mixture.    
		See Table~\ref{tab:glossary} for a glossary of used terms.
	}
	\label{fig:uqt:pre:mcest:total:unc}
\end{figure}

\begin{figure}[!ht]
	\centering
	\includegraphics[width=.7\linewidth]{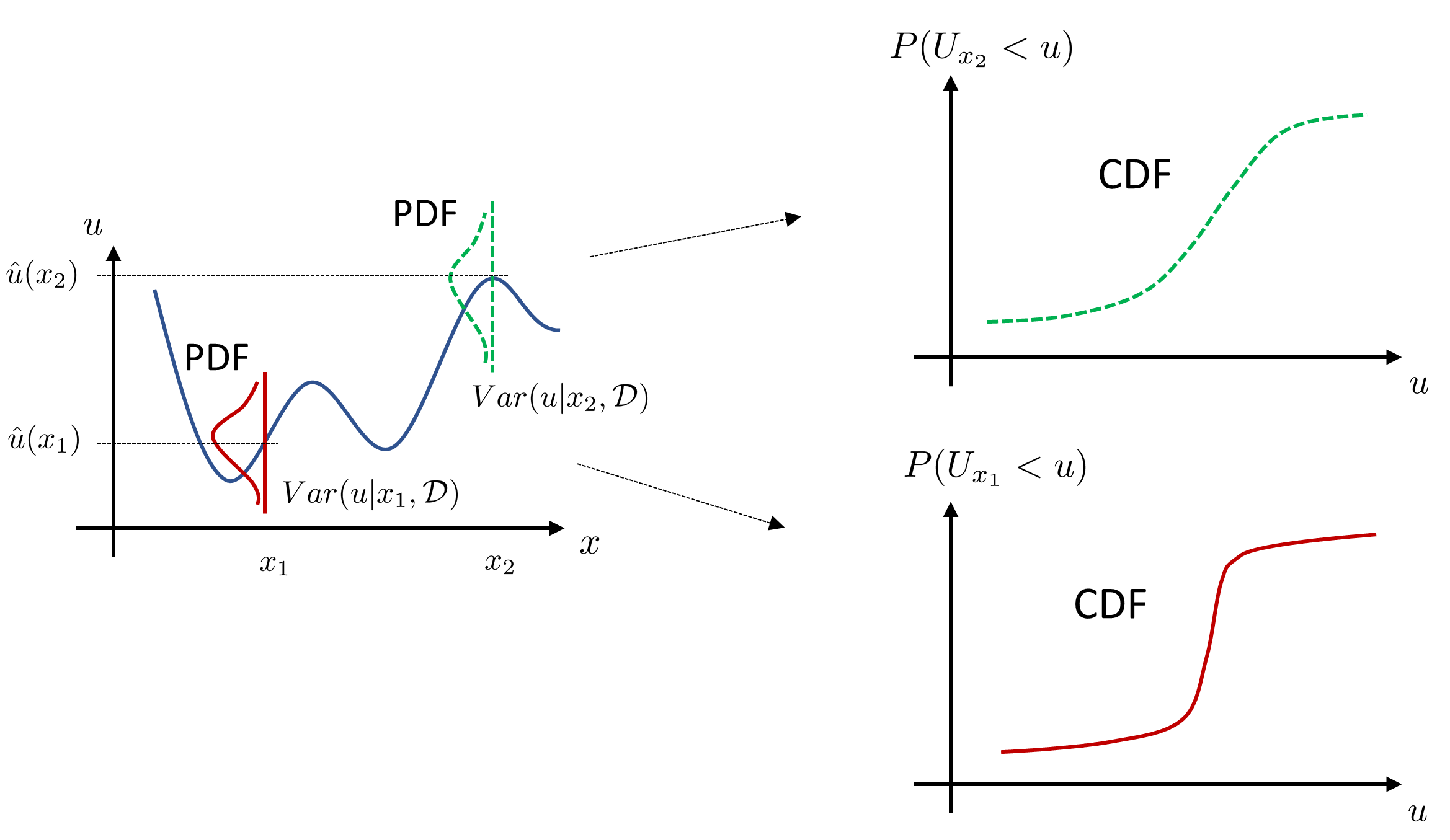}
	\caption{Predictions for $u$ corresponding to two arbitrary $x$ locations, $x_1$ (red solid lines) and $x_2$ (green dashed lines): given data $\cD$, all techniques presented in this paper output a predictive probability density function (PDF), or equivalently a cumulative distribution function (CDF), for each value of $x$ via Eq.~\eqref{eq:uqt:pre:bma:new:1}. 
	The most straightforward way to make PDF/CDF evaluations is to fit a Gaussian (or, e.g., Student-t) distribution to the posterior samples $\{u_{\hat{\theta}_j}(x)\}_{j=1}^M$ for each $x$.
	First- and second-order statistics can be approximated via Eqs.~\eqref{eq:uqt:pre:mcestmc:mean} and \eqref{eq:uqt:pre:mcestmc:totvar}, respectively.
	Note that, given left-out data referred to as calibration dataset, the post-training calibration techniques presented in Section~\ref{sec:eval:calib} modify the predictive PDFs/CDFs to improve statistical consistency of the predictions. See Table~\ref{tab:glossary} for a glossary of used terms.
	}
	\label{fig:eval:eval:outputs}
\end{figure}

Obtaining the posterior exactly via Eq.~\eqref{eq:uqt:pre:bma:post} is computationally and analytically intractable. 
To address this issue, approximate inference aims to approximate the posterior by another distribution and/or obtaining samples from the posterior. 
``Standard NN training'', which can be construed as approximating the posterior by a point estimate as described in Section~\ref{app:modeling:point}, does not take into account epistemic uncertainty.
In this regard, in Sections~\ref{sec:uqt:bnns}-\ref{sec:uqt:ens} we review and integrate into SciML the most commonly used posterior inference approaches.
All methods obtain a set of $\theta$ samples denoted as $\{\hat{\theta}_j\}_{j=1}^M$.
Following the posterior inference phase, the BMA of Eq.~\eqref{eq:uqt:pre:bma:new:1} can be approximated using Monte Carlo (MC) estimation as
\begin{equation}\label{eq:uqt:pre:mcestmc:mcest}
	p(u|x, \cD) = \mathbb{E}_{\theta|\cD}[p(u|x, \theta)] \approx \bar{p}(u|x) =  \frac{1}{M}\sum_{j=1}^Mp(u|x, \hat{\theta}_j).
\end{equation}
This equation provides the approximate \textit{total uncertainty} of $(u|x, \cD)$ in the form of a predictive PDF $\bar{p}(u|x)$; see Fig.~\ref{fig:uqt:pre:mcest:total:unc} for an illustration.
The mean of $(u|x, \cD)$ is modeled via $\hat{u}(x) = \mathbb{E}[u|x, \cD]$ of Eq.~\eqref{eq:uqt:pre:bma:new:4} and approximated by $\bar{\mu}(x)$ as
\begin{equation}\label{eq:uqt:pre:mcestmc:mean}
	\hat{u}(x) \approx \bar{\mu}(x) = \frac{1}{M}\sum_{j=1}^Mu_{\hat{\theta}_j}(x),
\end{equation}
where $\{u_{\hat{\theta}_j}(x)\}_{j=1}^M$ is the set of NN predictions corresponding to the samples $\{\hat{\theta}_j\}_{j=1}^M$.
For obtaining the variance of $(u|x, \cD)$, substituting the Gaussian likelihood of Eq.~\eqref{eq:uqt:pre:bma:Glike} into Eq.~\eqref{eq:uqt:pre:mcestmc:mcest} yields a Gaussian mixture with the diagonal part of its covariance matrix derived analytically and given by
\begin{equation}\label{eq:uqt:pre:mcestmc:totvar}
	Var(u|x, \cD) \approx \bar{\sigma}^2(x) =  \underbrace{\Sigma_{u}^2}_{\bar{\sigma}_a^2(x)}  + \underbrace{\frac{1}{M}\sum_{j=1}^M(u_{\hat{\theta}_j}(x)-\bar{\mu}(x))^2}_{\bar{\sigma}_e^2(x)}. 
\end{equation} 
In this equation, $\bar{\sigma}^2(x)$ represents the approximate total uncertainty of $(u|x, \cD)$, while $\bar{\sigma}_a^2(x)$ and $\bar{\sigma}_e^2(x)$ denote the aleatoric and the epistemic parts of the total uncertainty, respectively.
Note that up to a constant $\frac{M}{M-1}$, the term $\bar{\sigma}_e^2(x)$ is equal to the sample variance of $\{u_{\hat{\theta}_j}(x)\}_{j=1}^M$.
An alternative derivation of Eq.~\eqref{eq:uqt:pre:mcestmc:totvar} based on the law of total variance can be found in Eq.~\eqref{eq:uqt:pre:bma:totvar:1}.
Overall, after obtaining $\{\hat{\theta}_j\}_{j=1}^M$, a predictive model in the form of a PDF $\bar{p}(u|x)$ via Eq.~\eqref{eq:uqt:pre:mcestmc:mcest} is available for each value of $x$;
see Fig.~\ref{fig:eval:eval:outputs}.
The corresponding cumulative distribution function (CDF) denoted by $\bar{P}(u|x)$ can also be obtained in a straightforward manner. 
In practice, however, instead of using Eq.~\eqref{eq:uqt:pre:mcestmc:mcest}, a Gaussian (or, e.g., Student-t) distribution is commonly fitted to the posterior samples and used for both PDF and CDF evaluations.
For example, we denote as $\pazocal{N}(\bar{\mu}(x), \bar{\sigma}^2(x))$ the Gaussian approximation to the samples $\{u_{\hat{\theta}_j}(x)\}_{j=1}^M$ at an arbitrary location $x$, where $\bar{\mu}(x)$ and $\bar{\sigma}^2(x)$ are obtained via Eqs.~\eqref{eq:uqt:pre:mcestmc:mean} and \eqref{eq:uqt:pre:mcestmc:totvar}, respectively.

For the Gaussian likelihood of Eq.~\eqref{eq:uqt:pre:bma:Glike}, a homoscedastic data noise model has been considered so far in this paper, with a diagonal covariance matrix and constant, in terms of $x$, variance $\sigma_{u}^2$ for each dimension of $u(x)$.
An extension that accounts for heteroscedastic noise relates to considering a location-dependent variance vector $\Sigma_{u}^2$.
Specifically, $\Sigma_{u}^2$ can be modeled as part of the NN output by augmenting the NN output $u_{\theta}(x)$ by $\Sigma_{u, \theta}^2(x) = [\sigma_{u, \theta}^2(x)_1, \dots, \sigma_{u, \theta}^2(x)_{D_u}]$, where $\sigma_{u, \theta}^2(x)$ can also be shared among dimensions (see also \cite{lakshminarayanan2017simple}).
For making predictions given posterior samples $\{\hat{\theta}_j\}_{j=1}^M$, each $p(u|x, \hat{\theta}_j)$ in Eq.~\eqref{eq:uqt:pre:mcestmc:mcest} not only has a different mean, but also a different variance, and Eq.~\eqref{eq:uqt:pre:mcestmc:totvar} becomes   
\begin{equation}\label{eq:modeling:noise:model:totvar}
	Var(u|x, \cD) \approx \bar{\sigma}^2(x) =  \underbrace{\frac{1}{M}\sum_{j=1}^M\Sigma_{u, \hat{\theta}_j}^2(x)}_{\bar{\sigma}_a^2(x)} + \underbrace{\frac{1}{M}\sum_{j=1}^M(u_{\hat{\theta}_j}-\bar{\mu}(x))^2}_{\bar{\sigma}_e^2(x)}. 
\end{equation} 
Similarly, we can obtain the complete covariance matrix of $(u|x,\cD)$; see, e.g., \cite{tanno2021uncertainty}.

\subsection{Uncertainty in PINNs}\label{sec:uqt:pre:pinn}

The observations in the mixed PDE problem dataset $\cD = \{\cD_f, \cD_b, \cD_u, \cD_{\lambda}\}$ are typically obtained with a data acquisition device and are noisy. 
In this regard, we employ four likelihood functions for approximating the $f, b, u$ and $\lambda$ distributions, denoted as $p(f|x, \theta), p(b|x, \theta), p(u|x, \theta)$, and $p(\lambda|x, \theta)$, respectively.
The total likelihood of the dataset $\cD$ becomes
\begin{equation}\label{eq:uqt:pre:pinn:like}
	p(\cD|\theta) = p(\cD_f|\theta)p(\cD_b|\theta)p(\cD_{u}|\theta)p(\cD_{\lambda}|\theta),
\end{equation}
where 
\begin{equation}
\begin{aligned}\label{eq:uqt:pre:pinn:likes}
	p(\cD_f|\theta) & = \prod_{i=1}^{N_f}p(f_i|x_i, \theta) \text{; } & p(\cD_b|\theta) & = \prod_{i=1}^{N_b}p(b_i|x_i, \theta);\\
	p(\cD_u|\theta) & = \prod_{i=1}^{N_u}p(u_i|x_i, \theta) \text{; } & p(\cD_{\lambda}|\theta) & = \prod_{i=1}^{N_{\lambda}}p(\lambda_i|x_i, \theta).
\end{aligned}
\end{equation}

Combining point estimation of $\theta$ (Section~\ref{app:modeling:point}) with the Gaussian likelihood function of Eq.~\eqref{eq:uqt:pre:bma:Glike} for all data categories leads to the PINN minimization problem of Eq.~\eqref{eq:intro:piml:pinn:loss:1}.
The noise scales $\sigma_u^2$ in the assumed Gaussian likelihoods are in general different for each data category and are either known or learned together with $\theta$ during optimization.
The PDF of the solution random variable $(u|x, \cD)$ contains only aleatoric uncertainty and is given as $p(u|x, \hat{\theta})$, where $\hat{\theta}$ is the optimal $\theta$.
For accounting for epistemic uncertainty as well, posterior inference (Sections~\ref{sec:uqt:bnns}-\ref{sec:uqt:ens}) is required for obtaining multiple $\theta$ samples.
In this case, the PDF of the solution random variable $(u|x, \cD)$ contains both epistemic and aleatoric uncertainties and is computed via Eq.~\eqref{eq:uqt:pre:mcestmc:mcest}.

\subsection{Uncertainty in DeepONets}\label{sec:uqt:pre:don}

As in Section~\ref{sec:intro:piml:don}, we assume that $f$ and $b$ in Eq.~\eqref{eq:intro:piml:pinn:pde} are known. 
Next, we assume that the dataset $\cD_{\lambda}$ consists of clean data.
Because the data of $\lambda$ is used as input to DeepONet (Fig.~\ref{fig:intro:piml:don:arch}), it is not straightforward to consider noisy data during the pre-training procedure; see, e.g., \cite{wright1998neural} for accounting for input uncertainty in standard NNs.
Nevertheless, we assume that $\cD_u$ consists of noisy data in general.
In this regard, a likelihood function is employed for the $u$ data, denoted by $p(u|x, \xi, \theta)$, and the total likelihood is given as
\begin{equation}\label{eq:uqt:pre:don:like}
    p(\cD|\theta) =\prod_{i=1}^{N}\prod_{j=1}^{N_u}p(u_j^{(i)}|x_j^{(i)}, \xi_i, \theta). 
\end{equation}

Combining point estimation of $\theta$ (Section~\ref{app:modeling:point}) with the Gaussian likelihood function of Eq.~\eqref{eq:uqt:pre:bma:Glike} for the $u$ data leads to the DeepONet minimization problem of Eq.~\eqref{eq:intro:piml:don:min}.
The noise scale $\sigma_u^2$ in the assumed Gaussian likelihood is either known or learned together with $\theta$ during optimization.
During inference for a new value $\xi' \in \Xi$, the vanilla DeepONet considers a clean dataset $\cD_{\lambda}' = \{x_i, \lambda_i\}_{i=1}^{N_{\lambda}'}$; see Section~\ref{sec:uqt:fpriors} for overcoming this limitation. 
The PDF of the solution random variable $(u|x,\xi', \cD)$ contains only aleatoric uncertainty and is given as $p(u|x,\xi', \hat{\theta})$, where $\hat{\theta}$ is the optimal $\theta$.
For accounting for epistemic uncertainty as well, posterior inference (Sections~\ref{sec:uqt:bnns}-\ref{sec:uqt:ens}) is required for obtaining multiple $\theta$ samples.
In this case, the PDF of the solution random variable $(u|x, \xi', \cD)$ contains both epistemic and aleatoric uncertainties and is computed via Eq.~\eqref{eq:uqt:pre:mcestmc:mcest}.
Note that concurrently with the present work, Markov chain Monte Carlo has been employed for posterior inference in conjunction with DeepONet-based operator learning \cite{lin2021accelerated}.

%% file: IN_UQ_methods.tex
Following the uncertainty modeling presented in Section~\ref{sec:uqt:pre}, a solution method is required for obtaining a set of NN parameter posterior samples $\{\hat{\theta}_j\}_{j=1}^M$.
These samples are used in the approximate BMA of Eqs.~\eqref{eq:uqt:pre:mcestmc:mcest} and \eqref{eq:uqt:pre:mcestmc:mean}-\eqref{eq:uqt:pre:mcestmc:totvar} for obtaining total uncertainty predictions in the form of distribution and first-/second-order statistics, respectively.
In this section, we briefly review and integrate into SciML the most commonly used methods for obtaining $\{\hat{\theta}_j\}_{j=1}^M$; see Table~\ref{tab:uqt:over} for an overview.
Next, we demonstrate how to extend these methods for improving predictions when historical data is available, as well as for representing stochastic processes and solving stochastic differential equations.
Finally, we present a formulation towards a unified UQ framework that addresses the problems of Fig.~\ref{fig:intro:all:problems} by seamlessly combining physics with novel and historical data, which could be contaminated with various types of noise. A glossary of various terms we use in this section is provided in Table~\ref{tab:glossary}.
The interested reader is also directed to Section~\ref{app:methods} for supplementary material to this section.

\subsection{Bayesian methods}\label{sec:uqt:bnns}

Bayesian methods approximate the posterior $p(\theta|\cD)$ of Eq.~\eqref{eq:uqt:pre:bma:post}, either implicitly or explicitly.
In the following, we provide an overview of the most widely used Bayesian methods (see also Table~\ref{tab:uqt:over}), as well as brief information regarding prior/data uncertainty optimization and integration into SciML.
A detailed description of the considered methods can be found in Section~\ref{app:methods:bnns}. 

Markov-chain Monte Carlo (MCMC) techniques, reviewed in Section~\ref{app:methods:bnns:mcmc}, draw parameter $\theta$ samples from the posterior by using a Markov chain with $p(\theta|\cD)$ as its invariant, i.e., stationary, distribution.
We employ two such methods, namely Hamiltonian Monte Carlo (HMC) and Langevin Dynamics (LD), in the comparative study of Section~\ref{sec:comp}.
Variational inference (VI) techniques, reviewed in Section~\ref{app:methods:bnns:vi}, approximate $p(\theta|\cD)$ by employing a variational distribution $q_{\omega}(\theta)$ parametrized by $\omega$, that is optimized by maximizing a lower bound of the marginal likelihood of Eq.~\eqref{eq:uqt:pre:bma:marglike}.
Mean-field VI (MFVI)
utilizes a factorized Gaussian variational distribution.  
Monte Carlo dropout (MCD), which is reviewed in Section~\ref{app:methods:bnns:vi:mcd} and can be interpreted as a special case of VI, first requires standard training with dropout \cite{srivastava2014dropout}.
Subsequently, at inference time $\theta$ samples are drawn by randomly dropping for each sample a fixed percentage of the NN parameters.
Laplace approximation (LA), reviewed in Section~\ref{app:methods:bnns:lapl}, involves standard NN training, and subsequently fitting a Gaussian distribution to the discovered mode $\hat{\theta}$, with a width determined by the Hessian of the training error. 

For the techniques considered in this section, evaluations of the likelihood and prior terms, $p(\cD|\theta)$ and $p(\theta)$, respectively, for different values of $\theta$ are required.
In the PINN method, the required likelihood is given by Eqs.~\eqref{eq:uqt:pre:pinn:like}-\eqref{eq:uqt:pre:pinn:likes}, whereas in the DeepONet method by Eq.~\eqref{eq:uqt:pre:don:like}. 
Evaluating the prior $p(\theta)$ does not differ from standard function approximation in either PINN or DeepONet cases.	
The obtained set of samples $\{\hat{\theta}_j\}_{j=1}^M$ is subsequently combined with the approximate BMAs in Eqs.~\eqref{eq:uqt:pre:mcestmc:mcest}-\eqref{eq:uqt:pre:mcestmc:totvar}.
In this paper, we refer to PINN and DeepONet combined with Bayesian methods as U-PINN and U-DeepONet, where U stands for ``uncertain''.

\input{IN_table_techniques}

\begin{figure}[!ht]
	\centering
	\includegraphics[width=.7\linewidth]{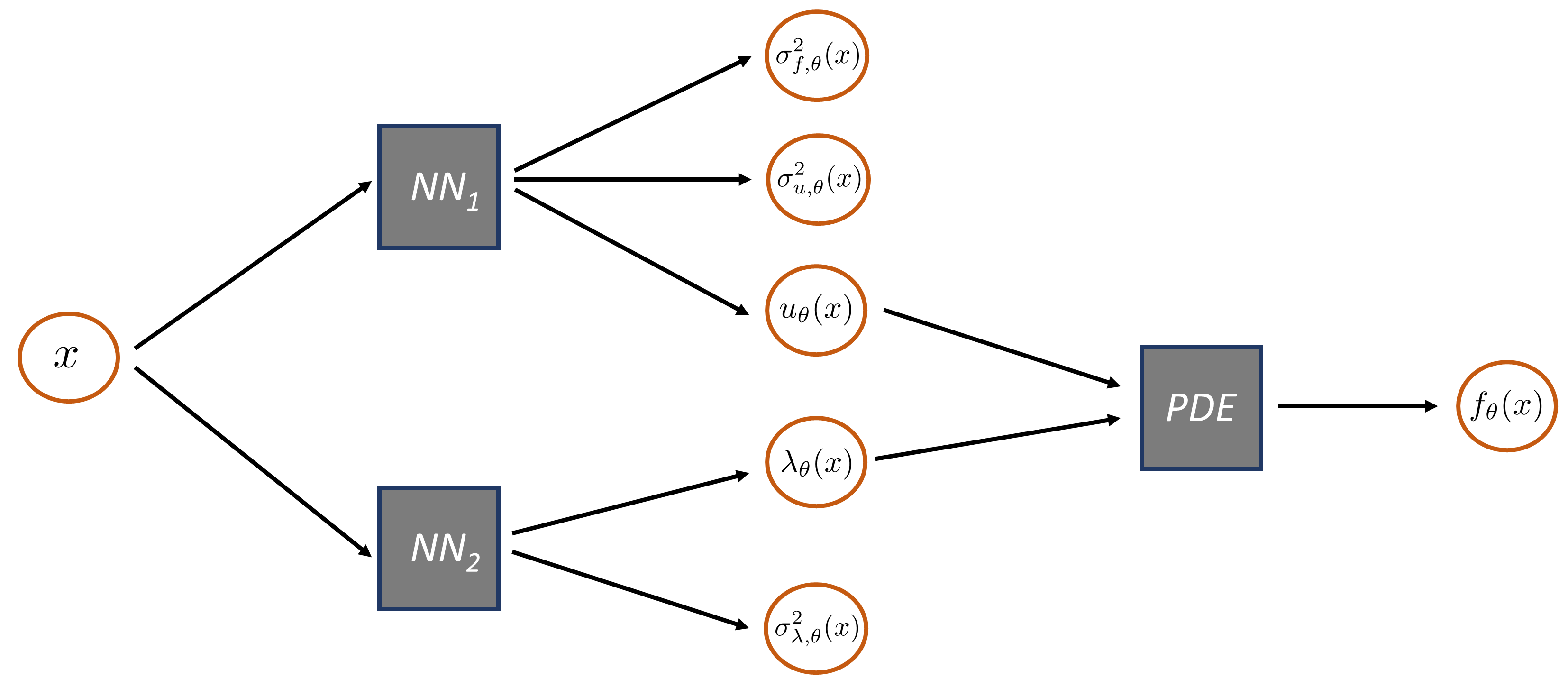}
	\caption{PINN consisting of two NNs, one for the solution $u$ and one for the parameter field $\lambda$.
	The parameters of the NNs are denoted by $\theta$.
	We also consider heteroscedastic Gaussian noise in the data of $f$, $u$, and $\lambda$, whose variances are denoted by $\sigma_{f, \theta}^2$, $\sigma_{u, \theta}^2$, and $\sigma_{\lambda, \theta}^2$, respectively. 
	We refer to this model as h-U-PINN in Table~\ref{tab:uqt:over}; i.e., PINN with heteroscedastic noise modeling and parameters $\theta$ obtained through posterior inference.
	Related computational results can be found in Section~\ref{sec:comp:pinns:hetero}.
		The model depicted in this figure can be construed as a special case of Fig.~\ref{fig:intro:all:techniques} with additional NN outputs for the noise variances.}
	\label{fig:uqt:bnns:dataunc:heteroscedastic:upinn}
\end{figure}

For the case of Section~\ref{app:modeling:hypers} pertaining to unknown prior and noise model, the considered Bayesian methods can be combined with alternating optimization schemes.
We construe these approaches as \textit{online learning methods}.
Further, cross-validation or meta-learning, in conjunction with the evaluation metrics of Section~\ref{sec:eval} and validation data, can be utilized.
These approaches are construed as \textit{offline learning methods}.
The interested reader can find detailed descriptions in Sections~\ref{app:methods:bnns:mcmc:hypers}, \ref{app:methods:bnns:vi:hypers}, and \ref{app:methods:bnns:lapl:hypers}, pertaining to MCMC, VI, and LA, respectively.
Lastly, aleatoric uncertainty can alternatively be considered as an additional output of the NN; see Section~\ref{sec:uqt:pre:bma}.
In that case, the posterior $p(\theta|\cD) \propto p(\cD|\theta)p(\theta)$ depends on $\sigma_{u, \theta}(x)$ through the likelihood term $p(\cD|\theta)$, and thus $\sigma_{u, \theta}(x)$ is automatically taken into account by all the posterior inference techniques of this section.
We refer to a posterior inference method denoted by X as h-X, when it is combined with heteroscedastic noise modeling.
In Fig.~\ref{fig:uqt:bnns:dataunc:heteroscedastic:upinn}, we illustrate how this idea can be applied to PINNs. 

\subsection{Ensembles}\label{sec:uqt:ens}

Instead of approximating the posterior $p(\theta|\cD)$ as performed with Bayesian methods, the aim of ensembles is to obtain multiple modes of $p(\theta|\cD)$, or equivalently, multiple minima of the NN parameter loss landscape. 
This set of parameters $\{\hat{\theta}_j\}_{j=1}^M$ can be used exactly as the posterior samples obtained with Bayesian techniques, i.e., in conjunction with an approximate BMA as delineated in Section~\ref{sec:uqt:pre:bma}.
In the following, we provide an overview of the considered techniques (see also
Table~\ref{tab:uqt:over}), as well as brief information regarding regularization/data uncertainty optimization and integration
into SciML.

Approximate inference by deep ensembles (DEns) pertains to obtaining $M$ different settings of parameters $\theta$ by standard NN training, independently $M$ times using different NN parameter initializations.
More information can be found in Section~\ref{app:methods:ens:deep}. 
On the contrary, snapshot ensembles (SEns) obtain $\{\hat{\theta}_j\}_{j=1}^M$ without incurring any additional cost as compared to standard training.
To this end, standard algorithms are let to converge to $M$ different local optima during a single optimization trajectory, and each optimum is used as a sample in $\{\hat{\theta}_j\}_{j=1}^M$.  
More information can be found in Section~\ref{app:methods:ens:snap}. 
Stochastic weight averaging-Gaussian (SWAG) extends SEns by also fitting a Gaussian distribution to the aforementioned local optima. 
This distribution is used for drawing the samples in $\{\hat{\theta}_j\}_{j=1}^M$.
More information can be found in Section~\ref{app:methods:ens:swag}.

For the techniques considered in this section, standard NN training is performed.
To integrate ensemble techniques into SciML, standard training of PINNs and DeepONets, as described in Sections~\ref{sec:intro:piml:pinn} and \ref{sec:intro:piml:don}, respectively, is performed either multiple times (DEns) or once with an appropriate learning rate schedule (SEns and SWAG) for obtaining $\{\hat{\theta}_j\}_{j=1}^M$.
This set of samples is, subsequently, combined with the approximate BMAs in Eqs.~\eqref{eq:uqt:pre:mcestmc:mcest}-\eqref{eq:uqt:pre:mcestmc:totvar}.
As in the case of Bayesian methods, we refer to PINN and DeepONet combined with ensembles as U-PINN and U-DeepONet.

Similarly to Bayesian methods, online and offline methods can be used in the context of ensembles for identifying the constant noise scale $\sigma_u^2$.
This requires using the negative log-likelihood as a loss function in NN training (Eqs.~\eqref{eq:modeling:point:mle}-\eqref{eq:modeling:point:map:2}), instead of the standard MSE loss (Eqs.~\eqref{eq:intro:piml:pinn:loss:1} and \eqref{eq:intro:piml:don:min}).
Specifically, the loss function $\pazocal{L}(\theta)$ of Eqs.~\eqref{eq:modeling:point:mle}-\eqref{eq:modeling:point:map:2} is also a function of $\sigma_{u}^2$ through the likelihood term $\log p(\cD|\theta)$. 
In this context, $\sigma_{u}^2$ can be identified \textit{online} by alternating between update steps for the NN parameters $\theta$ and steps for $\sigma_{u}^2$.
Clearly, these steps can also be combined into one step.
Further, $\sigma_{u}^2$ can be identified \textit{offline} by using cross-validation or meta-learning, in conjunction with the evaluation metrics of Section~\ref{sec:eval} and with validation data.
Furthermore, the equivalent of prior in ensembles with standard NN training is regularization, which can be optimized offline.
Lastly, aleatoric uncertainty can alternatively be considered as an additional output of the NN; see Section~\ref{sec:uqt:pre:bma}.
In that case, the log likelihood loss function of Eqs.~\eqref{eq:modeling:point:mle}-\eqref{eq:modeling:point:map:2} is utilized for standard NN training, either once (SEns and SWAG) or $M$ times (DEns).
This loss function depends on $\sigma_{u, \theta}(x)$ through the likelihood term $\log p(\cD|\theta)$.

\subsection{Functional priors (FPs)}\label{sec:uqt:fpriors}

In this section we describe an approach for harnessing historical data in order to improve performance and reduce the computational cost of posterior inference in function approximation and SciML problems.

\begin{figure}[!ht]
	\centering
	\includegraphics[width=.7\linewidth]{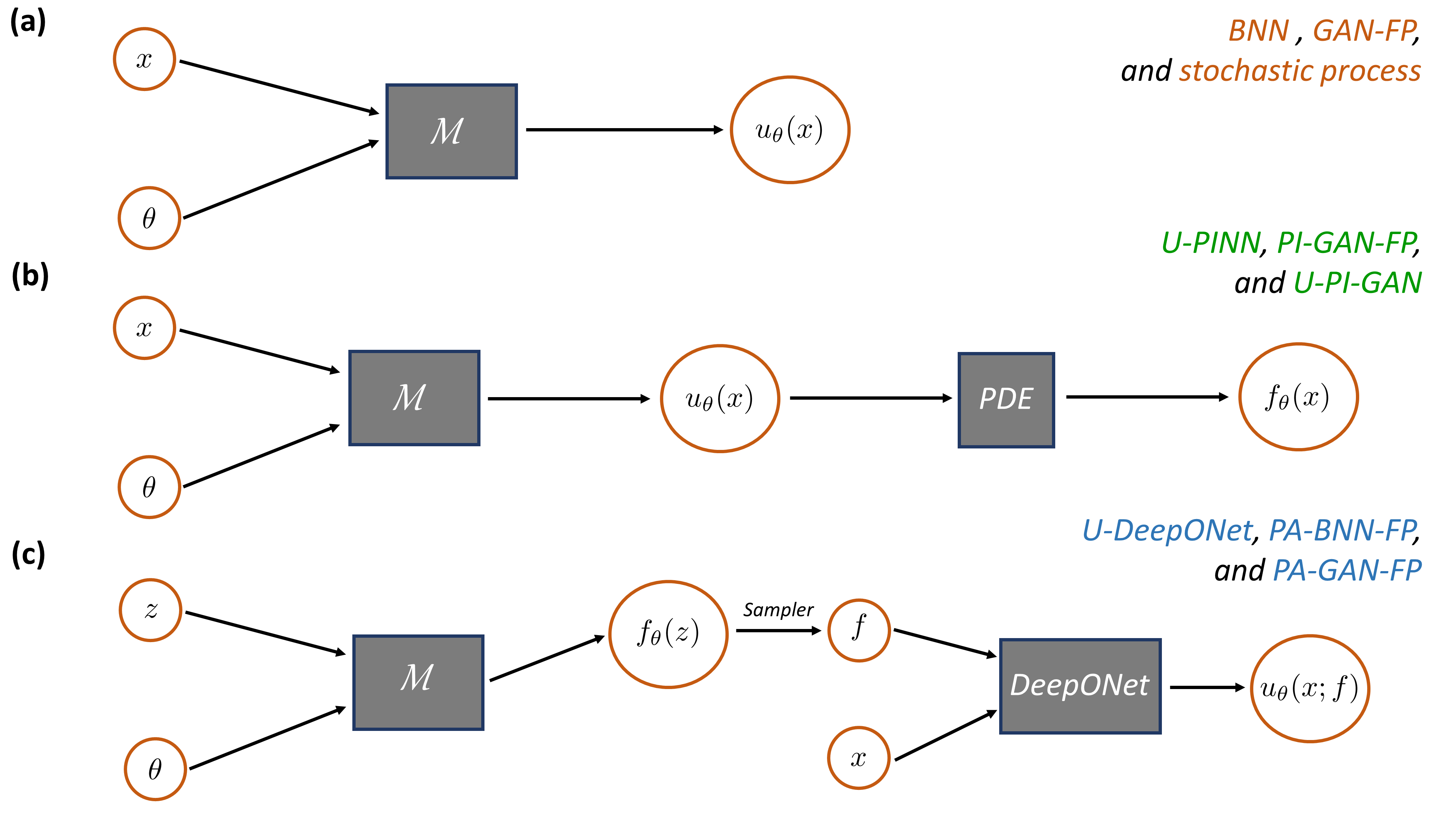}
	\caption{\textbf{(a)} BNN, GAN functional prior (GAN-FP) and stochastic process representation.
	\textbf{(b)} U-PINN, PI-GAN-FP, and U-PI-GAN.
	\textbf{(c)} U-DeepONet, PA-BNN-FP, and PA-GAN-FP.
	$\pazocal{M}$ expresses how $x$ and $\theta$ are combined to yield $u_{\theta}(x)$. 
		In standard NN models (BNN, U-PINN, U-DeepONet), $\pazocal{M}$ is a BNN.
		In generator-based models (stochastic process, FP, PI-GAN-FP, U-PI-GAN, and PA-GAN-FP), $\pazocal{M}$ is the generator of a GAN that is trained using stochastic realizations or pre-trained using historical data.
		All models depicted in this figure can be construed as special cases of Fig.~\ref{fig:intro:all:techniques}; note, however, the different meaning of $\theta$ for each technique.
		Further, in part (c), $\lambda$ (or both $f$ and $\lambda$) could be used as input to the DeepONet instead of $f$, as in Fig.~\ref{fig:intro:piml:don:arch}.
		Furthermore, $u(x; f)$ is equivalent to $u(x; \xi)$, because each $\xi \in \Xi$ is associated with an input $f$. 
		The additional input $z$ is used to emphasize that the locations in the input domains of $f$ and $u$ can be different and the ``sampler'' that the function $f_{\theta}(z)$ must be sampled at pre-specified locations before entered as input to the DeepONet. 
		See Table~\ref{tab:uqt:over} for an overview of the UQ methods and Table~\ref{tab:glossary} for a glossary of used terms.
	}
	\label{fig:uqt:fpriors:FP}
\end{figure}

Consider Fig.~\ref{fig:uqt:fpriors:FP}a, where $x$ and $u_{\theta}(x)$ are the input and the output of the NN, respectively, and $\theta$ are the NN parameters.
Further, $\pazocal{M}$ is a black-box model that combines $x$ and $\theta$ for producing $u_{\theta}(x)$. 
Typically, $\pazocal{M}$ is the NN architecture (connections and activation functions), which defines what operations are applied to $x$ and $\theta$.
In this regard, the model of Fig.~\ref{fig:uqt:fpriors:FP}a defines a functional prior (FP); i.e., $\theta$ is drawn from a known prior probability distribution and it induces a distribution of corresponding functions $u_{\theta}(x)$.
We refer to this as a Bayesian NN (BNN) or BNN-FP.
In BNNs, posterior inference for $\theta$ may be performed either using the Bayesian methods of Section~\ref{sec:uqt:bnns} or the ensembles of Section~\ref{sec:uqt:ens}.

Note that both different $\pazocal{M}$ models and $\theta$ priors lead to different FPs.
In this context, a FP can be constructed by training $\pazocal{M}$ in Fig.~\ref{fig:uqt:fpriors:FP} such that it encodes historical data information, as proposed by \cite{meng2021learning}.
To this end, we consider the dataset $\pazocal{D}_h = \{U_i\}_{i=1}^{N_h}$, with $U_i = \{(x_j^{(i)}, u_j^{(i)})\}_{j=1}^{N_u}$, containing $N_h$ historical realizations. 
Specifically, $u_{\theta}(x)$ is the output of the generator of a pre-trained generative adversarial network (GAN), whose distribution over $\theta$ matches the historical data distribution.
We refer to this as GAN-FP.
Following the pre-training phase, posterior inference for $\theta$ may be performed either using the standard Bayesian techniques of Section~\ref{sec:uqt:bnns} or the ensembles of Section~\ref{sec:uqt:ens}.
We refer to a posterior inference method denoted by X as X+BNN\footnote[1]{We use ``+'' here to emphasize that the same posterior inference methods can be used regardless of the
selected model, which can be a BNN or a GAN-FP.} when it is combined with a BNN-FP (default setting), and as X+FP when it is combined with a GAN-FP.
Note that inference for $\theta$ using the pre-trained GAN-FP can be performed faster, with less amount of data, and by using a significantly reduced dimension for the parameters $\theta$.
The interested reader is directed to Section~\ref{app:methods:fpriors} for additional details.

The above concepts are integrated into SciML, as shown in Figs.~\ref{fig:uqt:fpriors:FP}b and c.
Specifically, $\pazocal{M}$ in Fig.~\ref{fig:uqt:fpriors:FP}b is either a BNN, which is the case of standard U-PINNs, or the  generator of a pre-trained GAN. We refer to the latter as physics-informed GAN-FP (PI-GAN-FP).
The GAN is pre-trained using a dataset $\pazocal{D}_h$, $N_h$ historical realizations of $u$, each one denoted as $U_i = \{(x_j^{(i)}, u_j^{(i)})\}_{j=1}^{N_u}$, and/or of $f$, each one denoted as $F_i = \{(x_j^{(i)}, f_j^{(i)})\}_{j=1}^{N_f}$.
See the original paper \cite{meng2021learning} and Section~\ref{app:methods:fpriors} for detailed information, as well as Section~\ref{sec:comp:pinns:stand} for an example.  
Next, if $\pazocal{M}$ in Fig.~\ref{fig:uqt:fpriors:FP}c is a BNN and the DeepONet is pre-trained with noiseless data, the model can be used for treating cases with noisy data during inference; see Section~\ref{sec:intro:problem:form} for problem formulation details.
We refer to this as physics-agnostic BNN-FP (PA-BNN-FP) and note that it is a novel contribution of this paper. 
Further, if $\pazocal{M}$ is a pre-trained generator and the DeepONet is pre-trained with noiseless data, the model of Fig.~\ref{fig:uqt:fpriors:FP}c corresponds to a physics-agnostic GAN-FP (PA-GAN-FP).
This is used for treating cases with noisy data during inference similarly to PA-BNN-FP. 
The historical dataset for training PA-GAN-FP can be the same as the one used for pre-training the DeepONet. 
After pre-training, PA-BNN-FP and PA-GAN-FP can be used during inference to infer $\theta$ given new $(x, u)$ and/or $(x, f)$ data.
Further, in Fig.~\ref{fig:uqt:fpriors:FP}c, $\lambda$ (or both $f$ and $\lambda$) could be used as input to the DeepONet instead of $f$.
For example, in Section~\ref{sec:comp:don:fp} we solve an operator learning problem with $\lambda$ as the only input to the DeepONet and we consider historical data of $\lambda$ and $u$. 
See the original paper \cite{meng2021learning} and Section~\ref{app:methods:fpriors} for more information. 
In passing, note that in Fig.~\ref{fig:uqt:fpriors:FP}c we have used an additional input $z$ as well as a ``sampler''. 
The additional input $z$ is used to emphasize that the locations in the input domains of $f$ and $u$ can be different, i.e., $z$ and $x$, respectively.
In addition, the ``sampler'' signifies that the function $f_{\theta}(z)$ must be sampled at pre-specified locations before entered as input to the DeepONet. 

\subsection{Solving stochastic PDEs (SPDEs)}\label{sec:uqt:sdes}

In this section we extend the function approximation and PDE solution techniques of the previous sections for representing stochastic processes and for solving SPDEs.

The model in Fig.~\ref{fig:uqt:fpriors:FP}a can also be used for stochastic process representation if $\pazocal{M}$ is the generator of a GAN.
In this regard, $\pazocal{M}$ is trained such that its distribution over $\theta$ matches the distribution of the available stochastic realizations.
Specifically, the generator $u_{\theta}$ in Fig.~\ref{fig:uqt:fpriors:FP}a takes as input a random $\theta$ from a known distribution, e.g., Gaussian, and its parameters $\beta$ are trained using the dataset $\pazocal{D} = \{U_i\}_{i=1}^N$ with $U_i = \{(x_j^{(i)}, u_j^{(i)})\}_{j=1}^{N_u}$. 
After training of $\beta$, we can generate samples of $\theta$ from its distribution, and thus samples of the stochastic process $u_{\theta}$, which by construction approximately follow the same distribution as the data. 

Next, consider the forward SPDE problem of Section~\ref{sec:intro:problem:form} with known $b$ in Eq.~\eqref{eq:intro:piml:pinn:pde}, without loss of generality.
For known $\lambda$, unknown $u$, and stochastic realizations of $f$, the problem can be solved by training the model of Fig.~\ref{fig:uqt:fpriors:FP}b with $\pazocal{M}$ being the generator of a GAN. 
This GAN is used for modeling $u$, while $f$ is given by substituting the output of the GAN into the PDE.
The dataset for this forward case is expressed as $\pazocal{D} = \{F_i\}_{i=1}^{N}$, with $F_i = \{(x_j^{(i)}, f_j^{(i)})\}_{j=1}^{N_{f}}$, and contains $N$ realizations of $f$.
The same model of Fig.~\ref{fig:uqt:fpriors:FP}b can be used for solving a mixed SPDE problem with partially unknown $u$.
If $\lambda$ is also partially unknown, an additional GAN is used for modeling $\lambda$. 
The dataset in the latter case is expressed as $\pazocal{D} = \{\{F_i\}_{i=1}^{N}, \{U_i\}_{i=1}^{N}, \{\Lambda_i\}_{i=1}^{N}\}$, with $F_i = \{(x_j^{(i)}, f_j^{(i)})\}_{j=1}^{N_{f}}$, $U_i = \{(x_j^{(i)}, u_j^{(i)})\}_{j=1}^{N_u}$, and $\Lambda_i = \{(x_j^{(i)}, \lambda_j^{(i)})\}_{j=1}^{N_{\lambda}}$.
After training the generator parameters $\beta$, we can generate new samples of the stochastic process $u_{\theta}(x)$ by generating samples of $\theta$ from its known distribution.
We refer to this method as PI-GAN.
Further, this technique can be combined with the posterior inference techniques of Sections~\ref{sec:uqt:bnns}-\ref{sec:uqt:ens} for obtaining multiple samples of the generator parameters $\beta$.
This is referred to as U-PI-GAN in this paper. Overall, total uncertainty in this case refers to aleatoric and epistemic uncertainties, as well as uncertainty due to stochasticity.
That is, for $M$ posterior samples of the parameters $\beta$ expressed as $\{\hat{\beta}_j\}_{j=1}^M$ and $M'$ stochastic realizations produced from each one, the mean prediction for each $x$ is given by
\begin{equation}\label{eq:methods:sdes:mean}
	\bar{\mu}(x) = \frac{1}{MM'}\sum_{j=1}^M\sum_{k=1}^{M'}u_{\theta_k}(x;\hat{\beta}_j ),
\end{equation} 	
and the total variance $\bar{\sigma}^2(x)$ for each $x$ is given by
\begin{equation}\label{eq:methods:sdes:var}
	\bar{\sigma}^2(x) =  \bar{\sigma}_a^2(x) + \underbrace{\frac{1}{MM'}\sum_{j=1}^M\sum_{k=1}^{M'}(u_{\theta_k}(x;\hat{\beta}_j )-\bar{\mu}(x))^2}_{\bar{\sigma}_e^2(x) + \bar{\sigma}_{stoc}^2(x)} .
\end{equation} 
For stochastic and partially unknown $\lambda$, the additional generator $\lambda_{\theta}(x; \beta)$ also parametrized by $\beta$ for simplicity, takes the same random input $\theta$ and is trained simultaneously with $u_{\theta}(x; \beta)$.
See the original paper \cite{yang2020physicsinformed} and Section~\ref{app:methods:fpriors} for more information, as well as Section~\ref{sec:comp:stochastic} for an example.

Further, PI-GAN can be combined with Gaussian process (GP) regression (Section~\ref{app:methods:gps}) for solving deterministic forward PDE problems. 
We refer to this approach as GP+PI-GAN\footnote{We use ``+'' here to emphasize that a GP is first employed for fitting $f$, and subsequently, a PI-GAN is \textit{separately} employed for fitting $u$ by treating the problem as stochastic.} and note that it is a novel contribution of this paper.
Specifically, first we fit a GP given the noisy $f$ data (Table~\ref{tab:problem:form}, part 1). 
Subsequently, we produce realizations of $f$ using this GP, which are used for fitting a PI-GAN as described above and as if the problem was stochastic. This produces statistics of $u$ that satisfy the PDE of Eq.~\eqref{eq:intro:piml:pinn:pde} and include epistemic uncertainty due to limited and noisy data of $f$. 

\begin{figure}[!ht]
	\centering
	\includegraphics[width=.7\linewidth]{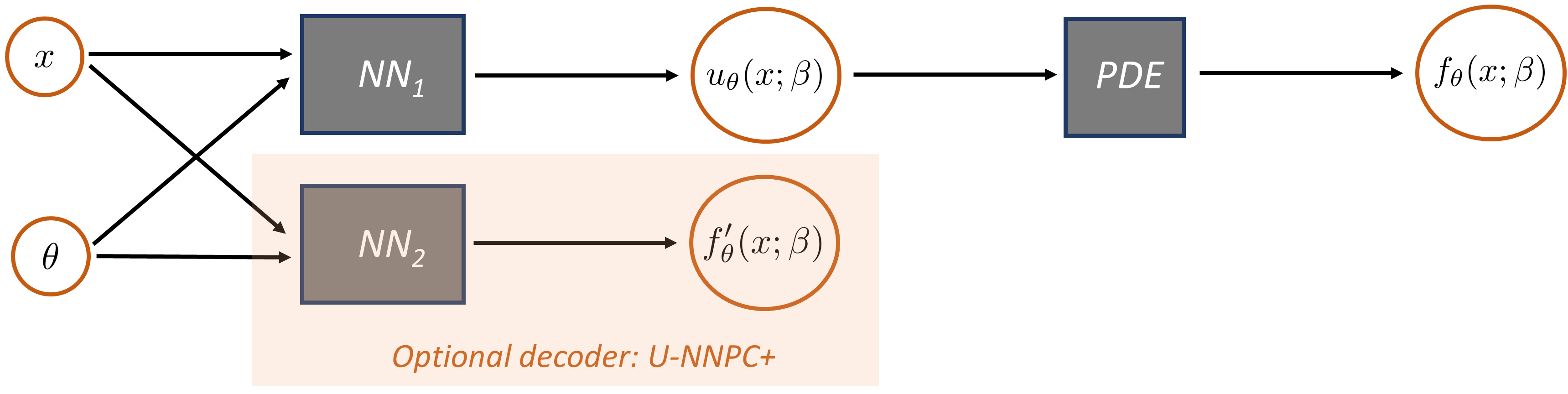}
	\caption{NNPC, developed by \cite{zhang2019quantifying}, and NNPC+ (orange box), proposed herein, for solving SPDEs.
		NNs 1 and 2, parametrized by $\beta$, combine the location in the domain $x$ to produce the modes of the stochastic representations of $u_{\theta}(x; \beta)$ and $f'_{\theta}(x; \beta)$, respectively.
		The model depicted in this figure can be construed as a special case of Fig.~\ref{fig:intro:all:techniques}; note, however, the different meaning of $\theta$ in this technique.
	}
	\label{fig:methods:sdes:nnpc}
\end{figure}

An alternative approach for solving SPDEs employs an arbitrary polynomial chaos expansion for the solution $u_{\theta}(x)$, and a PINN, parametrized by $\beta$, for approximating the corresponding stochastic modes.
Fig.~\ref{fig:methods:sdes:nnpc} provides an illustration of the model for the forward SPDE problem. 
For solving the mixed SPDE problem with partially unknown $\lambda$ and $u$, an additional NN is used for modeling the stochastic modes of $\lambda$. 
This technique can also be combined with the posterior inference techniques of Sections~\ref{sec:uqt:bnns}-\ref{sec:uqt:ens} for obtaining multiple samples of the parameters $\beta$ expressed as $\{\hat{\beta}_j\}_{j=1}^M$.
In this case, the mean for each $x$ and the total uncertainty including aleatoric, epistemic, and stochasticity uncertainties are given by Eqs.~\eqref{eq:methods:sdes:mean} and \eqref{eq:methods:sdes:var}, respectively.
This is referred to as U-NNPC in this paper. See the original papers \cite{zhang2019quantifying,zhang2020learning} and Section~\ref{app:methods:sdes} for more information, as well as Section~\ref{sec:comp:stochastic} for an example.
An extension to U-NNPC that we propose in this paper and refer to as U-NNPC+ pertains to employing an additional decoder NN specific to $f$.
This NN takes as input the encoding of the data of $f$, i.e., its low-dimensional representation expressed by $\theta$, and outputs $f'_{\theta}$ in Fig.~\ref{fig:methods:sdes:nnpc}.
More information regarding the training procedure of NNPC+ can be found in Section~\ref{app:methods:sdes}.

\subsection{Towards a unified view of the presented methods}\label{sec:uqt:uni}

\begin{figure}[!ht]
	\centering
	\includegraphics[width=.85\linewidth]{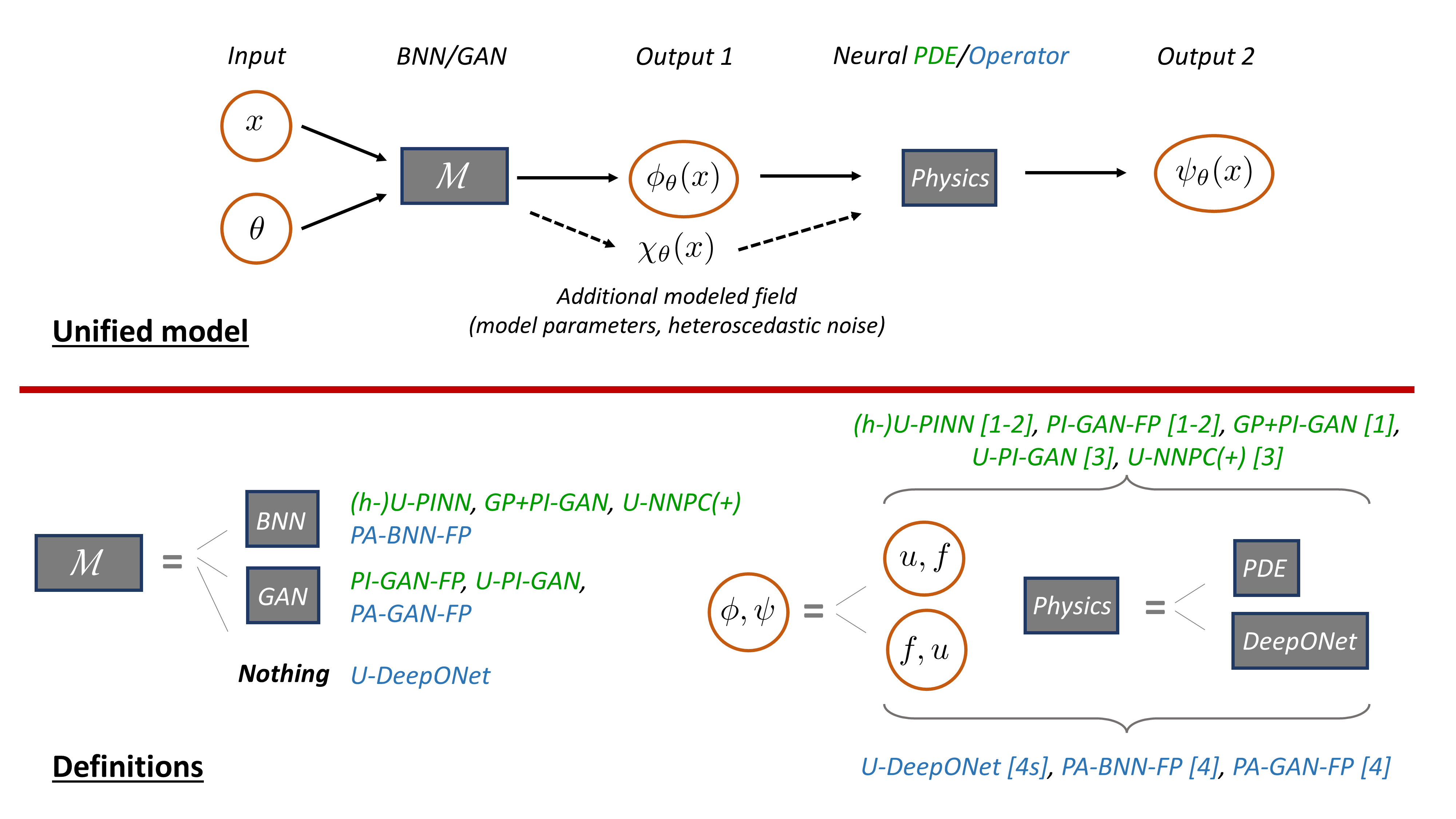}
	\caption{A conceptual model towards a unified view of the presented techniques for addressing the problems of Fig.~\ref{fig:intro:all:problems}.
	We divide the methods into the ones that utilize a PDE describing the underlying physics \textit{(neural PDEs)} and the ones that utilize a DeepONet as a surrogate model for the physics \textit{(neural operators)}.
	U-PINN and PI-GAN-FP address problems 1-2; GP+PI-GAN addresses problem 1; U-PI-GAN and U-NNPC(+) address problem 3; U-DeepONet, PA-BNN-FP, and PA-GAN-FP address problem 4.
	The \textbf{model} $\pazocal{M}$ can be a BNN or the generator of a GAN, or nothing.
	The \textbf{physics} can be a PDE or a pre-trained DeepONet.
	The \textbf{quantities} $\phi$, $\psi$ can be the solution $u$ and the source $f$, or vice-versa; $x$ represents the space-time input domain; and $\theta$ represents the parameters of the BNN or the input noise to the GAN generator.
	Further, ``$\chi$'' can represent the model parameters $\lambda$ in Eq.~\eqref{eq:intro:piml:pinn:pde} and/or heteroscedastic noise.
    The numbers in the brackets correspond to the problems of Fig.~\ref{fig:intro:all:problems}, with ``4s'' referring to the special case of problem 4 (Section~\ref{sec:intro:problem:form}).
	}
	\label{fig:intro:all:techniques}
\end{figure}

In this section, we summarize our methodology.
An overview of the presented methods for addressing the problems of Fig.~\ref{fig:intro:all:problems} is provided in Fig.~\ref{fig:intro:all:techniques}.
The neural PDE methods utilize a PDE describing the underlying physics of the problem, whereas the neural operator methods utilize a DeepONet as a surrogate model for the physics.
All methods utilize the posterior inference algorithms of Sections~\ref{sec:uqt:bnns}-\ref{sec:uqt:ens}.

\subsubsection{Neural PDEs: U-PINN, PI-GAN-FP, GP+PI-GAN, U-PI-GAN, and U-NNPC(+)}\label{sec:uqt:uni:pinn}

U-PINN (Sections~\ref{sec:uqt:bnns}-\ref{sec:uqt:ens}) combines the PINN methodology with posterior inference for solving mixed PDE problems.  
U-PINN is depicted in Fig.~\ref{fig:intro:all:techniques} with ``$x$'' denoting the space-time input domain, ``$\phi$'' the solution $u$, ``$\psi$'' the source term $f$, ``physics'' the PDE of Eq.~\eqref{eq:intro:piml:pinn:pde}, and ``$\pazocal{M}$'' a BNN parametrized by $\theta$.
The additional modeled field ``$\chi$'' can be used to represent the model parameters $\lambda$ in Eq.~\eqref{eq:intro:piml:pinn:pde} and/or heteroscedastic noise; see h-U-PINN in Table~\ref{tab:uqt:over} and Fig.~\ref{fig:uqt:bnns:dataunc:heteroscedastic:upinn}.
To solve a mixed PDE problem given $f$, $u$, and $\lambda$ data, we perform posterior inference for the PINN parameters $\theta$.

PI-GAN-FP (Section~\ref{sec:uqt:fpriors}) combines the PINN methodology with a GAN-FP and with posterior inference for solving mixed PDE problems.
PI-GAN-FP is depicted in Fig.~\ref{fig:intro:all:techniques} with ``$x$'', ``$\phi$'', ``$\psi$'', and ``physics'' as in U-PINN.
Further, ``$\pazocal{M}$'' is the generator of a GAN with random input $\theta$ that has a known distribution, e.g., Gaussian.
PI-GAN-FP is pre-trained using historical data.
Subsequently, given new limited and noisy $f$, $u$, and $\lambda$ data, the goal is to reconstruct $f$, $u$, and $\lambda$.
To solve the problem, we perform posterior inference for the GAN generator input $\theta$.

U-PI-GAN (Section~\ref{sec:uqt:sdes}) combines the PINN methodology with the GAN-based stochastic process representation of Section~\ref{sec:uqt:sdes} and with posterior inference for solving mixed SPDE problems.
U-PI-GAN is depicted in Fig.~\ref{fig:intro:all:techniques} with ``$x$'', ``$\phi$'', ``$\psi$'', ``physics'', ``$\pazocal{M}$'', and ``$\theta$'' as in PI-GAN-FP.
U-PI-GAN is given stochastic realizations of $f$, and potentially $u$, and the goal is to obtain the statistics of $u$ and of any unknown parameters. 
To solve the problem, we perform posterior inference for the parameters $\beta$ of U-PI-GAN.
Subsequently, we produce samples of $u$ by drawing $\theta$ samples from its known distribution and calculate the sought statistics. 

GP+PI-GAN (Section~\ref{sec:uqt:sdes}) combines a PI-GAN with GP regression for solving forward PDE problems.
Specifically, it fits a GP using noisy data of $f$, and subsequently produces realizations of $f$ and treats the problem as stochastic using PI-GAN.

Finally, U-NNPC(+) (Section~\ref{sec:uqt:sdes}) combines the PINN methodology with polynomial chaos and with posterior inference for solving mixed SPDE problems. 
U-NNPC(+) is depicted in Fig.~\ref{fig:intro:all:techniques} with ``$x$'', ``$\phi$'', ``$\psi$'', and ``physics'' as in U-PINN.
Further, ``$\pazocal{M}$'' is a BNN parametrized by $\beta$ and combines the location in the domain $x$ to produce the stochastic modes of $u_{\theta}(x; \beta)$.
Furthermore, $\theta$ represents a low-dimensional representation of the $f$-data.
U-NNPC(+) is given stochastic realizations of $f$, and potentially $u$, and the goal is to obtain the statistics of $u$ and of any unknown parameters. 
To solve the problem we perform posterior inference for the parameters $\beta$ of U-NNPC(+).
Subsequently, we calculate the statistics of the reconstructed $u$ realizations and of the unknown parameters. 

\subsubsection{Neural operators: U-DeepONet, PA-BNN-FP and PA-GAN-FP}\label{sec:uqt:uni:don}

U-DeepONet (Sections~\ref{sec:uqt:bnns}-\ref{sec:uqt:ens}) combines the DeepONet methodology with posterior inference for learning neural operators.
U-DeepONet is depicted in Fig.~\ref{fig:intro:all:techniques} with ``$x$'' denoting the space-time input domain, ``$\phi$'' the source term $f$, ``$\psi$'' the solution $u$, and ``physics'' the DeepONet parametrized by $\theta$.
Further, ``$\pazocal{M}$'' is discarded.
To solve the problem given a dataset of paired $u$ and $f$ realizations, we perform posterior inference for the DeepONet parameters $\theta$. 

PA-BNN-FP (Section~\ref{sec:uqt:fpriors}) combines the DeepONet methodology with an additional BNN and with posterior inference for learning neural operators.
PA-BNN-FP is depicted in Fig.~\ref{fig:intro:all:techniques} with ``$x$'', ``$\phi$'', and ``$\psi$'' as in U-DeepONet, while ``physics'' is a pre-trained DeepONet and ``$\pazocal{M}$'' is a BNN parametrized by $\theta$. 
After pre-training DeepONet with noiseless data, we are given new limited and noisy data of $f$ and $u$, and the goal is to reconstruct $f$ and $u$. 
To solve the problem, we perform posterior inference for the BNN parameters $\theta$.

Finally, PA-GAN-FP (Section~\ref{sec:uqt:fpriors}) combines the DeepONet methodology with a GAN-FP and with posterior inference for learning neural operators. 
PA-GAN-FP is depicted in Fig.~\ref{fig:intro:all:techniques} with ``$x$'', ``$\phi$'', ``$\psi$'', and ``physics'' as in PA-BNN-FP.
Further, ``$\pazocal{M}$'' is the generator of a GAN with random input $\theta$. 
After pre-training the DeepONet and the GAN-FP separately with noiseless data, we are given new limited and noisy data of $f$ and $u$, and the goal is to reconstruct $f$ and $u$. 
To solve the problem, we perform posterior inference for the GAN generator input $\theta$.

%% file: IN_table_techniques.tex
\begin{landscape}
\thispagestyle{mylandscape} 
	\begin{table}
	\centering
	\footnotesize
	\begin{tabular}{r|c|l|c|c|c|c|c|c}
		\toprule
		\multicolumn{9}{c}{\textbf{Methods for obtaining NN parameter posterior samples} $\{\hat{\theta}_j\}_{j=1}^M$ (or $\{\hat{\beta}_j\}_{j=1}^M$)} \\
		\midrule
		\multirow{2}{*}{Method full name }&\multirow{2}{*}{Method} & \hfil\multirow{2}{*}{Short description}\hfil & \multirow{2}{*}{\S} & In & In & In & In & In
		\\
		&&& &\S~\ref{sec:comp:func}& \S~\ref{sec:comp:pinns}& \S~\ref{sec:comp:stochastic}& \S~\ref{sec:comp:pinns:forw}& \S~\ref{sec:comp:don}
		\\
		\midrule
		Hamiltonian Monte Carlo &HMC &  Two-step MCMC $\vert$ simulates Hamiltonian dynamics  &  \ref{sec:uqt:bnns} &All & All &&All&\ref{sec:comp:don:fp}\\
		Langevin dynamics &LD &  One-step MCMC $\vert$ stochastic gradient descent with added noise & \ref{sec:uqt:bnns} & \ref{sec:comp:func:homosc:kno}&&&&\\
		Mean-field variational inference&MFVI & Approximates $p(\theta|\cD)$ by a Gaussian variational distribution $q_{\omega}(\theta)$ & \ref{sec:uqt:bnns} & All & \ref{sec:comp:pinns:stand}&&&\\ 
		Monte Carlo dropout &MCD & Standard training with some NN parameters set to zero in each step& \ref{sec:uqt:bnns} &\ref{sec:comp:func:homosc:kno}  & \ref{sec:comp:pinns:stand}&&&\\
		Laplace approximation &LA &Standard training with a Gaussian fitted to the discovered mode $\hat{\theta}$& \ref{sec:uqt:bnns} &\ref{sec:comp:func:homosc:kno}, \ref{sec:comp:func:homosc:unk} &&&&\\
		Deep ensemble &DEns & Standard training $M$ independent times & \ref{sec:uqt:ens}& 
		\ref{sec:comp:func:homosc:kno}, \ref{sec:comp:func:hetero}
		& \ref{sec:comp:pinns:stand}  &All&&\ref{sec:comp:don:dens}\\
		Snapshot ensemble &SEns &  Snapshots during standard training with learning rate schedule & \ref{sec:uqt:ens} &\ref{sec:comp:func:homosc:kno}&&&&\\
		Stochastic weight averaging-Gaussian &SWAG & SEns with a Gaussian distribution fitted to the snapshots & \ref{sec:uqt:ens}&\ref{sec:comp:func:homosc:kno}&&&&\\
		Heteroscedastic-X&h-X & Combines above methods with heteroscedastic modeling of noise & \ref{sec:uqt:pre:bma} & \ref{sec:comp:func:hetero}, \ref{app:comp:func:results:student} & \ref{sec:comp:pinns:hetero} &&& \\
		X+Bayesian neural network &X+BNN & Combines above methods with a BNN | \textit{Default setting} & \ref{sec:uqt}& All & All &All&All&All\\
		X+Functional prior &X+FP & Combines above methods with GAN functional prior instead of BNN & \ref{sec:uqt:fpriors}& \ref{sec:comp:func:hetero} & \ref{sec:comp:pinns:stand} &&&\ref{sec:comp:don:fp}\\
		\midrule
		\midrule
		\multicolumn{9}{c}{\textbf{Methods for learning neural (S)PDEs and neural operators}} \\
		\midrule
		\multirow{2}{*}{Method full name}&\multirow{2}{*}{Method} & \hfil\multirow{2}{*}{Short description}\hfil & \multirow{2}{*}{\S} & In & In & In & In & In
		\\
		&&&  & \S~\ref{sec:comp:func}& \S~\ref{sec:comp:pinns}& \S~\ref{sec:comp:stochastic}& \S~\ref{sec:comp:pinns:forw}& \S~\ref{sec:comp:don}
		\\
		\midrule
		Uncertain PINN &U-PINN & PINN combined with X+BNN  $\vert$ \textit{solves PDEs [1-2]} & \ref{sec:uqt:uni:pinn} & & All &&All&\\
		Heteroscedastic uncertain PINN &h-U-PINN & PINN combined with h-X+BNN  $\vert$ \textit{solves PDEs (heteroscedastic) [1-2]} & \ref{sec:uqt:bnns} & & \ref{sec:comp:pinns:hetero} &&&\\
		Physics-informed GAN functional prior &PI-GAN-FP & PINN combined with X+FP  $\vert$ \textit{solves PDEs [1-2]} & \ref{sec:uqt:uni:pinn} &&\ref{sec:comp:pinns:stand}&&&\\
		Gaussian process physics-informed GAN &GP+PI-GAN & Gaussian process combined with PINN and GAN  $\vert$ \textit{solves PDEs [1]} & \ref{sec:uqt:uni:pinn} &&&&All&\\
		Uncertain physics-informed GAN &U-PI-GAN & PINN combined with GAN and X+BNN $\vert$ \textit{solves SPDEs [3]} & \ref{sec:uqt:uni:pinn}&& &All&&\\
		Uncertain NN polynomial chaos (+) &U-NNPC(+) & PINN combined with polynomial chaos and X+BNN $\vert$ \textit{solves SPDEs [3]}& \ref{sec:uqt:uni:pinn}&&&All&&\\
		Uncertain DeepONet &U-DeepONet & DeepONet combined with X+BNN $\vert$ \textit{learns operators [4s]}& \ref{sec:uqt:uni:don}&& &&&\ref{sec:comp:don:dens}\\
		Physics-agnostic BNN functional prior &PA-BNN-FP & Pre-trained DeepONet combined with X+BNN $\vert$ \textit{learns operators [4]} & \ref{sec:uqt:uni:don}&& &&&\ref{sec:comp:don:dens}\\
		Physics-agnostic GAN functional prior &PA-GAN-FP & Pre-trained DeepONet combined with X+FP $\vert$ \textit{learns operators [4]} & \ref{sec:uqt:uni:don}&& &&&\ref{sec:comp:don:fp}\\
		\bottomrule
	\end{tabular}
	\caption{Overview of the UQ methods considered in this paper:
	The upper part corresponds to methods for obtaining NN parameter samples $\{\hat{\theta}_j\}_{j=1}^M$ (or $\{\hat{\beta}_j\}_{j=1}^M$).
	X represents one of the methods listed in the second column between entries ``HMC'' and ``SWAG''.
	The lower part corresponds to methods for learning (S)PDE solutions and neural operators by combining the methods in the upper part with PINNs and DeepONets.
	See also the conceptual model of Fig.~\ref{fig:intro:all:techniques}. 
	The numbers in the brackets correspond to the problems of Fig.~\ref{fig:intro:all:problems}, with ``4s'' referring to the special case of problem 4 (Section~\ref{sec:intro:problem:form}).
	The fourth column indicates the sections with the detailed descriptions of the methods.
	The right five columns serve to navigate through the comparative study of Section~\ref{sec:comp}.
	For example, ``All'' in the first row of the fifth column indicates that HMC has been used in all subsections of \ref{sec:comp:func}, while ``6.1.1'' in the second row means that LD has been used only in \ref{sec:comp:func:homosc:kno}. 
	}
	\label{tab:uqt:over}
\end{table}
\end{landscape}

%% file: IN_UQ_evaluation.tex
\subsection{Evaluation: Accuracy and uncertainty quality evaluation}\label{sec:eval:eval}

\begin{figure}[!ht]
	\centering
	\includegraphics[width=1\linewidth]{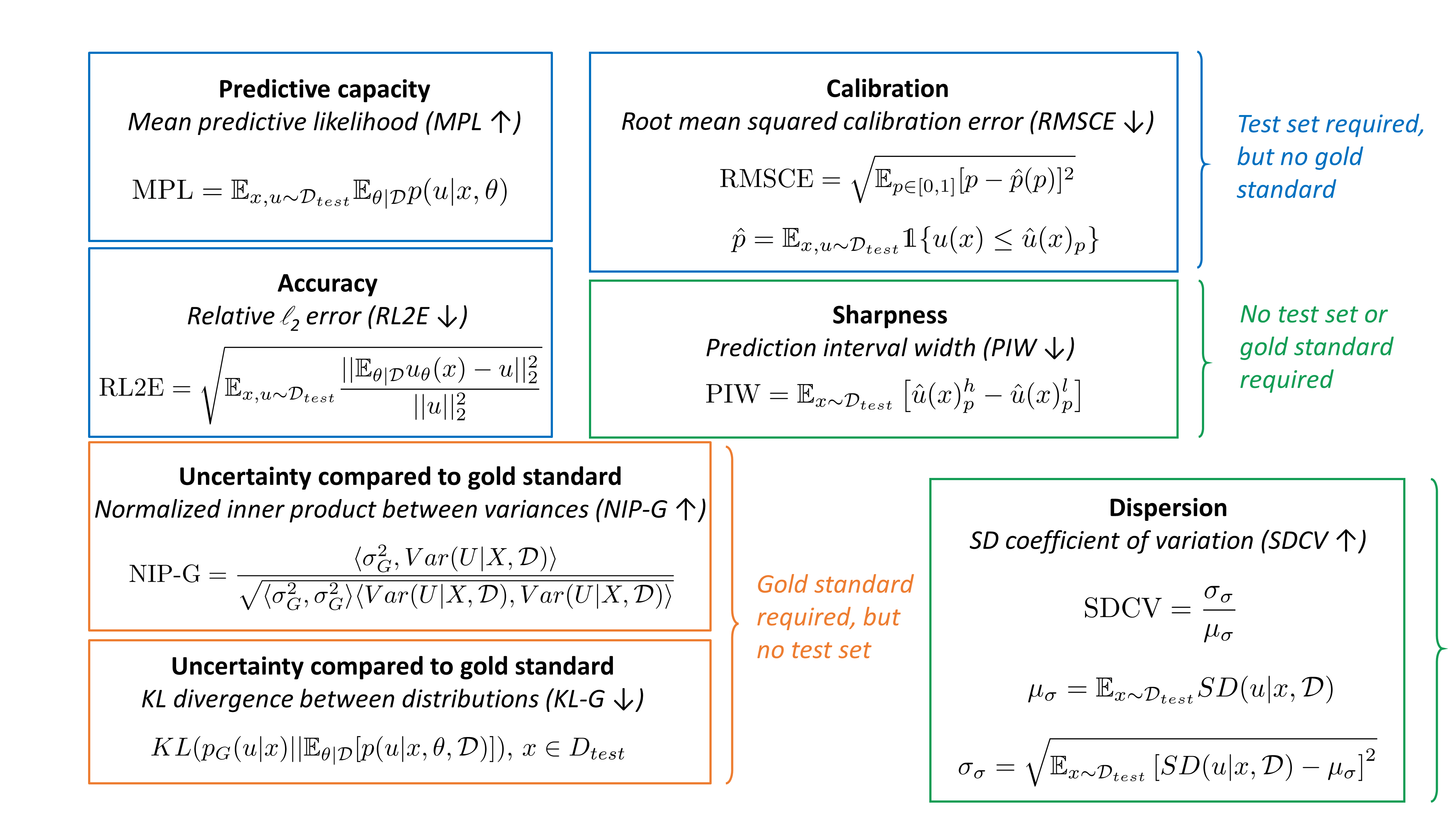}
	\caption{Accuracy and uncertainty quality evaluation metrics used in this paper.
		Blue boxes correspond to evaluation metrics that require a test set, orange boxes to metrics that require a ``gold standard'', e.g., for testing novel techniques, and green boxes to metrics that require neither a test set nor a gold standard.
		MPL measures general predictive capability, RL2E accuracy, and RMSCE statistical consistency between predictions and data.
		Further, the secondary metrics PIW and SDCV, presented in Section~\ref{app:eval:metrics:second}, measure prediction sharpness and dispersion across $x$ of uncertainty predictions, respectively.
		Finally, NIP-G and  KL-G, presented in Section~\ref{app:eval:metrics:gold} (see also \cite{rudy2021outputweighted}), measure the inner product between the predicted variance and the variance of gold standard, and the average distance between the predictive and gold standard distributions, respectively.
		A gold standard can be an accurate but computationally expensive technique or the exact solution if the problem is amenable to one.
		The up and down arrows next to the abbreviated names of the metrics indicate whether larger or smaller metric values correspond to better performance, respectively.
		}
	\label{fig:eval:eval:metrics}
\end{figure}

In this section we introduce a set of evaluation metrics, summarized in Fig.~\ref{fig:eval:eval:metrics}, to be used for model selection, e.g., prior/architecture optimization, comparison between UQ methods, and overall quality evaluation of the UQ design procedure.
Each of the metrics presented in this section may correspond to first- or second-order statistics of the predictive model, or to the whole distribution (PDF or CDF).
More information on evaluation metrics as well as comparative studies can be found in \cite{gneiting2007probabilistic,naeini2015obtaining, levi2019evaluating,ovadia2019can,yao2019quality,tran2020methods,ashukha2021pitfalls,chung2021uncertainty,chung2021pinball} and Appendix~\ref{app:eval}.

\subsubsection{Accuracy and predictive capacity}

As explained throughout the present paper, the BMA not only provides uncertainty estimates in the form of $p(u|x, \pazocal{D})$, but can also improve the accuracy of the mean predictions.
The mean for an arbitrary location $x$ is expressed as $\hat{u}(x)$ in Eq.~\eqref{eq:uqt:pre:bma:new:4} and is approximated by $\bar{\mu}(x)$ via Eq.~\eqref{eq:uqt:pre:mcestmc:mean} using samples $\{u_{\hat{\theta}_j}(x)\}_{j=1}^M$. 
In contrast, standard NNs predict $\hat{u}(x)$ as $u_{\hat{\theta}}$, where $\hat{\theta}$ is obtained through a point estimate as presented in Section~\ref{app:modeling:point}.
For evaluating the accuracy of the mean we use the relative $\ell_2$ error averaged over validation/test data, i.e., 
\begin{equation}\label{eq:eval:eval:rl2e}
	\text{RL2E}  = \sqrt{\mathbb{E}_{x, u \sim \cD_{test}}\frac{||\mathbb{E}_{\theta|\cD}u_{\theta}(x) - u||_2^2}{||u||_2^2}} 
	\approx \sqrt{\frac{\sum_{i=1}^{N_{test}}(\bar{\mu}(x_i)-u_i)^2}{\sum_{i=1}^{N_{test}}u_i^2}}.
\end{equation}
The validation/test data in Eq.~\eqref{eq:eval:eval:rl2e} can either be clean or noisy. 

Next, for predictive capacity evaluation a widely used metric is {\em mean predictive likelihood} given and approximated as
\begin{equation}\label{eq:eval:eval:MPL}
	\text{MPL} = \mathbb{E}_{x, u \sim \cD_{test}}\mathbb{E}_{\theta|\cD}p(u|x, \theta) 
	\approx
	\frac{1}{N_{test}}\sum_{i=1}^{N_{test}}
	\pazocal{N}(\bar{\mu}(x_i), \bar{\sigma}^2(x_i))[u_i].
\end{equation}
In Eq.~\eqref{eq:eval:eval:MPL} and for each $i \in \{1,\dots, N_{test}\}$, $\pazocal{N}(\bar{\mu}(x_i), \bar{\sigma}^2(x_i))[u_i]$ denotes the PDF value of the tested datapoint $u_i$ in the bracket, based on the Gaussian approximation $\bar{p}(u_i|x_i) \approx \pazocal{N}(\bar{\mu}(x_i), \bar{\sigma}^2(x_i))$ at the location $x_i$. 
High MPL values can be interpreted as the validation/test data being highly probable under the predictive distribution.

\subsubsection{Statistical consistency}\label{sec:eval:eval:stat}

\begin{figure}[!ht]
	\centering
	\includegraphics[width=.6\linewidth]{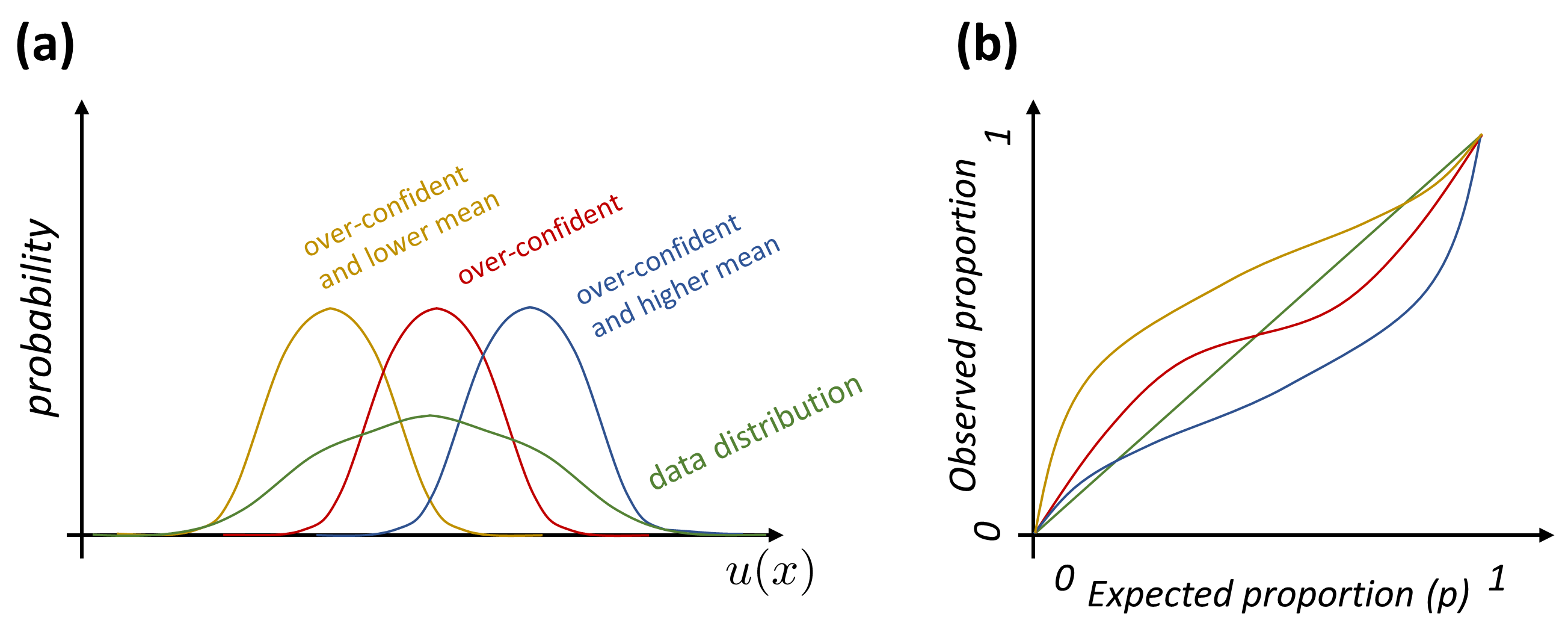}
	\caption{
	Sketch of a data-generating distribution and indicative over-confident predictive models (a), and the corresponding calibration plots (b), for a given $x$.
	The calibration plot, for a given $x$, can be used to visualize the relationship between ``expected proportion'' of datapoints inside a statistical interval, which is $p$, and ``observed proportion'', based on the predictive model, which is $P(U_x \leq \bar{P}_x^{-1}(p))$.
    Note, however, that in practice for each $x$ at most one sample $u(x)$ is available, and thus the calibration plot is produced ``on average'' for different values of $x$. The observed proportions can be approximated via $\mathbb{E}_{x, u \sim \cD_{test}} \mathds{1}\{u \leq \hat{u}(x)_p\}$, where $\hat{u}(x)_p$ is the inverse CDF value of the predictive model at $x$.
    An ``ideal'' predictive model matches the data distribution and its calibration plot is a straight line (green). An over-confident model (red) is typically above the green line for small values of $p$, and below the green line for larger values of $p$.
    The opposite is typically true for an under-confident model. Predictive models with different means compared to the data distribution (yellow, blue), may yield calibration plots that do not follow the aforementioned trends.
		}
	\label{fig:eval:eval:misscal}
\end{figure}

For evaluating the whole predictive distribution, we adopt a calibration metric that evaluates the statistical consistency between our predictions and data.
In the following, we assume that data is generated as $X, U_X \sim \mathbb{P}$, where $X$, $U_X$ are the random variables corresponding to the samples $x$, $u(x)$, and $\mathbb{P}$ is the data-generating distribution.  
A model that produces a CDF denoted by $\bar{P}(U_X|X)$ and abbreviated as $\bar{P}_X$ is called ``ideal'' if it matches the data-generating process, i.e., $\bar{P}(U_X|X) = \mathbb{P}(U|X)$ \cite{gneiting2007probabilistic}.
If $\bar{P}(U_X|X) \neq \mathbb{P}(U|X)$, the model may be over- or under-confident, depending on whether it predicts smaller or larger confidence intervals, respectively, as compared to the data-generating distribution.
However, because $\mathbb{P}$ is not known, a set of calibration conditions can be used in practice. 
The most commonly used condition is probabilistic calibration, which requires that $P(U_X \leq \bar{P}_X^{-1}(p)) = p$ is satisfied for any $p \in [0, 1]$ \cite{gneiting2007probabilistic,kuleshov2018accurate}.
Indicatively, for a given $x$ and for $p = 20 \%$, this condition requires that $20 \%$ of the observed datapoints $u(x)$ fall within the $20^{\text{th}}$ percentile of the predictive model. 
For a given $x$, the calibration plot can be used to visualize the relationship between ``expected proportion'' of datapoints inside a statistical interval, which is $p$, and ``observed proportion'', based on the predictive model.
The observed proportion is $P(U_x \leq \bar{P}_x^{-1}(p))$, where $U_x$ is the random variable of the sample $u(x)$, for a fixed $x$.
In Fig.~\ref{fig:eval:eval:misscal}, we provide a sketch of indicative over-confident predictive models, for a given $x$, and the corresponding calibration plots.
Note, however, that in practice for each $x$ at most one sample $u(x)$ is available. The observed proportions can be approximated via $\mathbb{E}_{x, u \sim \cD_{test}} \mathds{1}\{u \leq \hat{u}(x)_p\}$, where $\hat{u}(x)_{p} \approx \pazocal{N}^{-1}(\bar{\mu}(x), \bar{\sigma}^2(x))[p]$, and $\pazocal{N}^{-1}(\bar{\mu}(x), \bar{\sigma}^2(x))[p]$ denotes the inverse CDF value of the Gaussian predictive distribution $\pazocal{N}(\bar{\mu}(x), \bar{\sigma}^2(x))$ at a location $x$. 
As a result, the calibration plot is not produced for each $x$ separately, but ``on average'' for different values of $x$.
As a metric to measure calibration, we adopt the root mean squared calibration error (RMSCE) given and approximated as 
\begin{equation}\label{eq:eval:eval:approx:rmsce}
	\text{RMSCE} = \sqrt{\mathbb{E}_{p \in [0, 1]} (p - \hat{p}(p))^2} \approx \sqrt{\frac{1}{N_p}\sum_{j=1}^{N_p}\left[p_j - 
		\frac{1}{N_{test}}\sum_{i=1}^{N_{test}}
		\mathds{1}\{u_i \leq \hat{u}(x_i)_{p_j}\}
		\right]^2},
\end{equation}
where $\hat{p}(p)$ denotes the proportion of points falling inside the interval defined by $p$ \cite{kuleshov2018accurate}.
The interested reader is directed to Section~\ref{app:eval:metrics:second} for secondary metrics evaluating statistical consistency and to Section~\ref{app:eval:metrics:gold} for metrics comparing with a ``gold standard'' solution, e.g., an accurate but computationally expensive technique.

\subsection{Post-training improvement: Calibration}\label{sec:eval:calib}

Suppose that an additional dataset $\cD_r = \{x_i, u_i\}_{i=1}^{N_r}$ is available after training.
This can be a left-out calibration dataset, similar to validation and test sets.
In this section, we present three approaches for improving the calibration of function approximation predictive models after training, using the calibration dataset. Further, we extend these approaches for addressing SciML problems.  
These approaches can also prove useful for addressing model misspecification cases, e.g., when a wrong data noise model has been utilized (see also Section~\ref{app:modeling:postemp}).
More information on the active area of research related to calibration approaches for function approximation can be found in \cite{kuleshov2018accurate,levi2019evaluating,song2019distribution,cui2020calibrated,zelikman2020crude,zhao2020individual,chung2021uncertainty, rahaman2021uncertainty} and Appendix~\ref{app:eval}. 
Computational results can be found in Sections~\ref{sec:comp:func}, \ref{sec:comp:stochastic}, and \ref{sec:comp:don}.

A calibration approach proposed by \citet{levi2019evaluating} utilizes the calibration data for re-weighting optimally all predicted variances. 
Specifically, given a Gaussian predictive distribution $\pazocal{N}(\bar{\mu}(x), \bar{\sigma}^2(x))$ for every $x$ using samples $\{u_{\hat{\theta}_j}(x)\}_{j=1}^M$, the optimal constant value of $s$ to multiply the uncalibrated total uncertainty prediction is sought for. This is achieved by optimizing one of the related metrics of Section~\ref{sec:eval:eval}.
As a result, the calibrated predictive distribution for each $x$ becomes $\pazocal{N}(\bar{\mu}(x), s^2\bar{\sigma}^2(x))$.
Note that by using this approach only the variance is altered for each $x$ and not the mean prediction.
Next, an alternative approach that has become standard in function approximation calibration has been proposed by \citet{kuleshov2018accurate}.
Assume that the obtained predictive model is miscalibrated, i.e, it satisfies $Q(p) = P(U_X \leq \bar{P}_X^{-1}(p)) \neq p$ for every $p \in [0, 1]$.
It can be shown that by approximating $Q$ using calibration data, we can calibrate our model by applying the approximate $Q$ to all the predictive CDFs.
Clearly, this approach modifies both the mean and the variance of the predicted $u$ at each $x$.
More details can be found in Section~\ref{app:eval:calib:cdf}.
Finally, recall that when a Gaussian distribution $\pazocal{N}(\bar{\mu}(x), \bar{\sigma}^2(x))$ is used for making predictions as described in Section~\ref{sec:uqt:pre:bma}, the scaled residuals $\epsilon_u(x) = (u - \bar{\mu}(x)) / \bar{\sigma}(x)$ for arbitrary $x$, $u$ from the data-generating distribution are implicitly assumed to follow $\pazocal{N}(0, 1)$.
In this context, according to a calibration approach proposed by \citet{zelikman2020crude}, referred to as CRUDE, an empirical distribution is fitted to the scaled residuals that correspond to the calibration dataset.
By doing so, the mean and the variance of all the predicted distributions can be updated.
More details can be found in Section~\ref{app:eval:calib:crude}.
Clearly, all of the above methods modify the predictions of all values of $x$ in the same way, i.e., they calibrate the predictions for all points in the domain ``on average''. 
Also the calibration dataset is sampled from a given data distribution, which is usually the same as the training data distribution.
For these reasons, there are cases in which performance may deteriorate for either in-distribution evaluation, i.e., for test data following the training data distribution, or out-of-distribution evaluation, i.e., for test data following a different distribution.
See, for example, computational results in Table~\ref{tab:comp:func:homosc:kno:OOD}.

\begin{figure}[!ht]
	\centering
	\includegraphics[width=.9\linewidth]{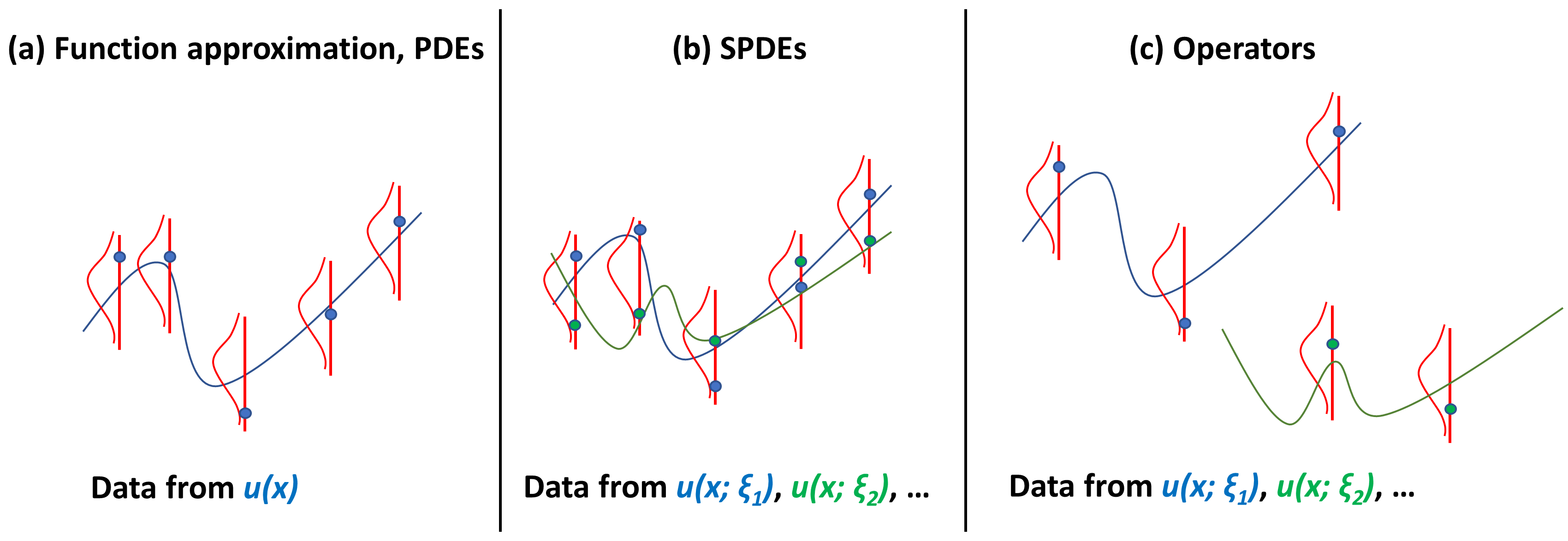}
	\caption{Post-training calibration of uncertainty predictions:
	(a) for function approximation and deterministic PDE problems, calibration can be applied using data from $u$ at different locations; 
	(b) for solving SPDEs, using data from realizations (denoted with $\xi$) of $u$ at different locations; 
	and (c) for neural operator learning, using data from $u$ functions evaluated at different locations and corresponding to different $\lambda$ inputs (or $f$, depending on which quantity is varying).}
	\label{fig:eval:calib:integ}
\end{figure}

For the case of solving mixed PDE problems with U-PINN, the calibration approaches can be separately applied to any of the quantities $f$, $b$, $u$, and $\lambda$ for which a left-out calibration dataset is available.
Next, for the case of solving mixed stochastic problems with U-PI-GAN or U-NNPC(+), the predictive distribution for each $x$ includes aleatoric and epistemic uncertainties as well as uncertainty from stochasticity of the modeled quantities, as explained in Sections~\ref{sec:uqt:sdes} and \ref{app:methods:sdes}. 
In this case, the calibration dataset includes data for different $x$ locations and from different stochastic realizations $\xi \in \Xi$. 
The calibration approaches can be separately applied to any of the quantities $f$, $b$, $u$, and $\lambda$ for which a calibration dataset containing left-out stochastic realizations is available;
see Section~\ref{sec:comp:stochastic} for an example.
Finally, we consider two cases of neural operator learning as explained in Section~\ref{sec:intro:problem:form}.
First, for the case where $f, b$, and $\lambda$ data are available at the complete set of locations used in pre-training, and the only sought outcome is the solution $u$, U-DeepONet produces a different $u$ predictive distribution for each $x$ and for each input $\lambda$ (for fixed and known $f, b$).
The calibration dataset must include $u$ data for various $x$ locations and corresponding to different inputs $\lambda$. 
Second, for the case where limited noisy data is available for $f, b, u$, and $\lambda$, PA-BNN-FP and PA-GAN-FP produce separate predictive distributions for each of the $f, b, u$, and $\lambda$ quantities. 
In this case, the calibration approaches can be separately applied to any of the quantities $f$, $b$, $u$, and $\lambda$ for which a left-out calibration dataset is available; see Section~\ref{sec:comp:don:fp} for an example.

%% file: IN_comp_intro.tex
In this section, we test the presented UQ methods for solving prototype problems in computational science and engineering.
In Section~\ref{sec:comp:func}, we consider a discontinuous function approximation problem, which we solve with more than ten methods. 
This pedagogical example serves as a testbed for demonstrating various features of our UQ methods, such as prior learning, heteroscedastic noise modeling, posterior tempering, functional priors, and post-training calibration.
In Section~\ref{sec:comp:pinns}, we consider a mixed PDE problem involving a time-dependent diffusion-reaction equation.
To solve this problem, we compare five different methods as well as four diverse problem setups, including extrapolation and heteroscedastic noise.
In Section~\ref{sec:comp:stochastic}, we consider a mixed problem involving a stochastic elliptic equation.
To solve this problem we compare and calibrate three methods, including the herein proposed U-NNPC+.  
In Section~\ref{sec:comp:pinns:forw}, we consider a forward PDE problem.
To solve this problem we compare two methods, including the herein proposed GP+PI-GAN. 
In Section~\ref{sec:comp:don}, we consider an operator learning problem with two different problem setups, as described in Section~\ref{sec:intro:problem:form}.
We use and calibrate various methods depending on the problem case.
We also propose methods to detect {\em out-of-distribution data}, which is critical for
risk-related applications. 
Finally, in Section~\ref{sec:comp:accvscost}, we briefly discuss the accuracy versus cost trade-off of the presented methods and also provide quantitative results pertaining to the function approximation example.
The hyperparameters we used as well as additional problem cases we tested can be found in Appendix~\ref{app:comp}.
In all problems, unless otherwise stated, a standard feed-forward NN architecture is used; see Section~\ref{app:comp:architecture} for architecture details.
Tables~\ref{tab:uqt:over} and \ref{tab:comp:contents} serve to navigate through the comparative study.

\input{IN_comp_contents}

To evaluate the performance of the employed UQ methods using the metrics of Section~\ref{sec:eval}, we use either in-distribution (ID) data, i.e., data sampled from the same distribution as the one used for training, and out-of-distribution (OOD) data, i.e., data sampled from an unseen distribution during training. Unless otherwise stated, the evaluation data is ID.
In addition, we present epistemic, aleatoric, and total uncertainties using the corresponding standard deviations and the $95 \%$ confidence intervals (CIs), which correspond to  approximately two standard deviations for Gaussian predictive distributions.
Lastly, we use calibration plots as illustrated in Fig.~\ref{fig:eval:eval:misscal}, and refer to UQ methods as over- or under-confident, depending on whether they
predict smaller or larger confidence intervals compared to the data-generating distribution, respectively.

%% file: IN_comp_contents.tex
\begin{table}[ht]
	\centering
	\footnotesize
	\begin{tabular}{c|c|c|c|c|c}
		\toprule
		\multicolumn{6}{c}{\textbf{Comparative study contents}}\\
		\midrule
		Problem & Function approximation & Mixed PDE & Mixed SPDE & Forward PDE & Operator learning \\
		\midrule
		Sections & \ref{sec:comp:func} $\&$ \ref{app:comp:func:results} & \ref{sec:comp:pinns} $\&$ \ref{app:comp:pinns:results}& \ref{sec:comp:stochastic} $\&$ \ref{app:comp:stochastic:results} & \ref{sec:comp:pinns:forw} $\&$ \ref{app:comp:pinns:forw:results} & \ref{sec:comp:don} $\&$ \ref{app:comp:don:lognormal}-\ref{app:comp:don:results:fp}   \\
		\midrule
		 \parbox[t]{2mm}{\multirow{14}{*}{\rotatebox[origin=c]{90}{Problem cases}}}& Known/homoscedastic & Standard case & \parbox[t]{2mm}{\multirow{14}{*}{\rotatebox[origin=c]{90}{Single case}}} & 15 datapoints & Noisy/limited  \\
		  &  Gaussian noise & $\sigma_u = \sigma_f = \sigma_{\lambda} = 0.05$&  & of source term $f$  & input data  \\
		 & \ref{sec:comp:func:homosc:kno} $\&$ \ref{app:comp:func:results:homosc:kno} & \ref{sec:comp:pinns:stand} $\&$ \ref{app:comp:pinns:results:standard}&  &  \ref{sec:comp:pinns:forw} & \ref{sec:comp:don:fp} $\&$ \ref{app:comp:don:results:fp} \\
		 \cline{2-3}\cline{5-6}
		 & Unknown/homoscedastic & Heteroscedastic noise  &  & 6 datapoints & Clean/complete   \\
		 &  Gaussian noise  & $\sigma_u = \sigma_f = \sigma_{\lambda} = 0.1|x|$  &  & of source term $f$ &  input data  \\
		 & \ref{sec:comp:func:homosc:unk} & \ref{sec:comp:pinns:hetero} &  & \ref{app:comp:pinns:forw:results} & \ref{sec:comp:don:dens}  \\
		 \cline{2-3}\cline{5-6}
		 & Heteroscedastic & Large noise   &  &  & OOD data \\
		  & Gaussian noise  & $\sigma_u = \sigma_f = \sigma_{\lambda} = 0.1$  &  &  & detection  \\
		 & \ref{sec:comp:func:hetero} & \ref{app:comp:pinns:results:large} &  &  &  \ref{sec:comp:don:ood}  \\
		 \cline{2-3}
		 \cline{6-6}
		 & Heteroscedastic & Extrapolation case &  &  &   \\
		 
 		 & Student-t noise& \ref{app:comp:pinns:results:extra}  &  &  &   \\
 		 \cline{3-3}
  		 &  \ref{app:comp:func:results:student} & Steep boundary layers &  &  &   \\
  		 \cline{2-2}
  		 & & \ref{app:comp:pinns:results:steep} &  &  &   \\
		\bottomrule
	\end{tabular}
	\caption{Problems and corresponding sections of the comparative study. For each example, we provide the sections that include the problem description, results, and hyperparameters, as well as the sections that include the various problem cases that we considered and the corresponding sub-sections, while ``single case'' corresponds to examples with only one problem case.
	This table in conjunction with Table~\ref{tab:uqt:over} serves to navigate through the comparative study. 
	In Appendix~\ref{app:comp}, we include additional results, implementation details, as well as the hyperparameters and NN architectures that we used.
	}
	\label{tab:comp:contents}
\end{table}

%% file: IN_function_example.tex
We consider the following one-dimensional discontinuous function
\begin{equation}\label{eq:comp:func:func}
	u(x) = \left\{
	\begin{aligned}
		&\frac{1}{2}[\sin^3(2 \pi x) - 1], ~ -1 \le x < 0, \\
		&\frac{1}{2}[\sin^3(3 \pi x) + 1], ~ 0 \le x \le 1.
	\end{aligned}
	\right.
\end{equation}
The training dataset $\pazocal{D} = \{x_i, u_i\}_{i=1}^{N}$ consists of $N = 32$, unless otherwise specified, equidistant measurements of Eq.~\eqref{eq:comp:func:func} at $x \in [-0.8, -0.2] \cup [0.2, 0.8]$.
The data is contaminated with zero-mean noise. 
In Section~\ref{sec:comp:func:homosc:kno}, the data noise is considered Gaussian with constant along $x$ (homoscedastic) and known scale; in Section~\ref{sec:comp:func:homosc:unk}, Gaussian with unknown constant scale; in Section~\ref{sec:comp:func:hetero}, Gaussian with varying along $x$ (heteroscedastic) and unknown scale; and in Section~\ref{app:comp:func:results:student}, Student-t with unknown varying scale. 
The hyperparameters and the NN architectures that we used are summarized in Section~\ref{app:comp:hyperparameters} and \ref{app:comp:architecture}, respectively.

\subsubsection{Known homoscedastic noise}\label{sec:comp:func:homosc:kno}

In this section, we solve the function approximation problem of Eq.~\eqref{eq:comp:func:func} with known and constant noise that follows $\cN(0, 0.1^2)$. 
The considered techniques include HMC, LD, MFVI, and LA from Section~\ref{sec:uqt:bnns} with fixed prior distribution $\pazocal{N}(0, 1)$ for all NN parameters; MCD with tuned dropout rate; as well as DEns, SEns, and SWAG from Section~\ref{sec:uqt:ens} with tuned weight decay constant, learning rate, and cycle length, where applicable. 
For obtaining a form of reference solution, we consider the ``analogous'' GP that corresponds to BNN with $\pazocal{N}(0, 1)$ as prior distribution, i.e., we numerically obtain a GP kernel that induces a FP that approximates the BNN-FP (see, e.g., \cite{yang2021bpinns}).
Using this derived kernel we can perform exact Bayesian inference using GP regression with the available data; see Section~\ref{app:methods:gps}.
Note, however, that this does not mean that a GP with a carefully selected kernel can replace BNNs in all cases. 
For example, BNNs can provide improved performance by learning a representation of the data and result in significant computational savings for large datasets; see, e.g., \cite{mackay1997gaussian,matthews2018gaussian, nalisnick2018priors} for extended discussions. 
Further, BNNs can be integrated into SciML in a straightforward manner without linearization of the problem; see \cite{pang2020physicsinformed} and Section~\ref{sec:uqt}. 

\begin{table}[!ht]
	\centering
	\footnotesize
	\begin{tabular}{c|ccccccccc}
		\toprule
		Metric ($\times 10^2$)
		& GP & HMC & LD & MFVI & MCD & LA & DEns & SEns & SWAG \\
		\midrule
		RL2E ($\downarrow$) & 22.1 & 22.9 & 23.5 & 29.2 & \textbf{21.9} & 23.1 & 23.1 & 22.1 & 26.2 \\ 
		MPL ($\uparrow$) & 228.3 & 225.8 & 103.3 & 121.6 & 237.8 & 209.5 & 235.0 & \textbf{241.6} & 109.0 \\ 
		RMSCE ($\downarrow$) & \textbf{6.9} & 7.2 & 15.5 & 8.8 & 8.6 & 7.1 & 8.3 & 7.2 & 13.6 \\ 
		\midrule
		Cal RMSCE ($\downarrow$) & 6.8 & 6.7 & 6.4 & \textbf{3.8} & 8.0 & 6.5 & 6.5 & 6.9 & 5.7 \\ 
		PIW ($\downarrow$) & 45.8 & 45.2 & 142.0 & 116.2 & 42.5 & 51.9 & \textbf{40.5} & 41.9 & 132.0 \\ 
		SDCV ($\uparrow$) & 3.3 & 4.6 & 2.8 & \textbf{38.2} & 4.6 & 29.6 & 6.0 & 6.3 & 8.8 \\ 
		\bottomrule
	\end{tabular}
	\caption{
		Function approximation problem of Eq.~\eqref{eq:comp:func:func} | \textit{Known homoscedastic noise}: the most competitive techniques are the GP, HMC, MCD, DEns and SEns.
		Post-training calibration makes all techniques approximately equally calibrated.
		To evaluate the performance of the employed UQ methods we split the $x$ domain into in-distribution (ID) data, i.e., a union of sub-domains $[-0.8, -0.2] \cup [0.2, 0.8]$ that was used for training, and out-of-distribution (OOD) data, i.e., $[-1.0, -0.8] \cup [-0.2, 0.2] \cup [0.8, 1.0]$ that was unseen during training.
		Here we evaluate ID performance of nine UQ methods based on the metrics of Section~\ref{sec:eval}, using noisy test data and uncalibrated predictions, except for calibrated (Cal) RMSCE.
		See also Figs.~\ref{fig:comp:func:homosc:kno:epist:hmc} and \ref{fig:comp:func:homosc:kno:epist:mcd} for the disandvantages of MCD compared to HMC, as well as Table~\ref{tab:comp:func:homosc:kno:OOD} for OOD evaluation.
		The up and down arrows next to the names of the metrics indicate whether larger or smaller metric values correspond to better performance, respectively.
	}
	\label{tab:comp:func:homosc:kno:ID}
\end{table}

\begin{figure}[!ht]
	\centering
	\subcaptionbox{}{}{\includegraphics[width=0.32\textwidth]{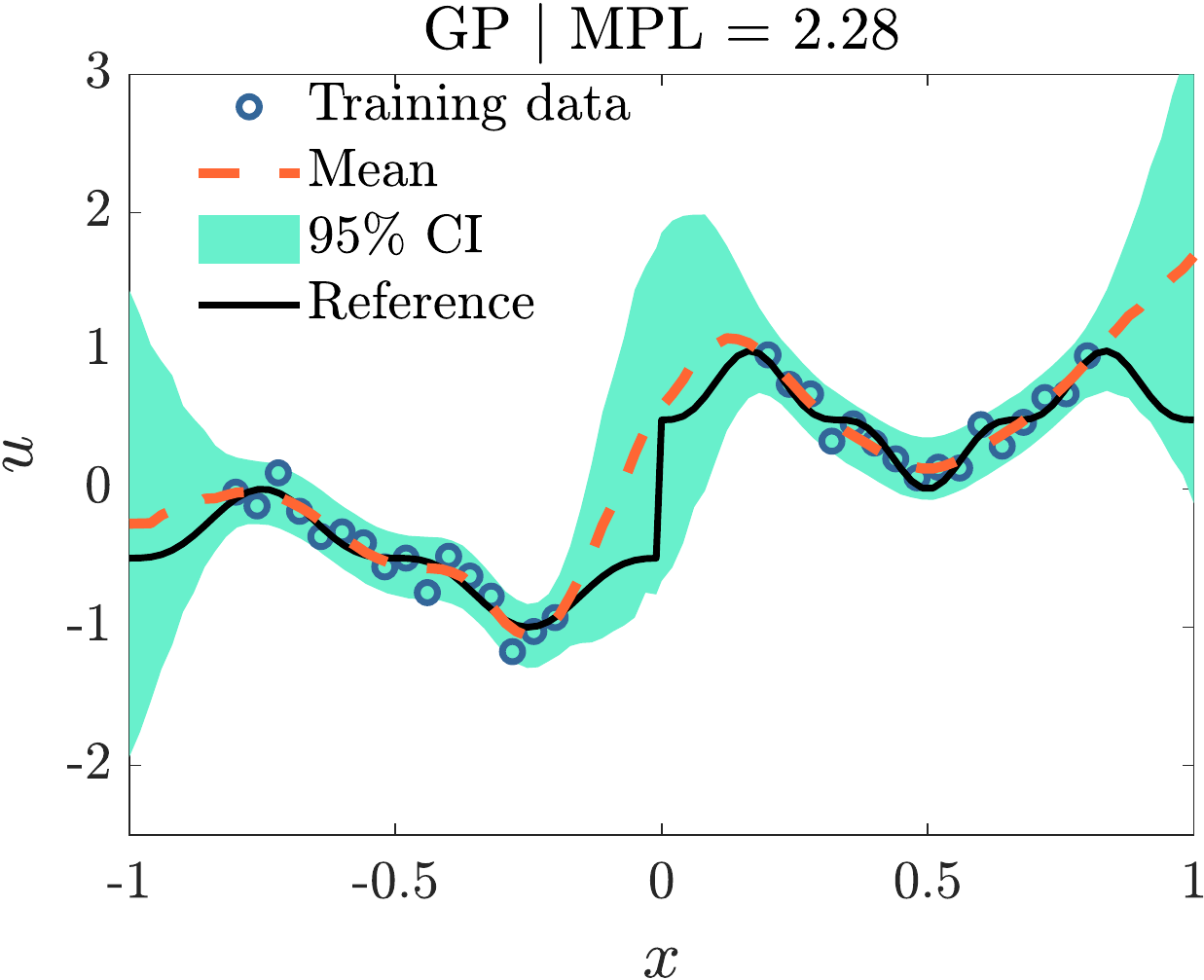}}
	\subcaptionbox{}{}{\includegraphics[width=0.32\textwidth]{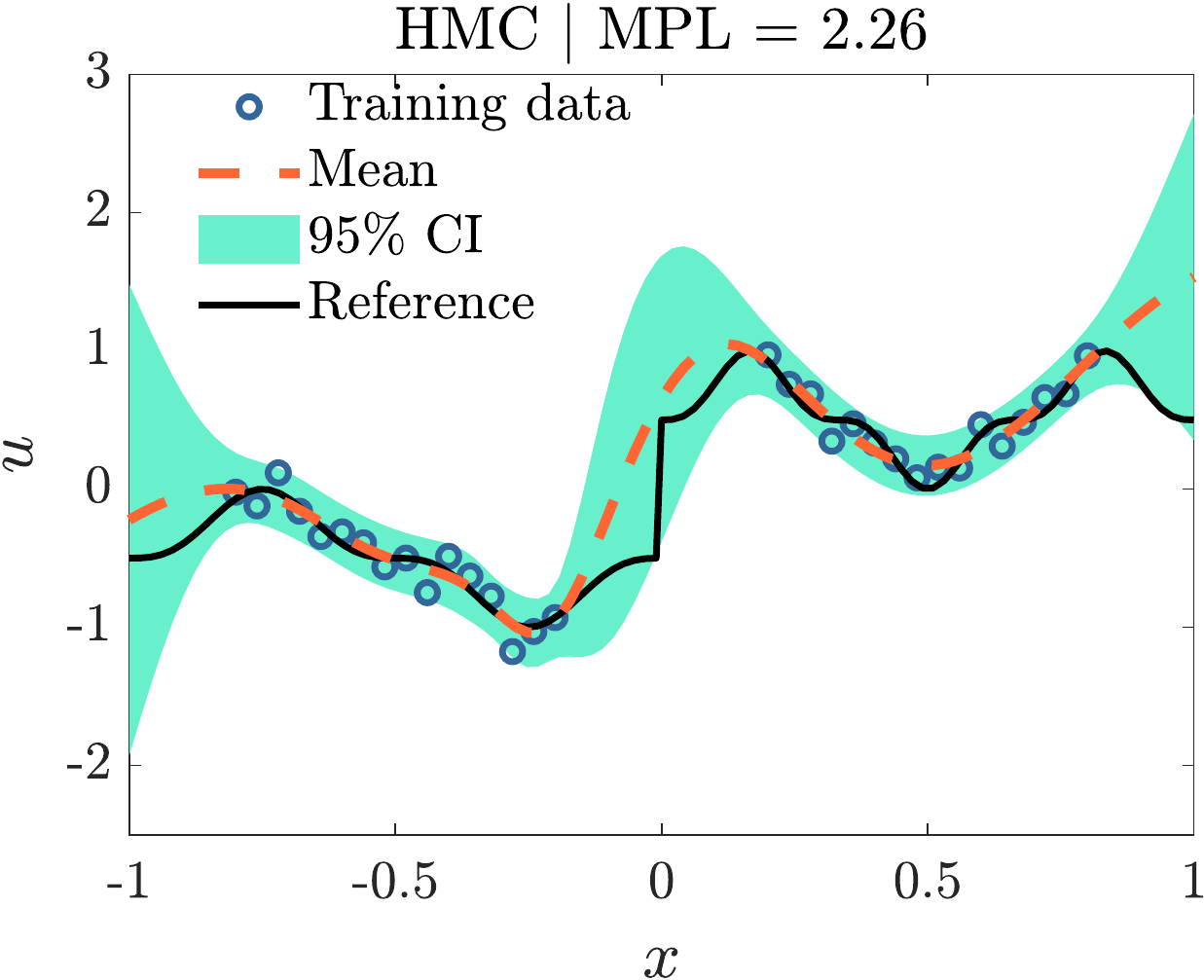}}
	\subcaptionbox{}{}{\includegraphics[width=0.32\textwidth]{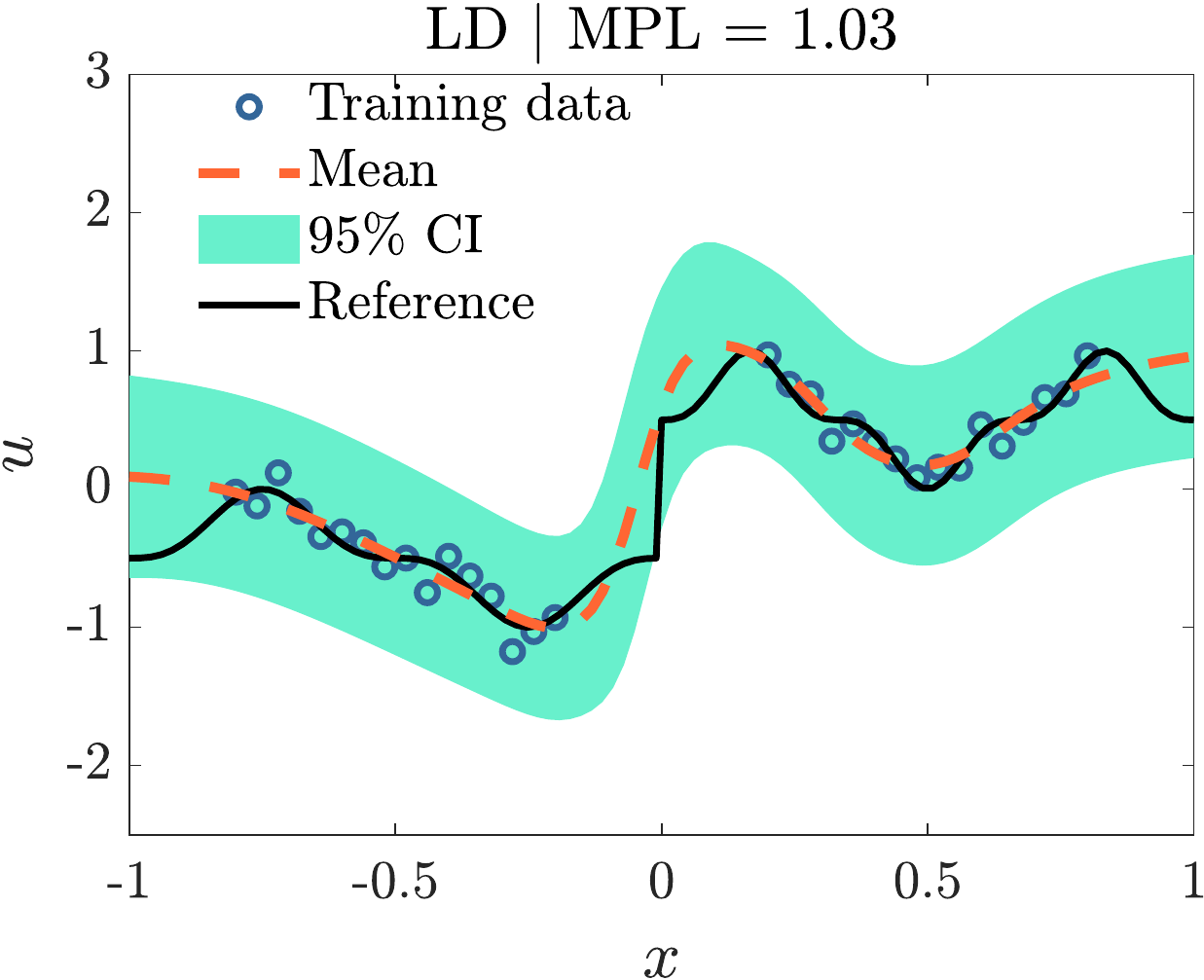}}
	\subcaptionbox{}{}{\includegraphics[width=0.32\textwidth]{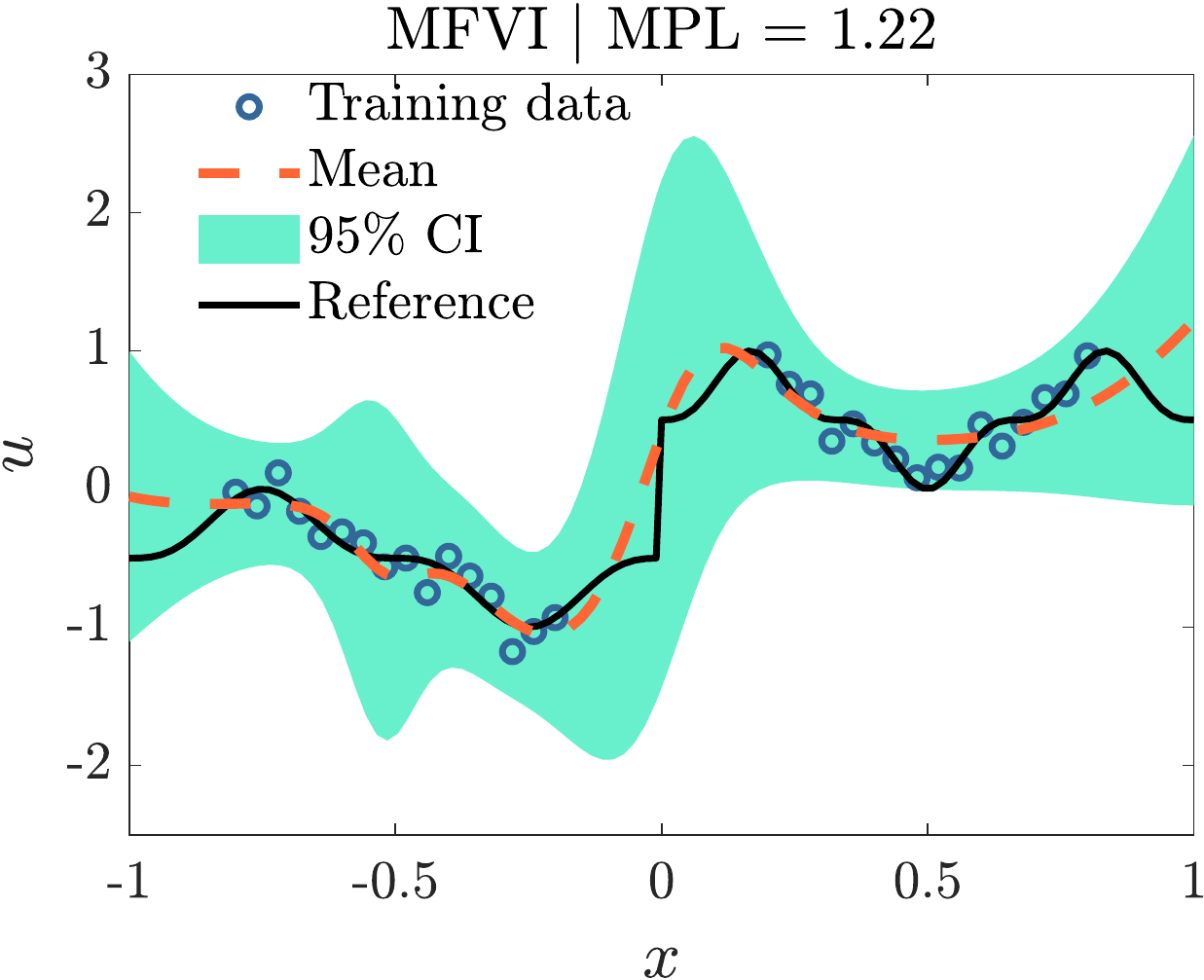}}
	\subcaptionbox{}{}{\includegraphics[width=0.32\textwidth]{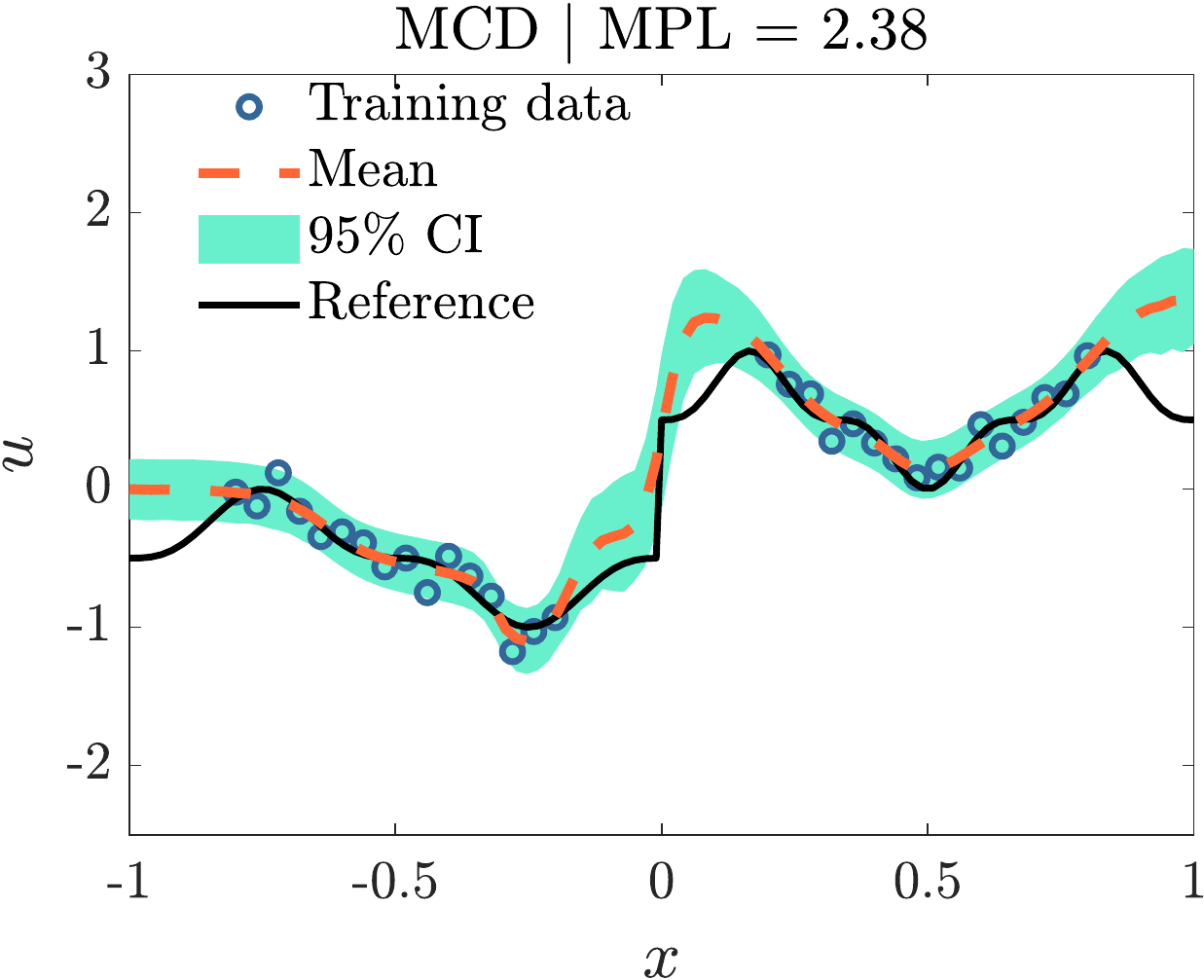}}
	\subcaptionbox{}{}{\includegraphics[width=0.32\textwidth]{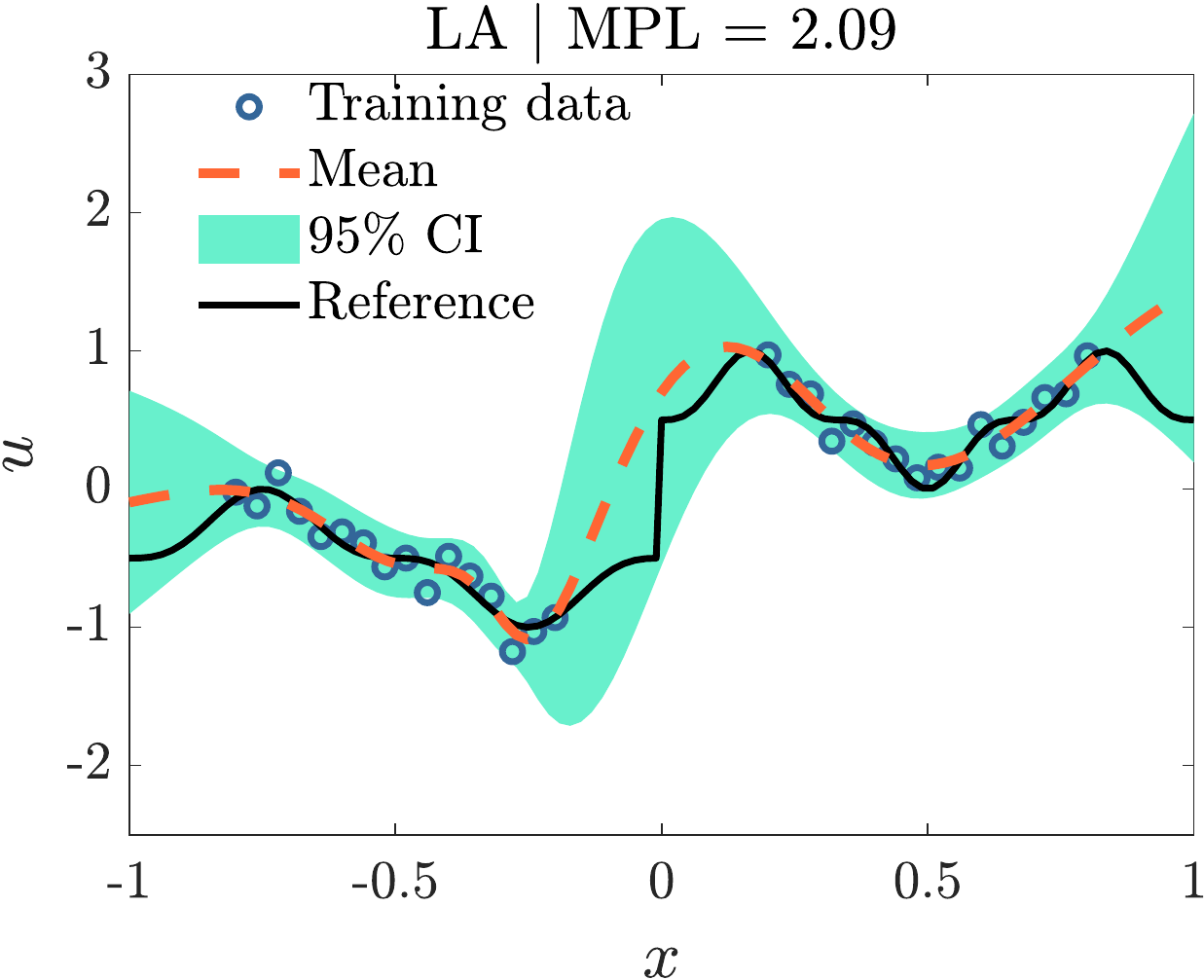}}
	\subcaptionbox{}{}{\includegraphics[width=0.32\textwidth]{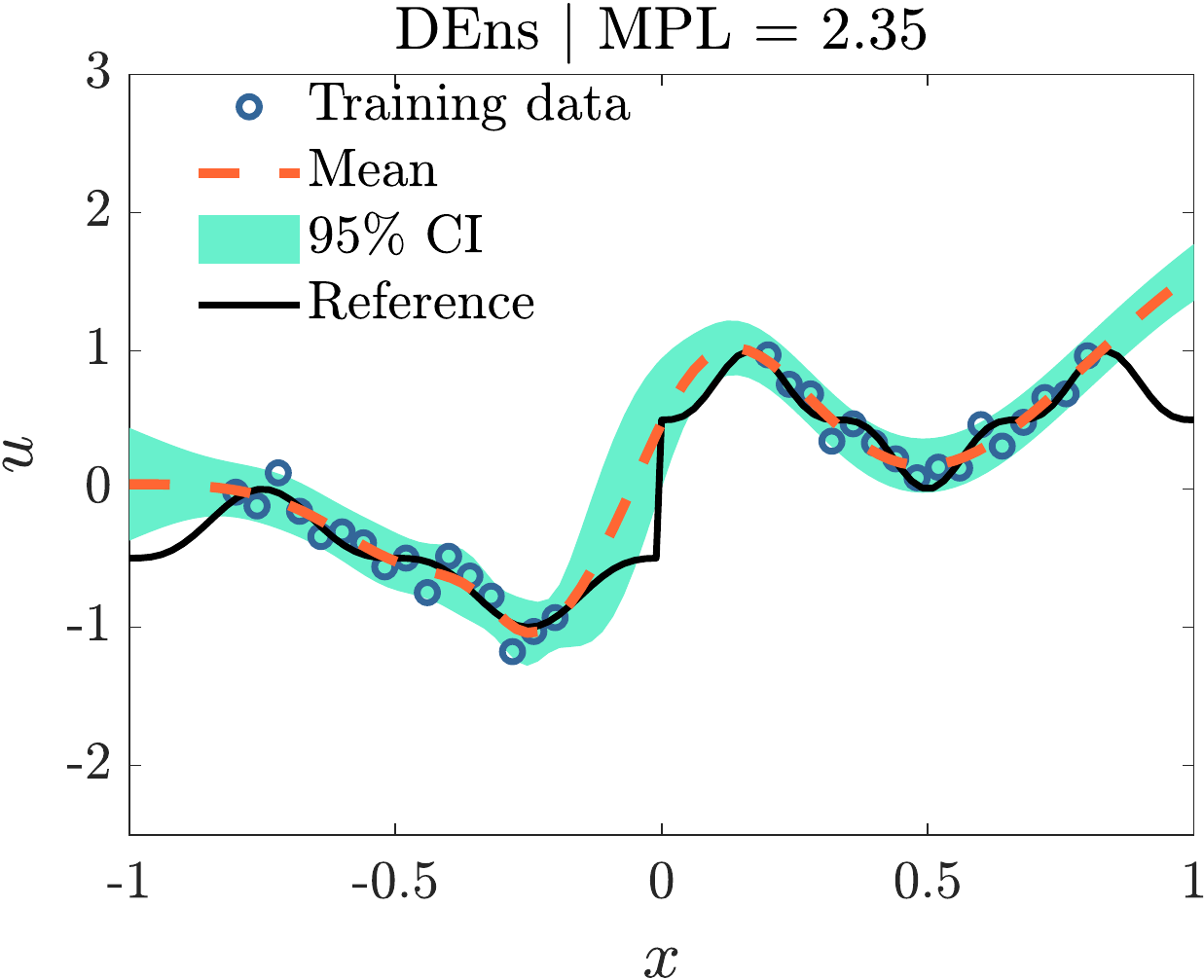}}
	\subcaptionbox{}{}{\includegraphics[width=0.32\textwidth]{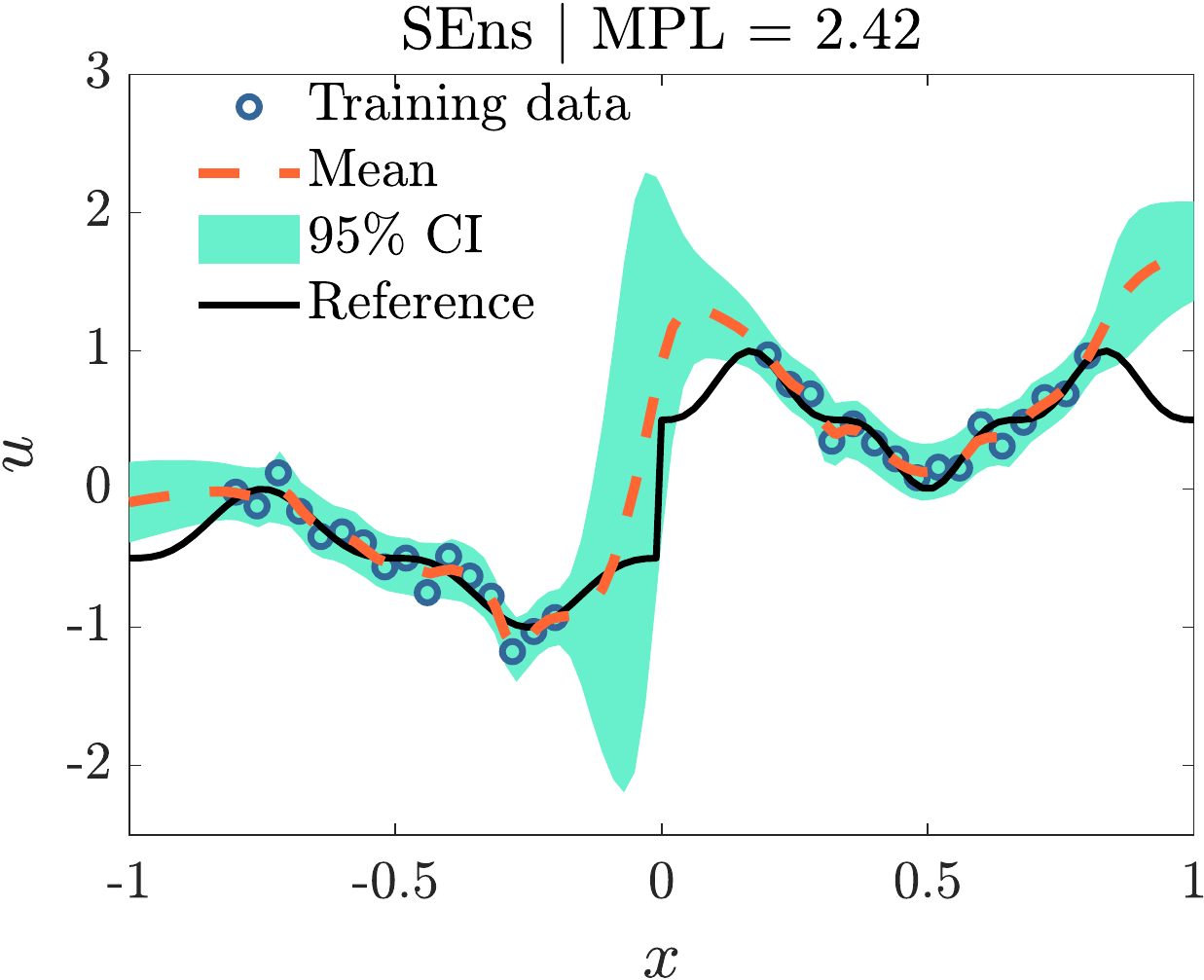}}
	\subcaptionbox{}{}{\includegraphics[width=0.32\textwidth]{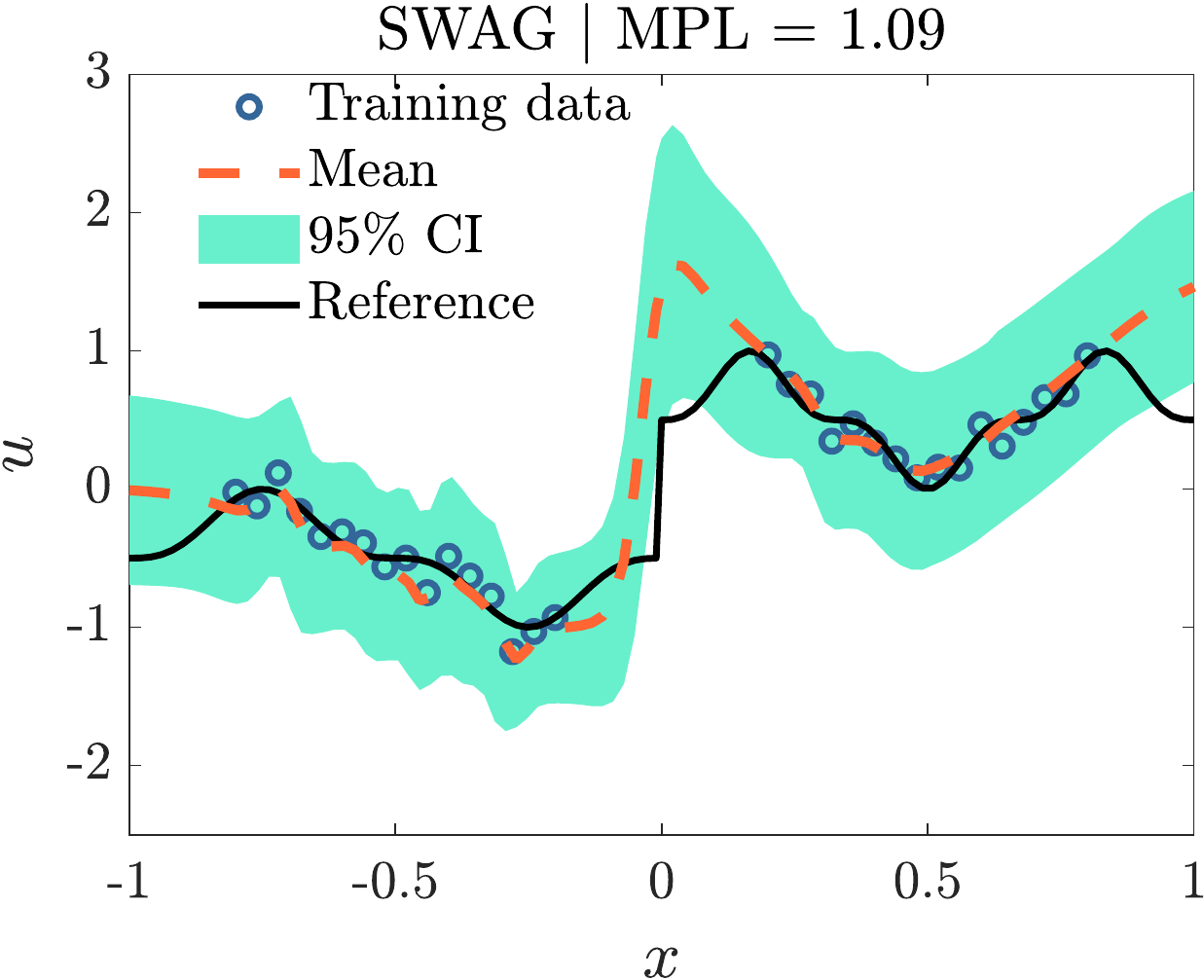}}
	\caption{
		Function approximation problem of Eq.~\eqref{eq:comp:func:func} | \textit{Known homoscedastic noise}:
		the most competitive techniques are the GP, HMC, MCD, and SEns.
		However, the uncertainties of MCD and SEns do not increase consistently with OOD data. 
		Shown here are the training data and exact function, as well as the mean and total uncertainty ($95\%$ confidence interval, CI) predictions without post-training calibration, as well as corresponding MPL values (higher is better). 
		The results correspond to the nine posterior inference methods of Table~\ref{tab:uqt:over}.
		Note that the total uncertainty estimates cover in most cases the point-wise errors within the $95\%$ CIs (approximately two standard deviations).
		See also accompanying Figs.~\ref{fig:comp:func:homosc:kno:res:1}-\ref{fig:comp:func:homosc:kno:res:3} including calibration plots and predictions with post-training calibration.
	}
	\label{fig:comp:func:homosc:kno:res}
\end{figure}

In Fig.~\ref{fig:comp:func:homosc:kno:res} we present the mean and total uncertainty predictions obtained by nine UQ methods. 
In Table~\ref{tab:comp:func:homosc:kno:ID} we also provide the corresponding evaluation metrics for ID evaluation, i.e., for values of $x \in [-0.8, -0.2] \cup [0.2, 0.8]$, whereas OOD evaluation, i.e., for values of $x \in [-1.0, -0.8] \cup [-0.2, 0.2] \cup [0.8, 1.0]$, is provided in Table~\ref{tab:comp:func:homosc:kno:OOD}.
The most competitive techniques for this problem are the GP,
HMC, MCD, DEns, and SEns.
However, the uncertainties of MCD and SEns do not increase consistently with OOD data.
This is important in risk-sensitive applications, where computational methods should be able to distinguish between parts of the input domain corresponding to interpolation and parts corresponding to extrapolation.
OOD evaluation as well as predictions following post-training calibration can be found in Section~\ref{app:comp:func:results:homosc:kno}. 
In addition, the secondary metrics PIW and SDCV of Section~\ref{app:eval:metrics:second} can be used for comparisons between equally calibrated techniques.
For example, the RMSCEs of MFVI and MCD are 8.8 and 8.6, respectively, while their SDCVs are 38.2 and 4.6. That is, the uncertainty predictions of MFVI vary more along $x$ compared to the predictions of MCD. 
This is used as a diagnostic tool for assessing which method is expected to perform better for OOD predictions (see, e.g., \cite{gneiting2007probabilistic,levi2019evaluating}).
In this regard, by comparing parts (d) and (e) of Fig.~\ref{fig:comp:func:homosc:kno:res}, we note that the uncertainty of MFVI increases for OOD data, whereas the uncertainty of MCD does not.
Similarly, LD has low SDCV score and its uncertainty is the same for both ID and OOD data.
Nevertheless, MFVI has a much larger PIW score compared to MCD, which indicates that its predictions may be under-confident. 
This property may be considered as safe and desirable in some applications, or as wasteful in some others. 

\begin{figure}[!ht]
	\centering
	\subcaptionbox{}{}{\includegraphics[width=0.32\textwidth]{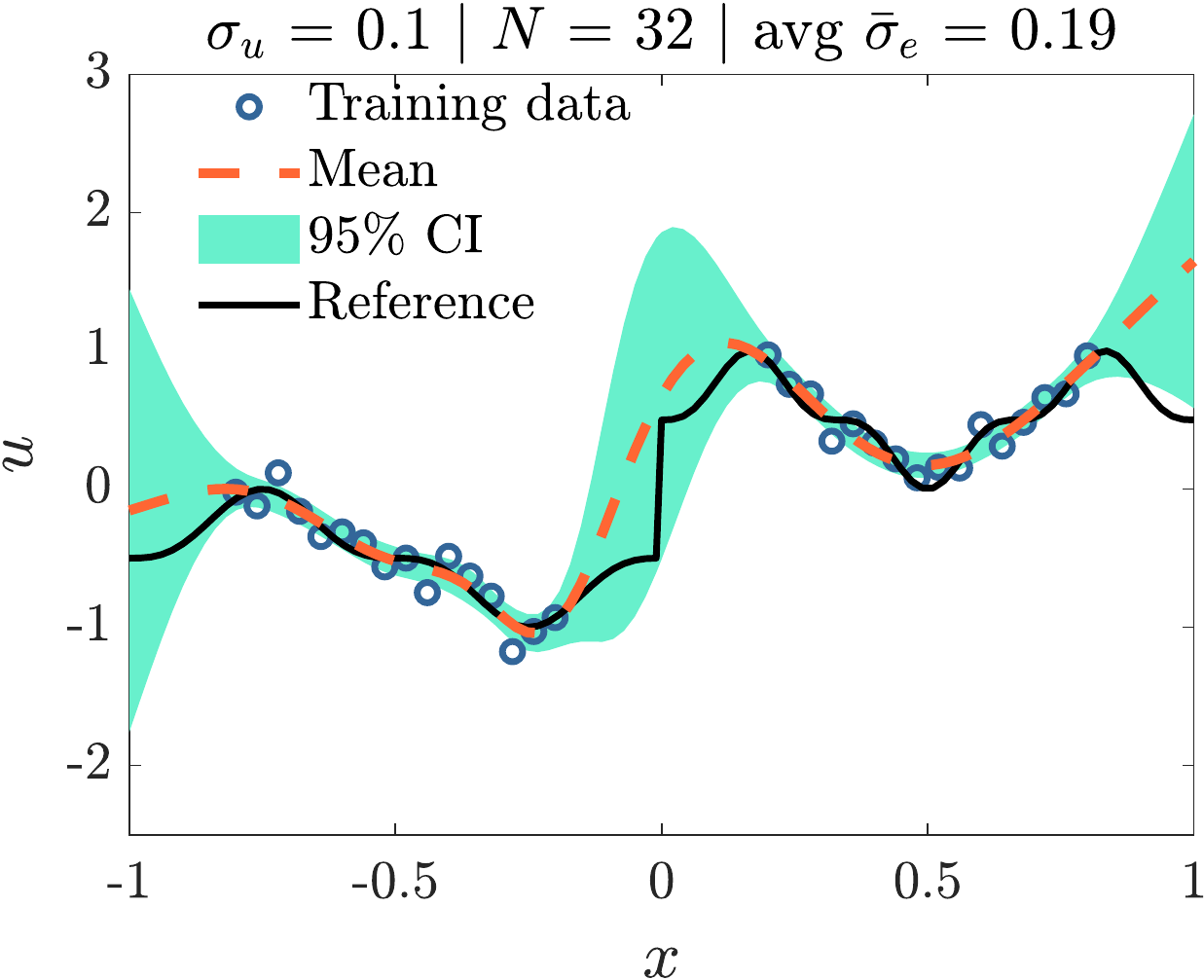}}
	\subcaptionbox{}{}{\includegraphics[width=0.32\textwidth]{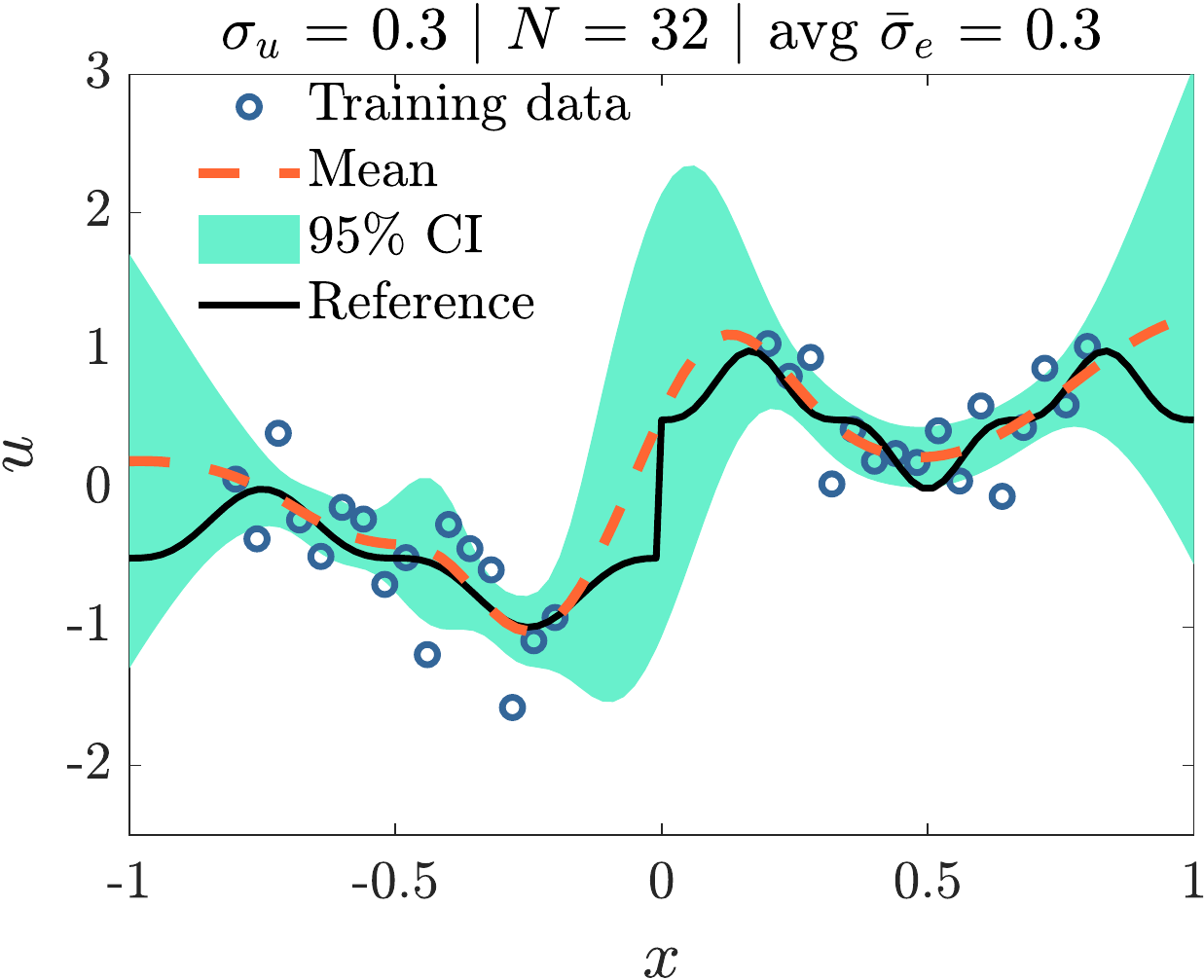}}
	\subcaptionbox{}{}{\includegraphics[width=0.32\textwidth]{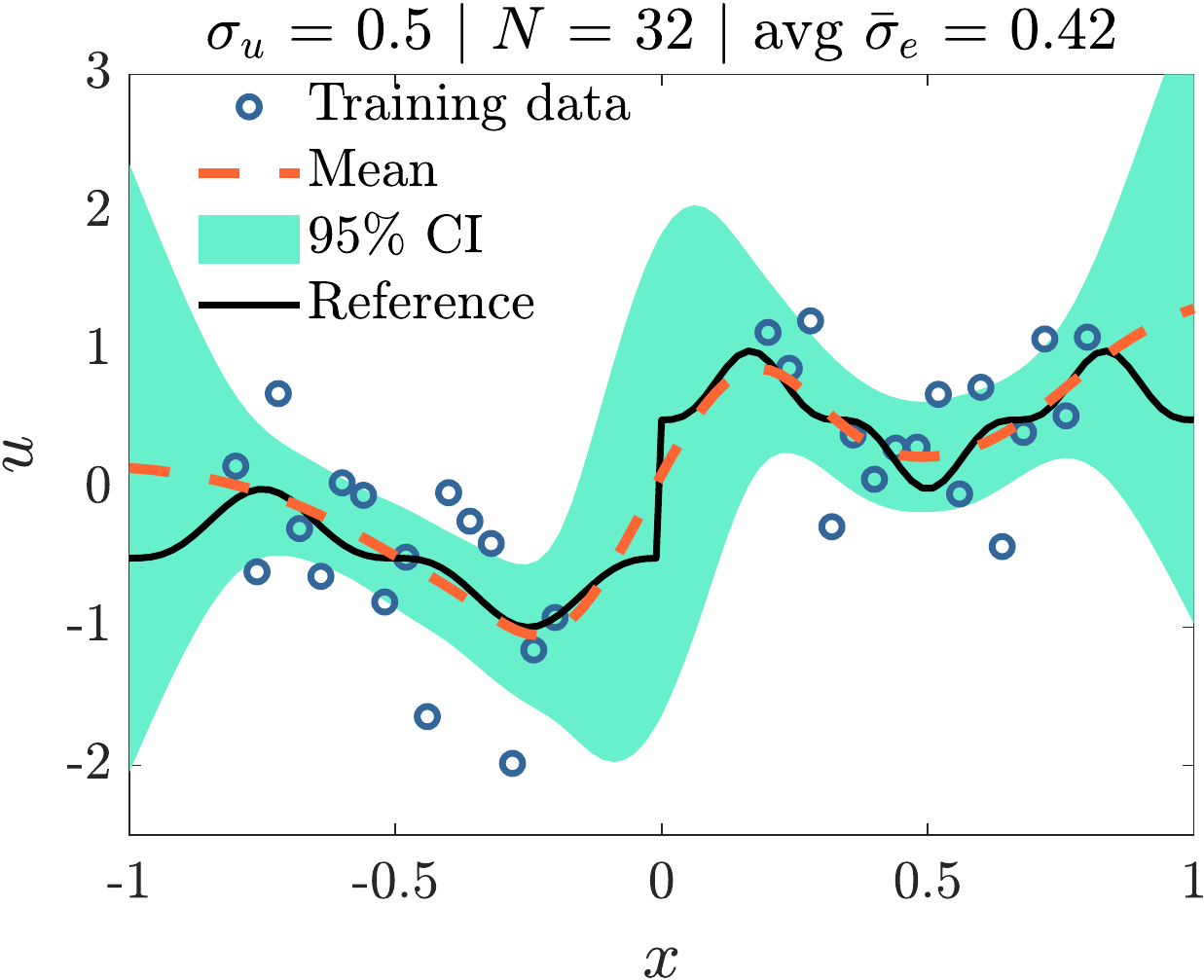}}
	\subcaptionbox{}{}{\includegraphics[width=0.32\textwidth]{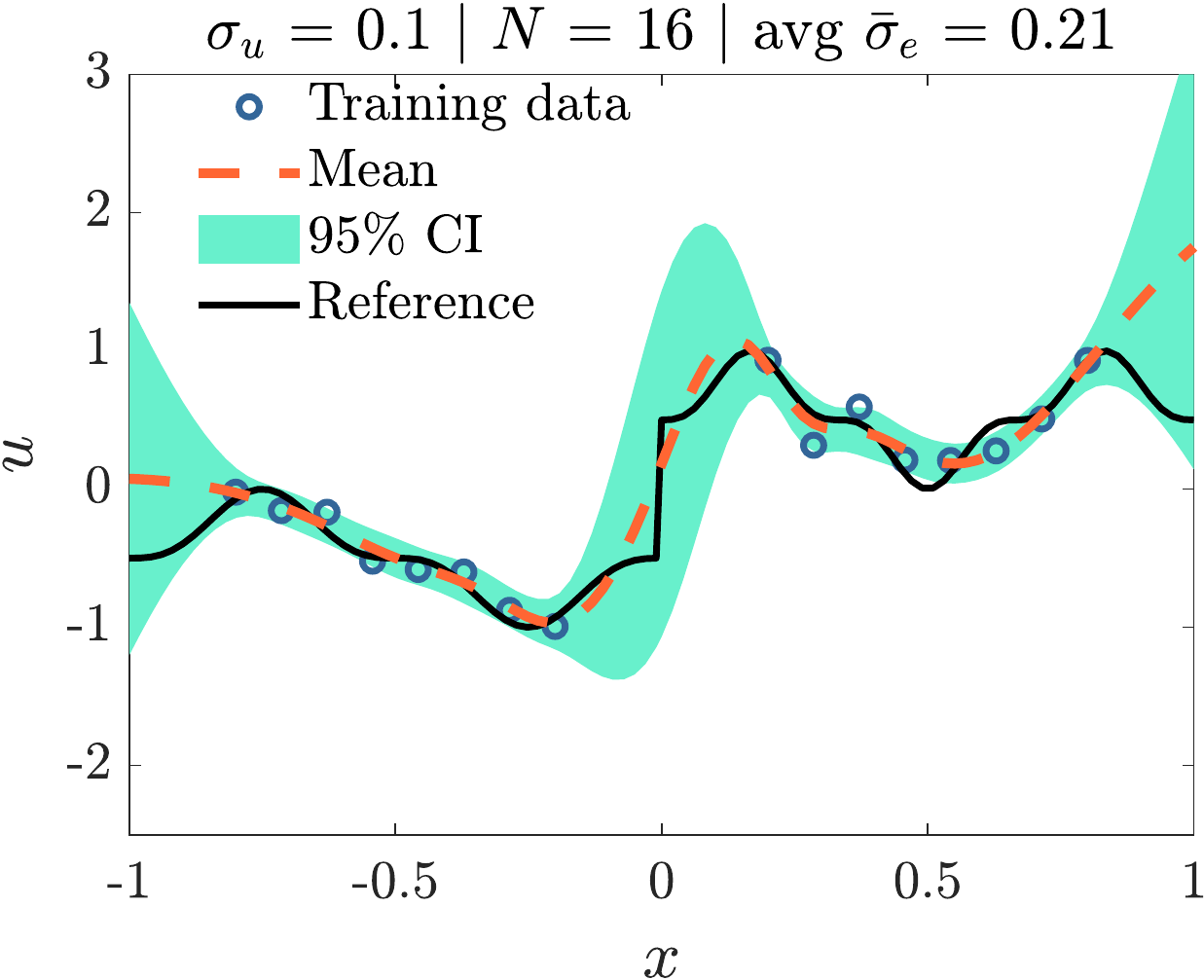}}
	\subcaptionbox{}{}{\includegraphics[width=0.32\textwidth]{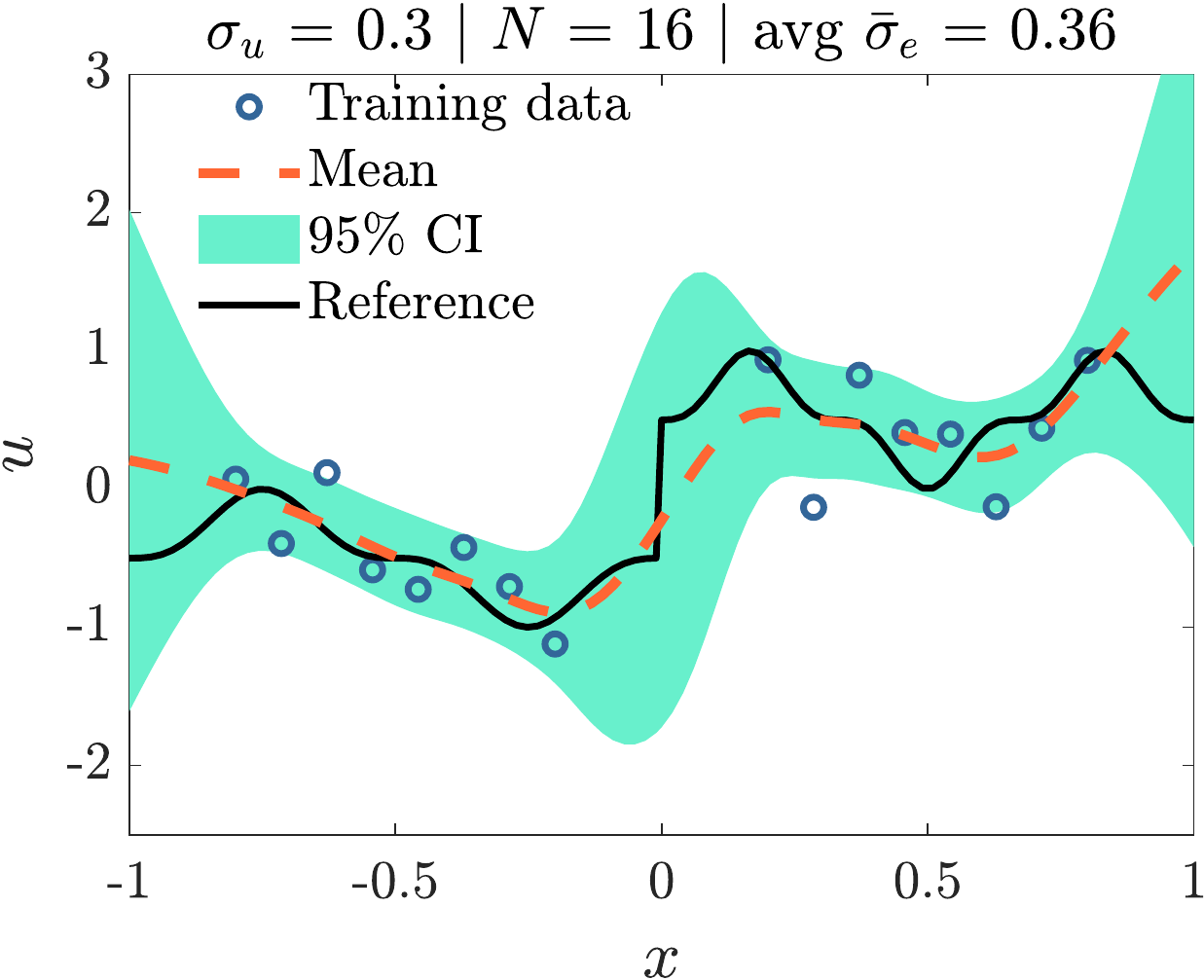}}
	\subcaptionbox{}{}{\includegraphics[width=0.32\textwidth]{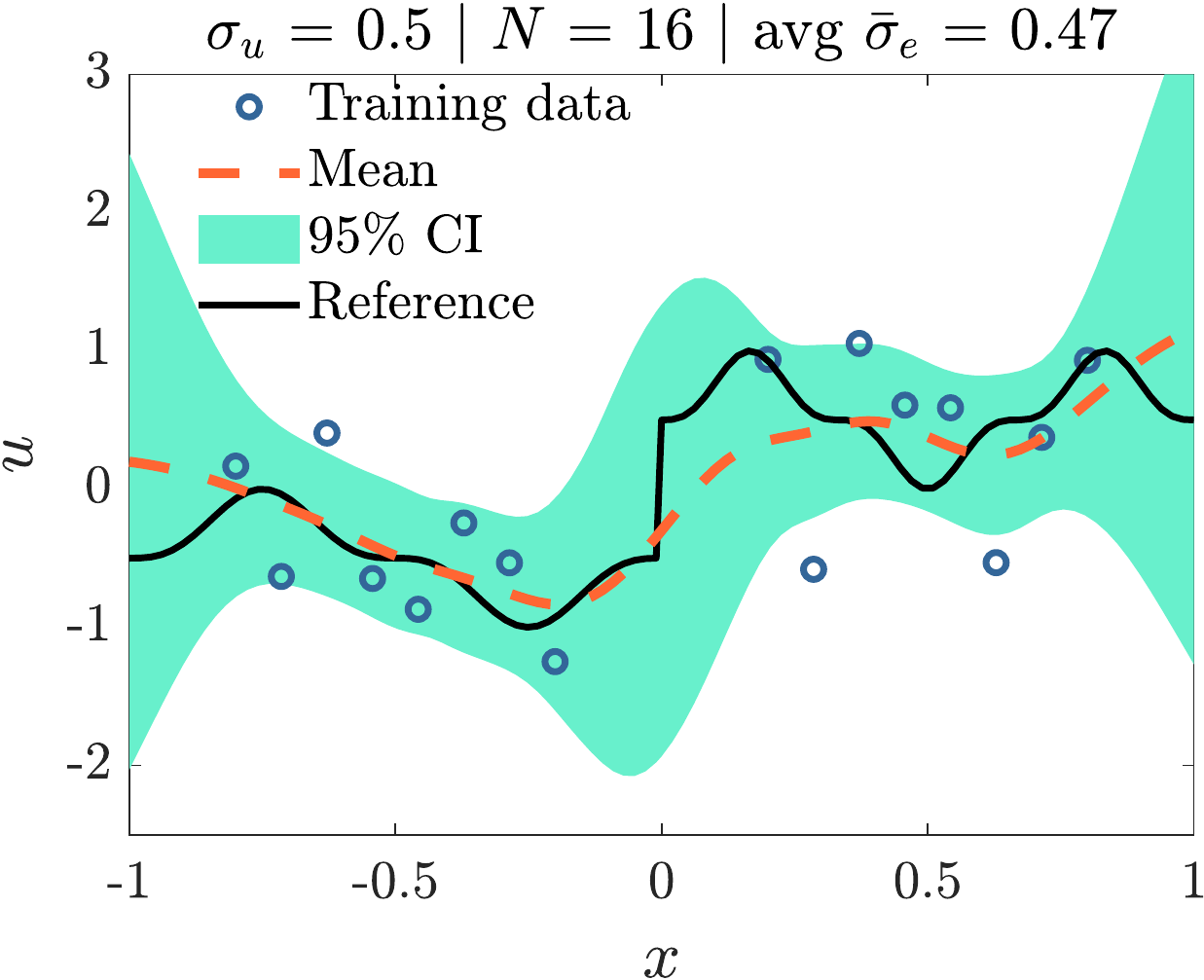}}
	\caption{
		Function approximation problem of Eq.~\eqref{eq:comp:func:func} | \textit{Known homoscedastic noise}:
		epistemic uncertainty of HMC (avg $\bar{\sigma}_e^2$ along $x$) increases with increasing noise magnitude and with decreasing dataset size.
		Shown here are the training data and exact function, as well as the mean and epistemic uncertainty ($95\%$ CI) predictions, as obtained by HMC in conjunction with three noise scales ($\sigma_u=$ 0.1: left, 0.3: middle, 0.5: right) and two dataset sizes ($N=$ 32: top, 16: bottom).
	}
	\label{fig:comp:func:homosc:kno:epist:hmc}
\end{figure}

Next, we compare HMC, which is in general computationally expensive, with MCD, which is as costly as standard NN training, for three difference noise scale values (0.1, 0.3, and 0.5) and two dataset sizes (32 and 16 points).
In general, epistemic uncertainty is expected to increase with increasing aleatoric uncertainty and decreasing dataset size (see also Section~\ref{app:methods:gps} for related discussion in GP regression).
As shown in Figs.~\ref{fig:comp:func:homosc:kno:epist:hmc} and \ref{fig:comp:func:homosc:kno:epist:mcd}, unlike HMC, epistemic uncertainty of MCD does not increase with increasing noise magnitude and with decreasing dataset size.

\begin{table}[!ht]
	\centering
	\footnotesize
	\begin{tabular}{c|c|c|c|c|c|c}
		\toprule
		\multirow{2}{*}{} & \multicolumn{6}{c}{$(N, \sigma_u)$} \\ \cline{2-7}
		& $(32, 0.1)$ & $(32, 0.3)$ & $(32, 0.5)$ & $(16, 0.1)$ & $(16, 0.3)$ & $(16, 0.5)$ \\
		\midrule
		Metric ($\times 10^2$) & \multicolumn{6}{c}{HMC}\\
		\midrule
		RL2E ($\downarrow$)& 22.9 & \textbf{45.2} & \textbf{63.6} & \textbf{23.9} & \textbf{53.1} & \textbf{72.8} \\ 
		MPL ($\uparrow$)& 225.8 & 91.8 & 55.8 & 216.8 & 83.9 & 52.2 \\ 
		RMSCE ($\downarrow$)& \textbf{7.2} & 4.2 & 4.3 & \textbf{3.9} & 3.6 & 4.2 \\ 
		\midrule
		Metric ($\times 10^2$) & \multicolumn{6}{c}{MCD}\\
		\midrule
		RL2E ($\downarrow$)& \textbf{21.9} & 46.1 & 65.2 & 24.5 & 53.7 & 73.4 \\ 
		MPL ($\uparrow$)& \textbf{237.8} & \textbf{94.9} & \textbf{56.9} & \textbf{222.0} & \textbf{89.9} & \textbf{56.7} \\ 
		RMSCE ($\downarrow$) & 8.6 & \textbf{3.0} & \textbf{3.5} & 4.6 & \textbf{2.9} & \textbf{2.0} \\
		\bottomrule
	\end{tabular}
	\caption{
		Function approximation problem of Eq.~\eqref{eq:comp:func:func} | \textit{Known homoscedastic noise}:
		MCD performs better than HMC for some magnitudes of noise and dataset sizes based on the evaluation metrics. However, unlike HMC, MCD does not provide consistently larger epistemic uncertainties for increasing noise or decreasing dataset size, as shown in Figs.~\ref{fig:comp:func:homosc:kno:epist:hmc} and \ref{fig:comp:func:homosc:kno:epist:mcd}.
		See also \cite{yao2019quality,ashukha2021pitfalls} regarding the limitations of the evaluation metrics.
		Here we evaluate ID performance of HMC and MCD in conjunction with three noise scales (0.1, 0.3, 0.5) and two dataset sizes (32, 16 points), using noisy test data and uncalibrated predictions.
		Comparisons are made in the vertical direction.
	}
	\label{tab:comp:func:homosc:kno:epist}
\end{table}

\noindent Nevertheless, this is not depicted in the corresponding evaluation Table~\ref{tab:comp:func:homosc:kno:epist}, where MCD performs better than HMC for some magnitudes of noise and dataset sizes based on the used metrics.
In fact, MCD performs better than HMC in terms of predictive likelihood (MPL) for all noise and dataset cases.
This can be explained by the fact that HMC, having large epistemic uncertainty, is quite under-confident in terms of total uncertainty predictions.
However, this lack of confidence of HMC is desirable in this case: more noise and less data should make the predictions less confident.  
This fact shows that a plurality of evaluation metrics is necessary for comparing different methods.
The interested reader is directed to \cite{yao2019quality,ashukha2021pitfalls} for discussions related to the limitations of evaluation metrics.

\begin{figure}[!ht]
	\centering
	\subcaptionbox{}{}{\includegraphics[width=0.32\textwidth]{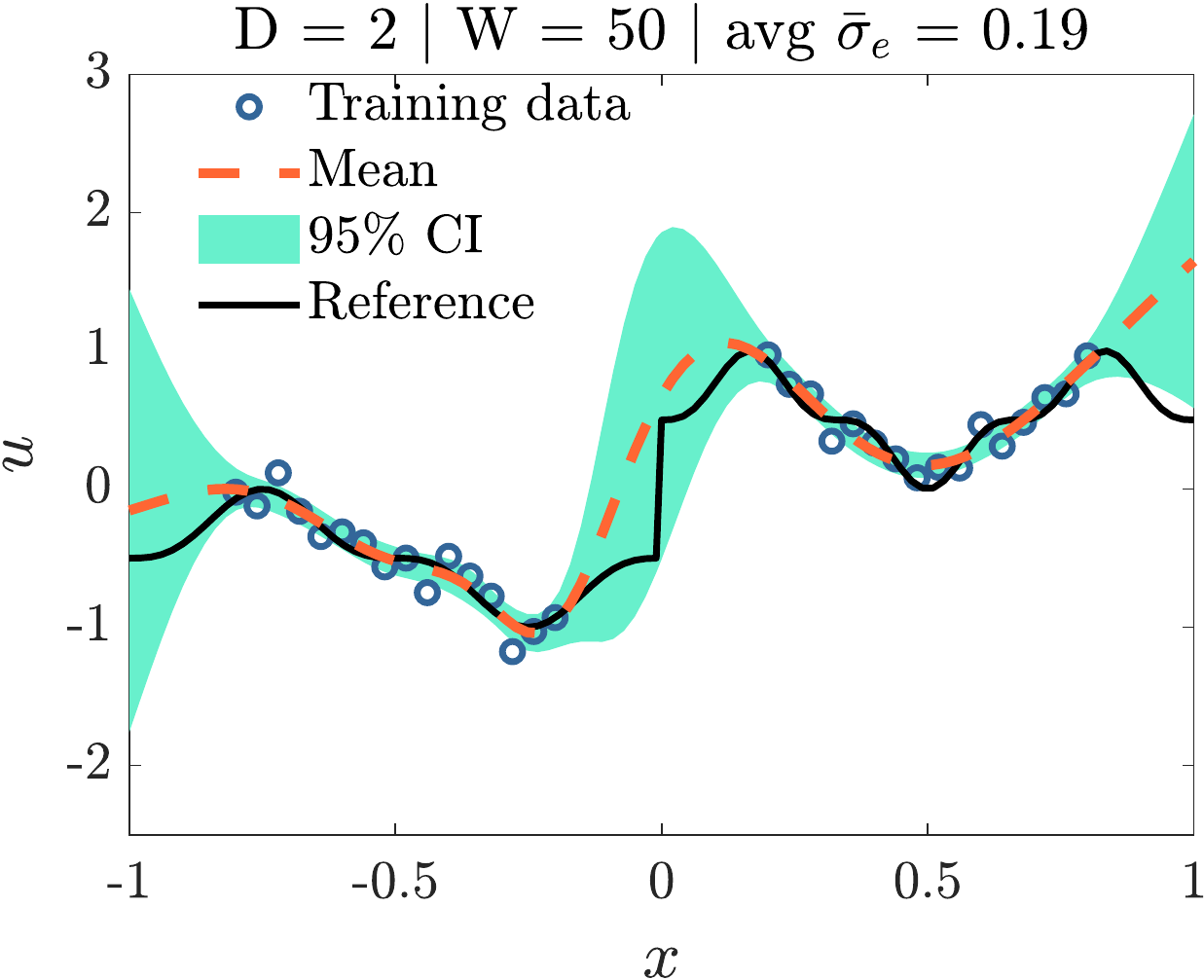}}
	\subcaptionbox{}{}{\includegraphics[width=0.32\textwidth]{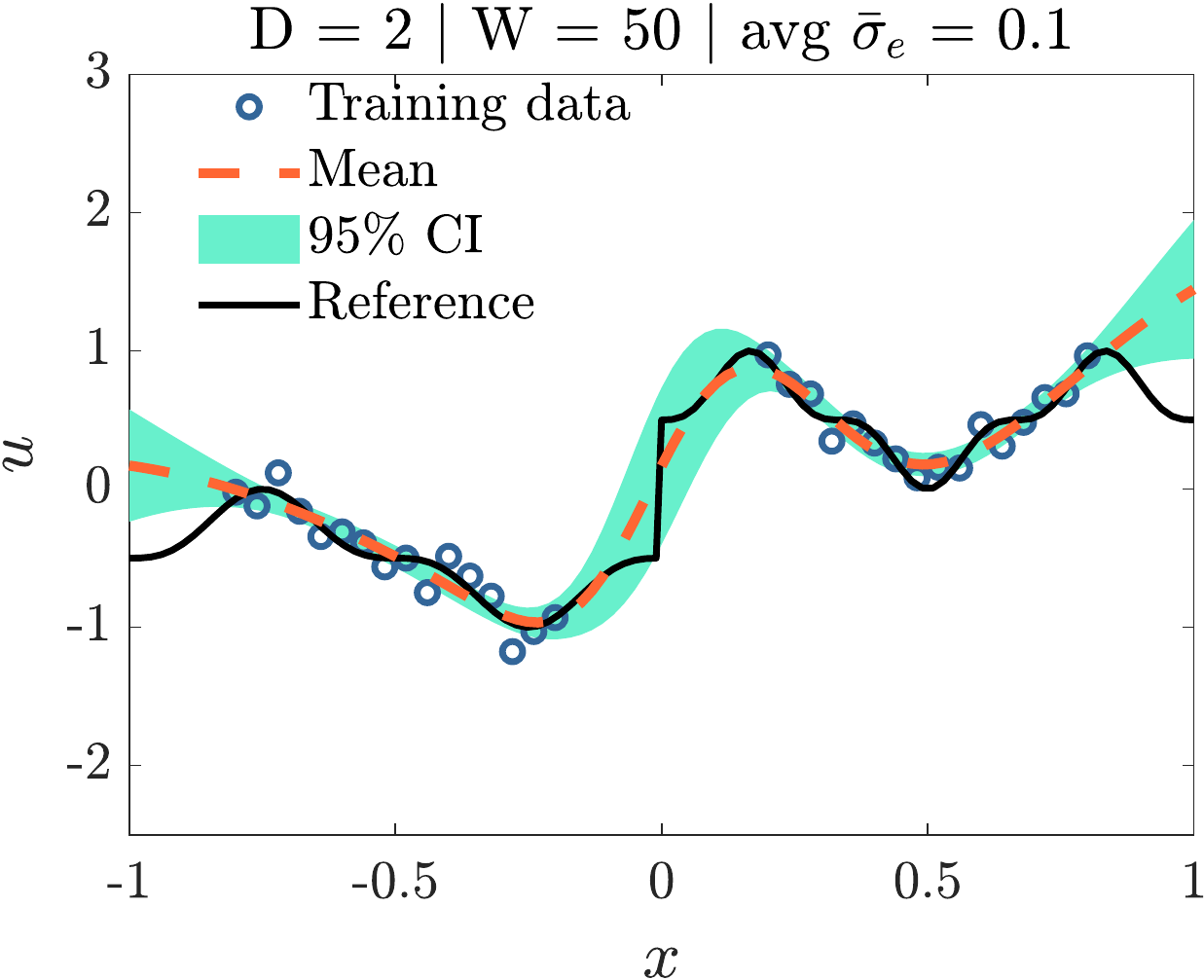}}
	\subcaptionbox{}{}{\includegraphics[width=0.32\textwidth]{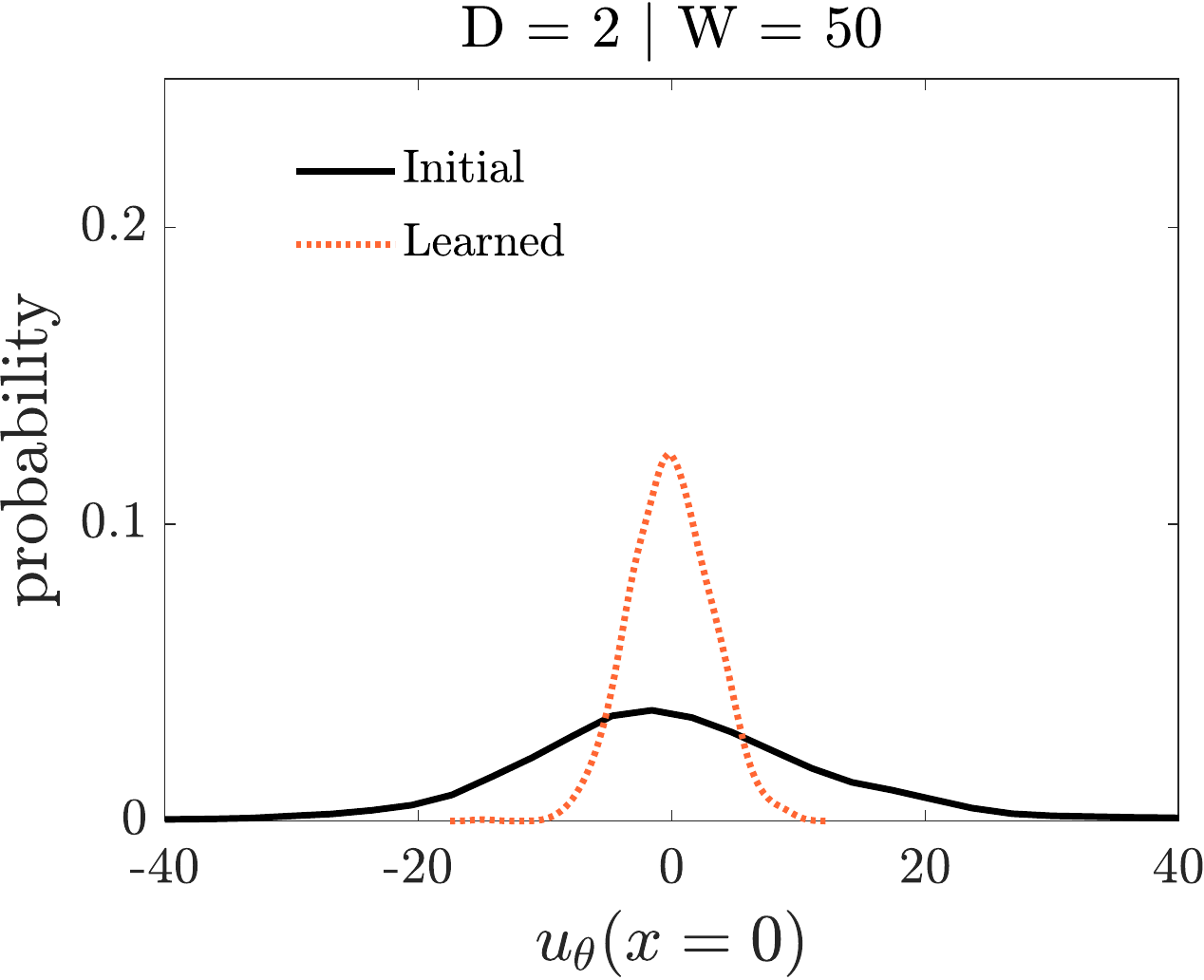}}
	\subcaptionbox{}{}{\includegraphics[width=0.32\textwidth]{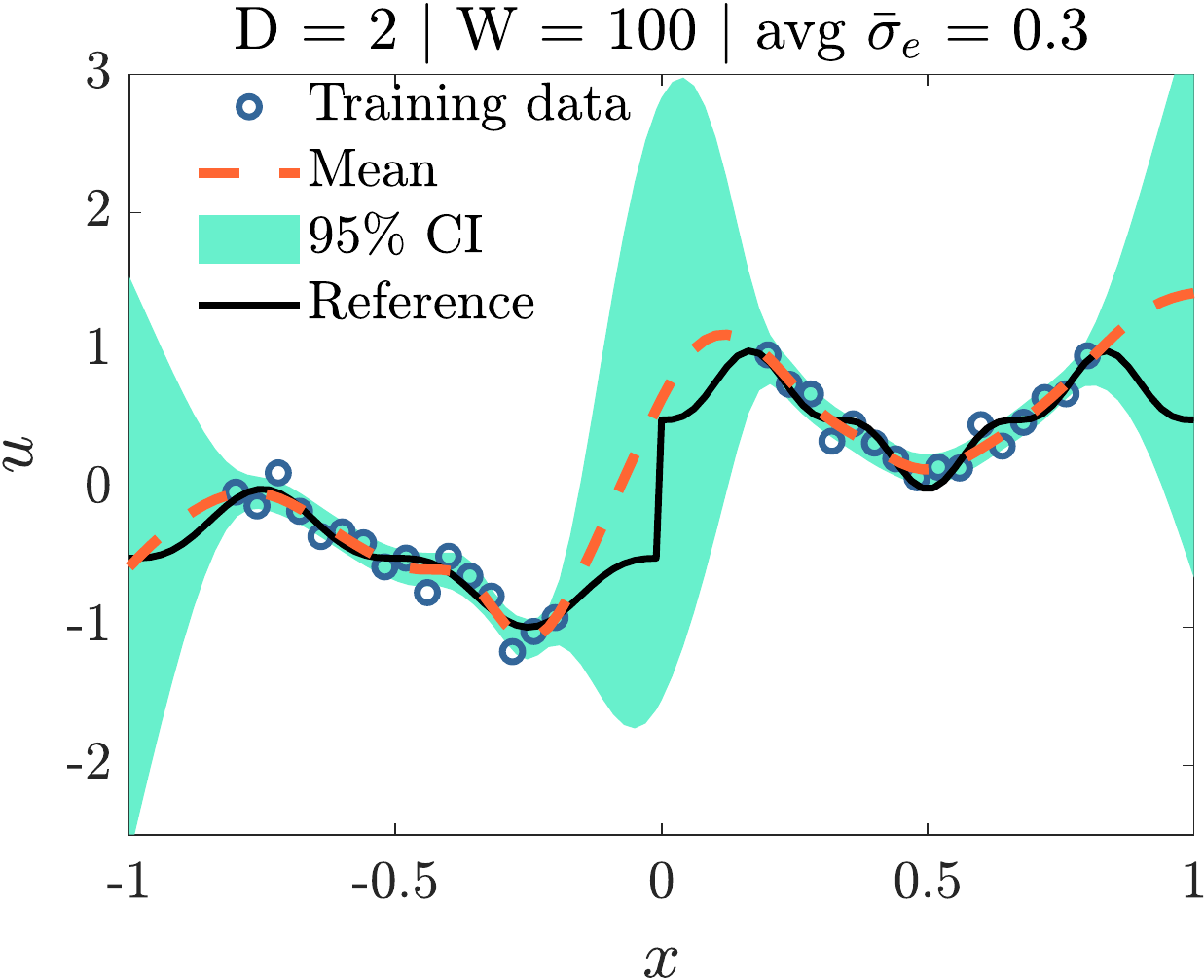}}
	\subcaptionbox{}{}{\includegraphics[width=0.32\textwidth]{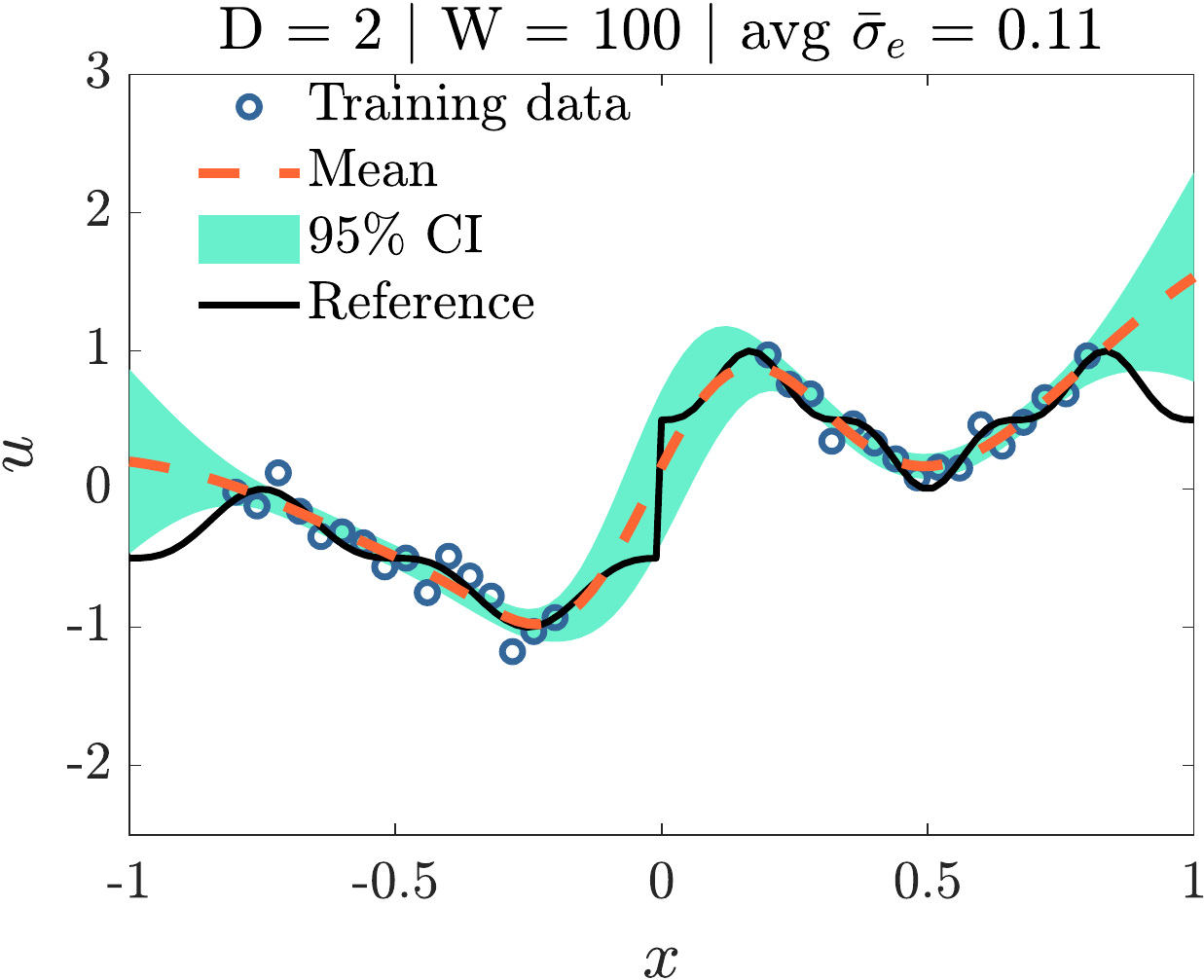}}
	\subcaptionbox{}{}{\includegraphics[width=0.32\textwidth]{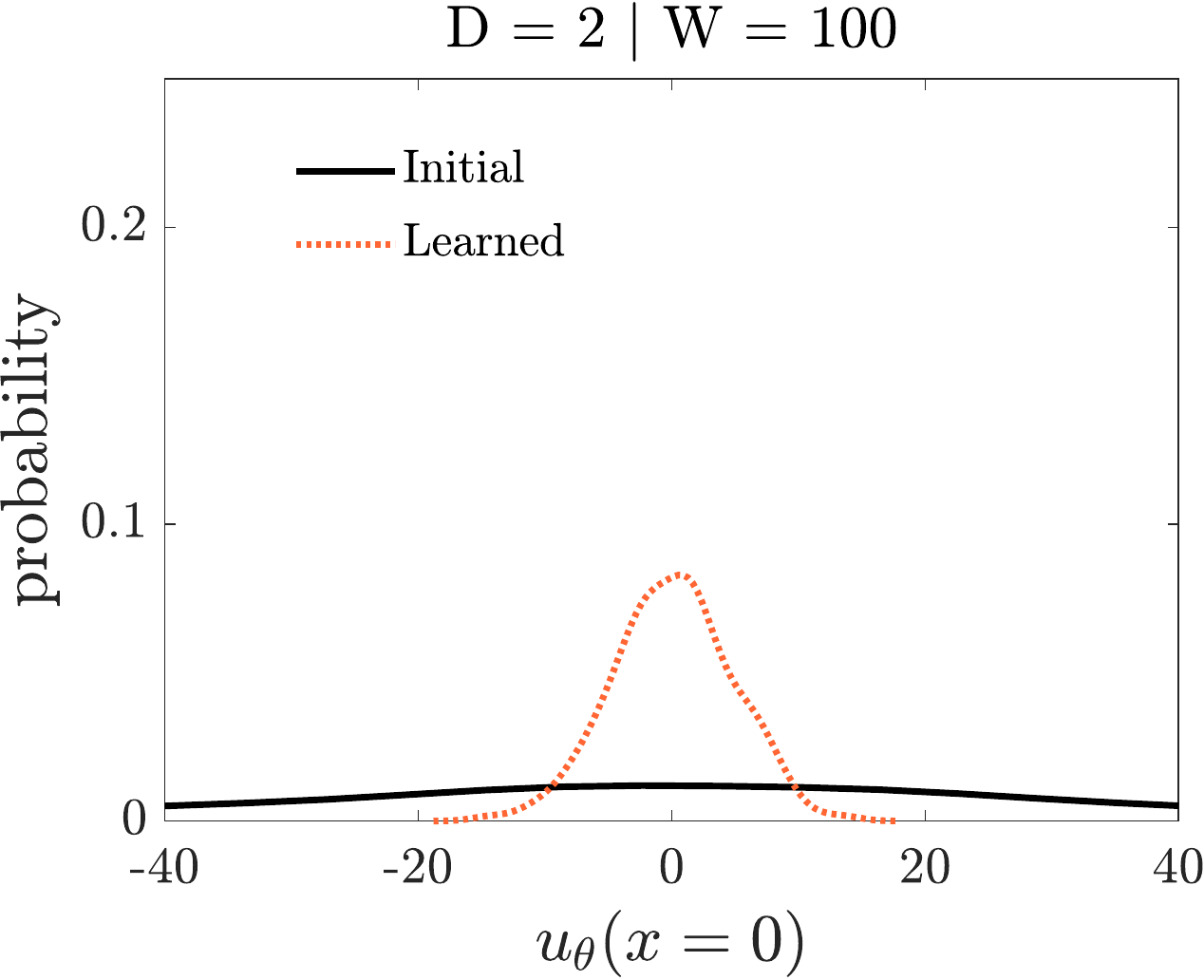}}
	\subcaptionbox{}{}{\includegraphics[width=0.32\textwidth]{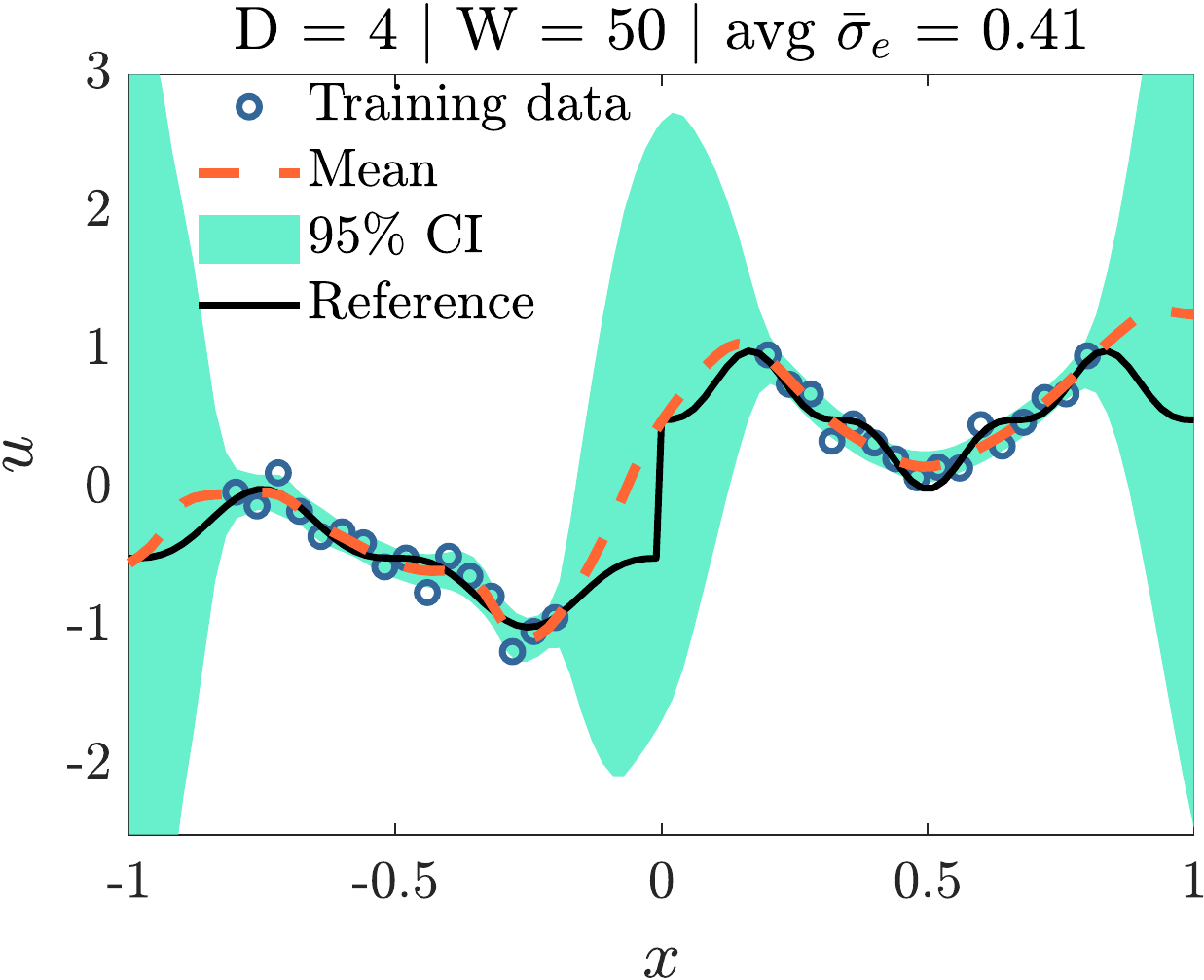}}
	\subcaptionbox{}{}{\includegraphics[width=0.32\textwidth]{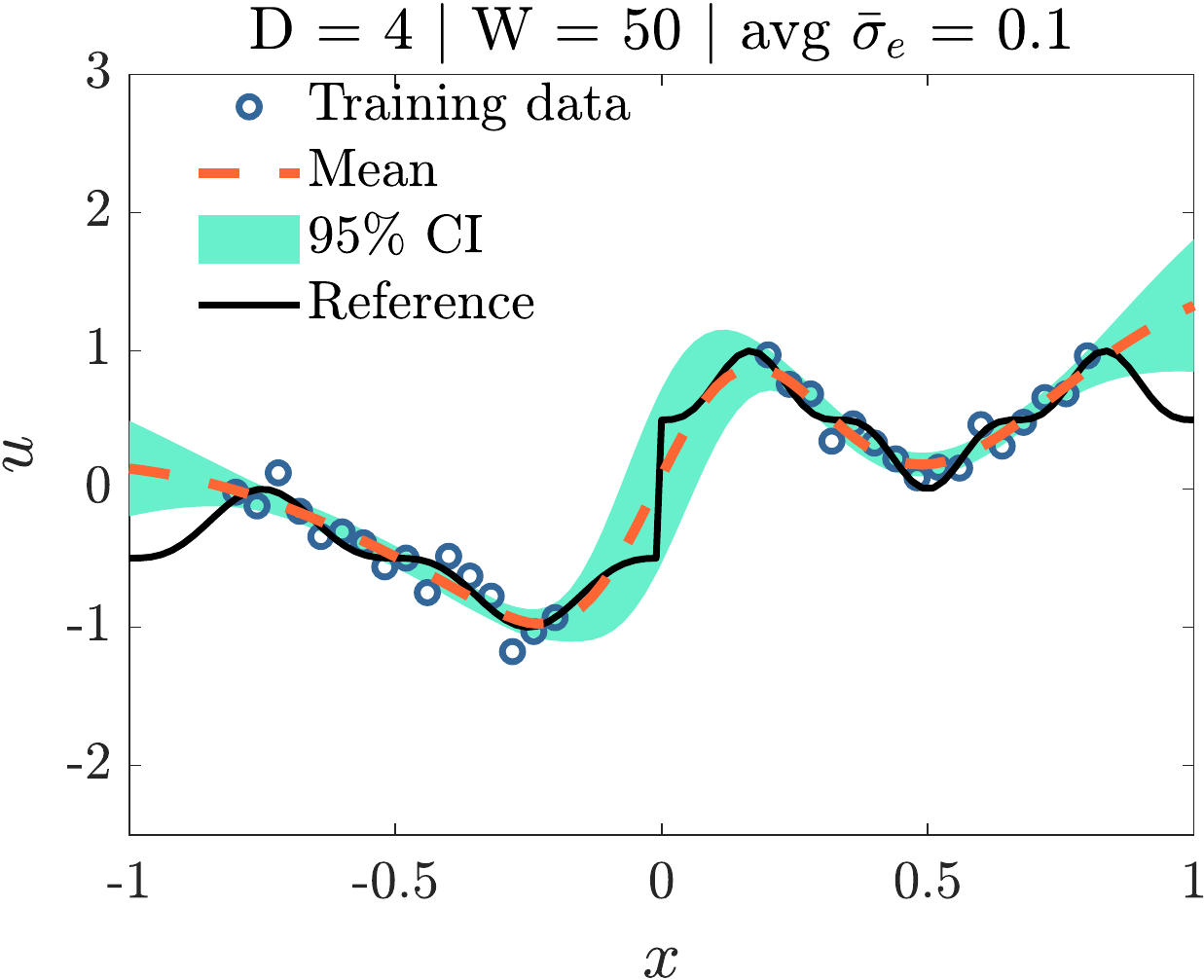}}
	\subcaptionbox{}{}{\includegraphics[width=0.32\textwidth]{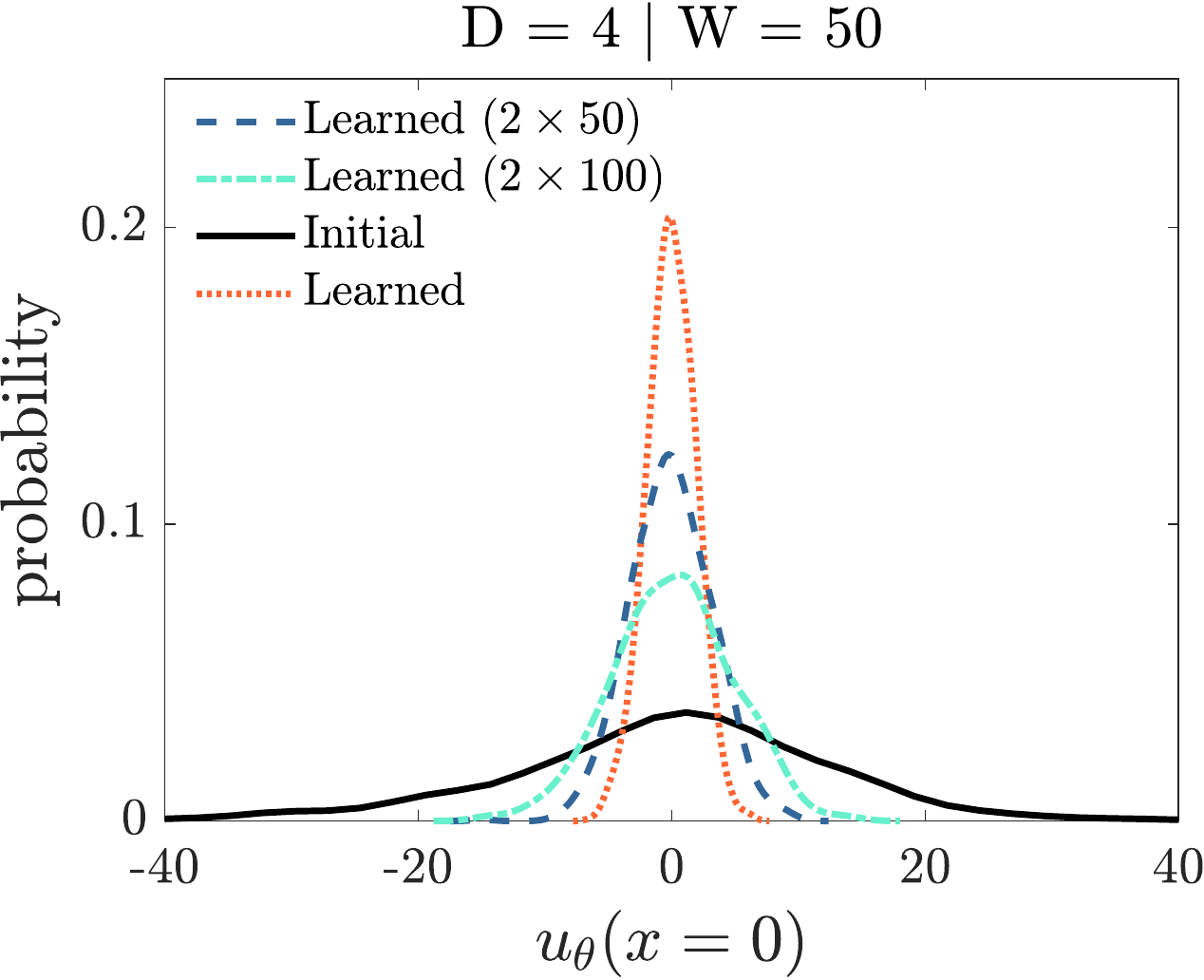}}
	\caption{
		Function approximation problem of Eq.~\eqref{eq:comp:func:func} | \textit{Known homoscedastic noise}:
		epistemic uncertainty (avg $\sigma_e^2$ along $x$) increases with increasing NN expressivity.
		By learning the prior this effect can be partially reverted.
		This reduces the burden of thoroughly selecting the NN architecure, as long as the selected one is expressive enough for the considered problem.
		Shown here are the training data and exact function, as well as the mean and epistemic uncertainty ($95\%$ CI) predictions, as obtained by HMC in conjunction with three NN architectures defined by their depth (D) and width (W).
		Cases considered are (a)-(c) D, W = 2, 50; (d)-(f) D, W = 2, 100; (g)-(i) D, W = 4, 50. 
		\textbf{Left:} HMC with fixed prior $\cN(0, 1)$.
		\textbf{Middle:} HMC with prior learning.
		\textbf{Right:} effect of prior of learning on the distribution of $u_{\theta}(0)$ with $\theta$ following the initial and learned prior distribution.}
	\label{fig:comp:func:homosc:kno:netsize}
\end{figure}

In Fig.~\ref{fig:comp:func:homosc:kno:netsize}, we compare the HMC results for different NN architectures; namely, with widths (number of neurons in each layer) in the set $\{50, 100\}$ and with depths (number of layers) in the set $\{2, 4\}$.
It is shown that epistemic uncertainty increases with increasing NN expressivity, i.e., larger number of trainable parameters. 
Nevertheless, this effect can be partially reverted by learning the prior. 
Specifically, as we demonstrate in the right column of Fig.~\ref{fig:comp:func:homosc:kno:netsize}, the distributions of $u_{\theta}(x=0)$ with $\theta$ drawn from the priors corresponding to the different architectures, concentrate to similar distributions after learning the prior.
Having similar learned priors, the different architectures yield similar posteriors, as shown in the middle column of Fig.~\ref{fig:comp:func:homosc:kno:netsize}.

\begin{figure}[!ht]
	\centering
	\subcaptionbox{}{}{\includegraphics[width=0.32\textwidth]{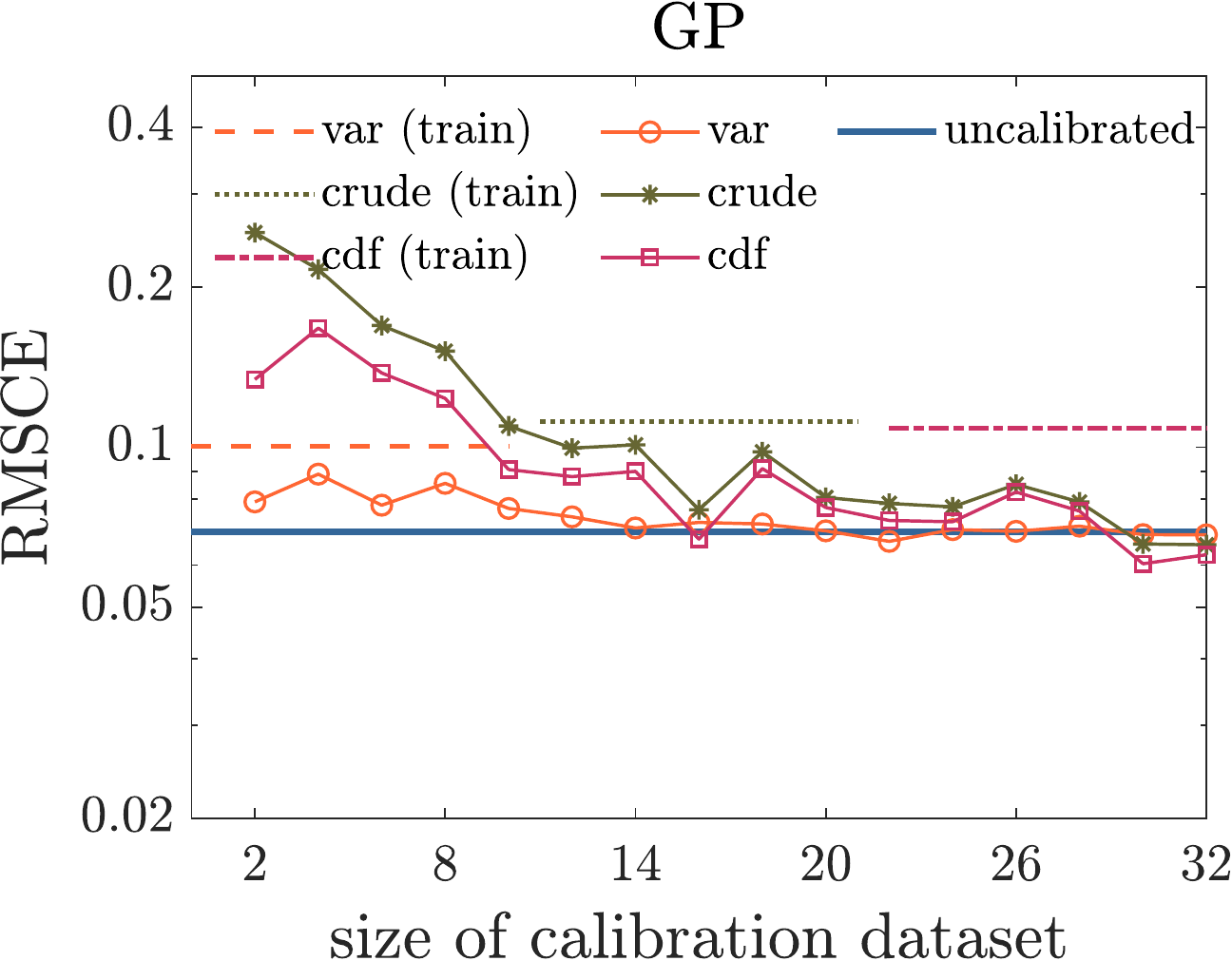}}
	\subcaptionbox{}{}{\includegraphics[width=0.32\textwidth]{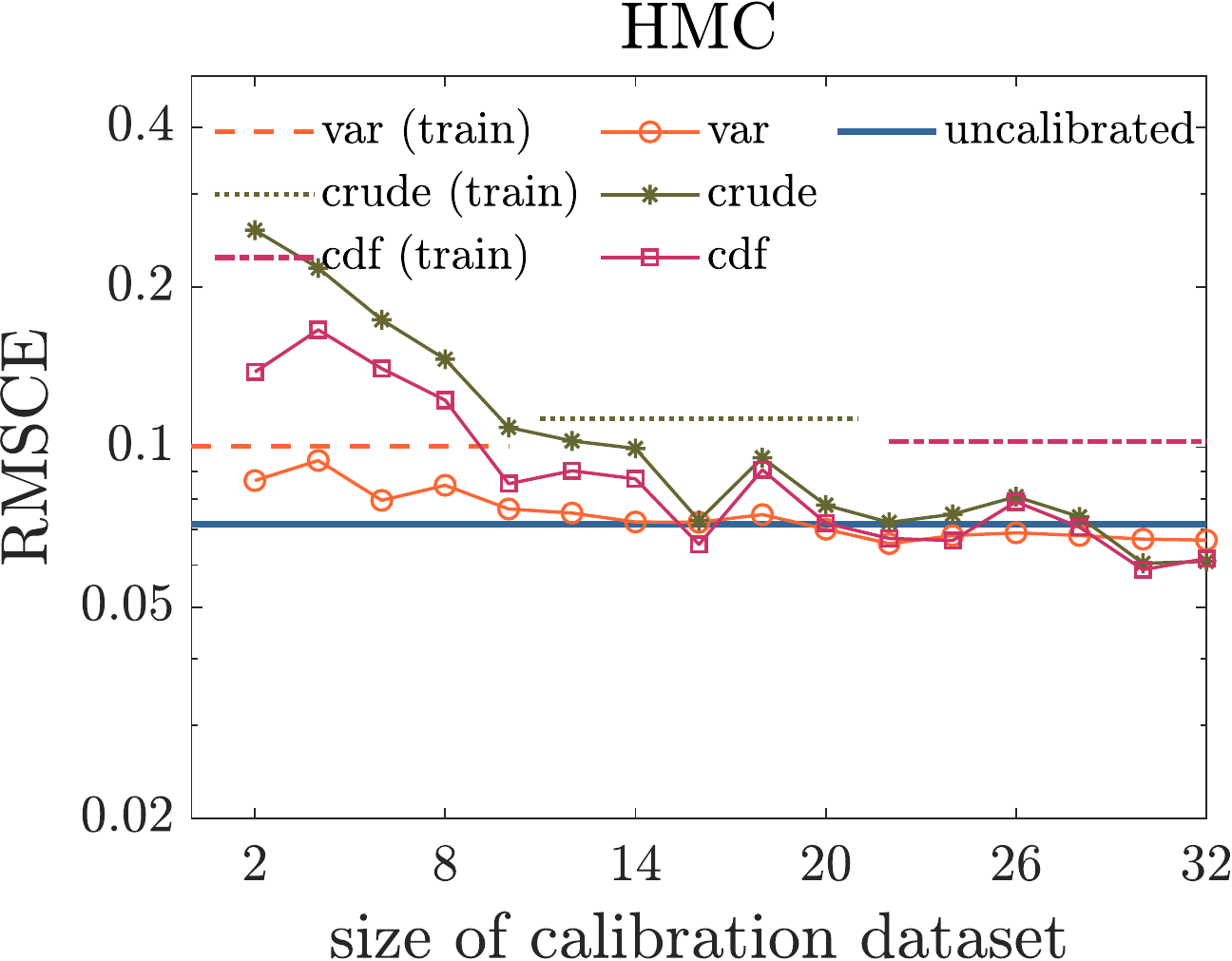}}
	\subcaptionbox{}{}{\includegraphics[width=0.32\textwidth]{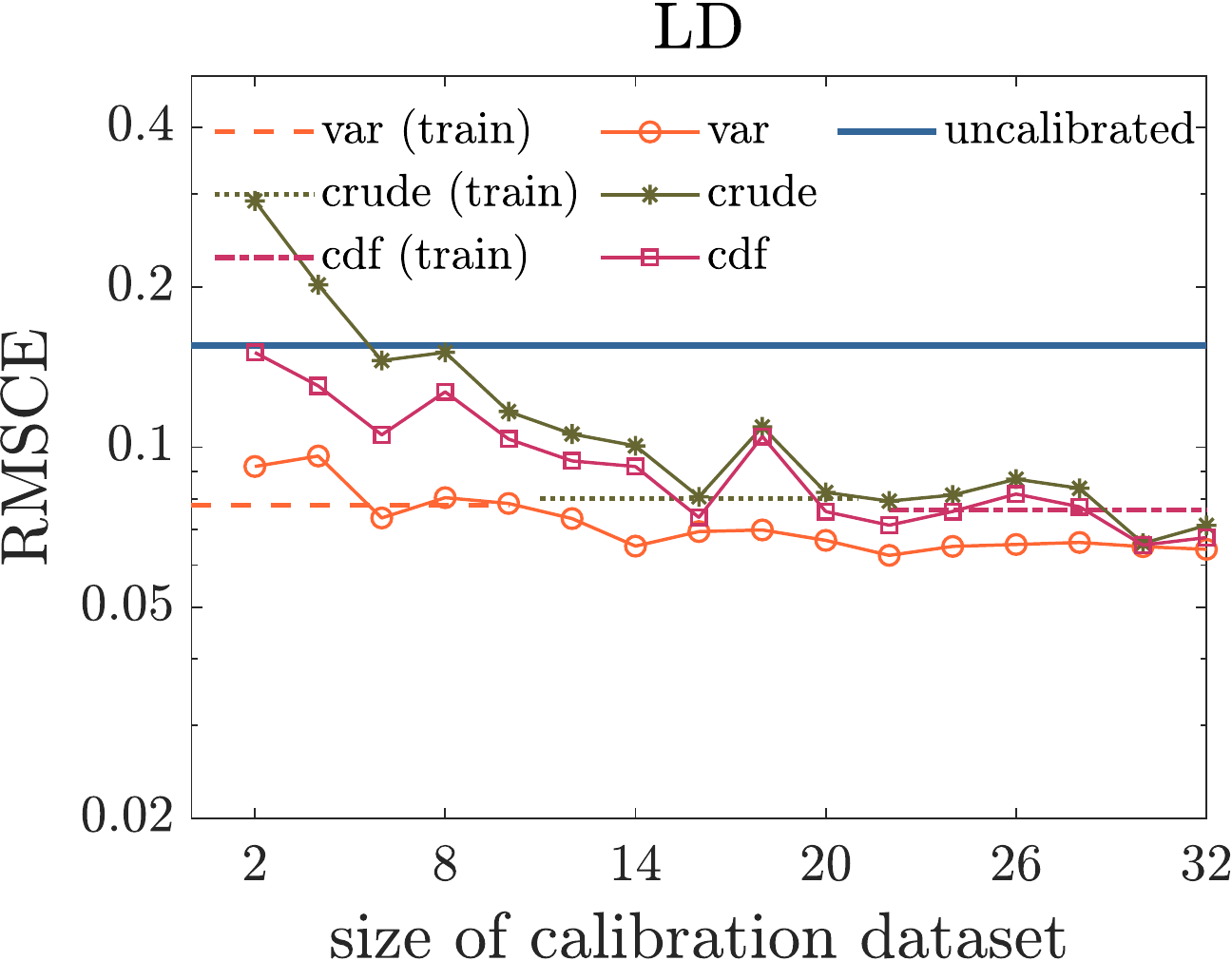}}
	\subcaptionbox{}{}{\includegraphics[width=0.32\textwidth]{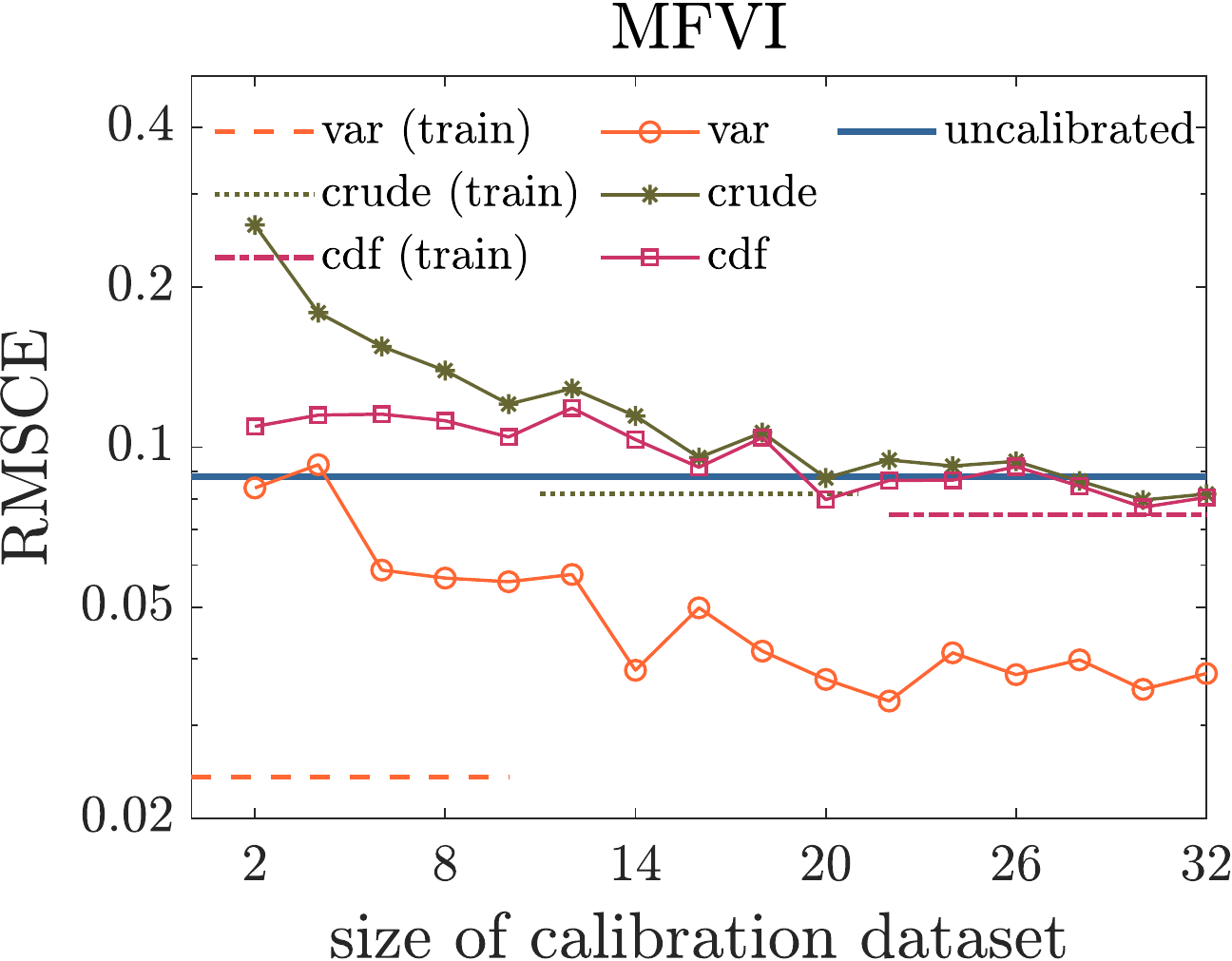}}
	\subcaptionbox{}{}{\includegraphics[width=0.32\textwidth]{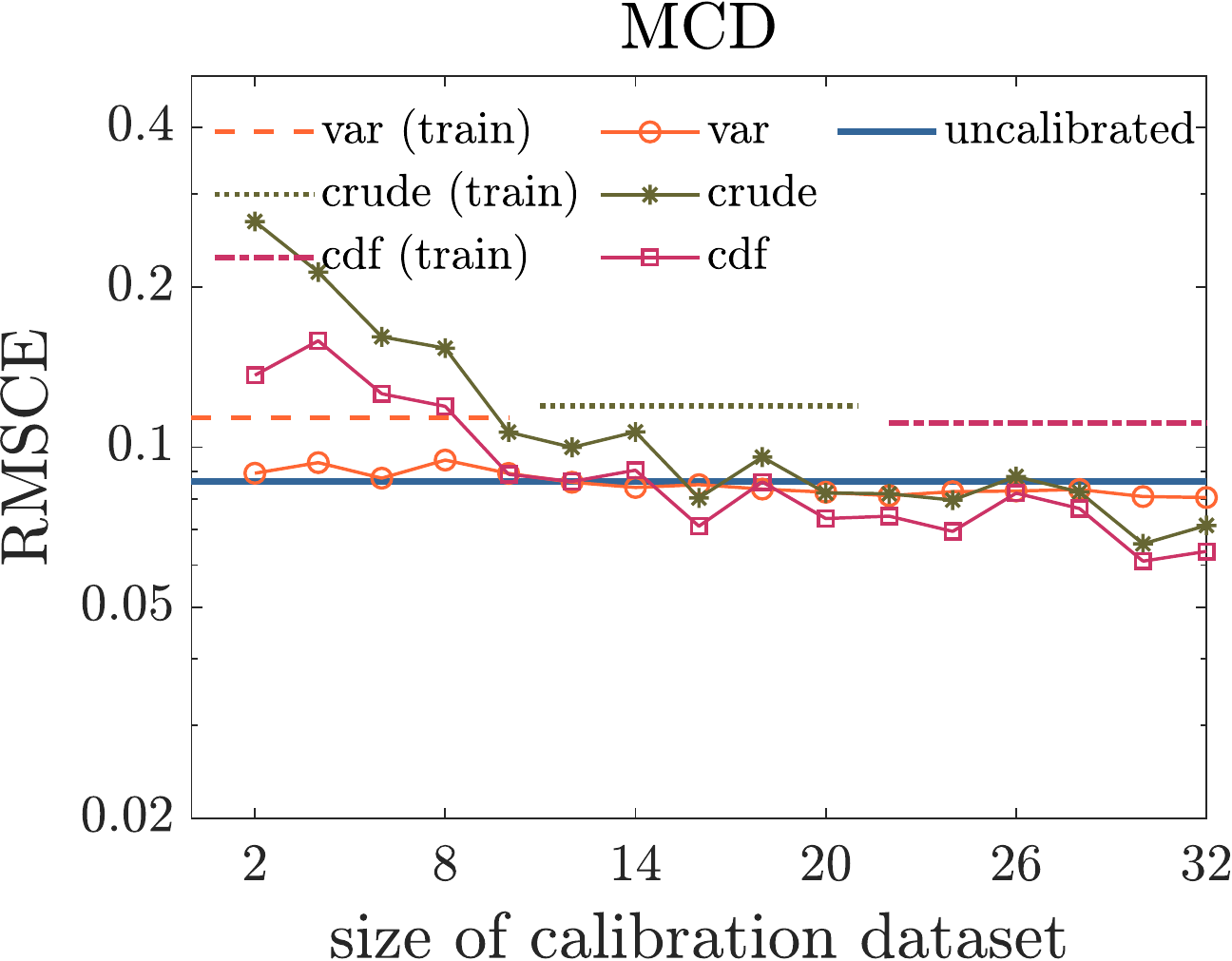}}
	\subcaptionbox{}{}{\includegraphics[width=0.32\textwidth]{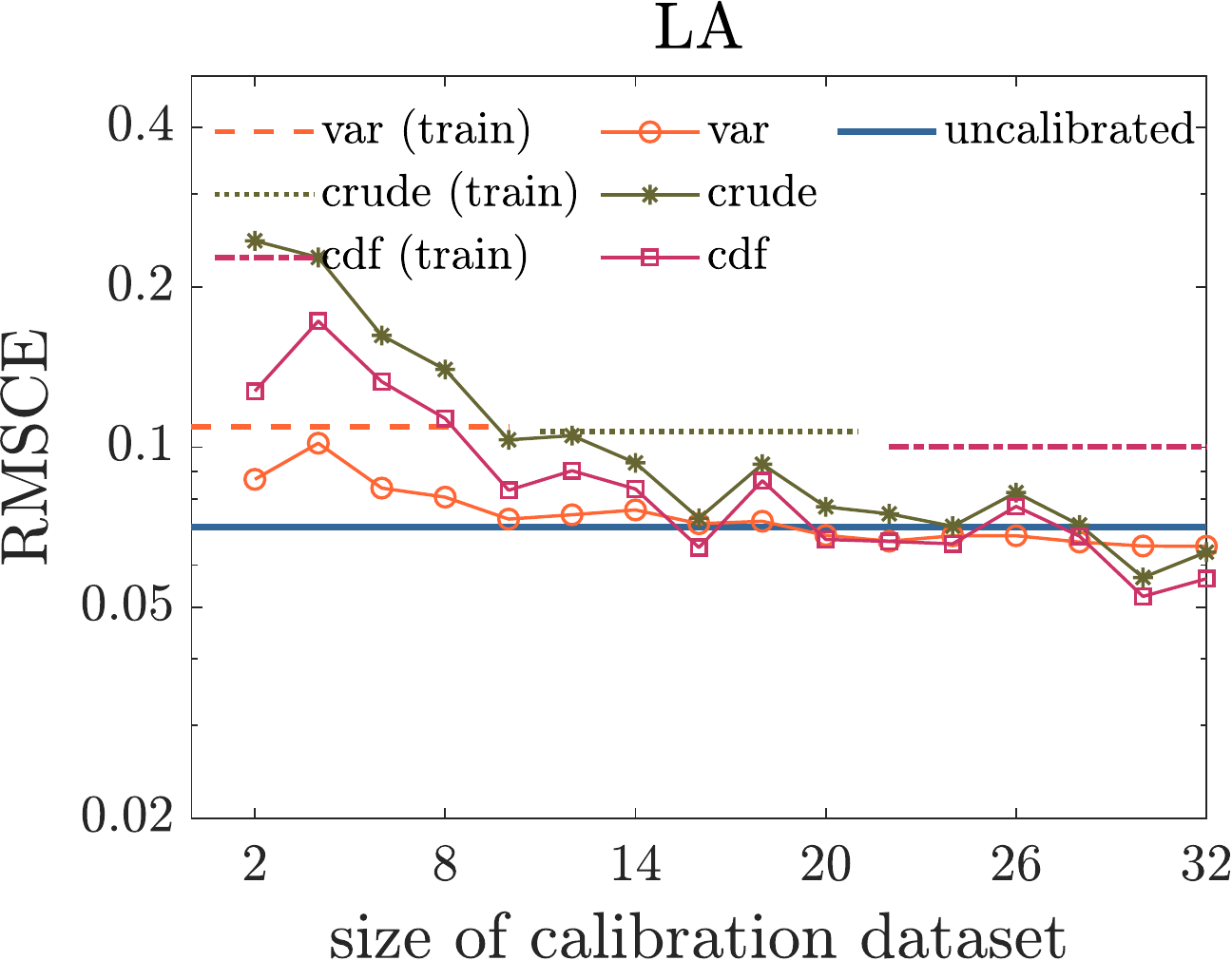}}
	\subcaptionbox{}{}{\includegraphics[width=0.32\textwidth]{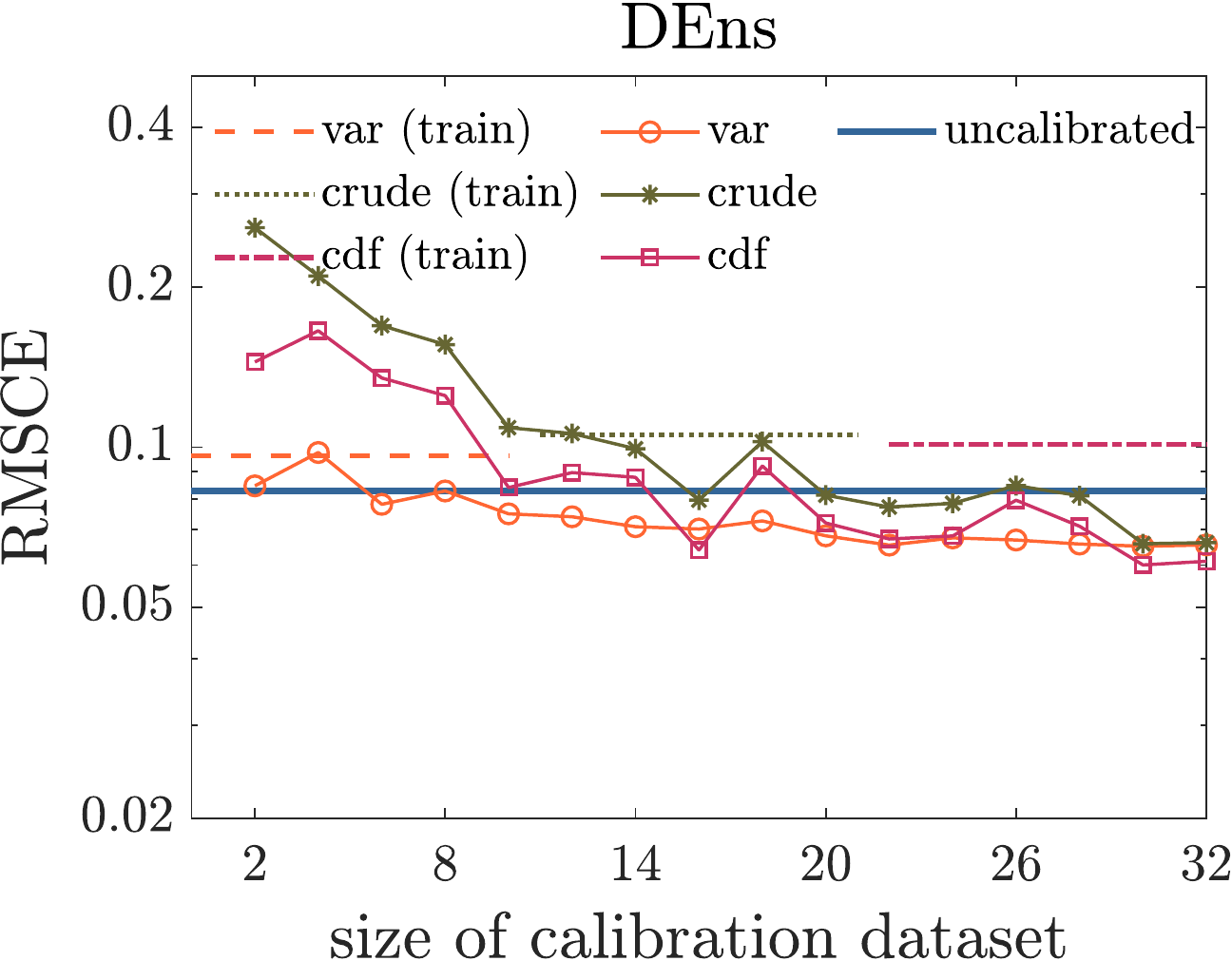}}
	\subcaptionbox{}{}{\includegraphics[width=0.32\textwidth]{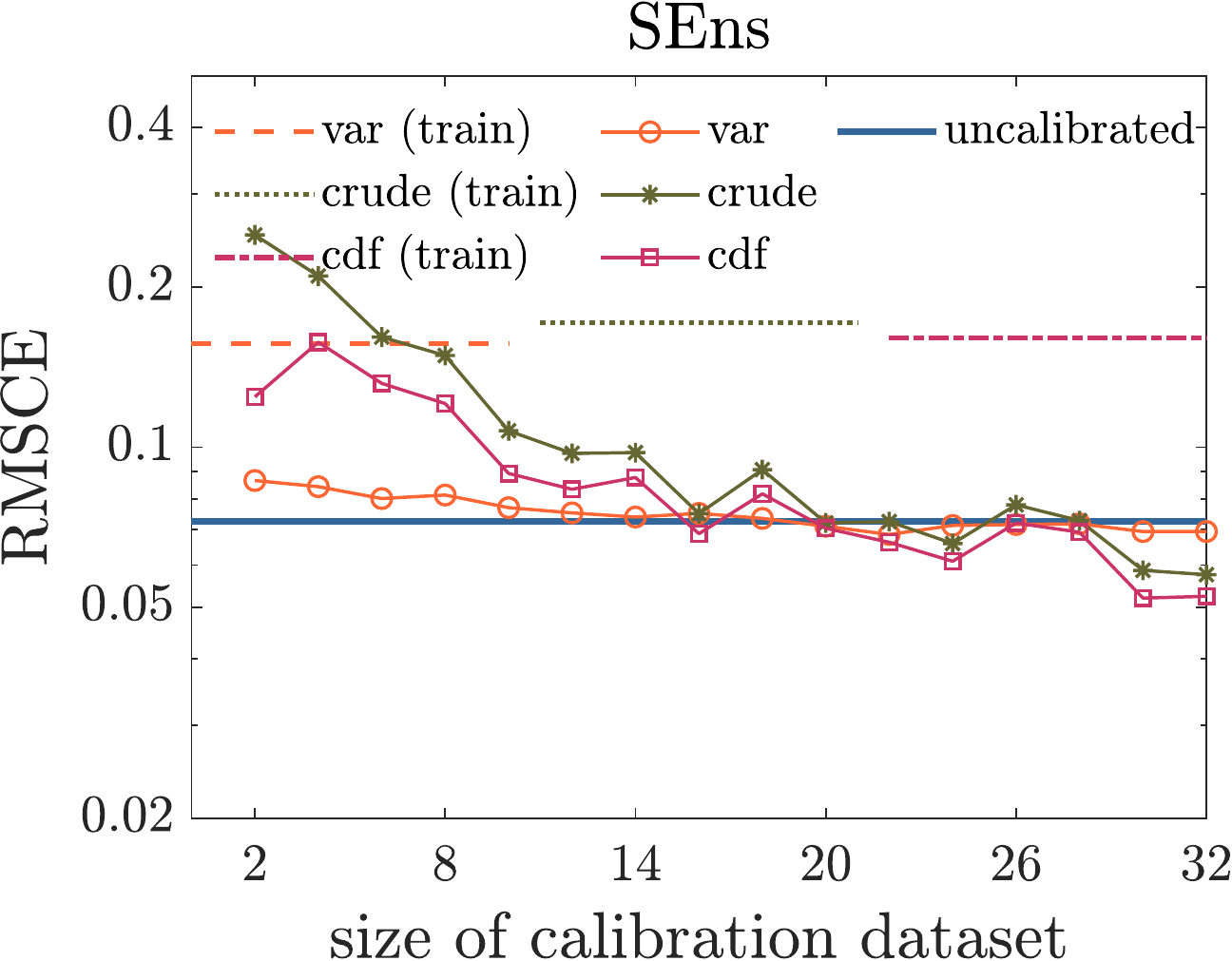}}
	\subcaptionbox{}{}{\includegraphics[width=0.32\textwidth]{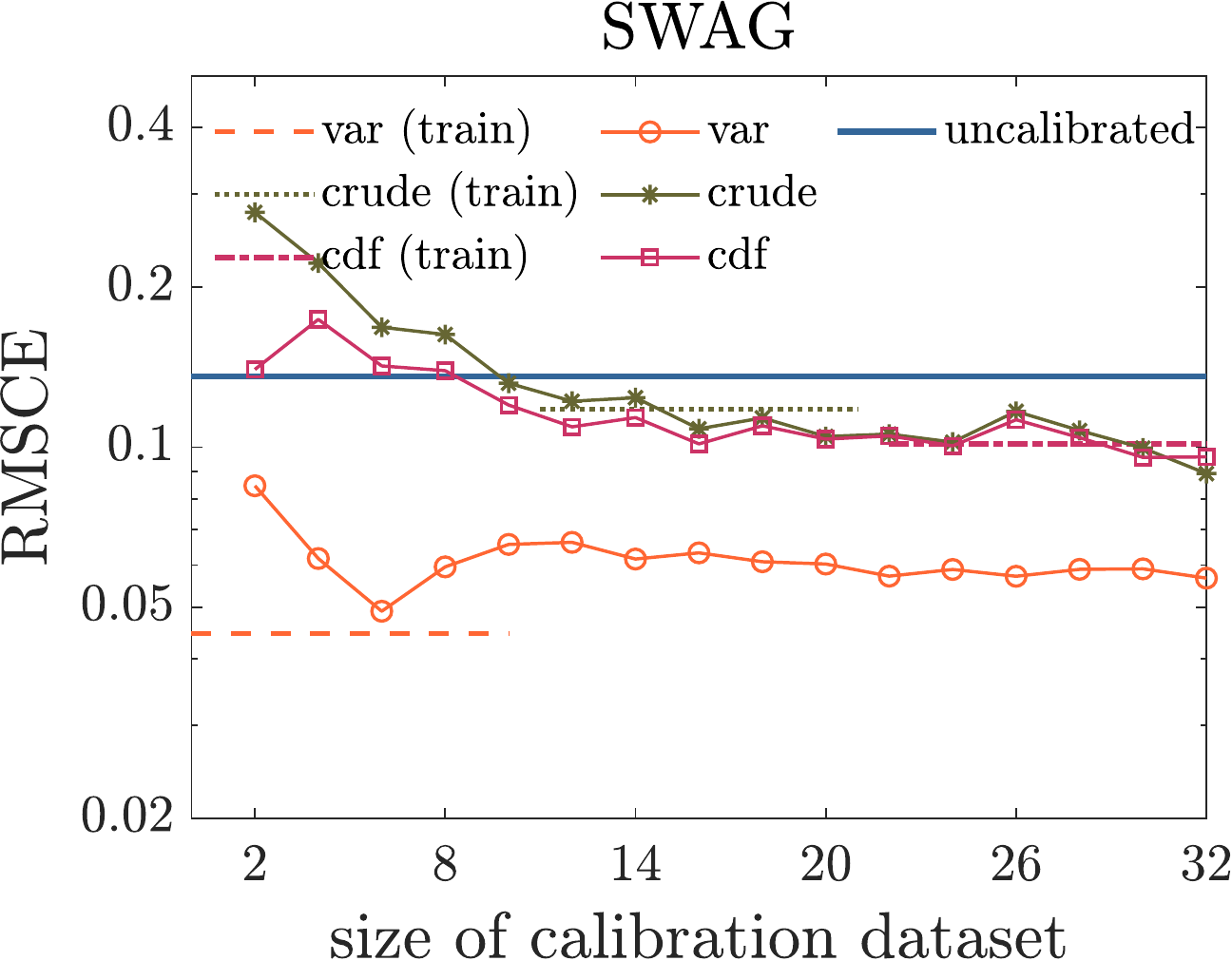}}
	\caption{
		Function approximation problem of Eq.~\eqref{eq:comp:func:func} | \textit{Known homoscedastic noise}:
		post-training calibration, even with only a few calibration datapoints (2-14), can reduce significantly RMSCE; see Figs.~\ref{fig:comp:func:homosc:kno:res:1}-\ref{fig:comp:func:homosc:kno:res:3} for the effect on the predictions.
		Here we demonstrate the effect of different calibration approaches and calibration dataset sizes on RMSCE for all considered UQ methods.
		Variance re-weighting (var), CDF modification using an auxiliary model (cdf) and CRUDE are considered (Section~\ref{sec:eval:calib}), whereas the calibration set is either the training set or a separate set with varying size.
		Calibration using the training dataset typically leads to overfitting; see, e.g., part (a).
		Note that calibration is performed using ID data, hence OOD performance may deteriorate; see, e.g., SWAG in Table~\ref{tab:comp:func:homosc:kno:OOD}. 
	}
	\label{fig:comp:func:homosc:kno:calsize}
\end{figure}

In Fig.~\ref{fig:comp:func:homosc:kno:calsize}, we demonstrate the effect of different
post-training calibration approaches and dataset sizes on calibration error (RMSCE).
It is shown that post-training calibration, even with only a few calibration datapoints (2-14), can reduce significantly RMSCE, especially for initially less calibrated methods, such as SWAG.
It is also shown that calibration using the training dataset typically leads to
overfitting; i.e., RMSCE after calibration is higher, and thus a separate calibration dataset should be used.
Lastly, note that calibration is performed using ID data, hence OOD performance may deteriorate; see, e.g., SWAG in Table~\ref{tab:comp:func:homosc:kno:OOD}.

\subsubsection{Unknown homoscedastic noise}\label{sec:comp:func:homosc:unk}

\begin{figure}[!ht]
	\centering
	\subcaptionbox{}{}{\includegraphics[width=0.32\textwidth]{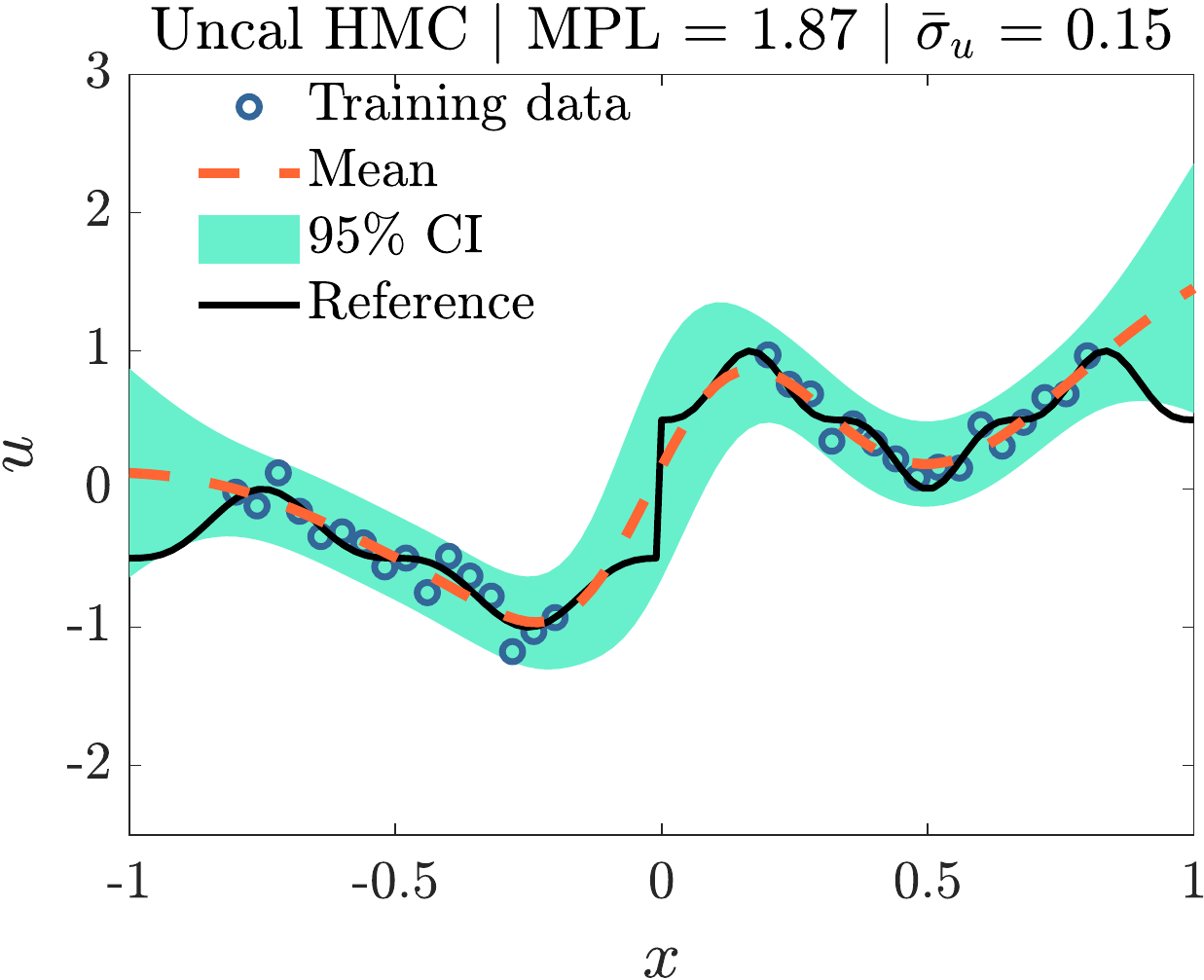}}
	\subcaptionbox{}{}{\includegraphics[width=0.32\textwidth]{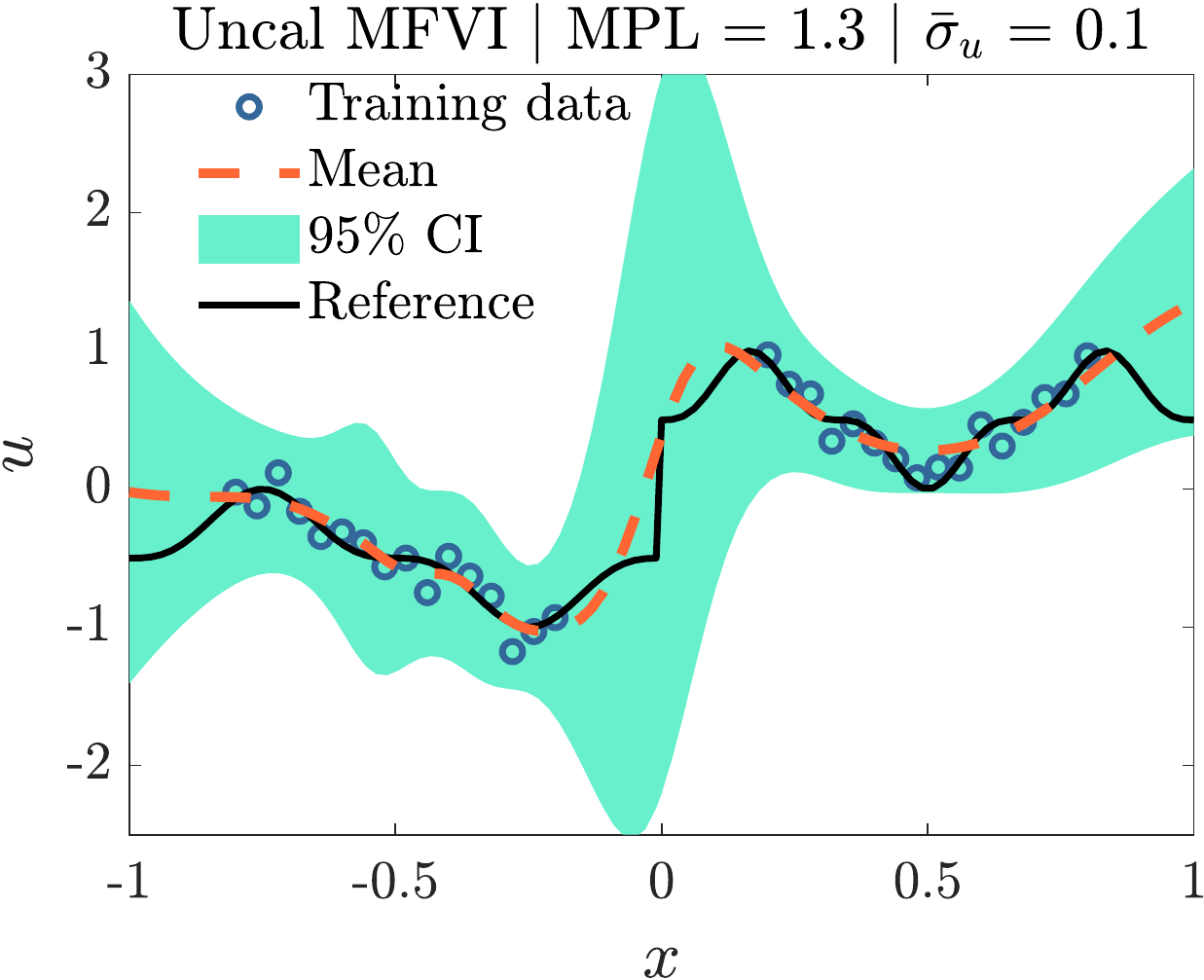}}
	\subcaptionbox{}{}{\includegraphics[width=0.32\textwidth]{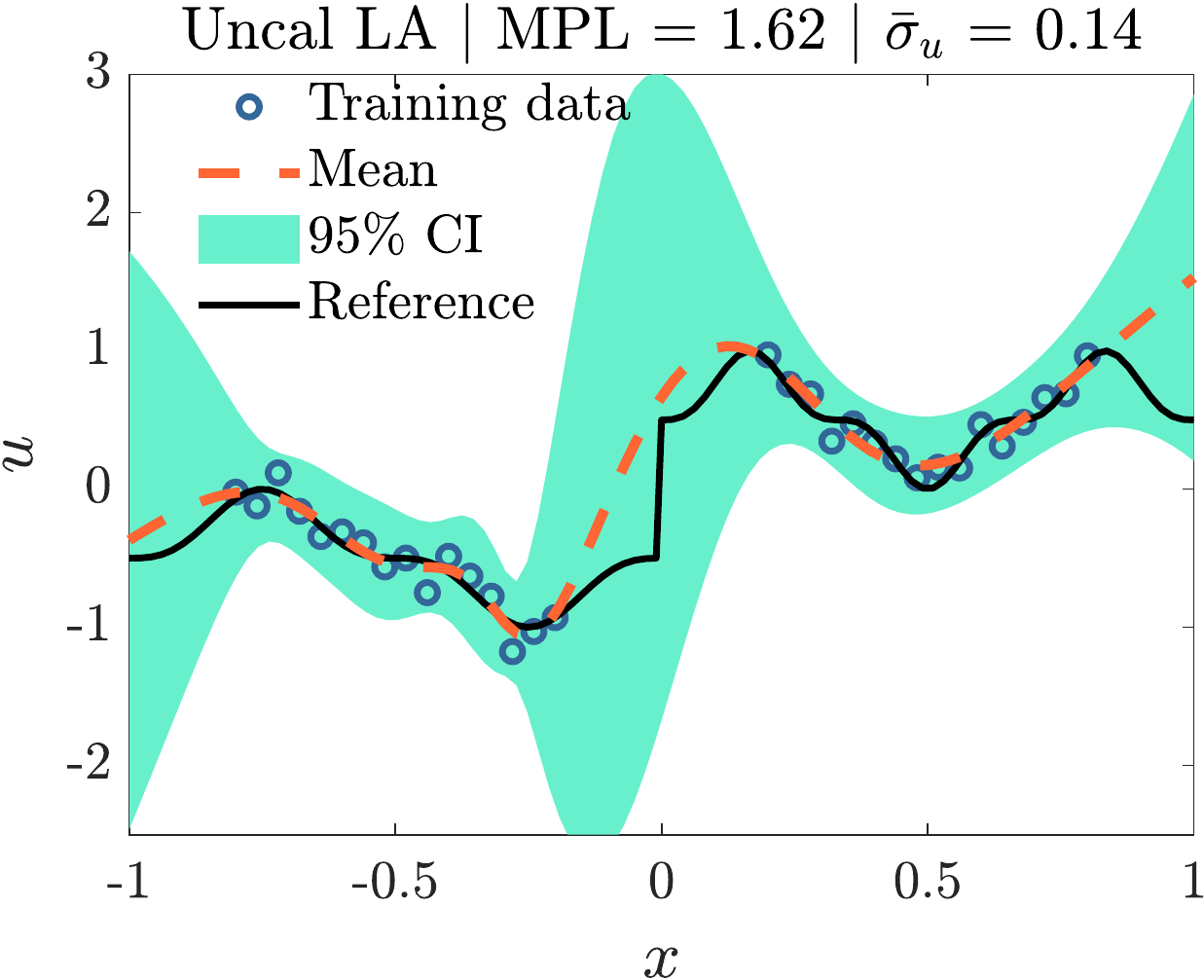}}
	\subcaptionbox{}{}{\includegraphics[width=0.32\textwidth]{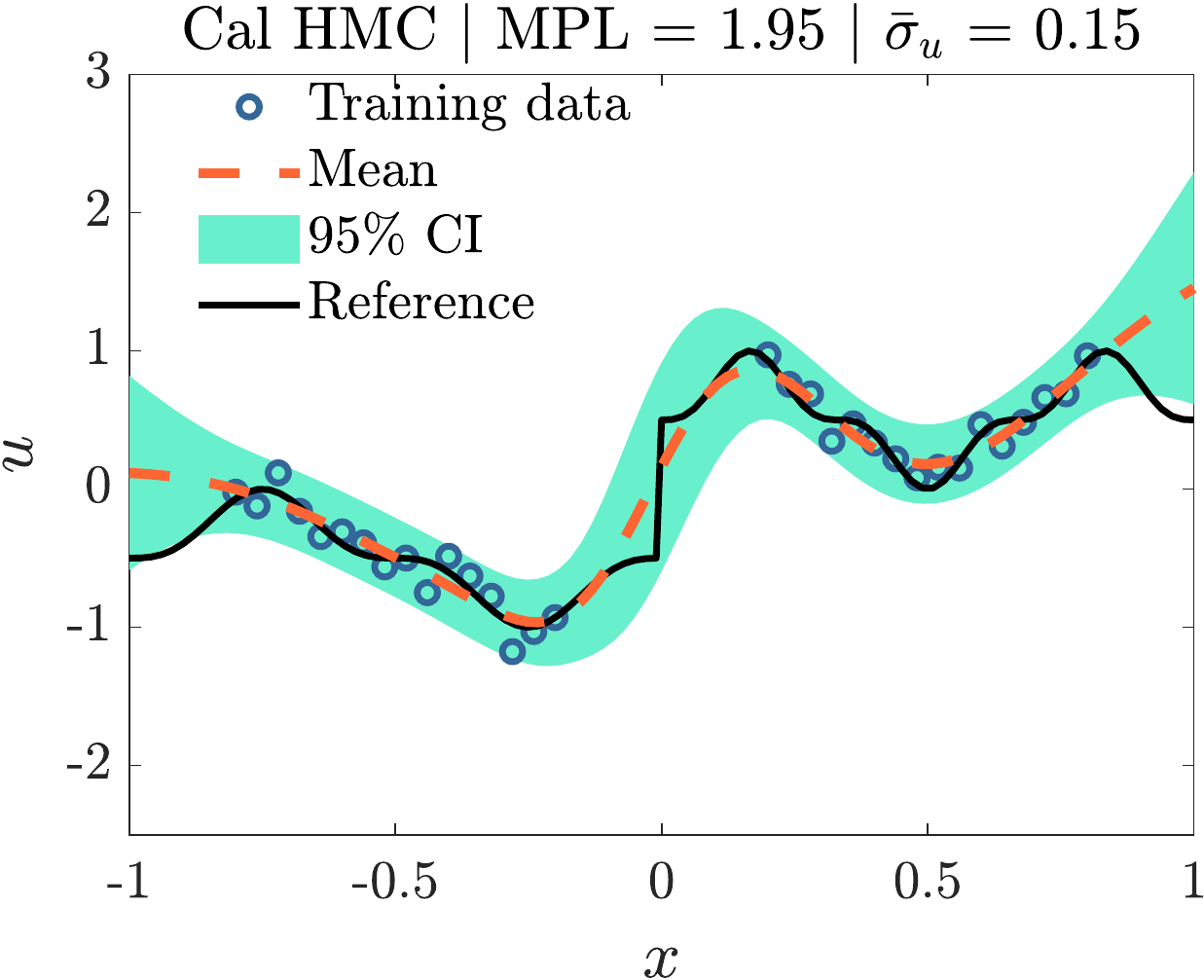}}
	\subcaptionbox{}{}{\includegraphics[width=0.32\textwidth]{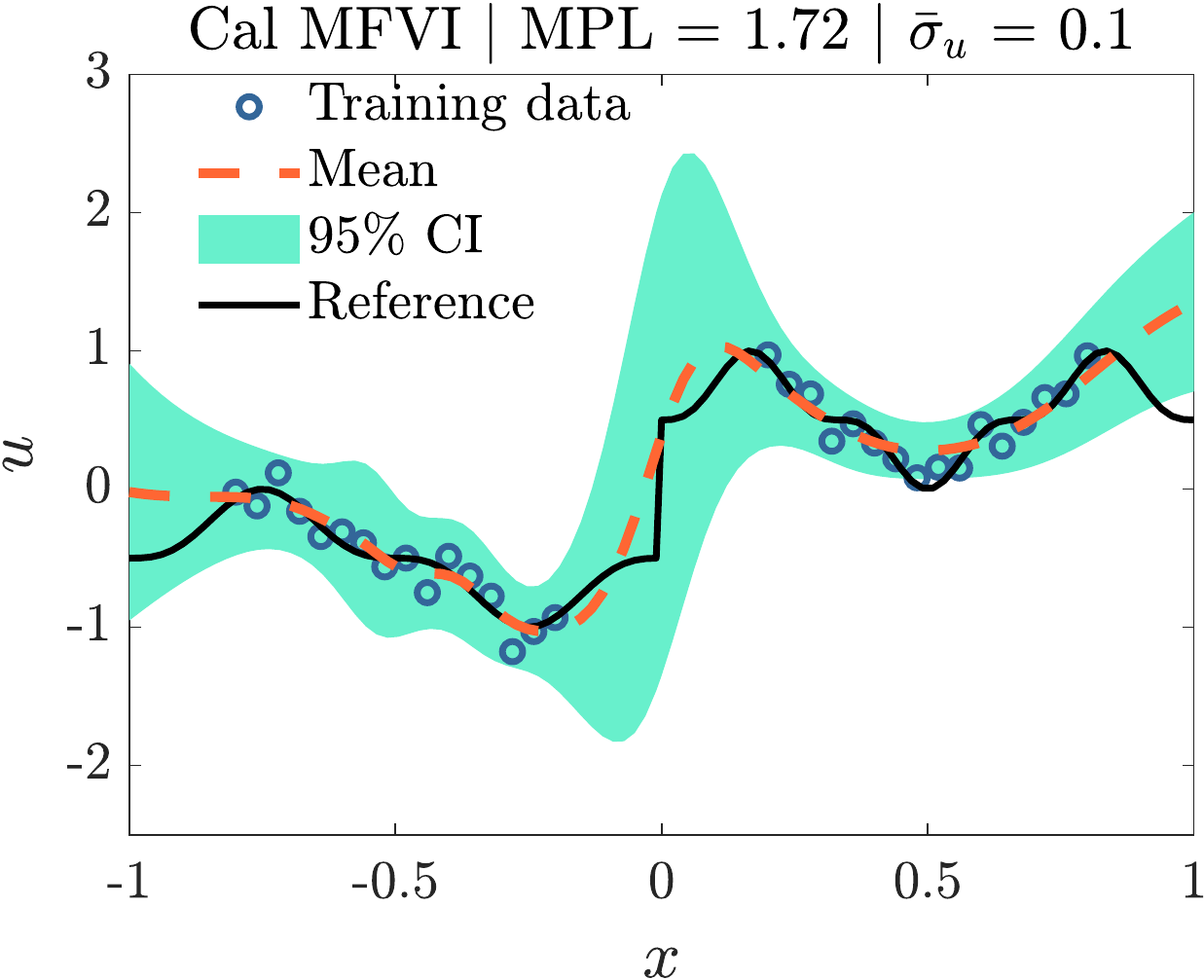}}
	\subcaptionbox{}{}{\includegraphics[width=0.32\textwidth]{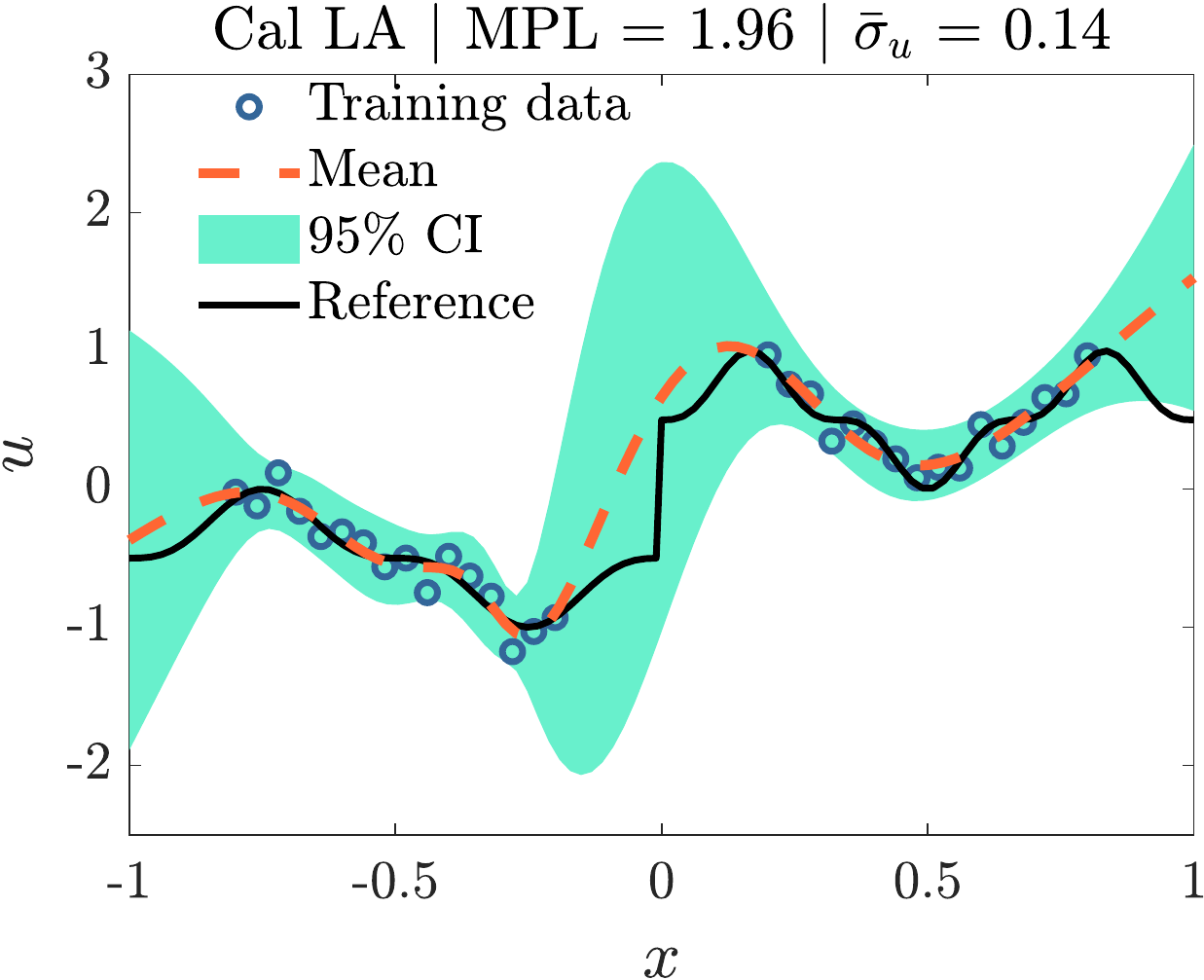}}
	\subcaptionbox{}{}{\includegraphics[width=0.32\textwidth]{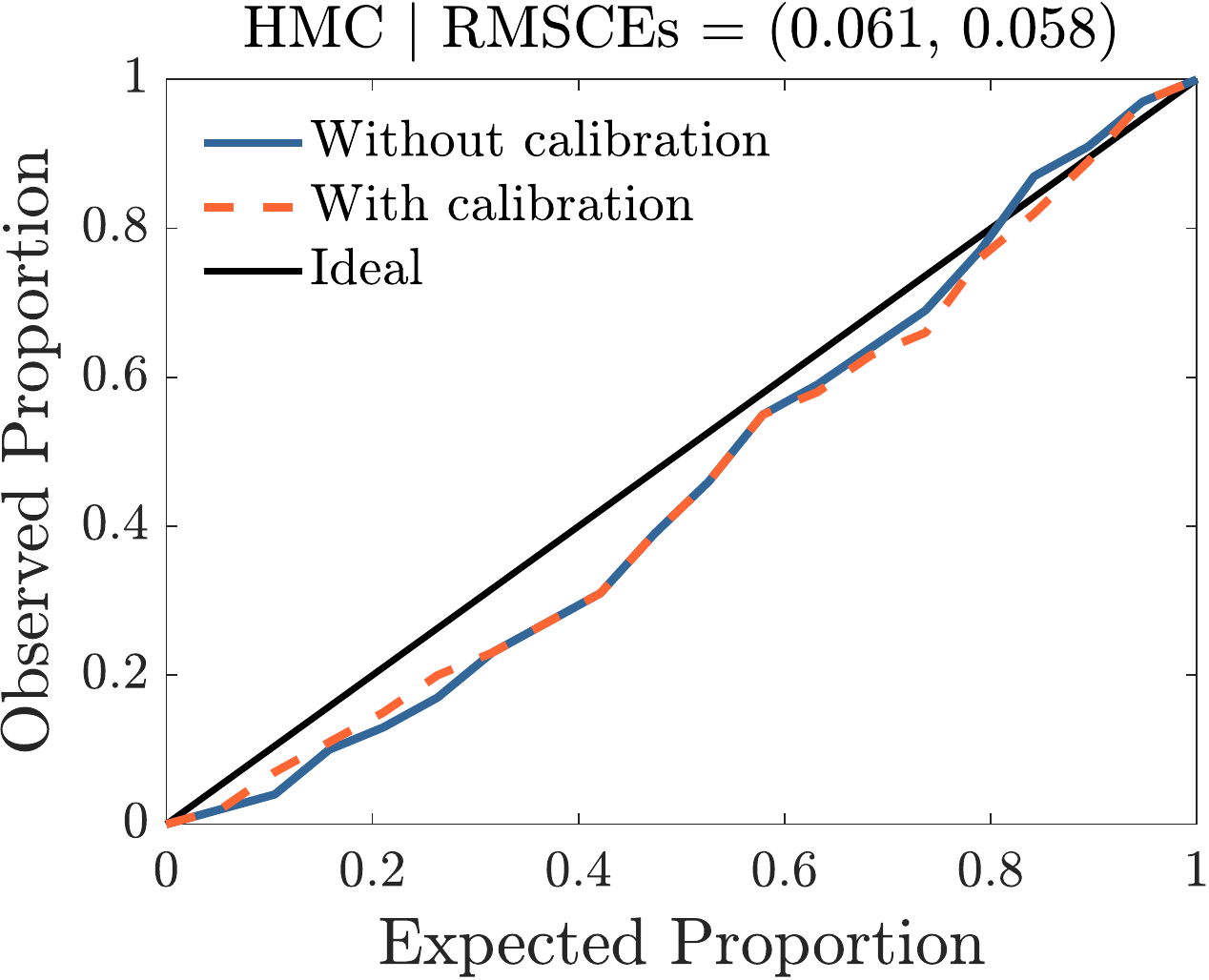}}
	\subcaptionbox{}{}{\includegraphics[width=0.32\textwidth]{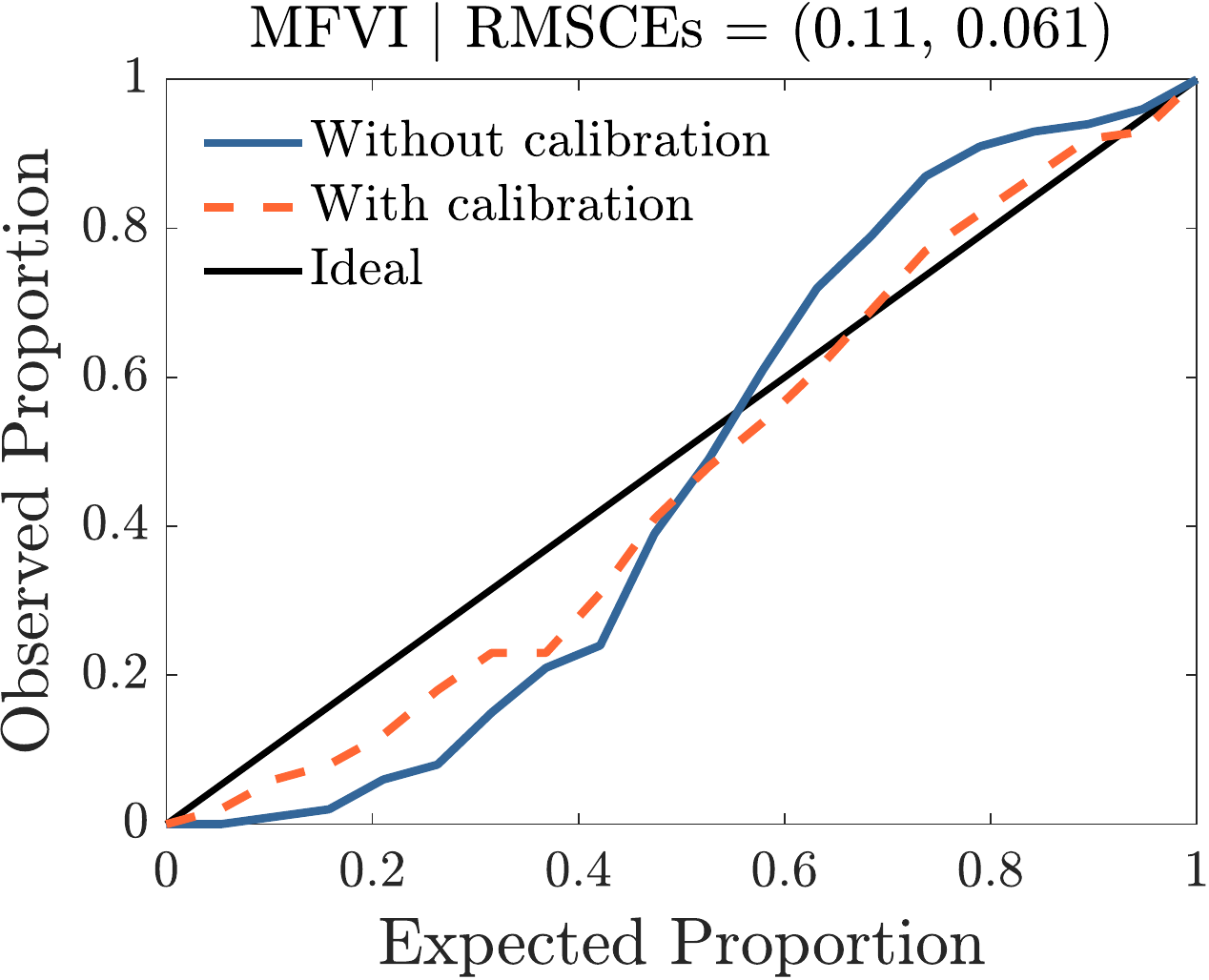}}
	\subcaptionbox{}{}{\includegraphics[width=0.32\textwidth]{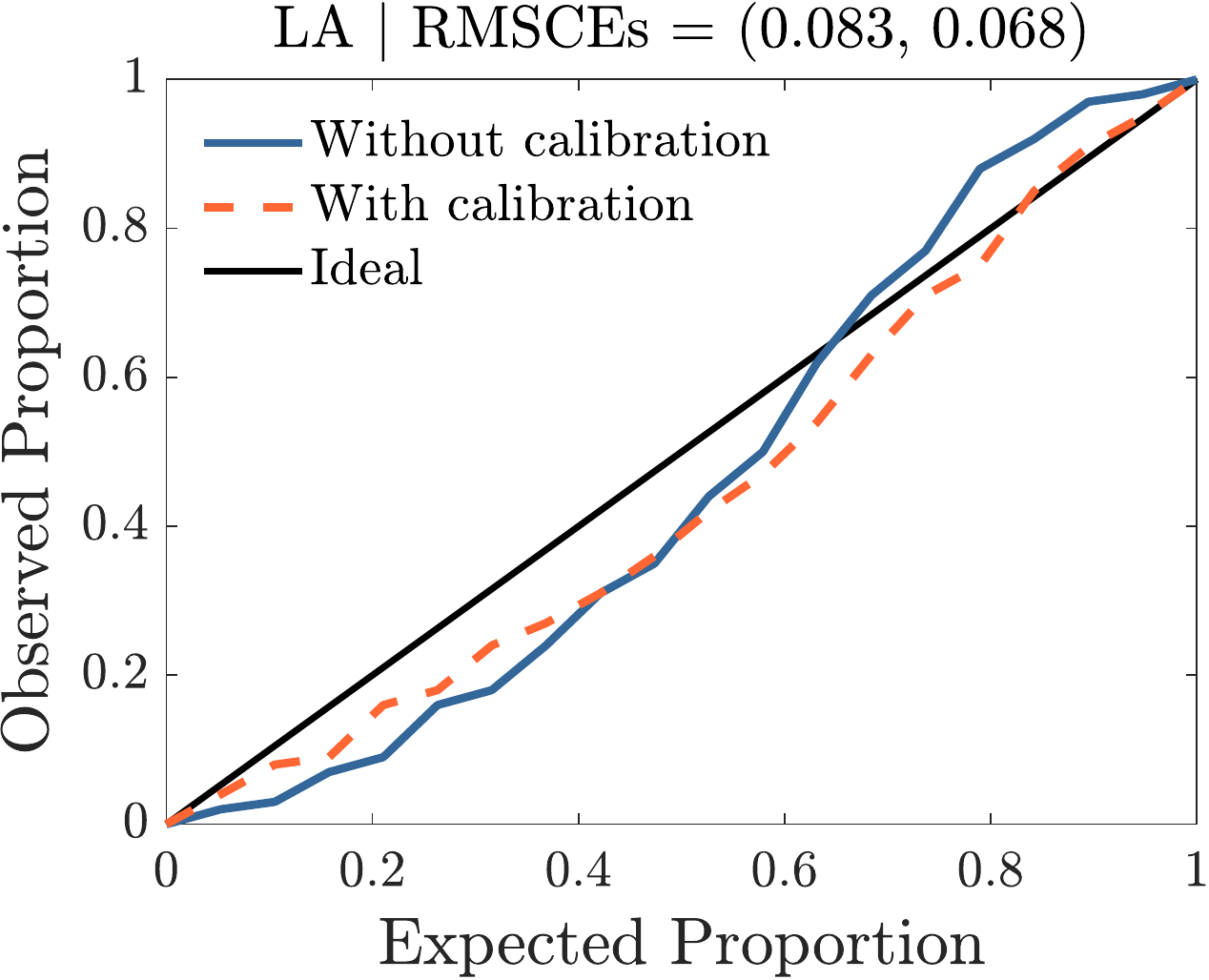}}
	\caption{
		Function approximation problem of Eq.~\eqref{eq:comp:func:func} | \textit{Unknown homoscedastic noise}:
		computationally cheaper techniques (MFVI and LA) perform comparably with HMC if their priors are learned.
		HMC remains the most calibrated candidate even following post-training calibration of the cheaper methods.
		Shown here are the training data and exact function, as well as the mean and total uncertainty ($95\%$ CI) predictions of HMC, MFVI, and LA with learned prior and noise scale.
		\textbf{Top row:} uncalibrated predictions. 
		\textbf{Middle row:} calibrated predictions (Section~\ref{sec:eval:calib}). 
		\textbf{Bottom row:} calibration plots (see also Fig.~\ref{fig:eval:eval:misscal}) and RMSCEs before and after post-training calibration (in the parentheses).  
		Learned noise scales are denoted by $\bar{\sigma}_u$, while the reference is $\sigma_u = 0.1$.
		``Uncal'' and ``Cal'' refer to results obtained before and after post-training calibration, respectively. 
	}
	\label{fig:comp:func:homosc:unk:res}
\end{figure}

In this section, the data noise follows $\cN(0, 0.1^2)$ and the noise scale is assumed to be unknown.
In Fig.~\ref{fig:comp:func:homosc:unk:res}, we present the mean and total uncertainty predictions obtained by HMC, MFVI and LA, as well as the corresponding calibration plots. 
In this figure, we include results before (``uncalibrated'') and after (``calibrated'') post-training calibration.
To obtain these results, the noise scale $\sigma_u$ and the prior $p(\theta)$ are learned online, as explained in Section~\ref{sec:uqt:bnns}.
In
Table~\ref{tab:comp:func:homosc:unk} we also provide the corresponding evaluation metrics for ID evaluation.
Overall, in this case computationally cheaper techniques, such as MFVI and LA, perform comparably with HMC, if their priors are learned. 
Nevertheles, HMC remains the method with the smallest calibration error for ID data evaluation, even following post-traing calibration of the cheaper methods.

\begin{figure}[!ht]
	\centering
	\subcaptionbox{}{}{\includegraphics[width=0.32\textwidth]{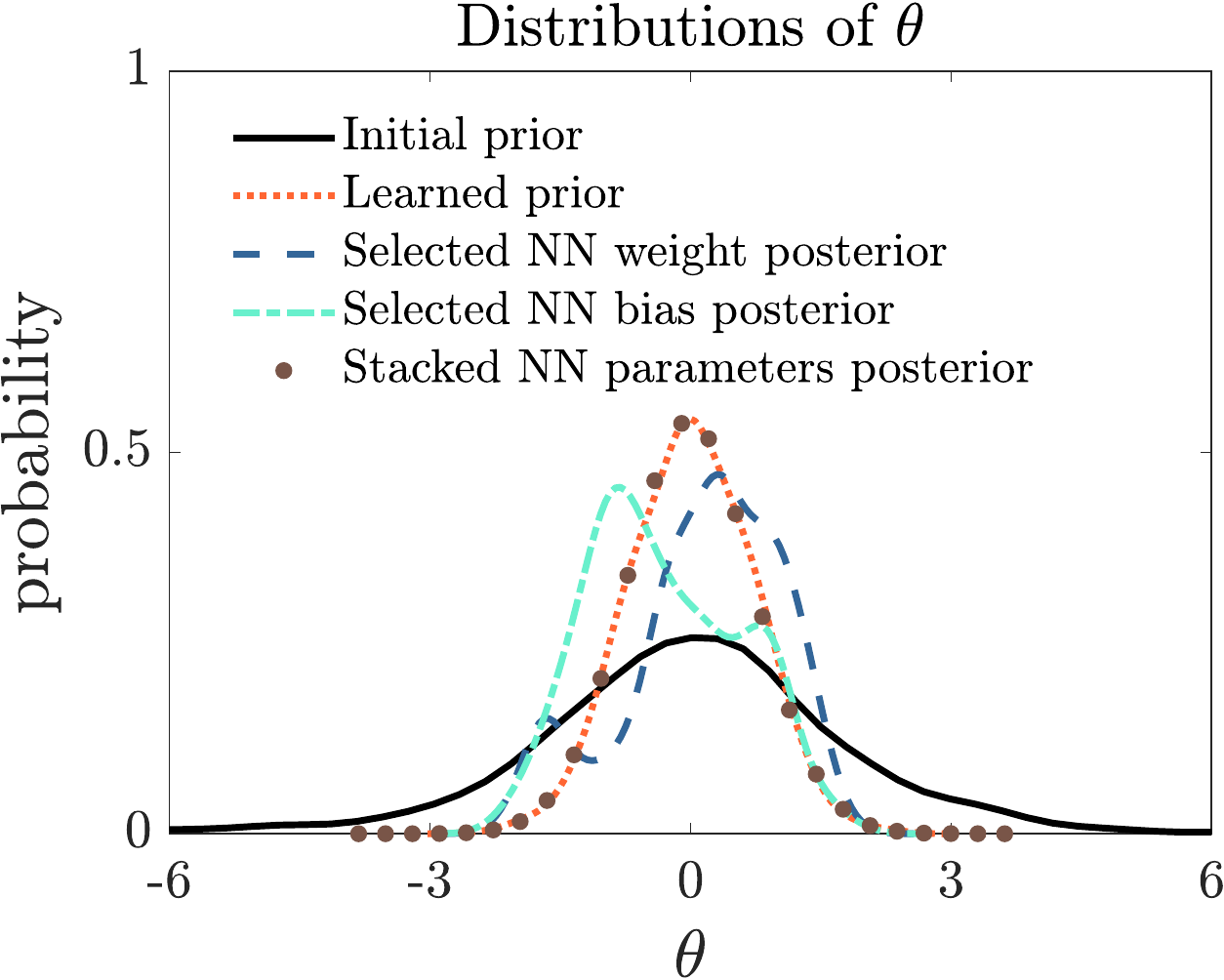}}
	\subcaptionbox{}{}{\includegraphics[width=0.32\textwidth]{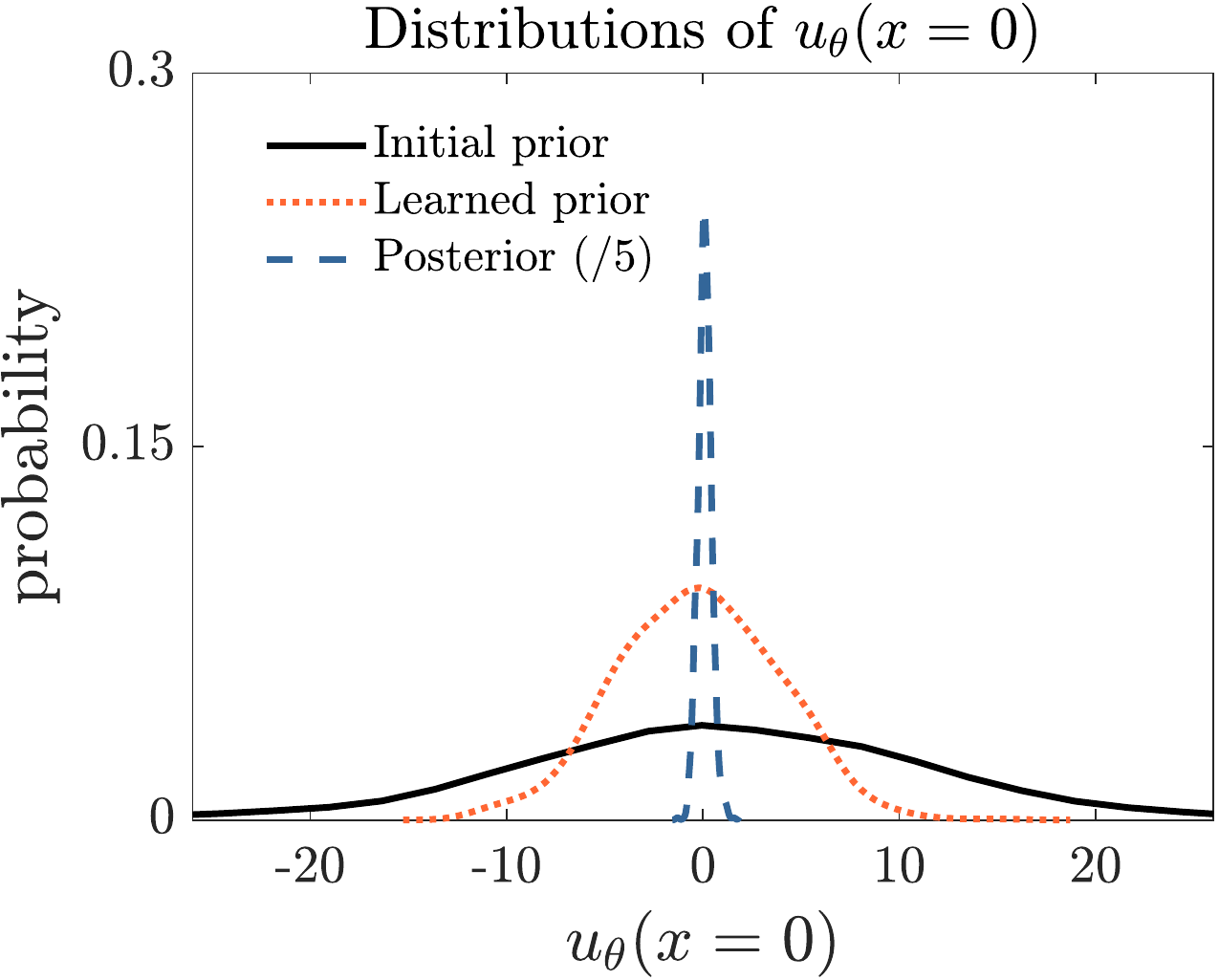}}
	\subcaptionbox{}{}{\includegraphics[width=0.32\textwidth]{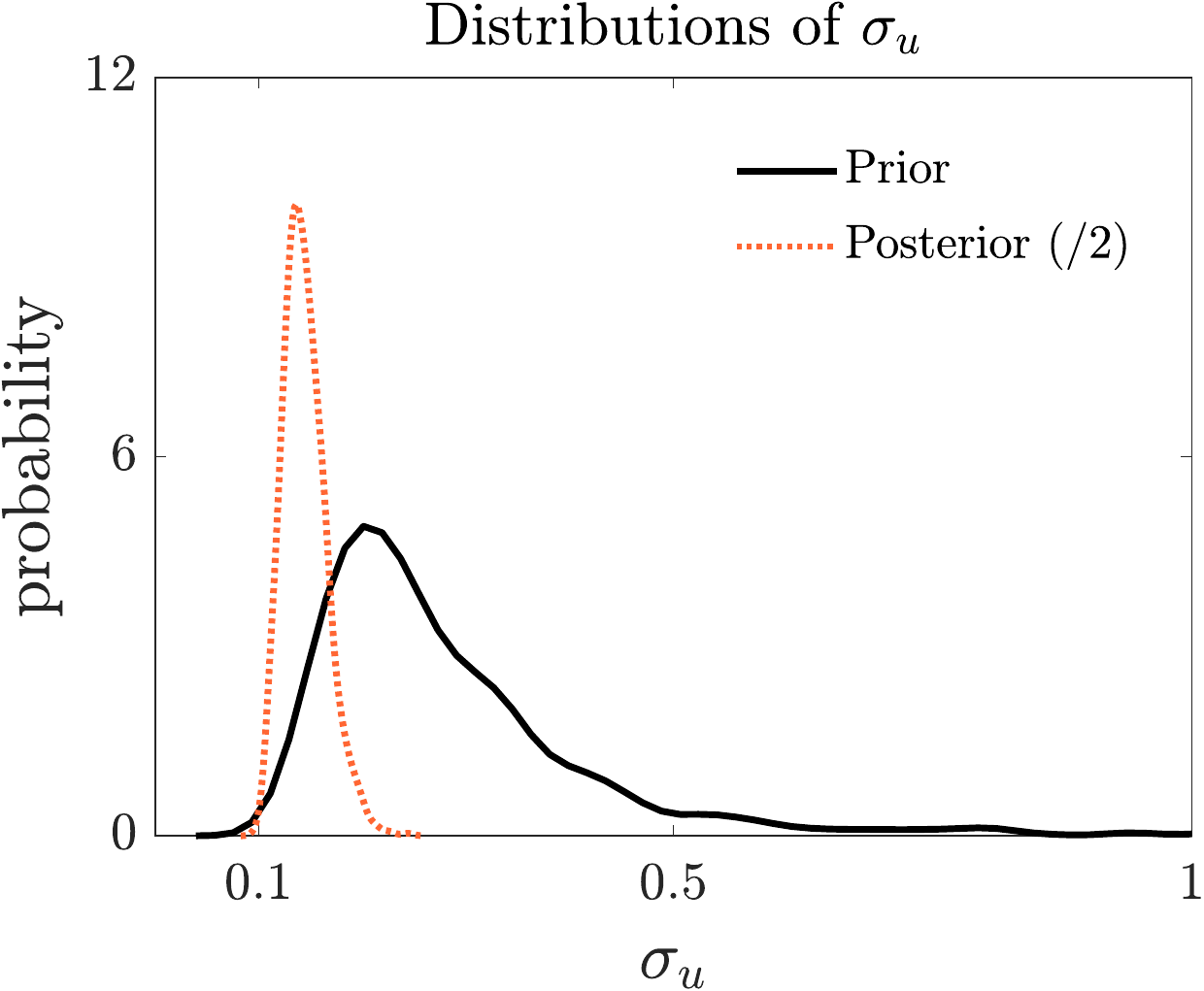}}
	\caption{
		Function approximation problem of Eq.~\eqref{eq:comp:func:func} | \textit{Unknown homoscedastic noise}:
		the initially spread out prior and noise scale distributions concentrate while the corresponding quantities are learned during HMC combined with Gibbs sampling.
		\textbf{(a)} Initial and learned prior $p(\theta)$, as well as posterior $p(\theta|\cD)$. For visualization, $p(\theta|\cD)$ is shown both for two randomly ``picked'' parameters, and for all dimensions of $\theta$ ``stacked'' together.
		Learned prior approximates the ``stacked'' posterior as explained in Section~\ref{app:methods:bnns:mcmc:hypers}.
		\textbf{(b)} Distributions of $u_{\theta}(0)$ based on prior and posterior distributions of $\theta$ from part (a) of the figure. 
		\textbf{(c)} Prior $p(\sigma_u)$ and learned posterior $p(\sigma_u|\cD)$ of the noise scale.
	}
	\label{fig:comp:func:homosc:unk:dists}
\end{figure}

\begin{table}[!ht]
	\centering
	\footnotesize
	\begin{tabular}{c|c|ccc|ccc}
		\toprule
		\multirow{5}{*}{\textbf{a}} & \multirow{2}{*}{Metric ($\times 10^2$)} & \multicolumn{3}{c|}{ID} & \multicolumn{3}{c}{OOD}\\ \cline{3-8}
		&& HMC& MFVI& LA& HMC& MFVI & LA \\
		\cline{2-8}
		&RL2E ($\downarrow$) & 23.7 & 25.3 & \textbf{22.9} & 56.4 & \textbf{51.7} & 66.3 \\ 
		&MPL ($\uparrow$) & \textbf{187.3} & 130.2 & 162.4 & \textbf{100.9} & 61.4 & 57.2 \\ 
		&RMSCE ($\downarrow$) & \textbf{6.1} & 11.0 & 8.3 & 17.0 & \textbf{15.6} & 21.9 \\
		\midrule
		\midrule	
		\multirow{3}{*}{\textbf{b}} &Reference &\multicolumn{6}{c}{Learned noise (aleatoric uncertainty)} \\
		\cline{2-8}	
		& \multirow{2}{*}{ $\sigma_u= 0.1$} & \multicolumn{2}{c}{HMC} & \multicolumn{2}{c}{MFVI} & \multicolumn{2}{c}{LA} 
		\\
		\cline{3-8}
		&& \multicolumn{2}{c}{0.15} & \multicolumn{2}{c}{\textbf{0.1}} & \multicolumn{2}{c}{0.14} \\
		\bottomrule
	\end{tabular}
	\caption{
		Function approximation problem of Eq.~\eqref{eq:comp:func:func} | \textit{Unknown homoscedastic noise}:
		HMC, MFVI, and LA with learned priors and noise scales are comparable in terms of accuracy (RL2E) and predictive capacity (MPL) for ID evaluation.
		Here we evaluate ID and OOD performance of HMC, MFVI, and LA, using noisy test data and uncalibrated predictions (part a), and the accuracy of the learned noise scale (part b).
	}
	\label{tab:comp:func:homosc:unk}
\end{table}

Next, Fig.~\ref{fig:comp:func:homosc:unk:dists} shows the prior and posterior distributions of the NN parameters $\theta$, of the prediction $u_{\theta}(x=0)$, and of the predicted noise scale.
They are obtained using HMC combined with Gibbs sampling for prior and noise scale learning (Section~\ref{app:methods:bnns:mcmc:hypers}).
In all cases, the initially spread out prior and noise scale distributions concentrate while the corresponding quantities are learned during HMC combined with Gibbs sampling.
Further, as explained in Section~\ref{app:methods:bnns:mcmc:hypers} and corroborated in Fig.~\ref{fig:comp:func:homosc:unk:dists}a, the learned prior approximates the ``stacked'' posterior during prior learning with Gibbs sampling, which is an one-dimensional distribution of the NN parameters stacked together.
Detailed descriptions of the plots in Fig.~\ref{fig:comp:func:homosc:unk:dists} can be found in Section~\ref{app:comp:func:explain:dists}.

\subsubsection{Unknown Gaussian heteroscedastic noise}\label{sec:comp:func:hetero}

\begin{figure}[!ht]
	\centering
	\subcaptionbox{}{}{\includegraphics[width=0.3\textwidth]{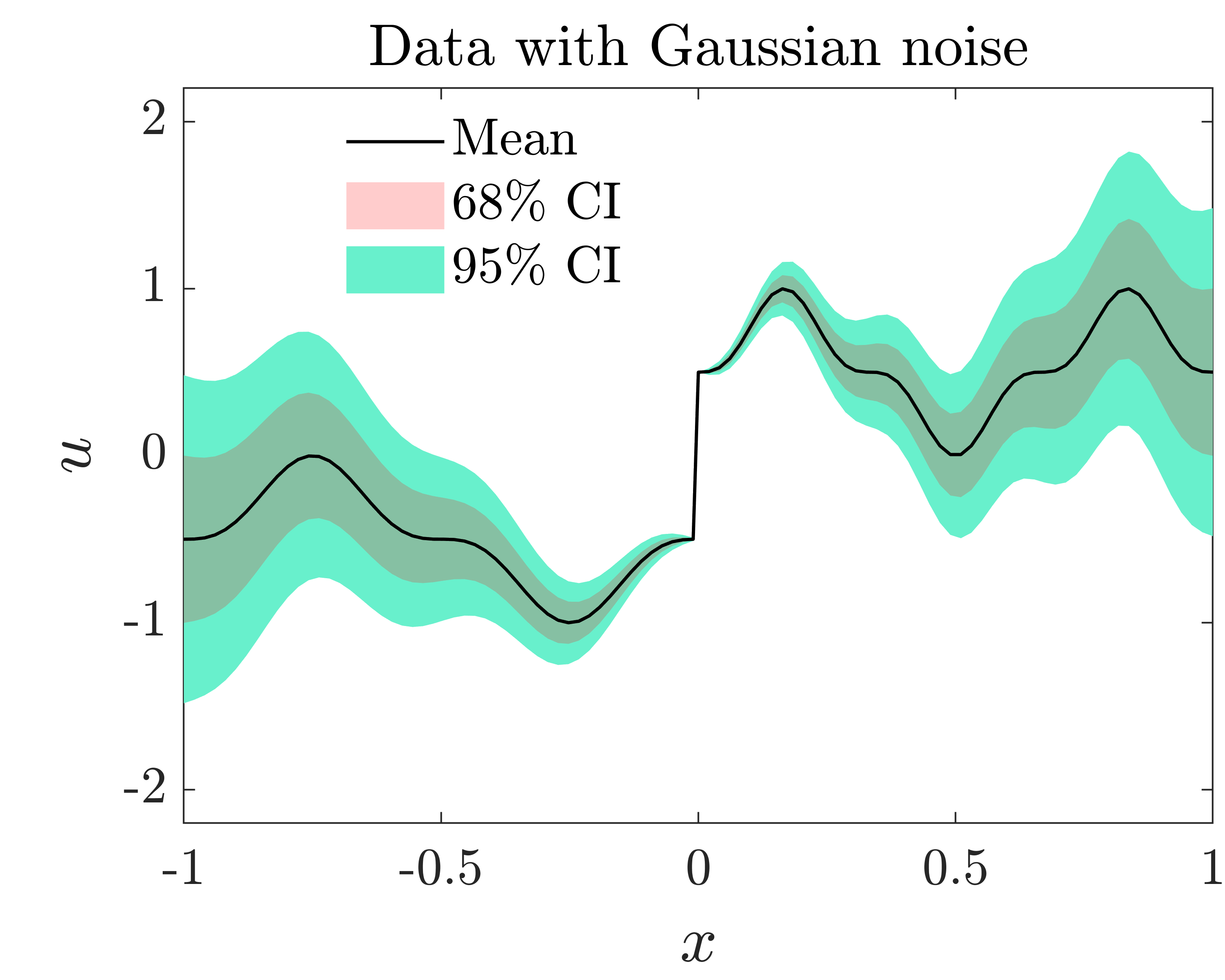}}
	\subcaptionbox{}{}{\includegraphics[width=0.3\textwidth]{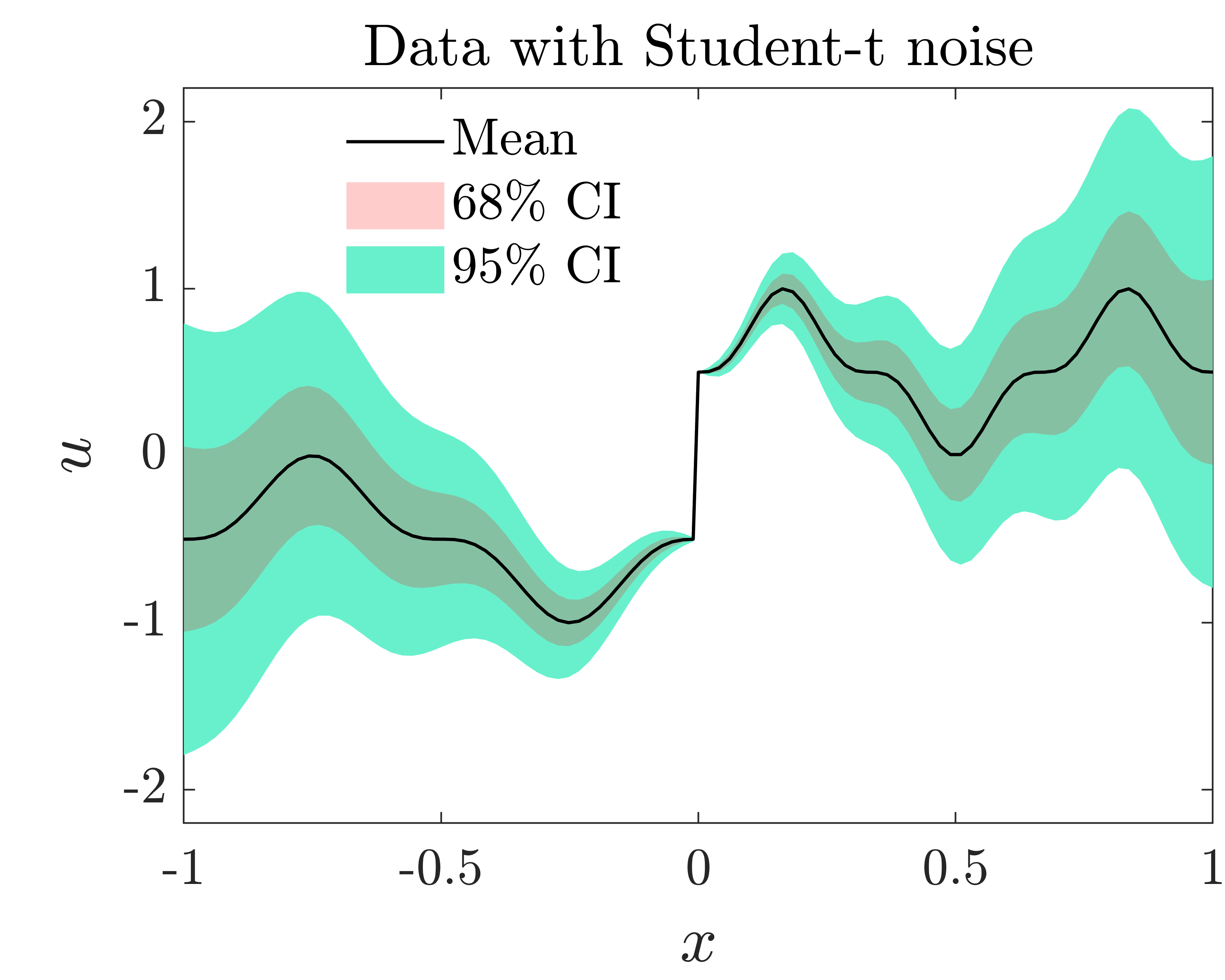}}
	\caption{
		Function approximation problem of Eq.~\eqref{eq:comp:func:func} | \textit{Unknown heteroscedastic noise}:
		data-generating process for Gaussian (a) and Student-t with 5 degrees of freedom (b) distributions.
		Although $68 \%$ noise CIs are comparable for the Gaussian and Student-t cases, the $95 \%$ intervals differ considerably because of the heavy tails of Student-t distribution.
		See Section~\ref{app:comp:func:results:student} for computational results pertaining to data with Student-t noise.
	}
	\label{fig:comp:func:hetero:datagen}
\end{figure}

In this section, the data noise for each $x$ follows $\cN(0, (0.5 |x|)^2)$; see Fig.~\ref{fig:comp:func:hetero:datagen}. 
For implementing h-HMC, h-MFVI, and h-DEns, the NN approximator has two outputs; one for the mean $u_{\theta}(x)$ and one for the location-dependent variance $\sigma^2_u(x)$ of the Gaussian likelihood. 
Subsequently, we perform posterior inference using each one of HMC, MFVI, DEns to obtain samples $\{\hat{\theta}_j\}_{j=1}^M$, which we use in Eq.~\eqref{eq:modeling:noise:model:totvar} to obtain total uncertainty including heteroscedastic noise.
Further, for implementing h-HMC+FP, we pre-train a GAN-FP that learns, using historical data, the heteroscedastic aleatoric uncertainty.
In this case, the historical data comprise of realizations from the BNN-FP, each one contaminated with a different realization from the noise following $\cN(0, (0.5 |x|)^2)$.
For each $x$ and for each random input $\theta$, the GAN generator outputs a random output that depends on both $x$ and $\theta$, and a deterministic output that depends on $x$ only; see Section~\ref{app:comp:architecture} for more details.
The random and deterministic parts are combined and compared with the historical data for training the GAN.
The GAN-FP learns the clean FP via the random output, and the heteroscedastic noise via the deterministic output.
Subsequently, the NN approximator in h-HMC+FP has only one output, i.e., $u_{\theta}(x)$, and posterior inference for $\theta$ is performed using HMC.

\begin{table}[!ht]
	\centering
	\footnotesize
	\begin{tabular}{c|cccc|cccc}
		\toprule
		Metric & \multicolumn{4}{c|}{ID} & \multicolumn{4}{c}{OOD}\\ \cline{2-9}
		($\times 10^2$) & h-HMC & h-MFVI & h-DEns &h-HMC+FP  & h-HMC & h-MFVI & h-DEns &h-HMC+FP \\
		\midrule
		RL2E ($\downarrow$) & 48.9 & 49.1 & 48.5 & \textbf{47.9} & 59.0 & 73.1 & \textbf{86.1} & 66.8 \\ 
		MPL ($\uparrow$) & 96.6 & 110.1 & \textbf{133.8} & 96.1 & 60.5 & \textbf{96.6} & 57.7 & 68.4 \\ 
		RMSCE ($\downarrow$) & \textbf{6.1} & 8.5 & 9.5 & 6.9 & 17.5 & 26.3 & 40.1 & \textbf{13.5} \\ 
		\bottomrule
	\end{tabular}
	\caption{
		Function approximation problem of Eq.~\eqref{eq:comp:func:func} | \textit{Unknown Gaussian heteroscedastic noise}:
		combining HMC with GAN using historical data to learn the heteroscedastic noise (h-HMC+FP) outperforms the UQ methods that use a BNN prior (h-HMC, h-MFVI, h-DEns).
		Here we evaluate ID and OOD performance using noisy test data and uncalibrated predictions.
		``h'' indicates that heteroscedastic noise modeling has be used.
		See also Section~\ref{app:comp:func:results:student} for computational results pertaining to data with Student-t noise.
	}
	\label{tab:comp:func:hetero}
\end{table}

\begin{figure}[!ht]
	\centering
	\subcaptionbox{}{}{\includegraphics[width=0.24\textwidth]{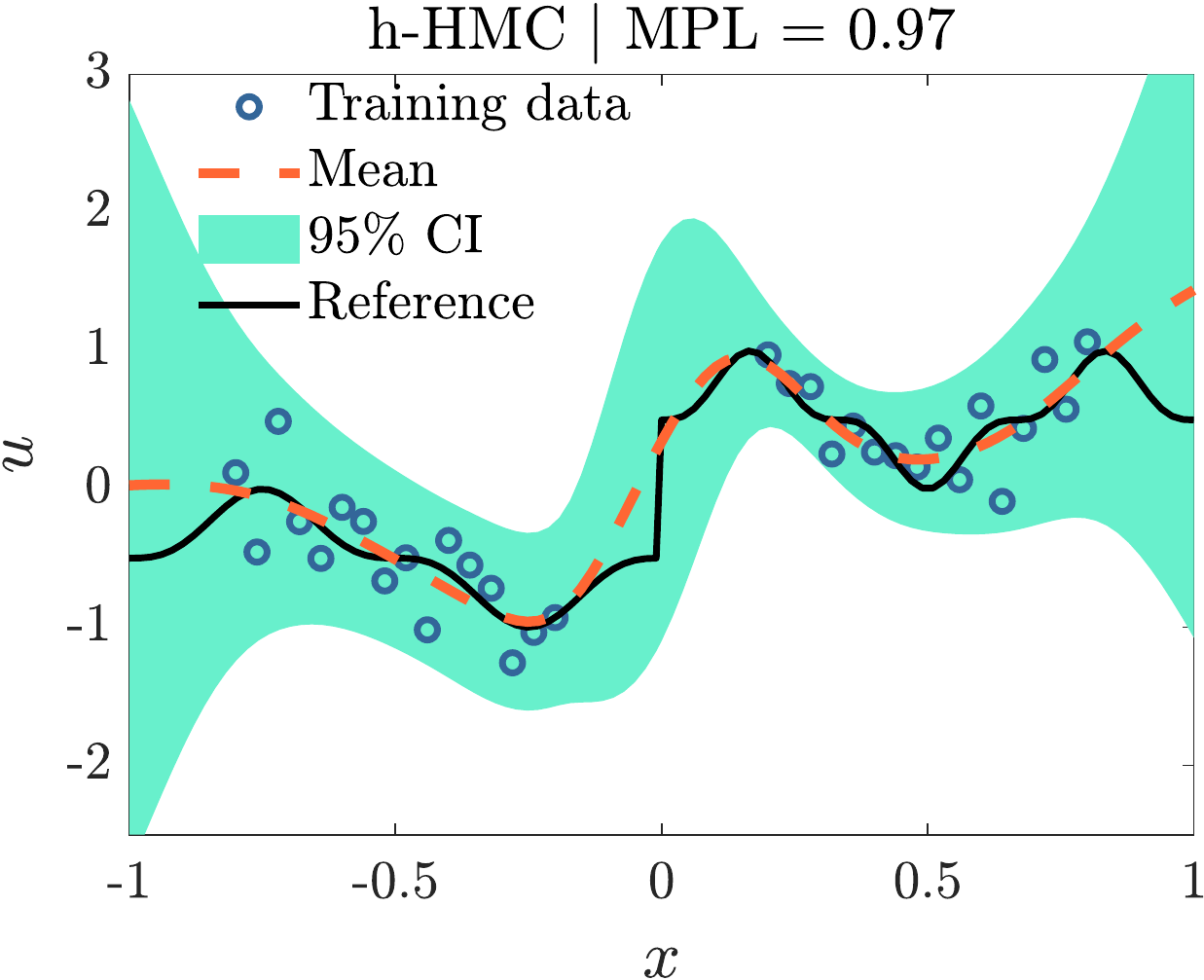}}
	\subcaptionbox{}{}{\includegraphics[width=0.24\textwidth]{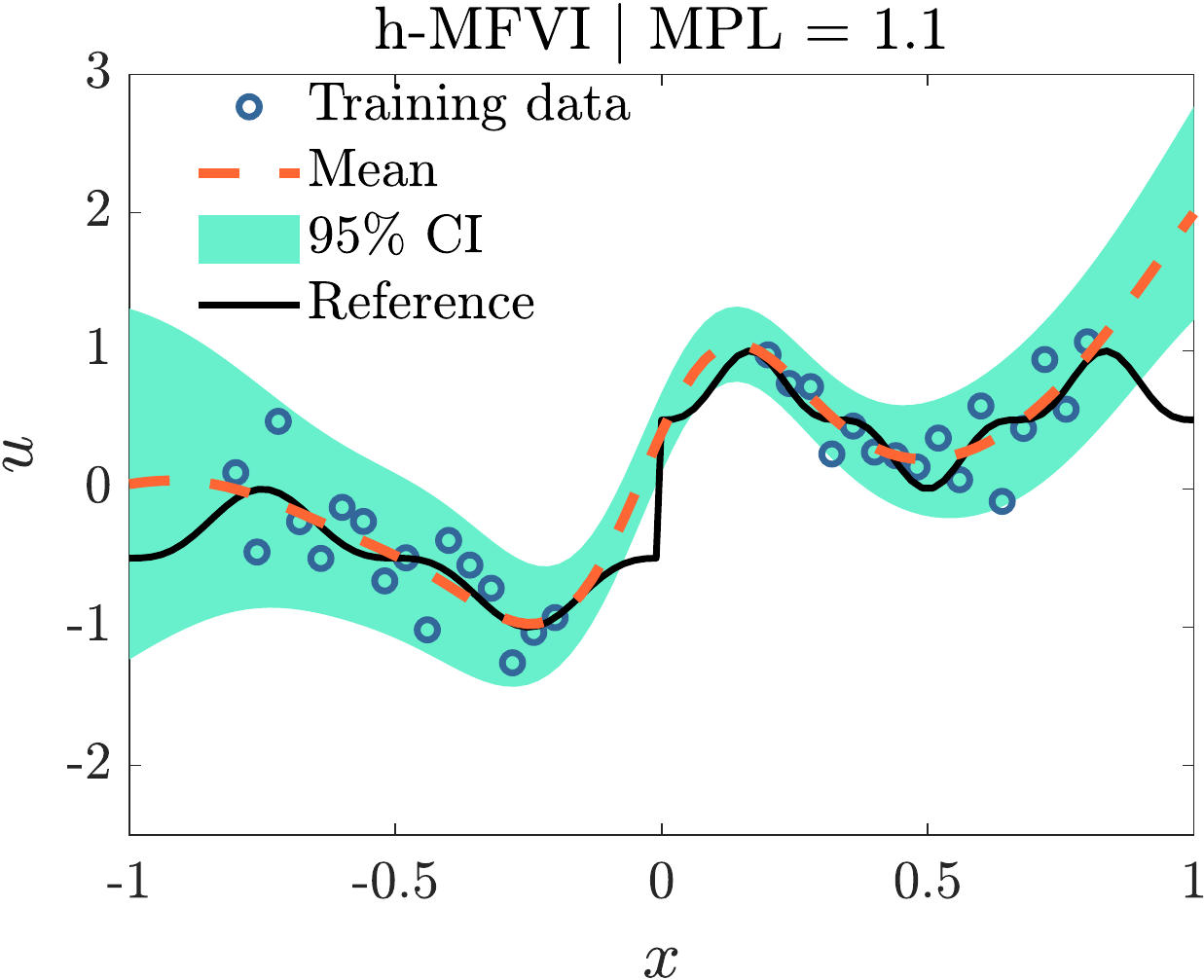}}
	\subcaptionbox{}{}{\includegraphics[width=0.24\textwidth]{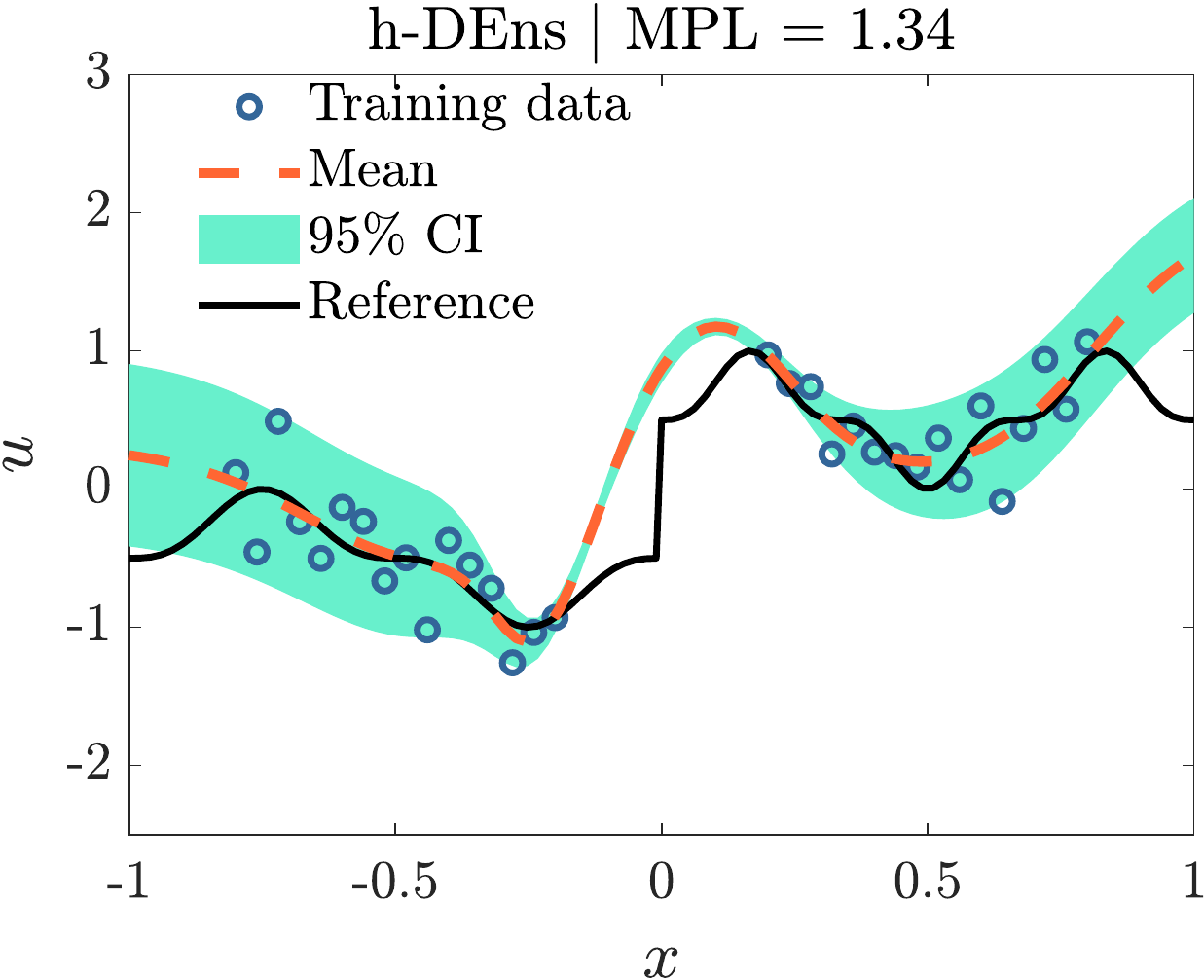}}
	\subcaptionbox{}{}{\includegraphics[width=0.24\textwidth]{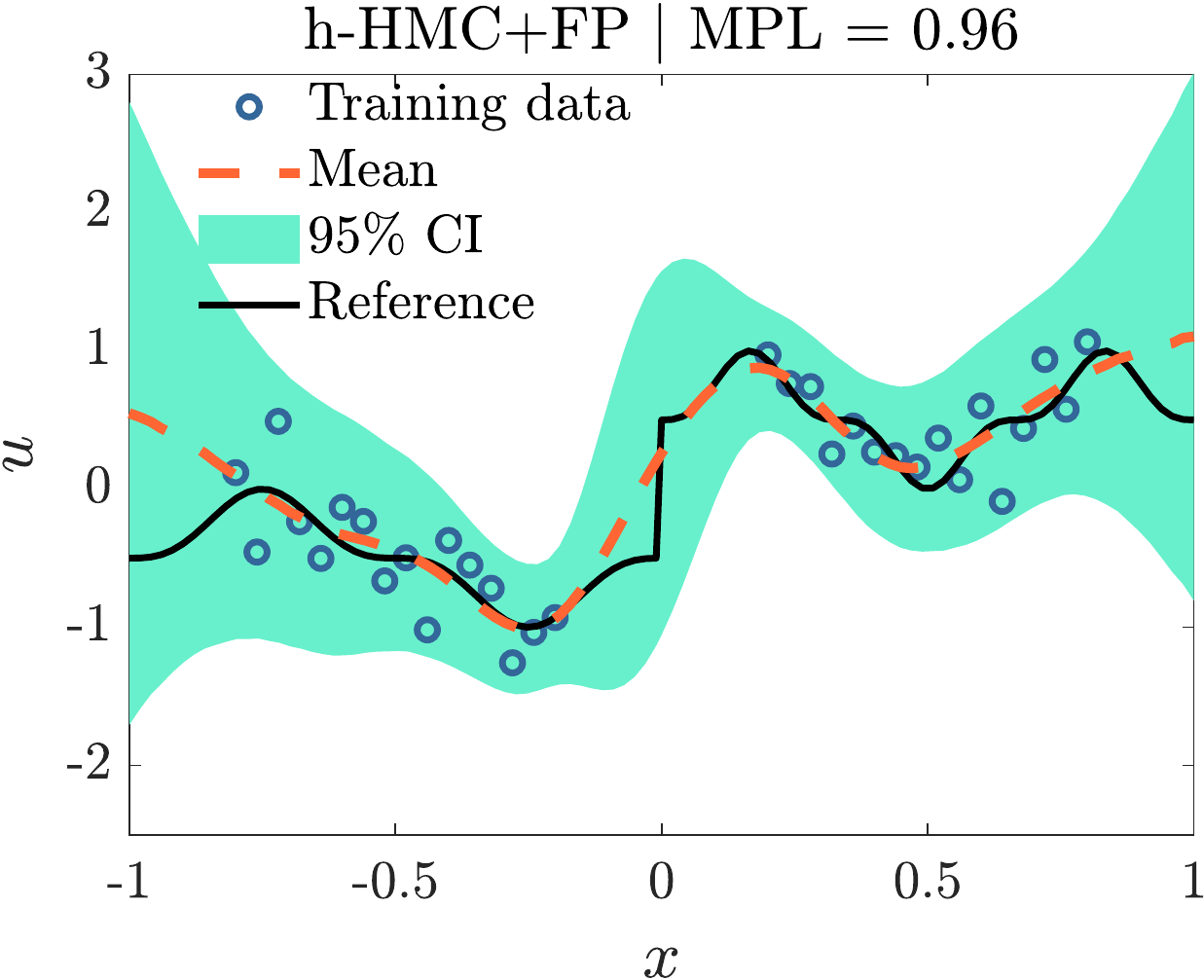}}
	\subcaptionbox{}{}{\includegraphics[width=0.24\textwidth]{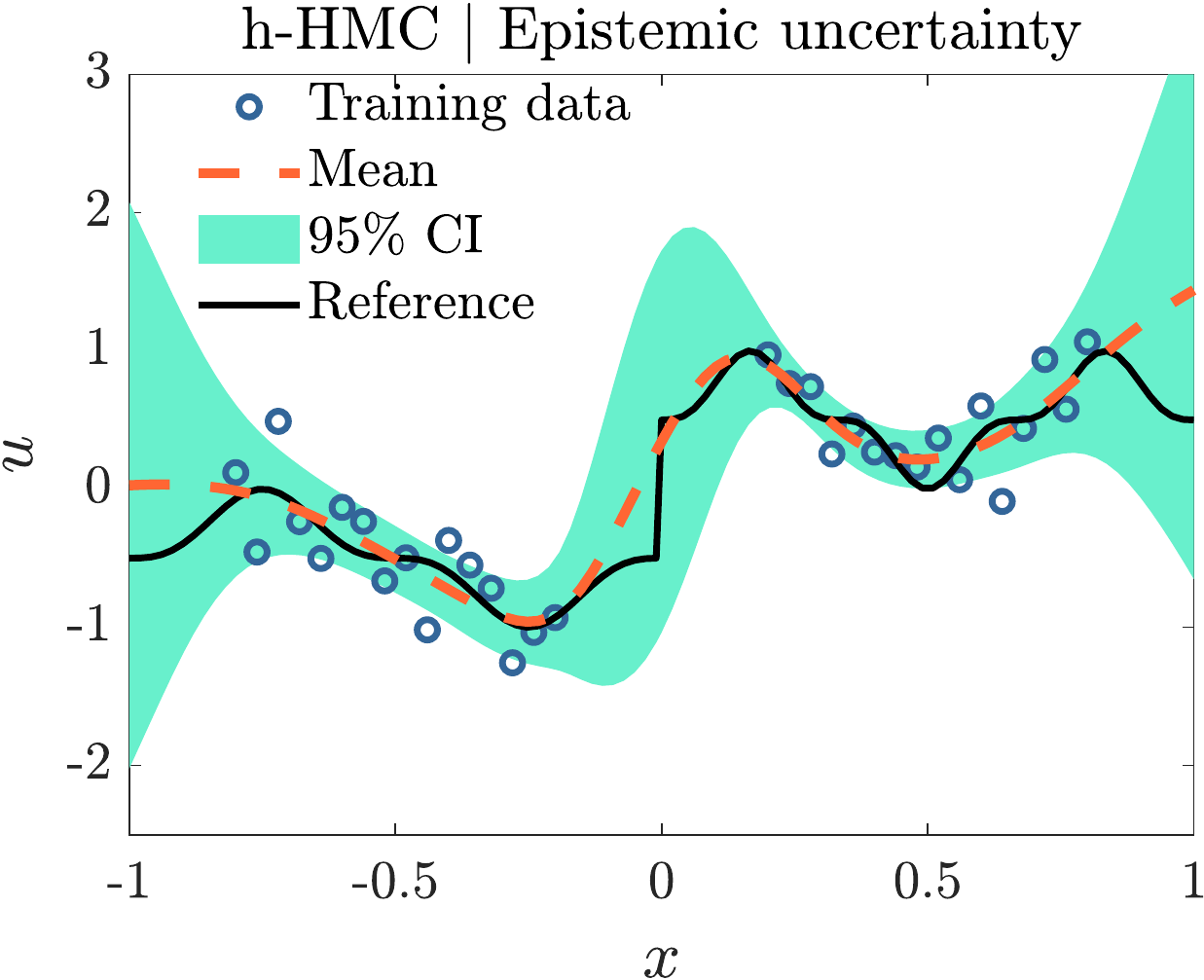}}
	\subcaptionbox{}{}{\includegraphics[width=0.24\textwidth]{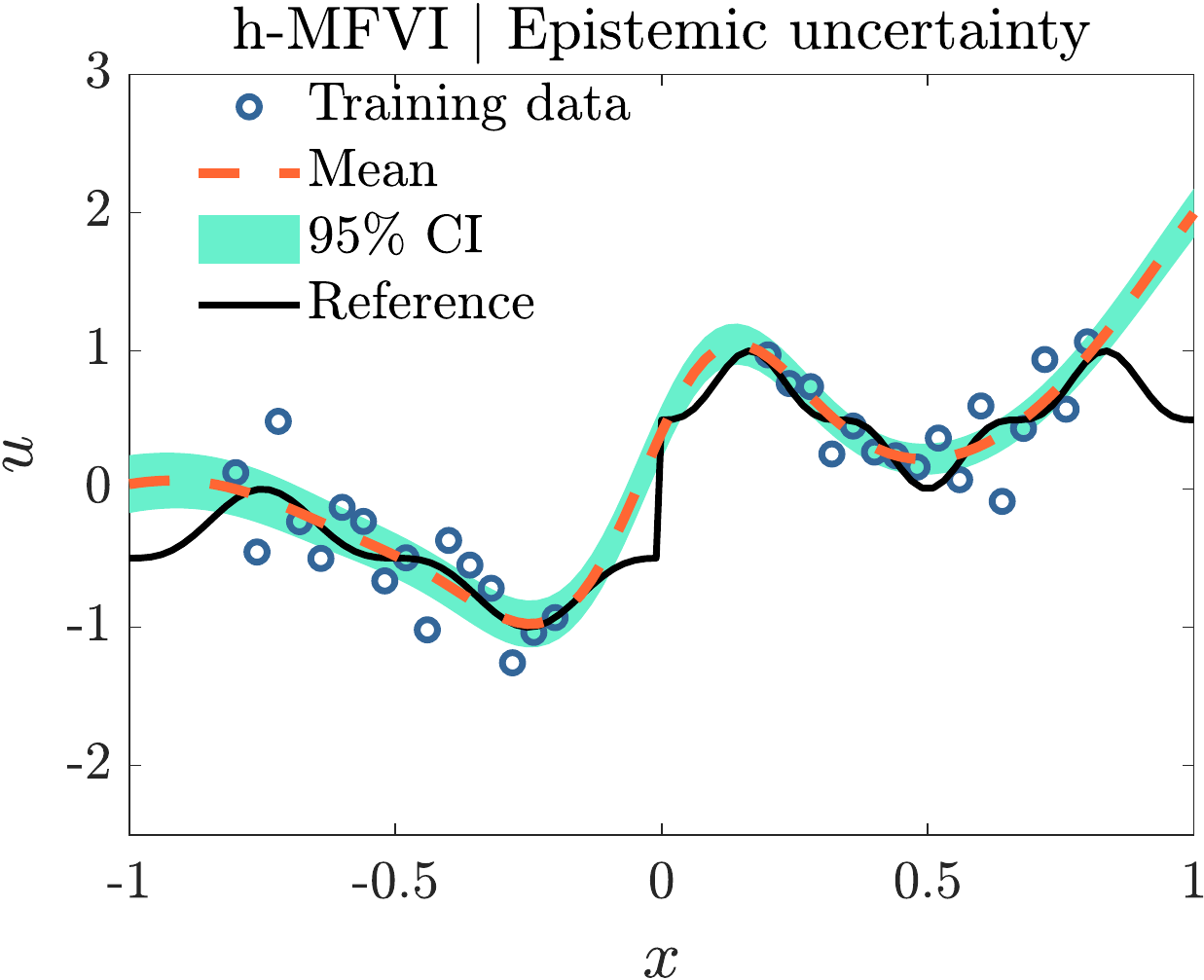}}
	\subcaptionbox{}{}{\includegraphics[width=0.24\textwidth]{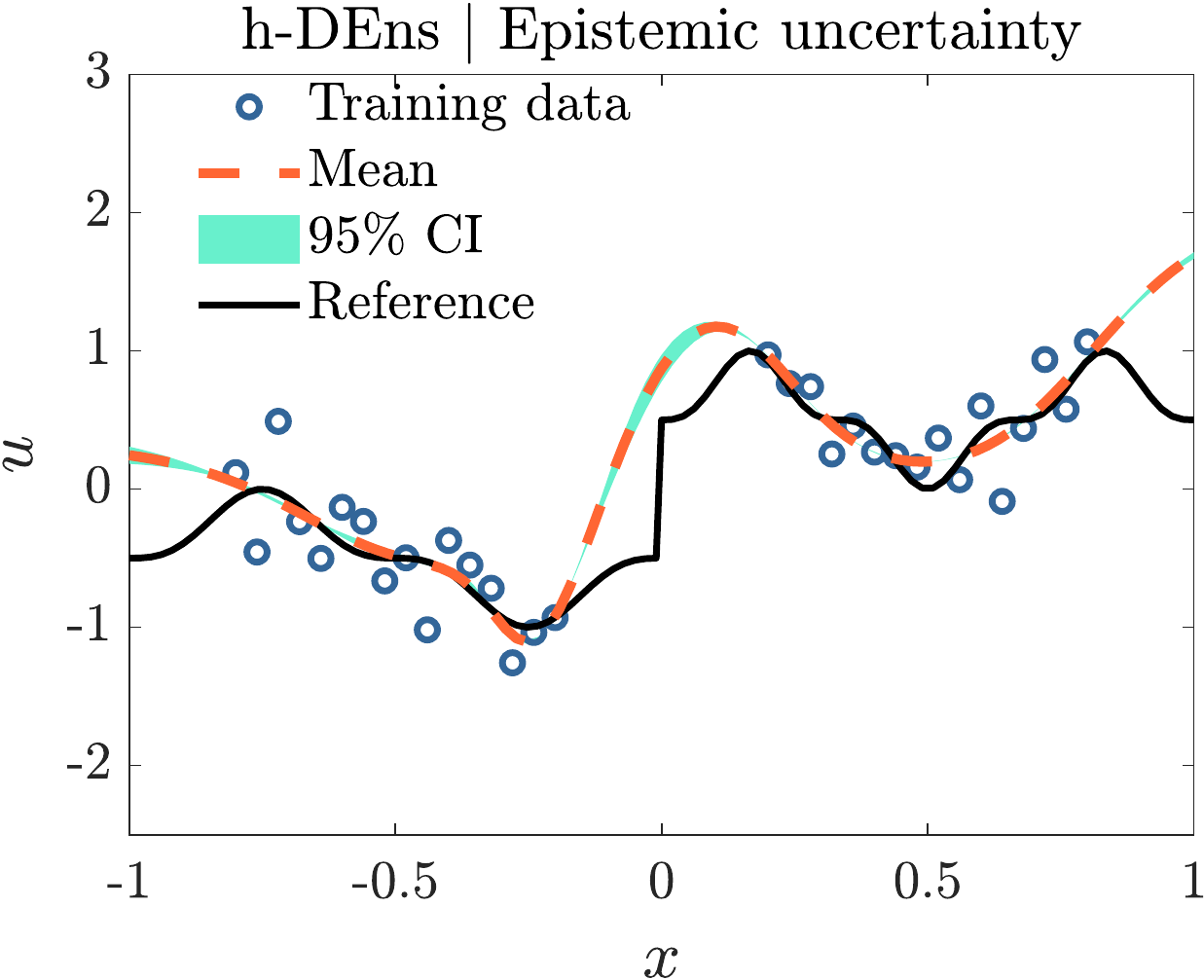}}
	\subcaptionbox{}{}{\includegraphics[width=0.24\textwidth]{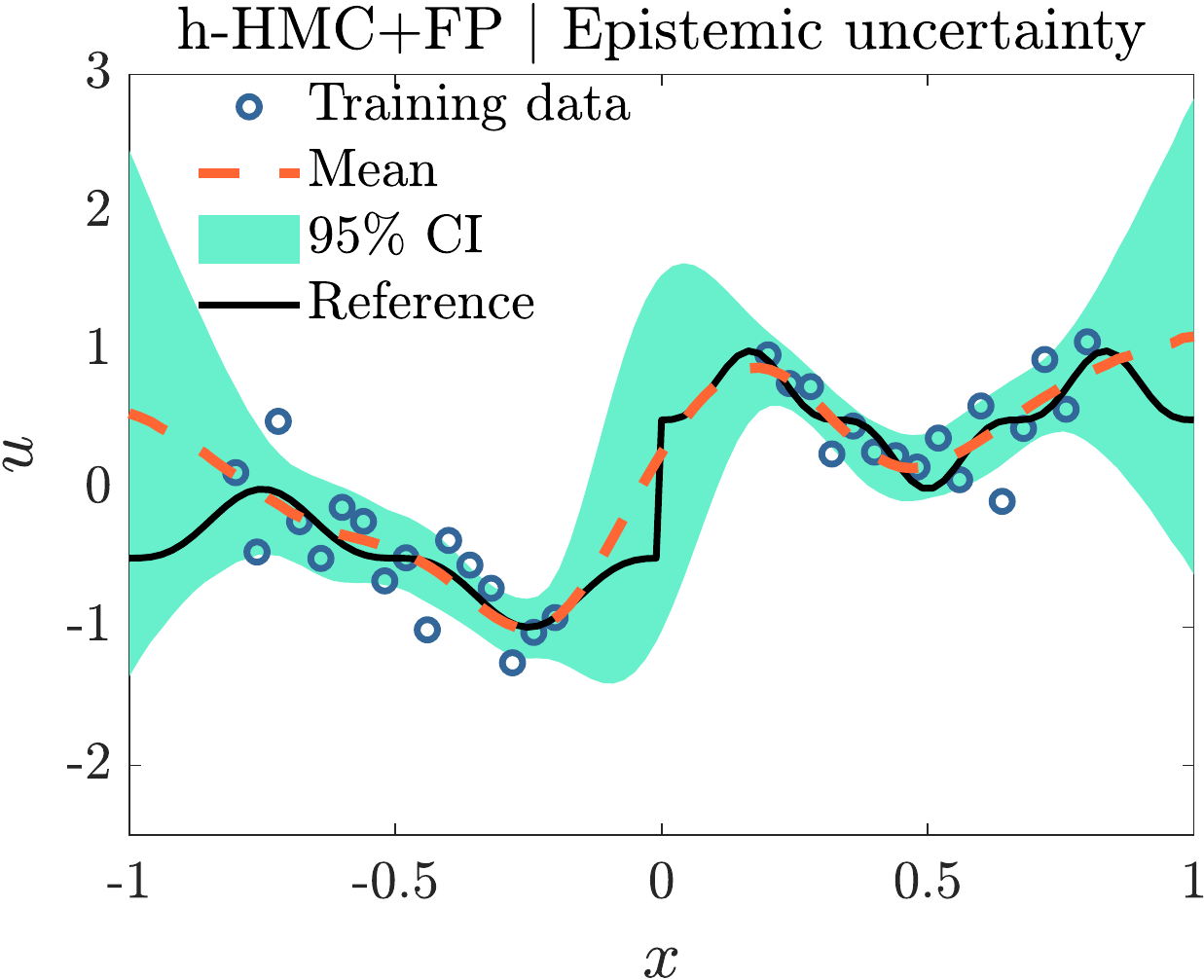}}
	\subcaptionbox{}{}{\includegraphics[width=0.24\textwidth]{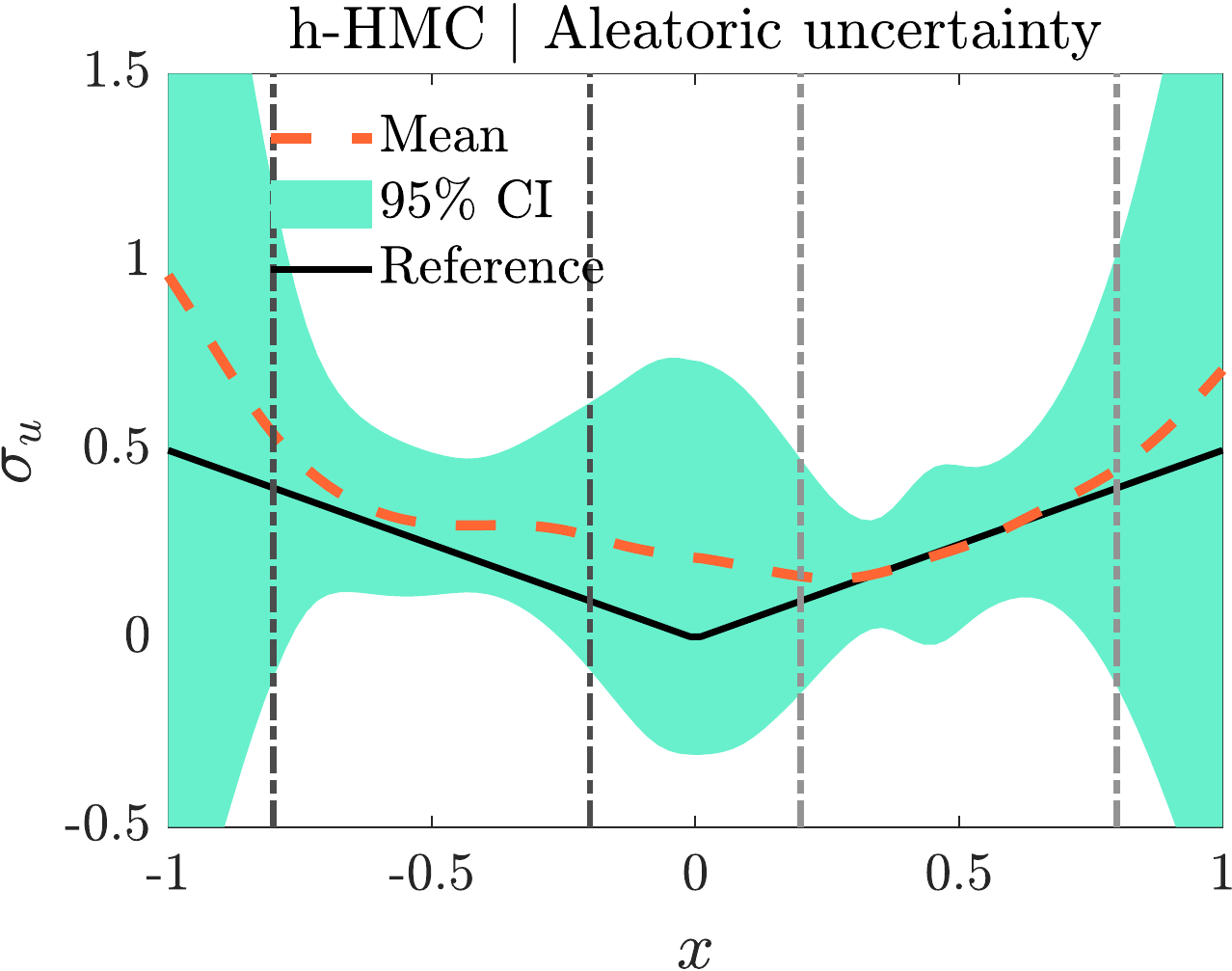}}
	\subcaptionbox{}{}{\includegraphics[width=0.24\textwidth]{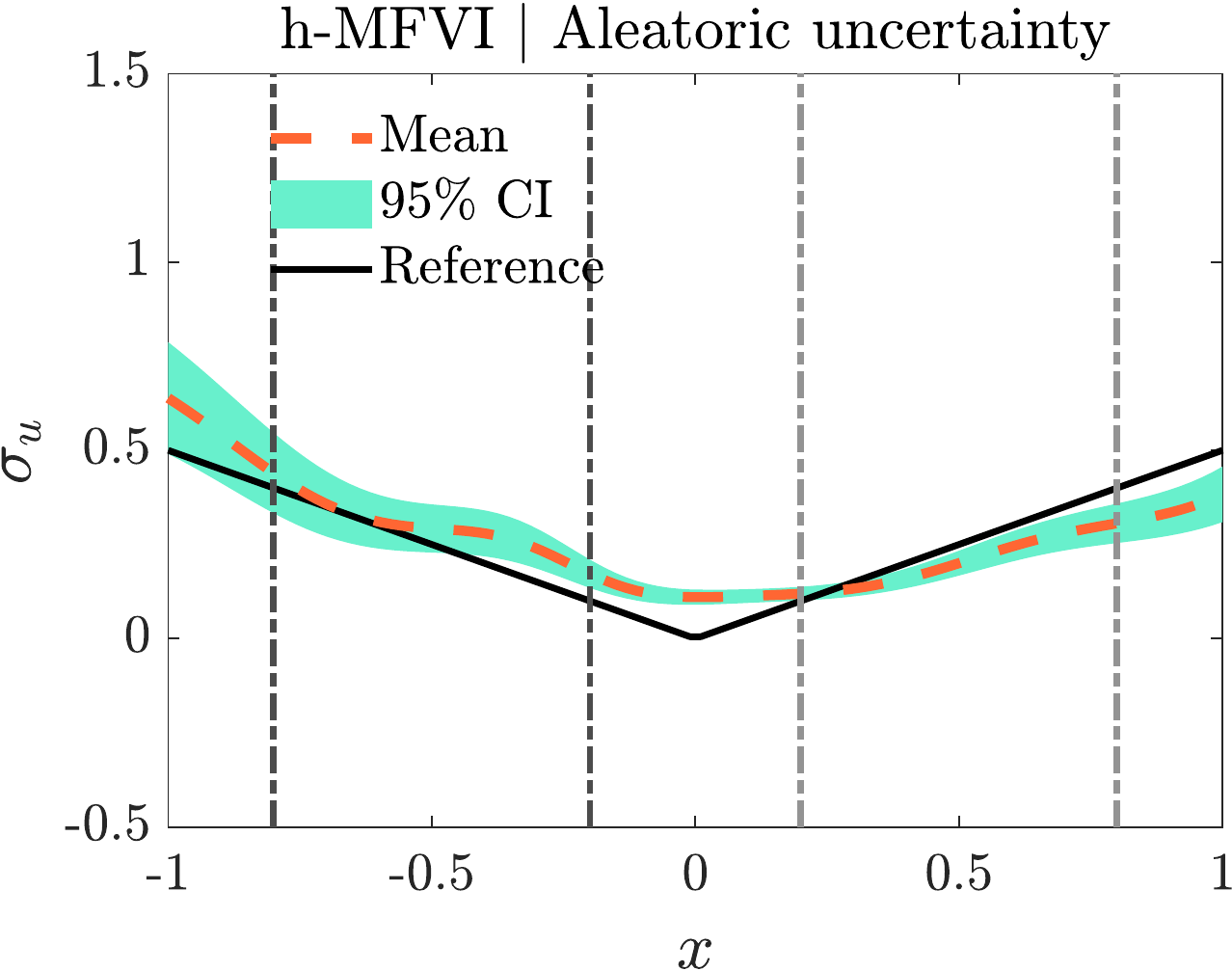}}
	\subcaptionbox{}{}{\includegraphics[width=0.24\textwidth]{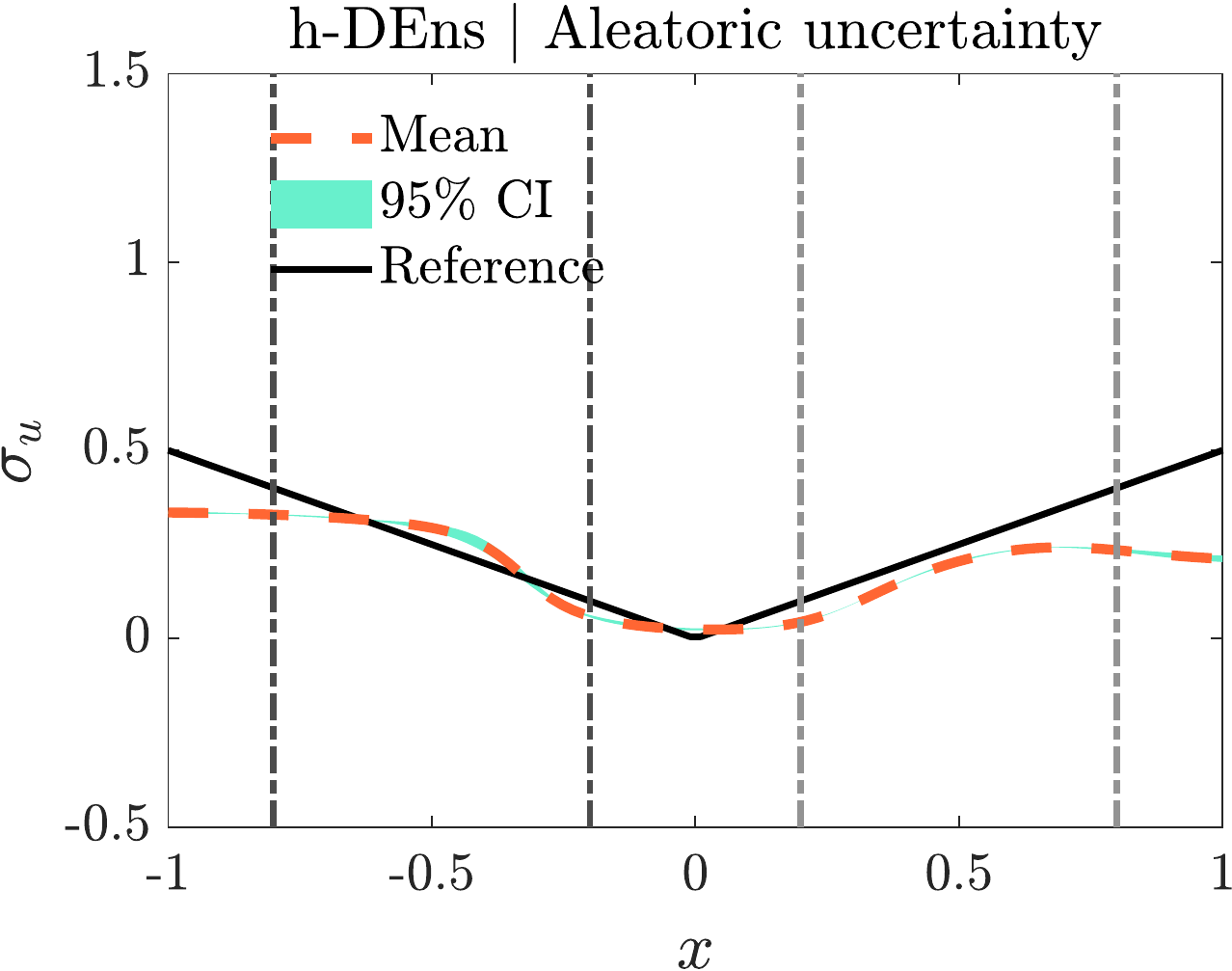}}
	\subcaptionbox{}{}{\includegraphics[width=0.24\textwidth]{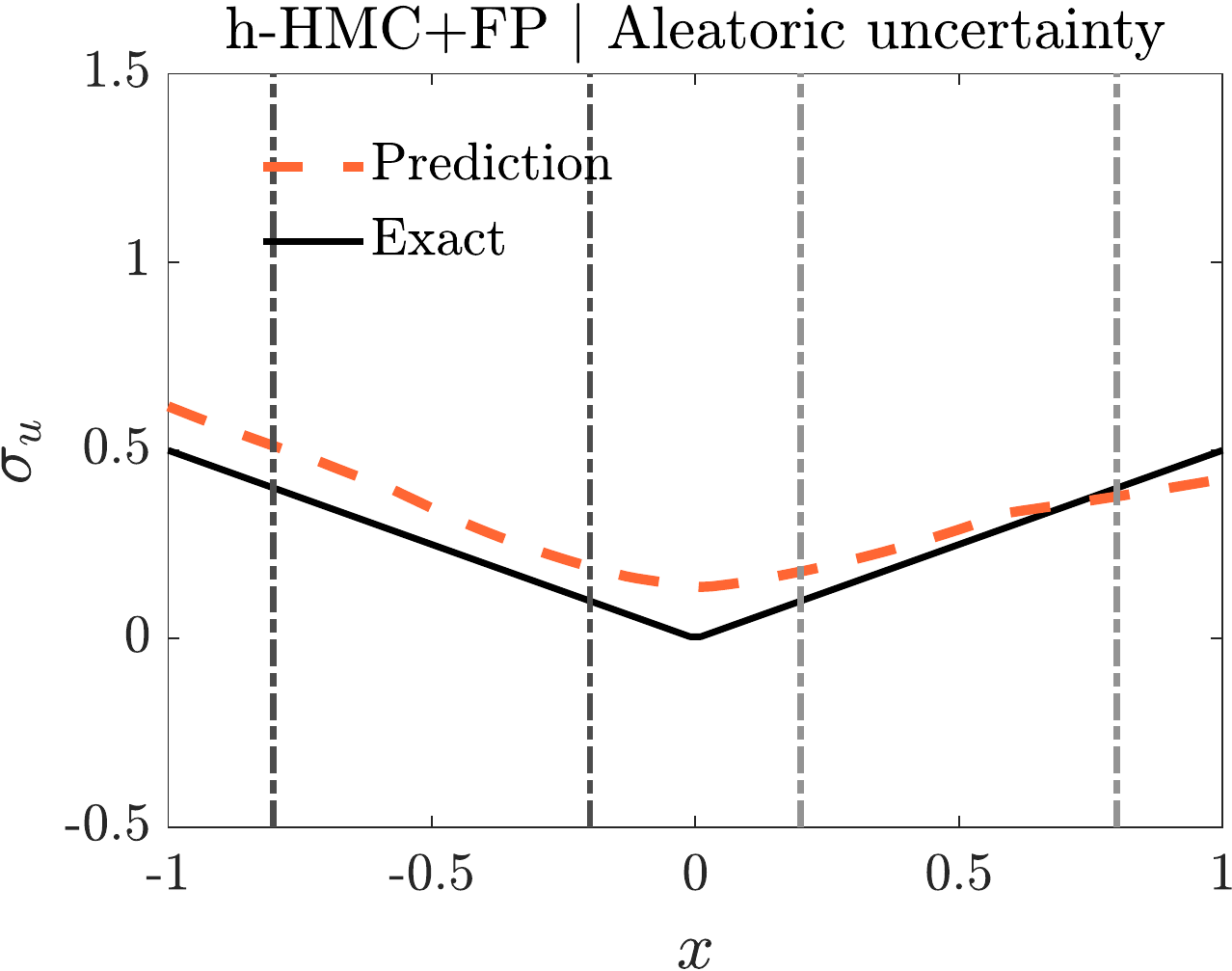}}
	\caption{
		Function approximation problem of Eq.~\eqref{eq:comp:func:func} | \textit{Unknown Gaussian heteroscedastic noise}:
		combining HMC with GAN using historical data to learn the heteroscedastic noise (h-HMC+FP) outperforms the UQ methods that use a BNN prior (h-HMC, h-MFVI, h-DEns).
		\textbf{Top row:} mean and total uncertainty of $u(x)$. 
		\textbf{Middle row:} mean and epistemic uncertainty of $u(x)$.    
		\textbf{Bottom row:}
		mean and epistemic uncertainty of predicted aleatoric uncertainty $\sigma_u(x)$.
		In the top two rows the training datapoints are also included, while in the bottom row the black and gray dot-dashed lines mark the training data intervals.
		Note that the aleatoric uncertainty predictions of h-HMC, h-MFVI, h-DEns are satisfactory, especially given the limited amount of data for such task.
		For example, note that the data on the left side happen to be more noisy and this is reflected in the noise predictions. 
		``h'' indicates that heteroscedastic noise modeling has been used.
	}
	\label{fig:comp:func:hetero:res}
\end{figure}

In Fig.~\ref{fig:comp:func:hetero:res} and Table~\ref{tab:comp:func:hetero}, we present the results obtained with h-HMC, h-MFVI, h-DEns, and h-HMC+FP. 
For ID data evaluation, it is shown that h-HMC+FP, which learns the heteroscedastic noise using a GAN-FP and historical data, outperforms the UQ methods that use a BNN-FP, in terms of accuracy of mean predictions (RL2E).
Further, h-DEns has the highest predictive capacity (MPL), while h-HMC is the most calibrated one (RMSCE).
Furthermore, the aleatoric uncertainty predictions of h-HMC, h-MFVI, h-DEns are satisfactory, especially given the limited
amount of data for such task.
For example, note that the data on the left side ``happen'' to be more noisy and this is reflected in the noise
predictions. 
Note that the epistemic uncertainty of DEns is quite small. In our experiments, we found that this uncertainty depends on the weight decay parameter. 
In this case, higher weight decay parameter values led to less epistemic uncertainty, but better predictions for the noise.
Finally, see also Section~\ref{app:comp:func:results:student} for computational results pertaining to data with heteroscedastic Student-t noise.

\subsubsection{Summary}

In this section, we presented a comparative study pertaining to function approximation. For constant aleatoric uncertainty along $x$ (homoscedastic noise) with known scale and Gaussian distribution, we compared nine methods, considered various noise scales and dataset sizes, and showed the effect of NN architecture on epistemic uncertainty. For homoscedastic noise with unknown scale, we compared three methods with learned priors, for approximating the function and learning the noise scale. For unknown aleatoric uncertainty with varying along $x$ (heteroscedastic noise) Gaussian distribution, we compared four methods, including a method involving a GAN-FP, for approximating the function and learning the heteroscedastic noise.
For Student-t heteroscedastic noise, we compared two methods for approximating the function and learning the noise using also posterior tempering.

Overall, we found that UQ methods with smaller computational cost and learned priors (e.g., MFVI and LA), can yield comparable performance with more expensive methods (e.g., HMC and DEns). 
Further, based on our experiments, post-training calibration using a small left-out dataset can reduce significantly the calibration error.
We also showed that learning the NN parameter prior during posterior inference helps to reduce the burden of thorough NN architecture selection.
An additional finding that has also been reported in other studies, e.g., \cite{yao2019quality,ashukha2021pitfalls}, is that performance evaluation and comparison of different UQ methods requires a plurality of evaluation metrics.
Further, we demonstrated that heteroscedastic noise modeling and learning with online or offline methods, e.g., functional priors, exhibits satisfactory accuracy. 

%% file: IN_mixed_PINN.tex
Next, we consider the deterministic 1D nonlinear time-dependent diffusion-reaction equation
\begin{subequations}\label{eq:comp:pinns:stand}
	\begin{align}
		\partial_t u = D\partial^2_x u - \lambda(x) u^3 + f(x), ~ t \in [0, 1], ~ x \in [-1, 1], \label{eq:comp:pinns:stand:pde}\\
		u(t, -1) = u(t, 1) = 0.5, \text{ and } u(0, x) = 0.5\cos^2(\pi x), \label{eq:comp:pinns:stand:bcs}\\
		\lambda(x) = 0.2 + \exp(x)\cos^2(2x), \label{eq:comp:pinns:stand:k}\\
		f(x) = \exp(-\frac{(x - 0.25)^2}{2l^2}) \sin^2(3 x), ~l = 0.4, \label{eq:comp:pinns:stand:f}
	\end{align}
\end{subequations}
where $t$ and $x$ are the time and space coordinates, respectively, $u(t, x)$ is the sought solution, $\lambda(x)$ and $f(x)$ are the space-dependent reaction rate and source term, respectively, and $D=0.01$ is the diffusion coefficient.
Eq.~\eqref{eq:comp:pinns:stand} can be viewed as a special case of Eq.~\eqref{eq:intro:piml:pinn:pde} with $\xi$ fixed and dropped; $\pazocal{F}$ and $\pazocal{B}$ given by Eqs.~\eqref{eq:comp:pinns:stand:pde} and \eqref{eq:comp:pinns:stand:bcs}, respectively, and with $x$ given by $(t, x)$.
The objective of this example is to compare different UQ methods for solving a mixed problem involving Eq.~\eqref{eq:comp:pinns:stand}, given noisy data of $f$, $u$, and $\lambda$. 
To this end, we consider the dataset of problem 2 in Table~\ref{tab:problem:form}, consisting of $N_f=15$ and $N_{\lambda}=6$ measurements of $f(x)$ and $\lambda(x)$, respectively, as well as $N_u=66$ measurements of $u(t, x)$ (11 locations along $x$ and 6 locations along $t$). 
For the data of $u$, the locations along $t$ are equidistant and the boundary/initial conditions of Eq.~\eqref{eq:comp:pinns:stand:bcs} are considered as noisy measurements of $u$ at the respective locations.
Further, all of $f$, $u$, and $\lambda$ data are contaminated with the same noise distribution, which is zero-mean Gaussian with different scales depending on the cases presented below.
We emphasize that the UQ methods we study in this paper do not distinguish between data of $u$ inside the domain and on the boundaries.
That is, the simple boundary/initial conditions of Eq.~\eqref{eq:comp:pinns:stand:bcs} are treated in exactly the same way as the rest of the noisy measurements of $u$.
As a result, these measurements can be noisy, limited, or even not available.
This flexibility enables these methods to be used in multiple contexts and problem scenarios. 
Clearly, however, the amount of available data and the locations in the domain affect the well-posedness of the problem, as well as the decision about what method should ultimately be employed.
Here, we use the {\it Matlab PDE toolbox} with a uniform grid $x \times t = 101  \times 101$ to obtain the reference solutions as well as the training data that we present in the following results.

\begin{table}[!ht]
	\centering
	\footnotesize
	\begin{tabular}{c|c|ccccc}
		\toprule
		\multicolumn{7}{c}{Standard case}\\
		\midrule
		\multirow{6}{*}{\textbf{a}}&Metric ($\times 10^2$) for {\color{Orange} $u(t, x)$}& HMC&MFVI&MCD&DEns&HMC+FP \\ 
		\cline{2-7}
		&RL2E ($\downarrow$)  & 3.7 & 6.1 & 13.5 & 4.1&\textbf{1.6} \\
		\cline{2-7}
		&Metric ($\times 10^2$) for {\color{Orange} $u(t=1, x)$}& HMC&MFVI&MCD&DEns&HMC+FP \\ 
		\cline{2-7}
		&RL2E ($\downarrow$)  & 2.5 & 4.2 & 6.8 & 2.8 & \textbf{1.9} \\ 
		&MPL ($\uparrow$) & 545.5 & 435.7 & 386.5 & \textbf{612.8} & 568.4 \\ 
		&RMSCE ($\downarrow$)  & \textbf{2.5} & 7.0 & 6.1 & 3.9 & 3.1 \\ 
		\midrule
		\multirow{4}{*}{\textbf{b}}&Metric ($\times 10^2$) for {\color{RoyalBlue} $f(x)$}& HMC&MFVI&MCD&DEns&HMC+FP \\ 
		\cline{2-7}
		&RL2E ($\downarrow$)  & 8.3 & 10.0 & 10.9 & 8.2 & \textbf{4.8} \\
		&MPL ($\uparrow$)  & 474.4 & 459.2 & 441.0 & \textbf{578.8} & 554.4 \\ 
		&RMSCE ($\downarrow$)  & 10.5 & 8.8 & 5.9 & 12.3 & \textbf{1.6} \\ 
		\midrule
		\multirow{4}{*}{\textbf{c}}&Metric ($\times 10^2$) for {\color{ForestGreen} $\lambda(x)$}& HMC&MFVI&MCD&DEns&HMC+FP\\ 
		\cline{2-7}
		&RL2E ($\downarrow$) & 17.3 & 15.3 & 19.5 & 14.8 & \textbf{2.6} \\
		&MPL ($\uparrow$) & 242.0 & 380.3 & 310.0 & 395.1 & \textbf{559.8} \\ 
		&RMSCE ($\downarrow$) & 9.7 & 8.1 & 11.0 & 11.1 & \textbf{2.5} \\ 
		\midrule
		\midrule
		\multirow{3}{*}{\textbf{d}}&Reference & \multicolumn{5}{c}{Learned noise (aleatoric uncertainty)}\\
		\cline{2-7}
		&\multirow{2}{*}{$\sigma_u = 0.05$} & HMC & MFVI & MCD & DEns&HMC+FP \\ 
		\cline{3-7}
		& & \textbf{0.051}&0.052&known&0.041&known \\
		\bottomrule
	\end{tabular}
	\caption{
		Mixed PDE problem of Eq.~\eqref{eq:comp:pinns:stand} | \textit{Standard case}:
		combining HMC with GAN using historical data to learn a FP for the reaction rate $\lambda(x)$ (HMC+FP) outperforms the UQ methods that use a BNN prior (HMC, MFVI, MCD, DEns) for learning $\lambda(x)$.
		Among HMC, MFVI, MCD, and DEns, the more expensive techniques (HMC and DEns) perform better.
		Here we evaluate the accuracy (RL2E) using clean test (reference) data, as well as the predictive capacity (MPL) and calibration error (RMSCE) using noisy test data, of the predictions for $u$ (part a), $f$ (part b), and $\lambda$ (part c). 
		We also evaluate the aleatoric uncertainty (noise) predictions (part c); ``known'' indicates that the noise was considered as known during training. 
	}
	\label{tab:pinns:standard}
\end{table}

\begin{table}[!ht]
	\centering
	\footnotesize
	\begin{tabular}{c|cccc}
		\toprule
		\multirow{2}{*}{RL2E ($\times 10^2$)} &  \multicolumn{4}{c}{Additional case solved with HMC} \\
		\cline{2-5}
		\multirow{2}{*}{of quantity ($\downarrow$)} & Heteroscedastic &Large noise &Extrapolation&Steep boundary layers \\
		& \S\ref{sec:comp:pinns:hetero} &\S\ref{app:comp:pinns:results:large}&\S\ref{app:comp:pinns:results:extra}&\S\ref{app:comp:pinns:results:steep} \\
		\midrule
		{\color{Orange} $u(t, x)$} & 4.7 & 7.6 & 3.9 & 4.2 \\
		{\color{Orange} $u(t=1, x)$} & 3.2 & 5.1 & 2.8 & 4.1 \\
		{\color{RoyalBlue} $f(x)$} & 9.0 & 15.1 & 11.1 & 40.3 \\ 
		{\color{ForestGreen} $\lambda(x)$} & 13.1 & 19.4 & 15.9 & 22.3 \\ 
		\bottomrule
	\end{tabular}
	\caption{
		Mixed PDE problem of Eq.~\eqref{eq:comp:pinns:stand} | \textit{Additional cases}:
		accuracy of the mean predictions (RL2E) deteriorates in the more challenging cases, as compared to the standard case of Table~\ref{tab:pinns:standard}. 
		However, using UQ we also obtain uncertainty estimates that cover in most cases the point-wise errors within the $95 \%$ CIs (approximately two standard deviations).
	}
	\label{tab:pinns:add}
\end{table}

We solve all problem cases described below using a U-PINN with different functional priors and posterior inference methods.
In Section~\ref{sec:comp:pinns:stand}, we solve the mixed problem, where the noise is homoscedastic (constant) with scales $\sigma_u = \sigma_f = \sigma_{\lambda} = 0.05$, with five methods of Table~\ref{tab:uqt:over}.
We refer to this case as ``standard'' because the additional considered cases have equivalent setups, except for various modifications that make them more challenging to treat.
Specifically, we consider a ``heteroscedastic'' case in Section~\ref{sec:comp:pinns:hetero} with space-dependent noise scale; a ``large noise'' case in Section~\ref{app:comp:pinns:results:large} with a larger noise scale compared to the standard case; an ``extrapolation'' case, where the measurements of $\lambda(x)$ are concentrated in the interval $x \in [0, 1]$ and we extrapolate for $x \in [-1, 0]$; and a ``steep boundary layers'' case in Section~\ref{app:comp:pinns:results:steep},  where the functions $u(t, x)$ and $\lambda(x)$ have large gradients for $x\approx -1$ and $x \approx 1$.
The hyperparameters and the NN architectures that we used are summarized in Section~\ref{app:comp:hyperparameters} and \ref{app:comp:architecture}, respectively.

\begin{figure}[!ht]
	\centering
	\subcaptionbox{}{}{\includegraphics[width=0.32\textwidth]{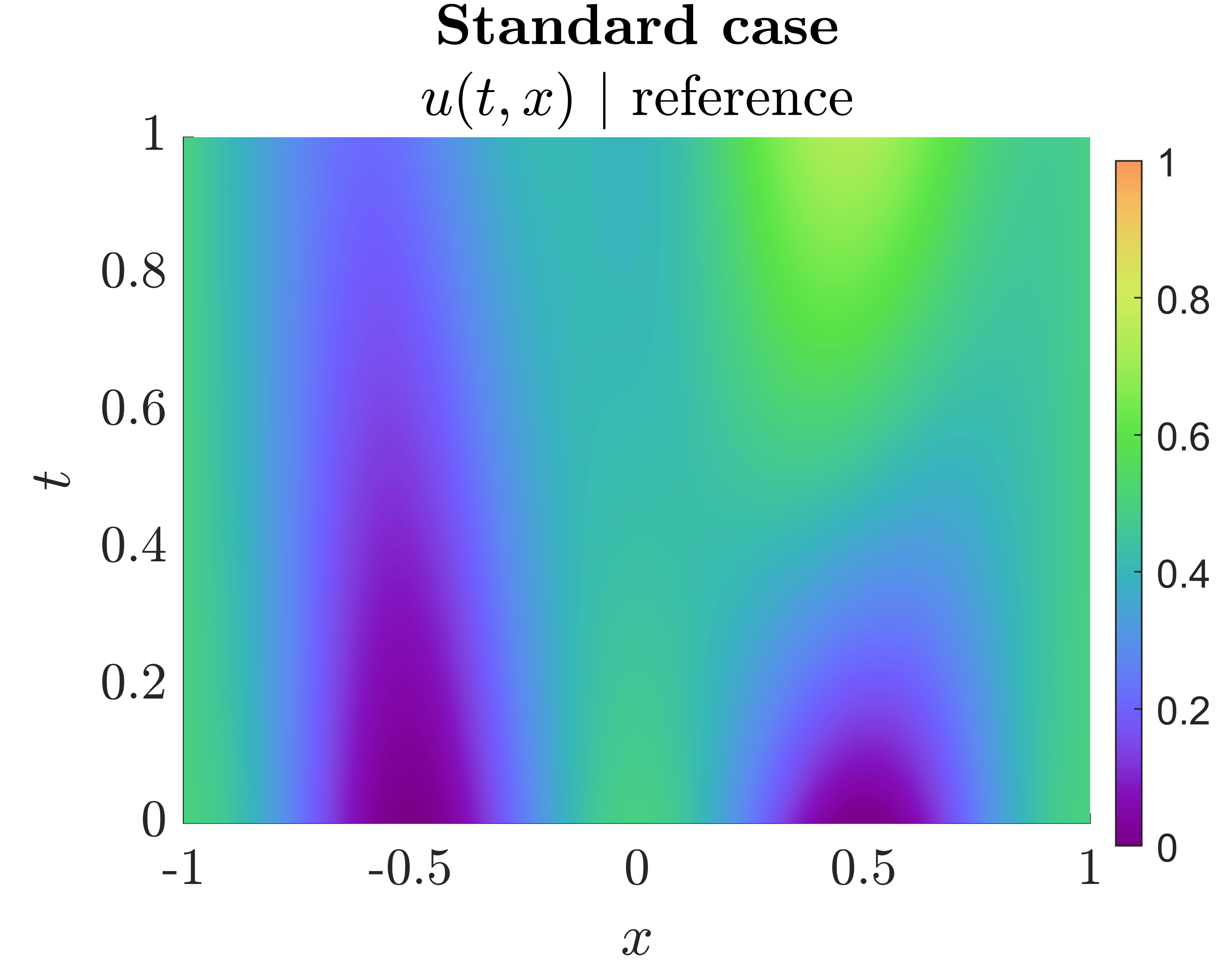}}
	\subcaptionbox{}{}{\includegraphics[width=0.32\textwidth]{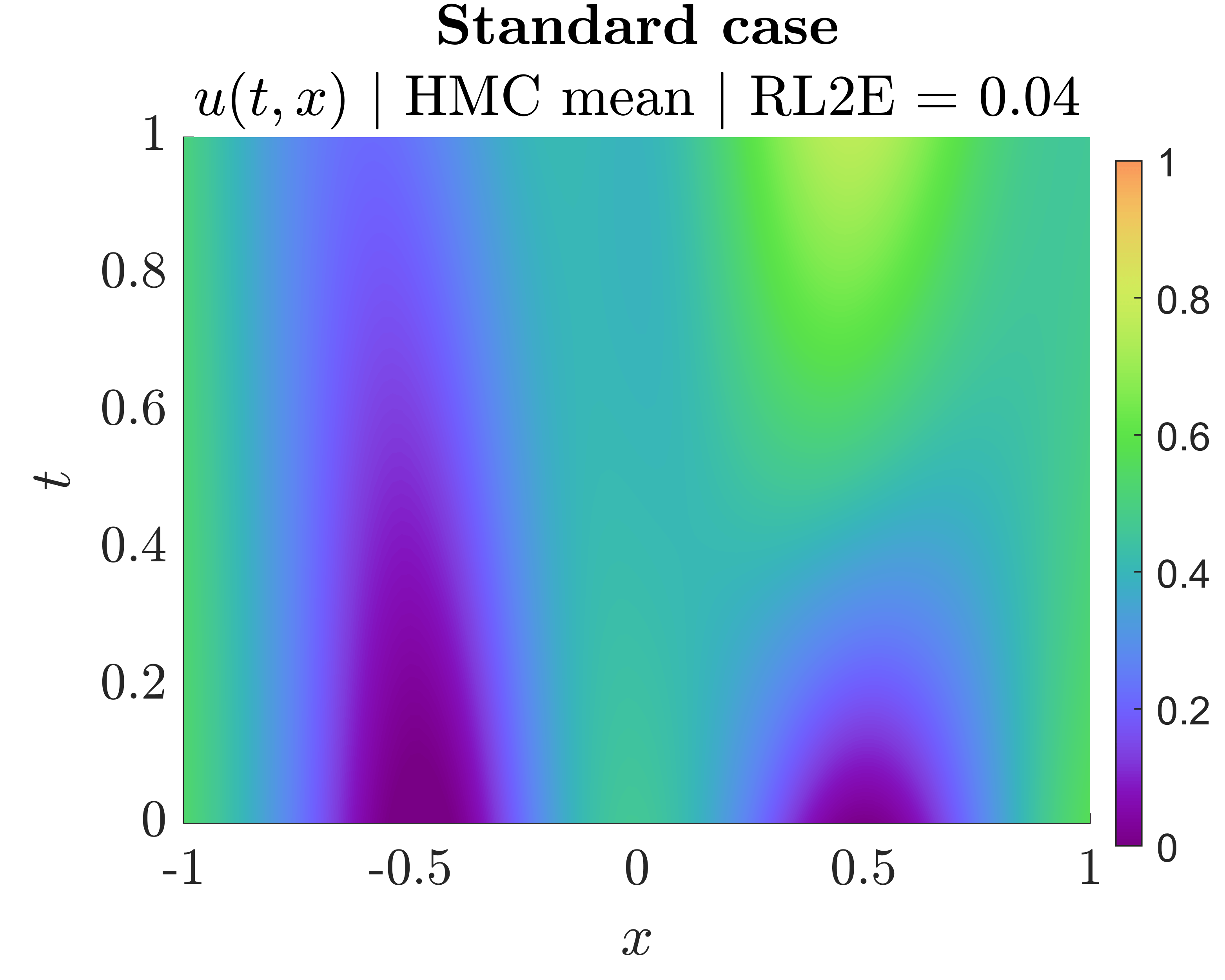}}
	\subcaptionbox{}{}{\includegraphics[width=0.32\textwidth]{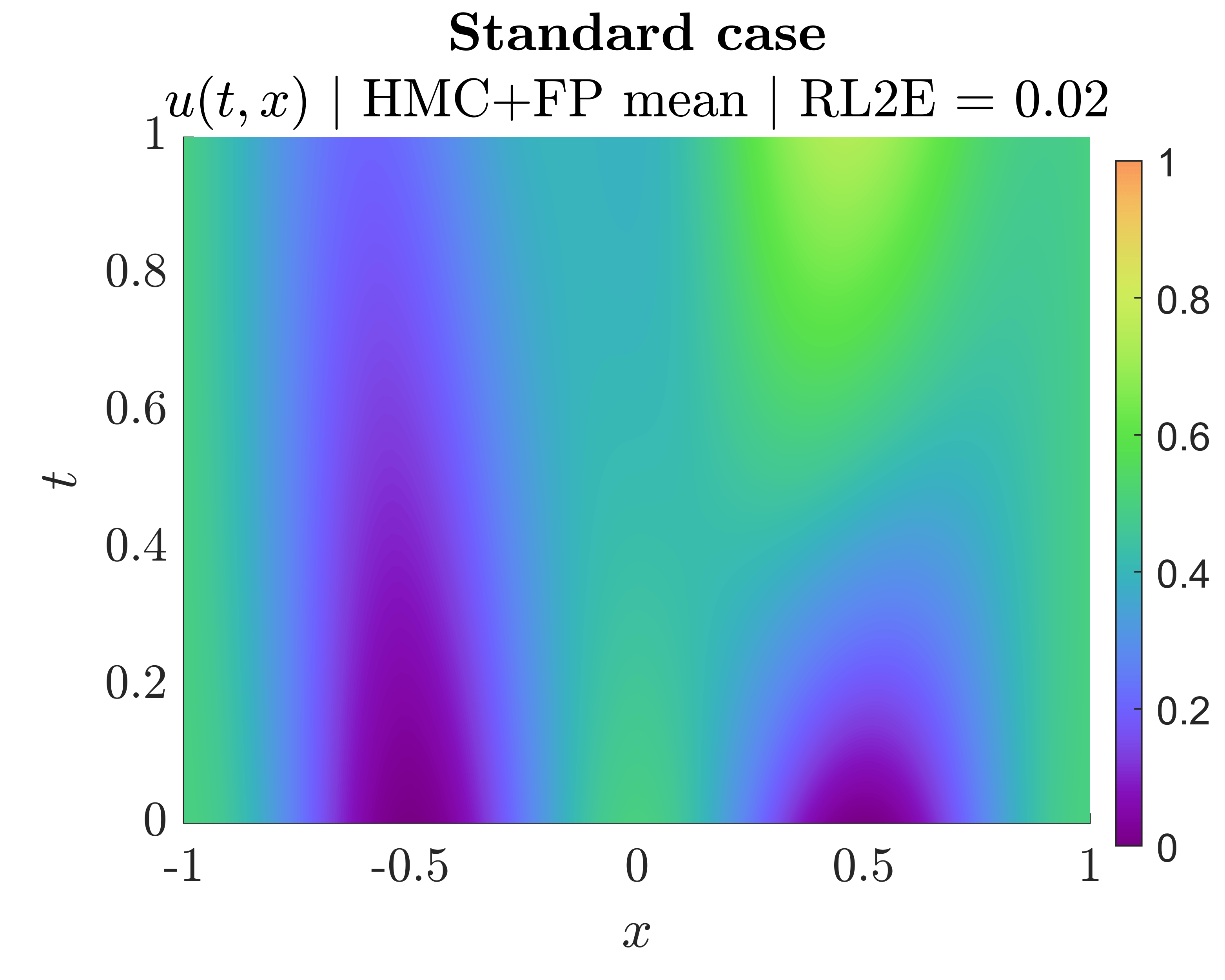}}
	\subcaptionbox{}{}{\includegraphics[width=0.24\textwidth]{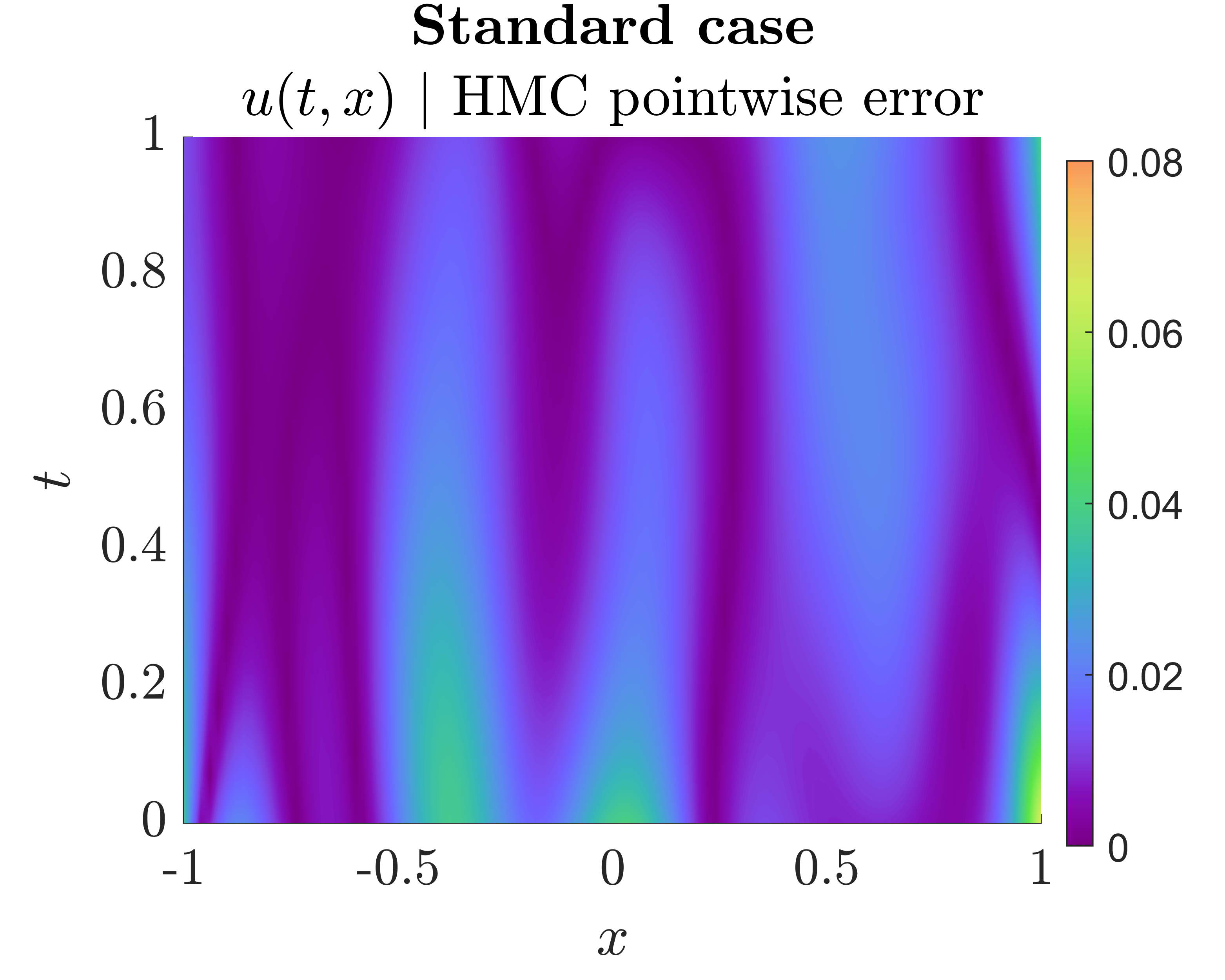}}
	\subcaptionbox{}{}{\includegraphics[width=0.24\textwidth]{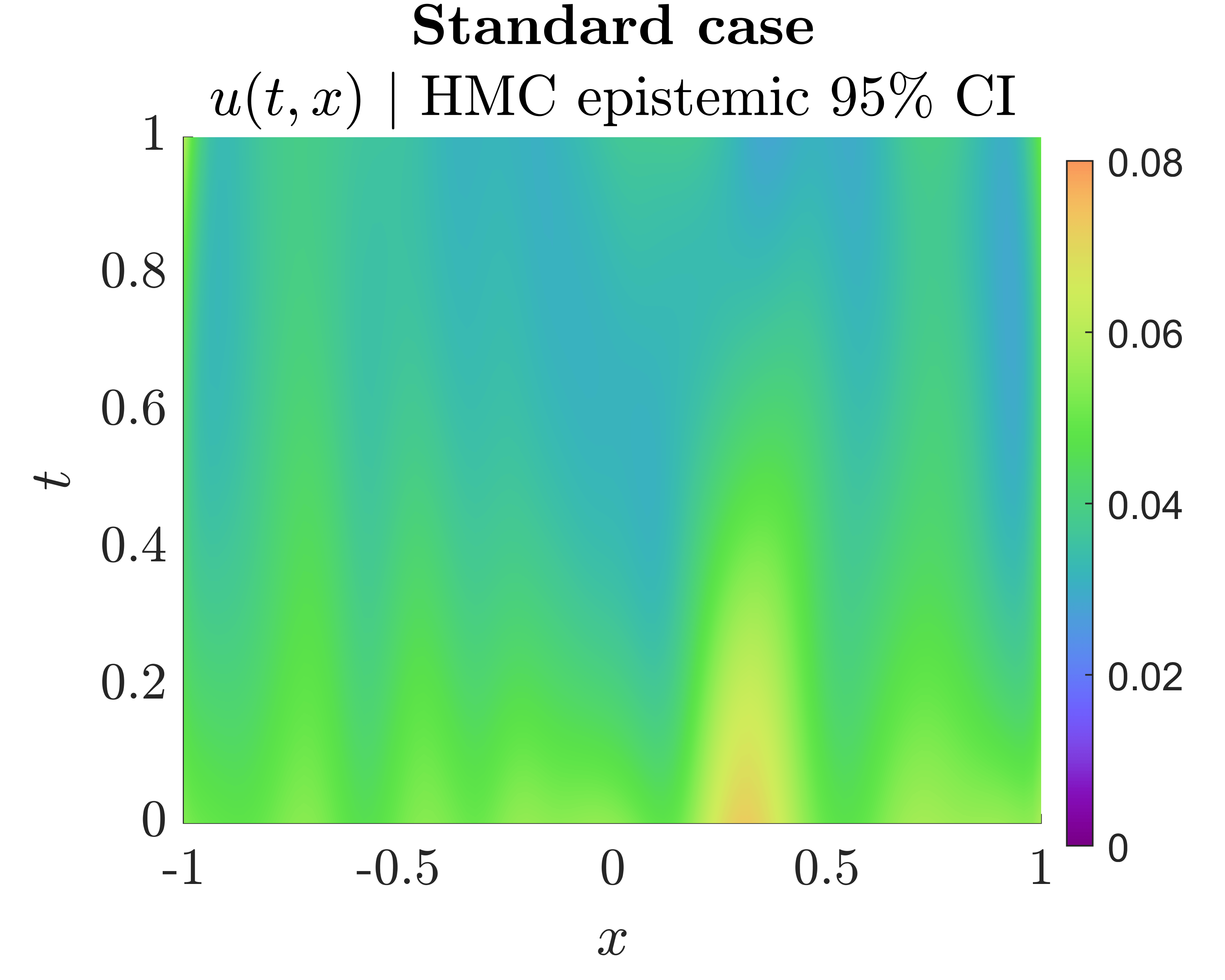}}
	\subcaptionbox{}{}{\includegraphics[width=0.24\textwidth]{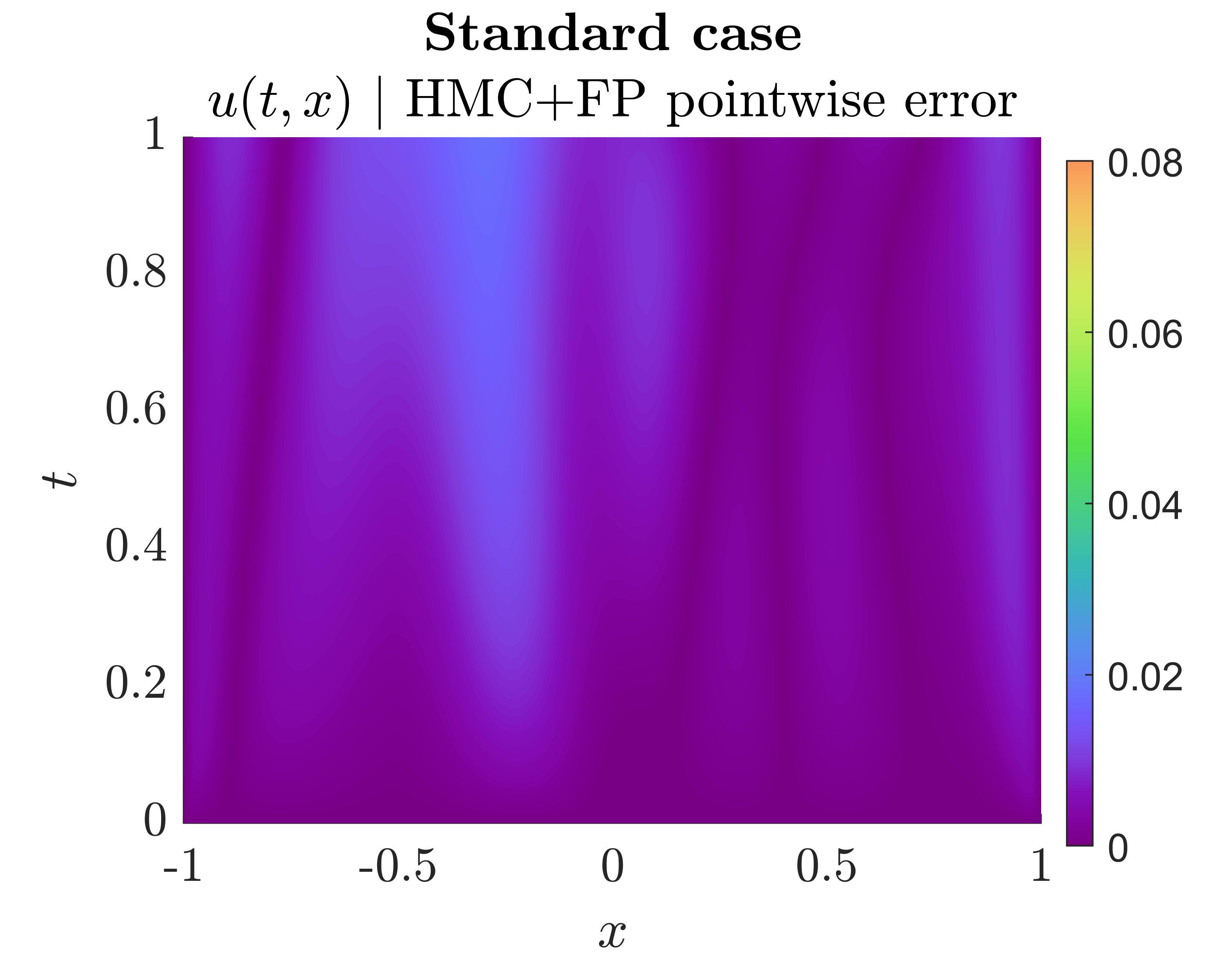}}
	\subcaptionbox{}{}{\includegraphics[width=0.24\textwidth]{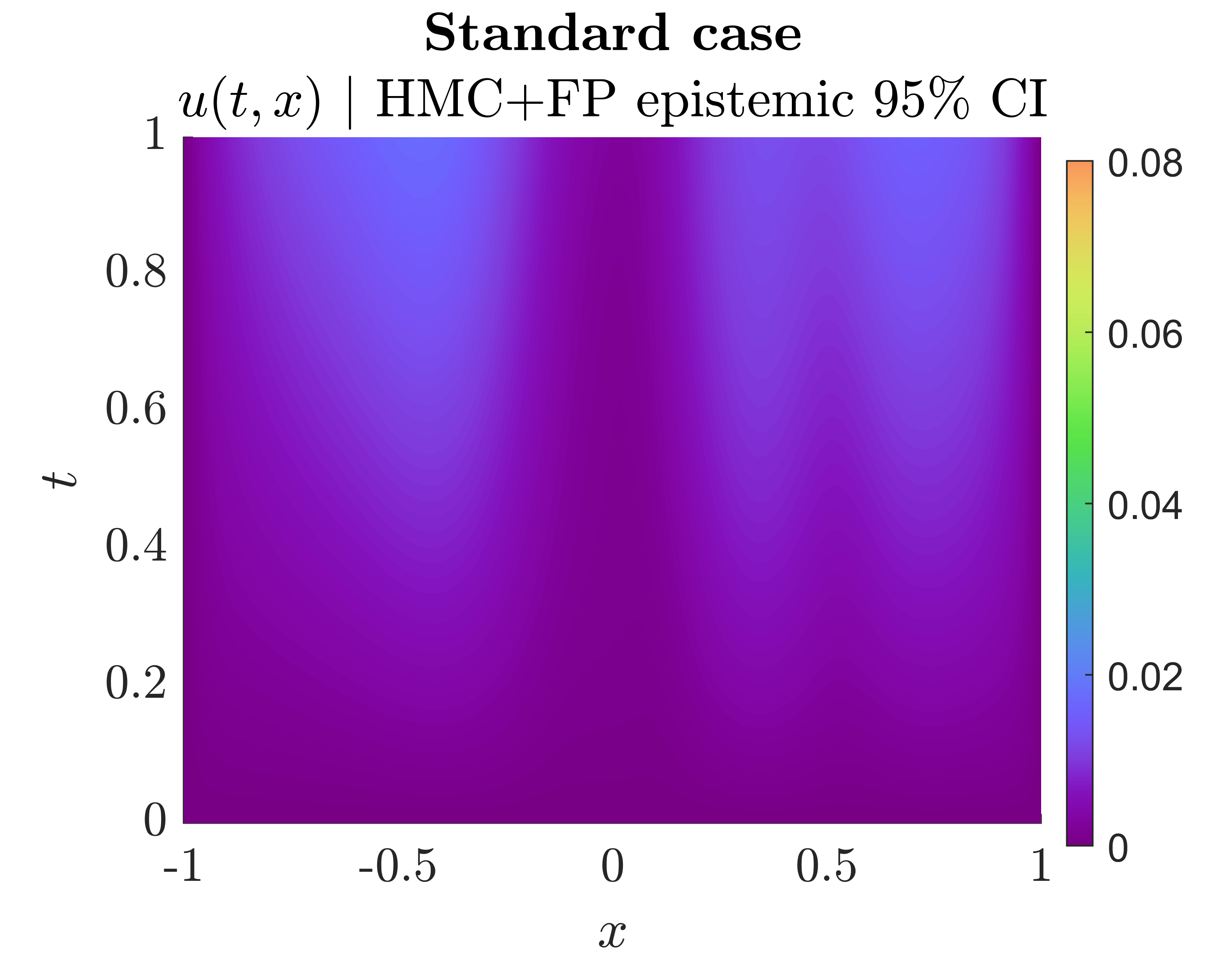}}
	\caption{
		Mixed PDE problem of Eq.~\eqref{eq:comp:pinns:stand} | \textit{Standard case}:
		for predicting $u(t, x)$, HMC (BNN prior) and HMC+FP (GAN prior) yield similar mean predictions, whereas HMC is less confident than HMC+FP (larger epistemic uncertainty).
		In both cases, however, the absolute error for each $x$ is within the $95 \%$ confidence interval (CI) of total (epistemic + aleatoric) uncertainty.
		\textbf{Top row:} reference solution $u(t, x)$, and mean prediction for $u(t, x)$ obtained by HMC and HMC+FP. 
		\textbf{Bottom row:} point-wise absolute error and epistemic uncertainty obtained by HMC and HMC+FP.
	}
	\label{fig:comp:pinns:stand:2d}
\end{figure}

\subsubsection{Standard case: homoscedastic noise and uniform measurements in space and time }\label{sec:comp:pinns:stand}

In this section, we solve the mixed PDE problem of Eq.~\eqref{eq:comp:pinns:stand} with data noise scales equal to $\sigma_u=\sigma_f=\sigma_{\lambda} = 0.05$.
For distributing uniformly the measurements along $x$, we have considered equidistant points for each of $u$, $f$ and $\lambda$, containing small perturbations in the respective coordinate $x$ values.
The objective of this section is to demonstrate and compare the capabilities of five UQ methods of Table~\ref{tab:uqt:over} in a problem where the noise scale and the NN parameter prior are unknown.
In this regard, we consider two NN approximators, one for $u$ and one for $\lambda$, as in Fig.~\ref{fig:uqt:bnns:dataunc:heteroscedastic:upinn} (without the outputs for the noise).
The outputs of the two NNs are substituted into Eq.~\eqref{eq:comp:pinns:stand:pde} for producing the approximated $f$.
In U-PINN with HMC, we learn the prior and the noise scale using Gibbs sampling (Section~\ref{app:methods:bnns:mcmc:hypers}).
In U-PINN with MFVI, we fix the prior, learn the mean of posterior Gaussian distribution with its standard deviation fixed and tuned, and learn the noise scale by optimization.
In U-PINN with MCD, we tune the dropout rate and assume known noise scale; for unknown noise scale the results were not meaningful.
In U-PINN with DEns, we tune the weight decay constant and learn the noise scale.
In U-PINN with HMC+FP, i.e., PI-GAN-FP of Table~\ref{tab:uqt:over}, we first pre-train a PI-GAN with synthetic historical data of $\lambda(x)$ and $f(x)$. In particular, we generate the historical data using 
\begin{align}
	\lambda(x) &= 0.2 + \exp(r_{\lambda,1} x) \cos^2(r_{\lambda,2} x), ~r_{\lambda, 1} = \frac{1}{2}(r_{k, 1} + 1), ~ r_{\lambda, 2} = r_{k, 2} + 3, \\
	r_{k, i} &\sim \pazocal{U}[0, 1], ~ i = 1,2, \\
	f(x) &= \exp(-\frac{(x - 0.25)^2}{2l^2}) \sin^2(r_{f} x), ~ l = 0.2 + 0.8 r_{f, 1}, ~ r_f = 1 + 3r_{f, 2}, \\
	r_{f, i} &\sim \pazocal{U} [0, 1], ~ i = 1,2,
\end{align}
and use 1,000 samples for $\lambda$ and $f$ in the training of PI-GANs.
Subsequently, we use the PI-GAN-FP to perform HMC posterior inference for $\theta$, which is the random input of the GAN generator.

Table~\ref{tab:pinns:standard} summarizes the performance evaluation of the considered UQ methods using the metrics of Section~\ref{sec:eval}. 
As expected, HMC+FP outperforms the rest of the methods in most comparisons. Among the methods that use a BNN prior, i.e., HMC, MFVI, MCD, and DEns, the more expensive techniques (HMC and DEns) perform better.

\newpage

\begin{figure}[!ht]
	\centering
	\subcaptionbox{}{}{\includegraphics[width=0.32\textwidth]{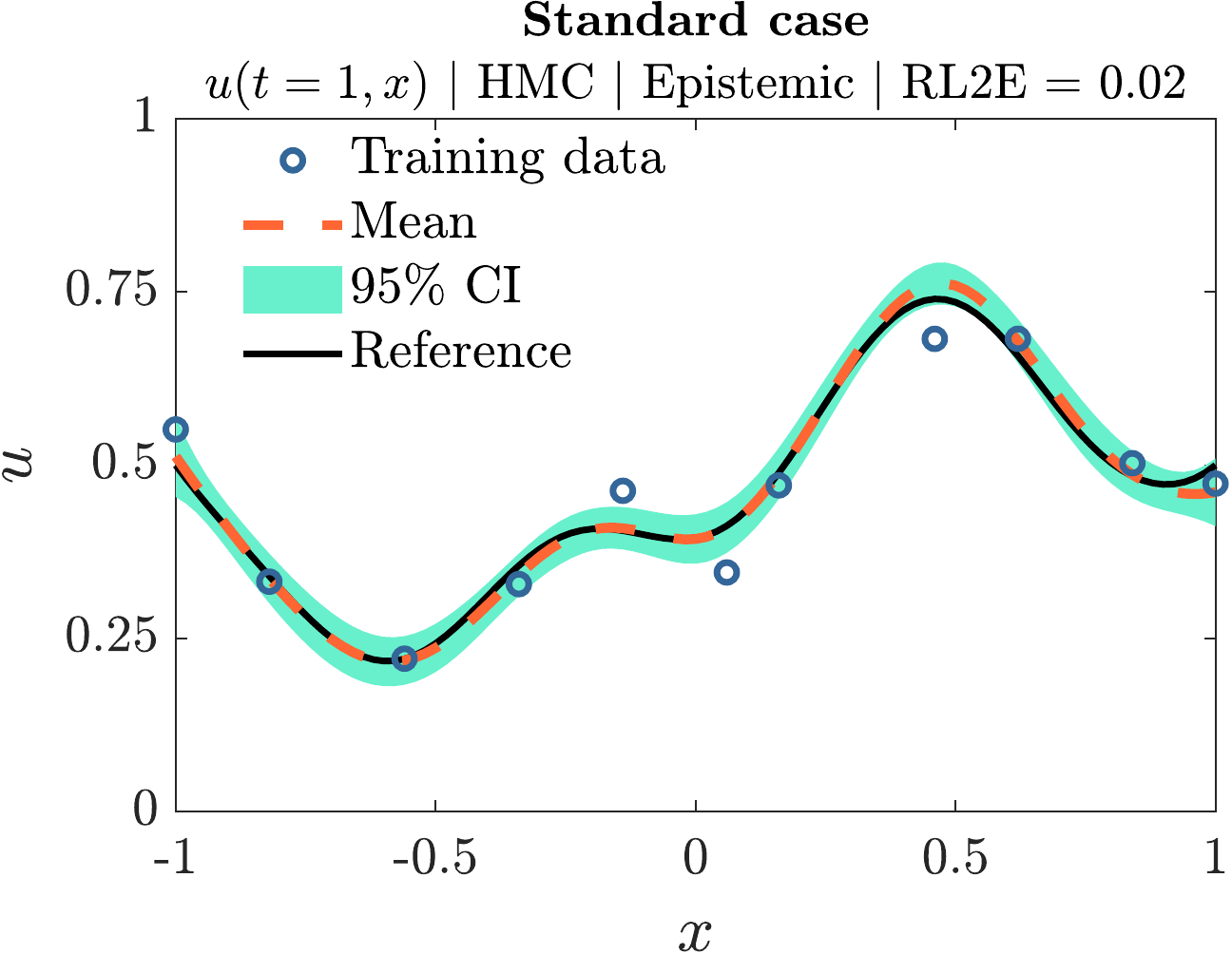}}
	\subcaptionbox{}{}{\includegraphics[width=0.32\textwidth]{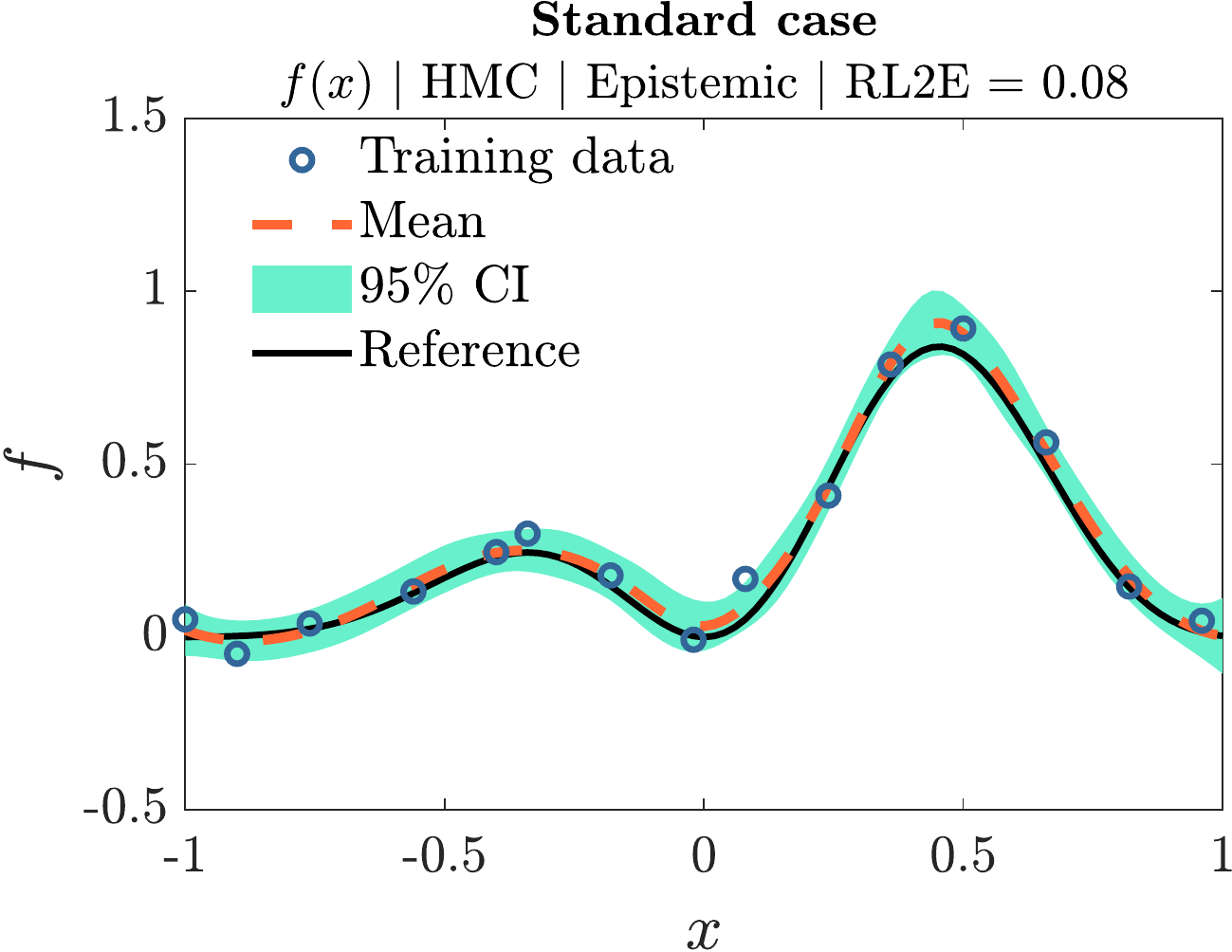}}
	\subcaptionbox{}{}{\includegraphics[width=0.32\textwidth]{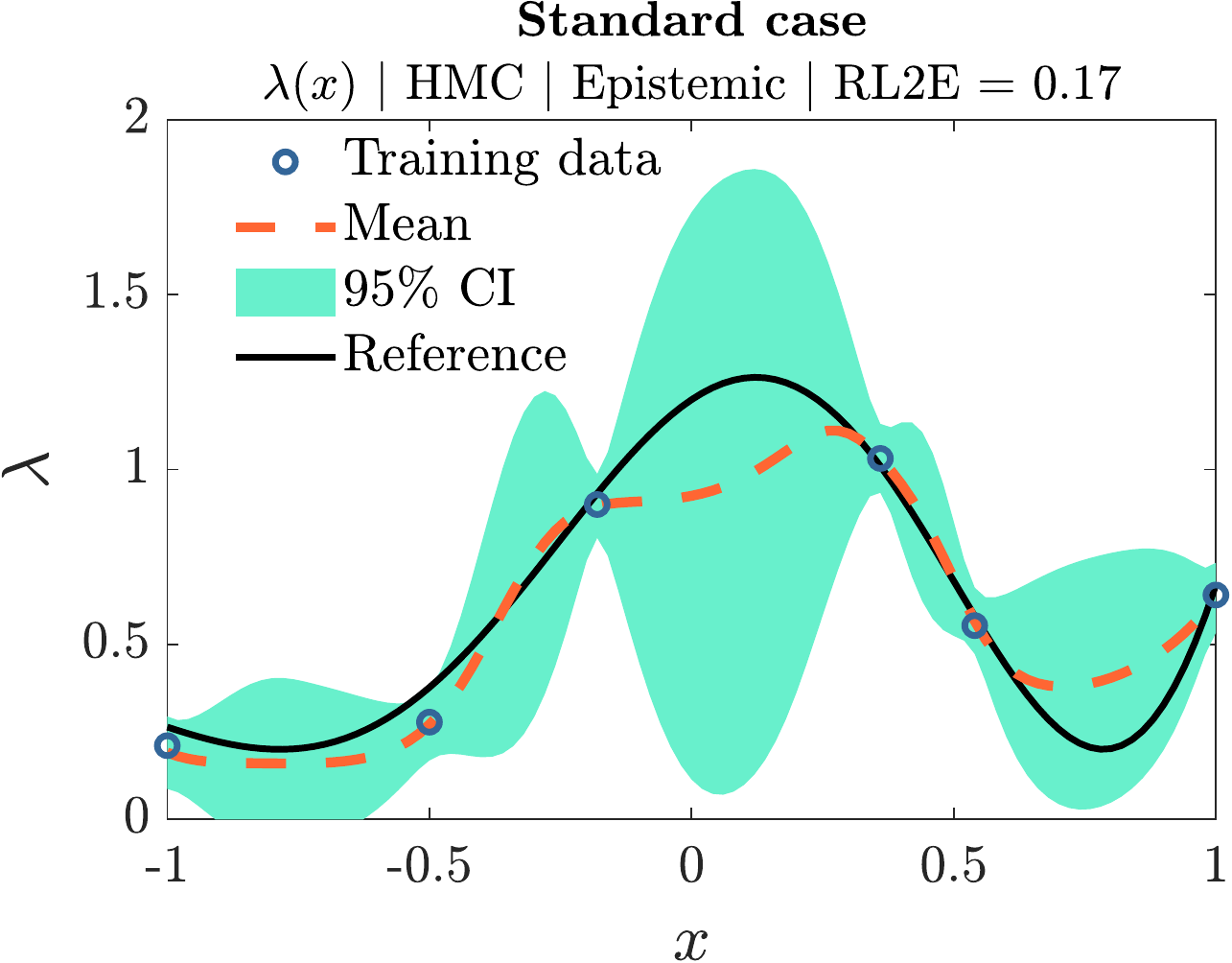}}
	\subcaptionbox{}{}{\includegraphics[width=0.32\textwidth]{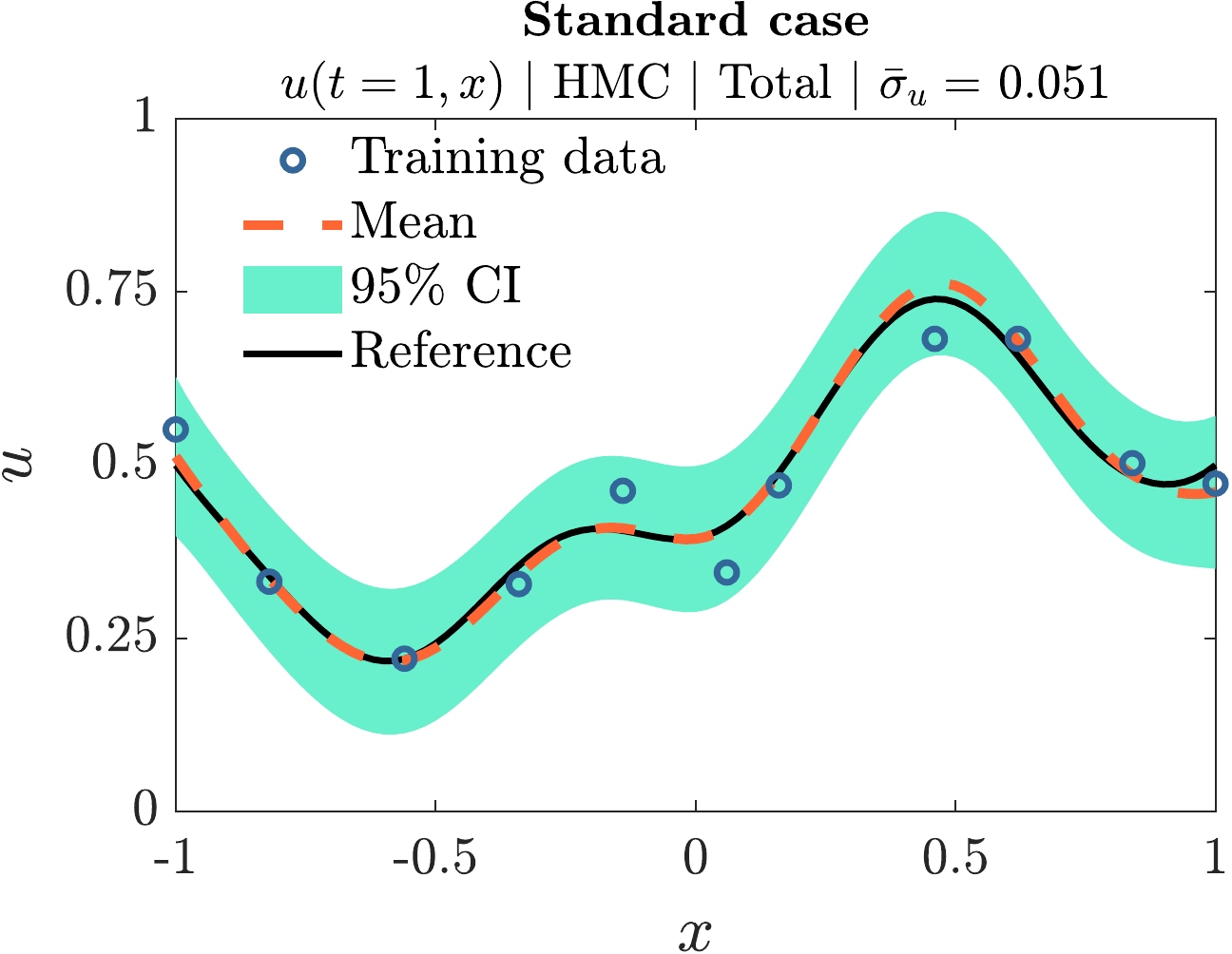}}
	\subcaptionbox{}{}{\includegraphics[width=0.32\textwidth]{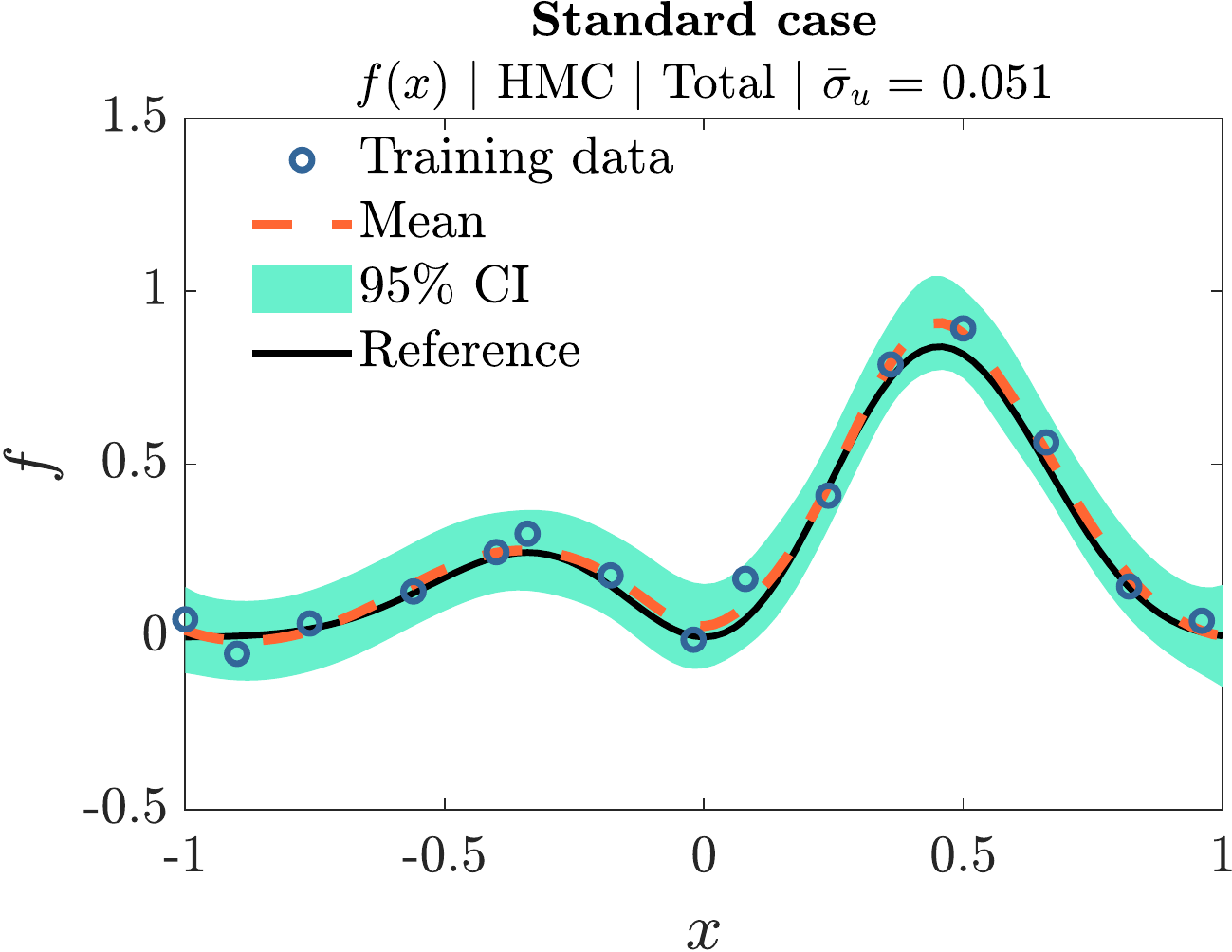}}
	\subcaptionbox{}{}{\includegraphics[width=0.32\textwidth]{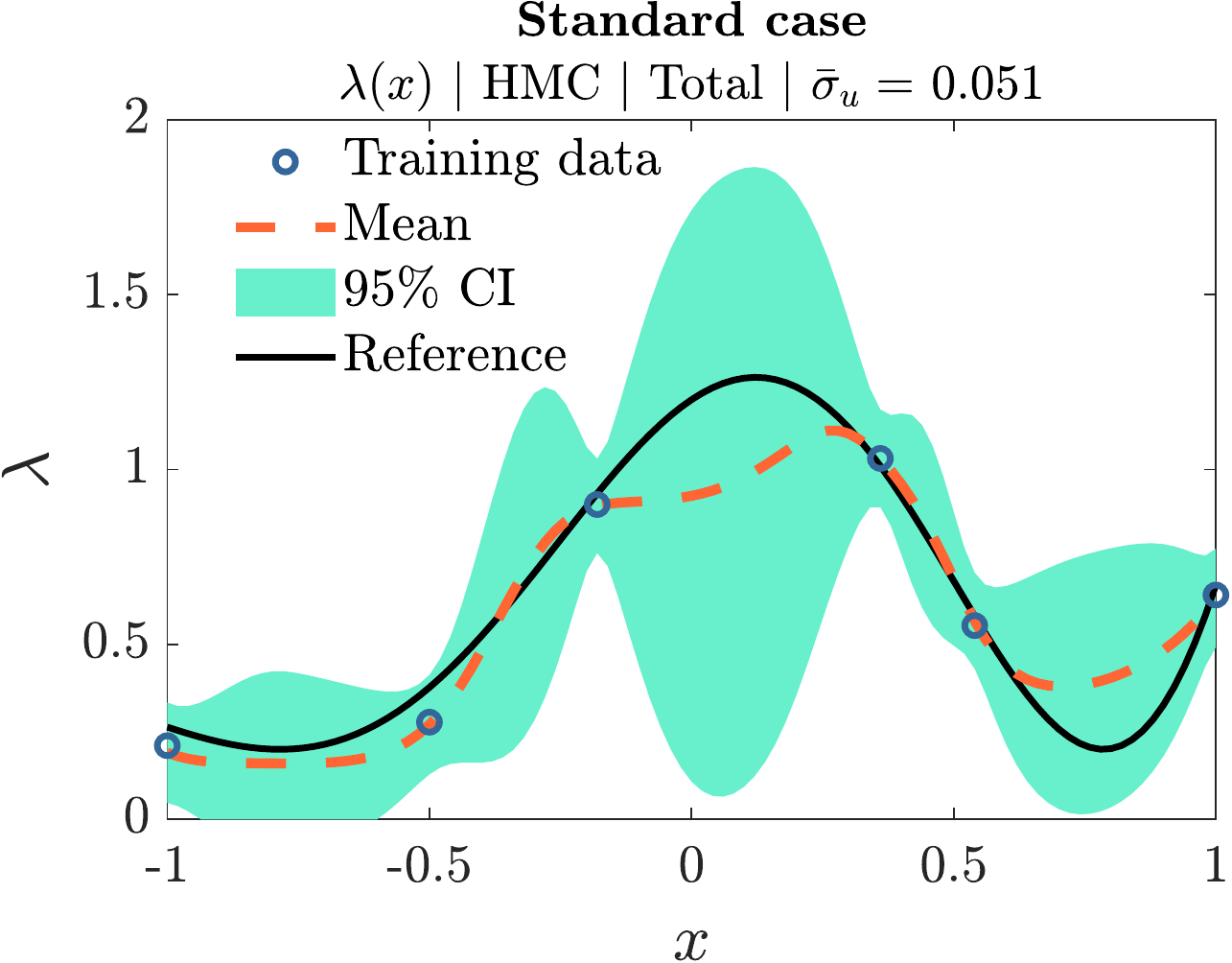}}
	\caption{
		Mixed PDE problem of Eq.~\eqref{eq:comp:pinns:stand} | \textit{Standard case}:
		epistemic uncertainty of HMC increases with more noise and with less amount of data, and is affected by the interdependence of $u$, $f$, and $\lambda$ through Eq.~\eqref{eq:comp:pinns:stand:pde}.
		Shown here are the training data, reference functions, as well as the mean and uncertainty ($95\%$ CI) predictions of HMC for $u$, $f$, and $\lambda$.
		We learn the prior via Gibbs sampling; see Section~\ref{sec:uqt:bnns}.
		\textbf{Top row:} epistemic uncertainty.
		\textbf{Bottom row:} total uncertainty including the \textit{learned} amount of aleatoric uncertainty, $\bar{\sigma}_u = 0.051$.
	}
	\label{fig:comp:pinns:stand:hmc}
\end{figure}

\begin{figure}[!ht]
	\centering
	\subcaptionbox{}{}{\includegraphics[width=0.32\textwidth]{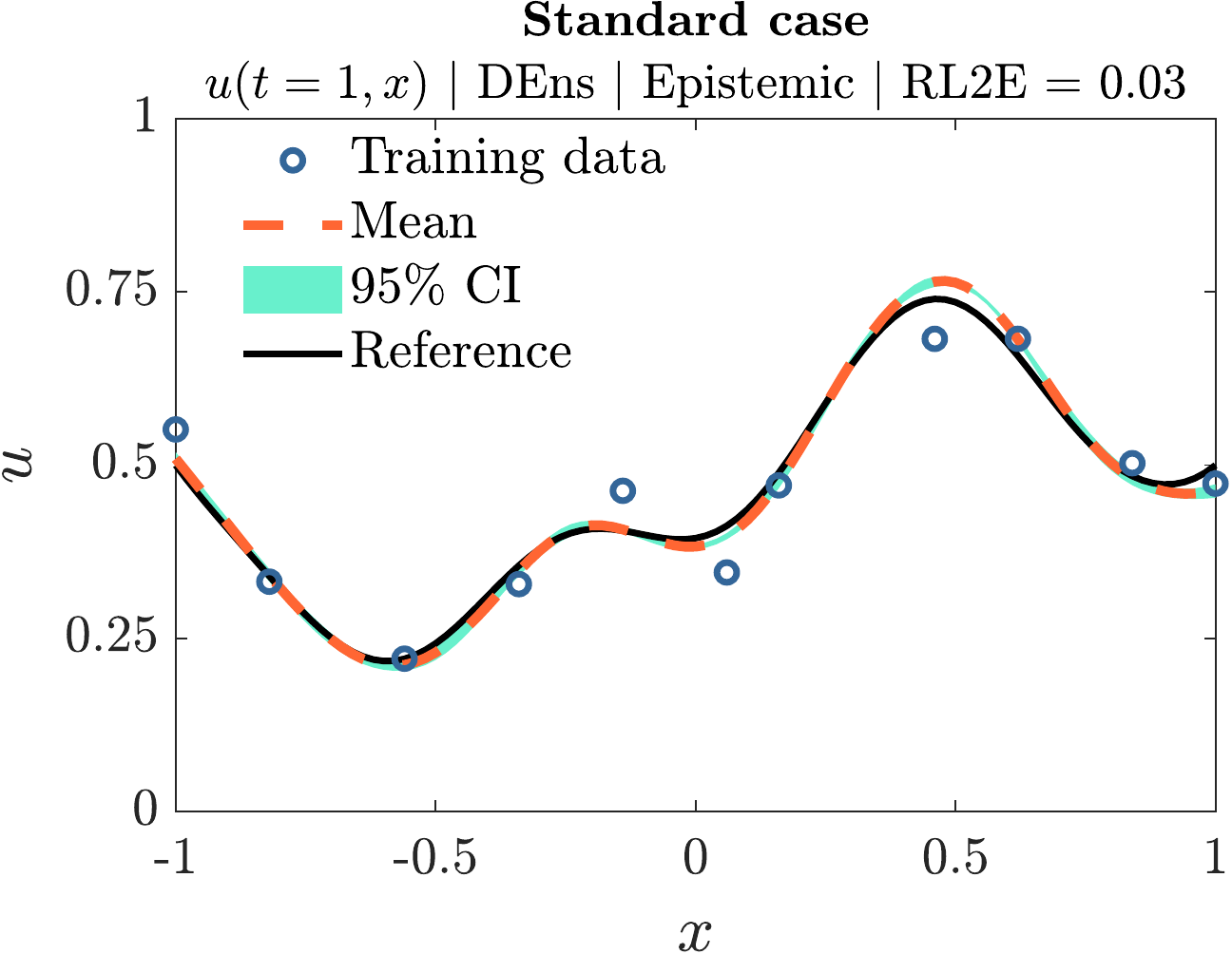}}
	\subcaptionbox{}{}{\includegraphics[width=0.32\textwidth]{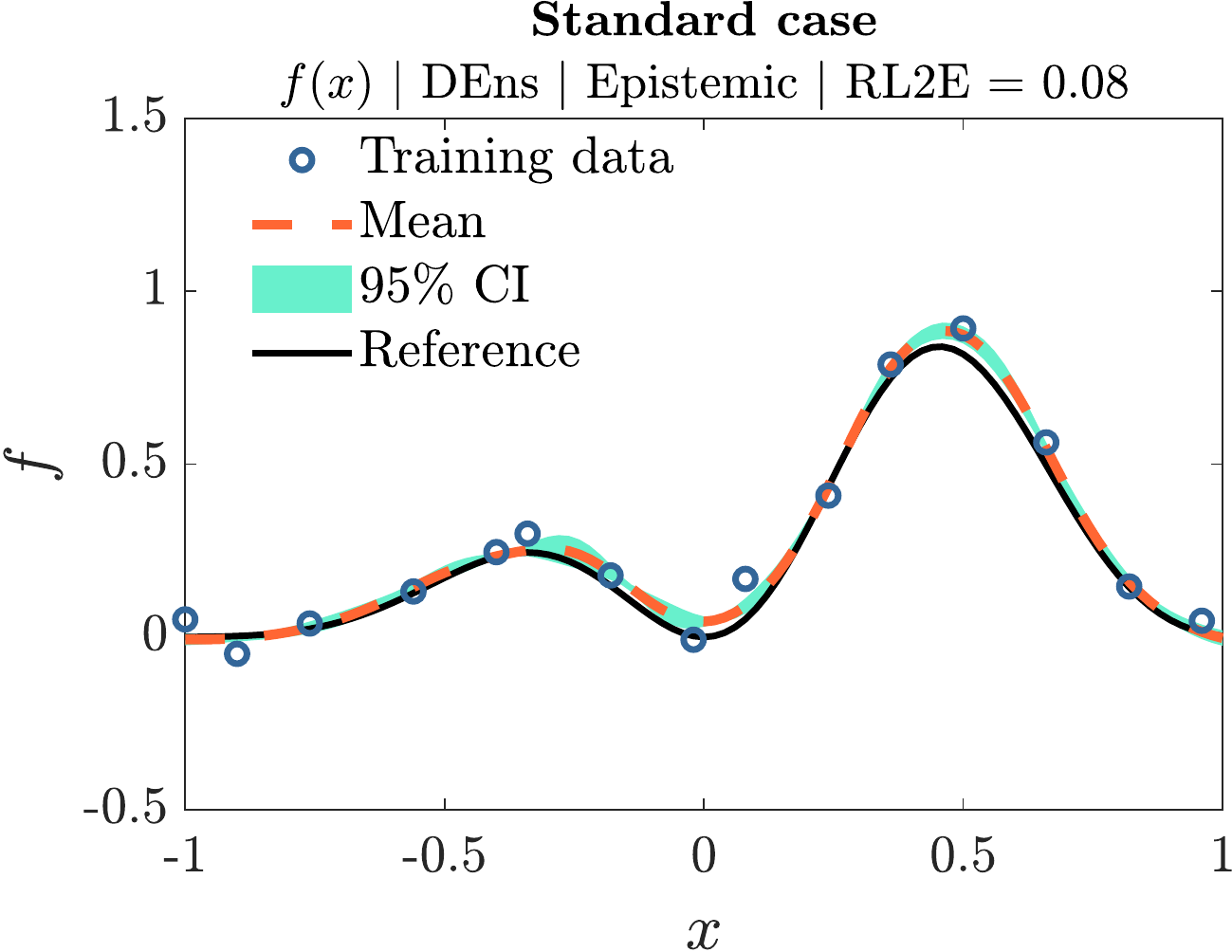}}
	\subcaptionbox{}{}{\includegraphics[width=0.32\textwidth]{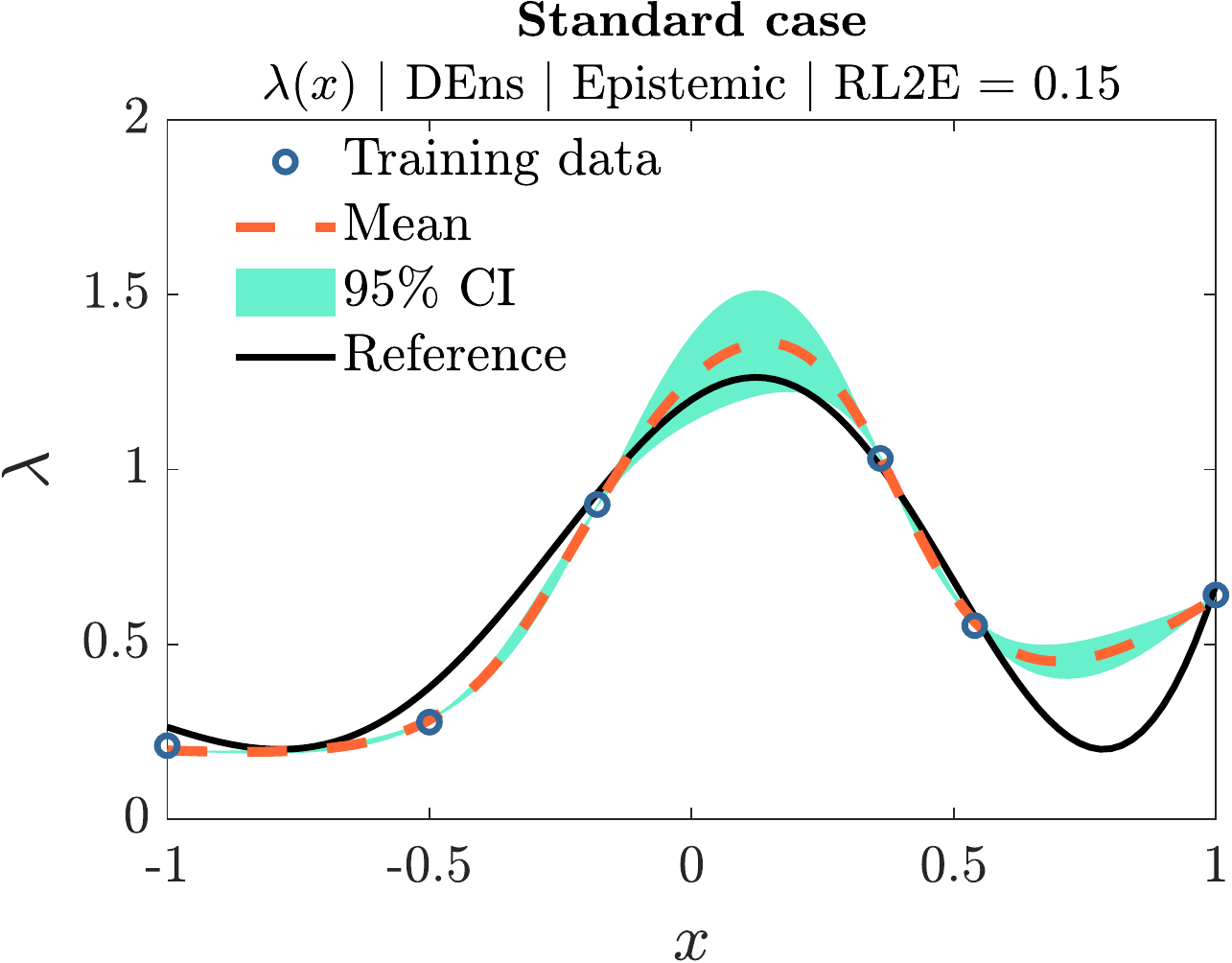}}
	\subcaptionbox{}{}{\includegraphics[width=0.32\textwidth]{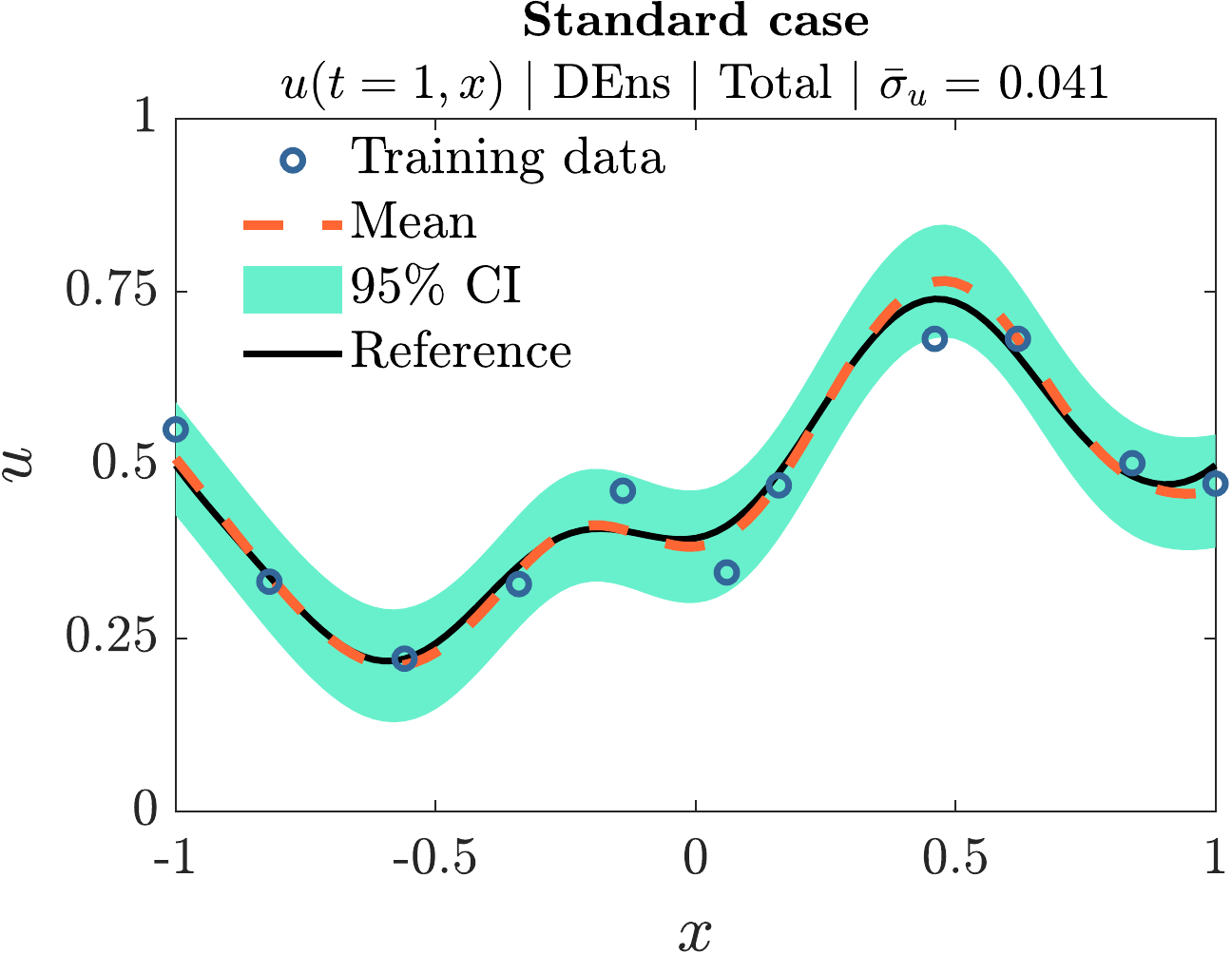}}
	\subcaptionbox{}{}{\includegraphics[width=0.32\textwidth]{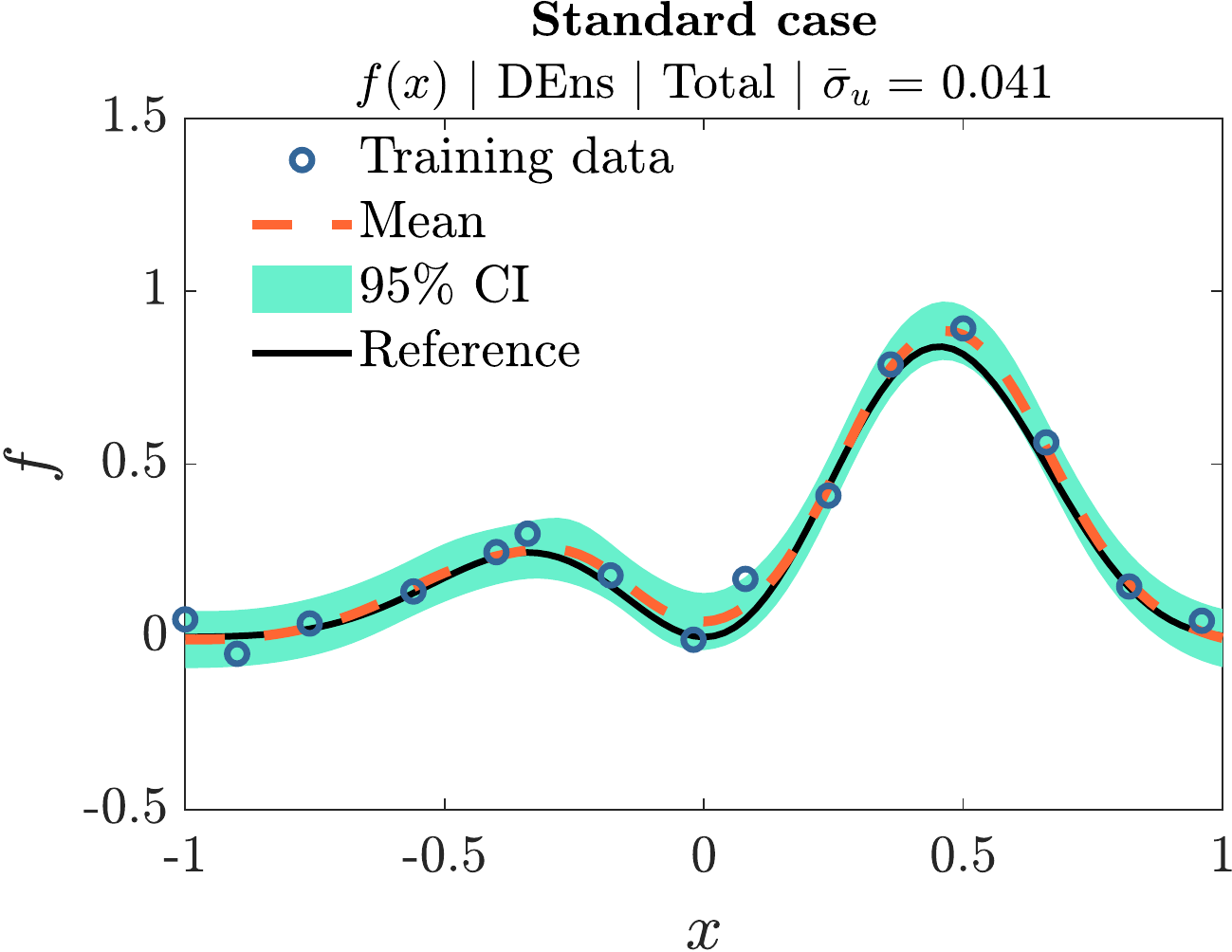}}
	\subcaptionbox{}{}{\includegraphics[width=0.32\textwidth]{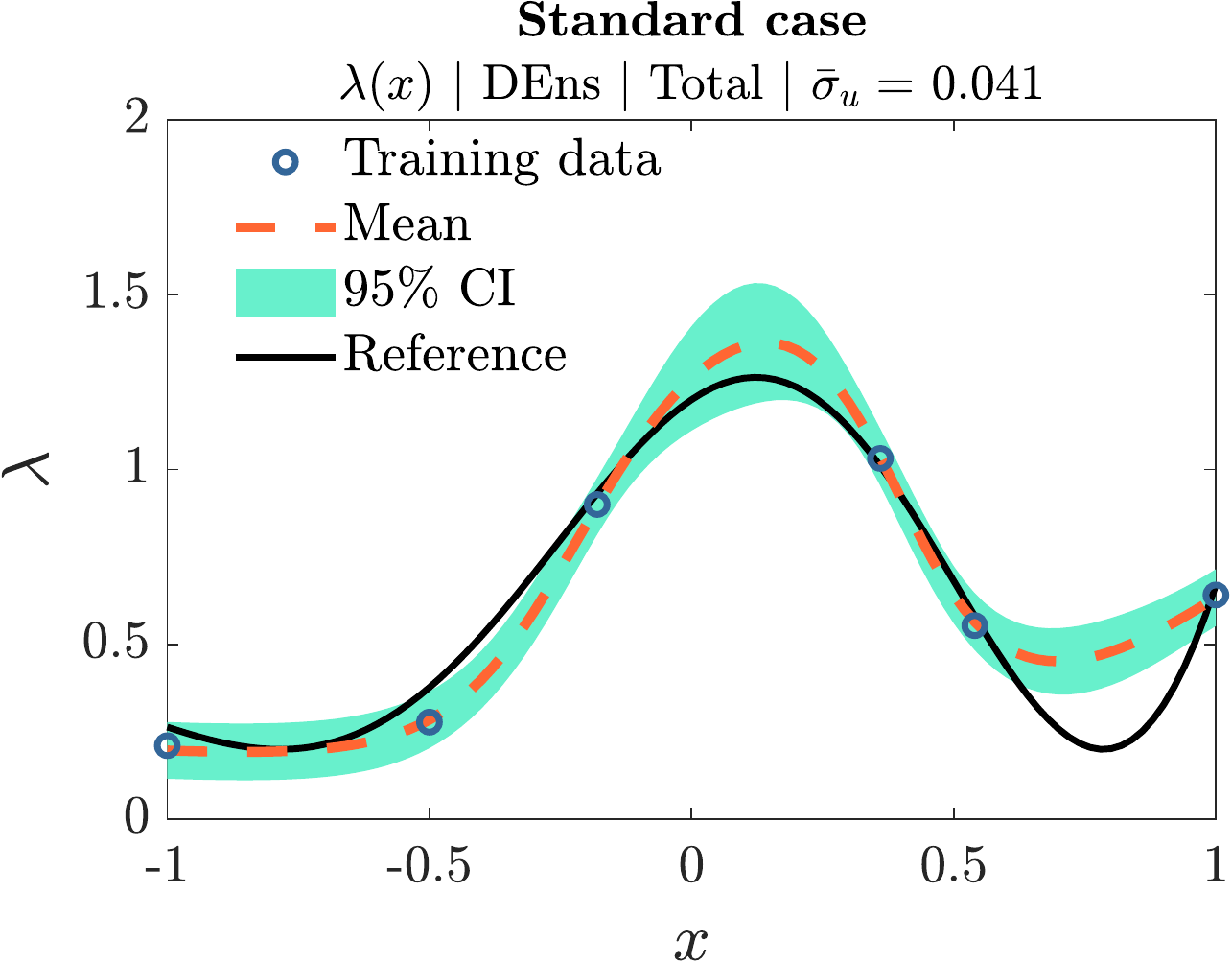}}
	\caption{
		Mixed PDE problem of Eq.~\eqref{eq:comp:pinns:stand} | \textit{Standard case}:
		DEns estimates $\lambda(x)$ more accurately than HMC, although it is over-confident (small epistemic uncertainty).
		Shown here are the training data, reference functions, as well as the mean and uncertainty ($95\%$ CI) predictions of DEns for $u$, $f$, and $\lambda$.
		\textbf{Top row:} epistemic uncertainty.
		\textbf{Bottom row:} total uncertainty including the \textit{learned} amount of aleatoric uncertainty, $\bar{\sigma}_u = 0.041$.
	}
	\label{fig:comp:pinns:stand:ens}
\end{figure}

\newpage

\noindent Further, in Fig.~\ref{fig:comp:pinns:stand:2d} we provide the space- and time-dependent predictions for $u(t, x)$ and compare with the reference solution. As shown in Fig.~\ref{fig:comp:pinns:stand:2d}e and g, although HMC and HMC+FP yield similar mean predictions for $u$, HMC is less confident than
HMC+FP, i.e., it has larger epistemic uncertainty.
This is due to the fact that HMC+FP has been pre-trained using pertinent historical data.

\begin{figure}[!ht]
	\centering
	\subcaptionbox{}{}{\includegraphics[width=0.32\textwidth]{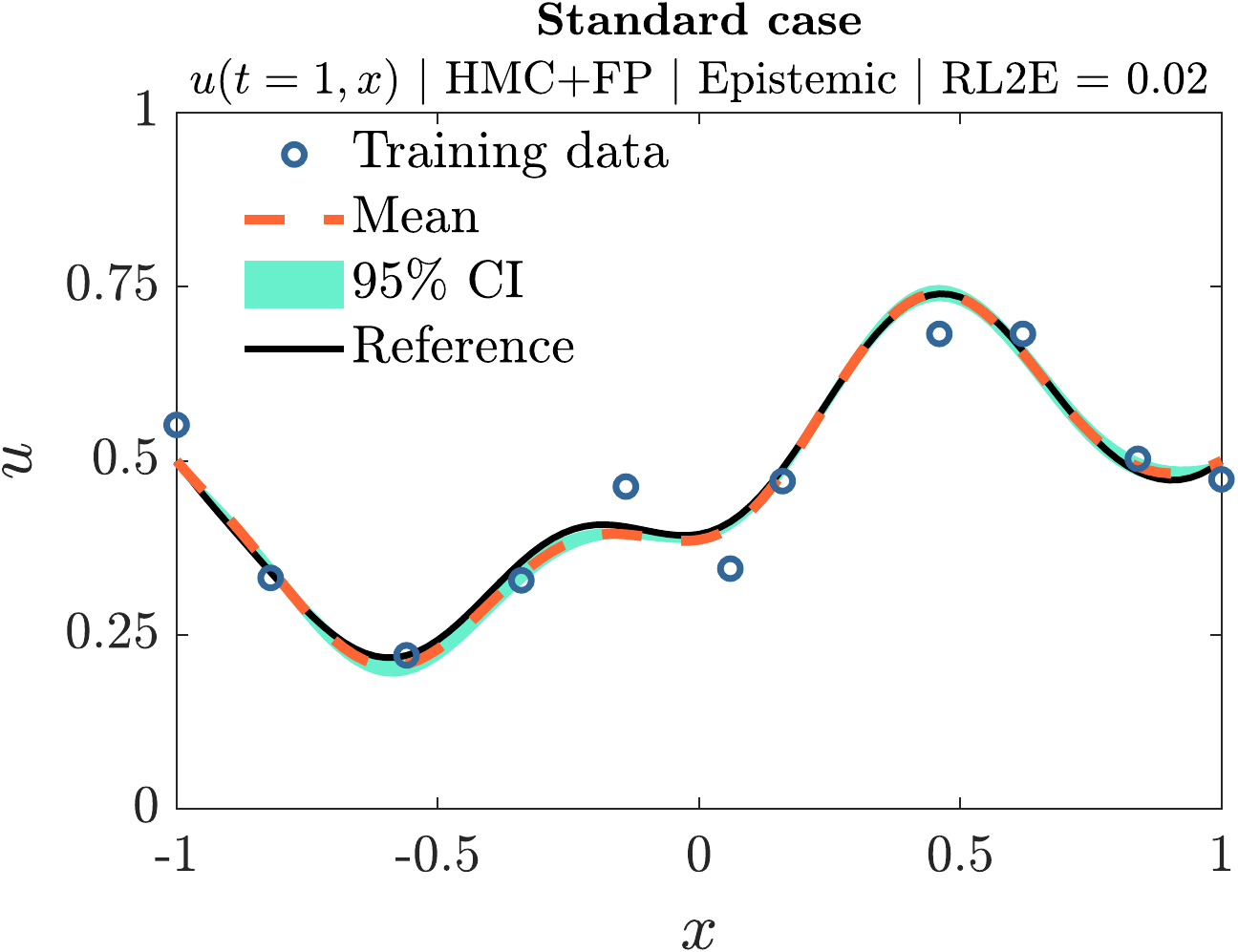}}
	\subcaptionbox{}{}{\includegraphics[width=0.32\textwidth]{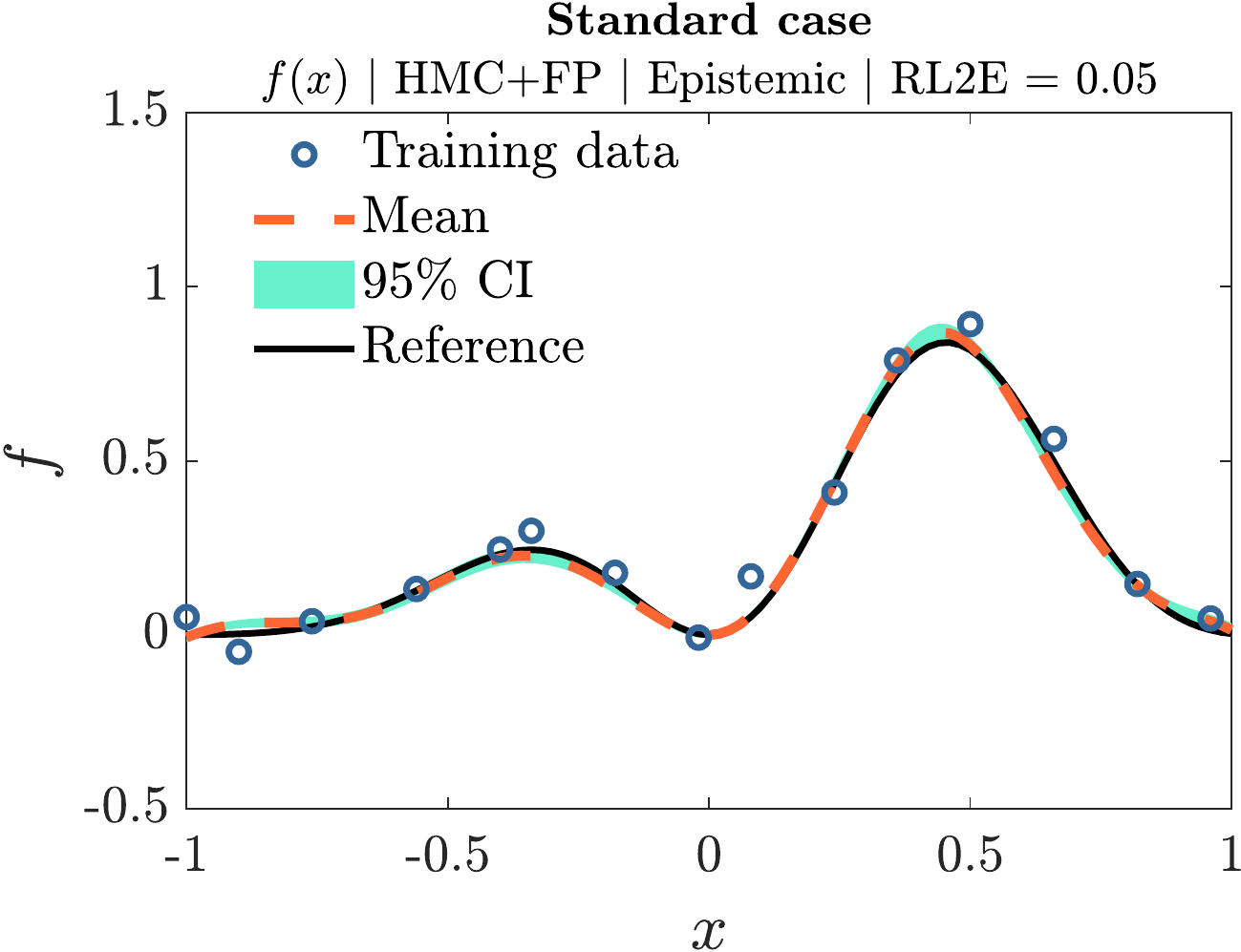}}
	\subcaptionbox{}{}{\includegraphics[width=0.32\textwidth]{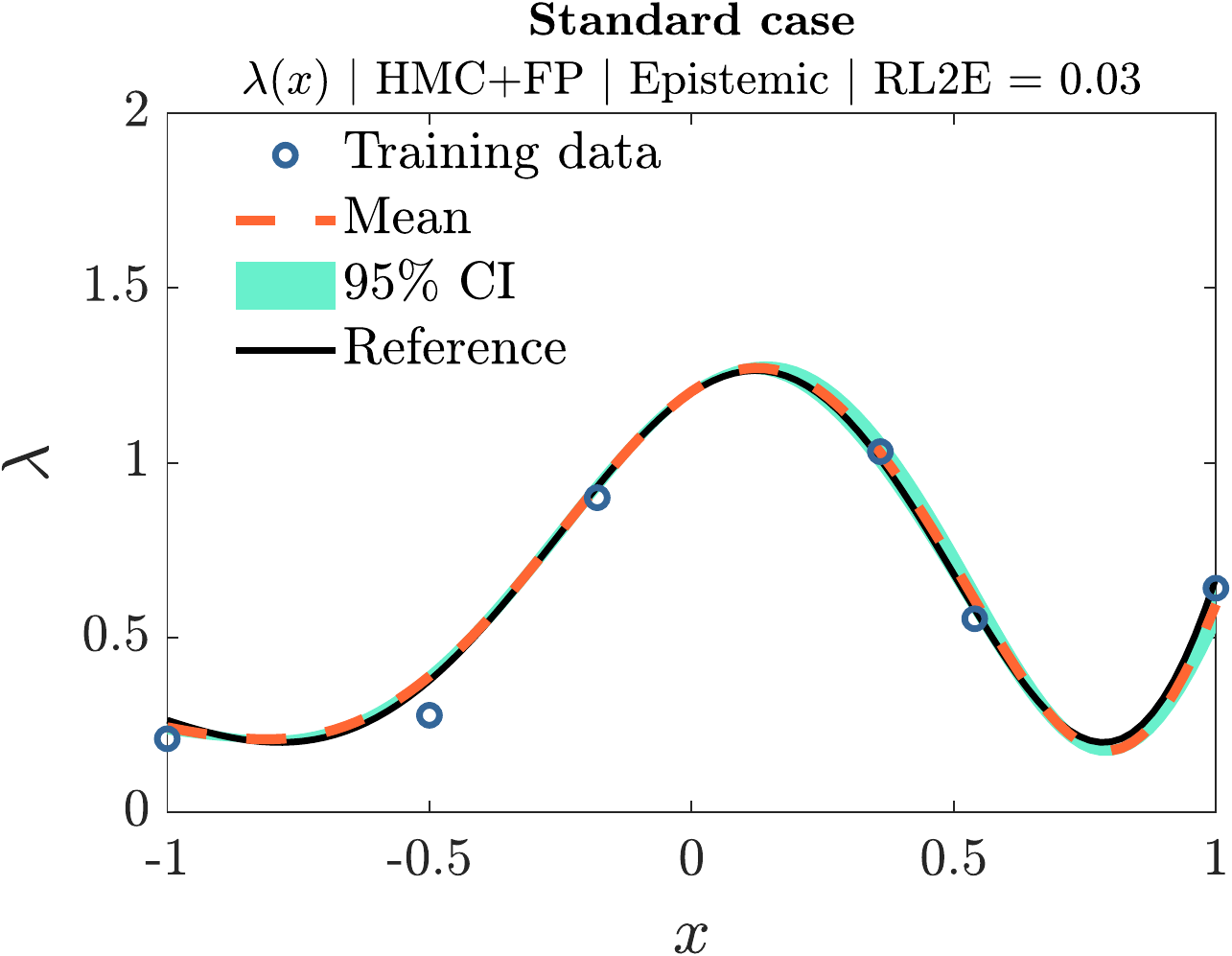}}
	\subcaptionbox{}{}{\includegraphics[width=0.32\textwidth]{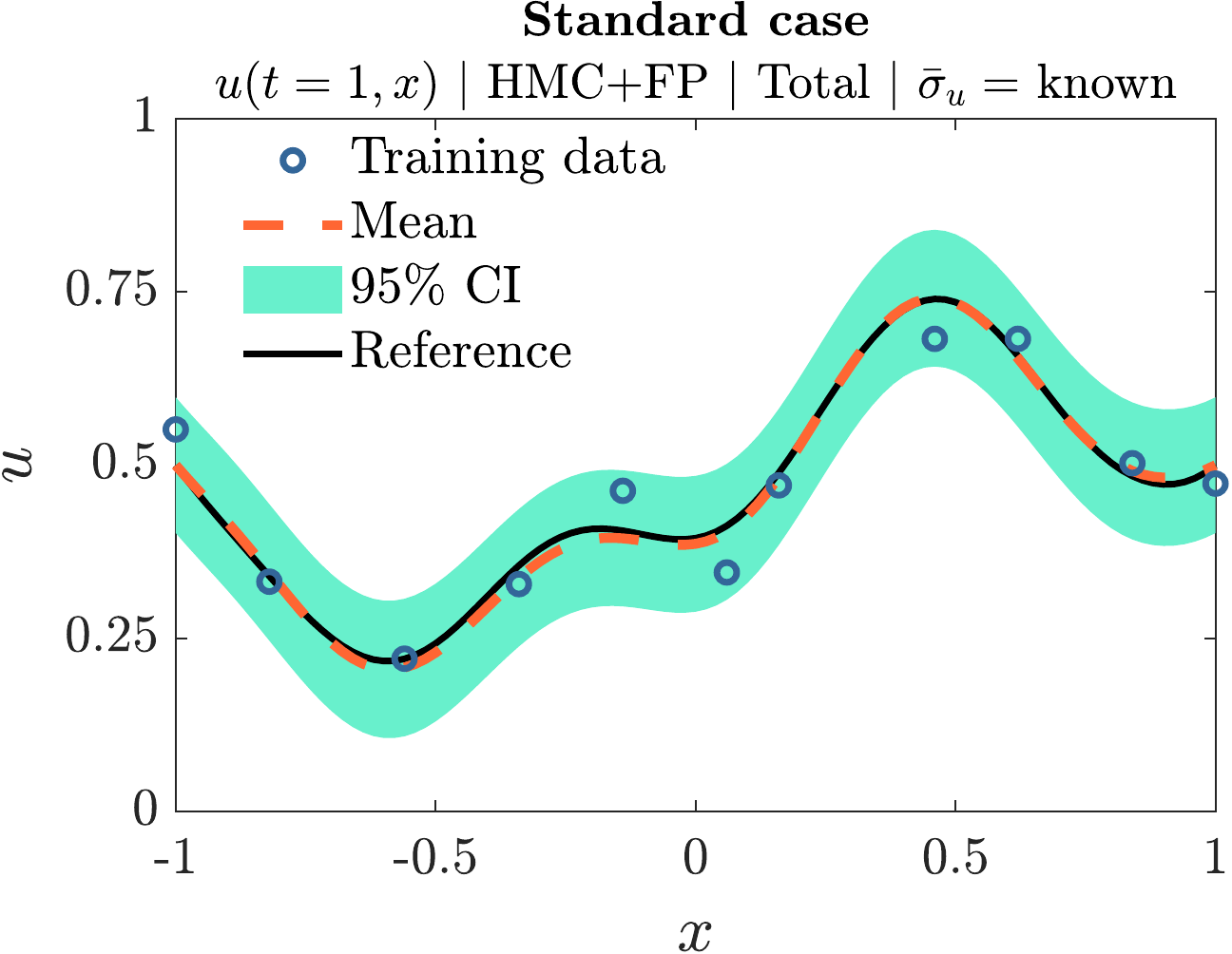}}
	\subcaptionbox{}{}{\includegraphics[width=0.32\textwidth]{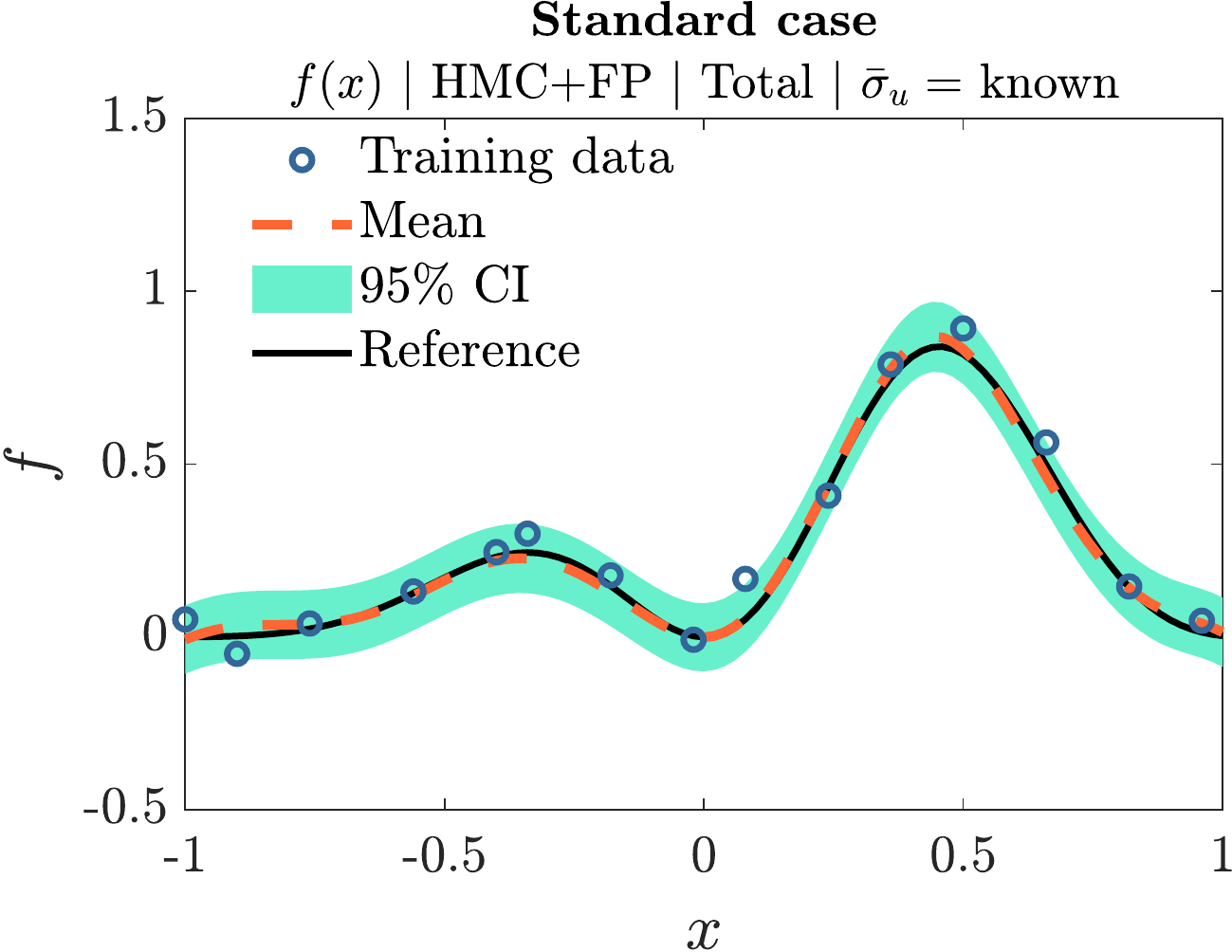}}
	\subcaptionbox{}{}{\includegraphics[width=0.32\textwidth]{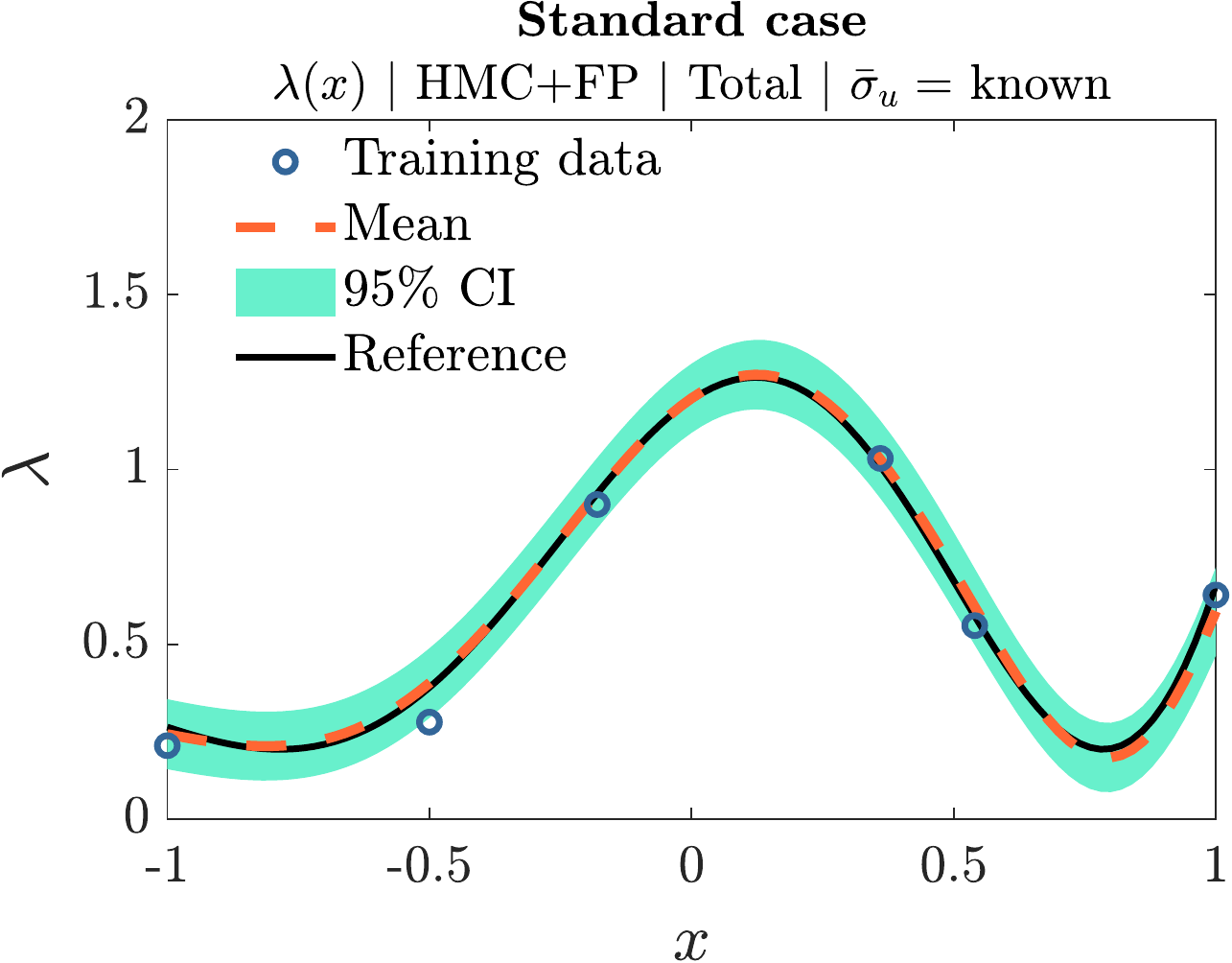}}
	\caption{
		Mixed PDE problem of Eq.~\eqref{eq:comp:pinns:stand} | \textit{Standard case}:
		HMC+FP estimates $\lambda(x)$ more accurately than both HMC and DEns, and it is almost perfectly calibrated (not over- or under-confident).
		Shown here are the training data, reference functions, as well as the mean and uncertainty ($95\%$ CI) predictions of HMC+FP for $u$, $f$, and $\lambda$.
		Results were obtained by pre-training a GAN-FP using historical data of $\lambda(x)$, and subsequently performing HMC posterior inference given the training data shown here; see Section~\ref{sec:uqt:fpriors}.
		\textbf{Top row:} epistemic uncertainty.
		\textbf{Bottom row:} total uncertainty including the considered as \textit{known} amount of aleatoric uncertainty, $\bar{\sigma}_u = 0.05$.
	}
	\label{fig:comp:pinns:stand:hmcfp}
\end{figure}

Next, in Figs.~\ref{fig:comp:pinns:stand:hmc}-\ref{fig:comp:pinns:stand:hmcfp} (and \ref{fig:comp:pinns:stand:mfvi}-\ref{fig:comp:pinns:stand:mcd}) we present the results obtained for $u(t=1, x)$, $f(x)$, and $\lambda(x)$, using HMC, DEns, and HMC+FP (and MFVI as well as MCD). 
It is shown that DEns estimates the mean of $\lambda(x)$ more accurately than HMC, although it is over-confident, i.e., it has point-wise errors that are not covered by the total uncertainty.
On the contrary, HMC+FP estimates $f(x)$ and $\lambda(x)$ more accurately than both HMC and DEns, and it is almost perfectly calibrated (not
over- or under-confident). 
This is reflected in the calibration error in Table~\ref{tab:pinns:standard}, where it is shown that the RMSCE of HMC+FP is at least six times smaller than the RMSCE of the rest of the methods.
Finally, for all UQ methods considered here, the absolute error for $u$ at each $x$ is, in most cases, within the $95 \%$ CI of total (epistemic+aleatoric) uncertainty, i.e., within approximately two standard deviations. 

\subsubsection{Heteroscedastic noise case: larger noise scales close to the boundaries of the space domain}\label{sec:comp:pinns:hetero}

\begin{figure}[!ht]
	\centering
	\subcaptionbox{}{}{\includegraphics[width=0.24\textwidth]{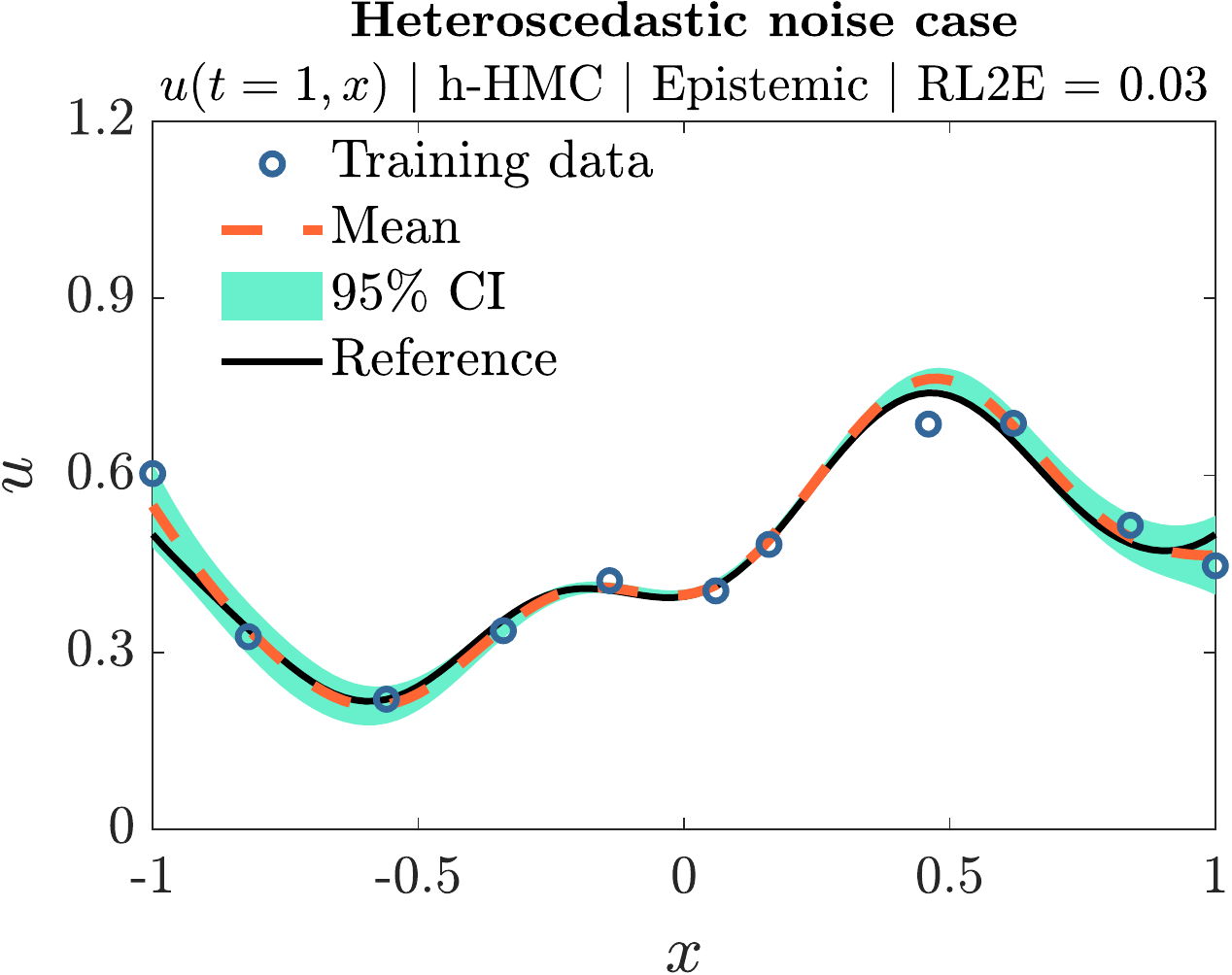}}
	\subcaptionbox{}{}{\includegraphics[width=0.24\textwidth]{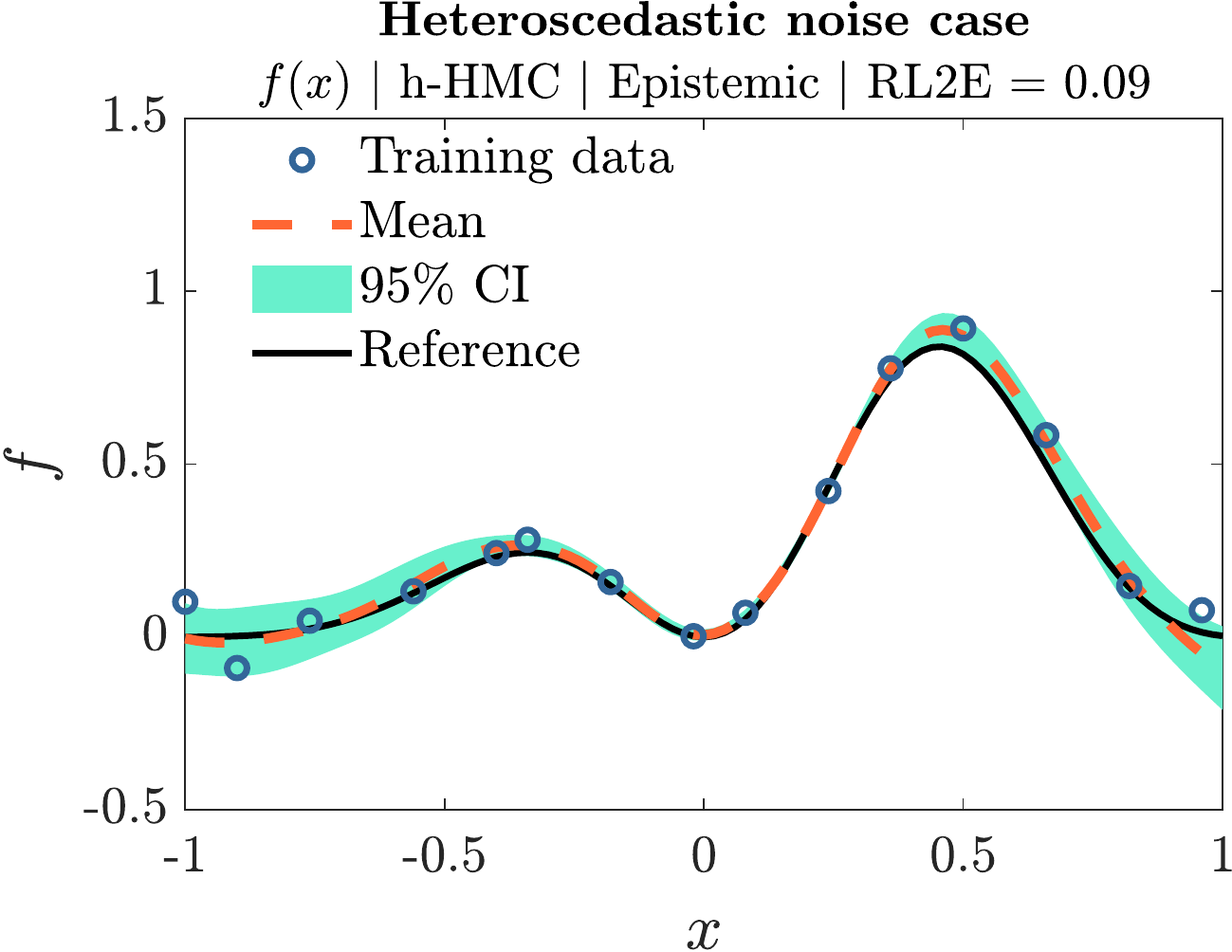}}
	\subcaptionbox{}{}{\includegraphics[width=0.24\textwidth]{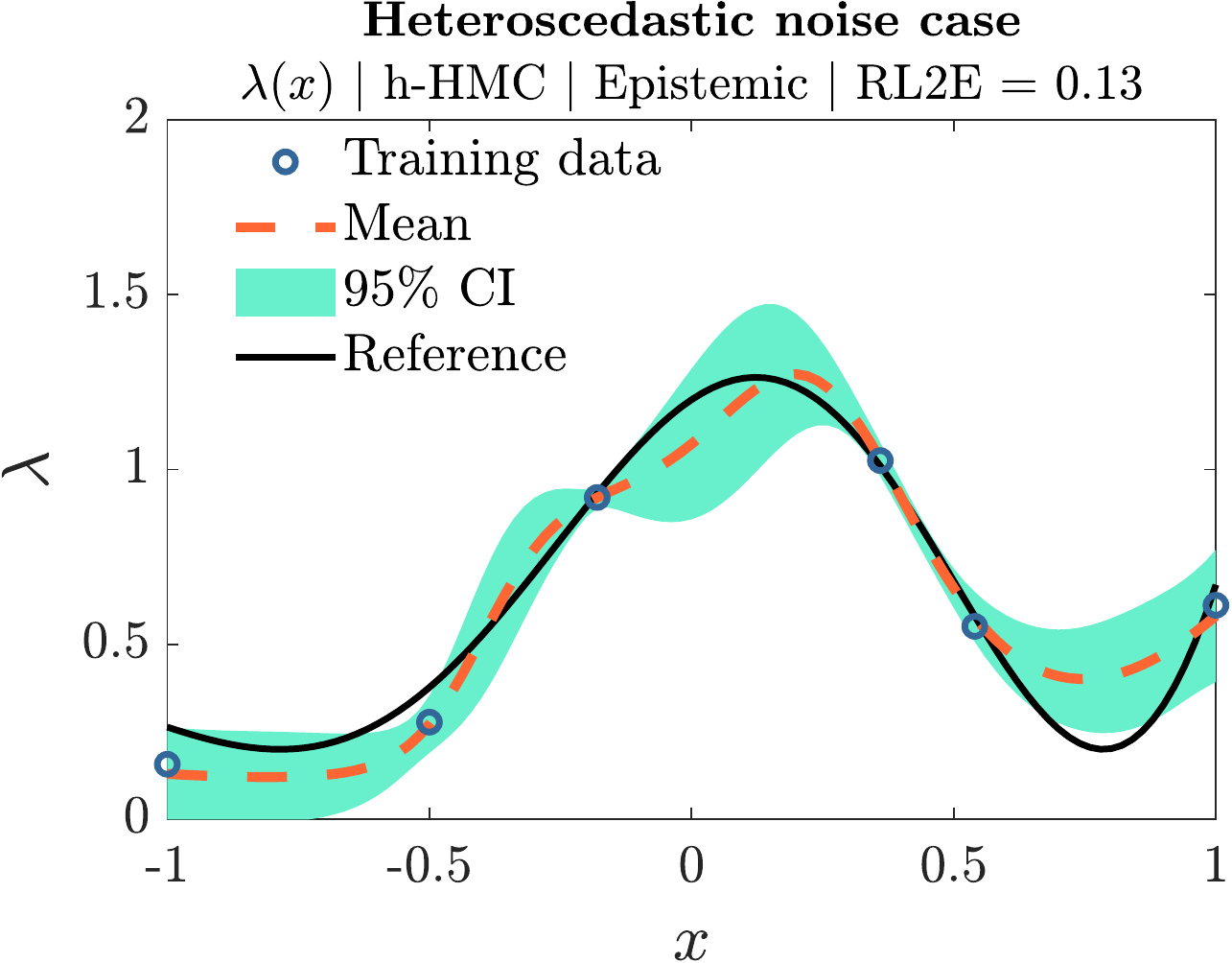}}
	\subcaptionbox{}{}{\includegraphics[width=0.24\textwidth]{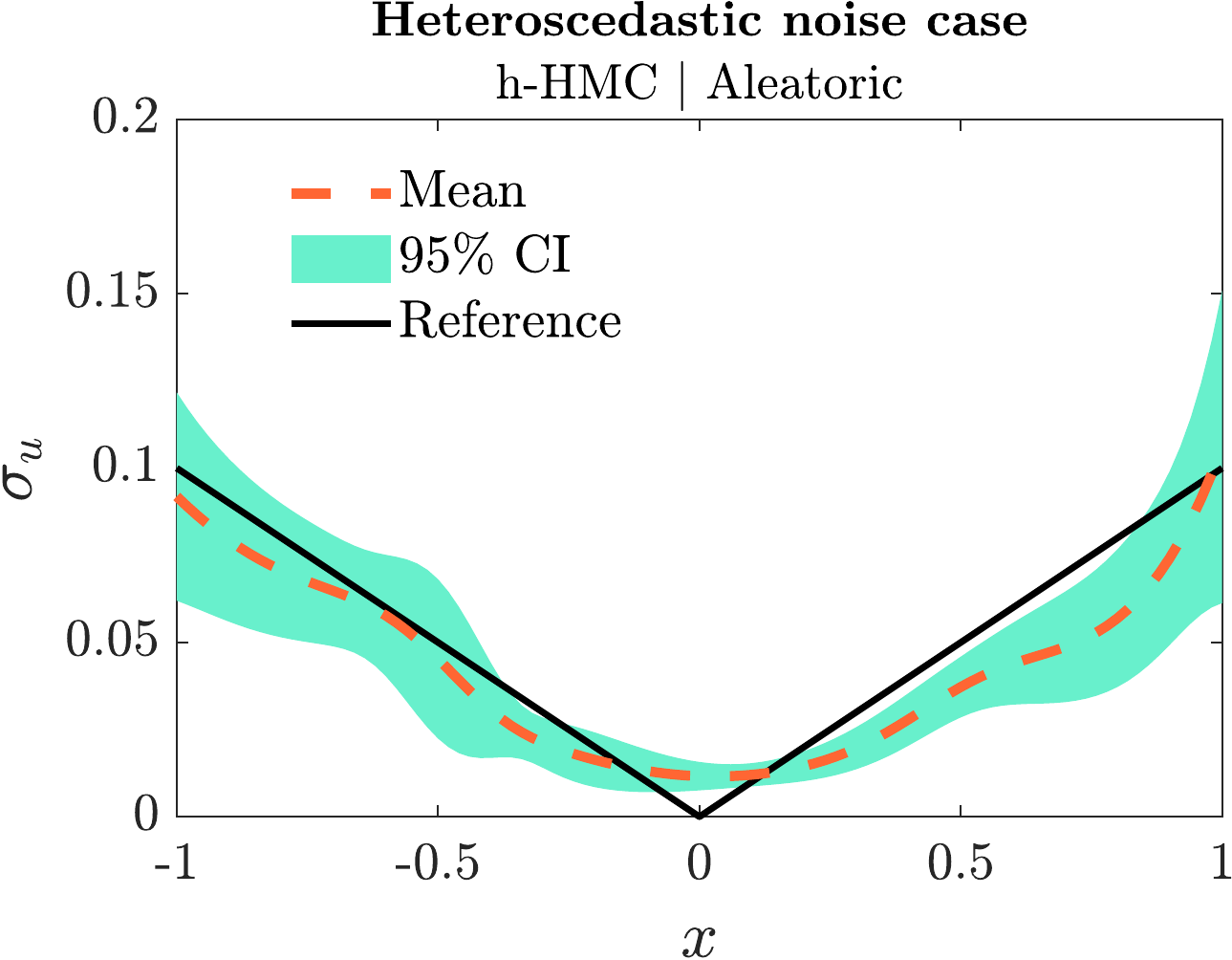}}
	\subcaptionbox{}{}{\includegraphics[width=0.32\textwidth]{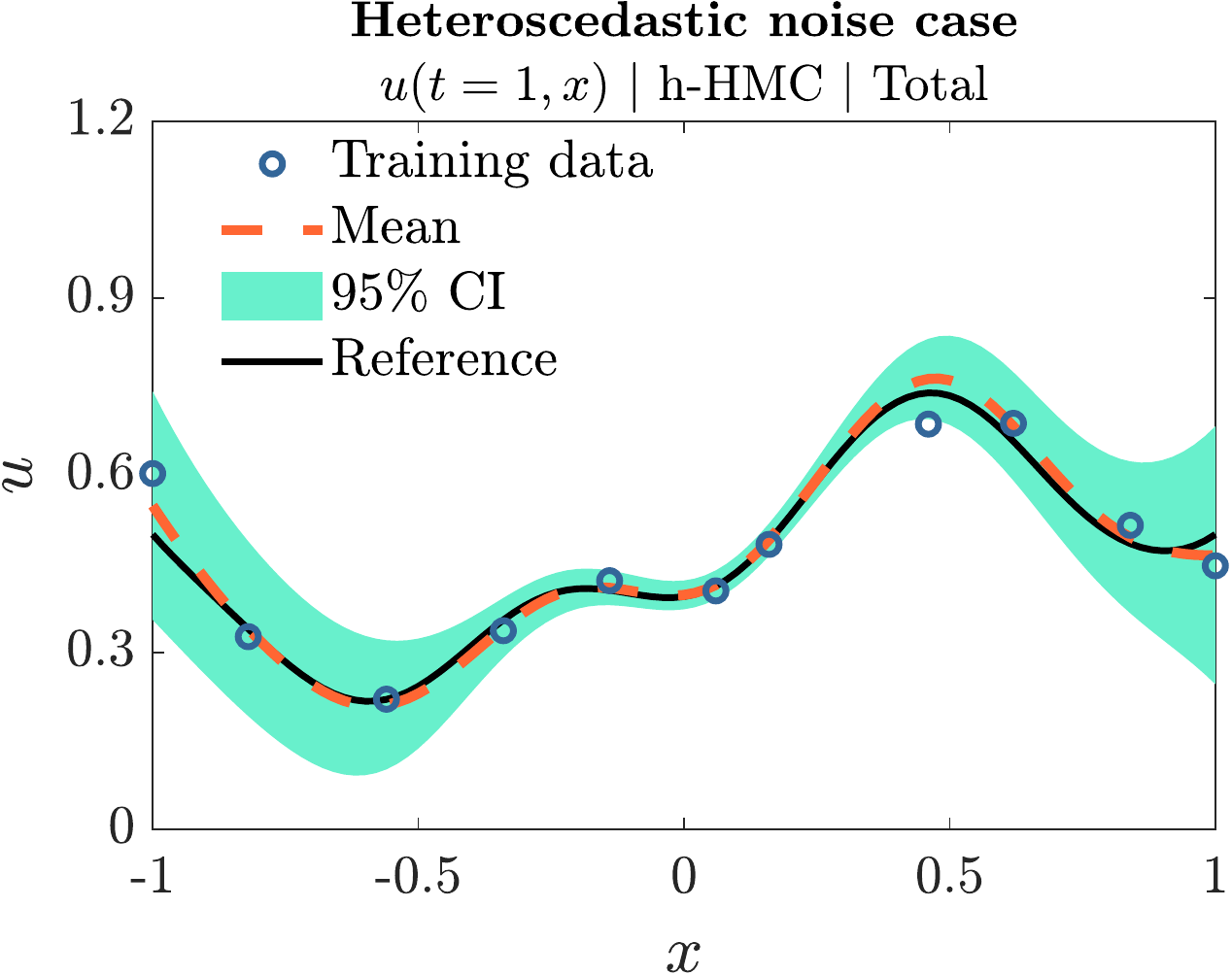}}
	\subcaptionbox{}{}{\includegraphics[width=0.32\textwidth]{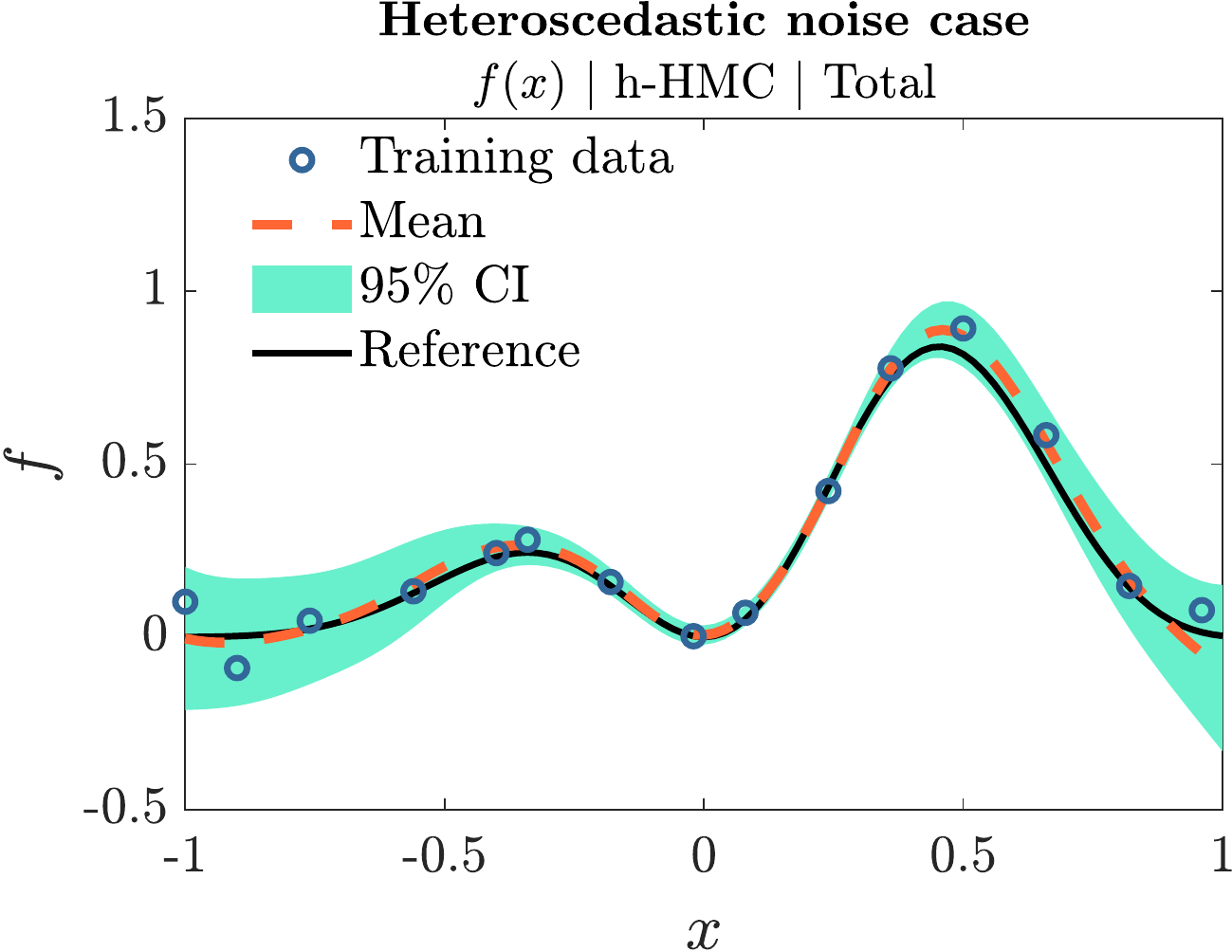}}
	\subcaptionbox{}{}{\includegraphics[width=0.32\textwidth]{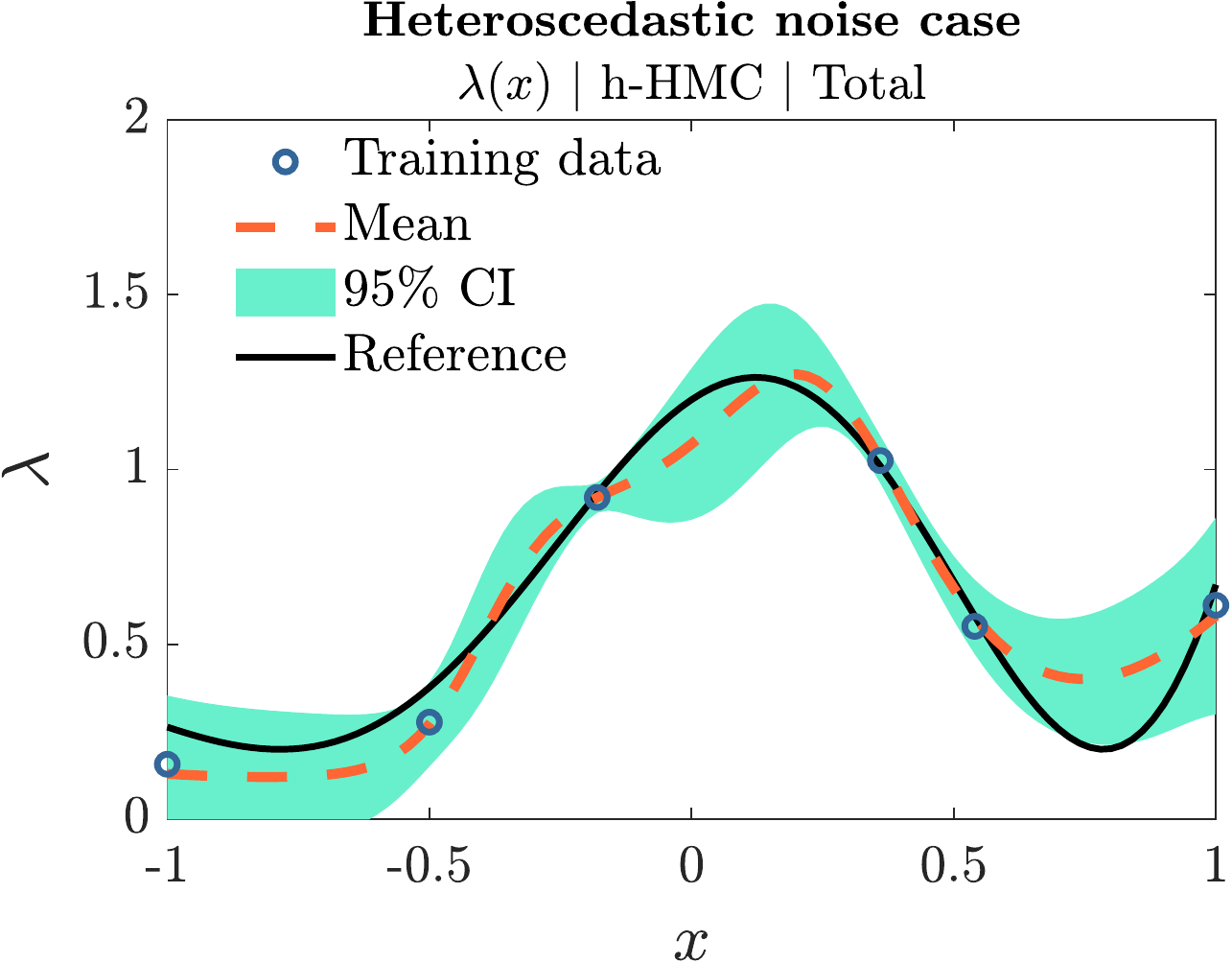}}
	\caption{
		Mixed PDE problem of Eq.~\eqref{eq:comp:pinns:stand} | \textit{Heteroscedastic noise case}:
		epistemic uncertainty of h-HMC increases with more noise ($x \approx -1$ and $x \approx 1$) and with less amount of data, and is affected by the interdependence of $u$, $f$, and $\lambda$ through Eq.~\eqref{eq:comp:pinns:stand:pde}.
		Shown here are the training data, reference functions, as well as the mean and uncertainty ($95\%$ CI) predictions of h-HMC for $u$, $f$, and $\lambda$.
		We fix the prior to $\cN(0, 1)$ for all parameters, and use an additional NN output for learning the space-dependent aleatoric uncertainty; see Fig.~\ref{fig:uqt:bnns:dataunc:heteroscedastic:upinn}.
		\textbf{Top row:} epistemic uncertainty.
		\textbf{Bottom row:} total uncertainty including the \textit{learned} amount of space-dependent aleatoric uncertainty, shown in part (d).
		The learned noise is relatively accurate, whereas the respective point-wise error is within the $95 \%$ CI of its corresponding uncertainty.
	}
	\label{fig:comp:pinns:hetero}
\end{figure}

In this section, we solve the mixed PDE problem of Eq.~\eqref{eq:comp:pinns:stand:pde} with space-dependent, i.e., heteroscedastic, noise with noise scales given as $\sigma_u=\sigma_f=\sigma_{\lambda} = 0.1|x|$.
That is, data is more noisy close to the boundaries of the space domain.
The training dataset locations are the same as in Section~\ref{sec:comp:pinns:stand} and we solve the problem using a U-PINN combined with HMC for posterior inference.
We refer to this method as h-U-PINN-HMC or just h-HMC, for simplicity; see Table~\ref{tab:uqt:over}.
In this regard, the output of the NN of $\lambda$ that we considered in Section~\ref{sec:comp:pinns:stand} is augmented by an additional output for the aleatoric uncertainty, as in Fig.~\ref{fig:uqt:bnns:dataunc:heteroscedastic:upinn}.
We perform posterior inference using HMC to obtain samples $\{\hat{\theta}_j\}_{j=1}^M$, which we use in Eq.~\eqref{eq:modeling:noise:model:totvar} to obtain total uncertainty including heteroscedastic noise.
In Fig.~\ref{fig:comp:pinns:hetero}, we present the mean predictions as well as epistemic and total uncertainties, as obtained for $u(t=1, x)$, $f(x)$, and $\lambda(x)$.
In Fig.~\ref{fig:comp:pinns:hetero}a-c we notice that epistemic uncertainty of h-HMC increases with more noise, i.e., close to the boundaries of the space domain.
It also increases with less amount of data, i.e., between datapoints and especially for $\lambda(x)$ in Fig.~\ref{fig:comp:pinns:hetero}c for which the amount of data is small.
Further, the learned noise is relatively accurate, whereas its respective point-wise error is within the $95 \%$ CI of its corresponding uncertainty (see Fig.~\ref{fig:comp:pinns:hetero}d).
In Table~\ref{tab:pinns:add}, we evaluate the accuracy of the mean predictions (RL2E) for all additional cases, including the heteroscedastic noise case of this section as well as the cases of Sections~\ref{app:comp:pinns:results:large}-\ref{app:comp:pinns:results:steep}.
Clearly, accuracy of the mean predictions deteriorates in the more challenging cases, as compared to the standard case of Table~\ref{tab:pinns:standard}. 
However, using UQ we also obtain uncertainty estimates that cover in most cases the point-wise errors within the $95 \%$ CIs.

\subsubsection{Summary}

In this section, we presented a comparative study pertaining to a mixed PDE problem. For the ``standard'' case, pertaining to constant noise and equidistant measurements, we compared five methods, including a method involving a GAN-FP, termed PI-GAN-FP, for solving the problem and learning the noise. For the ``additional cases'', pertaining to heteroscedastic noise, large constant noise, extrapolation and steep boundary layers, we presented results obtained by HMC.
Overall, we showed that employing a PI-GAN-FP, pre-trained with historical data of the source term and the problem parameters, exhibits higher accuracy than the BNN-based UQ methods (HMC, MFVI, MCD, and DEns).
Among HMC, MFVI, MCD, and DEns, the more expensive techniques (HMC and DEns) perform better, while we estimated the unknown noise satisfactorily with HMC, MFVI, and DEns.
For challenging cases involving large noise, data concentrated on one side of the space domain that require extrapolation, and functions with large gradients, the proposed UQ methods still perform satisfactorily, with uncertainties that cover the errors in most cases.

%% file: IN_stochastic.tex
In this section, we consider the stochastic elliptic equation 
\begin{equation}\label{eq:comp:stochastic:eq}
	-\frac{1}{10}\frac{d}{dx}\left(\lambda(x;\xi)\frac{d}{dx}u(x;\xi)\right) = f(x,\xi),
\end{equation}
where
\begin{subequations}
	\begin{align}
		f(x,\xi) &\sim GP\left(\frac{1}{2}, \frac{9}{400}\exp\left(-(x_1 - x_2)^2\right)\right),\\
		\lambda(x;\xi) & = \exp(\frac{1}{5} \sin(\frac{3}{2}\pi (x+1))+\tilde{\lambda}),\\
		\tilde{\lambda} & \sim GP\left(0, \frac{1}{100}\exp\left(-(x_1 - x_2)^2\right)\right).
	\end{align}
\end{subequations}
Eq.~\eqref{eq:comp:stochastic:eq} can be viewed as a special case of Eq.~\eqref{eq:intro:piml:pinn:pde} without the boundary term of Eq.~\eqref{eq:intro:piml:pinn:pde:b}, with $\pazocal{F}$ given by Eq.~\eqref{eq:comp:stochastic:eq}, and with each $\xi \in \Xi$ being a different random event corresponding to a different sample from the GPs of $f$ and $\lambda$.
The objective of this example is to compare different UQ methods for solving a mixed problem involving Eq.~\eqref{eq:comp:stochastic:eq}, given clean or noisy stochastic realizations of $f$, $u$, and $\lambda$. 
To this end, we consider the dataset of problem 3 in Table~\ref{tab:problem:form}, consisting of $N=$1,000 realizations of $f$, $u$, and $\lambda$, with each of the three quantities sampled at a different number of locations.
Specifically, $N_f=13$ sensors are considered for $f$, $N_u=9$ for $u$, and $N_{\lambda}=5$ for $\lambda$.
Further, there is no boundary term $b$ and the boundary data are included in the aforementioned datasets.
For the noisy data case, the data is contaminated with Gaussian homoscedastic (constant) noise with scales $\sigma_u = \sigma_f = \sigma_{\lambda} = 0.01$.
Next, we implement the three methods of Section~\ref{sec:uqt:sdes}, namely U-PI-GAN, U-NNPC, and U-NNPC+, to solve the problem.
All methods are combined with DEns of Section~\ref{sec:uqt:ens} for performing posterior inference of the $\beta$ parameters of each method. 
That is, we train each model $M=3$ times independently and with different NN initializations for obtaining three predictive models that we use in Eqs.~\eqref{eq:methods:sdes:mean}-\eqref{eq:methods:sdes:var}.
To obtain the reference solutions that we present in the following results, we use the finite difference method.
The hyperparameters and the NN architectures that we used are summarized in Section~\ref{app:comp:hyperparameters} and \ref{app:comp:architecture}, respectively.

For obtaining the first- and second-order statistics of the sought processes $u$ and $\lambda$, consider as an example the U-PI-GAN for predicting $u$. For each of the three PI-GANs that we have trained, we first produce 500 new samples of $u$.
Using these 500 samples, we can compute the mean and standard deviation for each $x$.
That is, given the three trained PI-GANs, we obtain three predictions for the mean and for the standard deviation of $u$.
We use them to compute the mean of the mean and standard deviation predictions.
The variability among these three predictions for the mean and the standard deviation corresponds to epistemic uncertainty.
We compare with the corresponding statistics from a set of 500 new, unseen during training, samples from $f$ and $\lambda$ and corresponding reference solutions $u$.  
Finally, by discretizing the domain of $x$ into 201 points, we also compute the covariance matrix of the predictions of each PI-GAN evaluated on the discrete points.
Subsequently, we obtain the 10 dominant eigenvalues for each of the three trained PI-GANs.
Clearly, the eigenvalues also have variability due to epistemic uncertainty; i.e., due to different optimizations of PI-GAN.
Note that U-NNPC and U-NNPC+ cannot produce new stochastic samples by sampling from a known distribution, as done by U-PI-GAN.
U-NNPC and U-NNPC+ require a new sample from the stochasticity-inducing functions, i.e., from $f$ and $\lambda$, for producing a new sample for $u$.
For this reason, we can obtain the statistics of the predicted stochastic processes based on the training realizations. 
This is equivalent to Monte Carlo simulation, as performed using traditional solvers; see, e.g., \cite{au2014engineering}.
Nevertheless, in this paper we use the $500$ unseen samples of $f$ and $\lambda$ for producing the corresponding predictions of $u$ and $\lambda$ using the trained U-NNPC and U-NNPC+.  

\begin{figure}[!ht]
	\centering
	\subcaptionbox{}{}{\includegraphics[width=0.24\textwidth]{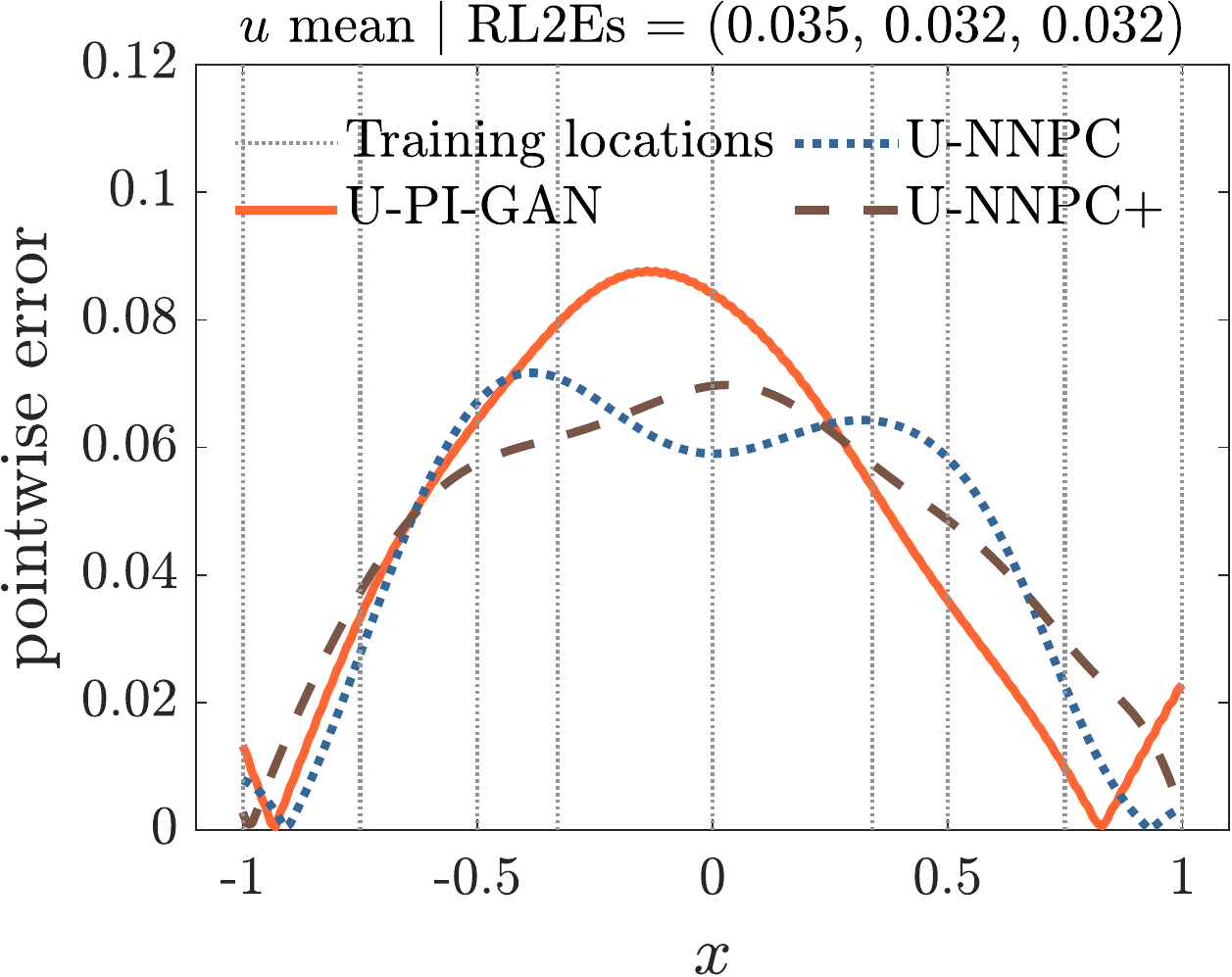}}
	\subcaptionbox{}{}{\includegraphics[width=0.24\textwidth]{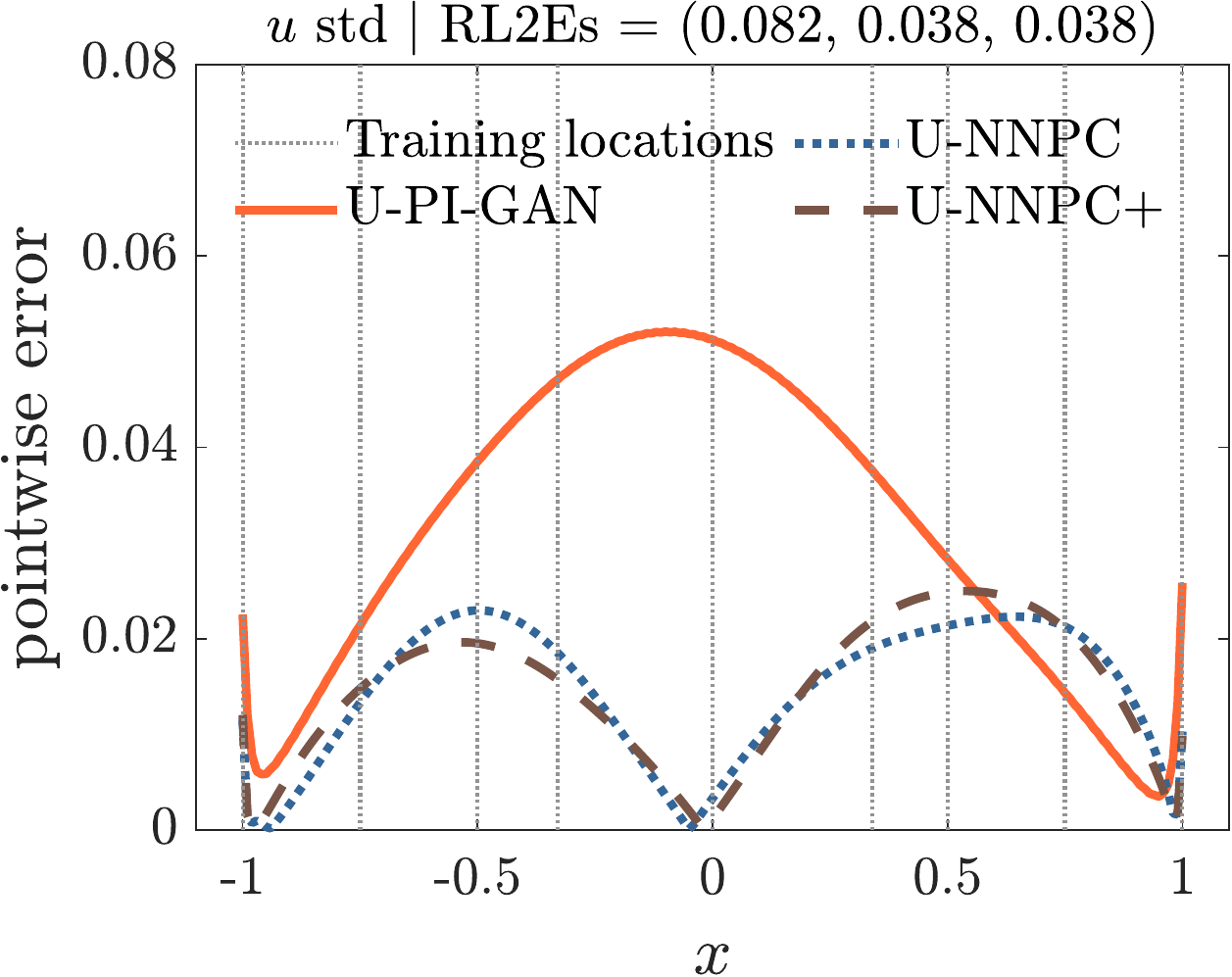}}
	\subcaptionbox{}{}{\includegraphics[width=0.24\textwidth]{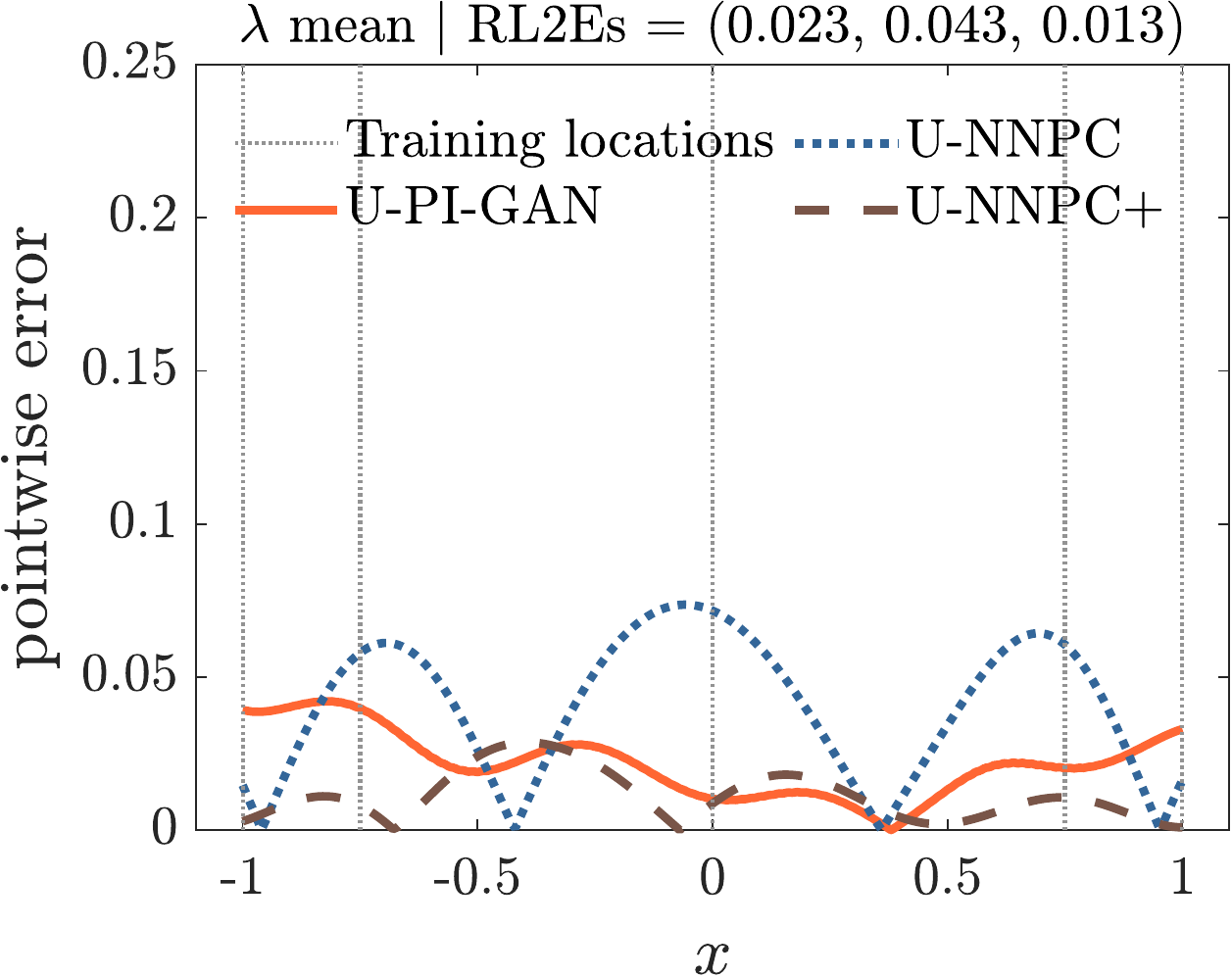}}
	\subcaptionbox{}{}{\includegraphics[width=0.24\textwidth]{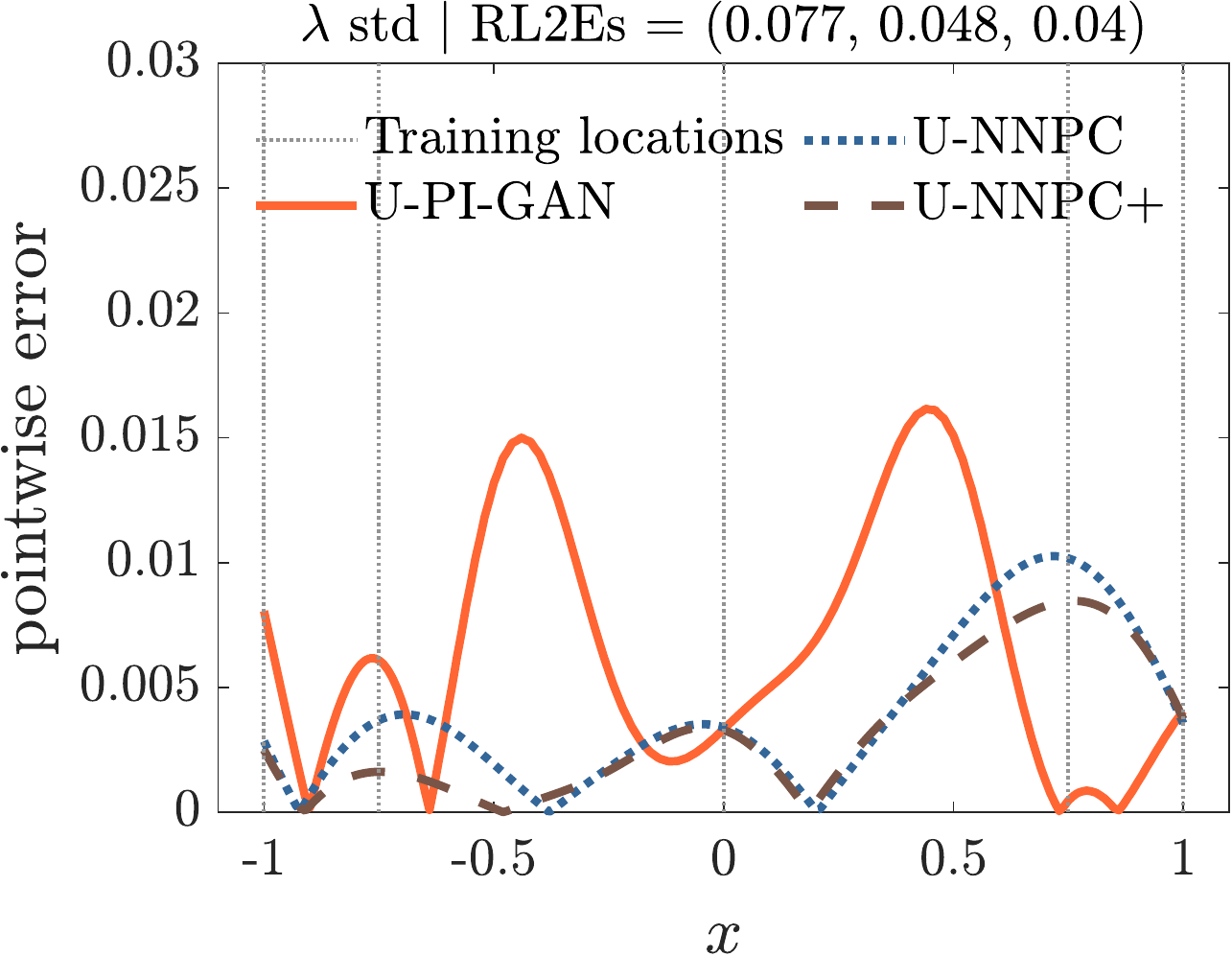}}
	\subcaptionbox{}{}{\includegraphics[width=0.24\textwidth]{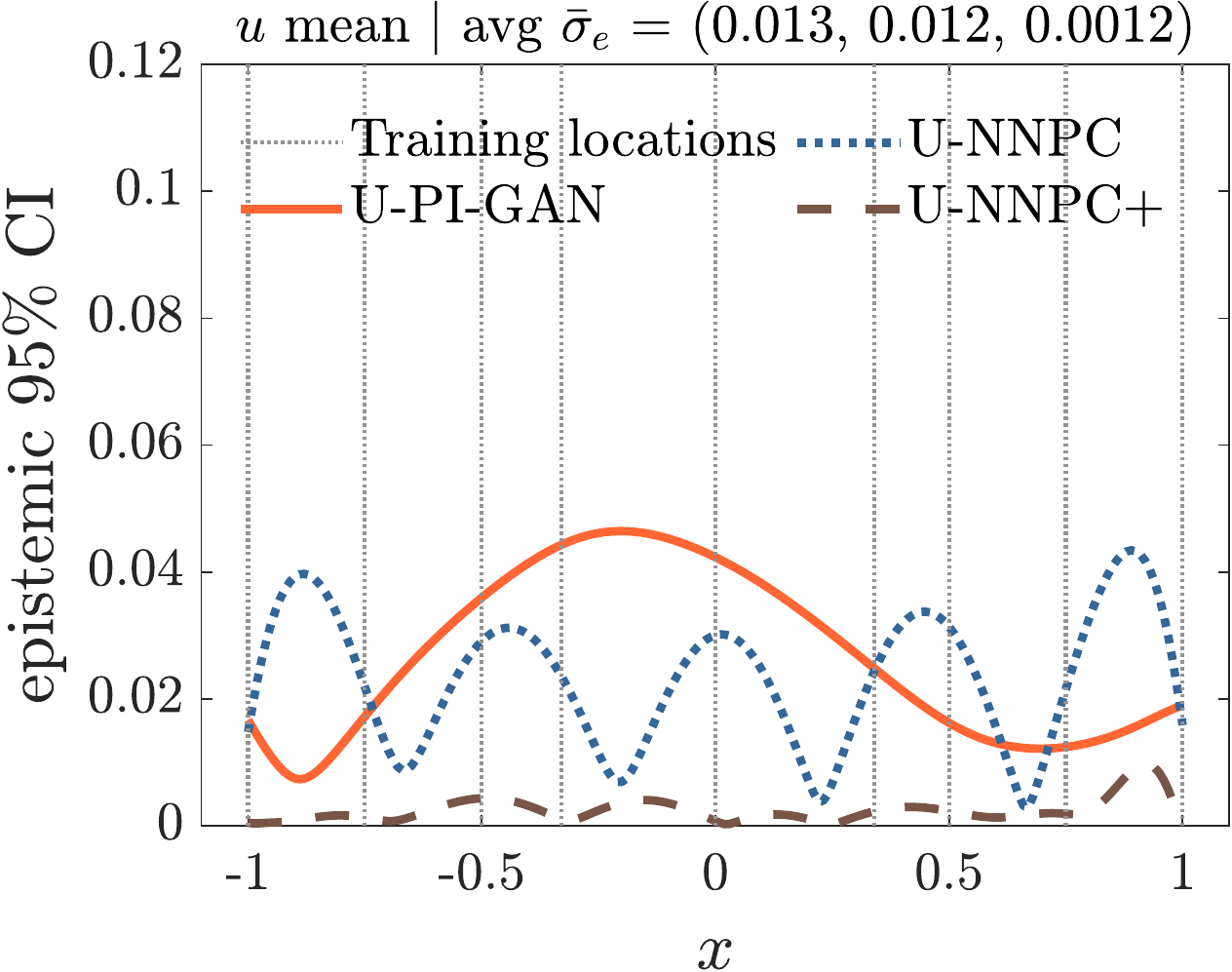}}
	\subcaptionbox{}{}{\includegraphics[width=0.24\textwidth]{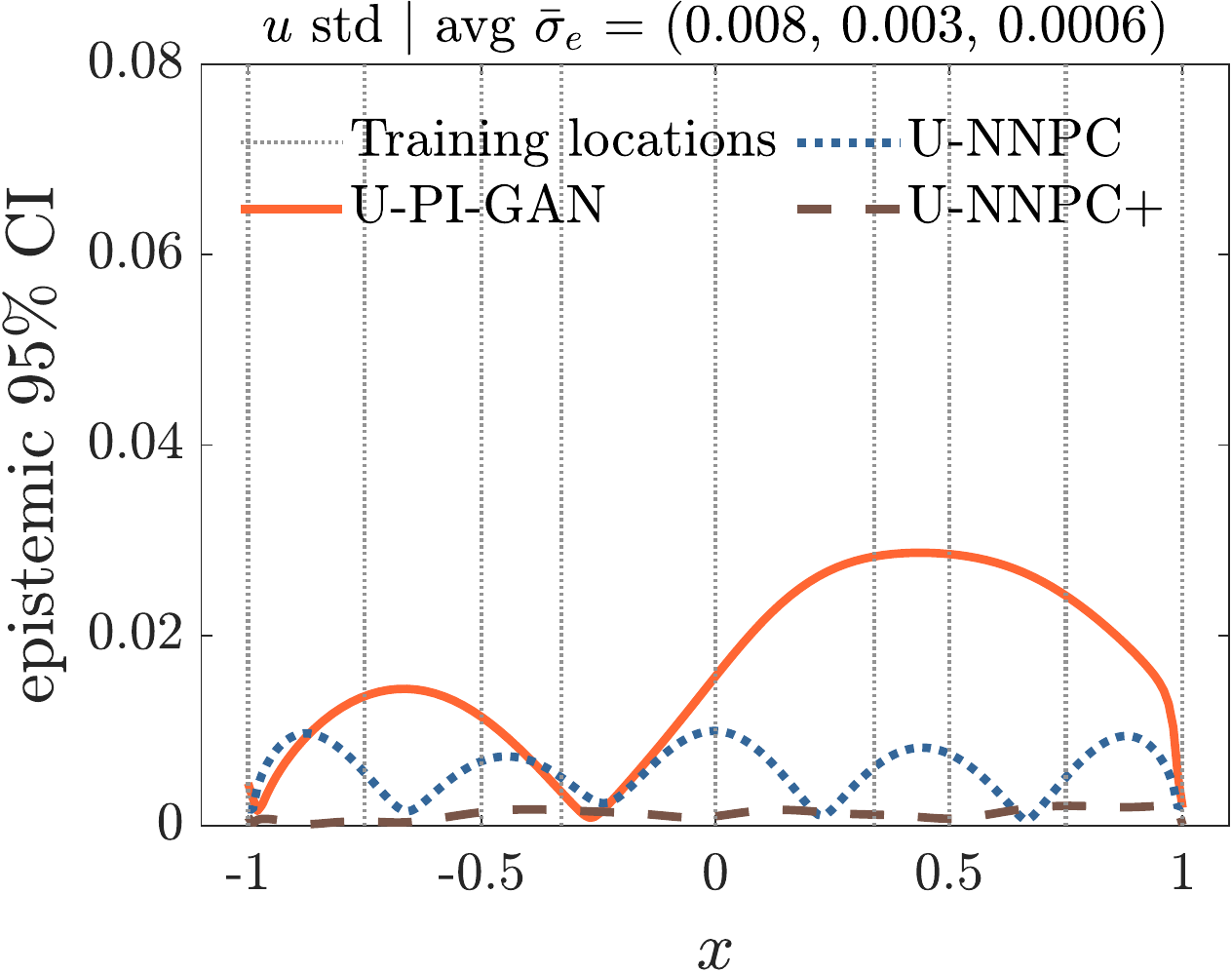}}
	\subcaptionbox{}{}{\includegraphics[width=0.24\textwidth]{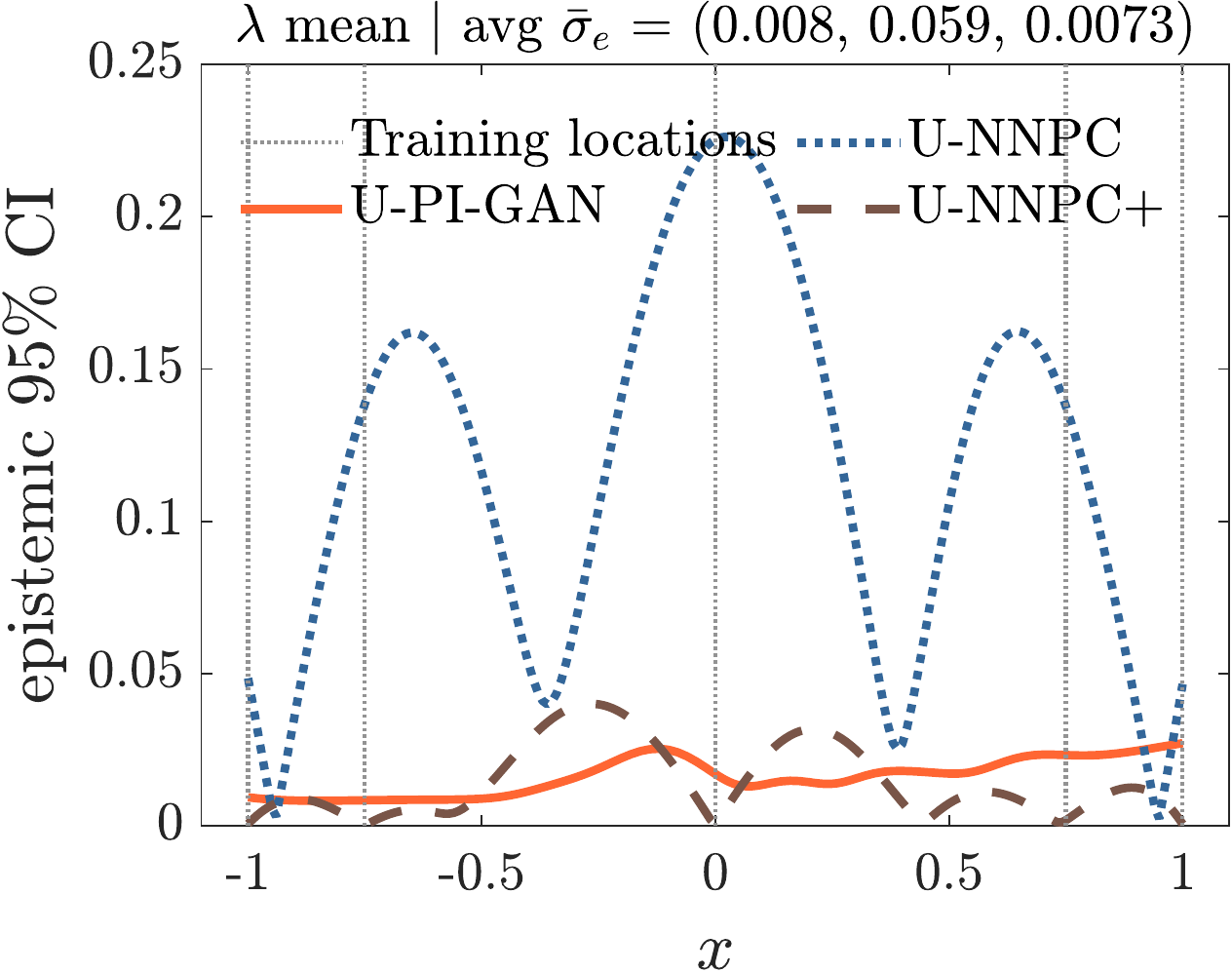}}
	\subcaptionbox{}{}{\includegraphics[width=0.24\textwidth]{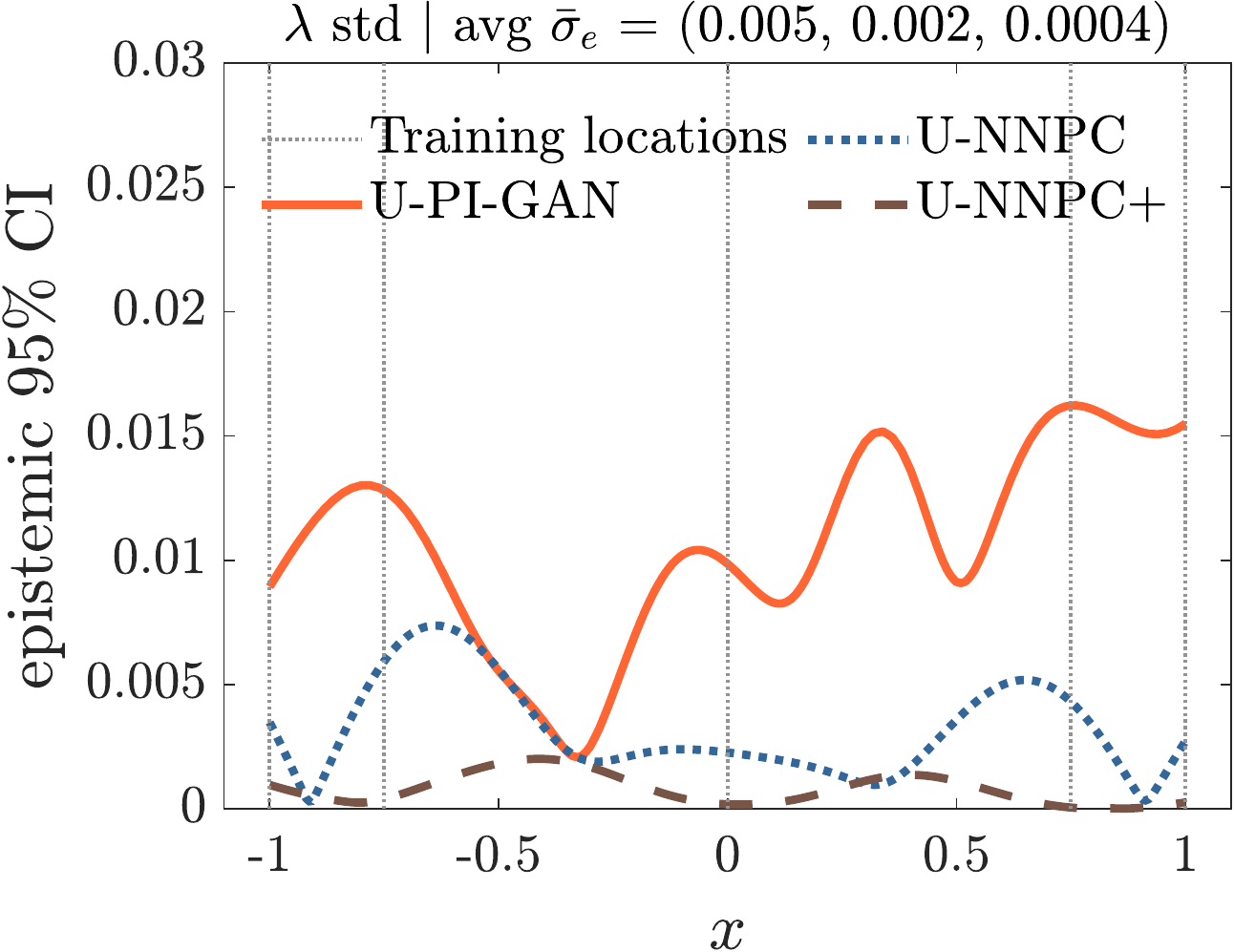}}
	\caption{
		Mixed stochastic problem of Eq.~\eqref{eq:comp:stochastic:eq}: the herein proposed U-NNPC+ performs better in this case and is robust to noisy data.
		\textbf{Top row:} Absolute point-wise errors of the mean and standard deviation (std) of $u(x; \xi)$ and $\lambda(x; \xi)$, as obtained by the unseen realizations of $\lambda$ (reference), the corresponding solutions $u$ (reference), and the predictions of U-PIGAN, U-NNPC, and U-NNPC+ (trained with noisy stochastic realizations). The RL2E values are also shown in the parentheses for U-PIGAN, U-NNPC, and U-NNPC+. 
		\textbf{Bottom row:} Corresponding epistemic uncertainty ($95 \%$ CI - approximately two standard deviations) of the mean and std of $u(x; \xi)$ and $\lambda(x; \xi)$, as obtained by combining the methods with DEns of Section~\ref{sec:uqt:ens}.
		The average along $x$ epistemic standard deviation (avg $\bar{\sigma}_e$) is also shown in the parentheses for U-PIGAN, U-NNPC, and U-NNPC+.
		In both rows, the sensor locations for the noisy data of $u$ and $\lambda$ are shown with vertical dashed lines.
		For both $u$ and $\lambda$, for all UQ methods employed, and for most $x$ locations, the errors are not covered within the $95 \%$ CIs. 
		Note that the epistemic uncertainty of U-NNPC+ is considerably smaller than the other two methods.
	}
	\label{fig:comp:stochastic:preds:noisy}
\end{figure}

\begin{figure}[!ht]
	\begin{subfigure}{.5\textwidth}
		\raggedleft
		\includegraphics[width=0.7\linewidth]{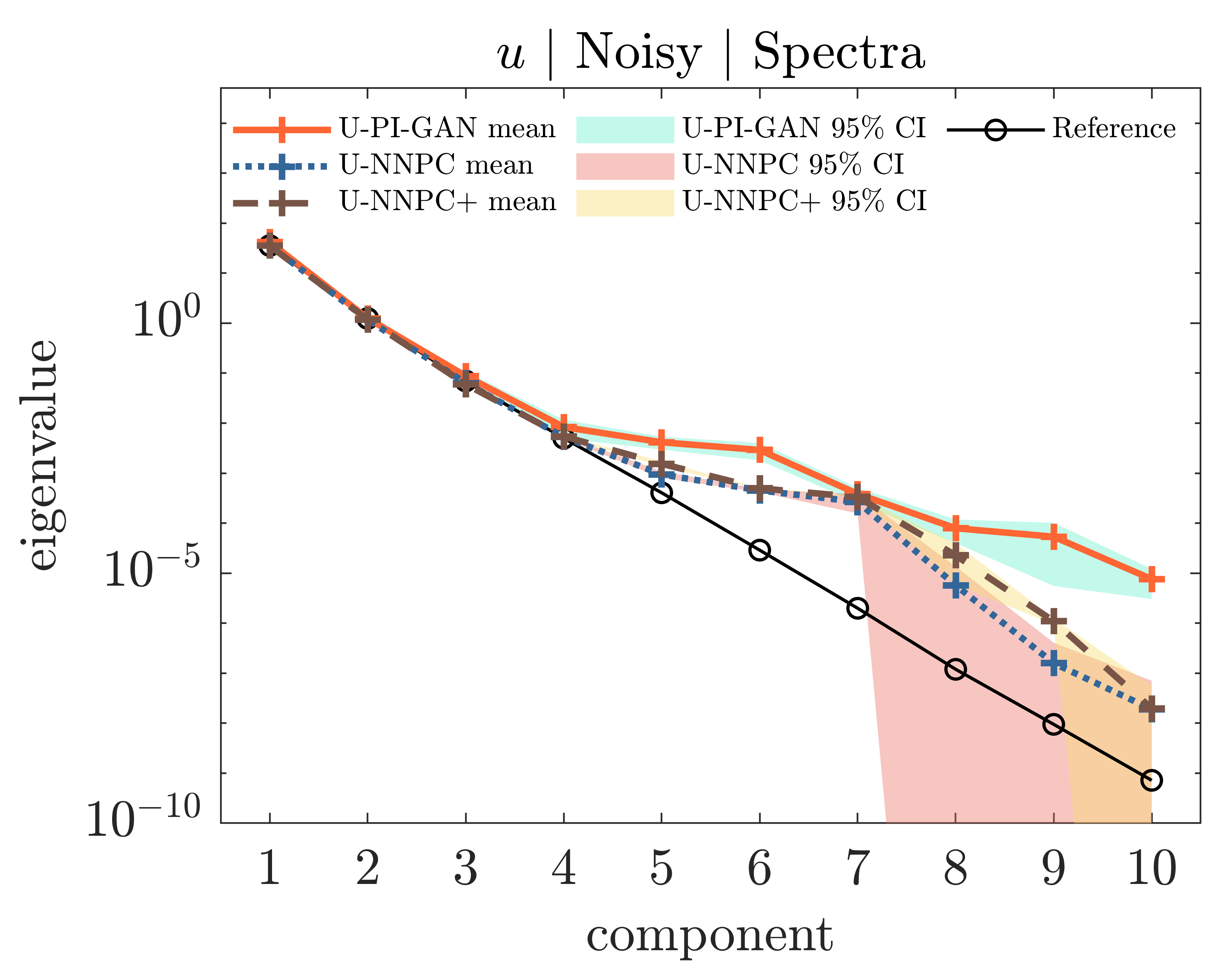}
	\end{subfigure}
	\begin{subfigure}{.5\textwidth}
		\raggedright
		\includegraphics[width=0.7\linewidth]{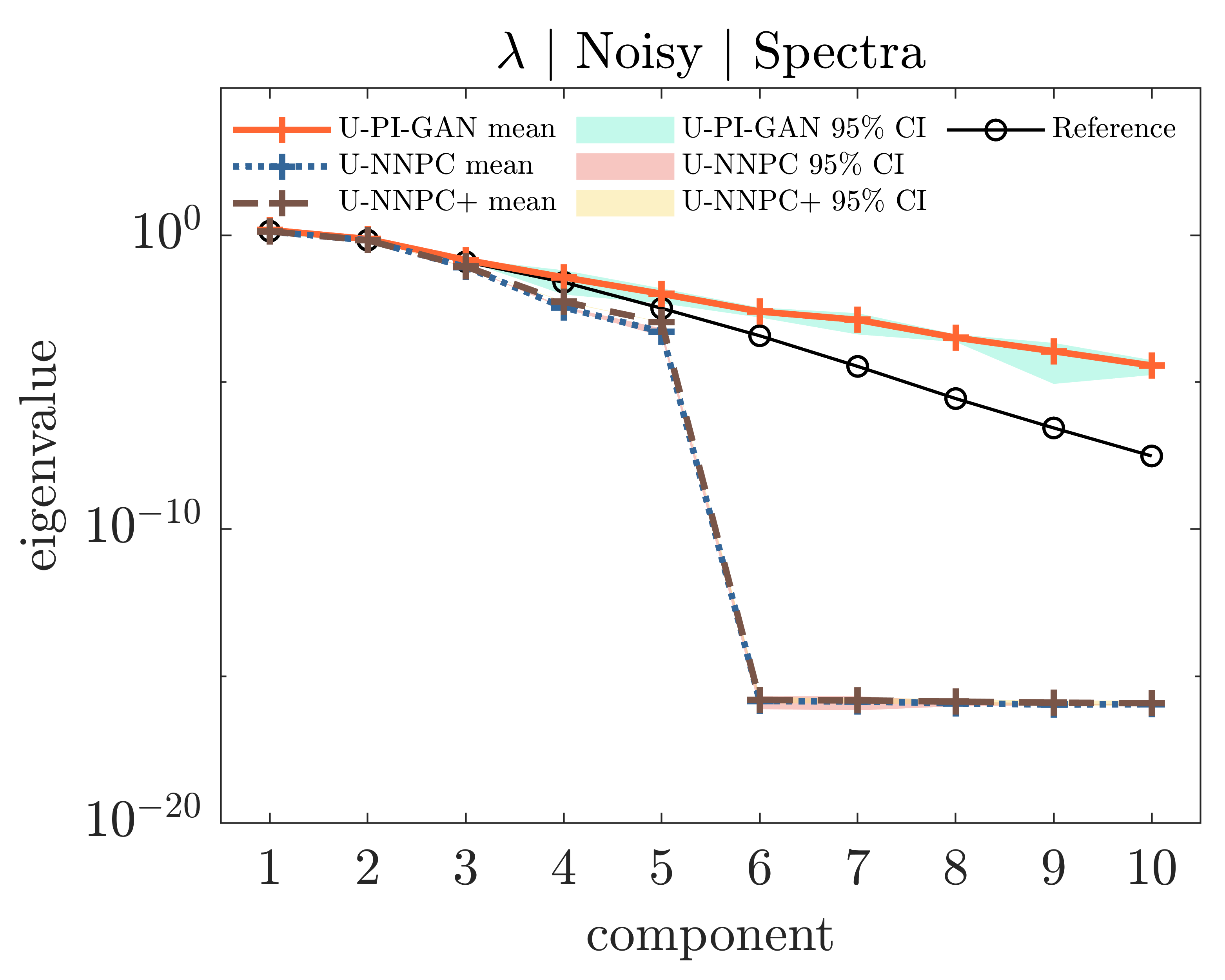}
	\end{subfigure}
	\caption{
		Mixed stochastic problem of Eq.~\eqref{eq:comp:stochastic:eq}:
		U-PI-GAN, U-NNPC, and U-NNPC+ approximate relatively accurately the most significant parts of the spectra of $u$ and $\lambda$.
		Shown here are the first 10 eigenvalues of the covariance matrices of $u(x; \xi)$ and $\lambda(x; \xi)$, as obtained by the unseen realizations of $\lambda$ (reference), the corresponding solutions $u$ (reference), and the predictions of U-PIGAN, U-NNPC, and U-NNPC+. The results were obtained using noisy training realizations.
		For each eigenvalue, the mean and the epistemic uncertainty 95$\%$ CI are provided, as obtained by combining the methods with DEns of Section~\ref{sec:uqt:ens}.
		All covariance matrices were obtained using 201 points for $x$.
	}
	\label{fig:comp:stochastic:eigens}
\end{figure}

Table~\ref{tab:comp:stochastic} summarizes the performance of the considered UQ methods for predicting the mean and the standard deviation (std) of the partially unknown stochastic processes $u(x; \xi)$ and $\lambda(x; \xi)$ of Eq.~\eqref{eq:comp:stochastic:eq}. 
In addition, Fig.~\ref{fig:comp:stochastic:preds:noisy} presents the errors of the mean and standard deviation predictions of the UQ methods, as obtained by training with noisy stochastic realizations.
Although, the herein proposed U-NNPC+ performs better in this case and is robust to noisy data, all three methods approximate relatively accurately the sought stochastic processes $u(x; \xi)$ and $\lambda(x; \xi)$. 
Finally, in Fig.~\ref{fig:comp:stochastic:eigens} we present the spectra of the learned stochastic processes.
Specifically, we provide the first 10 eigenvalues of the covariance matrices
of $u(x; \xi)$ and $\lambda(x; \xi)$, as obtained by the unseen realizations of $\lambda$, the corresponding reference solutions $u$, and the predictions of U-PIGAN, U-NNPC, and U-NNPC+.
The results were obtained using noisy training realizations. For each eigenvalue, the mean and the epistemic uncertainty $95 \%$
CI are provided.
It is seen that all methods approximate relatively accurately the most significant parts of the spectra of $u$ and $\lambda$.
In passing, note that in \ref{fig:comp:stochastic:calib} we present the results corresponding to a post-training calibration experiment. 
In some cases, even with a few (2-10) left-out stochastic realizations, we can reduce the calibration error (RMSCE) by a factor of 3; see, e.g., Fig.~\ref{fig:comp:stochastic:calib}c. 

\begin{table}[!ht]
	\centering
	\footnotesize
	\begin{tabular}{c|c|ccc|ccc}
		\toprule
		\multirow{4}{*}{\textbf{a}}&Metric ($\times 10^2$)& \multicolumn{3}{c|}{Clean data} & \multicolumn{3}{c}{Noisy data}\\ \cline{3-5}  \cline{6-8}
		&corresponding to $u$ & U-PI-GAN & U-NNPC & U-NNPC+ & U-PI-GAN & U-NNPC & U-NNPC+ \\
		\cline{2-8}
		&mean RL2E ($\downarrow$)  & 2.1 & \textbf{0.2} & \textbf{0.2} & 3.5 & \textbf{3.2} & \textbf{3.2} \\
		&std RL2E ($\downarrow$) & 6 & \textbf{0.6} & 0.7 & 8.2 & \textbf{3.8} & \textbf{3.8} \\ 
		\midrule
		\midrule
		\multirow{4}{*}{\textbf{b}}&Metric ($\times 10^2$)& \multicolumn{3}{c|}{Clean data} & \multicolumn{3}{c}{Noisy data}\\ \cline{3-5}  \cline{6-8}
		&corresponding to $\lambda$ & U-PI-GAN & U-NNPC & U-NNPC+ & U-PI-GAN & U-NNPC & U-NNPC+ \\
		\cline{2-8}
		&mean RL2E ($\downarrow$) & \textbf{1.5} & 2.2 & 1.6 & 2.3 & 4.3 & \textbf{1.3} \\ 
		&std RL2E ($\downarrow$) & 6.7 & 5.7 & \textbf{4.3} & 7.7 & 4.8 & \textbf{4} \\ 
		\bottomrule
	\end{tabular}
	\caption{
		Mixed stochastic problem of Eq.~\eqref{eq:comp:stochastic:eq}: the herein proposed U-NNPC+ performs better in this case and is robust to noisy data.
		Here we evaluate the accuracy (RL2E) of U-PI-GAN, U-NNPC, and U-NNPC+ in terms of mean and standard deviation (std) predictions corresponding to the partially unknown stochastic processes $u(x; \xi)$ and $\lambda(x; \xi)$ of Eq.~\eqref{eq:comp:stochastic:eq}.
		The training data consist of 1,000 clean or noisy realizations of $f$, $u$, and $\lambda$, while the test (reference) data we use to calculate RL2E are clean in all cases.
	}
	\label{tab:comp:stochastic}
\end{table}

\subsubsection{Summary}

In this section we solved a mixed SPDE problem. We considered three methods, namely U-PI-GAN, U-NNPC, and the herein proposed U-NNPC+, which were combined with an ensemble of three independent training sessions of the involved NNs. 
We observed that U-NNPC+ can outperform existing methods for treating cases with noisy stochastic realizations. Nevertheless, we found in our experiments that epistemic uncertainty obtained by combining deep ensembles with the considered methods may not cover the errors in predicting the first- and second-order statistics of the quantities of interest.
Further, the epistemic uncertainty of U-NNPC+ was reduced, as compared to U-NNPC, leading to over-confident predictions.

%% file: IN_forward_PINN.tex
In this section, we consider the deterministic 1D nonlinear static diffusion-reaction equation
\begin{equation}\label{eq:comp:pinns:forw:pde}
	D \partial^2_xu - \lambda u^3 = f,  ~x \in [-1, 1],
\end{equation}
where $D = 0.01$ is the diffusion coefficient, $\lambda$ is the constant reaction rate, while $f(x)$ and $u(x)$ are the source term and the sought solution, respectively. We set the exact solution as $u(x) = \sin^3(6x)$, and $f(x)$ can then be derived given the exact $u(x)$. 
Similarly to Section~\ref{sec:comp:pinns}, Eq.~\eqref{eq:comp:pinns:forw:pde} can be viewed as a special case of Eq.~\eqref{eq:intro:piml:pinn:pde}, with $\xi$ fixed and dropped, and $\pazocal{F}$ given by Eq.~\eqref{eq:comp:pinns:forw:pde}.
In this case, however, the model parameter $\lambda$ is known and equal to 0.7.
The objective of this example is to compare U-PINN with the herein proposed GP+PI-GAN for solving a forward PDE problem involving Eq.~\eqref{eq:comp:pinns:forw:pde}, given noisy data of $f$ as well as noisy boundary conditions for $u$. 
To this end, we consider the dataset of problem 1 in Table~\ref{tab:problem:form}, consisting of $N_f=15$ or 6 measurements of $f(x)$, and 2 measurements on $u(x)$ at $x = -1$ and 1. 
The data for $f(x)$ and $u(x)$ are both contaminated with noise following $\cN(0, 0.1^2) $. 
The hyperparameters and the NN architectures that we used are summarized in Section~\ref{app:comp:hyperparameters} and \ref{app:comp:architecture}, respectively.

\begin{figure}[!ht]
	\centering
	\subcaptionbox{}{}{\includegraphics[width=0.32\textwidth]{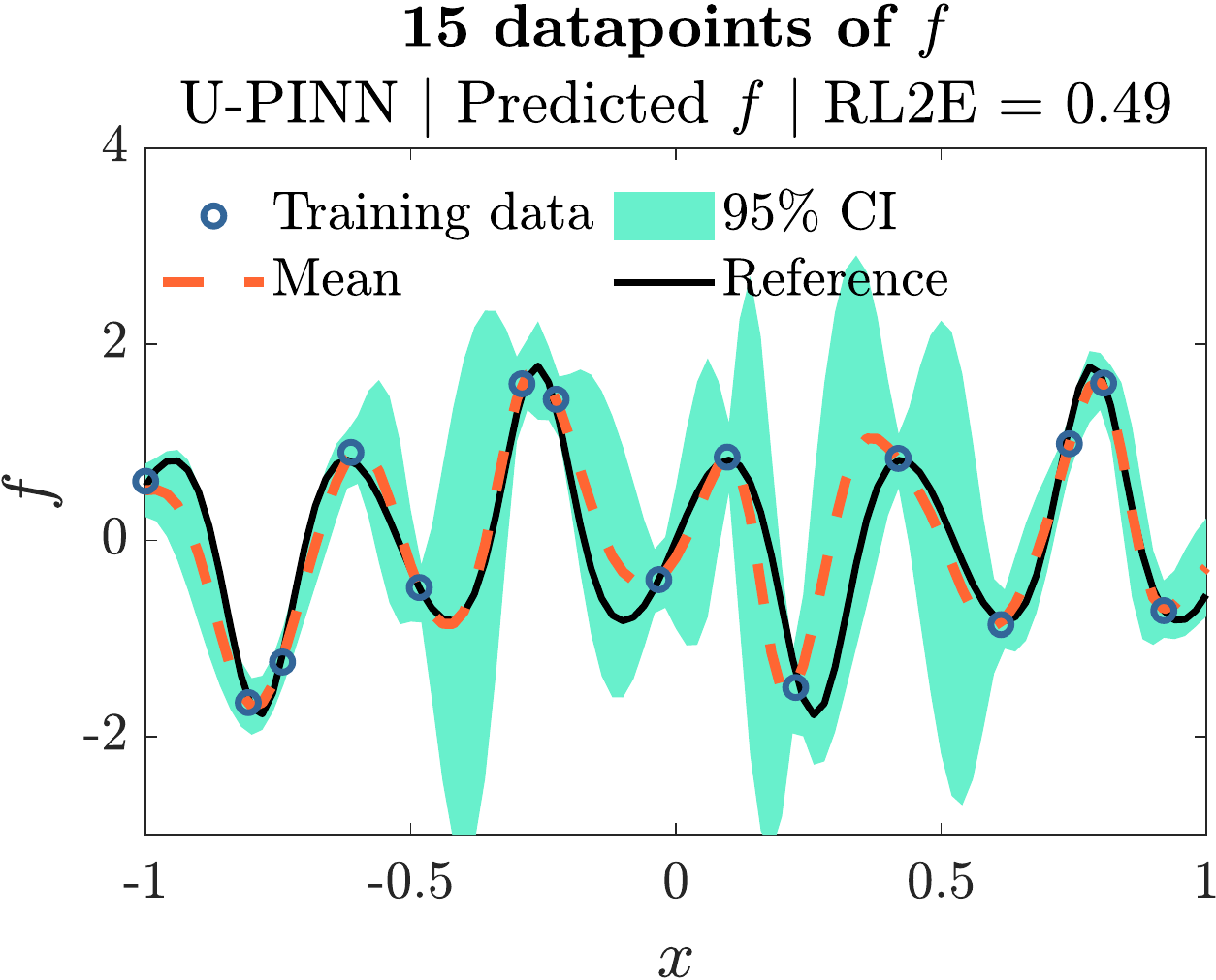}}
	\subcaptionbox{}{}{\includegraphics[width=0.32\textwidth]{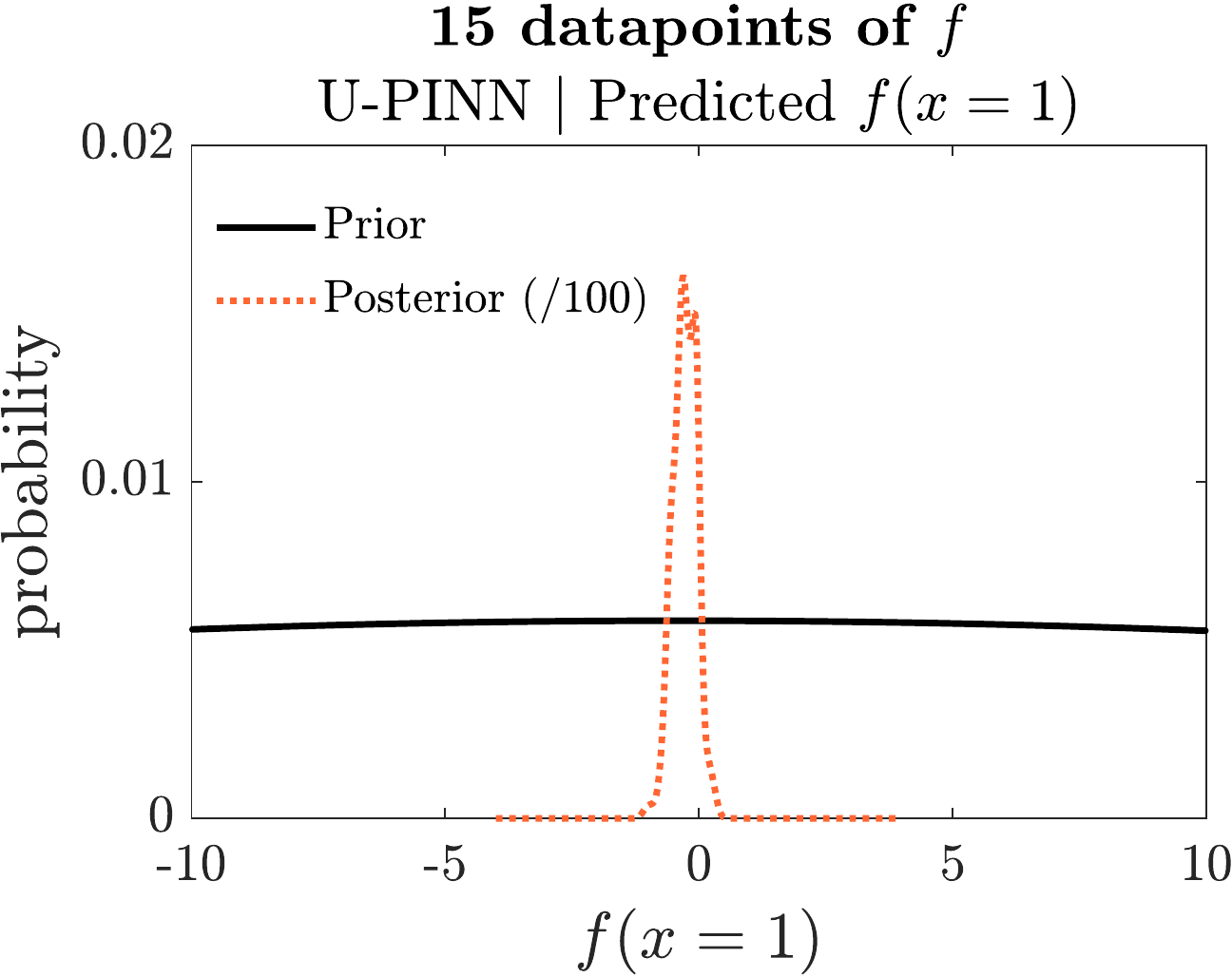}}
	\subcaptionbox{}{}{\includegraphics[width=0.32\textwidth]{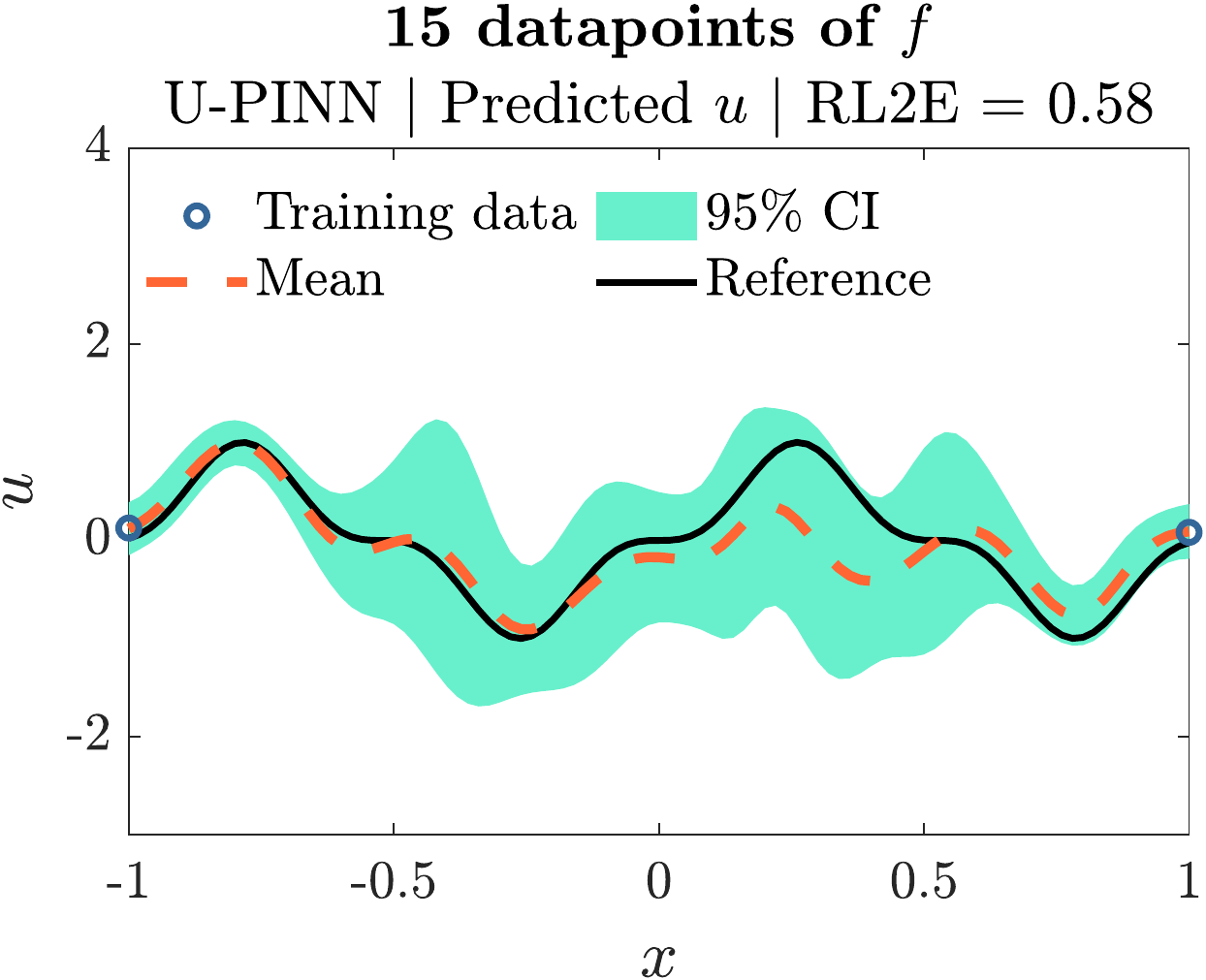}}
	\subcaptionbox{}{}{\includegraphics[width=0.32\textwidth]{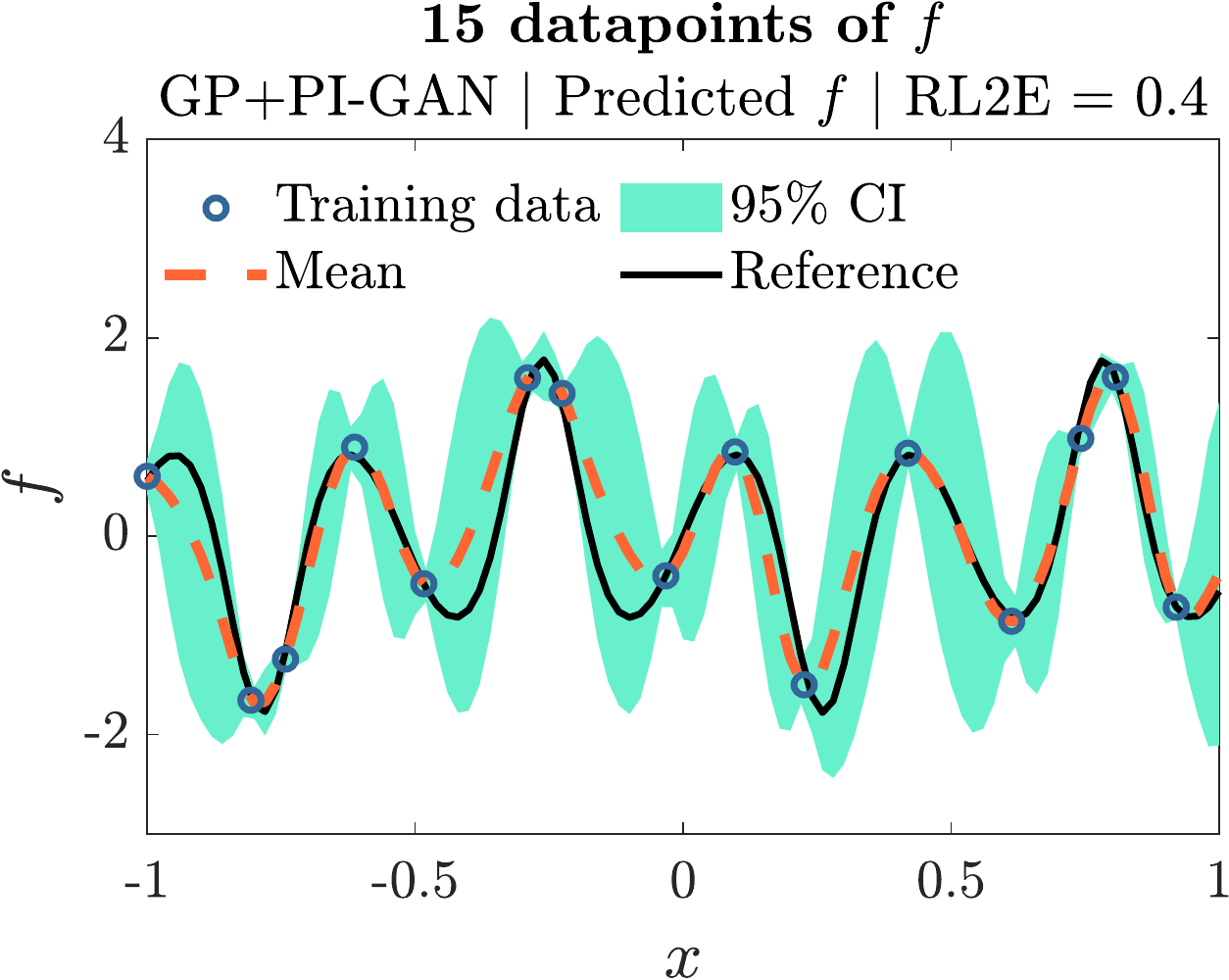}}
	\subcaptionbox{}{}{\includegraphics[width=0.32\textwidth]{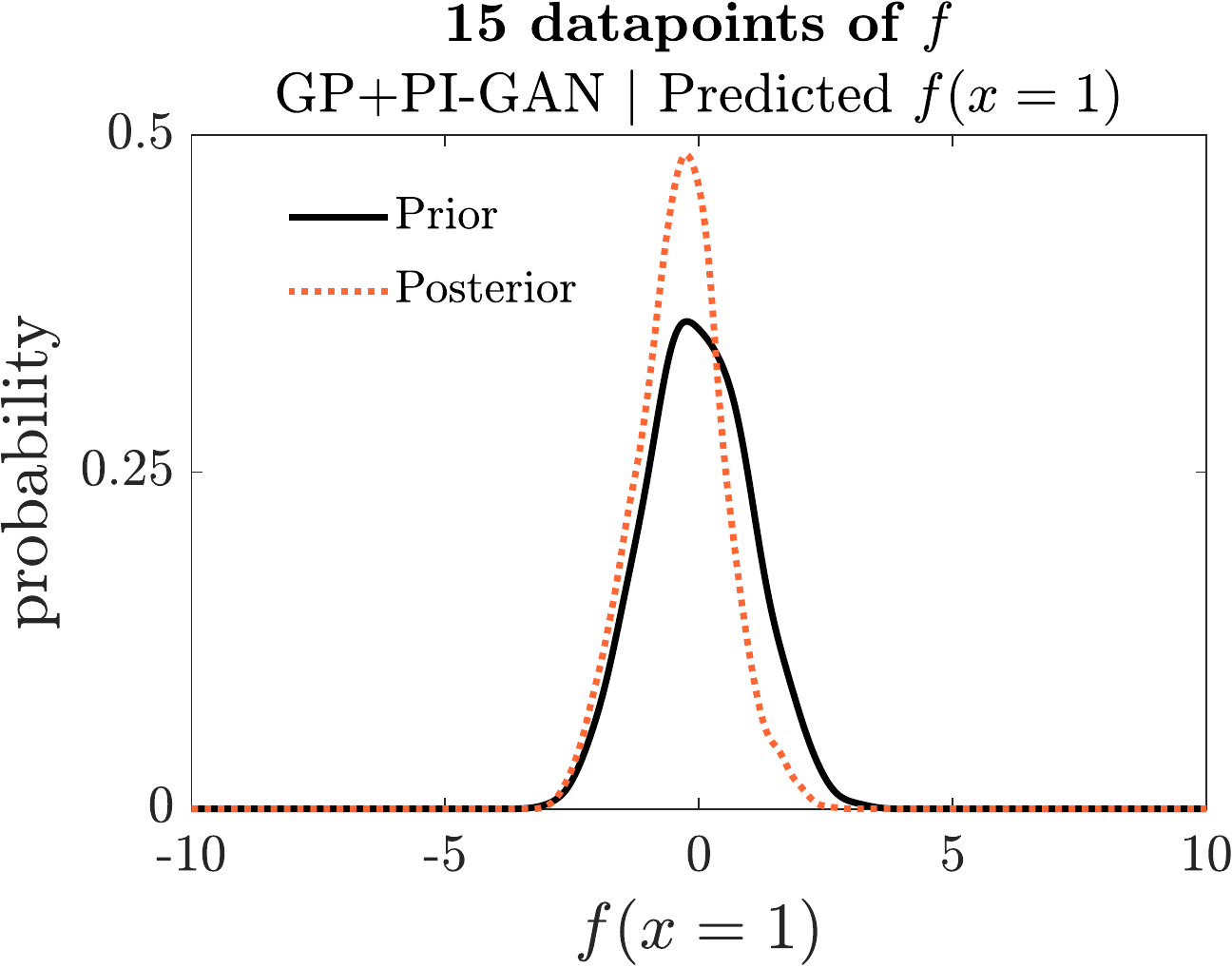}}
	\subcaptionbox{}{}{\includegraphics[width=0.32\textwidth]{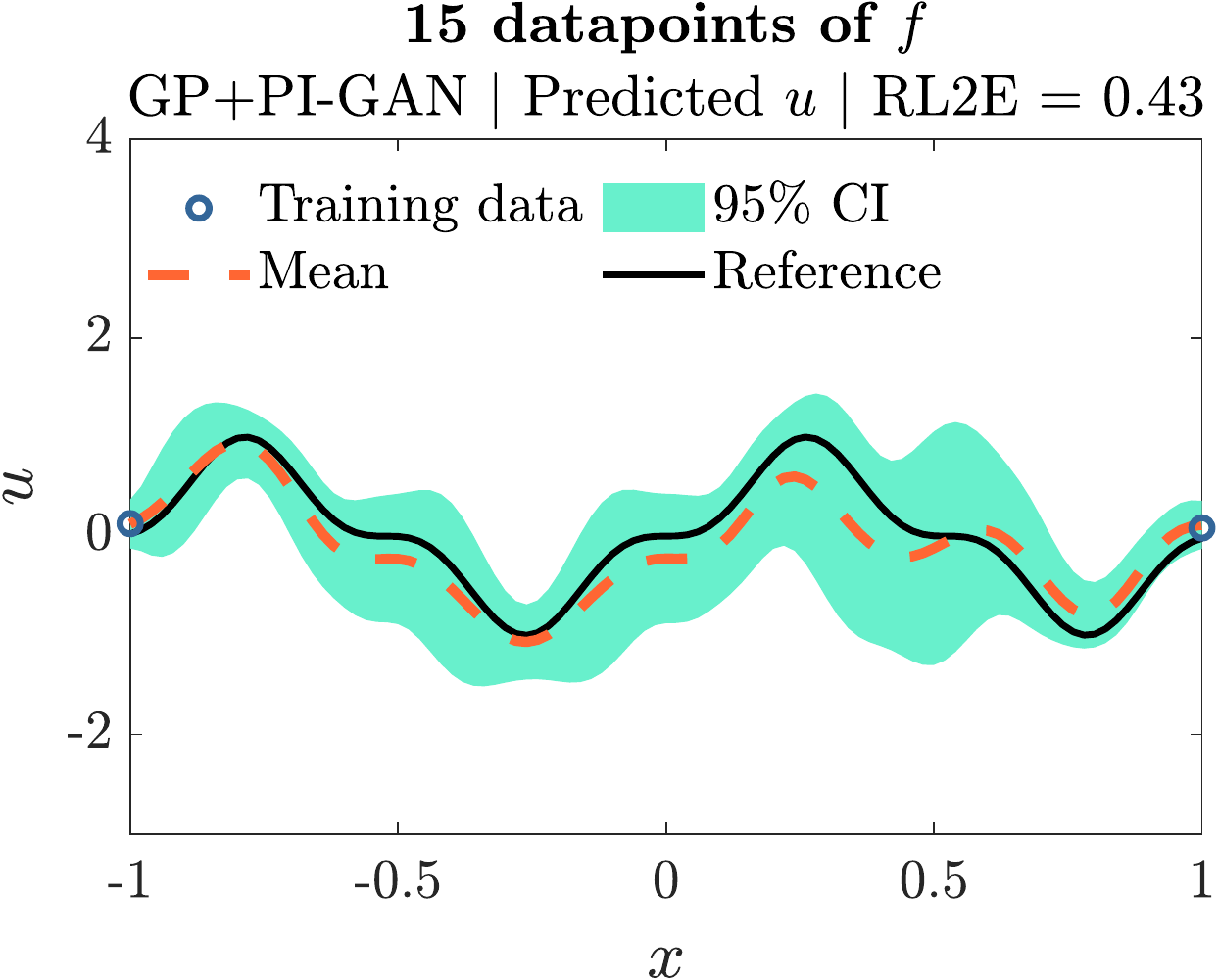}}
	\caption{
		Forward PDE problem of Eq.~\eqref{eq:comp:pinns:forw:pde} | $N_f=15$ \textit{datapoints of} $f$: we compare the U-PINN, which fits $f$ and $u$ simultaneously, with the herein proposed GP+PI-GAN, which first fits $f$ and then $u$.
		Note that because there is no $f$ datapoint at $x=1$ and fitting $f$ does not use the PDE of Eq.~\eqref{eq:comp:pinns:forw:pde}, the GP prior used for $f$ is similar to the posterior in the case of GP+PI-GAN.
		In contrast, the posterior of U-PINN concentrates significantly because $u$ and $f$ are coupled.
		Shown here are the training data, reference functions, as well as the mean and epistemic uncertainty ($95 \%$ CI) predictions of U-PINN (top) and GP+PI-GAN (bottom) for $u$ and $f$.
		The middle panels include the prior and posterior distributions of $f$ evaluated at $x=1$, i.e., at the right edge of the input domain where there is no available data for $f$. 
	}
	\label{fig:comp:forw:pinns:15}
\end{figure}

We solve the problem using a standard U-PINN, i.e., with a BNN-FP, and HMC for posterior inference, similarly to Section~\ref{sec:comp:pinns}.
The mean and total  uncertainty of $f(x)$ are shown in Figs.~\ref{fig:comp:forw:pinns:15}a ($N_f=15$) and \ref{fig:comp:forw:pinns:6}a ($N_f=6$), while the corresponding results for $u(x)$ are shown in Figs.~\ref{fig:comp:forw:pinns:15}c ($N_f=15$) and \ref{fig:comp:forw:pinns:6}c ($N_f=6$). 
To implement GP+PI-GAN, we first fit the 15 or 6 measurements of $f$ using a GP with a squared exponential kernel (Section~\ref{app:methods:gps}). Based on this GP, we produce 10,000 realizations of $f(x)$ at 101 equidistant locations in the domain of $x$.
The statistics of these realizations are shown in Figs.~\ref{fig:comp:forw:pinns:15}d ($N_f=15$) and \ref{fig:comp:forw:pinns:6}d ($N_f=6$).
Subsequently, we treat the problem as stochastic and employ a PI-GAN to solve it (Section~\ref{sec:uqt:sdes}). 
Following the training of PI-GAN, we produce 1,000 realizations of $u(x)$ and plot the statistics in Figs.~\ref{fig:comp:forw:pinns:15}f ($N_f=15$) and \ref{fig:comp:forw:pinns:6}f ($N_f=6$). 
The uncertainty of $u$ in Figs.~\ref{fig:comp:forw:pinns:15} and \ref{fig:comp:forw:pinns:6} can also be construed as epistemic uncertainty, because it includes the uncertainty due to the limited and noisy data of $f$.
However, it does not include the uncertainty due to fitting a PI-GAN; i.e., we trained PI-GAN only once, in contrast to U-PI-GAN of Section~\ref{sec:comp:stochastic}, where we trained and combined three PI-GANs.
Further, in the middle panels of Figs.~\ref{fig:comp:forw:pinns:15} and \ref{fig:comp:forw:pinns:6} we present the prior and posterior distributions of $f$ evaluated at $x=1$, i.e., at the right edge of the input domain where there is no available data for $f$. 
For U-PINN, the prior and posterior distributions are obtained using samples from the prior $\cN(0, 1)$ for the parameters $\theta$ of the NN and HMC samples from the posterior, respectively.
For GP+PI-GAN, they are obtained using samples from the GP prior and posterior.

In Fig.~\ref{fig:comp:forw:pinns:15} pertaining to $N_f=15$ datapoints of $f$, GP+PI-GAN outperforms U-PINN in terms of accuracy of the mean (RL2E).
In Fig.~\ref{fig:comp:forw:pinns:15}e pertaining to GP+PI-GAN, it is observed that because there is no $f$ datapoint at $x=1$ and fitting $f$ does not use the PDE of Eq.~\eqref{eq:comp:pinns:forw:pde}, the GP prior used for $f$ is similar with the posterior.
In contrast, the U-PINN posterior in Fig.~\ref{fig:comp:forw:pinns:15}b concentrates significantly because $u$ and $f$ are coupled. 
In Fig.~\ref{fig:comp:forw:pinns:6} pertaining to $N_f=6$ datapoints of $f$, we observe that the uncertainty of $u$ is underestimated by GP+PI-GAN, because the approximations of $u$ and $f$ are not coupled through the PDE of Eq.~\eqref{eq:comp:pinns:forw:pde}.
In contrast, in U-PINN the approximation of $u$ affects the approximation of $f$ and in this case the uncertainties of both $u$ and $f$ increase together. 

Overall, in this section we compared two techniques, namely U-PINN and the herein proposed GP+PI-GAN, for solving a forward PDE problem. We found that GP+PI-GAN, which first fits the source term data and subsequently propagates the uncertainty to the solution, can outperform U-PINNs in some cases.
Nevertheless, because it does not use the PDE for fitting the source term data, it can be over-confident in its predictions.
Further study is required to establish these new results. 
To illustrate the fact that, in addition to data, U-PINN also uses the PDE to fit the source term, we provided the prior and posterior distributions
of the source term evaluated at a location in the domain without available data, using both methods.
Although the GP posterior is similar to the prior in the case of GP+PI-GAN, the posterior of U-PINN concentrates significantly because of the coupling of the solution and the source term.

%% file: IN_DON.tex
Here we consider a two-dimensional Darcy flow in porous media, which is described by
\begin{align}\label{eq:comp:don:pde}
	\nabla \cdot (\lambda(x, y; \xi) \nabla u(x, y; \xi)) =  f, ~ 0 \le x, y \le 1, \xi \in \Xi, 
\end{align}
where $x, y$ are the space coordinates, the sought solution $u(x, y; \xi)$ denotes the hydraulic head, and we consider $f = 50$.
Further, $\lambda(x, y; \xi)$ denotes the hydraulic conductivity field and is determined by the pore structure. 
Furthermore, $\xi\in \Xi$ is an instance of the problem corresponding to a given field $\lambda(x, y, \xi)$.
We consider the boundary conditions
\begin{subequations}\label{eq:comp:don:bcs}
	\begin{align}
		u(x=0, y; \xi) = 1, ~ u(x=1, y; \xi) = 0, \\
		\partial_{\boldsymbol{n}} u(x, y=0; \xi) = \partial_{\boldsymbol{n}} u(x, y=1; \xi) = 0, \forall \xi \in \Xi,
	\end{align}
\end{subequations}
where $\boldsymbol{n}$ denotes the unit normal vector of the boundary.
In this example, we consider a stochastic model for $\lambda(x, y; \xi)$, accounting for different porous media. In particular, $\lambda(x, y; \xi) = \exp(\tilde{\lambda}(x, y; \xi))$, where $\tilde{\lambda}(x, y; \xi)$ is a truncated Karhunen-Lo\`{e}ve expansion of a GP with zero mean and kernel given by
\begin{subequations}\label{eq:comp:don:gp}
	\begin{align}
		k_{\tilde{\lambda}}(x, y, x', y') = \exp(-\frac{(x - x')^2}{2l^2} - \frac{(y - y')^2}{2l^2}), \\
		0 \le x, y, x', y' \le 1, ~ l = 0.25. 
	\end{align}
\end{subequations}
In the following, we only keep the first 100 leading terms of the expansion. 

\begin{table}[!ht]
	\centering
	\footnotesize
	\begin{tabular}{c|r|l|c}
		\toprule
		\multirow{5}{*}{\textbf{a}}&\multicolumn{3}{c}{\textbf{Pre-training phase of operator learning}}\\
		\cline{2-4}
		&\textbf{Problem case} & \hfil\textbf{Solution approach}\hfil & \textbf{Tested in \S}\\
		\cline{2-4}
		&Clean $\lambda$ and clean $u$ & (U-)DeepONet&\ref{sec:comp:don:fp}-\ref{sec:comp:don:dens}\\
		\cline{2-4}
		&Clean $\lambda$ and noisy $u$ & (U-)DeepONet&-\\
		\cline{2-4}
		&Noisy $\lambda$ and noisy $u$ & Future work &-\\
		\midrule
		\midrule
		\multirow{8}{*}{\textbf{b}}&\multicolumn{3}{c}{\textbf{Inference phase of operator learning}}\\
		\cline{2-4}
		&\textbf{Problem case} &\hfil \textbf{Solution approach}\hfil & \textbf{Tested in \S}\\
		\cline{2-4}
		&Clean $\lambda$ only & Perform simple interpolation to reconstruct&\multirow{2}{*}{\ref{sec:comp:don:dens}}\\
		&(complete$^*$ or limited)& $\lambda$, if limited. Then pass $\lambda$ to (U-)DeepONet.& \\
		\cline{2-4}
		&Noisy $\lambda$ only & Perform Bayesian inference using a GP, a BNN, or a GAN to &\multirow{2}{*}{-}\\
		&(complete$^*$ or limited)  & obtain realizations of $\lambda$. Then pass them to (U-)DeepONet.&\\
		\cline{2-4}
		&Both $\lambda$ and $u$ & PA-BNN-FP or PA-GAN-FP: Reconstruct $\lambda$ and $u$, together.&\multirow{2}{*}{\ref{sec:comp:don:fp}}\\
		&(limited and noisy)& Works with DeepONet only, not U-DeepONet.&\\
		\bottomrule
	\end{tabular}
	\caption{
		Overview of problem cases during pre-training (a) and inference (b) phases of operator learning using DeepONet, as well as pertinent solution approaches.
		The sections for each problem case and solution approach that are studied in this paper are provided in the third column.
		In the pre-training phase, we learn an operator mapping from $\lambda$ to $u$. To pre-train (U-)DeepONet, for each $\lambda$ in the pre-training dataset $u$ can be clean or noisy, whereas $\lambda$ must be clean and available on a pre-specified grid (``complete''). Interpolation is performed if $\lambda$ is not available on a grid (``limited'').
		A future research direction pertains to extending DeepONet for treating cases of noisy $\lambda$ during pre-training.
		In the inference phase, we employ the learned operator to make inferences given new data.
		Different approaches can be used depending on the quantity and quality of new data.
		See Section~\ref{sec:intro:problem:form} for more information regarding formulation of the operator learning problem.
		Note that the cases marked with an asterisk correspond to the special case described in Section~\ref{sec:intro:problem:form}. 
	}
	\label{tab:comp:don:problems}
\end{table}

Eqs.~\eqref{eq:comp:don:pde}-\eqref{eq:comp:don:bcs} can be viewed as a special case of Eq.~\eqref{eq:intro:piml:pinn:pde}, with each $\xi \in \Xi$ being a different random event corresponding to a different sample from the GP of $\lambda$; with $\pazocal{F}$ and $\pazocal{B}$ both considered unknown and given by Eqs.~\eqref{eq:comp:don:pde} and \eqref{eq:comp:don:bcs}, respectively; and with $x$ given by $(x, y)$.
The objective of this example is to compare different DeepONet-based UQ methods for solving an operator learning problem involving Eqs.~\eqref{eq:comp:don:pde}-\eqref{eq:comp:don:bcs}.
That is, for learning the operator mapping from $\lambda$ to $u$.
To this end, we consider the pre-training dataset of problem 4 in Table~\ref{tab:problem:form}, consisting of $N$ = 1,000 realizations of $\lambda$ and $u$.
Instead of using the original data of $\lambda$, we transform them using $\tilde{\lambda} = \log \lambda$, which corresponds to the GP of Eq.~\eqref{eq:comp:don:gp}.
We also standardize the data for each $(x, y)$ and for each of the quantities $\tilde{\lambda}$ and $u$, by subtracting the mean and dividing by the standard deviation, computed for each $(x, y)$ using the $N$ = 1,000 realizations.
The modified pre-training dataset is, thus, expressed as $\cD = \{\cD_{\hat{\lambda}}, \cD_{\hat{u}}\}$, where $\cD_{\hat{\lambda}} = \{\hat{\Lambda}_i\}_{i=1}^N$, $\cD_{\hat{u}} = \{\hat{U}_i\}_{i=1}^N$, each $\hat{\Lambda}_i$ is a transformed and standardized realization of $\lambda$, and each $\hat{U}_i$ is a standardized realization of $u$. 
Note that all plots in this section correspond to the results obtained for the standardized versions of $\tilde{\lambda} = \log\lambda$ and $u$.

We consider two problem scenarios, as described in Section~\ref{sec:intro:problem:form}, depending on the application of the learned operator for making inferences given new data, i.e., given inference data.
These are: a) a general case, which pertains to limited and noisy inference data of both $\lambda$ and $u$; and b) a special case, which pertains to complete data of only $\lambda$ for predicting the corresponding $u$.
Recall that DeepONet takes as input the function $\lambda$ in the branch net in the form of values on a pre-specified grid, as well as $(x, y)$ for predicting the corresponding $u(x,y)$ (see Fig.~\ref{fig:intro:piml:don:arch}).
Therefore, a complete or limited dataset refers to the case where $\lambda$ is available or not on the aforementioned pre-specified grid, on which the DeepONet has been pre-trained on, respectively.
We summarize the problem cases as well as the corresponding solution approaches in Table~\ref{tab:comp:don:problems}.
In passing, note that to obtain the reference solutions and the training data presented in the following, we use the \textit{Matlab PDE Toolbox}.
The hyperparameters and the NN architectures that we used are summarized in Section~\ref{app:comp:hyperparameters} and \ref{app:comp:architecture}, respectively.

\subsubsection{Physics-agnostic functional prior: combining a pre-trained DeepONet with BNN and GAN priors for treating noisy inference data}\label{sec:comp:don:fp}

In this section, we first use a DeepONet to learn the operator mapping from $\lambda$ to $u$, given the pre-training dataset $\cD = \{\cD_{\lambda}, \cD_u\}$ and following Section~\ref{sec:intro:piml:don}.
Next, during inference, we assume that new data arrives corresponding to an unseen random event $\xi' \in \Xi$, containing limited and noisy samples from $\lambda$ and $u$.
We denote these two noisy datasets by $\cD_{\lambda}' = \{x_i, y_i, \lambda_{i}\}_{i=1}^{N_{\lambda}'}$ and $\cD_u' = \{x_i, y_i, u_{i}\}_{i=1}^{N_u'}$.
The numbers of datapoints are $N_{\lambda}' = 20$ and $N_u'=10$.
Clearly, the points $(x,y)$ in the space domains of $\lambda$ and $u$ are in general different.
The objective of this section is to evaluate two alternative techniques for reconstructing $u$ and $\lambda$ based on the new data.
Note that this section can be construed as an extension of a similar problem considered in \cite{meng2021learning}.

\begin{table}[!ht]
	\centering
	\footnotesize
	\begin{tabular}{c|cc|cc||cc|cc}
		\toprule
		\multirow{2}{*}{Metric ($\times 10^2$)}& \multicolumn{2}{c|}{$\log \lambda$ $\vert$ ID data} & \multicolumn{2}{c||}{$u$ $\vert$ ID data}  & \multicolumn{2}{c|}{$\log \lambda$ $\vert$ OOD data} & \multicolumn{2}{c}{$u$ $\vert$ OOD data}\\
		\cline{2-5}\cline{5-9}
		& BNN & GAN& BNN & GAN  & BNN & GAN& BNN & GAN \\
		\midrule
		RL2E ($\downarrow$) & 26.7 & \textbf{16.3} & 14.6 & \textbf{7.7} & 49 & \textbf{42.8} & 48 & \textbf{35.5} \\ 
		MPL ($\uparrow$) & 91 & \textbf{154.8} & 168.8 & \textbf{217} & 65.6 & \textbf{131.2} & 99 & \textbf{127.4} \\ 
		RMSCE ($\downarrow$) & \textbf{8} & \textbf{8} & \textbf{2} & 8.6 & \textbf{5.1} & 11.4 & \textbf{9.7} & 20.3 \\ 
		\bottomrule
	\end{tabular}
	\caption{
		Operator learning problem of Eqs.~\eqref{eq:comp:don:pde}-\eqref{eq:comp:don:bcs} | \textit{Limited and noisy inference data of $\lambda$ and $u$ (ID or OOD)}:
		combining a pre-trained DeepONet with a pre-trained GAN using historical data performs better than combining with a BNN.
		Note, however, that combining with a BNN does not incur any additional computational cost and also the results for $u$ are more calibrated (smaller RMSCE), even for ID data.
		This can be attributed to the fact that we pre-trained GAN using historical data of $\lambda$ only.
		Shown here is the performance evaluation of BNN-FP (proposed herein) and GAN-FP, referred to as PA-BNN-FP and PA-GAN-FP, respectively, when combined with a pre-trained DeepONet.
		R2LE was calculated based on clean reference data, whereas MPL and RMSCE based on noisy data.
	}
	\label{tab:comp:don:fp}
\end{table}

\begin{figure}[!ht]
	\centering
	\subcaptionbox{}{}{\includegraphics[width=0.32\textwidth]{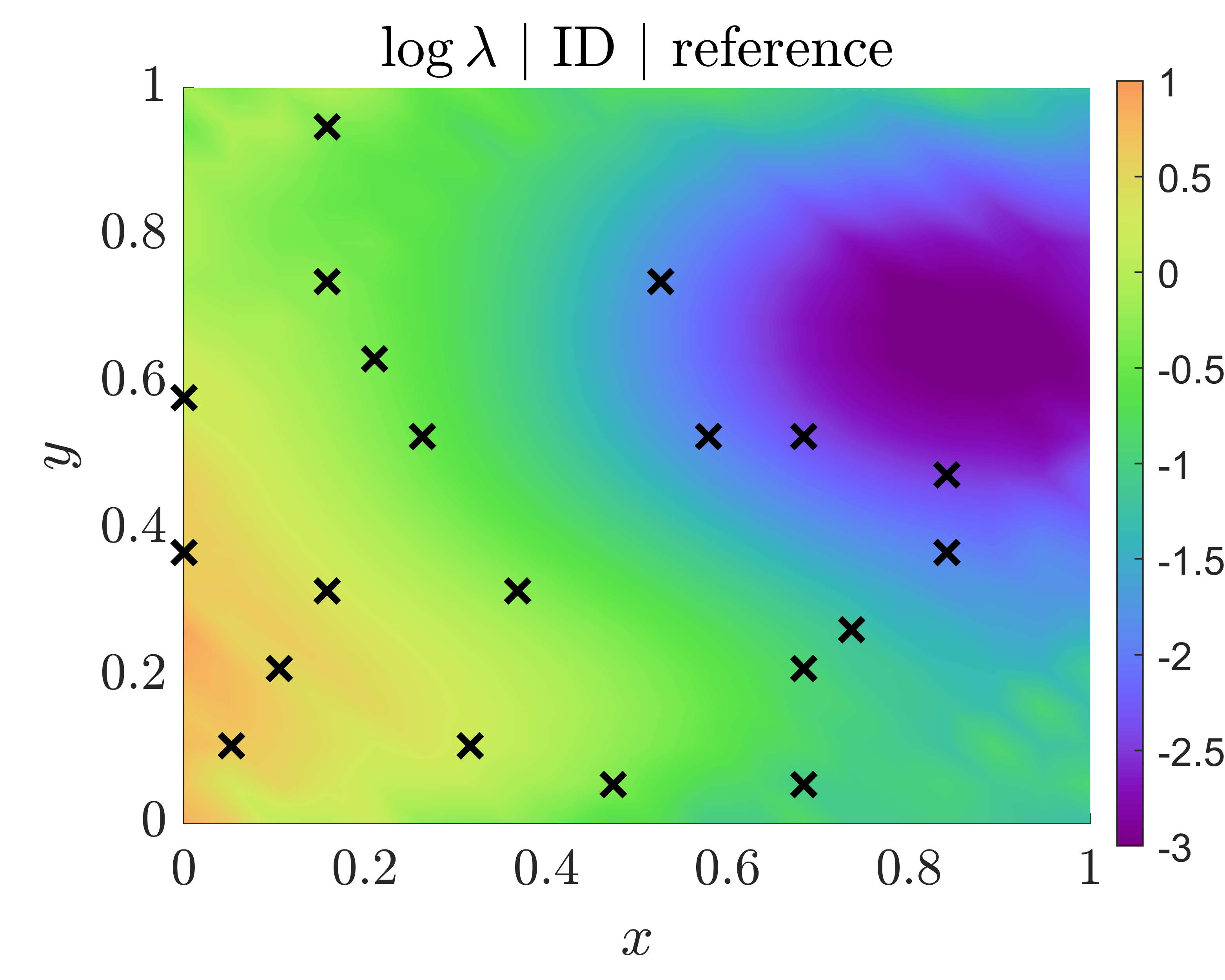}}
	\subcaptionbox{}{}{\includegraphics[width=0.32\textwidth]{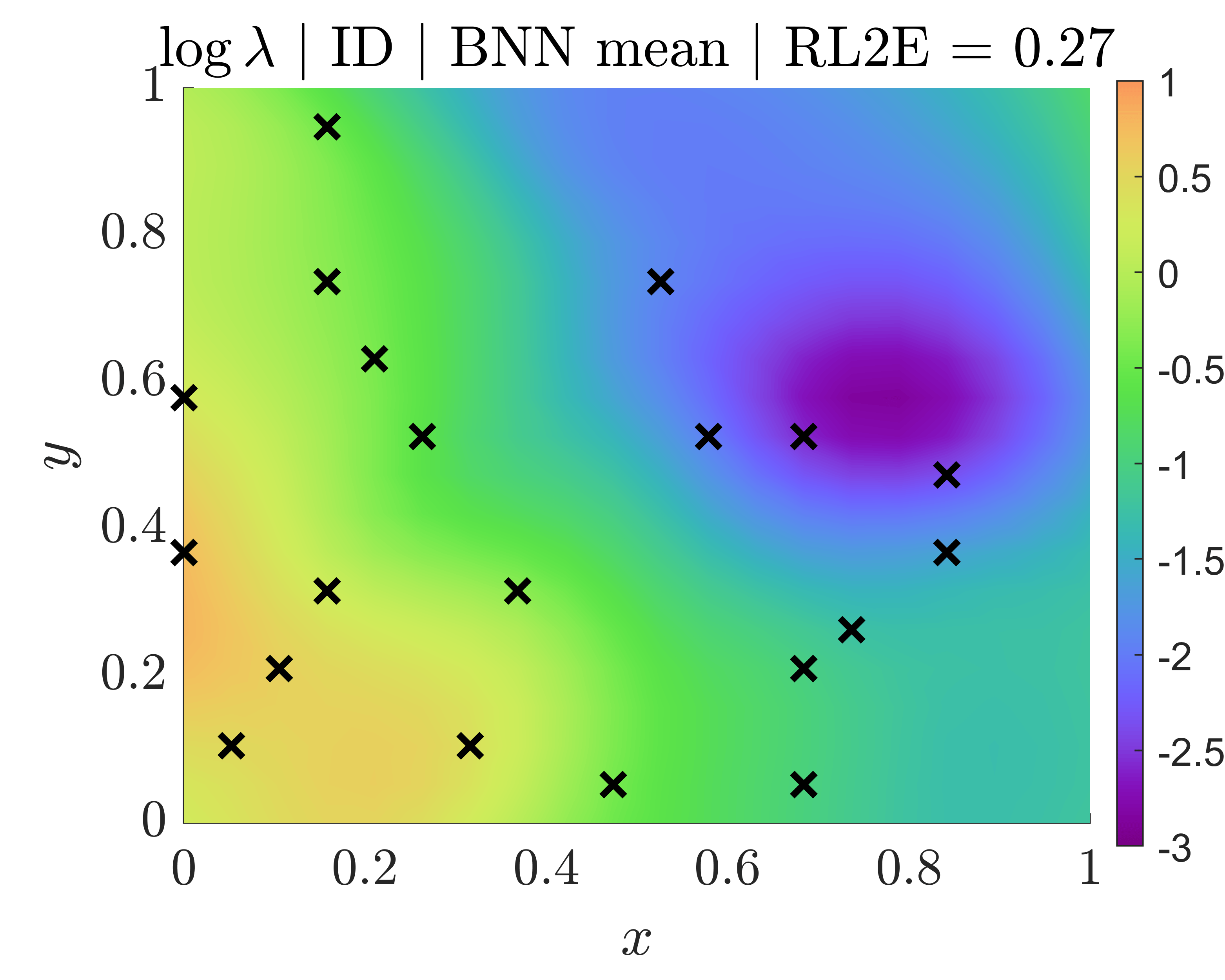}}
	\subcaptionbox{}{}{\includegraphics[width=0.32\textwidth]{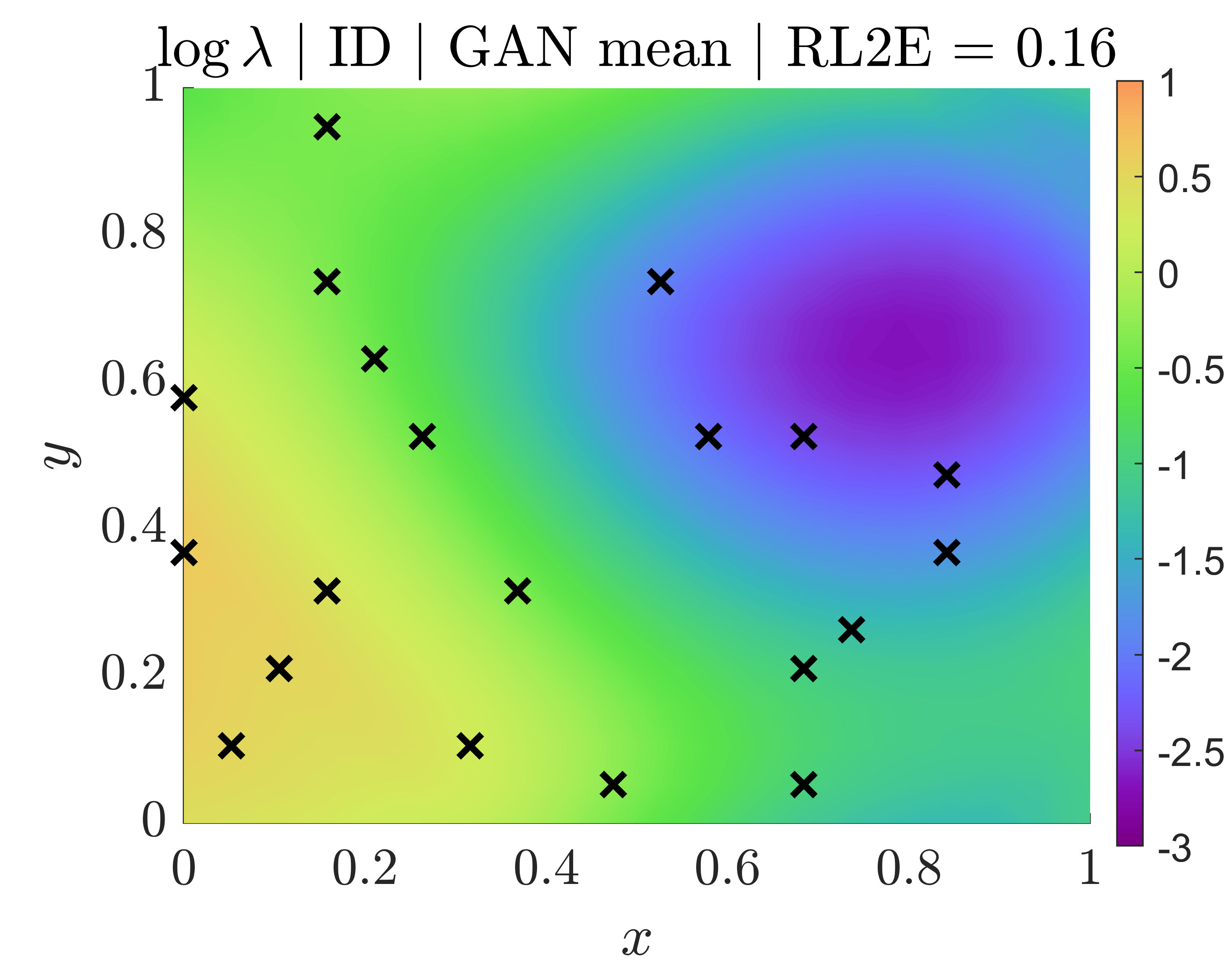}}
	\subcaptionbox{}{}{\includegraphics[width=0.32\textwidth]{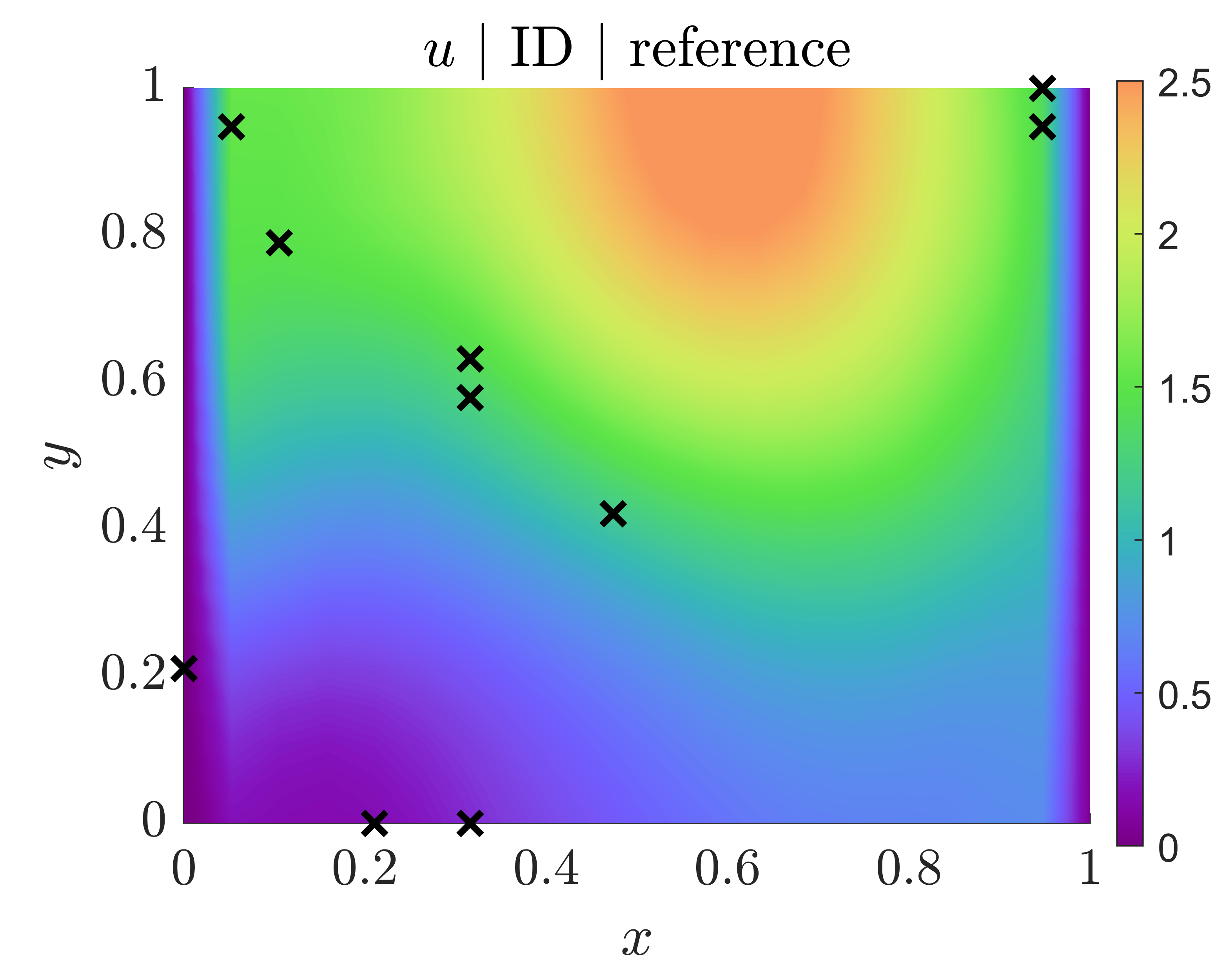}}
	\subcaptionbox{}{}{\includegraphics[width=0.32\textwidth]{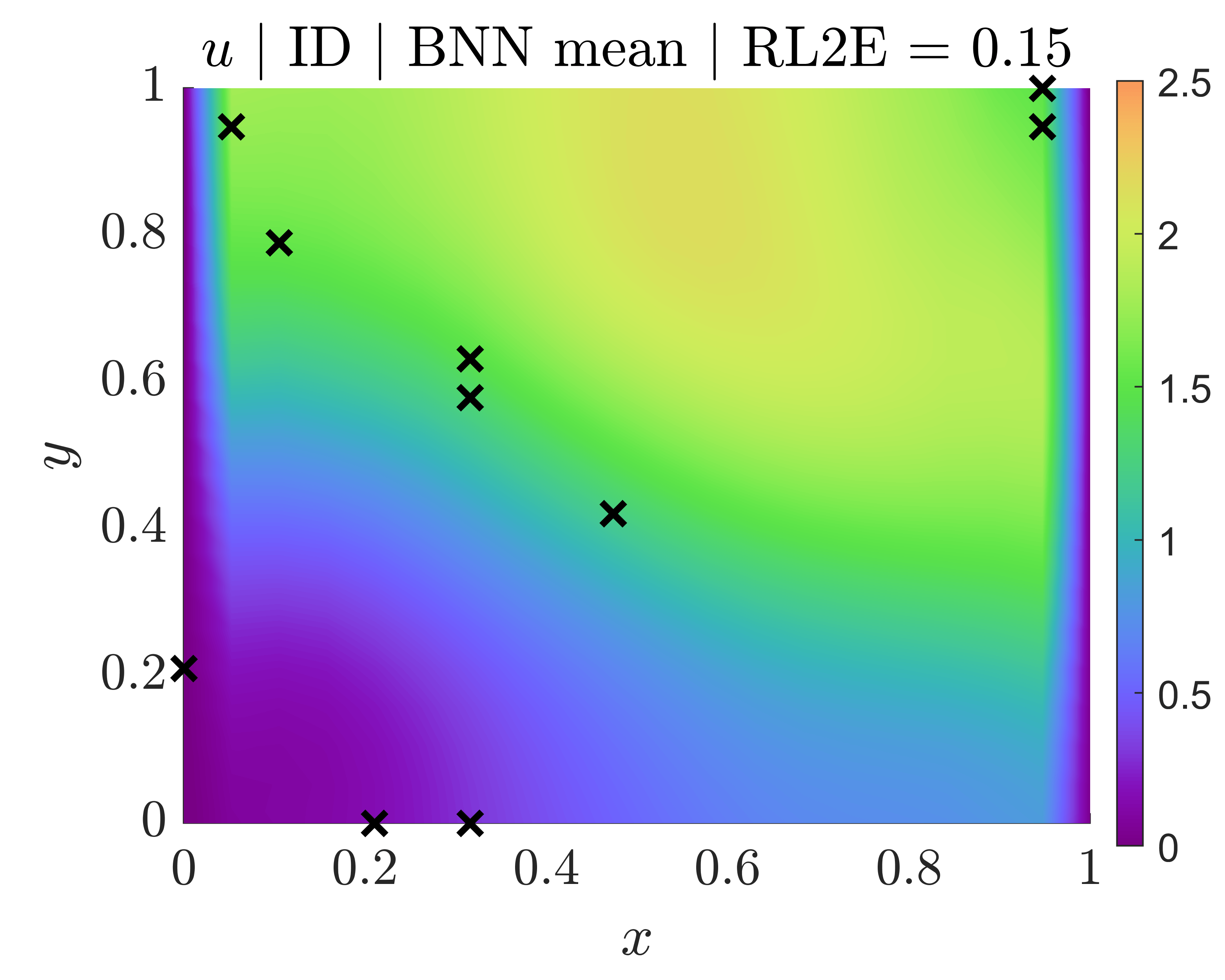}}
	\subcaptionbox{}{}{\includegraphics[width=0.32\textwidth]{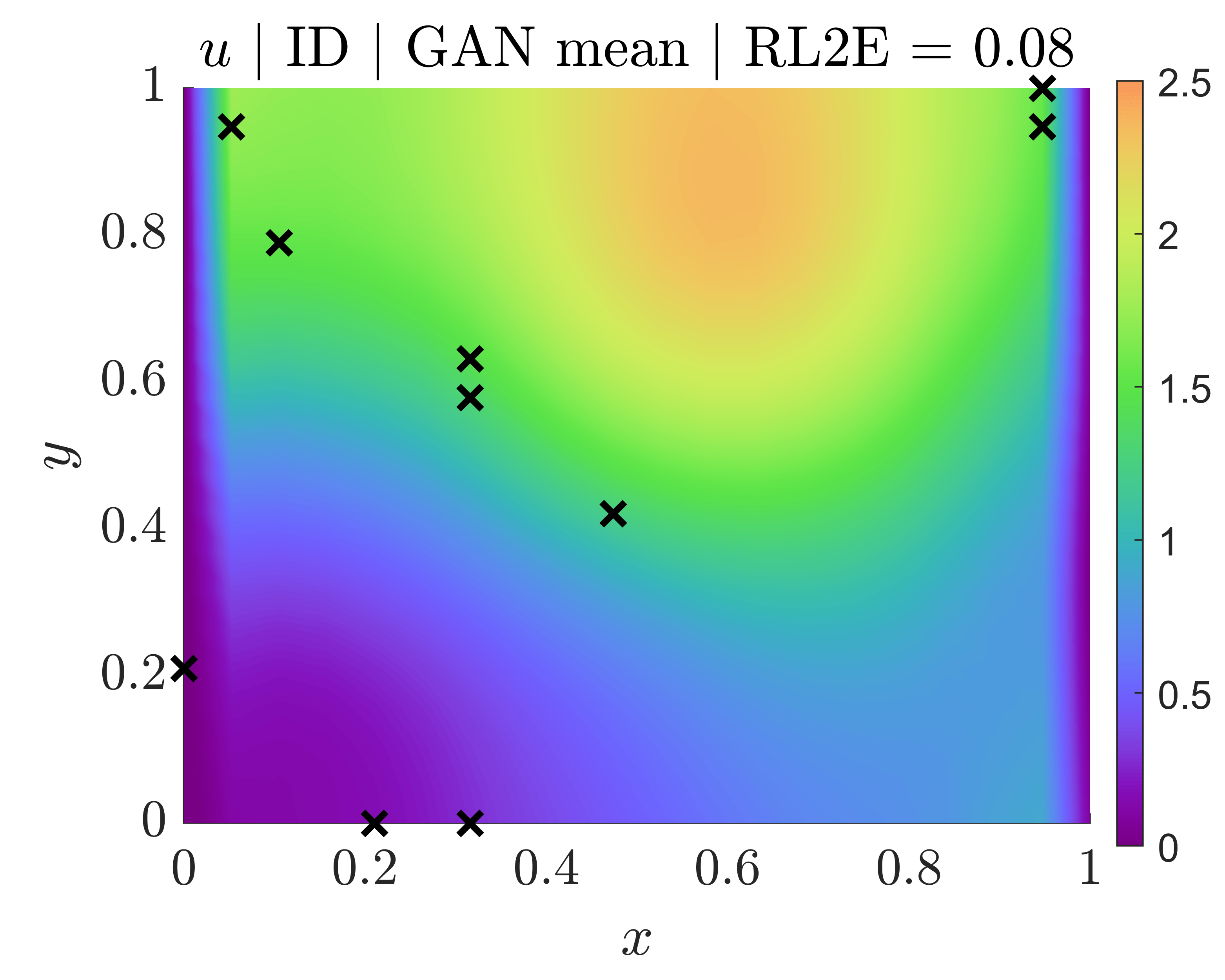}}
	\caption{
		Operator learning problem of Eqs.~\eqref{eq:comp:don:pde}-\eqref{eq:comp:don:bcs} | \textit{Limited and noisy inference data of $\lambda$ and $u$ (ID)}:
		the mean predictions of PA-GAN-FP for $\lambda$ and $u$ are more accurate than the predictions of the herein proposed PA-BNN-FP.   
		Note, however, that PA-BNN-FP does not incur any additional computational cost and its prediction can be more calibrated as shown in Fig.~\ref{fig:comp:don:fp:uncer:id}.
		Shown here are the reference input $\tilde{\lambda} = \log \lambda$ and solution $u$, the corresponding inference data locations (x markers), as well as the mean predictions by PA-BNN-FP and PA-GAN-FP.
		Top row results correspond to input $\log \lambda$, whereas bottom row to solution $u$.
	}
	\label{fig:comp:don:fp:mean:id}
\end{figure}

We employ an approximator $\lambda_{\theta}$ for fitting the transformed and standardized data of $\lambda$, which is denoted by $\hat{\lambda}$. 
The output of $\lambda_{\theta}$ becomes the input to the pre-trained DeepONet and produces an approximator $u_{\theta}$ for fitting the standardized data of $u$, denoted by $\hat{u}$. 
This is illustrated in Fig.~\ref{fig:uqt:fpriors:FP}c, where $\lambda_{\theta}$ is denoted by $\pazocal{M}$.
Because of data standardization (and transformation), the modified inference dataset is expressed as $\cD' = \{\cD_{\hat{\lambda}}', \cD_{\hat{u}}'\}$, where $\cD_{\hat{\lambda}}' = \{x_i, y_i, \hat{\lambda}_{i}'\}_{i=1}^{N_{\lambda}'}$ and $\cD_{\hat{u}}' = \{x_i, y_i, \hat{u}_{i}'\}_{i=1}^{N_{u}'}$.
As explained in Section~\ref{app:comp:don:lognormal}, the transformation $\tilde{\lambda} = \log \lambda$ corresponds to a log-normal noise model for $\lambda$ and to a Gaussian noise model for $\tilde{\lambda}$. 
Because we also standardize the data, according to Section~\ref{app:comp:don:lognormal} the noise for the transformed and standardized $\hat{\lambda}$ is Gaussian and space-dependent, and is given by Eq.~\eqref{eq:results:lognormal:stats}.  
However, for ease of implementation we consider a constant variance for the data of $\hat{\lambda}$ with value equal to 0.01. This corresponds to approximately $10 \%$ of noise in the original data of $\lambda$ for all $(x,y)$ points (in a log-normal noise model). 
We also consider the same constant variance for the data of $\hat{u}$, which corresponds to variance approximately equal to 0.01 in the original data of $u$ for all $(x,y)$ points (in a Gaussian noise model).

\begin{figure}[!ht]
	\centering
	\subcaptionbox{}{}{\includegraphics[width=0.24\textwidth]{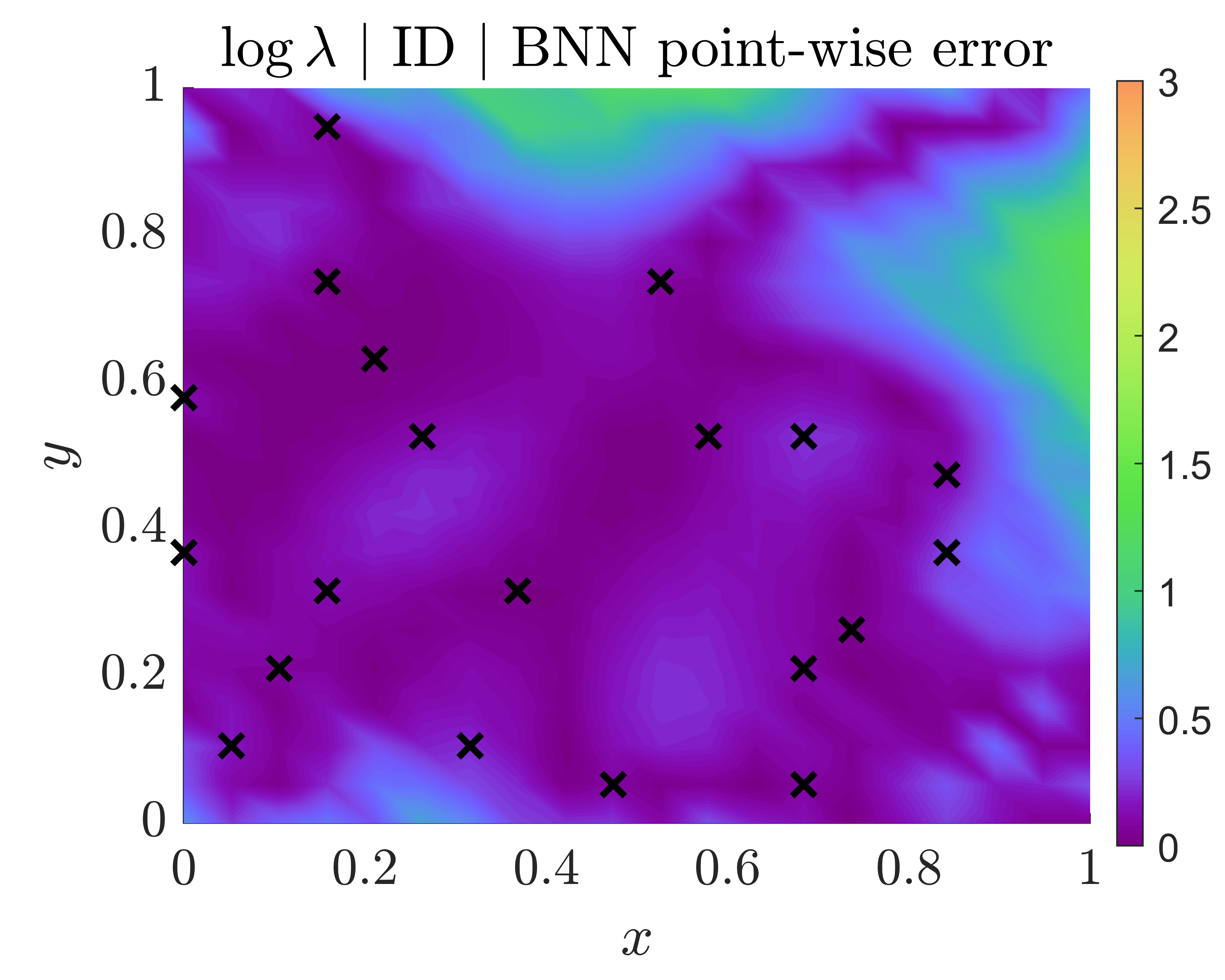}}
	\subcaptionbox{}{}{\includegraphics[width=0.24\textwidth]{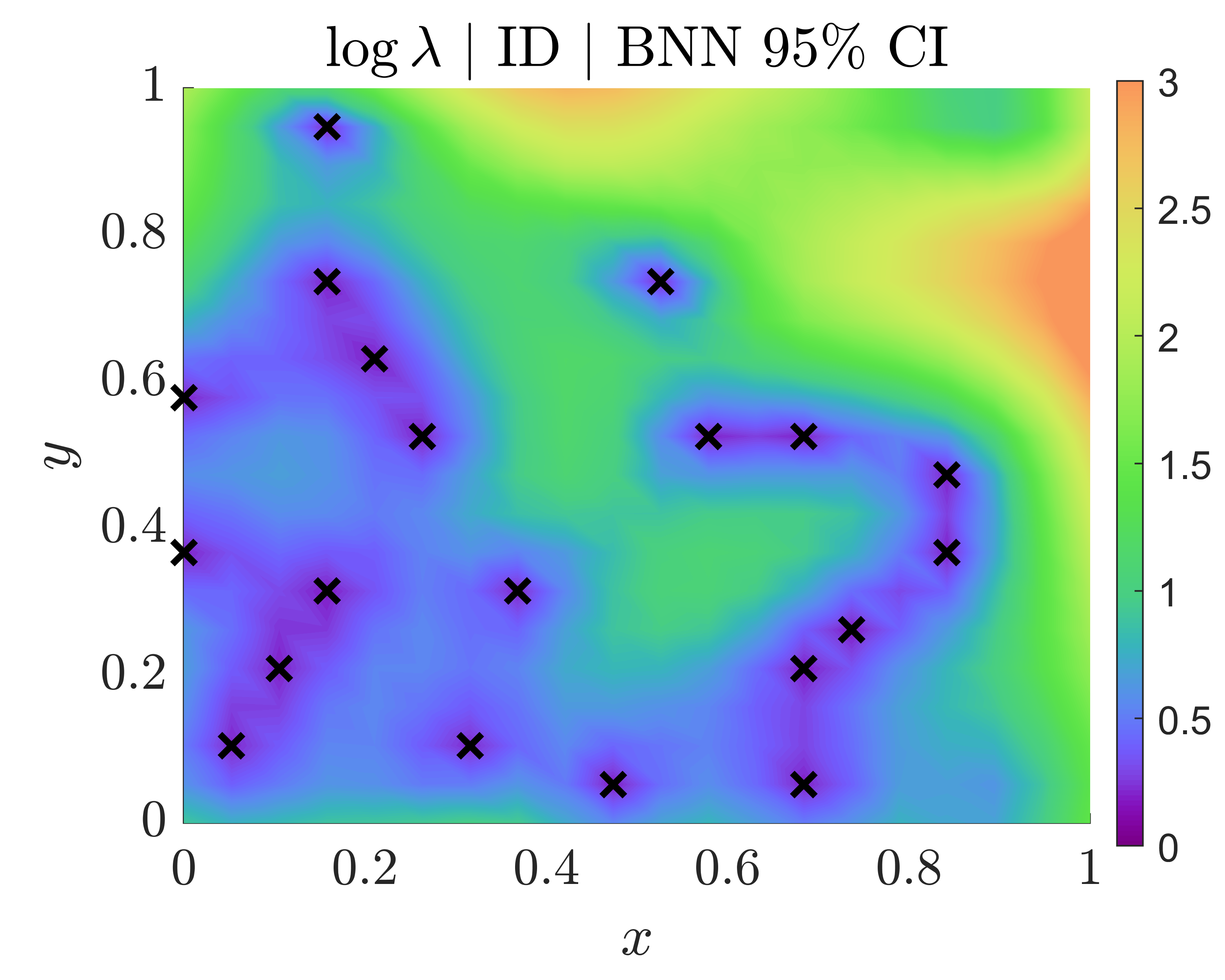}}
	\subcaptionbox{}{}{\includegraphics[width=0.24\textwidth]{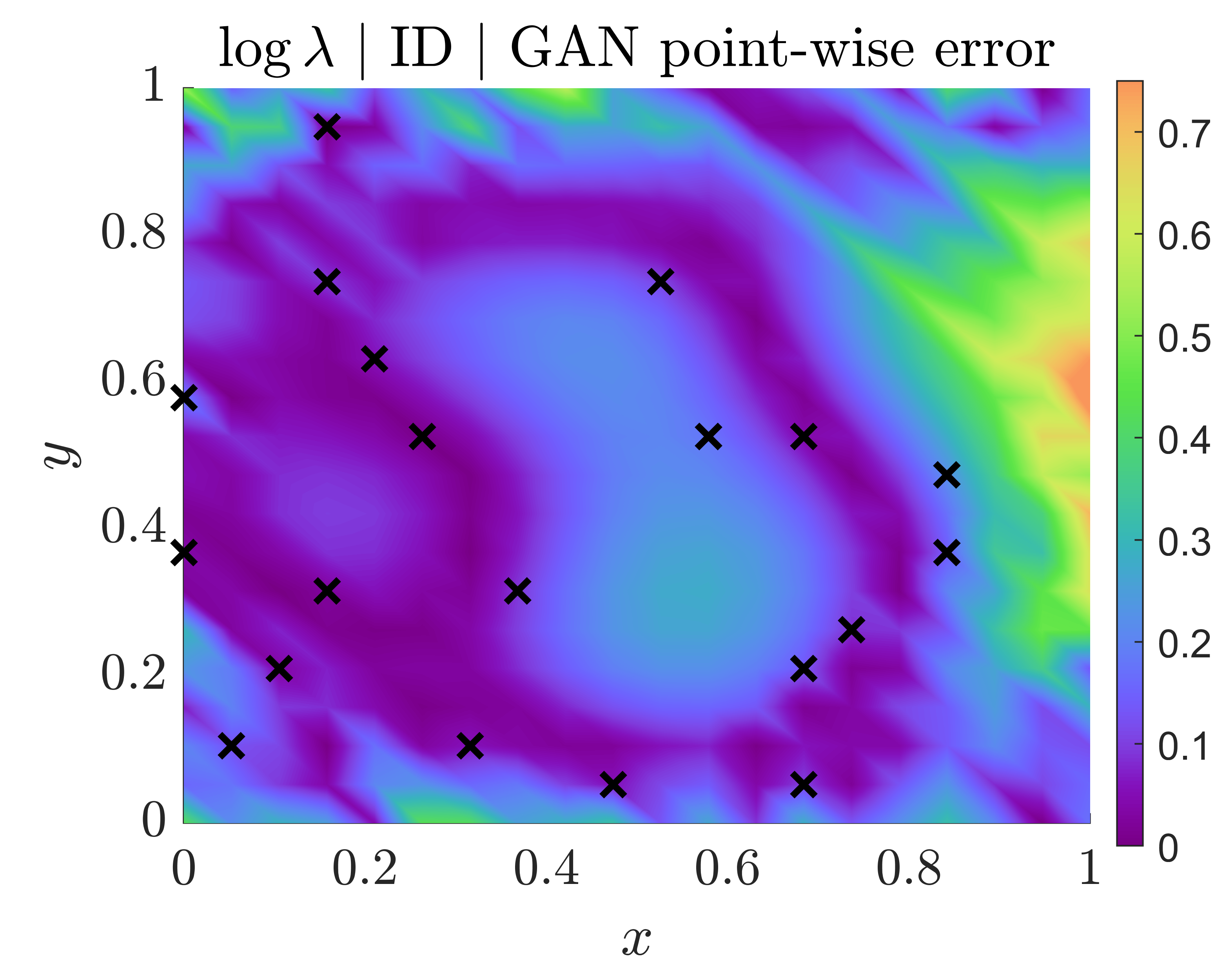}}
	\subcaptionbox{}{}{\includegraphics[width=0.24\textwidth]{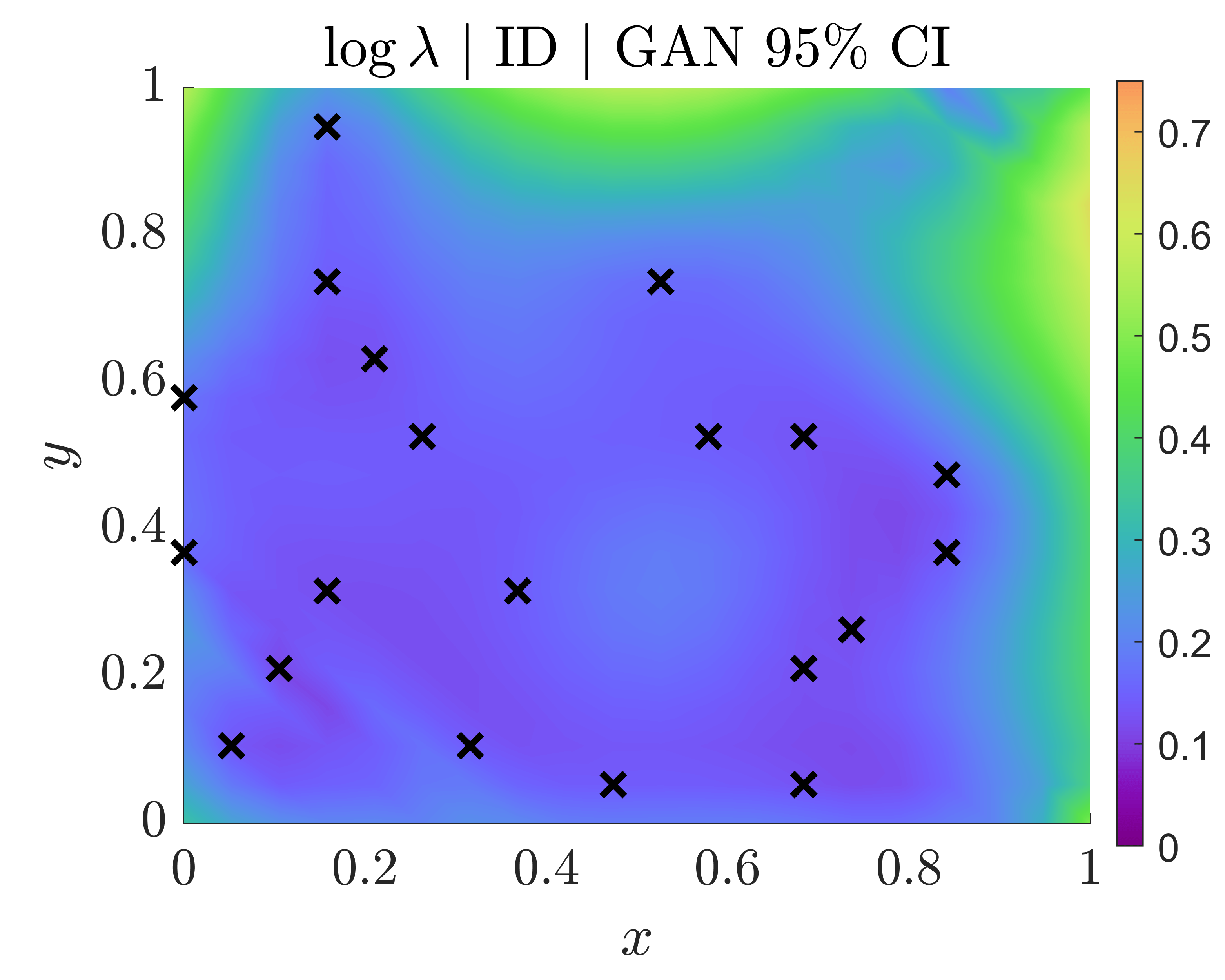}}
	\subcaptionbox{}{}{\includegraphics[width=0.24\textwidth]{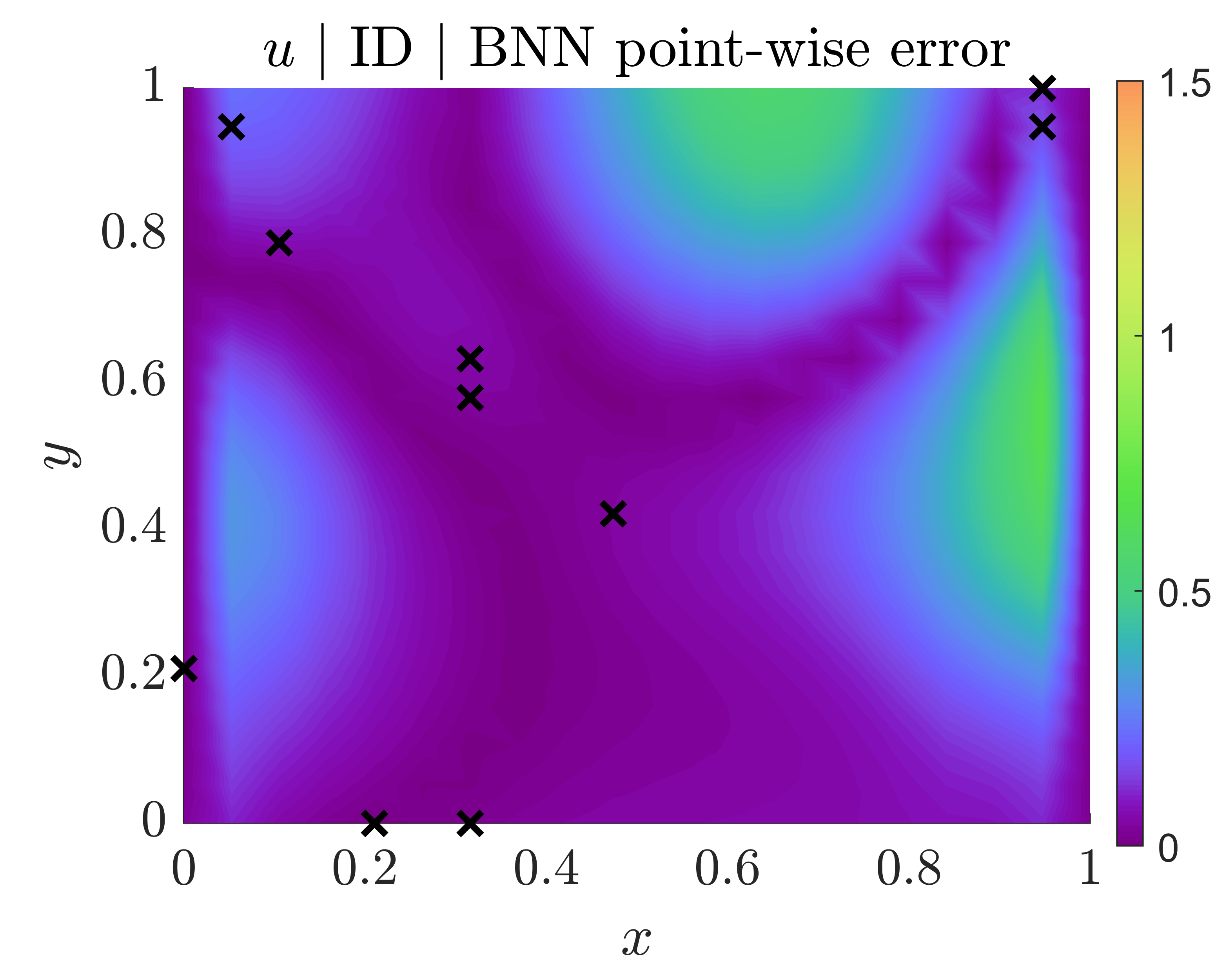}}
	\subcaptionbox{}{}{\includegraphics[width=0.24\textwidth]{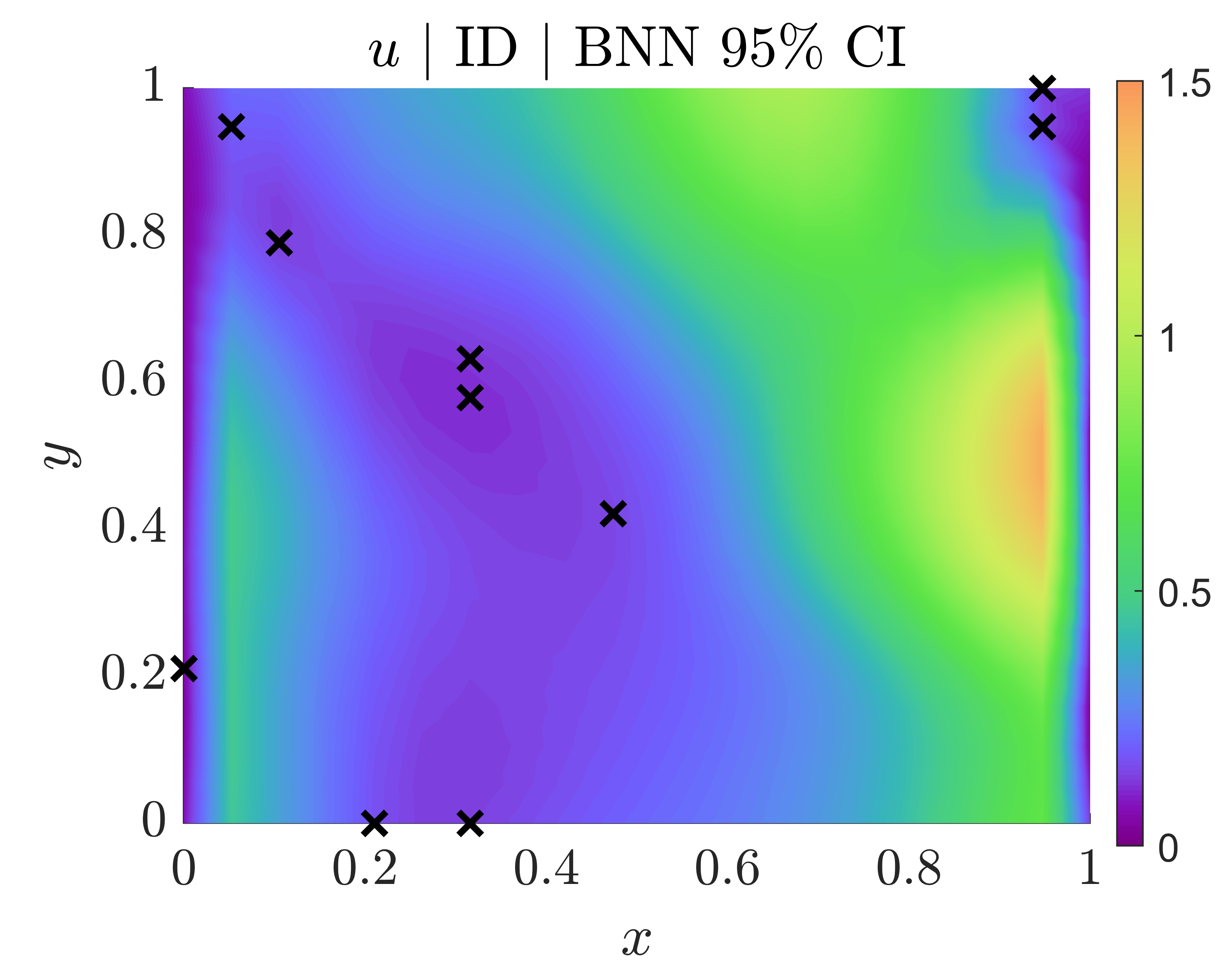}}
	\subcaptionbox{}{}{\includegraphics[width=0.24\textwidth]{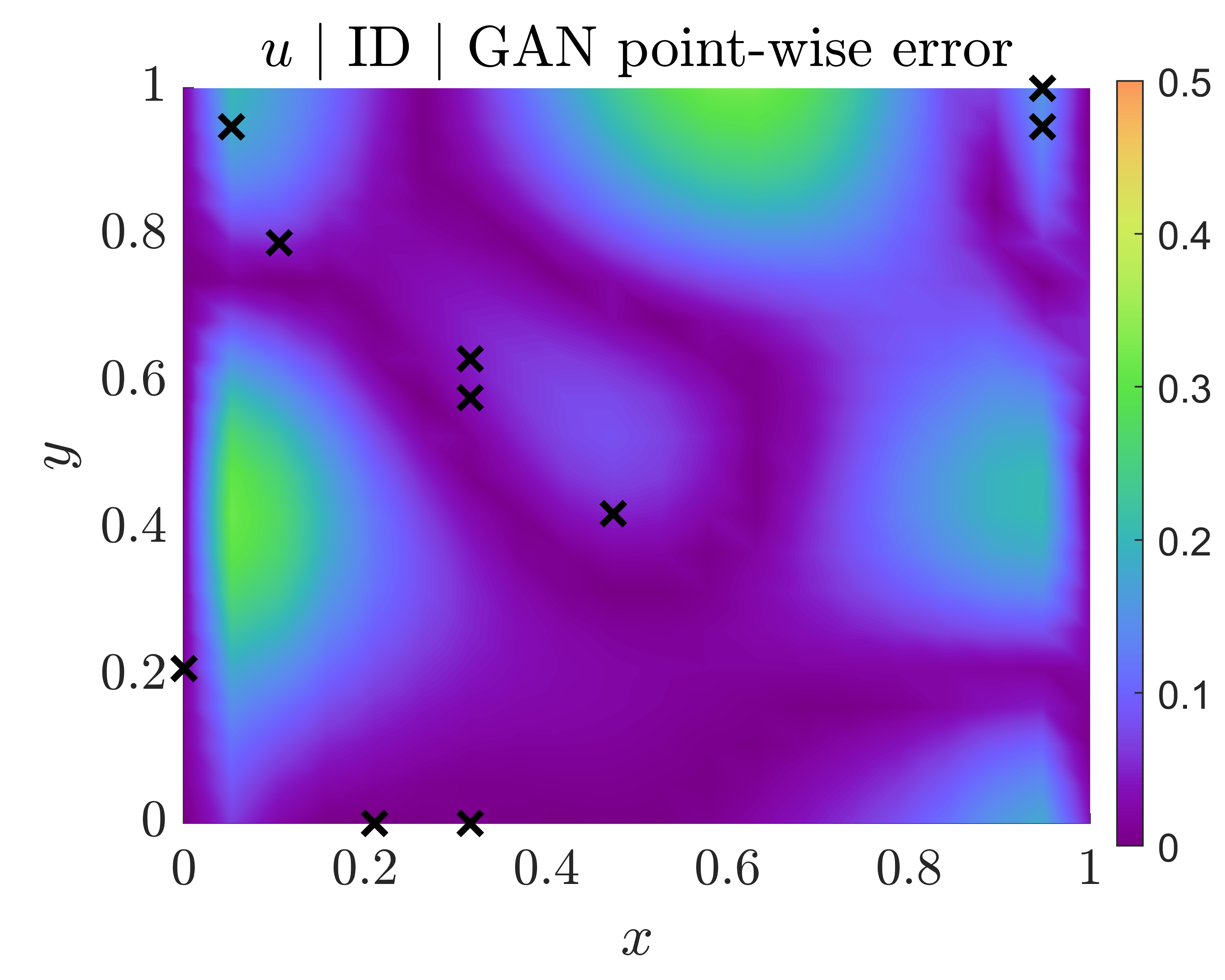}}
	\subcaptionbox{}{}{\includegraphics[width=0.24\textwidth]{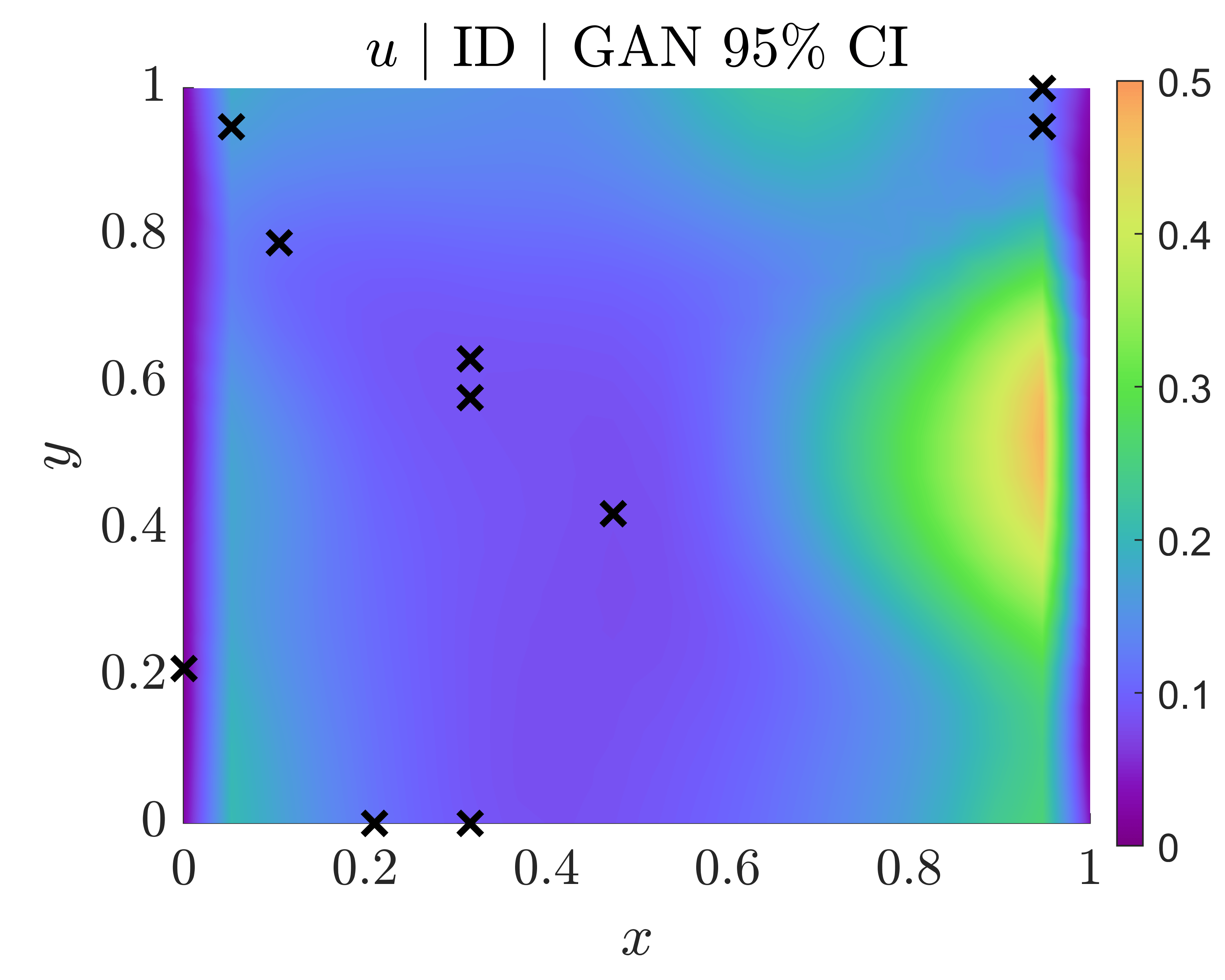}}
	\caption{
		Operator learning problem of Eqs.~\eqref{eq:comp:don:pde}-\eqref{eq:comp:don:bcs} | \textit{Limited and noisy inference data of $\lambda$ and $u$ (ID)}:
		although the mean of PA-GAN-FP is more accurate than the mean of the herein proposed PA-BNN-FP (Table~\ref{tab:comp:don:fp}), PA-GAN-FP is over-confident (less calibrated) for predicting $u$; compare, e.g., left sides of parts (g) and (h) and note that the error is not covered within the $95 \%$ CI.  
		This can be attributed to the fact that we pre-trained GAN using historical data of $\lambda$ only.
		Shown here are the absolute point-wise error and epistemic uncertainty predictions corresponding to input $\tilde{\lambda} = \log\lambda$ (a-d) and solution $u$ (e-f), as obtained by comparing PA-BNN-FP and PA-GAN-FP with reference solutions, as well as the corresponding inference data locations (x markers).
	}
	\label{fig:comp:don:fp:uncer:id}
\end{figure}

Next, to solve the problem we consider two versions of the model of Fig.~\ref{fig:uqt:fpriors:FP}c, namely the herein proposed PA-BNN-FP and the PA-GAN-FP proposed in \cite{meng2021learning}. 
In PA-BNN-FP, $\pazocal{M}$ in Fig.~\ref{fig:uqt:fpriors:FP}c is assumed to be a BNN; this corresponds to a BNN-induced functional prior.
In PA-GAN-FP, $\pazocal{M}$ in Fig.~\ref{fig:uqt:fpriors:FP}c is assumed to be the generator of a GAN as described in Section~\ref{sec:uqt:fpriors}.
We pre-train the GAN using clean historical data of $\lambda$, which in this case is the same as the data used for pre-training the DeepONet; i.e., the truncated GP of Eq.~\eqref{eq:comp:don:gp}. 
In both cases, we perform HMC posterior inference for obtaining samples $\{\hat{\theta_j}\}_{j=1}^M$, which corresponds to BNN parameters in the case of PA-BNN-FP, and to GAN generator random input in the case of PA-GAN-FP (see Fig.~\ref{fig:comp:don:gan:fp}).
The likelihood function used in both cases is exactly the same, i.e., Gaussian with mean being the output of each model and variance as described above.

In Table~\ref{tab:comp:don:fp}, we evaluate the performance of the two methods based on the metrics of Section~\ref{sec:eval}.
We perform the evaluation for ID and OOD data, i.e., for a field $\lambda$ from the same and from a different distribution, respectively, as compared to the one used for pre-training the DeepONet and the GAN.
Specifically, the OOD $\lambda$ corresponds to a sample from the truncated GP of Eq.~\eqref{eq:comp:don:gp} with correlation length $l = 0.2$ (100 leading terms).
It is shown that PA-GAN-FP performs better than PA-BNN-FP in terms of accuracy (RL2E) and predictive capacity (MPL) for both ID and OOD data.
This is also shown in Fig.~\ref{fig:comp:don:fp:mean:id}, where we plot the mean predictions for $\lambda$ and $u$ and compare with reference solutions.
Clearly, however, the performance of both methods deteriorates significantly for OOD data with RL2E values close to 40-50$\%$ (see also Figs.~\ref{fig:comp:don:ood:gan_bnn_mean}-\ref{fig:comp:don:ood:gan_bnn_std}).
Note also that the calibration error of PA-BNN-FP for $u$ is smaller for ID data.
This can be attributed to the fact that we pre-trained GAN using historical data of $\lambda$ only.
In Fig.~\ref{fig:comp:don:fp:uncer:id} we plot the absolute errors of the mean predictions as well as the epistemic uncertainties provided by PA-BNN-FP and PA-GAN-FP.
It is shown that the errors of both methods in most areas of the space domain are covered within the epistemic uncertainty $95 \%$ CIs (approximately two standard deviations).
Nevertheless, PA-GAN-FP is over-confident (less calibrated) for predicting $u$ in some areas of the space domain, as also indicated by its higher RMSCE value.
For example, by comparing the left sides of the plots in parts (g) and (h) of Fig.~\ref{fig:comp:don:fp:uncer:id}, where there is not much available data, we notice that the error is not covered with the $95 \%$ CI.
In passing, note that because comparisons in this example are made with clean reference solutions in Figs.~\ref{fig:comp:don:fp:mean:id}-\ref{fig:comp:don:fp:uncer:id}, we only plot epistemic uncertainty.
If the reference solutions used for comparisons are noisy, total uncertainty should be used instead.
In addition, we note that PA-BNN-FP is more calibrated than PA-GAN-FP also for the case of OOD data (see Table~\ref{tab:comp:don:fp} and Figs.~\ref{fig:comp:don:ood:gan_bnn_mean}-\ref{fig:comp:don:ood:gan_bnn_std})

Finally, in Fig.~\ref{fig:comp:don:fp:calib} we plot the RMSCE corresponding to $\tilde{\lambda} = \log \lambda$ and $u$ predictions, after post-training calibration with varying sizes of calibration datasets and calibration approach. 
We see that performing post-training calibration even with a few left-out noisy datapoints (2-10) can reduce the calibration error by half in some cases.
An interesting future research direction is to perform a cost-effectiveness study for comparing post-training calibration with active learning, i.e., with techniques that re-adjust the mean and uncertainty predictions in light of new data.
Although the computational cost of post-training calibration is approximately zero, there are cases in which it can make the predictions worse (e.g., Fig.~\ref{fig:comp:don:fp:calib}c), because it calibrates the predictions ``on average'', as explained in Section~\ref{sec:eval:calib}.
On the other hand, active learning may incur significant computational cost, but it is expected to perform better.

\subsubsection{U-DeepONet: combining DeepONet with deep ensemble for incorporating epistemic uncertainty}\label{sec:comp:don:dens}

In this section, we pre-train a DeepONet to learn the operator mapping from $\lambda$ to $u$, given the pre-training dataset $\cD = \{\cD_{\lambda}, \cD_u\}$.
In contrast to Section~\ref{sec:comp:don:fp}, here we employ a deep ensemble of five trained DeepONets; see DEns in Table~\ref{tab:uqt:over}. 
This corresponds to U-DeepONet of Table~\ref{tab:uqt:over}.
Next, during inference, we assume that new data arrives corresponding to an unseen random event $\xi' \in \Xi$, containing complete and clean samples from $\lambda$. 
This corresponds to the special case we described in Section~\ref{sec:intro:problem:form}; see also Table~\ref{tab:comp:don:problems}.
We denote the clean dataset as $\cD_{\lambda}' = \{x_i, y_i, \lambda_{i}\}_{i=1}^{N_{\lambda}'}$. The locations and the number of points are exactly the same as those of the points used in training of U-DeepONet, i.e., $N_{\lambda}'= 40 \times 40$ on a uniform grid. Note that this is an ill-posed problem because there is no available data of the solution $u$, not even in the boundaries of its domain.
The objective of this section is to demonstrate the advantage of U-DeepONet as compared to standard DeepONet, which does not provide epistemic uncertainty estimates.
Concurrently with the present work, \citet{lin2021accelerated} employed replica-exchange MCMC in conjunction with DeepONet for operator learning. 
Their model is a version of U-DeepONet, with MCMC as the posterior inference algorithm.

\begin{figure}[!ht]
	\centering
	\subcaptionbox{}{}{\includegraphics[width=0.32\textwidth]{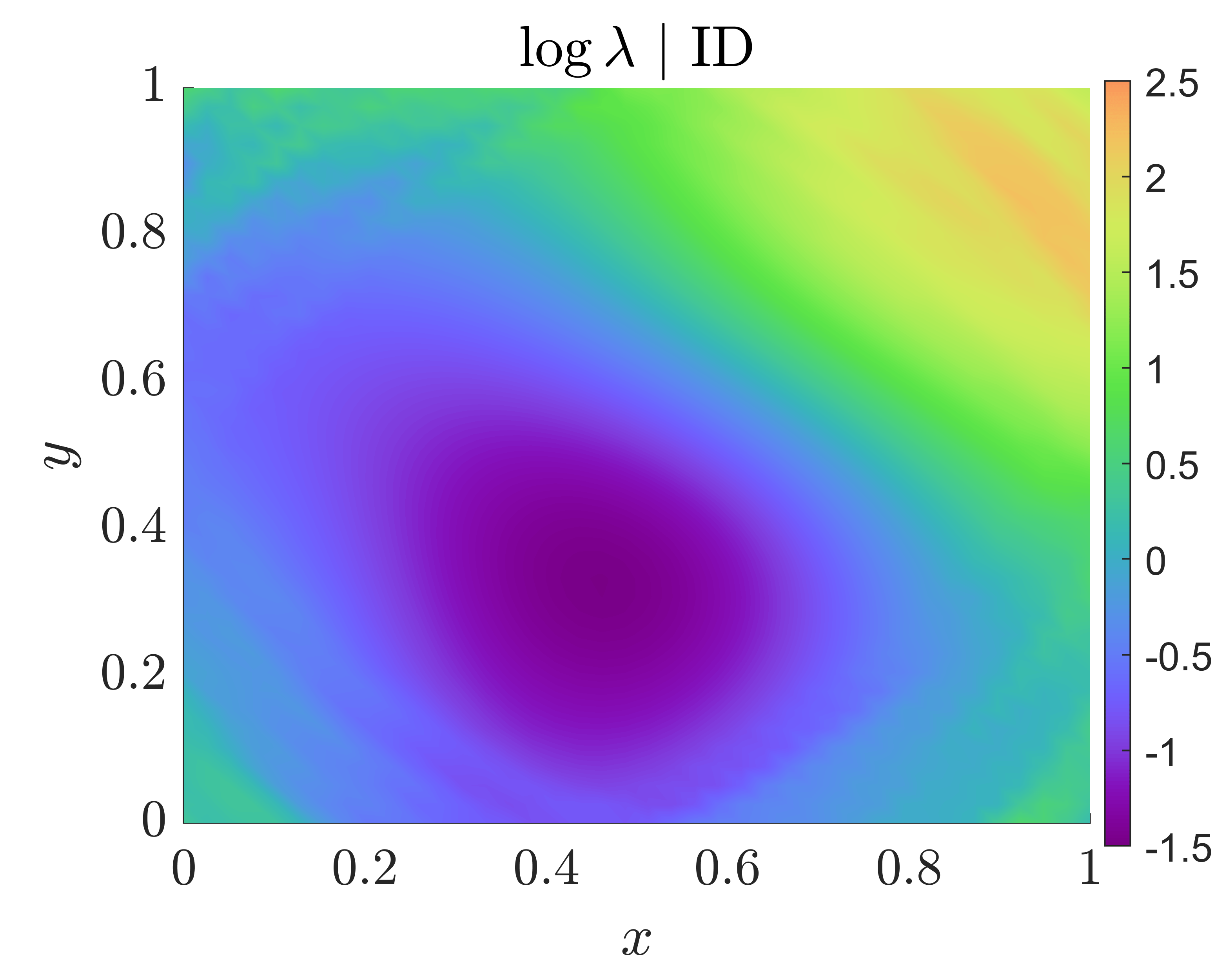}}
	\subcaptionbox{}{}{\includegraphics[width=0.32\textwidth]{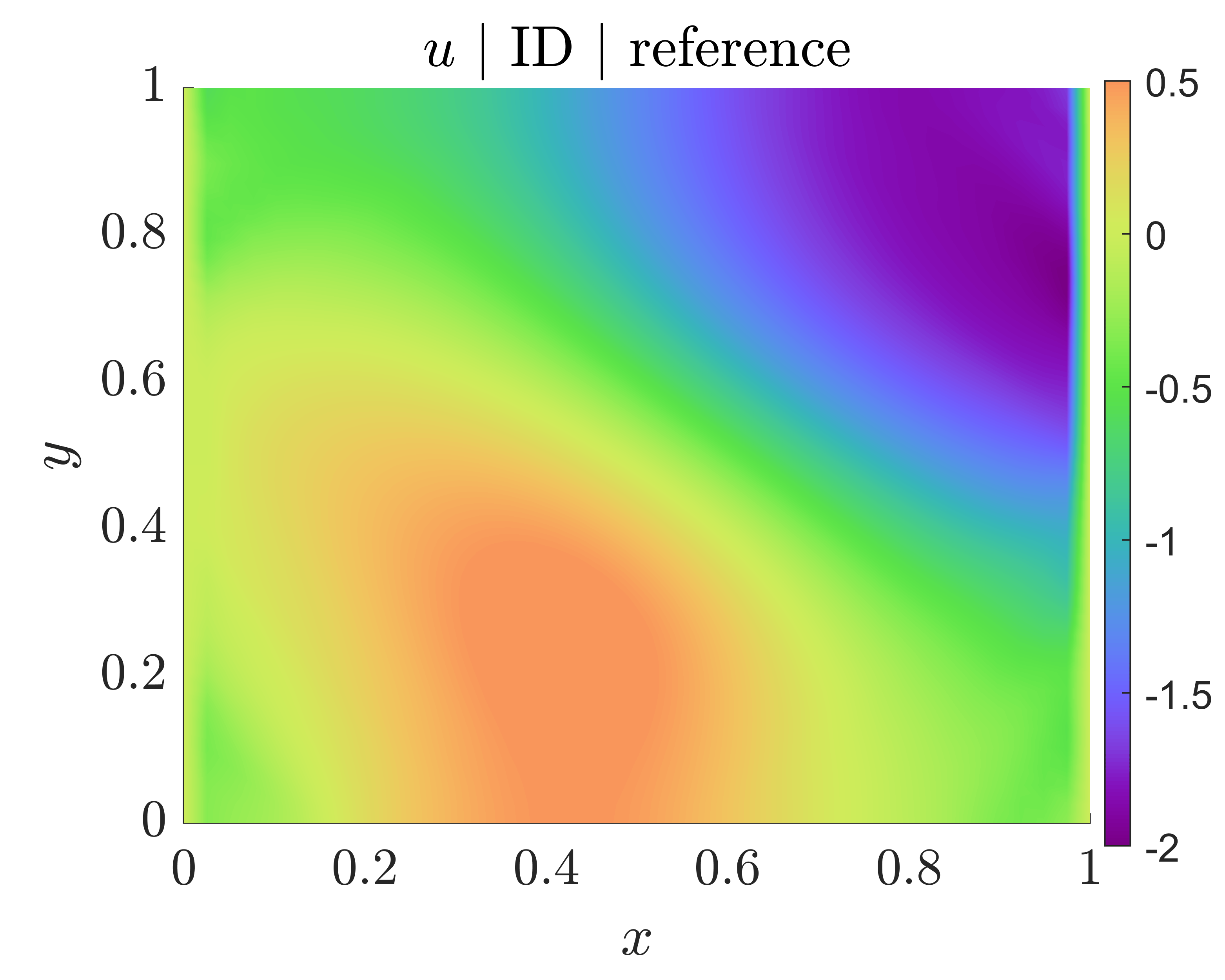}}
	\subcaptionbox{}{}{\includegraphics[width=0.32\textwidth]{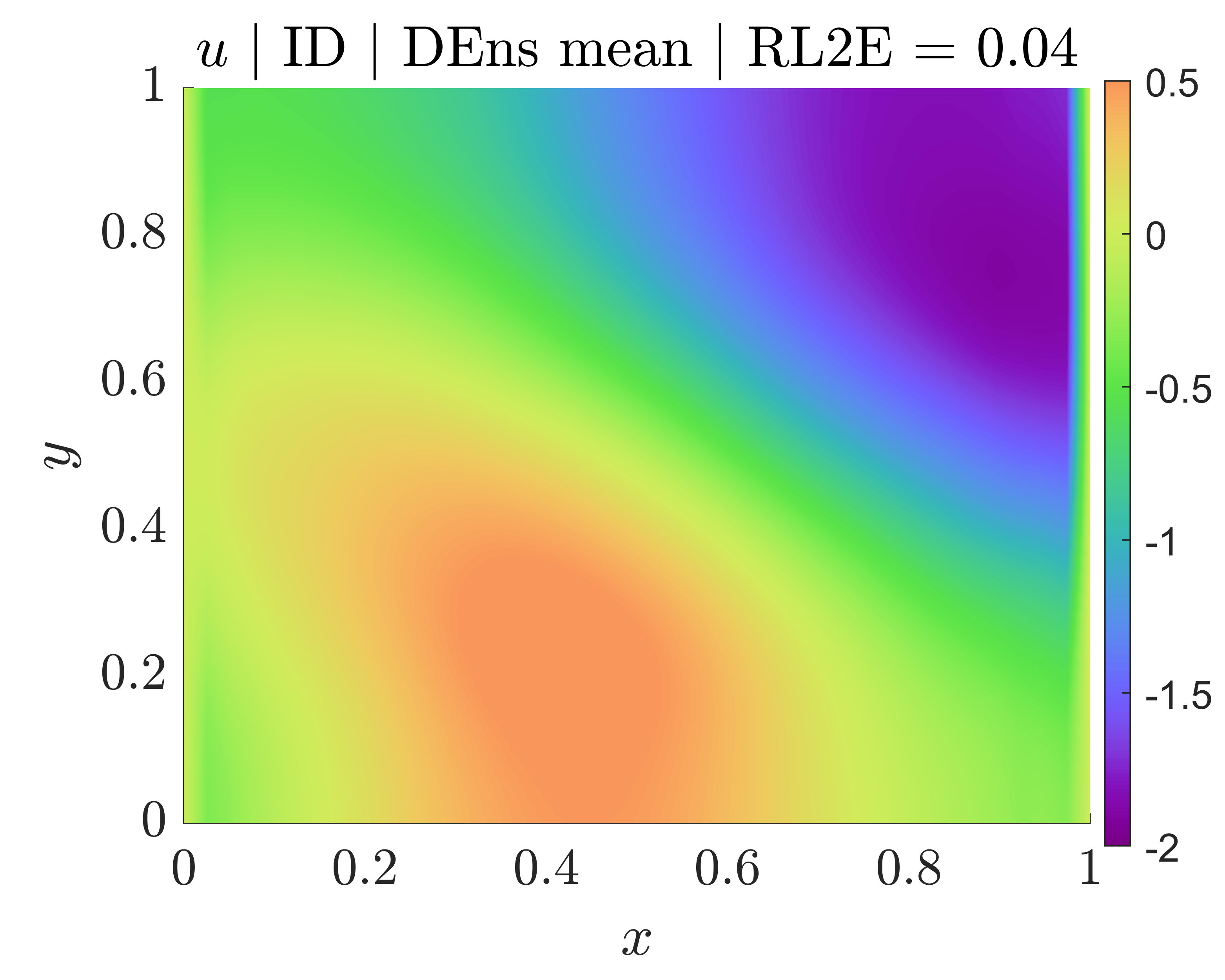}}
	\subcaptionbox{}{}{\includegraphics[width=0.32\textwidth]{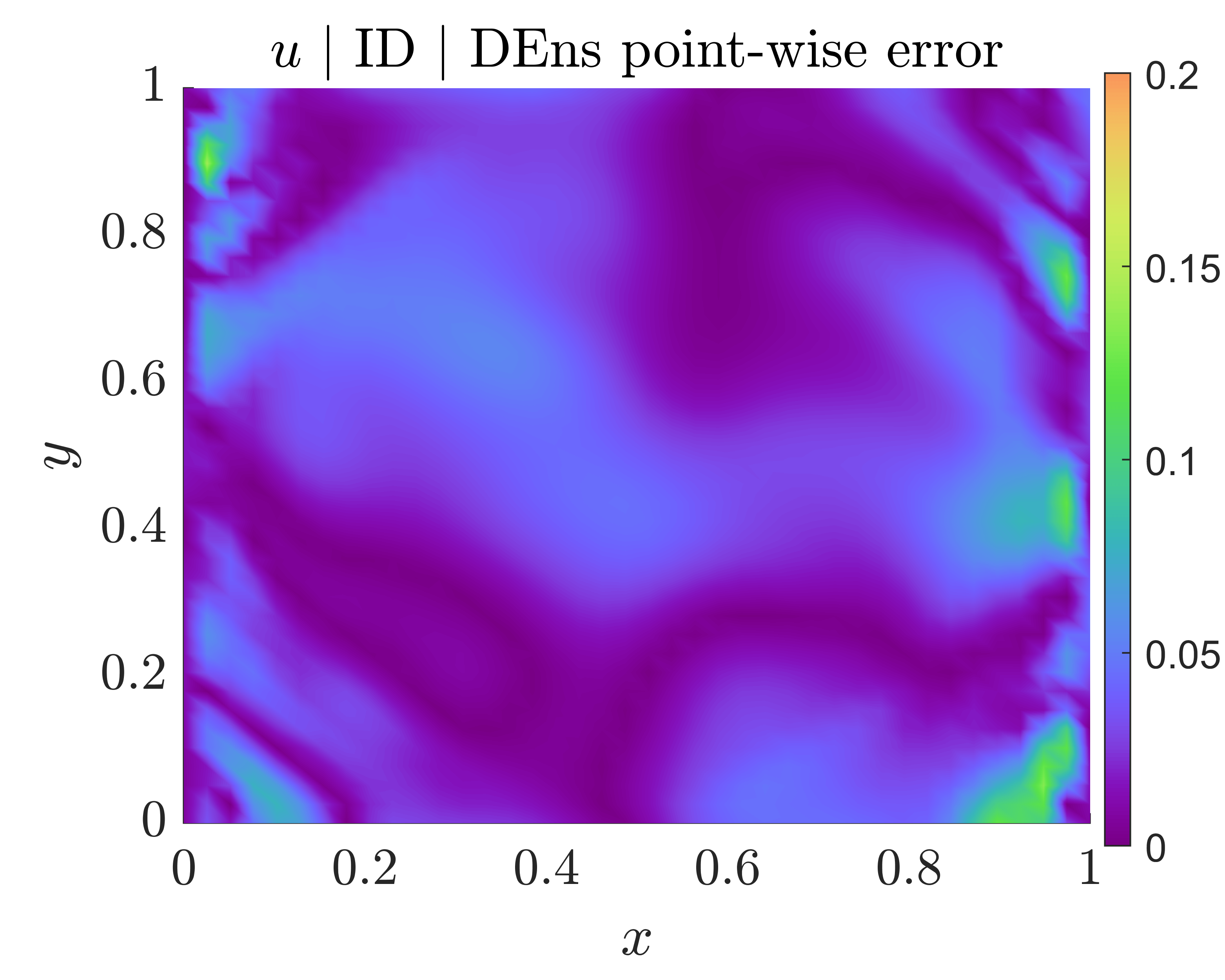}}
	\subcaptionbox{}{}{\includegraphics[width=0.32\textwidth]{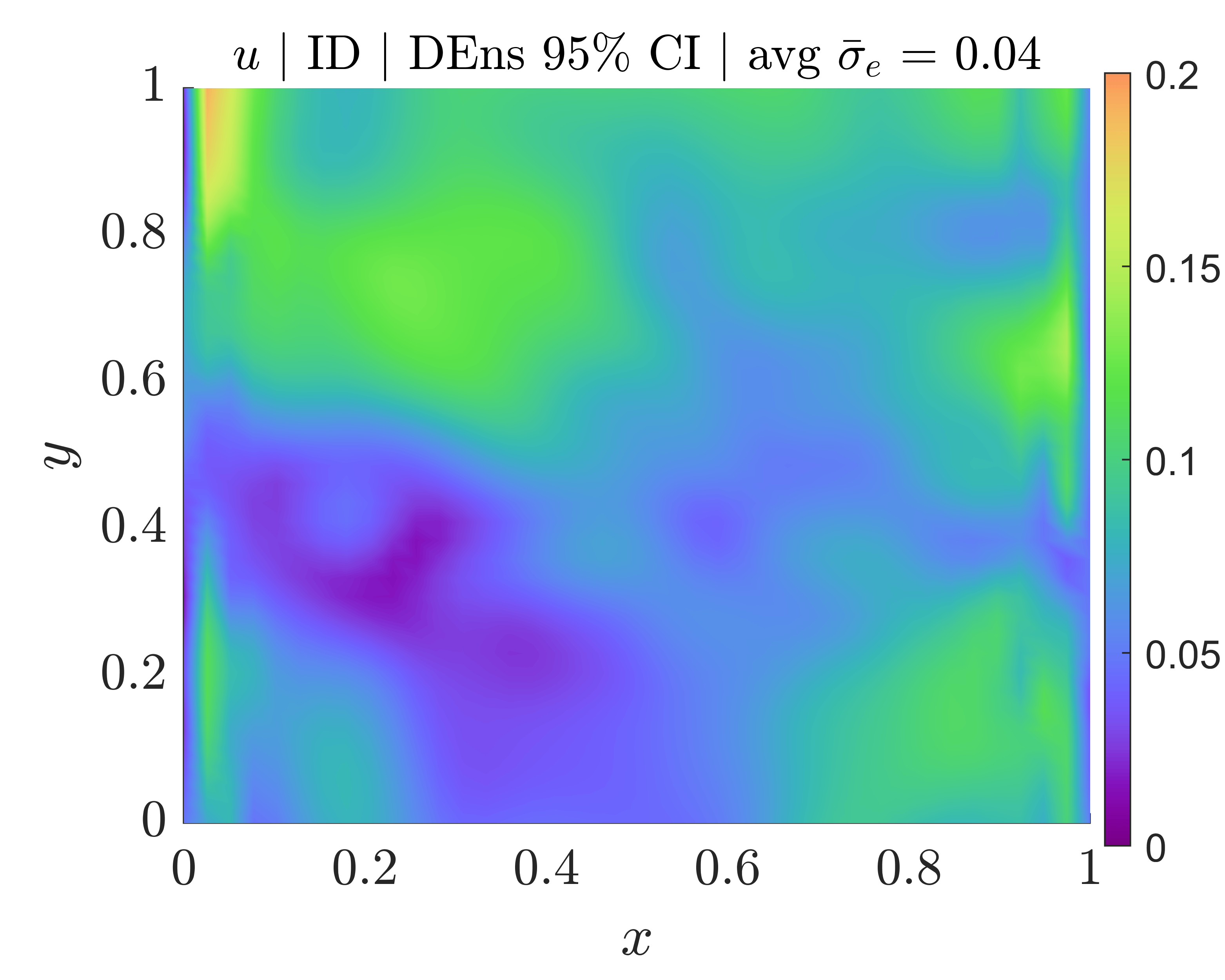}}
	\caption{
		Operator learning problem of Eqs.~\eqref{eq:comp:don:pde}-\eqref{eq:comp:don:bcs} | \textit{Complete and clean inference data of $\lambda$ (ID data)}: 
		the mean prediction of U-DeepONet, combined with DEns, approximates the solution $u$ accurately with RL2E = $4 \%$.
		Using each DeepONet separately, instead of a deep ensemble, would yield approximately the same error for ID data.
		Further, the $95 \%$ CI (approximately two standard deviations) of the epistemic uncertainty covers in all areas of the domain and follows spatially the absolute point-wise error of the mean.
		Here shown are the ID input $\tilde{\lambda}=\log \lambda$ (a), the reference solution $u$ (b), the mean of U-DeepONet (c), the absolute point-wise error of the mean (d), and the $95 \%$ CI of epistemic uncertainty (e).
	}
	\label{fig:comp:don:dens:id}
\end{figure}

\begin{figure}[!ht]
	\centering
	\subcaptionbox{}{}{\includegraphics[width=0.32\textwidth]{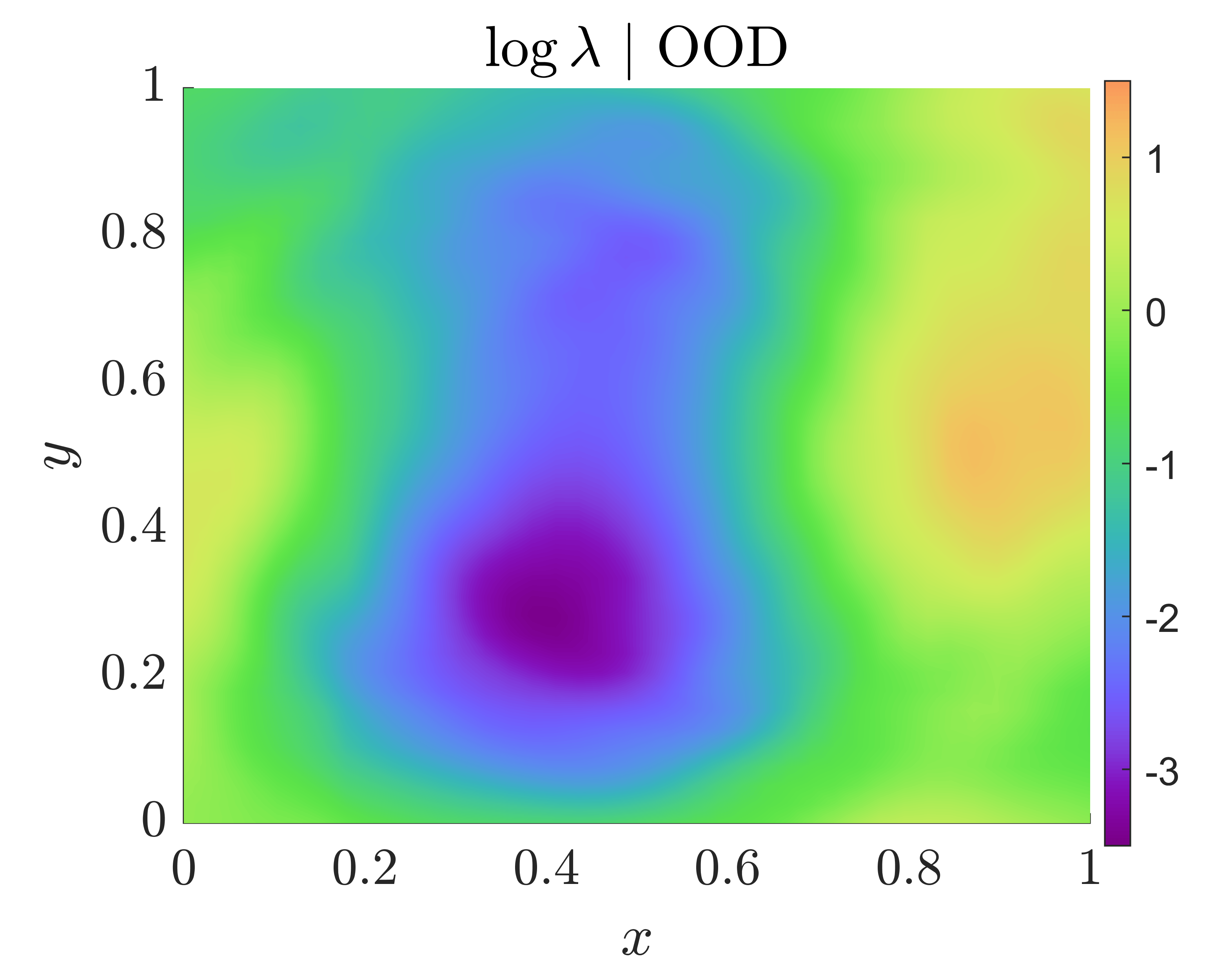}}
	\subcaptionbox{}{}{\includegraphics[width=0.32\textwidth]{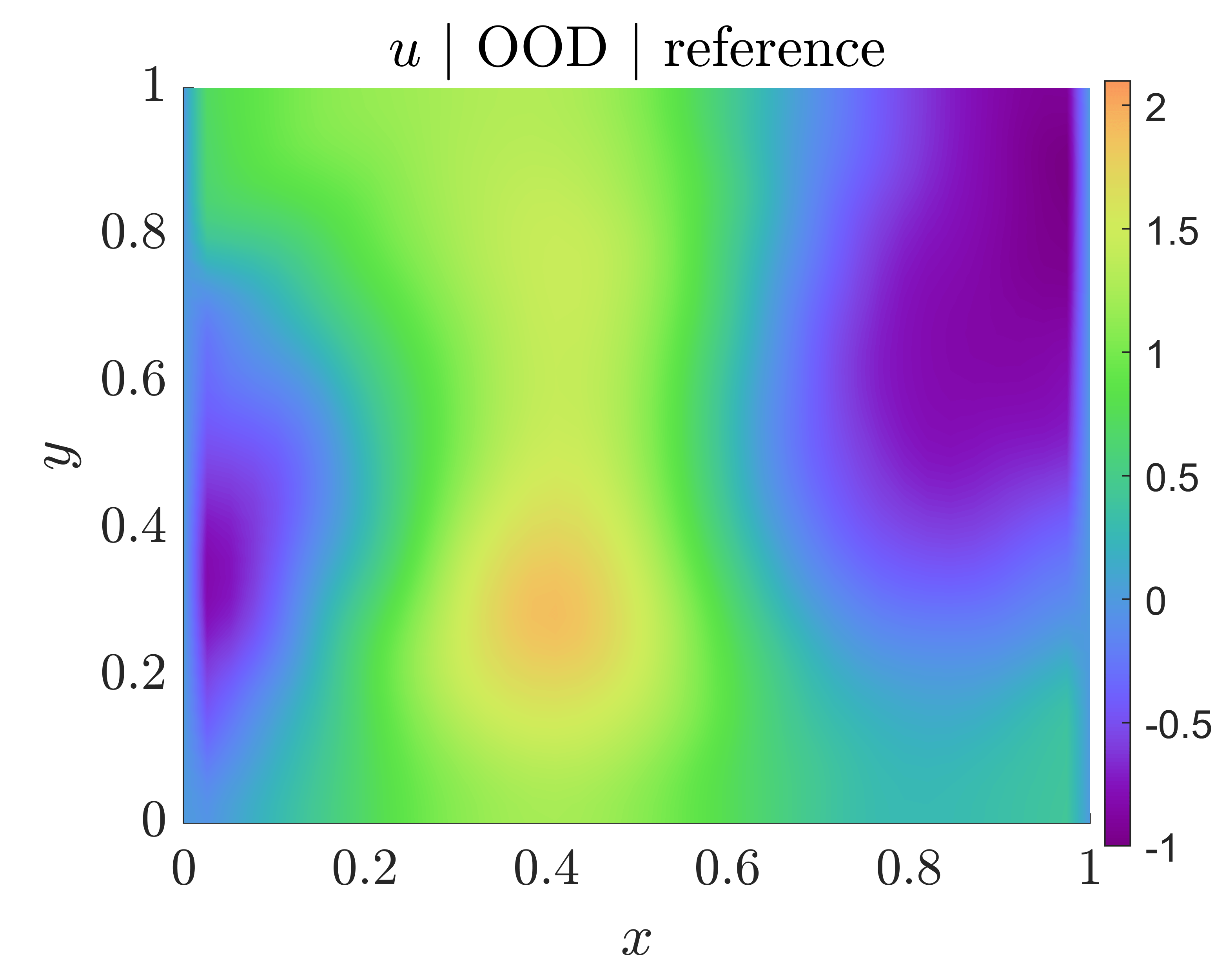}}
	\subcaptionbox{}{}{\includegraphics[width=0.32\textwidth]{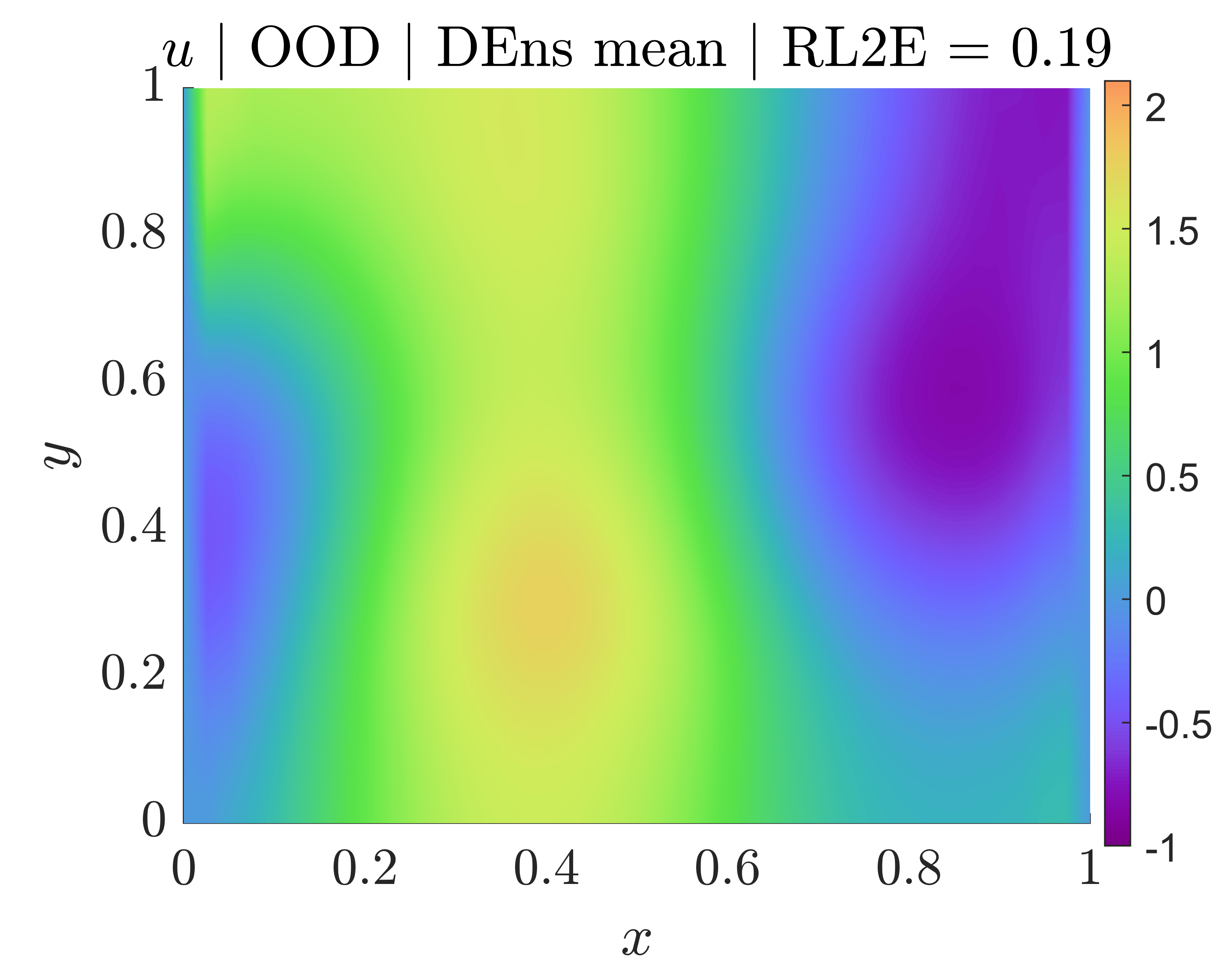}}
	\subcaptionbox{}{}{\includegraphics[width=0.32\textwidth]{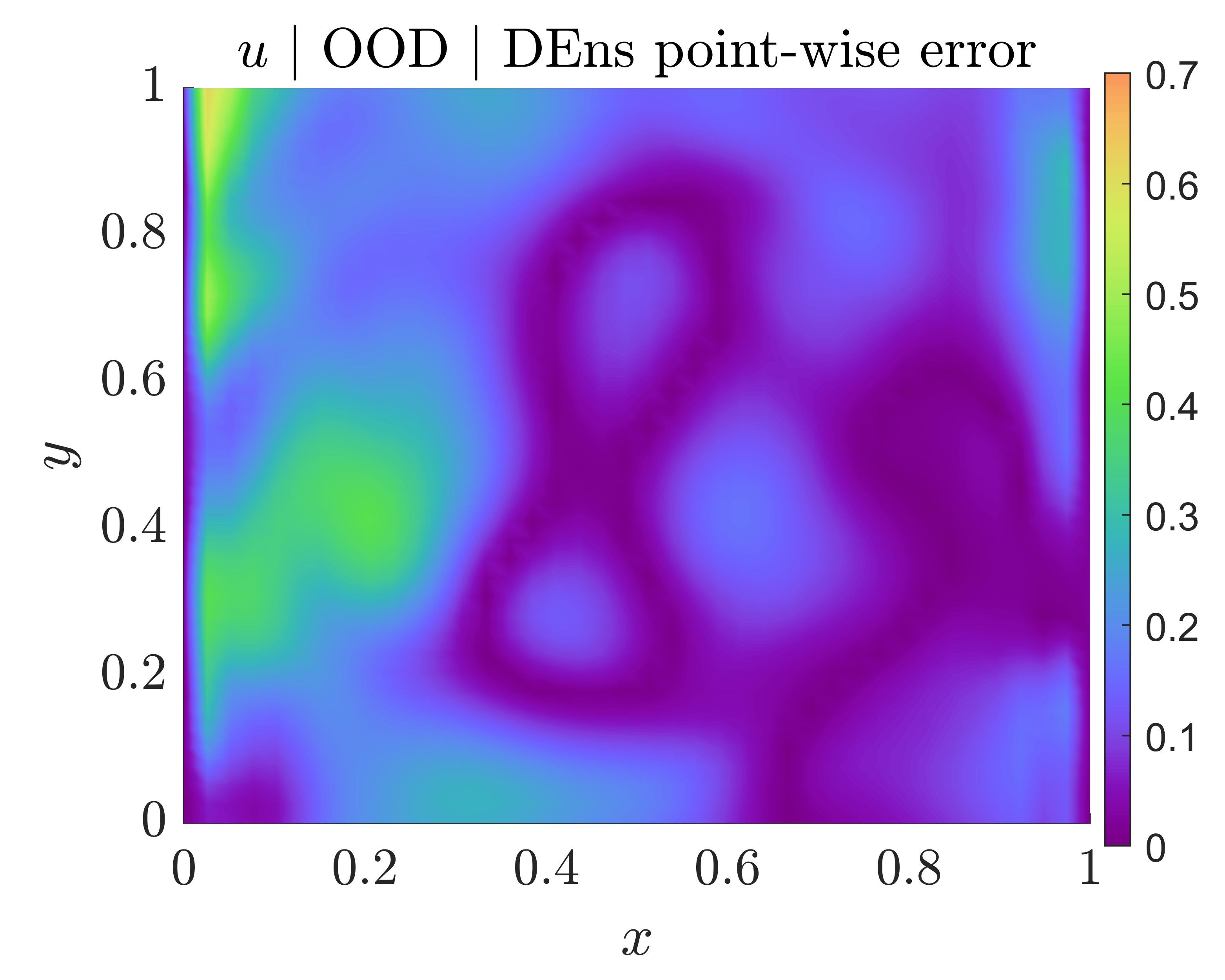}}
	\subcaptionbox{}{}{\includegraphics[width=0.32\textwidth]{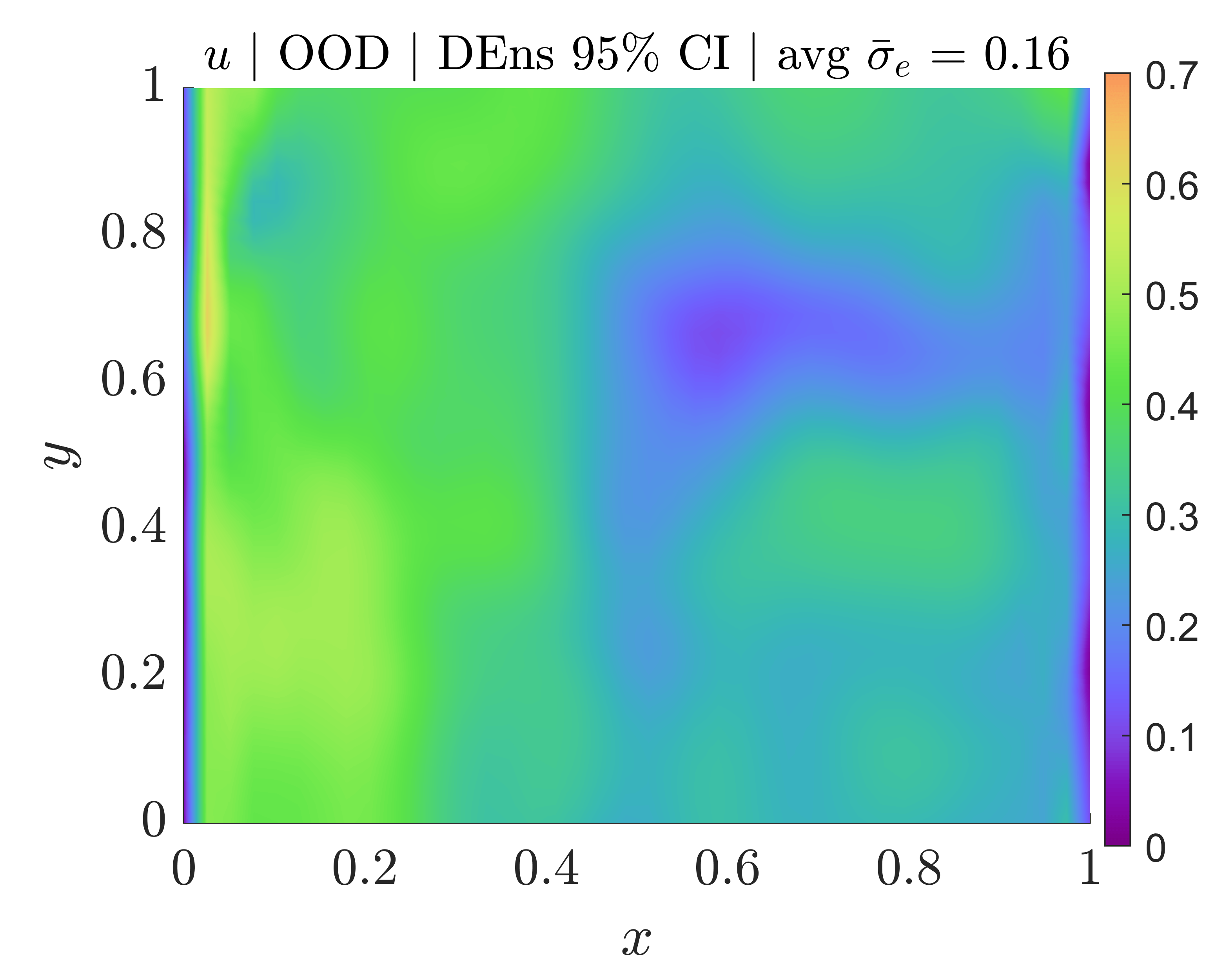}}
	\caption{
		Operator learning problem of Eqs.~\eqref{eq:comp:don:pde}-\eqref{eq:comp:don:bcs} | \textit{Complete and clean inference data of $\lambda$ (OOD data)}: 
		although the mean prediction of U-DeepONet, combined with DEns, has RL2E = 19$\%$, the value of the same metric would be 40$\%$ if a deep ensemble were not used.
		This fact demostrates the potency of U-DeepONet, as compared to standard DeepONet, for treating OOD data.
		Further, the $95 \%$ CI (approximately two standard deviations) of the epistemic uncertainty covers the absolute error of the mean in all areas of the domain.
		Here shown are the OOD input $\tilde{\lambda}=\log \lambda$ (a), the reference solution $u$ (b), the mean of U-DeepONet (c), the absolute point-wise error of the mean (d), and the $95 \%$ CI of epistemic uncertainty (e).
	}
	\label{fig:comp:don:dens:ood}
\end{figure}

In Figs.~\ref{fig:comp:don:dens:id}-\ref{fig:comp:don:dens:ood}, we present the neural operator predictions for representative examples of ID and OOD data, as obtained by U-DeepONet.
Specifically, for ID data, $\lambda$ corresponds to a sample from the same stochastic process used for pre-training the U-DeepONet, i.e., the truncated GP of Eq.~\eqref{eq:comp:don:gp}.
For OOD data, $\lambda$ corresponds to a sample from a truncated GP that follows Eq.~\eqref{eq:comp:don:gp} with correlation length $l=0.15$, instead of the value 0.25 used in pre-training. 
As shown in Fig.~\ref{fig:comp:don:dens:id}, the RL2E value for ID data is $4 \%$ and is approximately the same for each DeepONet of the ensemble separately.
However, U-DeepONet also provides epistemic uncertainty.
By comparing parts (d) and (e) of Fig.~\ref{fig:comp:don:dens:id} we observe that the $95 \%$ CI (approximately two standard deviations) of the epistemic uncertainty covers in all areas of the domain and follows spatially the absolute point-wise error of the mean.
Nevertheless, U-DeepONet flourishes in cases of OOD data; i.e, data that have not been seen during pre-training. 
Specifically, as shown in Fig.~\ref{fig:comp:don:dens:ood} for OOD data, the RL2E of U-DeepONet is approximately 19$\%$, whereas the error of each DeepONet of the ensemble used separately is approximately 40$\%$.
Further, similarly to the ID case, the $95 \%$ CI of the epistemic uncertainty covers the absolute point-wise error of the mean in all areas of the domain.
We also note that the overall epistemic uncertainty is higher for the case of OOD data, as compared to ID data. Specifically, the average along $(x,y)$ standard deviation of $u$ is 0.04 for ID data and 0.16 for OOD data.
This can be attributed to the fact that the five independently trained DeepONets ``disagree'' more on their predictions when the data has not been seen during training.
This additional property of U-DeepONet is harnessed in Section~\ref{sec:comp:don:ood} for detecting cases of OOD inputs to the learned neural operator.

In passing, note that a GP, or the BNN-FP and GAN-FP models of Section~\ref{sec:comp:don:fp}
can be combined with the U-DeepONet of this section for treating noisy, limited or complete, input data of $\lambda$ during inference, without access to $u$ data.
Specifically, given a noisy inference dataset $\cD_{\lambda}' = \{x_i, y_i, \lambda_{i}\}_{i=1}^{N_{\lambda}'}$, we can first perform posterior inference for obtaining, for instance, $M =$ 1,000 $\theta$ samples from the BNN, or GAN prior that explain the dataset.
In the BNN case, $\theta$ denotes the BNN parameters, and in the GAN case, $\theta$ denotes the GAN generator random input.
Subsequently, we use the $\theta$ samples to produce realizations of $\lambda$. 
These realizations are, finally, forward-passed through each of the five pre-trained DeepONets of the ensemble.
This results to $5\times1{,}000$ realizations of $u$. 
The uncertainty of these realizations corresponds to epistemic uncertainty due to DeepONet training (five models in the ensemble), as well as to uncertainty due to the noisy input $\lambda$.  
This is similar in spirit to the herein proposed GP+PI-GAN model that we tested for solving forward PDE problems in Section~\ref{sec:comp:pinns:forw}.
An interesting future research direction would be to extend the models we study in this paper for treating cases with limited and noisy input and output data, during both pre-training and inference in operator learning problems. 
We summarize the various problem cases and respective solution approaches in Table~\ref{tab:comp:don:problems}.

\subsubsection{Out-of-distribution (OOD) data detection}\label{sec:comp:don:ood}

\begin{figure}[!ht]
	\centering
	\includegraphics[width=.5\linewidth]{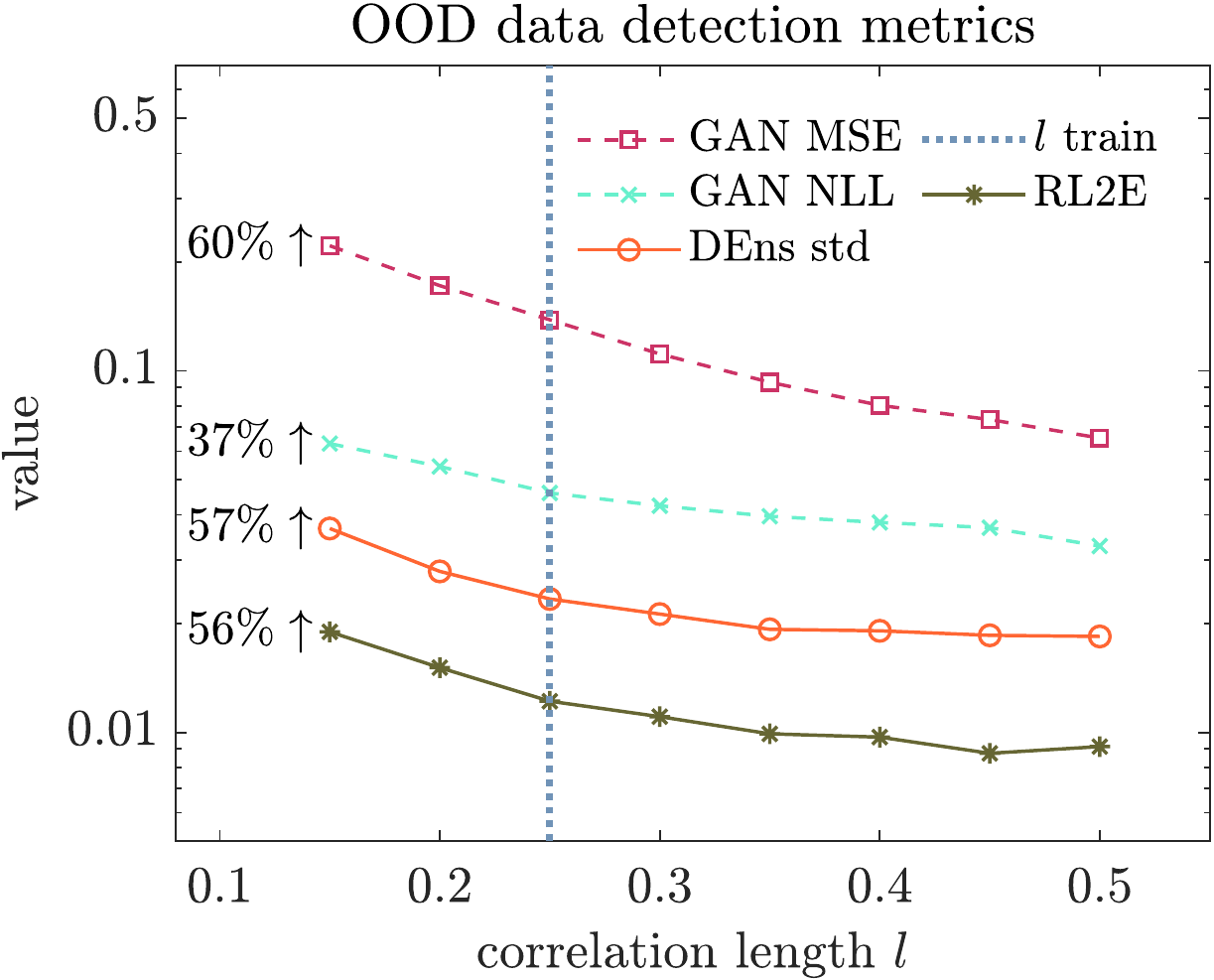}
	\caption{Operator learning problem of Eqs.~\eqref{eq:comp:don:pde}-\eqref{eq:comp:don:bcs} | \textit{OOD data detection}: the three proposed OOD data detection metrics of Table~\ref{tab:comp:don:ood:metrics}, namely, GAN MSE, GAN NLL, and DEns std, follow the RL2E trend.
		Here shown are the metric values for inputs $\lambda$ associated with a truncated GP of the form of Eq.~\eqref{eq:comp:don:gp} with different correlation lengths.
		The RL2E of U-DeepONet as well as the correlation length used in the U-DeepONet pre-training phase are also included for comparisons.
		All values are averaged over 100 $\lambda$ inputs.}
	\label{fig:comp:don:ood}
\end{figure}

In Section~\ref{sec:comp:don:fp}, we pre-trained one DeepONet and combined it with BNN-FP and GAN-FP for treating limited and noisy inference data of both $\lambda$ and $u$.
In Section~\ref{sec:comp:don:dens}, we pre-trained an ensemble of five DeepONets for treating complete and clean inference data of $\lambda$ only. 
In the end of Section~\ref{sec:comp:don:dens}, we explained that the methods of Sections~\ref{sec:comp:don:fp}-\ref{sec:comp:don:dens} can be combined for treating limited and noisy inference data of $\lambda$ only.
Nevertheless, these two methods can also be applied for OOD data detection.
Specifically, suppose that we have a pre-trained DeepONet (or an ensemble).
The question that we ask in this section is the following: \textit{given inference data of $\lambda$, which is the input of the DeepONet, what metric could be used for detecting whether this new data has been seen during pre-training?} Such a metric evaluates quantitatively the reliability of the DeepONet output based on the new data, and is useful for decision-making. 
As an example, consider Fig.~\ref{fig:comp:don:ood}.
The RL2E of the U-DeepONet output is approximately $1 \%$ on average based on 100 ID inputs $\lambda$; i.e., for $l=0.25$ in Eq.~\eqref{eq:comp:don:gp}.
However, the average RL2E increases significantly for OOD data with $l<0.25$ and the error for $l=0.15$ is approximately $60 \%$ higher than the error of $l=0.25$.
Nevertheless, for OOD data with $l>0.25$ the RL2E decreases.
That is, although U-DeepONet was not trained with $\lambda$ fields drawn from a GP with correlation lengths higher than $0.25$, it performs satisfactorily. 
Obviously, this a special case. Potential explanations are that a) the inputs $\lambda$ for $l=0.25$ used in training contain most of the information of the unseen inputs for $l>0.25$; and b) the outputs $u$ for $l=0.25$ have similar characteristics with the unseen outputs for $l>0.25$.
Next, note that the RL2E values shown in Fig.~\ref{fig:comp:don:ood} are not available in practice; i.e., we only have one shot for predicting the solution $u$ based on a single realization of the input $\lambda$.
The objective of this section is to propose metrics that utilize only the available information, which is a single input $\lambda$, and follow the RL2E trend in Fig.~\ref{fig:comp:don:ood}. 
Indicatively, more information on OOD detection can be found in \cite{charpentier2020posterior,liang2020enhancing,martin2020inspecting,martin2021detecting,wang2021you,hendrycks2018baseline}.

\begin{table}[!ht]
	\centering
	\footnotesize
	\begin{tabular}{c|c}
		\toprule
		\multicolumn{2}{c}{\textbf{Proposed OOD data detection metrics}}\\
		\midrule
		GAN MSE&$\frac{\sum_{i=1}^{N_{\lambda}'}||\lambda_i-\lambda_{\hat{\theta}}(x_i, y_i)||^2_2}{\sum_{i=1}^{N_{\lambda}'}||\lambda_i||^2_2}$\\
		\midrule
		GAN NLL&$||\hat{\theta}||^2_2 + 0.5K \log 2\pi$\\
		\midrule
		DEns std & $\mathbb{E}_{x, y} [\bar{\sigma}_e(x, y)]$, with $\bar{\sigma}_e$ given in Eq.~\eqref{eq:uqt:pre:mcestmc:totvar}  \\
		\bottomrule
	\end{tabular}
	\caption{OOD data detection metrics proposed herein. GAN MSE corresponds to the normalized final training MSE, following optimization of $\theta$ using a GAN-FP and new data of $\lambda$.  
		GAN NLL corresponds to the negative log-likelihood of optimal $\theta$ obtained via optimization as above.
		DEns std corresponds to the average (along $x,y$ in the domain) standard deviation of the output $u$, as obtained via U-DeepONet given new data of $\lambda$. 
	}
	\label{tab:comp:don:ood:metrics}
\end{table}

In this regard, we assume that we have a pre-trained GAN based on historical data of $\lambda$, exactly as in Section~\ref{sec:comp:don:fp}. This GAN learns the distribution of the data that the DeepONet has been trained based on. Next, we consider inference data of $\lambda$, limited or complete and noisy or clean. We can perform posterior inference for obtaining $\theta$ samples that explain the data, where $\theta$ is the input to the GAN generator. Because in this use case we are not interested in the uncertainty of $\theta$, we can obtain a point estimate $\hat{\theta}$ using standard optimization according to Section~\ref{app:modeling:point}, instead of costly posterior inference according to Sections~\ref{sec:uqt:bnns}-\ref{sec:uqt:ens}. 
Following optimization, an optimal value $\hat{\theta}$ is available, as well as the final MSE between the approximator $\lambda_{\hat{\theta}}$ evaluated on the data locations and the data of $\lambda$. 
The first metric that we propose is the normalized final training MSE, which we refer to as \textit{GAN MSE}, and is given in the first row of Table~\ref{tab:comp:don:ood:metrics}.
Clearly, if we work with standardized and/or transformed data as explained in the beginning of Section~\ref{sec:comp:don}, the respective modified values are used instead of the $\lambda$ values.
The rationale of this metric is that if we are unable to obtain a $\hat{\theta}$ through optimization that yields a small training MSE, the GAN-FP is unable to fit the given data, and thus it is expected that the GAN has not seen this data during pre-training.
As shown in Fig.~\ref{fig:comp:don:ood}, the GAN MSE metric follows the same trend as RL2E and the metric value for $l=0.15$ is $60 \%$ higher than the metric value for $l=0.25$.
In practice, we can obtain the average GAN MSE for 10, for example, ID inputs $\lambda$, and subsequently given a new input $\lambda$ we can compare its GAN MSE with the average GAN MSE from the ID data.

\begin{figure}[!ht]
	\centering
	\includegraphics[width=1\linewidth]{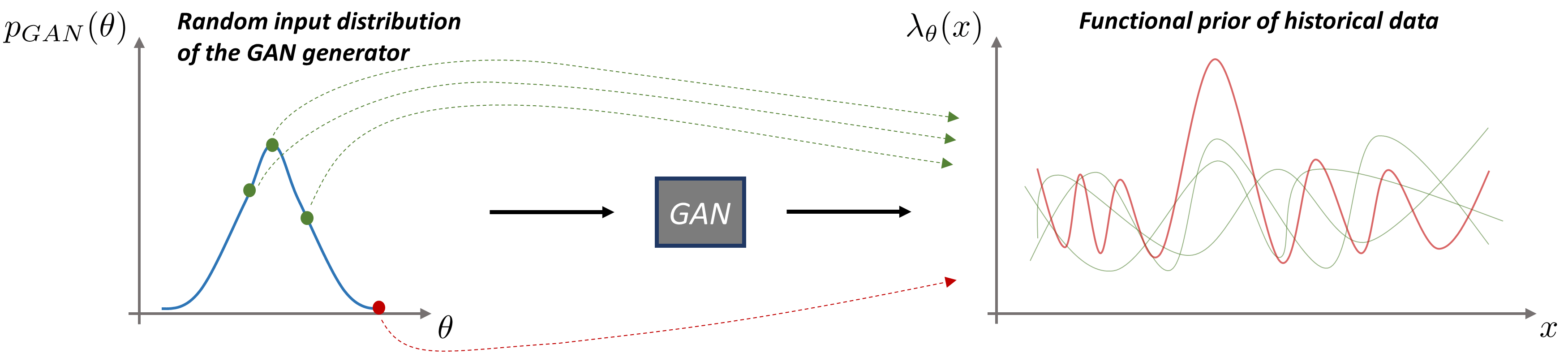}
	\caption{The GAN-FP is trained such that drawing samples of $\theta$ from $p_{GAN}(\theta)$ produces $\lambda$ functions that follow the historical data distribution. For OOD detection, we utilize the inverse of this relationship: a $\lambda$ function that is far from the historical data distribution is expected to correspond to a $\theta$ value that is far from the GAN input distribution. Values of $\theta$ shown with green color on the left side are more probable under the distribution $p_{GAN}$ and correspond to functions $\lambda$ on the right side that have similar characteristics. On the other hand, the $\theta$ value shown with red color is less probable and corresponds to a function $\lambda$ that is different from the rest.}
	\label{fig:comp:don:gan:fp}
\end{figure}

The second metric we propose, which we refer to as \textit{GAN NLL}, is the negative log-likelihood of $\hat{\theta}$ based on the known distribution of the GAN generator random input, and is given in the second row of Table~\ref{tab:comp:don:ood:metrics}.
In this paper, this distribution is denoted by $p_{GAN}$ and is a factorized Gaussian distribution, with $\cN(0,1)$ in each dimension.
Thus, GAN NLL is given as $-\log p_{GAN}(\hat{\theta}) = ||\hat{\theta}||^2_2 + 0.5K \log 2\pi $, where $K$ represents the size of $\hat{\theta}$. Note that the GAN has been trained such that drawing samples of $\theta$ from this Gaussian distribution produces $\lambda$ fields that follow the historical data distribution. For OOD data detection, we utilize the inverse of this relationship: a $\lambda$ field that is far from the historical data distribution is expected to correspond to a $\theta$ value that is far from the GAN input distribution.
This is illustrated in Fig.~\ref{fig:comp:don:gan:fp}, where values of $\theta$ with green color on the left side are more probable under the input distribution $p_{GAN}$ and correspond to functions $\lambda$ on the right side that have similar characteristics. On the other hand, the $\theta$ value shown with red color is less probable and corresponds to a function $\lambda$ that is quite different from the rest. If the function $\lambda$ shown with red is used in the optimization as explained above, it is expected to yield an optimal $\hat{\theta}$ that belongs in the tails of $p_{GAN}$, and thus it is associated with a high value of GAN NLL.
As shown in Fig.~\ref{fig:comp:don:ood}, the metric GAN NLL follows the same trend as RL2E and the metric value for $l=0.15$ is $37 \%$ higher than the metric value for $l=0.25$.

Next, consider a pre-trained ensemble of DeepONets, i.e., a U-DeepONet combined with DEns of Section~\ref{sec:uqt:ens}.
As shown in Figs~\ref{fig:comp:don:dens:id}-\ref{fig:comp:don:dens:ood} for a random ID or OOD input $\lambda$, respectively, the epistemic uncertainty of the output $u$ of U-DeepONet for OOD input $\lambda$ is higher (0.04 vs 0.16). 
In this regard, the third metric we propose, which we refer to as \textit{DEns std}, is the average along $(x,y)$ standard deviation of $u$ obtained via U-DeepONet.
It is given in the third row of Table~\ref{tab:comp:don:ood:metrics}.
As shown in Fig.~\ref{fig:comp:don:ood}, the metric DEns std follows the same trend as RL2E and the metric value for $l=0.15$ is $57 \%$ higher than the metric value for $l=0.25$.

\subsubsection{Summary}

In this section, we presented a comparative study pertaining to neural operator learning. Following the pre-training phase of DeepONet, we considered two inference cases, namely a case with noisy and limited inference data, and a case with clean and complete data. In the first case, we used the PA-GAN-FP and the herein proposed PA-BNN-FP. In the second case, we used a U-DeepONet combined with an ensemble of five independent training sessions of DeepONet.
Further, we proposed and tested three metrics for detecting OOD inference input data to U-DeepONet. This can be used to raise a ``red flag'' that the DeepONet output given a new input may not be accurate.

Overall, for treating limited and noisy inference data, combining a pre-trained DeepONet with a GAN-FP (PA-GAN-FP) performs better than the herein proposed BNN-FP (PA-BNN-FP), as expected. However, PA-BNN-FP does not incur any additional computational cost and is more calibrated in our experiments. 
For treating complete and clean inference data, U-DeepONet is more accurate (in terms of mean predictions) than standard DeepONet for OOD data, while the provided  epistemic uncertainty covers the errors for ID and OOD data. The epistemic uncertainty also increases for OOD data, a fact that is harnessed for OOD data detection.
For OOD data detection, the three proposed detection metrics follow the errors of performing inference with OOD data. Therefore, in practice, for a new input to the DeepONet, we can use the detection metrics to evaluate whether the input has been seen during training.

%% file: IN_accVScost.tex
This section provides qualitative comments and quantitative results regarding the computational cost of the employed UQ methods. In all cases, we compare the computational cost with that of standard training of NNs.
Note that we only compare the cost of obtaining the samples, and not the cost of using them for obtaining the sought statistics. The latter cost depends on the number of samples $M$ that each method collects and for the problems we considered herein is negligible. It becomes, however, important for larger NNs and number of evaluation points. 

For obtaining posterior samples $\{\hat{\theta}_j\}_{j=1}^M$ (or $\{\hat{\beta}_j\}_{j=1}^M$), we considered eight methods, as summarized in Table~\ref{tab:uqt:over}, excluding GP regression. In the following, we provide qualitative comments regarding their computational cost. HMC is a two-step MCMC approach that simulates the Hamiltonian dynamics of a fictitious particle by using a number of leapfrog steps and rejection sampling. Indicatively, for the function approximation problem and for 50 leapfrog steps the computational cost of HMC is 4-5 times higher than that of standard training. LD is a one-step MCMC approach that does not use rejection sampling and resembles stochastic gradient descent with added noise to the gradient. Its cost is approximately the same as that of standard training, although because of noise it may require a larger number of steps to converge. Unlike standard training that updates the NN parameters only, MFVI updates the mean and standard deviation of a NN variational distribution in each training step. Its cost is approximately the same as that of standard training, although in this study we used early stopping and the resulting cost can vary and be less than standard training without early stopping. MCD corresponds to standard training including dropout and its cost is approximately the same as that of standard training. LA consists of standard training and, subsequently, Hessian approximation. Depending on the method employed for Hessian approximation, the cost of LA is slightly higher than that of standard training.
DEns consists of $M$ standard training sessions and its cost is either the same (if performed in parallel) or $M$ times higher (if performed serially) than that of standard training. SEns corresponds to standard training with an appropriate learning rate schedule, and thus its cost is approximately the same as that of standard training. SWAG consists of SEns and, subsequently, fitting and sampling a Gaussian distribution using the optimization iterates. Its cost is slightly higher than that of standard training. 
In Table~\ref{tab:comp_cost}, we provide the indicative computational time required for performing posterior inference in the function approximation problem of Section~\ref{sec:comp:func:homosc:kno} using each of the methods described above.

\begin{table}
    \centering
    \begin{tabular}{c|ccccccccc}
    \hline 
      & HMC & LD & MFVI& MCD& LA& DEns& SEns& SWAG \\
      \hline 
     CPU (s) & 131.5 & 44.4 & 19.5 & 41.7& 60.3  & 359.8 & 31.9 & 36.6 \\
     
     GPU (s) & 322.7 & 269.7 & 31.2 & 55.4 & 77.6 & 494.7 & 56.1 & 61.5 \\
     \hline 
    \end{tabular}
    \caption{Indicative computational time for performing posterior inference in the function approximation problem of Section~\ref{sec:comp:func:homosc:kno} with each of the methods of Table~\ref{tab:uqt:over}, by using CPU and GPU.
    CPU details: 2 Intel(R) Xeon(R) CPU E-2643 @ 3.30 GHz (4 cores, total 8 threads);  GPU details: Nvidia's TITAN Xp. 
    Note that the computational time of DEns is the total time required for $M=10$ independent training sessions performed serially. This can also be done in parallel.
    Although, clearly, NN training on GPU is generally faster than on CPU, in this case the computational cost for all methods is higher on GPU.
    For HMC and LD, based on our experiments one reason that makes CPU more efficient is that we need to generate random numbers for the momentum at each iteration, which is more efficient on a simple architecture, such as CPU, than on GPU for {\emph{tensorflow probability}} \cite{lao2020tfp}.
    Further, the rejection sampling of HMC includes an if/else statement, which is more efficient on CPU than on GPU.
    For the remaining approaches, which are trained using the Adam optimizer, using GPU is typically preferable, especially in cases with big data. Specifically, matrix-vector multiplication, which is a heavy-lifter for NN training, has spatial locality of $O(n)$ and naturally maps to GPU architecture. However, for cases with small data, a CPU may have comparable performance to a GPU. The reason is that we need to transfer the dataset from the CPU memory to the GPU, and then transfer it back to the CPU memory. Transferring the data from CPU to GPU and vice versa is facilitated by PCI Express, which in our case is 3.0 and can provide only 8 GT/s (Giga Transmission per second). Thus, the computational time in small data cases is dominated by the memory latency.
    Overall, this table is not intended for performance comparison between CPU and GPU, but only for reporting the respective elapsed times on both architectures. 
    }
    \label{tab:comp_cost}
\end{table} 

Next, h-X combines the above methods with heteroscedastic noise modeling, i.e, the involved NNs have one additional output modeling the noise. If the same NN architecture is used as above, the computational costs remain approximately the same as described above. 
All the above methods assume by default a BNN-FP, i.e., a standard NN with unknown parameters following a prior distribution. If a GAN-FP is used instead of a BNN-FP, the cost of posterior inference can be smaller, because we draw samples of $\theta$ that corresponds to the GAN generator input. This input commonly has a significantly lower dimension compared to the parameters $\theta$ of a NN. However, GAN-FP requires pre-training using historical data. Indicatively, training the GAN-FP used in the function approximation problem with heteroscedastic noise (Fig.~\ref{fig:comp:func:hetero:res}) takes approximately one hour on one GPU.
Finally, these computational costs generalize accordingly when we combine the posterior inference methods with PINNs and DeepONets, as summarized in Table~\ref{tab:uqt:over}.

In terms of accuracy of mean predictions, predictive capacity, and statistical consistency (calibration), the most computationally expensive methods, namely HMC and DEns, are expected to exhibit the highest performance. Their performance can significantly be enhanced if the prior is learned in HMC or the hyperparameters of DEns are tuned. This has been corroborated with our experiments throughout the present study. In terms of the computationally cheaper methods, we found that they can provide accurate mean predictions and relatively calibrated predictions, especially if their priors (or hyperparameters) are learned (or tuned). However, with the exception of LA, they are less calibrated than HMC and DEns and their epistemic uncertainties do not increase consistently with increasing noise and decreasing dataset size (see Section~\ref{sec:comp:func:homosc:kno}). Nevertheless, we found that LA, MFVI, and SEns can provide \textit{a good trade-off between performance and cost}. We suggest to potential users of the presented methods to (a) compare various methods for solving a problem at hand; (b) use a plurality of evaluation metrics; (c) learn (or tune) the priors (or hyperparameters) using validation data (where applicable); and (d) calibrate, after training, the predictions using a small left-out calibration dataset. Lastly, we found that by incurring some additional pre-training computational cost using a GAN-FP, we can reduce the computational cost of posterior inference and improve performance. 

In terms of code implementation, MFVI, MCD, LA, DEns, SEns, and SWAG can be implemented by extending available standard NN training codes in a straightforward manner. For LA, an off-the-shelf Hessian approximation implementation can be used; see, e.g., \cite{daxberger2021laplace}. For HMC and LD the open-source package {\emph{tensorflow probability}} \cite{lao2020tfp} can be used, with the implementation of HMC being more involved, especially for learning the prior and noise scale. For implementing h-X, we simply include an additional NN output modeling the noise.
For GAN-FP, we used our own implementation following Section~\ref{app:methods:fpriors}. For post-training calibration, we used the package of \citet{chung2021uncertainty}. In the near future, we will provide additional implementation details in an upcoming manual and release our code in a public GitHub repository.

%% file: IN_discussion.tex
We have presented UQ methods for neural networks for function approximation, and for solving PDEs and learning operator mappings between
infinite-dimensional function spaces. 
In particular, we have presented a formulation towards a unified and comprehensive framework for UQ in SciML, which is summarized in Fig.~\ref{fig:intro:all:techniques}.
Further, we have integrated various uncertainty modeling, posterior inference, prior learning, and post-training calibration approaches into neural PDEs, SPDEs, and neural operators.
An overview of the presented solution methods is provided in Table~\ref{tab:uqt:over}.
Furthermore, we have performed an extensive comparative study, where we addressed a function approximation problem as well as the problems of Fig.~\ref{fig:intro:all:problems}, by employing the related UQ methods of Table~\ref{tab:uqt:over}.
The comparative study serves to demonstrate the applicability and reliability of the presented methods for treating limited and mixed input-output data that are contaminated with various types of noise; for seamlessly combining these data with available physics knowledge; and for harnessing historical data to reduce cost while increasing performance.  
An overview of the considered problems in the comparative study is provided in Table~\ref{tab:comp:contents}. 
In the following, we summarize our contributions, our findings, as well as proposed future research directions.

\input{IN_glossary}

\subsection{Novel contributions}

For each of the problems considered herein, we have also presented original contributions.
For \textit{function approximation}, we have demonstrated how to learn the heteroscedastic noise given historical data by using functional priors; see Section~\ref{sec:comp:func:hetero} for the method and Fig.~\ref{fig:comp:func:hetero:res} for results.
For \textit{forward PDEs}, we have proposed a new technique, termed GP+PI-GAN, which we compared with U-PINNs; see Section~\ref{sec:uqt:sdes} for the GP+PI-GAN method and Figs.~\ref{fig:comp:forw:pinns:15} and \ref{fig:comp:forw:pinns:6} for results.
For \textit{mixed PDEs}, we have trained and combined with existing methods a GAN-based functional prior, termed PI-GAN-FP, given historical data of the source term and the PDE space-dependent parameters for solving the problem faster and more accurately; see Section~\ref{sec:uqt:fpriors} for the method and Table~\ref{tab:pinns:standard} for results.
Further, we have demonstrated how to treat heteroscedastic noise in the source term, problem parameters, and solution data; see Fig.~\ref{fig:uqt:bnns:dataunc:heteroscedastic:upinn} for the method and Fig.~\ref{fig:comp:pinns:hetero} for results.
For \textit{mixed SPDEs}, we have proposed a new NNPC architecture, termed NNPC+, which performs better, and we have considered stochastic problems with noisy realizations; see Fig.~\ref{fig:methods:sdes:nnpc} for the method and Fig.~\ref{fig:comp:stochastic:preds:noisy} for results.
For \textit{operator learning}, we have demonstrated how to deal with noisy and incomplete inference data given a pre-trained neural operator on clean data; see Section~\ref{sec:uqt:fpriors} for the method and Figs.~\ref{fig:comp:don:fp:mean:id}-\ref{fig:comp:don:fp:uncer:id} for results.
In addition, we have proposed different metrics for detecting OOD data in neural operators, which is critical for risk-related applications; see Table~\ref{tab:comp:don:ood:metrics} for the metrics and Fig.~\ref{fig:comp:don:ood} for results.

\subsection{Findings}

Based on the comparative study of Section~\ref{sec:comp:func}, we have found that UQ methods with smaller computational cost (e.g., MFVI and LA), can yield comparable performance with more expensive methods (e.g., HMC and DEns); see Fig.~\ref{fig:comp:func:homosc:unk:res}. 
This is true especially if the NN parameter priors are learned during posterior inference and, clearly, depends on the problem.
Further, based on our experiments, post-training calibration using a small left-out dataset can reduce significantly the calibration error; see, e.g., Fig.~\ref{fig:comp:func:homosc:kno:calsize}.
This is achieved by calibrating the produced uncertainties so that they are neither under- or over-confident.
We have also shown that learning the NN parameter prior during posterior inference helps to reduce the burden of thorough NN architecture selection.
Specifically, we have shown that different architectures yield similar results if the priors are learned; see Fig.~\ref{fig:comp:func:homosc:kno:netsize}.
An additional finding that has also been reported in other studies, e.g., \cite{yao2019quality,ashukha2021pitfalls}, is that performance evaluation and comparison of different UQ methods requires a plurality of evaluation metrics. In particular, we have shown that there are cases in which MCD performs much better than HMC in terms of metrics employed, although it may lead to over-confident predictions, especially for OOD data; compare Fig.~\ref{fig:comp:func:homosc:kno:epist:hmc} (HMC) with \ref{fig:comp:func:homosc:kno:epist:mcd} (MCD). 
Further, we have demonstrated that heteroscedastic noise modeling and learning with online or offline methods, e.g., functional priors, exhibits satisfactory performance. This is true even for computationally less expensive methods, e.g., MFVI; see Fig.~\ref{fig:comp:func:hetero:res}.

In \textit{solving forward PDEs}, we found that the herein proposed GP+PI-GAN method, which first fits the source term data and subsequently propagates the uncertainty to the solution, can outperform U-PINNs in some cases; see Fig.~\ref{fig:comp:forw:pinns:15}.
Nevertheless, because it does not use the PDE for fitting the source term data, it can be over-confident in its predictions; see Fig.~\ref{fig:comp:forw:pinns:6}.
Further study is required to establish these new results.
In \textit{solving mixed PDEs}, we have shown that employing a PI-GAN-FP, pre-trained with historical data of the source term and the problem parameters, exhibits higher performance than the BNN-based UQ methods; compare Fig.~\ref{fig:comp:pinns:stand:hmcfp} (PI-GAN-FP) with Fig.~\ref{fig:comp:pinns:stand:hmc} (standard U-PINN with BNN-FP).
Among HMC, MFVI, MCD, and DEns, the more expensive techniques (HMC and DEns) perform better, while we estimated the unknown noise satisfactorily with HMC, MFVI, and DEns.
For challenging cases involving large noise, data concentrated on one side of the space domain that require extrapolation, and functions with large gradients, the proposed UQ methods still perform satisfactorily, with uncertainties that cover the errors in most cases; see Table~\ref{tab:pinns:add}.
In \textit{solving mixed SPDEs}, the herein proposed U-NNPC+ can outperform existing methods for treating cases with noisy stochastic realizations; see Table~\ref{tab:comp:stochastic}. Nevertheless, we found in our experiments that epistemic uncertainty obtained by combining deep ensembles with the considered methods may not cover the errors in predicting the first- and second-order statistics of the quantities of interest; see Fig.~\ref{fig:comp:stochastic:preds:noisy}.

In \textit{operator learning and treating limited and noisy inference data}, combining a pre-trained DeepONet with a GAN-FP, termed PA-GAN-FP, performs better than the herein proposed PA-BNN-FP, as expected; see Table~\ref{tab:comp:don:fp}. However, the PA-BNN-FP does not incur any additional computational cost and is more calibrated in our experiments. 
In \textit{operator learning and for treating complete and clean inference data}, U-DeepONet, i.e., DeepONet combined with deep ensembles, is more accurate (in terms of mean predictions) than standard DeepONet for OOD data, while the provided  epistemic uncertainty covers the errors for ID and OOD data. The epistemic uncertainty also increases for OOD data, a fact that is harnessed for OOD data detection; see Figs~\ref{fig:comp:don:dens:id}-\ref{fig:comp:don:dens:ood}.
In \textit{operator learning and OOD data detection}, the three proposed detection metrics follow the errors of performing inference with OOD data. Therefore, in practice, for a new input to the DeepONet, we can use the detection metrics to evaluate whether the input has been seen during training; see Fig.~\ref{fig:comp:don:ood}. This can be used to raise a ``red flag'' that the DeepONet output given a new input may not be accurate.

\subsection{Outlook and future research directions}

Neural networks are currently revolutionizing the computational science field by combining physics with data for solving problems of ever-increasing complexity,
and are the building blocks in developing digital twins of physical and biological systems. In such problems, the physics may be available only partially, the data may be limited, noisy, and contaminated with heteroscedastic noise of an unknown distribution, and extrapolation with OOD data may be required. In this context, reliability assessment by UQ is essential for employment of the NN-based methods in risk-related applications and for decision-making. To achieve this, a unified framework is required that classifies the problems of interest, and could provide methods, hyperparameter selection approaches, evaluation metrics, and post-training calibration approaches for avoiding making under- or over-confident predictions. Herein, we have proposed a formulation towards such a unified framework, based on which we have presented several novel contributions for addressing some of the aforementioned challenges of SciML. We have shown that by modifying, only slightly in some cases, standard optimization algorithms, output predictions can be endowed with uncertainty estimates.
These estimates follow the trend of the prediction error and can be used for evaluation of the NN methods and for detecting unseen data during training.  
 
In terms of future research, a direction that would be useful for practical applications is UQ scalability, i.e., extending the presented UQ methods for treating high-dimensional problems as well as large amounts of data. In such cases, the increased computational cost can be addressed by using cheaper methods that are also combined with stochastic gradients; see, e.g., \cite{ma2015complete}. This could be combined with a thorough cost-effectiveness study of the various posterior inference methods we discussed in this paper, as well as of the neural (S)PDE and operator methods.
Preliminary results included in Section~\ref{sec:comp:accvscost} suggest that even UQ methods with associated computational cost slightly higher than standard NN training, such as MFVI, LA, and SEns, can improve mean predictions and provide reliable uncertainty estimates, if properly combined with prior learning approaches, hyperparameter tuning, and post-training calibration.
Clearly, however, the levels of accuracy and confidence that can be tolerated are discipline- and application-dependent.
Further, it would be of interest to also perform a cost-effectiveness study for comparing post-training calibration with active learning approaches. The latter re-adjust the predictions in light of left-out or novel data and are expected to improve performance by incurring some additional computational cost; see, e.g., \cite{ren2021survey}. 
Active learning is particularly important for UQ of rare or extreme events for which very limited data may exist, and hence output-weighted criteria
should be exploited as in \cite{rudy2021outputweighted}. More broadly, UQ methods for multi-fidelity training of neural networks as well as for diverse data augmentation techniques that have to be used in these cases should be investigated in the near future.
This proposed research area is also important for operator learning, but in addition, a future research direction pertains to extending DeepONet for treating noisy input data during the pre-training phase, as shown in Table~\ref{tab:comp:don:problems}.
Furthermore, alternative UQ approaches that could be employed in the context of SciML are based on the evidential framework, e.g., \cite{malinin2018predictive,sensoy2018evidential,amini2020deep,charpentier2020posterior,kopetzki2020evaluating,malinin2020regression,charpentier2021natural,meinert2021multivariate,ulmer2021survey}, the variational information bottleneck, e.g., \cite{alemi2016deep,alemi2018uncertainty,goldfeld2020information}, and the conformal prediction framework, e.g., \cite{angelopoulos2021gentle,dewolf2021valid}.
Lastly, the capabilities of the presented UQ methods should be tested in the context of digital twins of multi-scale and multi-physics systems, where a plurality of methods could be used for different sub-systems, and it would be interesting to assess the synergistic or antagonistic effects of all UQ methods employed together in such a complex environment.

%% file: IN_glossary.tex
\begin{table}[ht]
	\centering
	\footnotesize
	\begin{tabular}{r|l|r|l}
		\toprule
		\multicolumn{4}{c}{\textbf{Glossary of terms frequently used in the paper}}\\
		\midrule
		\midrule
		\multicolumn{4}{c}{\textbf{Terms related to Machine Learning (ML)}}\\
		\midrule
		\multirow{2}{*}{\textbf{NN}} & Neural Network & \textbf{Standard} & Point estimates $\hat{\theta}$ (or $\hat{\beta}$)\\
		& with parameters $\theta$ (or $\beta$) & \textbf{NN training}& according to \S\ref{app:modeling:point}\\
		\midrule
		\multirow{2}{*}{\textbf{FP}} & Functional Prior induced by a model & \textbf{Posterior} & Samples $\{\hat{\theta}_j\}_{j=1}^M$ (or $\{\hat{\beta}_j\}_{j=1}^M$) \\
		 & with random parameters $\theta$ (or $\beta$) &\textbf{inference}& according to \S\ref{sec:uqt:bnns}-\ref{sec:uqt:ens}\\
		 \midrule
		 \multirow{2}{*}{\textbf{BMA}} & Bayesian Model Average of Eq.~\eqref{eq:uqt:pre:bma:new:1} (exact)  & \textbf{Predictive} & Predictive Probability/Cumulative \\
		 & or Eqs.~\eqref{eq:uqt:pre:mcestmc:mcest}-\eqref{eq:uqt:pre:mcestmc:totvar} (approximate) &\textbf{PDF/CDF}&  Density/Distribution Function (Fig.~\ref{fig:eval:eval:outputs})\\
		 \midrule
		 \multirow{2}{*}{\textbf{BNN}} & Bayesian NN - FP induced by a NN &  \textbf{Post-training} & Modified predictive PDF/CDF to \\
		 & with $\theta$ (or $\beta$) following prior distribution &  \textbf{calibration} & improve statistical consistency (\S\ref{sec:eval:calib})\\
		 \midrule
		 \textbf{GAN} & Generative Adversarial Network &  \textbf{MCMC} & Markov chain Monte Carlo \\
		 \midrule
		 \multirow{2}{*}{\textbf{GAN-FP}} & FP induced by a GAN generator &  \textbf{Epistemic} & Posterior uncertainty of $\theta$ (or $\beta$) \\
		 &with random input $\theta$ & \textbf{uncertainty}&quantified via $\{\hat{\theta}_j\}_{j=1}^M$ (or $\{\hat{\beta}_j\}_{j=1}^M$)\\
		 \midrule
		 \multirow{2}{*}{\textbf{h-X}} & Model X with heteroscedastic &  \textbf{Aleatoric} & Irreducible data uncertainty \\
		 &modeling of aleatoric uncertainty & \textbf{uncertainty}& considered known or learned  \\
		 \midrule
		 \multirow{2}{*}{\textbf{U-X}} & Model X with epistemic and aleatoric &  \textbf{Total}& Epistemic + Aleatoric uncertainty \\
		 & uncertainties taken into account & \textbf{uncertainty}& (PDF/CDF or moments) \\
		 \midrule
		 \midrule
		 \multicolumn{4}{c}{\textbf{Terms related to Scientific Machine Learning (SciML)}}\\
		\midrule
		 \multirow{2}{*}{\textbf{(S)PDE}} & (Stochastic) Partial  &  \multirow{2}{*}{\textbf{Operator}} & Mapping between \\
		 & Differential Equation && infinite-dimensional function spaces \\
		 \midrule
		 \textbf{Neural} & \multirow{2}{*}{Solves (S)PDE with a NN} & \textbf{Neural}  & \multirow{2}{*}{Learns operator with a NN}  \\
		  \textbf{(S)PDE} && \textbf{operator} &  \\
		 \midrule
		 \multirow{2}{*}{\textbf{PI-X}} & Physics-Informed method X - &  \multirow{2}{*}{\textbf{PA-X}} & Physics-Agnostic method X -  \\
		  &Uses (S)PDE during training & & Ignores/replaces (S)PDE w/ surrogate \\
		  \midrule
		 \multirow{2}{*}{\textbf{PINN}} & Physics-Informed Neural Network &  \multirow{2}{*}{\textbf{DeepONet}} & Deep Operator Network (PI or PA) \\
		  &for solving PDEs & & for learning operators \\
		\bottomrule
	\end{tabular}
	\caption{
	Glossary of terms frequently used in this paper. 
	The upper part corresponds to machine learning terms, whereas the lower part to scientific machine learning terms. 
	}
	\label{tab:glossary}
\end{table}

%% file: appendix/IN_app_modeling.tex
\section{Supplementary material to Section~\ref{sec:uqt:pre}: Additional topics in total uncertainty modeling}\label{app:modeling}

\begingroup
\etocsetstyle {subsection}
{\leftskip 0pt}
{\leftskip 0pt}
{\bfseries\footnotesize\makebox[1cm][l]{\etocnumber}%
\etocname\nobreak\hfill\nobreak\etocpage
\par
}
{}
\etocsetstyle {subsubsection}
{}
{\leftskip 30pt}
{\mdseries\footnotesize\makebox[1cm][l]{\etocnumber}%
\etocname\nobreak\hfill\nobreak\etocpage
\par
}
{}
\localtableofcontentswithrelativedepth{+3}
\endgroup

\subsection{Statistics of the predictive distribution}\label{app:modeling:stats}

The mean of the random variable $(u|x, \cD)$ is denoted by $\mathbb{E}[u|x, \cD]$ and is given by
\begin{equation}\label{eq:uqt:pre:bma:new:4}
	\hat{u}(x) = \mathbb{E}[u|x, \cD] = \int u_{\theta}(x)p(\theta|\cD) d\theta. 
\end{equation}
Clearly, even if a single output value $\hat{u}$ is required instead of a predictive distribution, its determination involves an integration over all plausible values of $\theta$ given (conditioned on) the data.
Next, utilizing the law of total variance we obtain the diagonal part of the covariance matrix of $(u|x, \cD)$ as (e.g., \cite{zhao2018empirical})
\begin{equation}\label{eq:uqt:pre:bma:totvar:1}
	Var(u|x, \cD) = \underbrace{\mathbb{E}_{\theta|\cD}\left[Var(u|x, \theta)\right]}_{\text{aleatoric}} + \underbrace{Var_{\theta|\cD}(\mathbb{E}[u|x, \theta])}_{\text{epistemic}},
\end{equation} 
which simplifies to 
\begin{equation}\label{eq:uqt:pre:bma:totvar:2}
	Var(u|x, \cD) = \underbrace{\mathbb{E}_{\theta|\cD}\left[\Sigma_{u}^2\right]}_{= \Sigma_{u}^2}  + Var_{\theta|\cD}(u_{\theta}(x))
\end{equation} 
for the Gaussian likelihood function of Eq.~\eqref{eq:uqt:pre:bma:Glike}.
Eqs.~\eqref{eq:uqt:pre:bma:totvar:1}-\eqref{eq:uqt:pre:bma:totvar:2} provide the \textit{total uncertainty} of $(u|x, \cD)$ in the form of total variance.
Note that in Eq.~\eqref{eq:uqt:pre:bma:totvar:1} there are two distinct random variables pertaining to $u$; i.e., $(u|x, \cD)$, which represents the sought value of $u(x)$, is conditioned on the dataset $\cD$, and does not depend on $\theta$, and $(u|x, \theta)$ which is the means for obtaining $(u|x, \cD)$ and depends on $\theta$.
Further, it is straightforward to express similarly the non-diagonal terms of the covariance matrix of $(u|x, \cD)$; see, e.g., \cite{tanno2021uncertainty}.
Furthermore, we emphasize that aleatoric uncertainty affects $Var(u|x, \cD)$ not only via the corresponding added term in Eq.~\eqref{eq:uqt:pre:bma:totvar:2}, but also via epistemic uncertainty, which is expected to increase with more noise.
Nevertheless, it is not generally straightforward to completely disentangle aleatoric and epistemic uncertainties; see the comparative study in Section~\ref{sec:comp} for related experiments as well as Section~\ref{app:methods:gps} for a related discussion pertaining to GP regression.
Finally, if no aleatoric uncertainty in the prediction of $(u|x, \cD)$ is considered, Eq.~\eqref{eq:uqt:pre:bma:new:1} becomes 
\begin{equation}\label{eq:uqt:pre:bma:new:2}
	p(u|x, \cD) = \int \delta(u-u_{\theta}(x))p(\theta|\cD) d\theta.
\end{equation}
Note that this equation is still a probability density with the diagonal part of the covariance matrix of the random variable $(u|x, \cD)$ given by Eqs.~\eqref{eq:uqt:pre:bma:totvar:1}-\eqref{eq:uqt:pre:bma:totvar:2} without the aleatoric uncertainty part. 

\subsection{Point estimates (standard neural network training)}\label{app:modeling:point}

Here we provide a brief and partial overview of standard NN training that serves for making our paper self-contained. 
By maximizing the data likelihood $p(\cD|\theta)$, the model parameters $\theta$ can be inferred. 
This approach is widely known as maximum likelihood estimation (MLE) and when combined with a Gaussian likelihood of the form of Eq.~\eqref{eq:uqt:pre:bma:Glike} it is expressed as
\begin{equation}\label{eq:modeling:point:mle}
	\hat{\theta} = \underset{\theta}{\mathrm{argmin}}~\pazocal{L}(\theta) \text{, where } 
	\pazocal{L}(\theta) = \sum_{i=1}^{N}||u_i-u_{\theta}(x_i)||^2_2.
\end{equation}
In standard function approximation, Eq.~\eqref{eq:modeling:point:mle} is referred to as mean squared error (MSE) minimization without regularization.
Combining MLE with a different likelihood function, e.g., Laplace distribution, yields a different minimization objective in Eq.~\eqref{eq:modeling:point:mle}, e.g., mean absolute error minimization. See also \cite{psaros2021metalearning} for meta-learning the loss function in Eq.~\eqref{eq:modeling:point:mle}.

Alternatively, if we consider the approximation
\begin{equation}\label{eq:modeling:point:map}
	p(\theta|\cD) \approx \delta(\theta-\hat{\theta}),
\end{equation}
we can infer $\theta$ by maximizing the posterior $p(\theta|\cD)$, i.e., 
\begin{equation}\label{eq:modeling:point:map:1}
	\hat{\theta} = \underset{\theta}{\mathrm{argmax}}~p(\theta|\cD),
\end{equation}
or, equivalently,
\begin{equation}\label{eq:modeling:point:map:2}
	\hat{\theta} = \underset{\theta}{\mathrm{argmax}}~\pazocal{L}(\theta) \text{, where } 
	\pazocal{L}(\theta) = \log p(\cD|\theta) + \log p(\theta).
\end{equation}
This is widely known as maximum a posteriori (MAP) estimation.
In conjunction with the Gaussian likelihood of Eq.~\eqref{eq:uqt:pre:bma:Glike}, Eq.~\eqref{eq:modeling:point:map:2} corresponds to regularized MSE minimization.
For example, a Gaussian prior $p(\theta)$ gives rise to $\ell_2$ regularization (also called weight decay in NN literature; e.g., \cite{goodfellow2016deep}).
Utilizing either one of Eqs.~\eqref{eq:modeling:point:mle} and \eqref{eq:modeling:point:map:2}, we end up with a deterministic estimate of $\theta$, and thus $\hat{u}$ is also deterministically estimated as $\hat{u} \approx u_{\hat{\theta}}(x)$.

Finally, if gradient descent with data sub-sampling is used, i.e., stochastic gradient descent (SGD), the update step of MAP at $t$-th iteration is given by
\begin{equation}\label{eq:modeling:point:map:3}
	\theta^{(t+1)} = \theta^{(t)} + \epsilon \left[\nabla_{\theta} \sum_{i\in S}^{} \frac{N}{|S|}\log p(u_i|x_i, \theta^{(t)}) + \nabla_{\theta}\log p(\theta^{(t)})\right],
\end{equation}
where only a mini-batch, i.e., datapoints in the index set $S$ of size $\abs{S}$, are considered and $\epsilon$ denotes the learning rate.
Note that by modifying Eq.~\eqref{eq:modeling:point:map:3} slightly, we show in Section~\ref{sec:uqt} how epistemic and aleatoric uncertainties can be incorporated.
An overview of gradient descent optimization methods can be found in \cite{ruder2017overview}.

\subsection{Hyperparameters and hyperpriors}\label{app:modeling:hypers}

We have assumed in Section~\ref{sec:uqt:pre:bma} that the prior $p(\theta)$ and the parameters of the noise model (i.e., $\sigma_u^2$ for the Gaussian likelihood of Eq.~\eqref{eq:uqt:pre:bma:Glike}) are known and fixed. 
If they are not known, according to the Bayesian framework, they should be integrated out, similarly to Eq.~\eqref{eq:uqt:pre:bma:new:1} involving $\theta$.
For convenience, assume that the prior is parametrized by $\sigma_{\theta}^2$; e.g., $\sigma_{\theta}^2$ can be the variance of each normally distributed component of the parameter vector $\theta$.
Other prior choices are the Laplace, Cauchy, and horseshoe distributions, where $\sigma_{\theta}^2$ can be construed as an unknown scale parameter; see relevant discussions in \cite{nalisnick2018priors,ghosh2019model,fortuin2021bayesian}.
Next, by integrating over $\sigma_{\theta}^2$ and $\sigma_{u}^2$, Eq.~\eqref{eq:uqt:pre:bma:new:1} becomes
\begin{equation}\label{eq:modeling:hypers:new}
	p(u|x, \cD) = \int p(u|x, \theta, \sigma_{u}^2)p(\theta, \sigma_{\theta}^2, \sigma_{u}^2|\cD) d\theta d\sigma_{\theta}^2 d\sigma_{u}^2,
\end{equation}
where $p(\theta, \sigma_{\theta}^2, \sigma_{u}^2|\cD) = p(\theta|\cD, \sigma_{\theta}^2, \sigma_{u}^2)p(\sigma_{\theta}^2, \sigma_{u}^2|\cD)$,
\begin{equation}\label{eq:modeling:hypers:post:1}
	p(\theta|\cD, \sigma_{\theta}^2, \sigma_{u}^2) =\frac{p(\cD|\theta, \sigma_{u}^2)p(\theta|\sigma_{\theta}^2)}{p(\cD|\sigma_{\theta}^2, \sigma_{u}^2)},
\end{equation}
and
\begin{equation}\label{eq:modeling:hypers:post:2}
	p(\sigma_{\theta}^2, \sigma_{u}^2|\cD) =\frac{p(\cD|\sigma_{\theta}^2, \sigma_{u}^2)p(\sigma_{\theta}^2)p(\sigma_{u}^2)}{p(\cD)}.
\end{equation}
This procedure is called hierarchical modeling, $\sigma_{\theta}^2$ and $\sigma_{u}^2$ are called hyperparameters in order to be distinguished from the NN parameters $\theta$, and $p(\sigma_{\theta}^2)$, $p(\sigma_{u}^2)$ are called hyperpriors.	
Clearly, exact integration in Eq.~\eqref{eq:modeling:hypers:new} for real world problems is infeasible, and thus approximation techniques are typically utilized. 
A standard approximation is type-II maximum likelihood (also known as empirical Bayes), in which, in the same spirit of MLE of Section~\ref{app:modeling:hypers}, best hyperparameters are not integrated out, but are selected based on maximizing the likelihood $p(\cD| \sigma_{\theta}^2, \sigma_{u}^2)$ of Eq.~\eqref{eq:modeling:hypers:post:1}.
For each of the BNN methods presented in Section~\ref{app:methods:bnns}, we demonstrate how the hyperparameters can be optimized in practice. 

So far in the discussion of this section we have considered a fixed model $\cH$; e.g., a fixed NN architecture. 
Similarly to Eq.~\eqref{eq:modeling:hypers:new}, different models can also, in principle, be integrated out. 
As was the case with the hyperparameters $\sigma_{\theta}^2$ and $\sigma_{u}^2$, exact integration is not feasible and type-II maximum likelihood can be considered as an approximate way for selecting between competing models.
In Section~\ref{sec:eval} we provide additional metrics for hyperparameter optimization and model design.

\subsection{Posterior tempering for model misspecification}\label{app:modeling:postemp}

In general, for overparametrized NNs that are trained by utilizing finite datasets, the Bayesian posterior $p(\theta|\cD)$ is the optimal candidate distribution, with respect to $\theta$, to be used in the model average of Eq.~\eqref{eq:uqt:pre:bma:new:1}.
Specifically, for a fixed prior, the Bayesian posterior minimizes the Probably Approximately Correct (PAC)-Bayes generalization bound \cite{germain2017pacbayesian,masegosa2020learning}.
This is also corroborated in practice, where the BMA approximated via Eq.~\eqref{eq:uqt:pre:mcestmc:mcest} outperforms point estimate procedures (see, e.g., \cite{wenzel2020how} and references therein).
Nevertheless, this result requires perfect model specification; i.e., that there exists a $\theta$ such that the predictive model $p(u|x, \cD)$ induced by $p(\theta|\cD)$ matches exactly the data-generating process \cite{masegosa2020learning}. 
Indicatively, it requires that the approximated function belongs to the function space of the selected NN architecture and the correct data noise model is used. 	

In this regard, it has been shown that in many cases the BMA performs worse than a point estimate; see, e.g., \cite{wenzel2020how,grunwald2017inconsistency}.
This is often attributed to model misspecification and is also studied in theoretical works \cite{masegosa2020learning}.
A case of model misspecification is when the Gaussian likelihood of Eq.~\eqref{eq:uqt:pre:bma:Glike} is used in calculations, but the data follows, for example, a Student-t distribution (see Section~\ref{app:comp:func:results:student} for an example).
The prior $p(\theta)$, which controls together with the NN architecture the function space of our approximation, may also lead to sub-optimal results using the Bayesian paradigm \cite{fortuin2021bayesian}.

In this regard, a technique that is often used in practice is posterior tempering, i.e., sampling $\theta$ values from $p(\theta|\cD)^{1/\tau}$ instead of the true posterior ($\tau = 1$), where $\tau$ is called temperature.
Specifically, it has been reported in the literature that ``cold'' posteriors, $\tau < 1$, perform better \cite{wenzel2020how,leimkuhler2020partitioned,zhang2019cyclical}, although using a more informed prior can potentially remove this effect \cite{fortuin2021bayesian}.
Cold posteriors can be interpreted as over-counting the available data using $1/\tau$ replications of it, thus, making the posterior more concentrated.
This is also related to calibration discussed in Section~\ref{sec:eval:calib}.
Finally, in Section~\ref{app:methods:bnns} we explain how posterior tempering can be used for each of the presented posterior inference techniques. 
In passing, note that the post-training calibration techniques of Section~\ref{sec:eval:calib} can also prove helpful for addressing model misspecification.   

%% file: appendix/IN_app_methods.tex
\section{Supplementary material to Section~\ref{sec:uqt}: Additional details of employed UQ methods}\label{app:methods}

\begingroup
\etocsetstyle {subsection}
{\leftskip 0pt}
{\leftskip 0pt}
{\bfseries\footnotesize\makebox[1cm][l]{\etocnumber}%
\etocname\nobreak\hfill\nobreak\etocpage
\par
}
{}
\etocsetstyle {subsubsection}
{}
{\leftskip 30pt}
{\mdseries\footnotesize\makebox[1cm][l]{\etocnumber}%
\etocname\nobreak\hfill\nobreak\etocpage
\par
}
{}
\localtableofcontentswithrelativedepth{+3}
\endgroup

\subsection{Gaussian processes (GPs)}\label{app:methods:gps}

Consider for simplicity a one-dimensional function approximation problem solved with GP regression.
We denote the noisy data as $u = u_c + \epsilon_u$, where $u_c$ is the function to be approximated and $\epsilon_u$ is a Gaussian noise with zero mean and variance $\sigma_u^2$.
Then, the data $U = [u_1,\dots,u_N]^T$ on locations $X = [x_1,\dots,x_N]^T$ together with the predicted function $U_c^* = [u^{*}_{c1},\dots,u^{*}_{cN'}]^T$ on test locations $X^* = [x^{*}_1,\dots,x^{*}_{N'}]^T$ follow a joint Gaussian distribution written as
\begin{equation}\label{eq:uqt:pre:bma:gpjoint}
	\begin{bmatrix}
		U\\
		U_c^*
	\end{bmatrix} \sim
	\cN\left(0, 
	\begin{bmatrix}
		K_u(X, X) + \sigma_u^2I & K_u(X, X^*)\\
		K_u(X^*, X) & K_u(X^*, X^*)
	\end{bmatrix}
	\right),
\end{equation}
where $K_u(\cdot, \cdot)$ denotes the GP covariance function evaluated on each dataset pair and $I$ the identity matrix.
Consequently, the mean vector and covariance matrix of $U_c^*$ are given as
\begin{equation}\label{eq:uqt:pre:bma:gpmean}
	mean(U_c^*) = K_u(X^*, X)\left[K_u(X, X) + \sigma_u^2I\right]^{-1}U,
\end{equation}
and
\begin{equation}\label{eq:uqt:pre:bma:gpcov}
	cov(U_c^*) = K_u(X^*, X^*) - K_u(X^*, X)\left[K_u(X, X) + \sigma_u^2I\right]^{-1}K_u(X, X^*),
\end{equation}
respectively.
Eq.~\eqref{eq:uqt:pre:bma:gpcov} provides the epistemic uncertainty of $u$ evaluated at $X^*$.
Further, the total uncertainty of $u$ at $X^*$, i.e., the covariance matrix of $U^* = [u^{*}_{1},\dots,u^{*}_{N'}]^T$, is given by adding $\sigma_u^2I$ to Eq.~\eqref{eq:uqt:pre:bma:gpcov}.

Overall, exactly as in Eq.~\eqref{eq:uqt:pre:bma:totvar:2}, aleatoric uncertainty is added to epistemic uncertainty. 
However, in GP regression we also have a relationship between epistemic and aleatoric uncertainties given by Eq.~\eqref{eq:uqt:pre:bma:gpcov}: higher values of $\sigma_u^2$ make the negative right-hand side term decrease, and thus make $cov(U_c^*)$ increase. 
For large $\sigma_u^2$ values, the negative right-hand side term becomes zero and epistemic uncertainty is given exclusively by the prior uncertainty; i.e., no new information is acquired from a dataset that is too noisy. 
Detailed information on GPs can be found in \cite{rasmussen2003gaussian}; see also \cite{damianou2013deep,raissi2016deep,jakkala2021deep,aitchison2021deep} for combining NNs with GPs.

\subsection{Bayesian methods}\label{app:methods:bnns}

\subsubsection{Markov chain Monte Carlo (MCMC)}\label{app:methods:bnns:mcmc}

\paragraph{Markov chain sampling}\label{app:methods:bnns:mcmc:sampl}

Parameter $\theta$ samples from the posterior can be generated using a Markov chain with $p(\theta|\cD)$ as its invariant (or stationary) distribution, an idea that is foundational for a family of techniques known as Markov chain Monte Carlo (MCMC) techniques \cite{neal1993probabilistic,gelman2015bayesian}. 
First, a Markov chain with initial state $\theta^{(0)}$ (following some initial distribution) and transition probability $Q_t(\theta^{(t+1)}|\theta^{(t)})$ for each state $t$ is defined. 
If $Q_t$ is independent of $t$, the Markov chain is called homogeneous with transition probability $Q$. 
An invariant distribution $q$ is one for which if $\theta^{(t)}$ follows a distribution $q$, then any $\theta^{(t')}$ with $t'>t$ will also follow the distribution $q$. 
Equivalently, any transition $Q_t$ applied to the invariant distribution $q$ leaves $q$ invariant. 
Further, an ergodic Markov chain is one that has a unique invariant distribution, called equilibrium distribution, and regardless of the initial state $\theta^{(0)}$ the same equilibrium distribution is reached in stationarity. 

The goal of MCMC techniques is to construct an ergodic Markov chain with equilibrium distribution $q(\theta)=p(\theta|\cD)$, and subsequently sample from $p(\theta|\cD)$ by sampling from $q(\theta)$; this can be performed by simulating a realization of the Markov chain and discarding a number of ``burn-in'' states until equilibrium is reached. 
After sampling a state, there is no need to start over the Markov chain because of ergodicity, i.e., regardless the sampled state, after a few steps the equilibrium distribution $q$ is reached again. 
We can thus resample from the same realization after a number of ``lag'' steps. 

One way to construct an ergodic Markov chain is by applying a set of base transitions $B_1,\dots,B_K$ in turn (e.g., by changing only a subset of the dimensions of $\theta^{(t)}$ in each transition). 
To this end, $q$ is required to remain invariant with respect to each transition, so that it remains invariant with respect to the overall transition $Q=B_1\cdots B_K$, where the center dot denotes sequential transitions.	
Gibbs sampling is an example of the above scheme.
Specifically, for a $K$-dimensional $\theta$ vector, $K$ transitions $B_1,\dots,B_K$ are considered; in each transition only one dimension of the $t$-th state $\theta^{(t)}$ is changed using the conditional distribution (under $p(\theta|\cD)$) of each dimension given the current values of all the rest.
Although Gibbs sampling is widely used in the context of latent variable models (e.g., \cite{blei2014build}), in deep learning the conditional distributions under the posterior are typically not available.
Nevertheless, Gibbs sampling is useful in the context of the stochastic dynamics method and of hyperparameter optimization.

A more general technique that unlike Gibbs sampling does not require sampling from complex distributions is the Metropolis algorithm. 
Specifically, a new state $\theta^{(t+1)}$ is generated from $\theta^{(t)}$ by first generating a candidate state using a proposal distribution $p_p(\theta^*|\theta^{(t)})$ and then deciding whether to accept the candidate $\theta^*$ or not. 
If it is not accepted, $\theta^{(t+1)}$ is taken to be previous state $\theta^{(t)}$.
The above scheme is called global Metropolis, whereas if each component of $\theta^{(t)}$ is separately modified the scheme is called local Metropolis.
Denoting $\log p(\theta|\cD)$ by $\cL(\theta)$, the steps of global Metropolis are:
\begin{enumerate}
	\item Generate candidate $\theta^*$ from a proposal distribution $p_p(\theta^*|\theta^{(t)})$,
	\item If $\frac{p(\theta^*|\cD)}{p(\theta^{(t)}|\cD)} = \exp(-\cL(\theta^*)+\cL(\theta^{(t)}))\geq 1$, then accept $\theta^*$, \\ otherwise accept $\theta^*$ with probability $\exp(-\cL(\theta^*)+\cL(\theta^{(t)}))$,
	\item If $\theta^*$ is accepted, then set $\theta^{(t+1)} = \theta^*$, otherwise set $\theta^{(t+1)} = \theta^{(t)}$. 
\end{enumerate}
The acceptance criterion at Step 2 is called Metropolis acceptance function and can be shown that, under certain conditions on $p_p$, leads to $p(\theta|\cD)$ as the invariant distribution.
The proposal distribution $p_p(\theta^*|\theta^{(t)})$, among many choices, can be a Gaussian distribution centered at $\theta^{(t)}$; see \cite{neal1993probabilistic}.
The Metropolis algorithm is an important step of Hamiltonian Monte Carlo presented below. 

\paragraph{The stochastic dynamics method}\label{app:methods:bnns:mcmc:stocdyn}

MCMC techniques have been widely used in statistical physics, and thus the terminology originating from their use in statistical physics often accompanies them even when used in domains other than physics. 
For example, in the MCMC literature $p(\theta|\cD) = \exp(-\cL(\theta))/Z$ is called canonical distribution over ``positions'' $\theta$, $\cL(\theta)$ is the energy of the system defined for each $\theta$, and $Z$, the normalization constant, is known as the partition function. 	
For formulating the stochastic dynamics method, a ``momentum'' variable $m$ is also introduced for which there is one-to-one correspondence with the components of positions $\theta$.
The canonical distribution for the phase space of $\theta$ and $m$ is defined as 
\begin{equation}
	p(\theta,m) \propto \exp(-H(\theta,m)),
\end{equation}
where $H(\theta,m) = \cL(\theta) + \pazocal{K}(m)$ is the total energy of the system and is known as the Hamiltonian function, and $\pazocal{K}$ denotes the ``kinetic'' energy. 
In this regard, $\theta$ and $m$ are independent, i.e., the marginal of $\theta$ under $p(\theta,m)$ is equal to $p(\theta|\cD)$, which is the distribution we seek to sample from. 
Thus, a Markov chain that converges to $p(\theta,m)$ can first be constructed, and subsequently $\theta$ samples can be collected by sampling from $p(\theta,m)$ and discarding the $m$ samples.

Overall, the stochastic dynamics method consists of two tasks: (1) sampling states $(\theta,m)$ with fixed energy $H$, and (2) sampling states with different values of $H$. 
In Task 1, we obtain different samples of $\theta$ by following the Hamiltonian dynamics of the system without visiting states of different probability.
This is done by simulating the Hamiltonian dynamics of the system, in which the state $(\theta,m)$ evolves in fictitious time $\tau$.
It can be shown that, in theory, $H$ remains the same for different states, and thus  $p(\theta,m)$ is invariant with respect to transitions related to following a trajectory based on the Hamiltonian dynamics equations. 
Next, in Task 2, we visit states with different probability by changing only the state of $m$. 
This is performed by Gibbs sampling updates of the momentum via its density which is proportional to $\exp(-\pazocal{K}(m))$. 
If, for example, $\pazocal{K}(m) = \norm{m}_2^2/2$, the conditional distributions required by Gibbs sampling are readily available (the components of $m$ have independent Gaussian distributions). 	

In practice, however, Hamiltonian dynamics are simulated using a discretization scheme and finite time steps (e.g., leapfrog method). 
To follow the dynamics for some period of fictitious time $\Delta \tau$, $T = \Delta \tau/\epsilon$ iterations are applied, with $\epsilon$ being the time step; this is the main computational bottleneck of the algorithm. 
As a result, $H$ does not stay exactly the same and MC estimates obtained via the stochastic dynamics method will have some systematic error that goes to zero as $\epsilon$ goes to zero. 
The systematic error is eliminated in the hybrid MC method (HMC; also known as Hamiltonian MC) by merging the stochastic dynamics method with the Metropolis algorithm. 
In contrast to the stochastic dynamics method, in HMC the state reached is only a candidate; it is accepted based on the change in total energy (as in the Metropolis algorithm).
These rejections eliminate the bias introduced by inexact simulation. 
Because of its high degree of accuracy, HMC is often used as the ``ground truth'' for comparing different approximate inference techniques. As reported in \cite{hoffman2013stochastic}, HMC with an acceptance rate around 0.6 exhibits satisfactory performance, and thus in this study we tune the time/leapfrog step with the objective of yielding an acceptance rate close to 0.6 in our computations.
For more theoretical and practical information on HMC the interested reader is directed to \cite{neal1993probabilistic,neal2012mcmc,betancourt2017conceptual,yang2021bpinns}. 	

Next, the Langevin dynamics (LD) method is a computationally cheap version of HMC with $T=1$, i.e., with a single leapfrog iteration.
In fact, in the uncorrected version of LD (all candidates accepted) there is no need for the two-step approach described above and for explicitly representing the momentum at all; one stochastic differential equation, the Langevin equation, is used for sampling new values of $\theta$.
Specifically, for the $t$-th state, the Langevin equation is given as
\begin{equation}\label{eq:methods:bnns:mcmc:stocdyn:sgld:1}
	\theta^{(t+1)} - \theta^{(t)} = -\frac{\zeta^2}{2} \nabla_{\theta} \cL(\theta^{(t)}) + \zeta n,
\end{equation}
where $n = [n_1,\dots,n_K]^T$ and $n_j \sim \cN(0, 1)$ for all $j \in \{1, \dots, K\}$. 
By substituting $\zeta = \sqrt{\epsilon}$ and $\cL(\theta^{(t)}) = -\log p(\cD|\theta^{(t)}) - \log p(\theta^{(t)})$, Eq.~\eqref{eq:methods:bnns:mcmc:stocdyn:sgld:1} becomes
\begin{equation}\label{eq:methods:bnns:mcmc:stocdyn:sgld:2}
	\theta^{(t+1)} = \theta^{(t)} + \frac{\epsilon}{2} \left[\nabla_{\theta} \sum_{i=1}^{N}\log p(u_i|x_i, \theta^{(t)}) + \nabla_{\theta}\log p(\theta^{(t)})\right] + \eta,
\end{equation}
where $\eta = [\eta_1,\dots,\eta_K]^T$ and $\eta_j \sim \cN(0, \epsilon)$ for all $j \in \{1, \dots, K\}$. 	
The interested reader is directed to \cite{xifara2014langevin} for a Metropolis-adjusted LD algorithm.
Note that all the MCMC methods described so far require computations over the whole dataset at every iteration, resulting in high computational cost for large datasets. 
Motivated by SGD, in which at each iteration only a mini-batch of the available data is used, \cite{welling2011bayesian} proposed a technique that combines LD and SGD, referred to as stochastic gradient Langevin dynamics (SGLD). 
Specifically, the resulting parameter update is computed using only a subset of the data and is the same as the update in SGD (Eq.~\eqref{eq:modeling:point:map:3}), except with added Gaussian noise, i.e., 
\begin{equation}\label{eq:methods:bnns:mcmc:stocdyn:sgld:3}
	\theta^{(t+1)} = \theta^{(t)} + \frac{\epsilon}{2} \left[\nabla_{\theta} \sum_{i\in S}^{} \frac{N}{|S|}\log p(u_i|x_i, \theta^{(t)}) + \nabla_{\theta}\log p(\theta^{(t)})\right] + \eta.
\end{equation}	
Further, also motivated by the scalability issues of MCMC techniques, \cite{chen2014stochastic} introduced stochastic gradients within the HMC framework; see also \cite{ma2015complete} for a general framework for constructing MCMC samplers with data sub-sampling.
Furthermore, for including posterior tempering, presented in Section~\ref{app:modeling:postemp}, the term inside the brackets in Eq.~\eqref{eq:methods:bnns:mcmc:stocdyn:sgld:2} can be divided by $\tau$ (the temperature), while the variance of $\eta$ remains unchanged; see \cite{wenzel2020how} for more information.
The optimal value of $\tau$ can be obtained via cross-validation using one of the metrics of Section~\ref{sec:eval} and validation data. 

Finally, see \cite{chen2020accelerating,deng2020nonconvex,deng2021accelerating} for a replica exchange MCMC technique that considers two $\theta$ particles that follow Langevin dynamics with two different temperatures.
Specifically, the sampler swaps between the two particle chains with a certain ratio and at each step samples from the current chain.
Concurrently with the present work, the replica-exchange MCMC technique has been employed in conjunction with DeepONet for operator learning \cite{lin2021accelerated}.
Additional UQ methods based on MCMC can be found, indicatively, in \cite{ahn2012bayesian,li2015preconditioned,ma2015complete,zhang2019cyclical,kim2020stochastic}.

\paragraph{Prior and data uncertainty optimization}\label{app:methods:bnns:mcmc:hypers}

If the likelihood function hyperparameter $\sigma_{u}^2$ is unknown, a deterministic way to optimize it is by alternating between MCMC steps for the parameters $\theta$ and maximum likelihood $p(\cD|\theta, \sigma_u^2)$ steps for $\sigma_u^2$; this is commonly referred to as MC expectation-maximization (e.g., \cite{delyon1999convergence}).  
Further, a Bayesian approach that can be used also if both the prior and likelihood function hyperparameters $\sigma_{\theta}^2$ and $\sigma_{u}^2$, respectively, are unknown, involves using Eq.~\eqref{eq:modeling:hypers:new} and integrating out $\sigma_{\theta}^2$ and $\sigma_{u}^2$.
In this context, the objective of MCMC becomes to sample from $p(\theta, \sigma_{\theta}^2, \sigma_{u}^2|\cD)$ instead of $p(\theta|\cD)$.
A straightforward way to do this is by sampling the parameters and hyperparameters simultaneously according to the posterior expressed as $p(\theta, \sigma_{\theta}^2, \sigma_{u}^2|\cD) \propto p(\cD|\theta, \sigma_{\theta}^2, \sigma_{u}^2)p(\theta| \sigma_{\theta}^2)p(\sigma_{\theta}^2)p(\sigma_{u}^2)$, where $p(\sigma_{\theta}^2), p(\sigma_{u}^2)$ are hyperpriors; see, e.g., \cite{gelman2015bayesian,yang2020bayesian}.
Note that, following this approach, the likelihood term $p(\cD|\theta, \sigma_{\theta}^2, \sigma_{u}^2)$ does not depend explicitly on $\sigma_{\theta}^2$, and thus $\sigma_{\theta}^2$ only appears in the prior part of the posterior. 

An alternative way used in this paper is based on Gibbs sampling; see \cite{gelfand1990illustration,neal1995bayesian,gelman2015bayesian,nalisnick2018priors}.
That is, for each step the parameters are first sampled based on the current hyperparameters, and subsequently the hyperparameters are sampled based on the current parameters.
Specifically, in step $t$, $\theta^{(t)}$ can be sampled based on $p(\theta^{(t)}|\cD, \sigma_{\theta}^{2(t-1)}, \sigma_{u}^{2(t-1)})$ (exactly as done so far with HMC, for example), and subsequently $\sigma_{\theta}^{2(t)}, \sigma_{u}^{2(t)}$ can be sampled based on $p(\sigma_{\theta}^{2(t)}, \sigma_{u}^{2(t)}|\cD, \theta^{(t)}) = p(\sigma_{\theta}^{2(t)}|\theta^{(t)})p(\sigma_{u}^{2(t)}|\cD, \theta^{(t)})$.  
Next, assuming Gaussian prior and likelihood functions, we utilize an inverse-Gamma hyperprior for each of $\sigma_{\theta}^{2}, \sigma_{u}^{2}$; note that the inverse-Gamma distribution is the conjugate prior for Gaussian distribution with unknown variance.
In more detail, $p(\sigma_{\theta}^{2})$ is expressed as
\begin{equation}
	p(\sigma_{\theta}^{2}) = \frac{\sigma_{\theta}^{-2(h_1+1)}}{\Gamma(h_1)h_2^{h_1}}\exp\left(-\frac{1}{\sigma_{\theta}^2h_2}\right),
\end{equation}
where $h_1, h_2$ are hyperparameters, and the conditional distribution $p(\sigma_{\theta}^2|\theta)$ is given as
\begin{equation}
	\begin{alignedat}{20}
		p(\sigma_{\theta}^2|\theta) & \propto p(\sigma_{\theta}^2)p(\theta|\sigma_{\theta}^2) \\
		& \propto \sigma_{\theta}^{-2(h_1+1) - K}\exp\left(-\frac{1}{\sigma_{\theta}^2}\left(\frac{1}{h_2}+\frac{\sum_{j=1}^{K}\theta_j^2}{2}\right)\right),
	\end{alignedat}
\end{equation}
where $K$ is the size of $\theta$.
Using standard transformation of random variables rules, it can be shown that a sample from $p(\sigma_{\theta}^2|\theta)$ can be drawn using a Gamma distribution with shape/scale parametrization according to the following steps: 
\begin{enumerate}
	\item A sample is drawn from a Gamma distribution with shape parameter value equal to $h_1 + K/2$ and scale parameter value equal to $(h_2^{-1}+0.5\sum_{j=1}^{K}\theta_j^2)^{-1}$,
	\item The reciprocal of the Gamma sample is used as a sample from $p(\sigma_{\theta}^2|\theta)$. 
\end{enumerate}

Overall, the initial prior of $\theta$, i.e., before performing Gibbs sampling, is expressed as
\begin{equation}
	p(\theta) = \int p(\theta|\sigma_{\theta}^2)p(\sigma_{\theta}^2)d\sigma_{\theta}^2,
\end{equation}
which, for the selected inverse-Gamma hyperprior, is a Student-t distribution (see, e.g., \cite{nalisnick2018priors}). 
Note that although $p(\theta)$ is nowhere used in the computations within the Gibbs sampling scheme, it is useful for visualization purposes (see Fig.~\ref{fig:comp:func:homosc:unk:dists}) as well as for selecting the hyperparameters $h_1, h_2$.
Further, the learned prior of $\theta$, i.e., after performing Gibbs sampling, is denoted as $\tilde{p}(\theta)$ and given by
\begin{equation}\label{eq:methods:bnns:mcmc:learned:prior}
	\tilde{p}(\theta) = \int p(\theta|\sigma_{\theta}^2)p(\sigma_{\theta}^2|\cD)d\sigma_{\theta}^2,
\end{equation}
where
\begin{equation}
	p(\sigma_{\theta}^2|\cD) = \int p(\sigma_{\theta}^2|\theta)p(\theta|\cD)d\theta.
\end{equation}
That is, the learned prior can be viewed as a simple Gaussian fit, parametrized by $\sigma_{\theta}^2$, to the posterior samples (see Fig.~\ref{fig:comp:func:homosc:unk:dists}).
Overall, during the alternating steps of Gibbs sampling, on the one hand $\tilde{p}(\theta)$ attempts to approximate the posterior $p(\theta|\cD)$ by a simple Gaussian fit, and on the other hand posterior $\theta$ samples attempt to both maximize likelihood, but also stay close to the continuously updated prior $\tilde{p}(\theta)$; see, e.g., Eq.~\eqref{eq:methods:bnns:mcmc:stocdyn:sgld:3} for the latter.

Similarly, for drawing a sample from $p(\sigma_u^2|\cD, \theta)$, the reciprocal of a sample from a Gamma distribution with shape parameter value equal to $h_3 + N/2$ and scale parameter value equal to $(h_4^{-1}+0.5\sum_{i=1}^{N}||u_i-u_{\theta}(x_i)||^2_2)^{-1}$ is used.
The hyperparameters $h_3, h_4$ define the inverse-Gamma hyperprior $p(\sigma_u^2)$ and $N$ is the number of datapoints.
The interested reader is directed to \cite{neal1995bayesian} and \cite{nalisnick2018priors} for extensions considering alternative distributions for the prior and the likelihood function.

\subsubsection{Variational inference (VI)}\label{app:methods:bnns:vi}

\paragraph{Theoretical concepts}\label{app:methods:bnns:vi:theo}

In variational inference (VI) the posterior is approximated by a so-called variational distribution $q_{\omega}(\theta)$ parametrized by $\omega$ \cite{kingma2019introduction}.
The objective used for optimizing $\omega$ is marginal likelihood maximization; i.e., $p(\cD)$ of Eq.~\eqref{eq:uqt:pre:bma:marglike}, which is the probability of observing the training dataset $\cD$ under the model $\cH$.
In this regard, noting that $\log p(\cD)$ is constant with respect to $q_{\omega}(\theta)$, $\log p(\cD)$ is expressed as 
\begin{equation}
	\log p(\cD) = \mathbb{E}_{q_{\omega}(\theta)}\left[\log p(\cD) \right],
\end{equation}
and using $p(\cD,\theta) = p(\cD)p(\theta|\cD)$ as
\begin{equation}
	\log p(\cD) = \mathbb{E}_{q_{\omega}(\theta)}\left[\log \left(\frac{p(\cD,\theta)}{p(\theta|\cD)}\frac{q_{\omega}(\theta)}{q_{\omega}(\theta)}\right) \right],
\end{equation}
or 
\begin{equation}
	\log p(\cD) = \text{ELBO}(\omega)  + KL\left(q_{\omega}(\theta)||p(\theta|\cD)\right),
\end{equation}
where $KL\left(q_{\omega}(\theta)||p(\theta|\cD)\right)$ denotes the KL divergence between the true and the approximate posterior, and the evidence lower bound (ELBO) is given as
\begin{equation}\label{eq:methods:bnns:vi:theo:ELBO:1}
	\text{ELBO}(\omega) = \mathbb{E}_{q_{\omega}(\theta)}\left[\log \left(\frac{p(\cD,\theta)}{q_{\omega}(\theta)}\right) \right] \leq \log p(\cD).
\end{equation}
Obviously, $KL\left(q_{\omega}(\theta)||p(\theta|\cD)\right)$ cannot be computed since $p(\theta|\cD)$ is the sought distribution.
However, by maximizing the ELBO, simultaneously the evidence of the model is maximized and the KL divergence between the true and the approximate posterior is minimized. 
Note, also, that Eq.~\eqref{eq:methods:bnns:vi:theo:ELBO:1} can equivalently be written as
\begin{equation}\label{eq:methods:bnns:vi:theo:ELBO:2}
	\text{ELBO}(\omega) = \int \log\left(p(\cD|\theta)\right)q_{\omega}(\theta) d\theta - KL\left(q_{\omega}(\theta)||p(\theta)\right),
\end{equation}
i.e., maximizing the ELBO encourages $q_{\omega}(\theta)$ to explain the data, while being as close to the prior as possible (penalizing overcomplex models).
Note, finally, that the quality of the approximation depends on the expressiveness of $q_{\omega}(\theta)$.
That is, the richer the family of $q$ distributions considered, the better the resulting bound will be; see, e.g., \cite{barber1998ensemble} for discussion and \cite{foong2019expressiveness} for theoretical results.
Relevant discussions and methods considering various distributions can be found, indicatively, in \cite{graves2011practical,hoffman2013stochastic,blundell2015weight,kingma2015variational,gal2016uncertainty,louizos2017multiplicative,mescheder2018adversarial,khan2018fast,zhao2020probabilistic,bai2020efficient,zhang2021metalearning,unlu2021variational}.

Having determined the optimal parameters $\omega$, $p(\theta|\cD)$ in Eq.~\eqref{eq:uqt:pre:bma:new:1} is substituted by $q_{\omega}(\theta)$, which is straightforward to sample from. 
A MC estimate of the predictive distribution can be obtained via Eq.~\eqref{eq:uqt:pre:mcestmc:mcest} by drawing $M$ samples from the approximate posterior $q_{\omega}(\theta)$.

\paragraph{Computational efficiency and tractability}\label{app:methods:bnns:vi:pract}

Eq.~\eqref{eq:methods:bnns:vi:theo:ELBO:2} can also be written as
\begin{equation}\label{eq:methods:bnns:vi:pract:ELBO:3}
	\cL_{VI}(\omega) = -\sum_{i=1}^{N}\int \log\left(p(u_i|x_i,\theta)\right)q_{\omega}(\theta) d\theta + KL\left(q_{\omega}(\theta)||p(\theta)\right),
\end{equation}
where $\cL_{VI}(\omega) = -\text{ELBO}(\omega)$ is the minimization objective. 	
In this section, we present an estimator of the integral term in Eq.~\eqref{eq:methods:bnns:vi:pract:ELBO:3} and, more specifically, for its gradient with respect to $\omega$ in order to optimize it \cite{gal2016uncertainty}.
This derivative has the general form 
\begin{equation}
	I(\omega) = \nabla_{\omega} \int G(\theta)q_{\omega}(\theta)d\theta.
\end{equation}	
Expressing
\begin{equation}
	q_{\omega}(\theta)=
	\int  q_{\omega}(\theta, \eta)d\eta = \int  q_{\omega}(\theta| \eta)p(\eta)d\eta,
\end{equation}
where $q_{\omega}(\theta| \eta) = \delta(\theta-F(\omega,\eta))$ for some transformation $\theta = F(\omega,\eta)$, we obtain
\begin{equation}
	\int G(\theta)q_{\omega}(\theta)d\theta = \int G(F(\omega,\eta))p(\eta)d\eta ,
\end{equation}
which we can differentiate with respect to $\omega$.
Specifically, $I(\omega)$ becomes
\begin{equation}\label{eq:methods:bnns:vi:pract:int:est}
	I(\omega) =
	\int \left.\nabla_{\theta}G(\theta) \right\vert_{\theta = F(\omega,\eta)} J_{F, \omega}p(\eta)d\eta ,
\end{equation}
where $J_{F, \omega}$ is the Jacobian of the transformation $\omega \mapsto F(\omega,\eta)$.
Therefore, for a one-sample MC estimate of the integral of Eq.~\eqref{eq:methods:bnns:vi:pract:int:est}, we can draw a sample of $\eta$ from $p(\eta)$ and estimate the derivative $I(\omega)$ by $\left.\nabla_{\theta}G(\theta) \right\vert_{\theta = F(\omega,\eta)} J_{F, \omega}F(\omega,\eta)$.
This is also known as path-wise estimator; see \cite{gal2016uncertainty} for a detailed discussion.

If, for computational efficiency, only a mini-batch, i.e., datapoints in the index set $S$ of size $\abs{S}$ are considered, Eq.~\eqref{eq:methods:bnns:vi:pract:ELBO:3} is approximated as
\begin{equation}\label{eq:methods:bnns:vi:pract:ELBO:4}
	\cL_{VI}(\omega) = -\frac{N}{\abs{S}}\sum_{i\in S}^{}\int \log\left(p(u_i|x_i,\theta)\right)q_{\omega}(\theta) d\theta + KL\left(q_{\omega}(\theta)||p(\theta)\right).
\end{equation}
The update step for $\omega$ at $t$-th iteration in a gradient descent optimization scheme, with $\epsilon$ as learning rate, is given by
\begin{equation}\label{eq:methods:bnns:vi:pract:vi:upd}
	\omega^{(t+1)} =  \omega^{(t)} + \epsilon \left[\nabla_{\omega} \sum_{i\in S}^{} \frac{N}{|S|}\log p(u_i|x_i,F(\omega, \eta_i)) - \nabla_{\omega} KL\left(q_{\omega}(\theta)||p(\theta)\right)\right],
\end{equation}
where $\{\eta_i\}_{i=1}^{|S|}$ are samples from $p(\eta)$.
In Eq.~\eqref{eq:methods:bnns:vi:pract:vi:upd} the one-sample estimator of Eq.~\eqref{eq:methods:bnns:vi:pract:int:est} has been considered, while the first derivative with respect to $\omega$ can be evaluated based on the integrand of Eq.~\eqref{eq:methods:bnns:vi:pract:int:est}.
Overall, as seen in Eq.~\eqref{eq:methods:bnns:vi:pract:vi:upd}, the update steps of VI resemble the ones of SGD and SGLD (see Eqs.~\eqref{eq:modeling:point:map:3} and \eqref{eq:methods:bnns:mcmc:stocdyn:sgld:3}, respectively).
Finally, for including posterior tempering, presented in Section~\ref{app:modeling:postemp}, the KL term in Eq.~\eqref{eq:methods:bnns:vi:pract:vi:upd} is commonly multiplied by $\tau$ in practice (e.g., \cite{wenzel2020how}).
Similarly to posterior tempering in MCMC techniques, the optimal value of $\tau$ can be obtained via cross-validation. 

\paragraph{Factorized Gaussian as a variational distribution}\label{app:methods:bnns:vi:mfvi}

A common choice for $q_{\omega}(\theta)$ is a factorized Gaussian posterior given as \cite{kingma2014autoencoding}
\begin{equation}
	q_{\omega}(\theta) = \cN(\theta|\mu, diag(\Sigma^2)) = \prod_{j = 1}^{K} \cN(\theta_j|\mu_j, \Sigma_j^2),
\end{equation} 
where $\mu, \Sigma$ are vectors of size $K$. 
The reparametrization for this case is given as
\begin{equation}
	\theta = F(\omega, \eta) = \mu + \Sigma \odot \eta,
\end{equation} 
and $\omega = \{\mu, \Sigma\}$.
Note that, in general, an approximation in which the variational distribution factorizes over the parameters is often referred to as mean-field approximation (MFVI). Further, in our experiments we found that early stopping can help to improve performance. Specifically, in Section~\ref{sec:comp} we employ validation data and the negative log-likelihood metric to determine the optimal number of training steps for early stopping; see, e.g., \cite{sohn2015learning} for a similar implementation.

\paragraph{Dropout as a Bernoulli variational distribution}\label{app:methods:bnns:vi:mcd}

Stochastic regularization relates to a family of techniques that are used to regularize NNs through the injection of stochastic noise into the model. 
One such technique, dropout, follows exactly standard NN training (including standard weight-decay regularization), but in every iteration some of the parameters are set to zero when computing the gradient descent step \cite{srivastava2014dropout}. 
Equivalently, the weight matrices are multiplied by diagonal matrices with 1's and 0's in the diagonal for randomly deleting parameters in each iteration. 

\citet{gal2016uncertainty} has shown that by expressing the above matrix multiplication as reparametrization, dropout is equivalent with BNNs trained with VI, given that the prior over the parameters and the approximate posterior in VI have a specific form. 
NNs trained with dropout can thus be used exactly as BNNs; they can be used for making predictions using Eq.~\eqref{eq:uqt:pre:mcestmc:mcest} (multiple forward passes), where $\{\hat{\theta}_j\}_{j=1}^M$ are obtained by multiplying the optimal parameters by binary vectors drawn from a Bernoulli distribution.
Equivalently, a set of optimal parameters are first obtained by training using dropout and then samples from the approximate posterior are just noisy versions of these parameters. 
This is called MC-dropout (MCD) in order to be distinguished from ``standard'' dropout.
Standard dropout approximates the model average of Eq.~\eqref{eq:uqt:pre:mcestmc:mcest} using only one forward pass with the obtained set of parameters multiplied by a deterministic factor.
Finally, different reparametrizations lead to different versions of dropout, e.g., Gaussian dropout.
See \cite{boluki2020learnable} for relevant discussion and an extension of MCD.

\paragraph{Prior and data uncertainty optimization}\label{app:methods:bnns:vi:hypers}

The prior $p(\theta)$ enters VI in the KL term of Eq.~\eqref{eq:methods:bnns:vi:theo:ELBO:2}, whereas the data uncertainty enters in the likelihood term. 
As explained in Section~\ref{app:modeling:hypers}, the prior likelihood function hyperparameters $\sigma_{\theta}^2$ $\sigma_{u}^2$ can be optimized using type-II maximum likelihood. 
Within the context of VI, marginal likelihood is approximated with the ELBO, which can be used for hyperparameter optimization as well. 
Employing an alternating optimization scheme, in each step of the VI optimization, the variational parameters $\omega$ can be updated based on ELBO with fixed $\sigma_{\theta}^2$ $\sigma_{u}^2$, and subsequently $\sigma_{\theta}^2$ $\sigma_{u}^2$ can be updated based on ELBO with fixed $\omega$.   

\subsubsection{Laplace approximation (LA)}\label{app:methods:bnns:lapl}

\paragraph{Theoretical concepts}

The Laplace method for approximating the posterior of a BNN was first proposed by \cite{denker1991transforming}. 
For fixed hyperparameters, it involves optimizing the NN parameters according to Section~\ref{app:modeling:point}, and subsequently fitting a Gaussian distribution to the discovered mode $\hat{\theta}$, with width determined by the Hessian of the training error. 
This is equivalent to using a quadratic approximation for the logarithm of the probability density. 
In this regard, the covariance matrix of the fitted distribution is the inverse of the Hessian $A$ of the loss at $\hat{\theta}$, which is expressed as $A = -\nabla_{\theta\theta}^2\cL(\hat{\theta}) $, where $\cL(\hat{\theta}) = \log p(\hat{\theta}|\cD)$.
Further, the posterior $p(\theta|\cD)$ is approximated as (see, e.g., \cite{mackay1995probable}))
\begin{equation}\label{eq:methods:bnns:lapl:post}
	q(\theta) = \frac{1}{Z} \exp(-\cL(\hat{\theta}) - \frac{1}{2}(\theta-\hat{\theta})^T A(\theta-\hat{\theta})),
\end{equation}
where $Z$ is a normalization constant.
Based on the approximation of Eq.~\eqref{eq:methods:bnns:lapl:post}, the log-evidence $\log p(\cD|\sigma_{\theta}^2, \sigma_{u}^2)$ of Eq.~\eqref{eq:uqt:pre:bma:marglike} can be approximated as
\begin{equation}\label{eq:methods:bnns:lapl:evi}
	\log p(\cD|\sigma_{\theta}^2, \sigma_{u}^2) \approx \cL(\hat{\theta}) - \frac{1}{2}\log \det A + \frac{K}{2}\log 2\pi - \log Z_{\theta}- \log Z_u,
\end{equation}
where $\log Z_{\theta}$, $\log Z_u$ are the normalization constants of the prior $p(\theta|\sigma_{\theta}^2)$ and the data likelihood $p(\cD|\theta, \sigma_{u}^2)$, respectively; in practice $\log Z_{\theta}$, $\log Z_u$ are omitted (see e.g., \cite{immer2021scalable, daxberger2021laplace}).

Because obtaining the exact Hessian is computationally expensive, the Fisher information matrix and the generalized Gauss-Newton matrix are commonly utilized as Hessian approximations. 
Indicatively, according to the generalized Gauss-Newton approximation, the Hessian $A = -\nabla_{\theta\theta}^2\cL(\hat{\theta}) $ is approximated using only first-order information as
\begin{equation}\label{eq:methods:bnns:lapl:GNN}
	A \approx -\nabla_{\theta}\cL(\hat{\theta})\left[\nabla_{\theta}\cL(\hat{\theta})\right]^T.
\end{equation}
Nevertheless, $A$ in Eq.~\eqref{eq:methods:bnns:lapl:GNN} is a full matrix that must be inverted for computing the covariance matrix.
To reduce the determinant and matrix inversion computational cost, in practice a diagonal factorization of the above approximations is considered.
Alternative approximations can be obtained by using block-diagonal factorizations (typically known as KFAC, standing for Kronecker-factored approximate curvature).
More information regarding Hessian approximations can be found in \cite{martens2014new,ritter2018scalable,ritter2018online,lee2020estimating,daxberger2021laplace,immer2021scalable,immer2021improving}. 
In passing, it is noted that an additional step to reduce the computational cost is to consider the approximate posterior of Eq.~\eqref{eq:methods:bnns:lapl:post} only for a subset of the NN parameters (e.g., only the last layer) and use the MAP estimate for the rest \cite{daxberger2021laplace, daxberger2021bayesian}. 

For making predictions, a MC estimate of the predictive distribution can be obtained by drawing $M$ samples from the approximate posterior $q(\theta)$, which is a Gaussian distribution with mean $\hat{\theta}$ and covariance matrix $A^{-1}$.
The obtained samples $\{\hat{\theta}_j\}_{j=1}^M$ can be used for making predictions according to Eq.~\eqref{eq:uqt:pre:mcestmc:mcest}. 
However, it has been reported in the literature that the accuracy of the MC estimate deteriorates if an approximation of the Hessian $A$ is considered (see, e.g., \cite{daxberger2021laplace}).
This can be attributed to the fact that the new approximate posterior corresponds to a linearized NN model, different from $u_{\theta}$.
Therefore, as proposed in \cite{immer2021improving}, this linearized model should instead be used for predictions. 
Following \cite{immer2021improving}, for one-dimensional outputs $u$, the epistemic part of the total variance of Eq.~\eqref{eq:uqt:pre:mcestmc:totvar} is computed based on the analytical expression
\begin{equation}\label{eq:methods:bnns:lapl:linear}
	\bar{\sigma}_e^2(x) = \left[\nabla_{\theta}u_{\hat{\theta}}(x)\right]^T A^{-1} \nabla_{\theta}u_{\hat{\theta}}(x).
\end{equation}
In the present study, we compute the full matrix of $A$ using Eq.~\eqref{eq:methods:bnns:lapl:GNN} to achieve optimal performance and also use Eq.~\eqref{eq:methods:bnns:lapl:linear} for obtaining epistemic uncertainty.
Finally, for including posterior tempering, presented in Section~\ref{app:modeling:postemp}, the Hessian $A$ in Eq.~\eqref{eq:methods:bnns:lapl:post} can be divided by $\tau$; i.e., small values of $\tau$ make the posterior more concentrated.
Similarly to posterior tempering in MCMC and VI techniques, the optimal value of $\tau$ can be obtained via cross-validation. 

\paragraph{Prior and data uncertainty optimization}\label{app:methods:bnns:lapl:hypers}

Type-II maximum likelihood based on Eq.~\eqref{eq:methods:bnns:lapl:evi} and as presented in Section~\ref{app:modeling:hypers} can be used to optimize the prior and the likelihood function hyperparameters $\sigma_{\theta}^2$ and $\sigma_{u}$, respectively. 
This approach leads to an alternating optimization scheme similar to the one presented for VI in Section~\ref{app:methods:bnns:vi:hypers}, where the hyperparameters are optimized online.
As an approximation to marginal likelihood, the ELBO of Eq.~\eqref{eq:methods:bnns:vi:theo:ELBO:2} can be used by replacing $q_{\omega}$ with $q$ of Eq.~\eqref{eq:methods:bnns:lapl:post}. 
However, in both cases computation of the Hessian $A$ is required for updating $\sigma_{\theta}^2$ and $\sigma_{u}^2$, which renders this approach too computationally expensive for high-dimensional $\theta$. 
To address this issue, approximations can be considered for $A$ or the hyperparameter updates can be performed only every several iterations and not at every iteration.	
Instead of updating the hyperparameters alternately with the parameters $\theta$, a different approach involves optimizing the hyperparameters after completing MAP optimization. 
In this case, the Hessian $A$ makes use of the optimized hyperparameters for making predictions.
Clearly, always hyperparameters can also, alternatively, be optimized in a cross-validation manner using validation data and one of the metrics of Section~\ref{sec:eval}, e.g., predictive log-likelihood.

\subsection{Ensembles}\label{app:methods:ens}

\subsubsection{Deep ensembles (DEns)}\label{app:methods:ens:deep}

Posterior inference by deep ensembles (DEns) pertains to obtaining $M$ different settings of parameters $\theta$ by training the approximator NN (Section~\ref{app:modeling:point}) independently $M$ times using different parameter initializations \cite{lakshminarayanan2017simple}. 
Because the philosophies of DEns and the Bayesian framework are similar \cite{wilson2020bayesian}, DEns can be interpreted as approximating the BMA of Eq.~\eqref{eq:uqt:pre:bma:new:1}. 
Precisely, DEns approximates the posterior $p(\theta|\cD)$ using a mixture of delta distributions located at $\{\hat{\theta}_j\}_{j=1}^M$, which are different MAP estimates arising from random initialization of the NN.
Therefore, Eq.~\eqref{eq:uqt:pre:bma:new:1} is approximated by Eq.~\eqref{eq:uqt:pre:mcestmc:mcest} with $M$ values of $\hat{\theta}$ which are not anymore samples from an approximate posterior, but rather $M$ modes of the true posterior obtained by $M$ MAP optimization runs.
In passing, it can be shown that for linear models DEns provides samples that follow the posterior, unlike for standard NNs \cite{matthews2018samplethenoptimize}.
Finally, \citet{pearce2020uncertainty} have proposed a modification to standard DEns that leads to approximate Bayesian inference, i.e., approximate samples from the posterior, even for NN models.

\subsubsection{Snapshot ensembles (SEns)}\label{app:methods:ens:snap}

\begin{figure}[!ht]
	\centering
	\includegraphics[width=.6\linewidth]{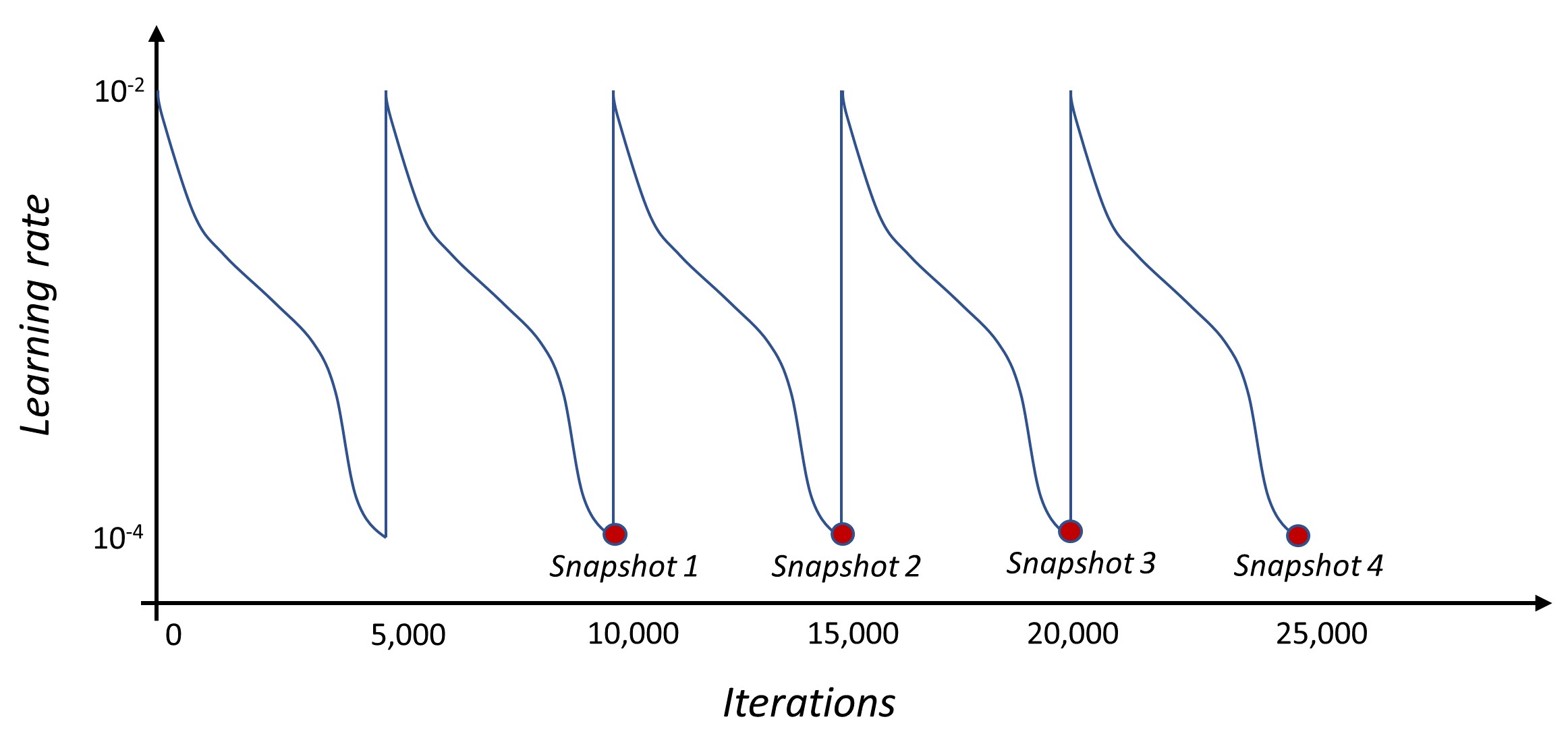}
	\caption{Learning rate schedule example for snapshot ensembles (SEns).
		The learning rate is initially decreased and after converging to a local minimum it is increased abruptly, letting the algorithm escape the current local minimum, and then decreased again for convergence at a different local minimum.
		After a certain number of iterations (e.g., 10,000 in the figure), a snapshot of NN parameters is captured before every restart to be used in the ensemble.}
	\label{fig:methods:ens:snap:lrsched}
\end{figure}

The purpose of snapshot ensembles (SEns) is to perform BMA without incurring any additional cost as compared to standard NN training.
Instead of training $M$ different networks, standard training algorithms (Section~\ref{app:modeling:point}) are let to converge to $M$ different local optima (see, e.g., \cite{huang2017snapshot}). 
The learning rate is initially decreased and after converging to a local minimum it is increased abruptly, letting the algorithm escape the current local minimum, and then decreased again for convergence at a different local minimum.
Before every restart a snapshot is captured. 
A convenient way to implement this cyclical learning rate is by using a cosine annealing learning rate schedule; the range of learning rates and the number of epochs until the next restart can be treated as hyperparameters.
Note that a larger restart learning rate leads to a stronger perturbation of the NN parameters and possibly to increased model diversity.
The total training budget (e.g., 25,000 iterations) is divided into $M$ cycles (e.g., $M=5$) and a snapshot is collected at the end of each cycle.
Fig.~\ref{fig:methods:ens:snap:lrsched} provides an illustration of the SEns training procedure.
Note that DEns and SEns can be combined for forming a multi-snapshot ensemble (Multi-SEns) of $M \times \tilde{M}$ models, which consists of $\tilde{M}$ training trajectories, each one consisting of $M$ snapshots. 

\subsubsection{Stochastic weight averaging (SWA) and SWA-Gaussian (SWAG)}\label{app:methods:ens:swag}

According to stochastic weight averaging (SWA) proposed by \citet{izmailov2019averaging}, parameter snapshots captured during standard training (Section~\ref{app:modeling:point}) are averaged for producing the final prediction; this is in contrast with SEns that involve ensembling on the prediction level.
Similarly to SEns, training involves a cyclical learning rate (but not necessarily) and the computational cost is the same as in standard NN training.
Clearly, the main goal of SWA is to improve accuracy through ensembling, although uncertainty estimates can, in principle, be obtained by using the various captured parameter settings as in SEns. 

An extension to SWA, called SWA-Gaussian (SWAG), that also provides uncertainty estimates has been proposed by \citet{maddox2019simple}.
In SWAG, a Gaussian distribution is fitted to the snapshots of SEns.
Specifically, SWA is used for estimating the mean of the Gaussian distribution and two different approaches for estimating the covariance matrix can be used. 
The first approach involves fitting a diagonal covariance matrix; a running average of the parameters and of the squares of the parameters is maintained (updated once per iterations) and the covariance matrix is computed at the end of training. 
The second approach involves estimating the covariance as the sum of the diagonal covariance and a low-rank matrix given as
\begin{equation}
	\Sigma = \frac{1}{Q-1}\sum_{j=1}^{Q}(\theta_j-\bar{\theta}_j)(\theta_j-\bar{\theta}_j)^T,
\end{equation}
where $\bar{\theta}$ is the running average of the parameters.
Note that $Q$ defines the rank of the matrix and can be regarded as a hyperparameter of the method.
In this study we used $Q=M$, i.e., the rank is equal to the number of snapshots obtained by using SEns.
After fitting a covariance matrix it is straightforward to draw samples of the NN parameters and perform BMA using the MC estimate of Eq.~\eqref{eq:uqt:pre:mcestmc:mcest} for making predictions.
By combining SWAG with DEns, referred to as Multi-SWAG, i.e., by performing the above fitting procedure to each DEns trajectory, we can obtain even better uncertainty estimates. 

\subsection{Functional priors (FPs)}\label{app:methods:fpriors}

\begin{figure}[!ht]
	\centering
	\includegraphics[width=.7\linewidth]{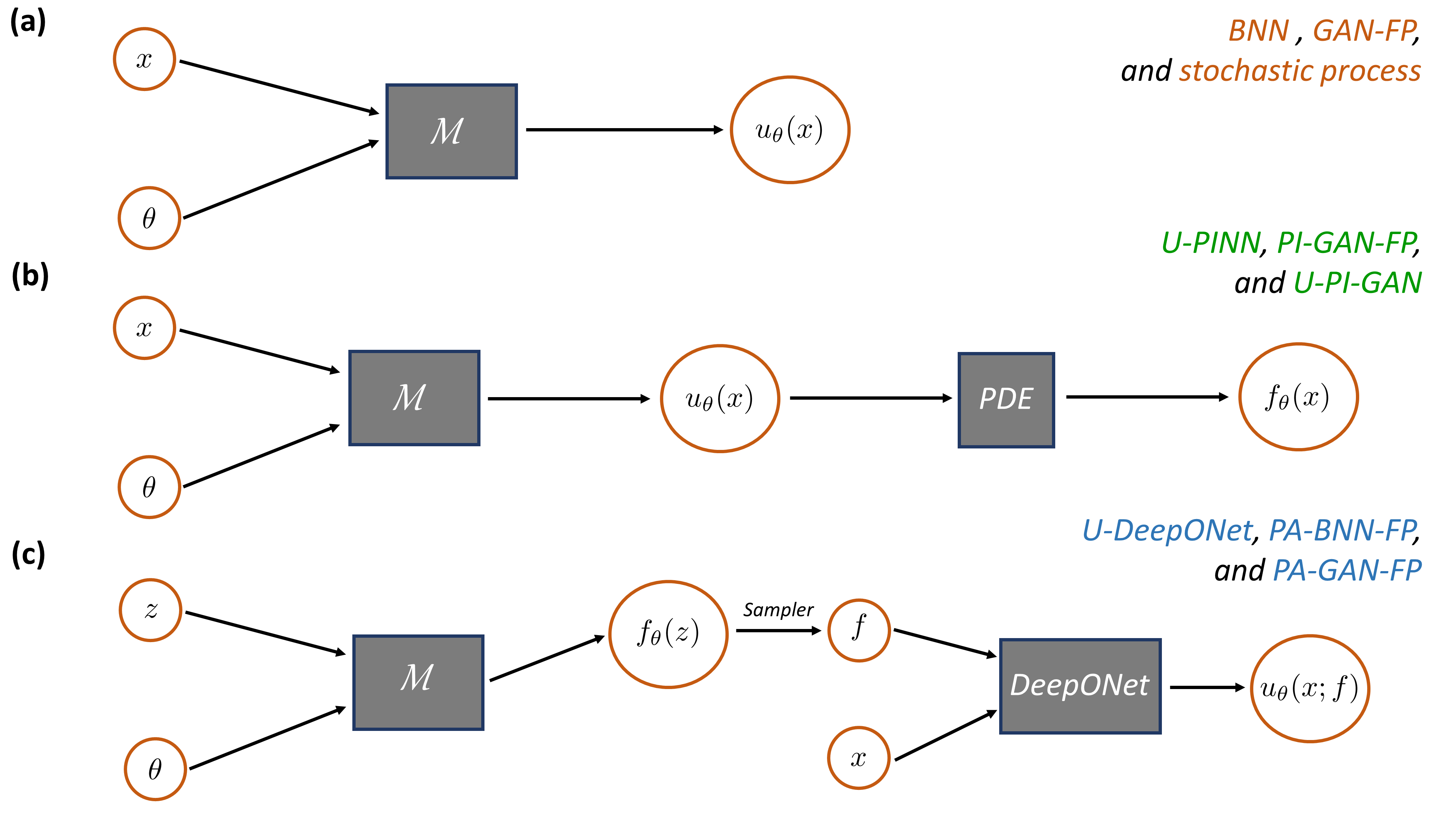}
	\caption{Repeated Fig.~\ref{fig:uqt:fpriors:FP} from main text for convenience of the reader. \textbf{(a)} BNN, GAN functional prior (GAN-FP) and stochastic process representation.
	\textbf{(b)} U-PINN, PI-GAN-FP, and U-PI-GAN.
	\textbf{(c)} U-DeepONet, PA-BNN-FP, and PA-GAN-FP.
	$\pazocal{M}$ expresses how $x$ and $\theta$ are combined to yield $u_{\theta}(x)$. 
		In standard NN models (BNN, U-PINN, U-DeepONet), $\pazocal{M}$ is a BNN.
		In generator-based models (stochastic process, FP, PI-GAN-FP, U-PI-GAN, and PA-GAN-FP), $\pazocal{M}$ is the generator of a GAN that is trained using stochastic realizations or pre-trained using historical data.
		All models depicted in this figure can be construed as special cases of Fig.~\ref{fig:intro:all:techniques}; note, however, the different meaning of $\theta$ for each technique.
		Further, in part (c), $\lambda$ (or both $f$ and $\lambda$) could be used as input to the DeepONet instead of $f$, as in Fig.~\ref{fig:intro:piml:don:arch}.
		Furthermore, $u(x; f)$ is equivalent to $u(x; \xi)$, because each $\xi \in \Xi$ is associated with an input $f$. 
		The additional input $z$ is used to emphasize that the locations in the input domains of $f$ and $u$ can be different and the ``sampler'' that the function $f_{\theta}(z)$ must be sampled at pre-specified locations before entered as input to the DeepONet. 
		See Table~\ref{tab:uqt:over} for an overview of the UQ methods and Table~\ref{tab:glossary} for a glossary of used terms.
	}
	\label{fig:uqt:fpriors:FP:addedapp}
\end{figure}

Consider Fig.~\ref{fig:uqt:fpriors:FP:addedapp}a.
For given $\pazocal{M}$ and paired data of $(x, u)$, the aim of standard NN training is to obtain optimal parameters $\theta$ to be used for making predictions $u_{\theta}(x)$. 
Next, for making training more efficient, meta-learning \cite{hospedales2020metalearning} can be used to obtain, among other choices, a better optimization algorithm, or a better $\pazocal{M}$ (architecture), or a better initialization of $\theta$. 
In cases where the procedure is performed offline, i.e., one of the above components is pre-trained, the objective of meta-learning is to maximize predictive performance at ``test-time'', i.e., for solving a novel problem; see also \cite{psaros2021metalearning} for a detailed view of meta-learning from a PINN perspective. For example, a better $\pazocal{M}$ may correspond to a pre-trained black-box function that takes as input $x$ and $\theta$, and outputs $u_{\theta}(x)$. 
Such a function has its own parameters, denoted herein by $\beta$, and can make learning $\theta$ at test-time more efficient.

Next, in the context of posterior inference according to Sections~\ref{sec:uqt:bnns}-\ref{sec:uqt:ens}, meta-learning relates to obtaining either a better posterior approximation algorithm, or a better $\pazocal{M}$, or a better prior distribution $p(\theta)$, indicatively.
For example, the online prior optimization procedures we presented in Sections~\ref{app:methods:bnns:mcmc:hypers}, \ref{app:methods:bnns:vi:hypers}, and \ref{app:methods:bnns:lapl:hypers} seek to make the learning procedure more efficient and the obtained predictive model better performing.
As for offline procedures, indicatively, the techniques presented in \cite{flam-shepherd2017mapping,hafner2018noise,yang2019outputconstrained} construct various different FPs by constructing $p(\theta)$ so that the BNN-induced FP (BNN-FP) behaves like a GP, or it assigns high probability to the data and to perturbations of it, or it constraints the values that the function is allowed to take in certain parts of the input space, respectively.
See also \cite{pearce2020expressive,nalisnick2021predictive,fortuin2021priors} for additional techniques and a review.

In the following, suppose that we have a dataset $\pazocal{D}_h = \{U_i\}_{i=1}^{N_h}$, with $U_i = \{(x_j^{(i)}, u_j^{(i)})\}_{j=1}^{N_u}$, containing $N_h$ historical realizations. 
The main idea of \citet{meng2021learning} is to utilize, as $\pazocal{M}$ in Fig.~\ref{fig:uqt:fpriors:FP:addedapp}, a pre-trained model parametrized by $\beta$ instead of just a NN architecture.
Specifically, it has been shown that if $u_{\theta}(x)$ is the output of a pre-trained generator $u_{\theta}(x; \beta)$, whose distribution over $\theta$ matches the historical data distribution $P_{data}$, posterior inference of $\theta$ at test-time can be performed faster and with less amount of data.
Additionally, such a procedure allows the use of much lower-dimensional $\theta$, even of size $10$-$100$, as opposed to the high-dimensional $\theta$ corresponding to a typical NN architecture.  
For example, by using a type of generative adversarial network (GAN), called Wasserstein GAN with gradient penalty (WGAN-GP), we can train a generator NN parametrized by $\beta$ whose underlying distribution $P_{gen}$ approximates $P_{data}$ by minimizing 
\begin{equation}\label{eq:methods:fpriors:gloss}
	\cL_G(\beta) = -\mathbb{E}_{\theta \sim p(\theta)}\left[D_{\gamma}(u_{\theta}(X; \beta))\right].
\end{equation}  
In this equation, $u_{\theta}(X; \beta)$ denotes the output of the generator for fixed $\theta$ and $\beta$ evaluated on $X = [x_1,\dots,x_{N_{u}}]$ and can be seen as a point in an $N_{u}$-dimensional vector space.
Further, in Eq.~\eqref{eq:methods:fpriors:gloss}, $D_{\gamma}(\cdot)$ denotes a discriminator NN parametrized by $\gamma$, which aims to approximate the Wasserstein-1 distance between $P_{data}$ and $P_{gen}$.
This is achieved by minimizing the objective
\begin{equation}\label{eq:methods:fpriors:dloss}
	\cL_D(\gamma) = \mathbb{E}_{\theta \sim p(\theta)}\left[D_{\gamma}(u_{\theta}(X; \beta))\right] - \mathbb{E}_{S\sim P_{data}}\left[D_{\gamma}(S)\right] + \zeta \mathbb{E}_{\hat{S}\sim P_i}\left(\norm{\nabla_{\hat{S}}D_{\gamma}(\hat{S})}_2 - 1\right)^2,
\end{equation}  	 
where $P_i$ is the distribution induced by uniform sampling on interpolation lines between independent samples of $S\sim P_{data}$ and $u_{\theta}(X; \beta)$, and $\zeta$ is the gradient penalty coefficient, set as equal to 10 in the present study; see more information in \cite{gulrajani2017improved}.
Because $u_{\theta}(\cdot; \beta)$ is assumed to be noiseless, in order for $P_{gen}$ to match $P_{data}$, we have to synthetically contaminate the functions produced by $u_{\theta}(\cdot; \beta)$ with the assumed amount of noise in the historical data, denoted by $\sigma_h^2$.
If $\sigma_h^2$ is unknown, it can be optimized simultaneously with the generator and discriminator parameters; see Fig.~\ref{fig:comp:func:hetero:res} for an example. 
Finally, regardless of the amount of noise in the historical data used to optimize $\beta$, at test-time the new dataset $\cD = \{x_i, u_i\}_{i = 1}^N$ can have a different amount of noise. 
This is facilitated by the fact that the generator output $u_{\theta}(\cdot; \beta)$ is noiseless and the fact that pre-training of the generator is a separate procedure than learning $\theta$ at test-time.  

\subsection{Solving stochastic PDEs (SPDEs) with U-NNPC(+)}\label{app:methods:sdes} 

\begin{figure}[!ht]
	\centering
	\includegraphics[width=.7\linewidth]{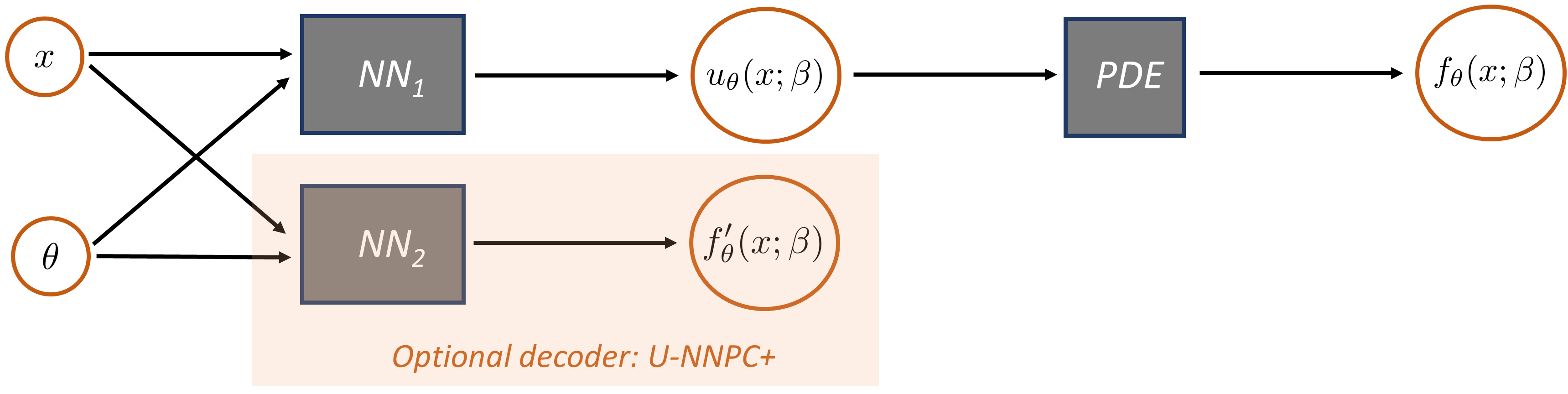}
	\caption{Repeated Fig.~\ref{fig:methods:sdes:nnpc} from main text for convenience of the reader.
	NNPC, developed by \cite{zhang2019quantifying}, and NNPC+ (orange box), proposed herein, for solving SPDEs.
		NNs 1 and 2, parametrized by $\beta$, combine the location in the domain $x$ to produce the modes of the stochastic representations of $u_{\theta}(x; \beta)$ and $f'_{\theta}(x; \beta)$, respectively.
		The model depicted in this figure can be construed as a special case of Fig.~\ref{fig:intro:all:techniques}; note however the different meaning of $\theta$ in this technique.
	}
	\label{fig:methods:sdes:nnpc:addedapp}
\end{figure}

NNPC, depicted in Fig.~\ref{fig:methods:sdes:nnpc:addedapp}, was proposed in \cite{zhang2019quantifying}; see also \cite{zhang2020learning}.
For the forward problem, given the dataset $\cD = \{F_i\}_{i=1}^N$ with $F_i = \{x_j^{(i)}, f_j^{(i)}\}_{j=1}^{N_f}$, Principal Component Analysis (PCA) is first performed for the $f$ data.
This yields the low dimensional representation of the realizations, which is expressed as $\theta = \Phi^T\sqrt{Q}^{-1}f = \{\theta_j\}_{j=1}^K$, where $\Phi$ and the diagonal matrix $Q$ contain the eigenvectors and the eigenvalues of the covariance matrix of the $N$ realizations of $f$, respectively; and a measurement $f$ is a vector of size $N_f$.
Further, given a value of $\theta$, $u_{\theta}(x; \beta)$ is expressed as
\begin{equation}\label{eq:methods:sdes:umodes}
	u_{\theta}(x; \beta) = \sum_{p = 0}^{P}u_p(x; \beta)\rho_{p}(\theta),
\end{equation}	
where each $\rho_{p}$ represents one of the $P+1$ considered polynomial basis functions and each $u_p(\cdot; \beta)$ represents one of the $P+1$ arbitrary polynomial chaos modes of $u$. 
Each $u_p(\cdot; \beta)$ is one of the outputs of a NN (specific to $u$) parametrized by $\beta$.
See \cite{zhang2019quantifying} for implementation information. 
Finally, $\beta$ is trained such that $f_{\theta}(x; \beta)$ in Fig.~\ref{fig:methods:sdes:nnpc:addedapp} approximates the data of $f$.

As an extension, a Karhunen-Lo\`{e}ve expansion in the form 
\begin{equation}\label{eq:methods:sdes:fmodes}
	f'_{\theta}(x; \beta) = f'_{0}(x; \beta) + \sum_{j = 1}^{S}\sqrt{q_j}f'_{j}(x; \beta)\theta_j,
\end{equation}
can also be utilized for $f$. 
In this equation, each $f'_{j}(\cdot; \beta)$ represents one of the mode functions of $f$ and is one of the outputs of an additional NN (specific to $f$), also parametrized by $\beta$ for notation simplicity.
This NN can be construed as a decoder NN, e.g., \cite{goodfellow2016deep}, because it takes as input the PCA encoding of $f$, which is expressed by $\theta$, and outputs $f'_{\theta}(x; \beta)$ in Fig.~\ref{fig:methods:sdes:nnpc:addedapp}, which is trained to approximate the data of $f$.
Finally, $\beta$ is also trained such that $f_{\theta}(x; \beta)$ in Fig.~\ref{fig:methods:sdes:nnpc:addedapp} approximates $f'_{\theta}(x; \beta)$.
Note that when performing PCA for obtaining $\theta$, some information of $f$ is lost. Therefore, the rationale of NNPC+ is that the output $f_{\theta}$ of the PDE is not compared, in a loss function during training, with the data of $f$. Instead, it is compared with the output of a decoder that is trained to approximate the data of $f$ given the low-dimensional input $\theta$.
Clearly, this added expressivity through $\text{NN}_2$ in Fig.~\ref{fig:methods:sdes:nnpc:addedapp} may deteriorate performance, and thus regularization should be used during training.
We refer to this technique as NNPC+ and provide computational results in Section~\ref{sec:comp:stochastic}.

For a mixed SPDE problem where $\lambda$ is also stochastic and partially unknown, we consider the dataset $\pazocal{D} = \{\{F_i\}_{i=1}^{N}, \{U_i\}_{i=1}^{N}, \{\Lambda_i\}_{i=1}^{N}\}$, with $F_i = \{(x_j^{(i)}, f_j^{(i)})\}_{j=1}^{N_{f}}$, $U_i = \{(x_j^{(i)}, u_j^{(i)})\}_{j=1}^{N_u}$, and $\Lambda_i = \{(x_j^{(i)}, \lambda_j^{(i)})\}_{j=1}^{N_{\lambda}}$.
Within NNPC(+), this additional stochastic term can be modeled similarly to Eq.~\eqref{eq:methods:sdes:fmodes} with a Karhunen-Lo\`{e}ve expansion as
\begin{equation}\label{eq:methods:sdes:lambdamodes}
	\lambda_{\theta_2}(x; \beta) = \lambda_{0}(x; \beta) + \sum_{j = 1}^{S_2}\sqrt{q_j}\lambda_{j}(x; \beta)\theta_{2j},
\end{equation}
where $\theta_2 = \Phi_2^T\sqrt{Q_2}^{-1}\lambda = \{\theta_{2j}\}_{j=1}^{K_2}$; $\Phi_2$ and the diagonal matrix $Q_2$ contain the eigenvectors and the eigenvalues of the covariance matrix of the $N$ realizations of $\lambda$, respectively; and a measurement $\lambda$ is a vector of size $N_{\lambda}$.
Similarly to Eq.~\eqref{eq:methods:sdes:fmodes}, each $\lambda_{j}(\cdot; \beta)$ represents one of the mode functions of $\lambda$ and is one of the outputs of a NN (specific to $\lambda$) parametrized by $\beta$.
Finally, $u_{\theta}(x; \beta)$ is given by Eq.~\eqref{eq:methods:sdes:umodes}, with $\theta = [\theta_1;\theta_2]$, where $\theta_1,\theta_2$ are the low dimensional representations of $f$ and $\lambda$, respectively.
See Section~\ref{sec:comp:stochastic} for an example.

%% file: appendix/IN_app_eval.tex
\section{Supplementary material to Section~\ref{sec:eval}: Additional metrics and calibration details}\label{app:eval}

\begingroup
\etocsetstyle {subsection}
{\leftskip 0pt}
{\leftskip 0pt}
{\bfseries\footnotesize\makebox[1cm][l]{\etocnumber}%
\etocname\nobreak\hfill\nobreak\etocpage
\par
}
{}
\etocsetstyle {subsubsection}
{}
{\leftskip 30pt}
{\mdseries\footnotesize\makebox[1cm][l]{\etocnumber}%
\etocname\nobreak\hfill\nobreak\etocpage
\par
}
{}
\localtableofcontentswithrelativedepth{+3}
\endgroup

\subsection{Additional evaluation metrics}\label{app:eval:metrics}

\subsubsection{Secondary metrics evaluating statistical consistency}\label{app:eval:metrics:second}

Because Eq.~\eqref{eq:eval:eval:approx:rmsce} averages predictions from different locations $x$, it is possible for a predictive model to be probabilistically calibrated but not ideal; e.g., a model with constant uncertainty for every $x$, although useless, may satisfy Eq.~\eqref{eq:eval:eval:approx:rmsce} if the predicted uncertainty is accurate for the entire data distribution.   
In this context, \citet{gneiting2007probabilistic} have shown that a metric measuring sharpness, which is independent of any data, can be used.
Specifically, between two models with the same RMSCE, we prefer the sharper one, where sharpness can be defined as $\text{PIW} = \mathbb{E}_{x \sim \cD_{test}}\left[\hat{u}(x)_p^h - \hat{u}(x)_p^l\right]$ for some value of $p$, e.g., $95 \%$.
Finally, a model with homogeneous across $x$ uncertainty predictions may still be performing well in terms of RMSCE and PIW, but be far from ideal. 
In this regard, references \cite{gneiting2007probabilistic,levi2019evaluating} have proposed to use a metric measuring dispersion in uncertainty predictions.
The rationale for this metric is that a more disperse model can be more robust in terms of predictions away from training data.
We denote the metric as $\text{SDCV}$ and is given as the coefficient of variation of the predicted standard deviation for different $x$ values, i.e., $\text{SDCV} = \frac{\sigma_{\sigma}}{\mu_{\sigma}}$, where $\mu_{\sigma}$, $\sigma_{\sigma}$ are given and approximated as 
\begin{equation} \label{}
	\begin{split}
		\mu_{\sigma} = \mathbb{E}_{x \sim \cD_{test}} SD(u|x, \cD) & \approx \frac{1}{N_{test}}\sum_{i=1}^{N_{test}}\bar{\sigma}(x_i), \\
		\sigma_{\sigma} = \sqrt{\mathbb{E}_{x \sim \cD_{test}}\left[ SD(u|x, \cD) - \mu_{\sigma}\right]^2} & \approx \sqrt{\frac{1}{N_{test}}\sum_{i=1}^{N_{test}} (\bar{\sigma}(x_i) - \mu_{\sigma})^2},
	\end{split}
\end{equation}
where $SD(u|x, \cD)= \sqrt{Var(u|x, \cD)}$.	
Both PIW and SDCV cannot be used as metrics by themselves.
Instead, for a given calibration value RMSCE, we prefer models with smaller PIW and larger SDCV values.	

\subsubsection{Comparison to gold standard}\label{app:eval:metrics:gold}

Further, when novel techniques are developed, they are expected to be more efficient or more accurate than alternatives. 
In those cases, usually a representative problem is solved with both the novel technique and a gold standard technique for comparison purposes. 
A gold standard can be an accurate but computationally expensive technique (e.g., HMC of Section~\ref{app:methods:bnns:mcmc:stocdyn}) or the exact solution if the problem is amenable to one.
A comparison metric to be used in such cases is the normalized inner product between the predicted variances (see also \cite{rudy2021outputweighted}), i.e., 
\begin{equation}\label{eq:eval:eval:nip}
	\text{NIP-G} = \frac{\langle \sigma_G^2, Var(U|X, \cD) \rangle}{\sqrt{\langle \sigma_G^2, \sigma_G^2 \rangle \langle Var(U|X, \cD), Var(U|X, \cD) \rangle}},
\end{equation} 
where $\sigma_G^2(X)$, $Var(U|X, \cD) \approx \bar{\sigma}^2(X)$ are the predicted variances of a set of values of $u$, denoted by $U = [u_1,\dots,u_{N_{test}}]^T$, at locations $X = [x_1,\dots,x_{N_{test}}]^T$, obtained by the gold standard and the technique to be tested, respectively.
Of course, NIP-G of Eq.~\eqref{eq:eval:eval:nip} is connected through a linear relationship with the distance $\langle{ \sigma_G^2 - Var(U|X, \cD), \sigma_G^2 - Var(U|X, \cD) \rangle}$, but it is advantageous in terms of interpretability because it takes values only between $0$ and $1$.
Nevertheless, NIP-G compares only the second-order statistics between the gold standard and the predictive distributions. 
A more comprehensive metric is the KL divergence between the two predictive distributions $p_G(u|x)$ and $\mathbb{E}_{\theta|\cD}[p(u|x, \theta,\cD)]$ at an arbitrary location $x$.
This is written as $KL(p_G(u|x)||\mathbb{E}_{\theta|\cD}[p(u|x, \theta,\cD)])$ and denoted as KL-G in this paper. 
Once again, $\mathbb{E}_{\theta|\cD}[p(u|x, \theta,\cD)]$ is approximated by $\bar{p}(u|x)$ via Eq.~\eqref{eq:uqt:pre:mcestmc:mcest}, and $\bar{p}(u|x)$ can be approximated by $\pazocal{N}(\bar{\mu}(x), \bar{\sigma}^2(x))$ given samples $\{u_{\hat{\theta}_j}(x)\}_{j=1}^M$.

\subsection{Post-training calibration}\label{app:eval:calib}

\subsubsection{CDF modification by fitting an auxiliary model}\label{app:eval:calib:cdf}

Following \cite{kuleshov2018accurate}, assume that our predictive model is miscalibrated and does not satisfy $P(U_X \leq \bar{P}_X^{-1}(p)) = p$, but instead satisfies $Q(p) = P(U_X \leq \bar{P}_X^{-1}(p)) \neq p$ for every $p \in [0, 1]$.
Subsequently, note that $p = Q^{-1}(p')$ is satisfied for some arbitrary $p' \in (0,1)$ and 
\begin{equation}
	p' = Q\circ Q^{-1}(p') = P(U_X \leq \bar{P}_X^{-1}\circ Q^{-1}(p') ).
\end{equation}
Using the property $(g \circ f)^{-1} = f^{-1}\circ g^{-1}$ we obtain
\begin{equation}
	p' = P(U_X \leq (Q\circ \bar{P}_X)^{-1}(p') ),
\end{equation}
which shows that $Q\circ \bar{P}_X$ yields a perfectly calibrated model; i.e., $P(U_X \leq (Q\circ \bar{P}_X)^{-1}(p)) = p$ for any $p \in [0, 1]$. 
Therefore, we can calibrate our model by applying $Q$ to the output CDFs.

Clearly, $Q$ is not known, but it can be approximated. 
Given calibration data $\cD_r = \{x_i, u_i\}_{i=1}^{N_r}$ and the CDFs $\bar{P}(u_i|x_i)$ for every $i$, we can construct an estimate of $Q(p)$ as
\begin{equation}
	Q_{N_r}(p) = \frac{1}{N_r}\sum_{i=1}^{N_r}\mathds{1}(\bar{P}(u_i|x_i) \leq p),
\end{equation}
where the fact that every CDF is an increasing function has been used.
We can, subsequently, train an auxiliary model that approximates $Q$ using the dataset 
\begin{equation}\label{eq:eval:eval:approx:recal:data}
	\cD_{Q} = \{\bar{P}(u_i|x_i), Q_{N_r}(\bar{P}(u_i|x_i))\}_{i = 1}^{N_r}.
\end{equation}
For example, isotonic regression, which is suitable for monotonic function fitting, can be used; an open-source implementation can be found in \cite{chung2021uncertainty}.
Note that if the model is calibrated, the dataset of Eq.~\eqref{eq:eval:eval:approx:recal:data} is simply a straight line (see Fig.~\ref{fig:eval:eval:misscal}).
Finally, clearly this approach modifies both the mean and the variance of the predicted $u$ at each $x$; the new mean and variance for each $x$ can be obtained using
\begin{equation}
	\begin{split}
		\bar{\mu}_{\text{new}}(x) & = \int_{0}^{\infty}\left[1 - 	Q(\bar{P}(u|x))\right]du - \int_{-\infty}^{0}Q(\bar{P}(u|x))du,\\
		\bar{\sigma}^2_{\text{new}}(x) & = \int_{0}^{\infty}\left[1 - 	\left[Q(\bar{P}(\sqrt{z}|x)) - Q(\bar{P}(-\sqrt{z}|x))\right]\right]dz - \bar{\mu}_{\text{new}}^2(x),
	\end{split}
\end{equation}
where the integrals can be evaluated numerically.

\subsubsection{CRUDE: CDF modification by fitting an empirical distribution}\label{app:eval:calib:crude}

\citet{zelikman2020crude} have proposed a calibration approach according to which an empirical distribution is fitted to the scaled residuals that correspond to the calibration dataset. 
This is performed by simply creating a vector of sorted residuals expressed as $E_u = sorted\{\epsilon_u(x_i)\}_{i=1}^{N_r}$. 
The sorted residuals $E_u$ can, subsequently, be combined with $\bar{\mu}(x)$ and $\bar{\sigma}(x)$ for each $x$ for defining a new predictive distribution that follows the empirical one.
Specifically, for any $p \in [0,1]$, the corresponding percentiles can be obtained using 
\begin{equation} \label{eq:eval:eval:approx:percs:crude}
	\begin{split}
		\hat{u}(x)_{p}^h & \approx \bar{\mu}(x) + \bar{\sigma}(x) E_u[int(p_h N_r)], \\
		\hat{u}(x)_{p}^l &  \approx \bar{\mu}(x) + \bar{\sigma}(x) E_u[int(p_l N_r)], 
	\end{split}
\end{equation}
where $int(\cdot)$ returns the closest integer to a scalar. 
This approach also modifies both the mean and the variance of the predicted $u$ at each $x$.
The new values can be obtained as
\begin{equation}
	\begin{split}
		\bar{\mu}_{\text{new}}(x) & = \bar{\mu}(x) + \bar{\sigma}(x)\bar{\mu}_{\epsilon},  \\
		\bar{\sigma}^2_{\text{new}}(x) & = \bar{\sigma}^2_{}(x)\bar{\sigma}^2_{\epsilon},
	\end{split}
\end{equation}
where
\begin{equation}
	\begin{split}
		\bar{\mu}_{\epsilon} & = \frac{1}{N_r}\sum_{i=1}^{N_r}\epsilon_u(x_i), \\
		\bar{\sigma}^2_{\epsilon} & = \frac{1}{N_r}\sum_{i=1}^{N_r}(\epsilon_u(x_i) - \bar{\mu}_{\epsilon})^2.
	\end{split}
\end{equation}

%% file: appendix/IN_app_results.tex
\section{Supplementary material to Section~\ref{sec:comp}: Details and additional computational results}\label{app:comp}

\begingroup
\etocsetstyle {subsection}
{\leftskip 0pt}
{\leftskip 0pt}
{\bfseries\footnotesize\makebox[1cm][l]{\etocnumber}%
\etocname\nobreak\hfill\nobreak\etocpage
\par
}
{}
\etocsetstyle {subsubsection}
{}
{\leftskip 30pt}
{\mdseries\footnotesize\makebox[1cm][l]{\etocnumber}%
\etocname\nobreak\hfill\nobreak\etocpage
\par
}
{}
\localtableofcontentswithrelativedepth{+3}
\endgroup

\input{appendix/IN_app_results_function}

\clearpage

\input{appendix/IN_app_results_mixed_PINN}
\clearpage

\input{appendix/IN_app_results_stochastic}
\clearpage

\input{appendix/IN_app_results_forward_PINN}
\clearpage

\input{appendix/IN_app_results_DON}

\input{appendix/IN_app_hyperparameters}
\input{appendix/IN_app_architecture}

%% file: appendix/IN_app_results_function.tex
\subsection{Details regarding Fig.~\ref{fig:comp:func:homosc:unk:dists} for the function approximation problem of Section~\ref{sec:comp:func}}\label{app:comp:func:explain:dists}

In all parts of Fig.~\ref{fig:comp:func:homosc:unk:dists}, we obtain the histograms as described below and use kernel density estimation for visualization.
In Fig.~\ref{fig:comp:func:homosc:unk:dists}a, we plot the ``initial prior'' of $\theta$, i.e., the histogram of $\theta$ values drawn from the Gaussian prior $p(\theta|\sigma_{\theta}^2)$, where $\sigma_{\theta}^2$ is the prior variance drawn from the Gamma hyperprior $p(\sigma_{\theta}^2)$ (see Section~\ref{app:methods:bnns:mcmc:hypers}); the ``learned prior'' of $\theta$, i.e., the histogram of $\theta$ values drawn from $p(\theta|\sigma_{\theta}^2)$, where $\sigma_{\theta}^2$ is learned and drawn from $p(\sigma_{\theta}^2|\cD)$ (see Eq.~\eqref{eq:methods:bnns:mcmc:learned:prior}); the posterior distributions of two randomly picked NN parameters, i.e., the histograms of samples of these two parameters obtained during posterior inference; and the ``stacked'' posterior of $\theta$, i.e., the histogram of all posterior samples $\theta$ stacked together.
In Fig.~\ref{fig:comp:func:homosc:unk:dists}b, we plot the initial prior, the learned prior, and the posterior of $u_{\theta}(x=0)$, i.e, the histograms of $u_{\theta}(x=0)$ given $\theta$ samples according to Fig.~\ref{fig:comp:func:homosc:unk:dists}a.
Finally, in Fig.~\ref{fig:comp:func:homosc:unk:dists}c, we plot the prior and the posterior of the predicted noise $\sigma_u$, i.e., the histograms of the $\sigma_u$ values drawn from the Gamma distributions $p(\sigma_u^2)$ and $p(\sigma_u^2|\cD)$; see Section~\ref{app:methods:bnns:mcmc:hypers}.

\subsection{Additional results for the function approximation problem of Section~\ref{sec:comp:func}}\label{app:comp:func:results}

In this section, we provide additional results pertaining to the known homoscedastic noise (Section~\ref{app:comp:func:results:homosc:kno}) and to the unknown Student-t heteroscedastic noise (Section~\ref{app:comp:func:results:student}) cases.

\subsubsection{Known homoscedastic noise and post-training calibration}\label{app:comp:func:results:homosc:kno}

\begin{figure}[H]
	\centering
	\subcaptionbox{}{}{\includegraphics[width=0.32\textwidth]{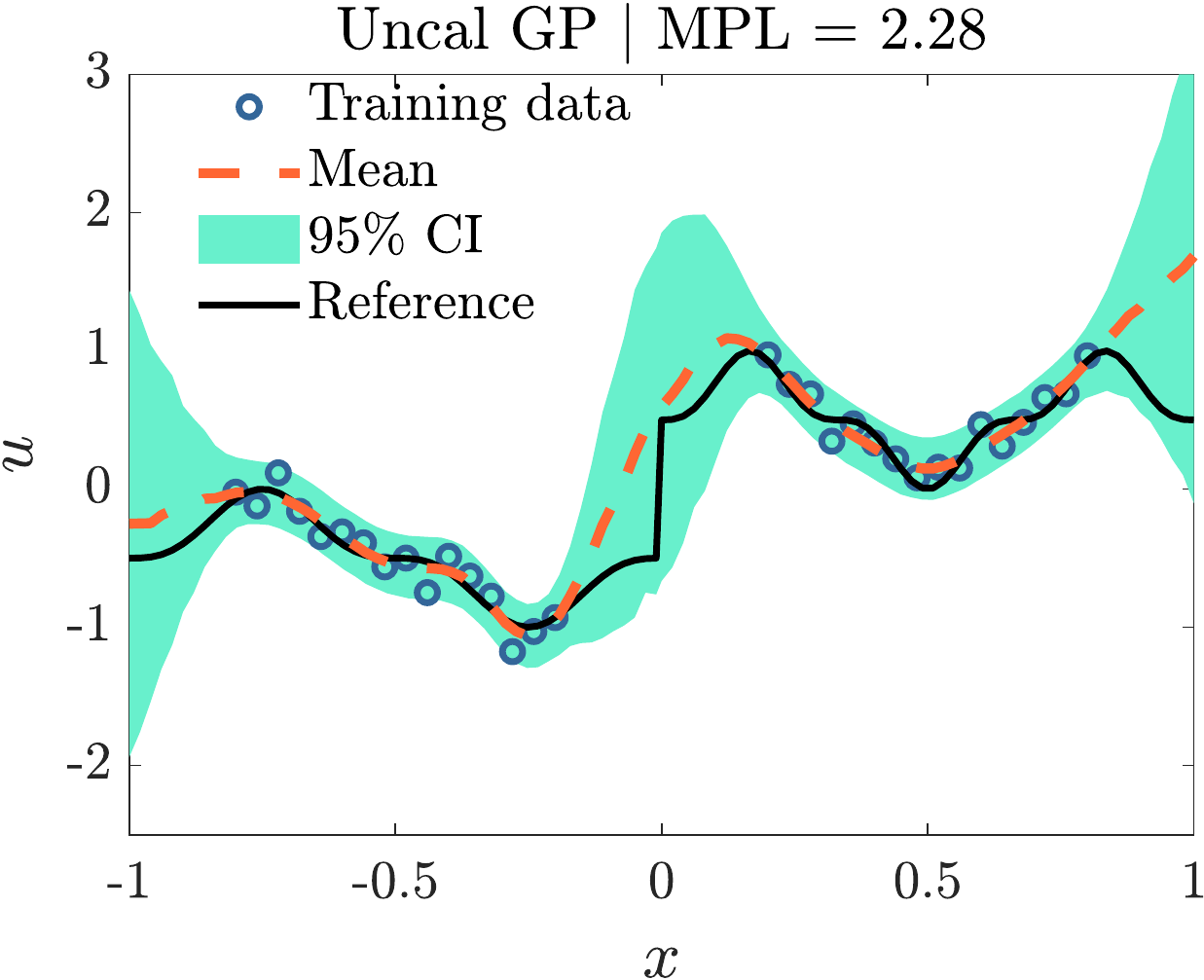}}
	\subcaptionbox{}{}{\includegraphics[width=0.32\textwidth]{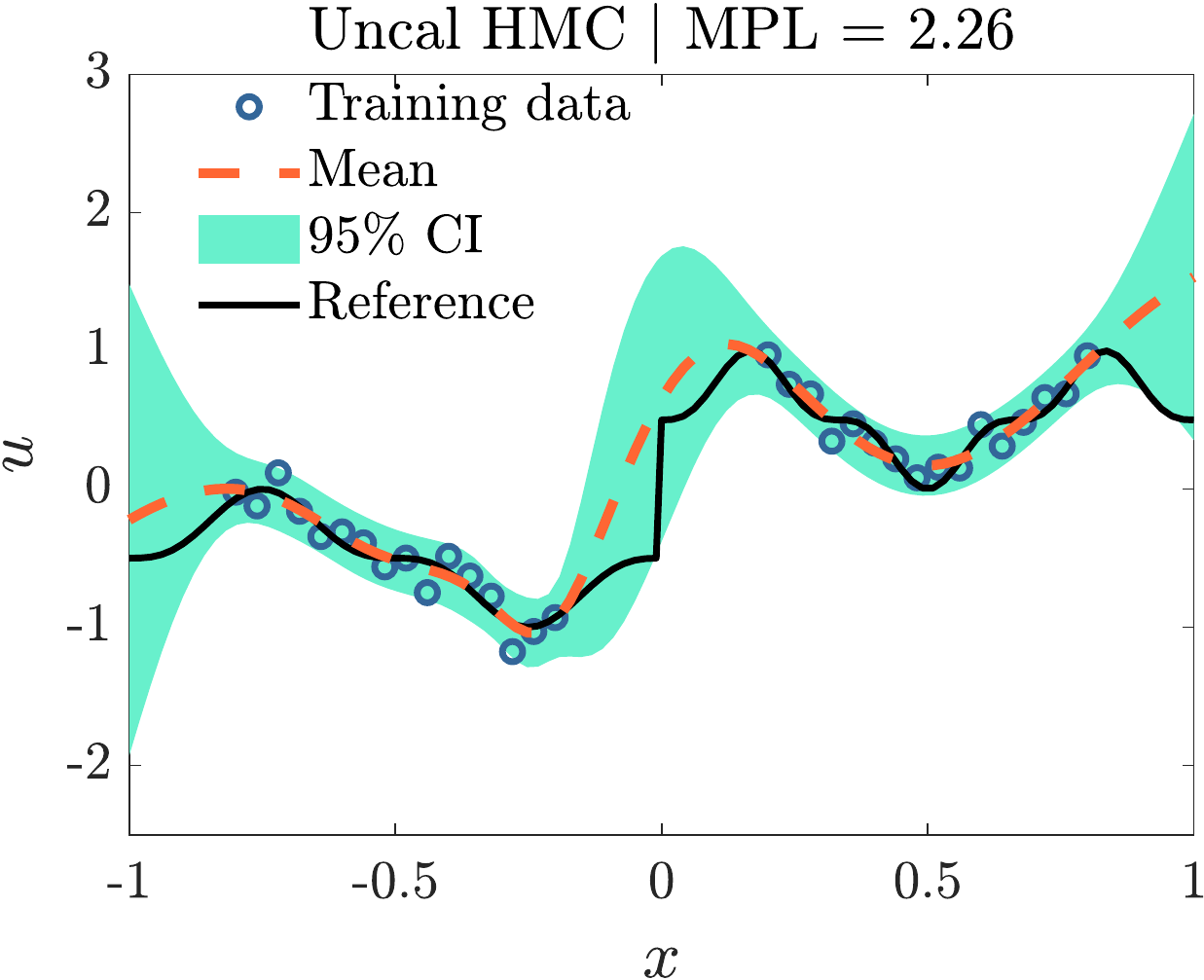}}
	\subcaptionbox{}{}{\includegraphics[width=0.32\textwidth]{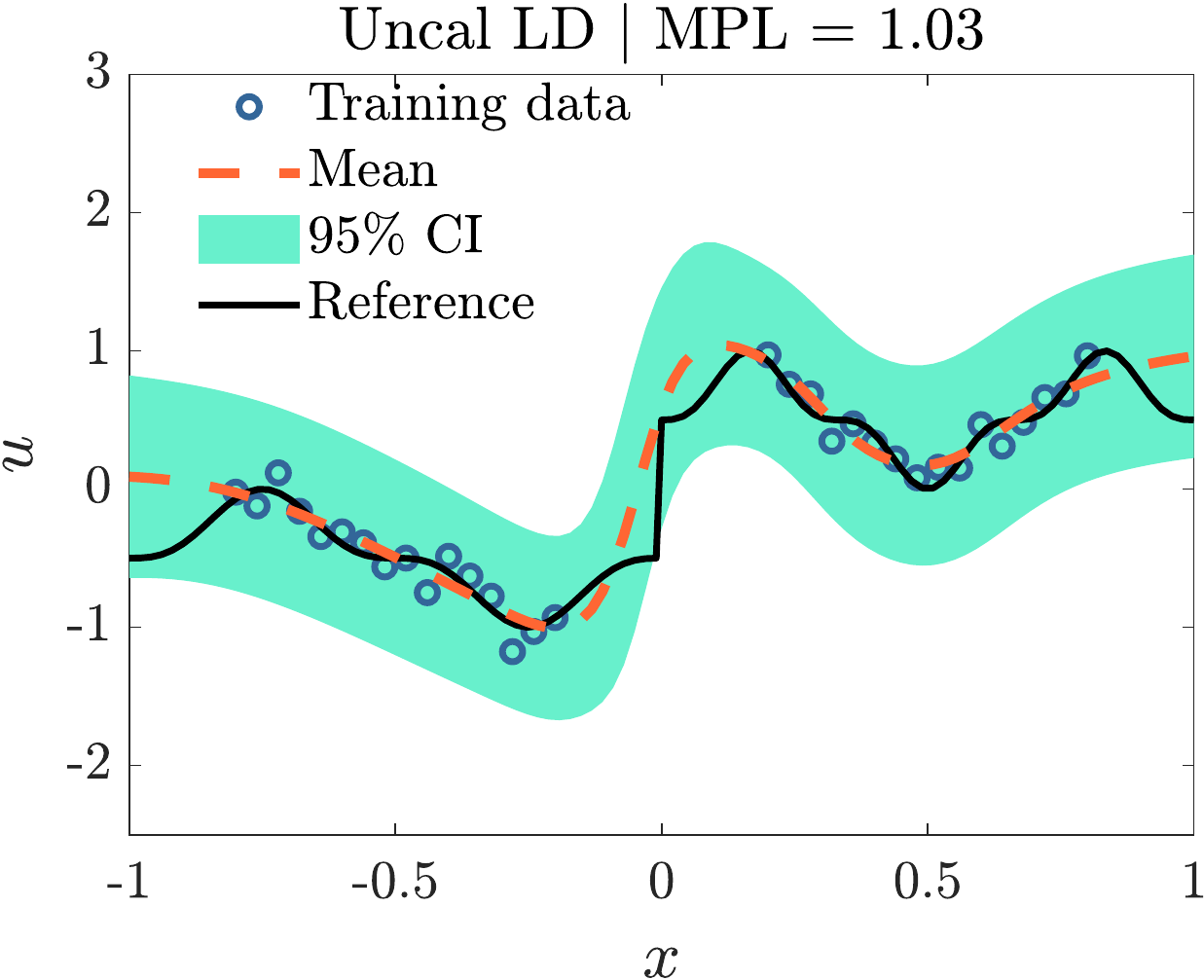}}
	\subcaptionbox{}{}{\includegraphics[width=0.32\textwidth]{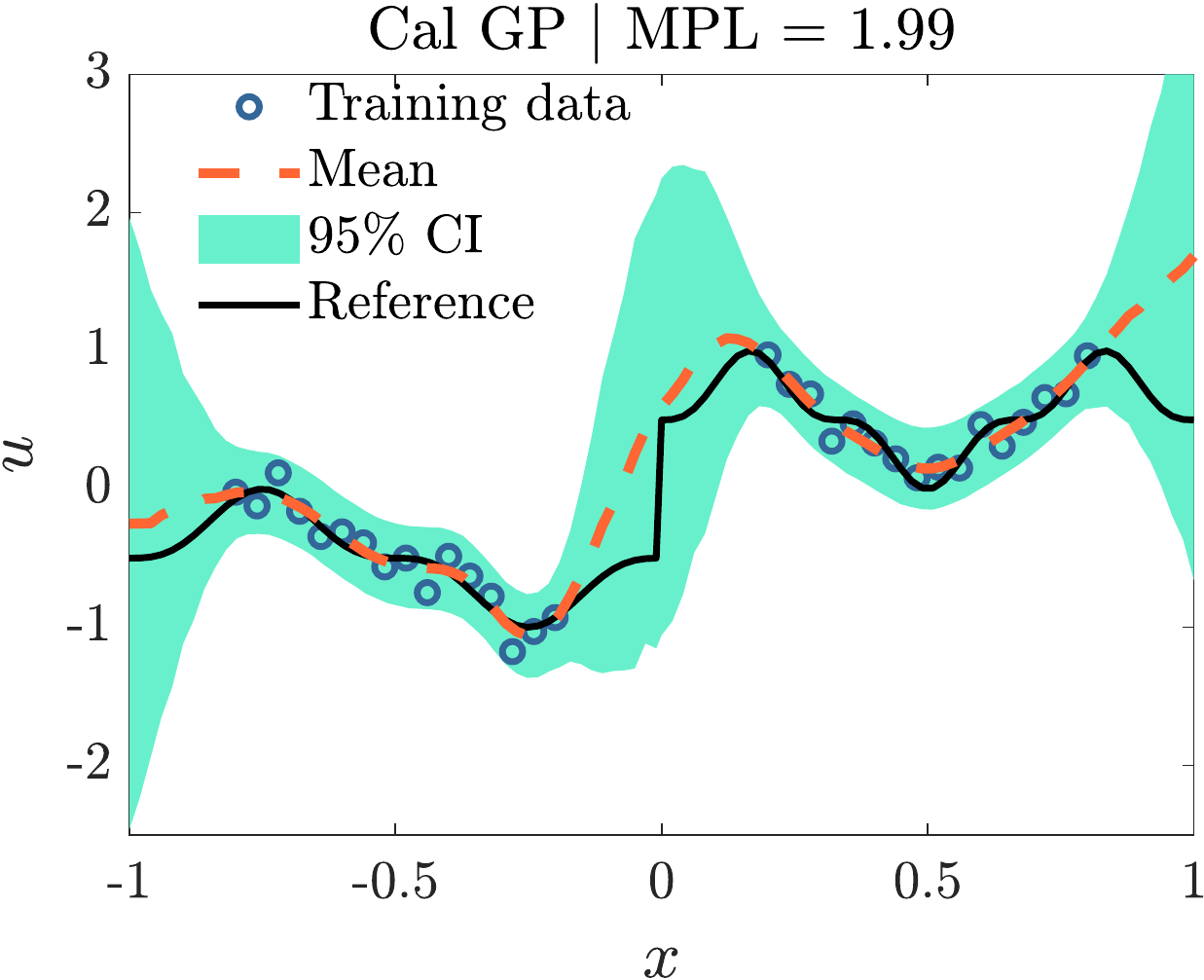}}
	\subcaptionbox{}{}{\includegraphics[width=0.32\textwidth]{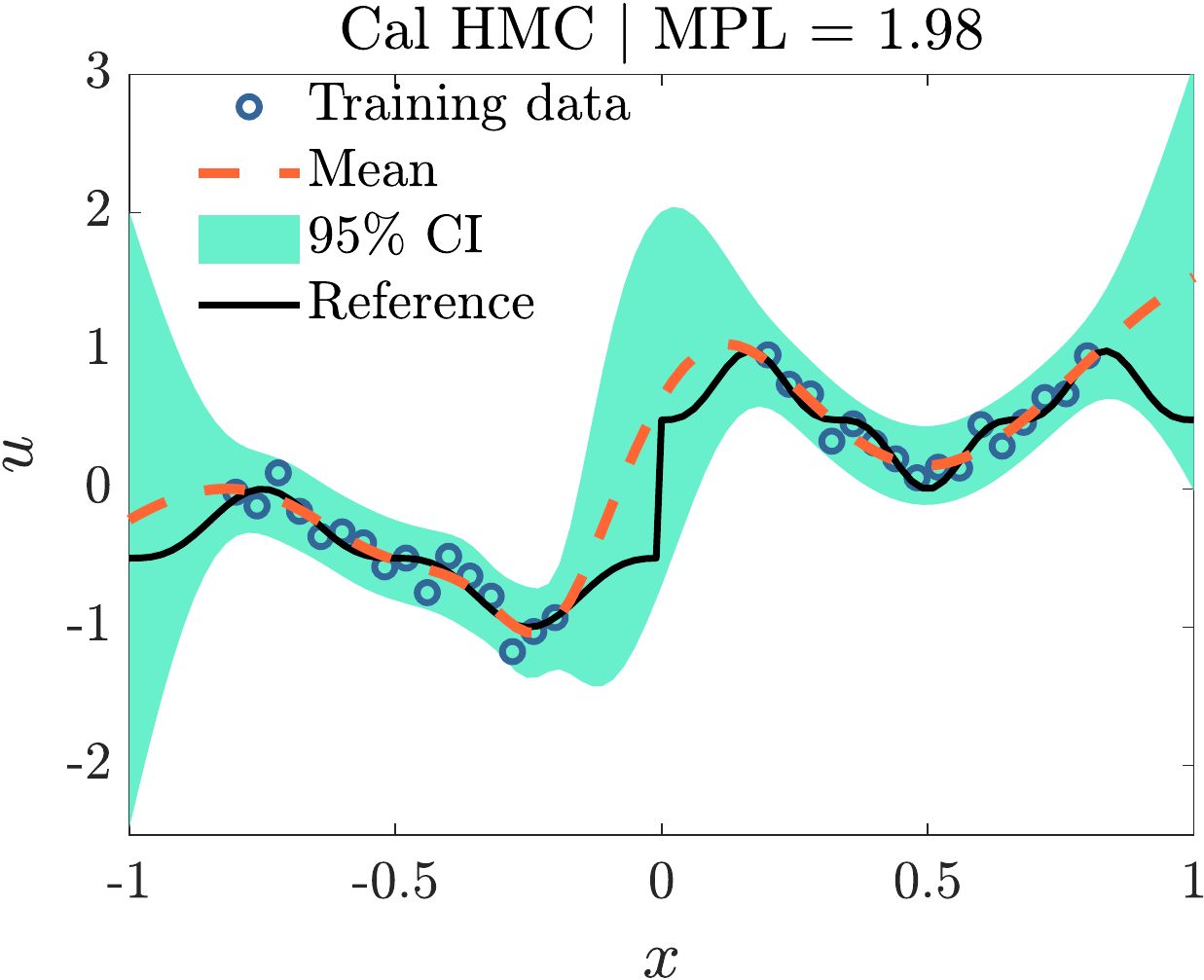}}
	\subcaptionbox{}{}{\includegraphics[width=0.32\textwidth]{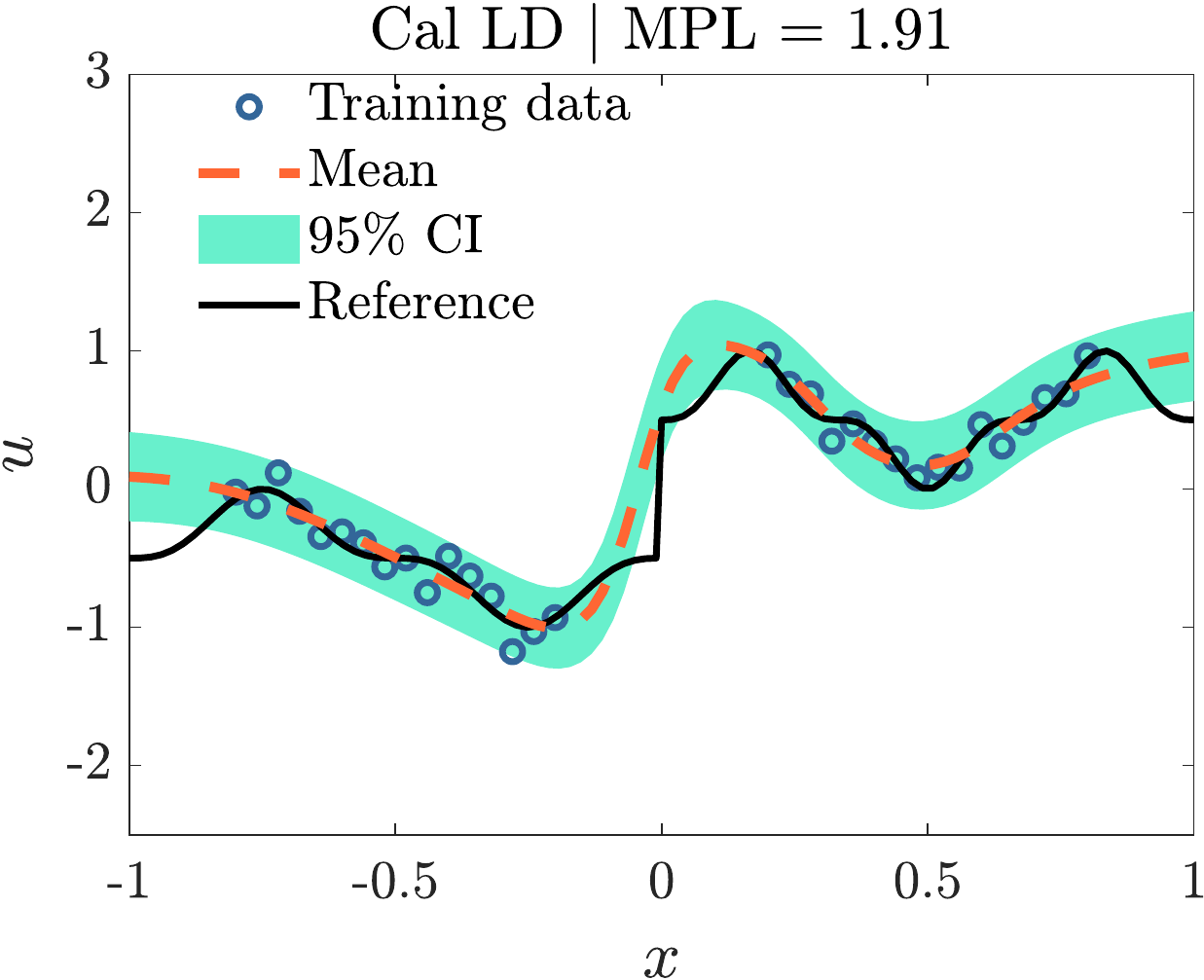}}
	\subcaptionbox{}{}{\includegraphics[width=0.32\textwidth]{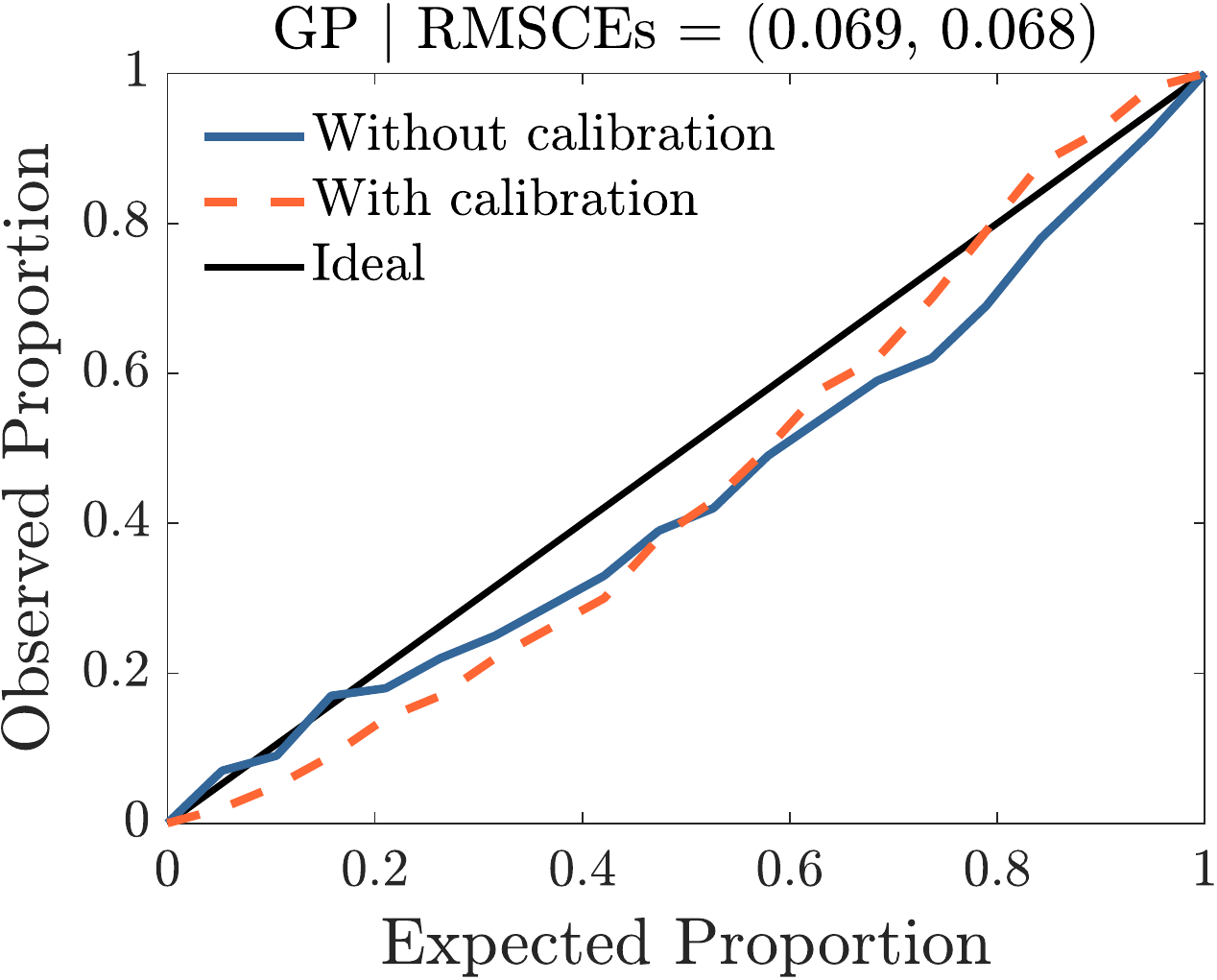}}
	\subcaptionbox{}{}{\includegraphics[width=0.32\textwidth]{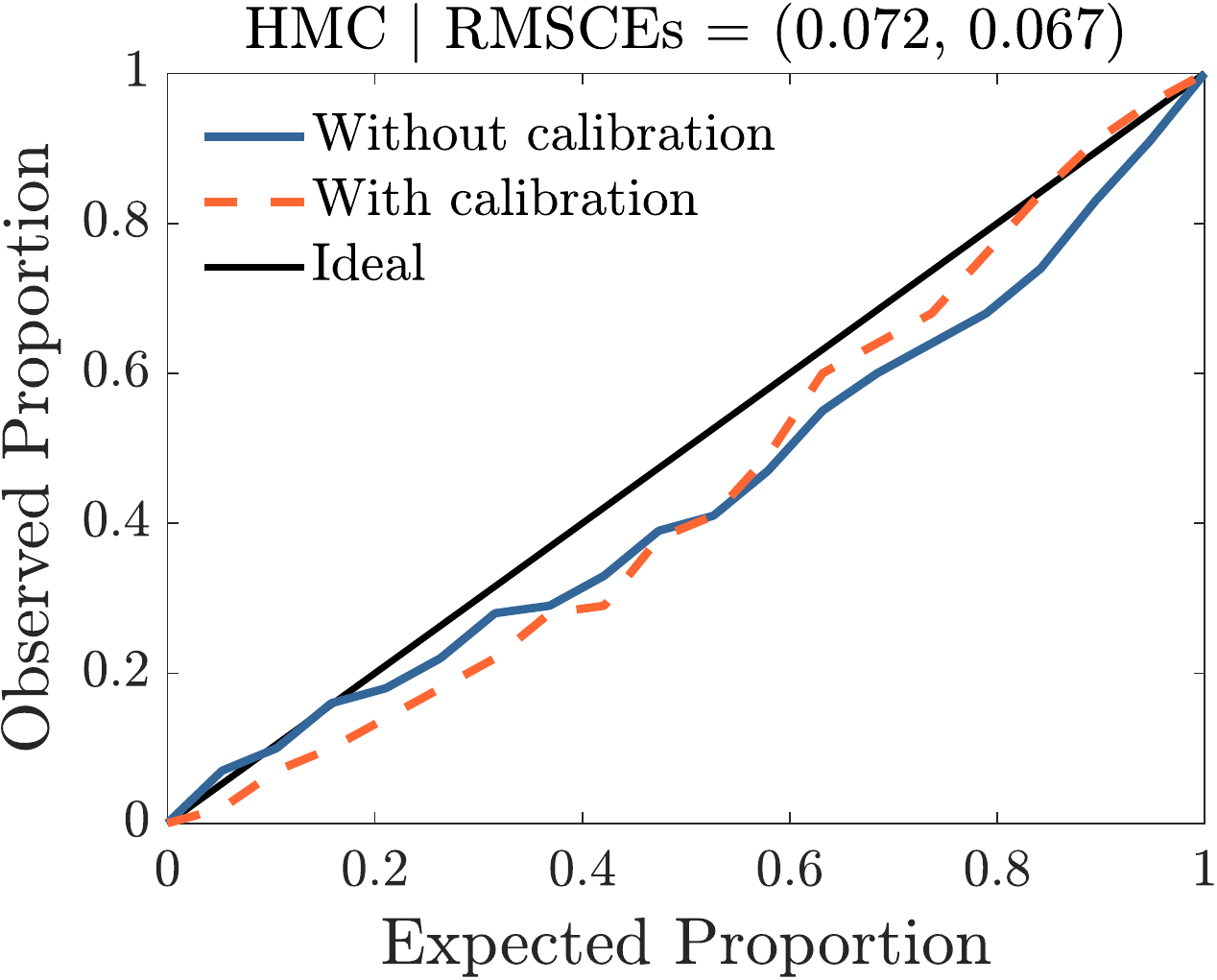}}
	\subcaptionbox{}{}{\includegraphics[width=0.32\textwidth]{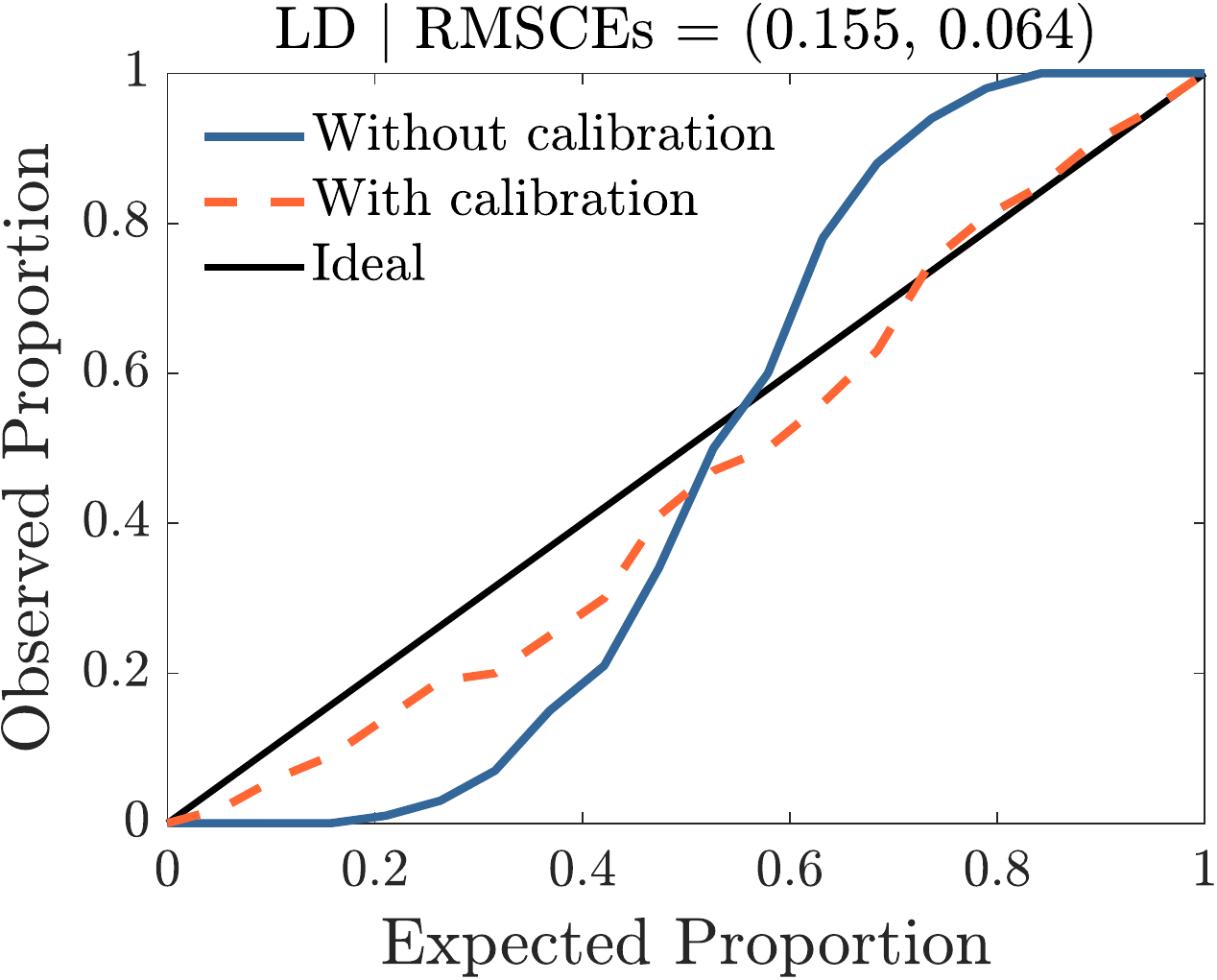}}
	\caption{
		Function approximation problem of Eq.~\eqref{eq:comp:func:func} | \textit{Known homoscedastic noise}: training data and exact function, as well as the mean and total uncertainty ($95\%$ CI) predictions of GP, HMC, and LD.
		\textbf{Top row:} uncalibrated predictions. 
		\textbf{Middle row:} calibrated predictions (Section~\ref{sec:eval:calib}). 
		\textbf{Bottom row:} calibration plots (see also Fig.~\ref{fig:eval:eval:misscal}) and RMSCEs before and after post-training calibration (in the parentheses). 
	}
	\label{fig:comp:func:homosc:kno:res:1}
\end{figure}

\begin{figure}[H]
	\centering
	\subcaptionbox{}{}{\includegraphics[width=0.32\textwidth]{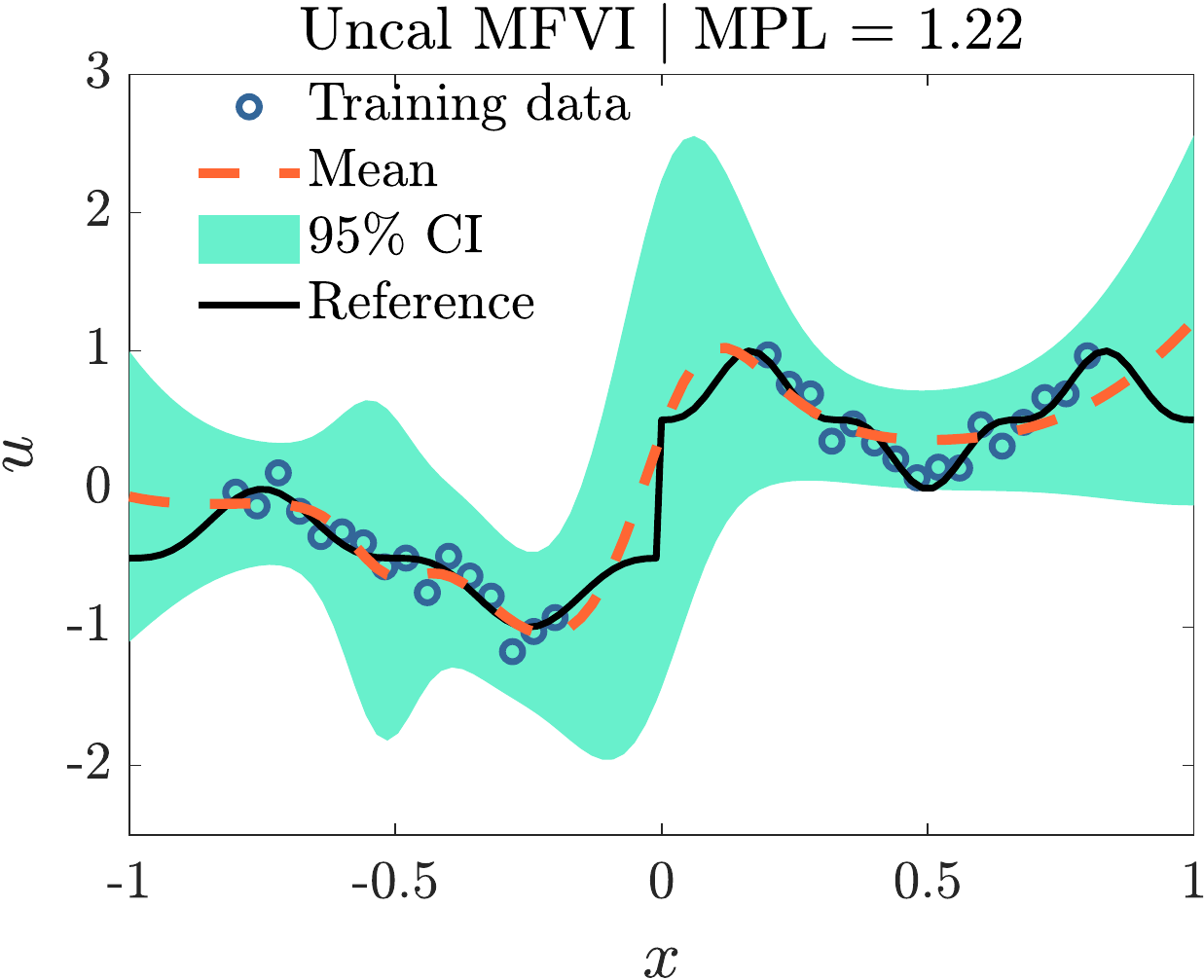}}
	\subcaptionbox{}{}{\includegraphics[width=0.32\textwidth]{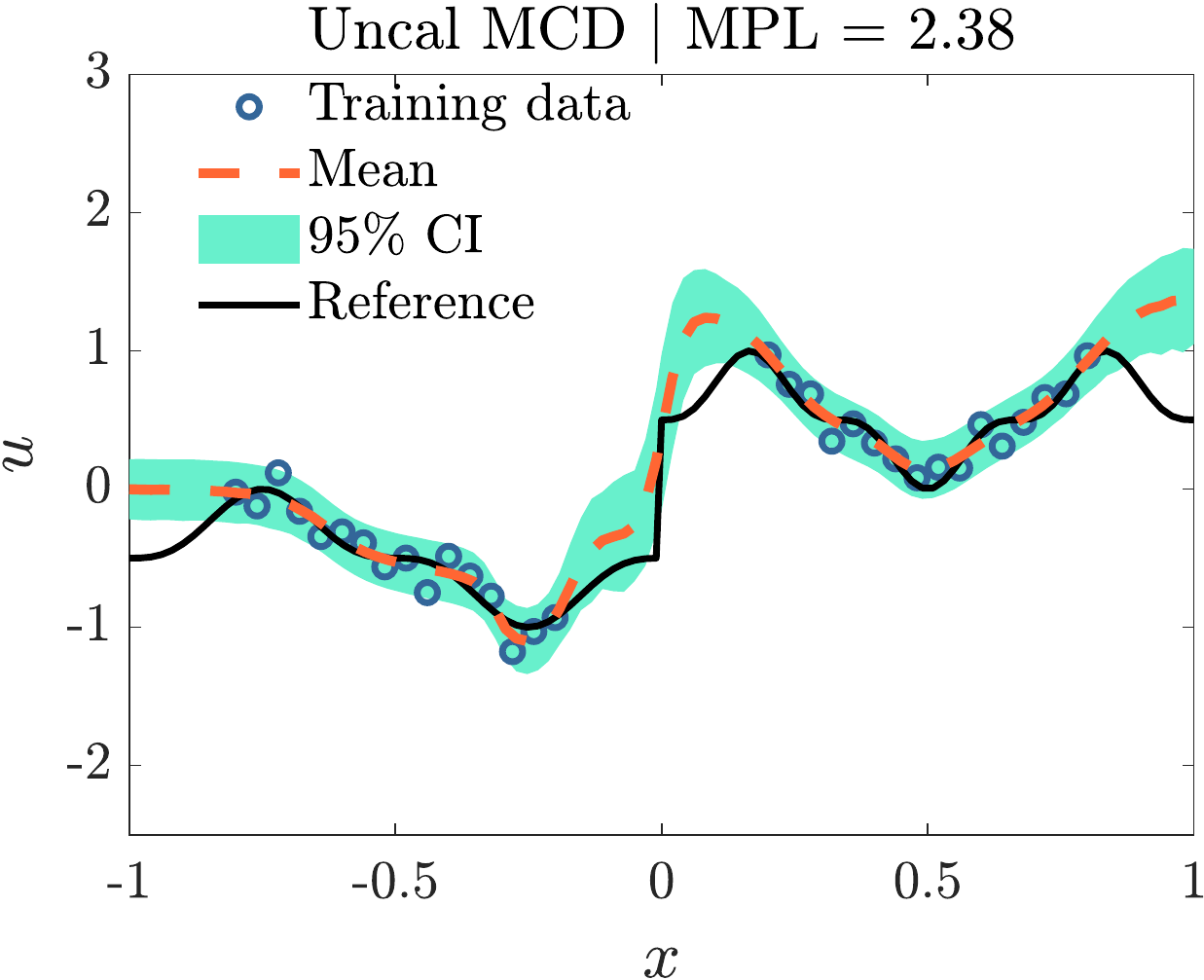}}
	\subcaptionbox{}{}{\includegraphics[width=0.32\textwidth]{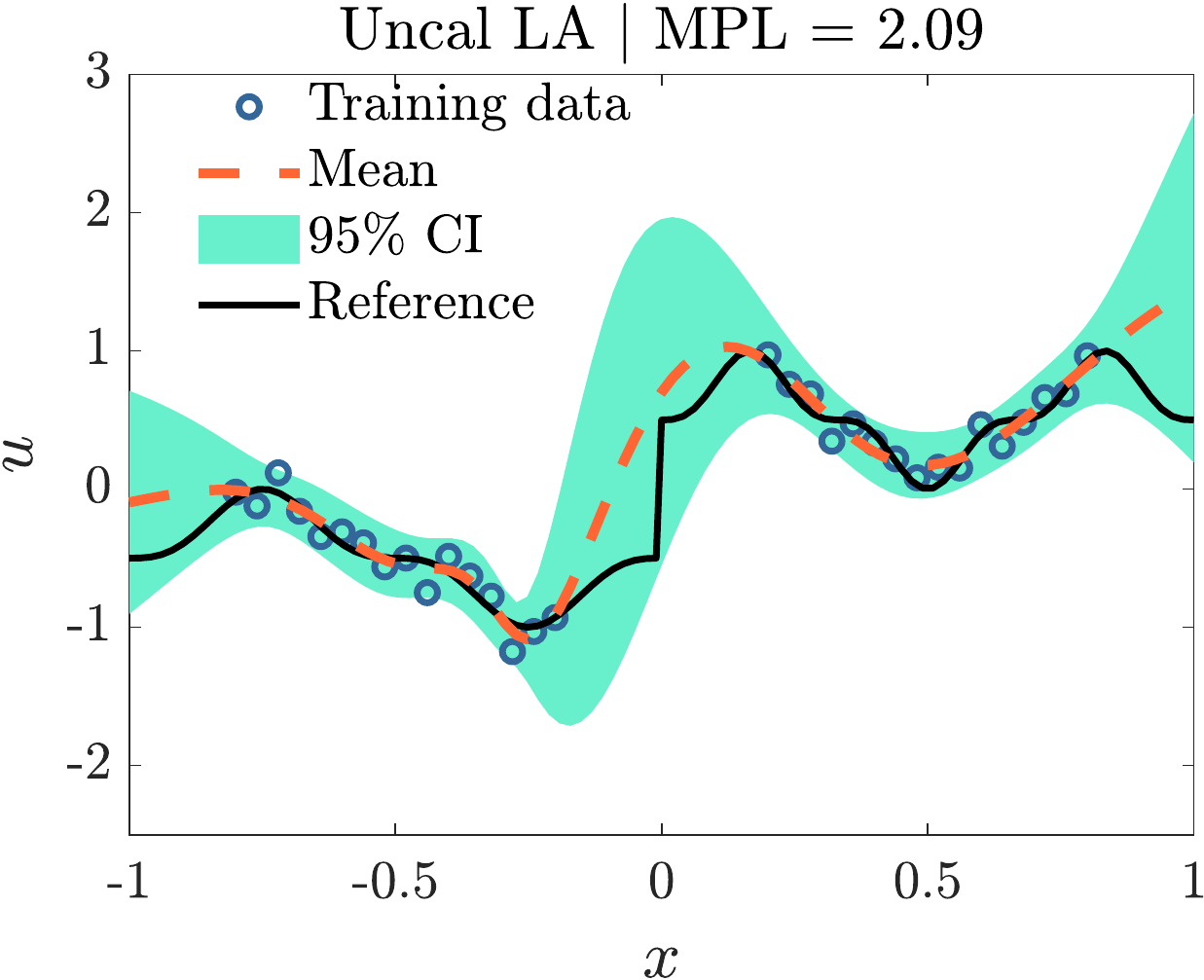}}
	\subcaptionbox{}{}{\includegraphics[width=0.32\textwidth]{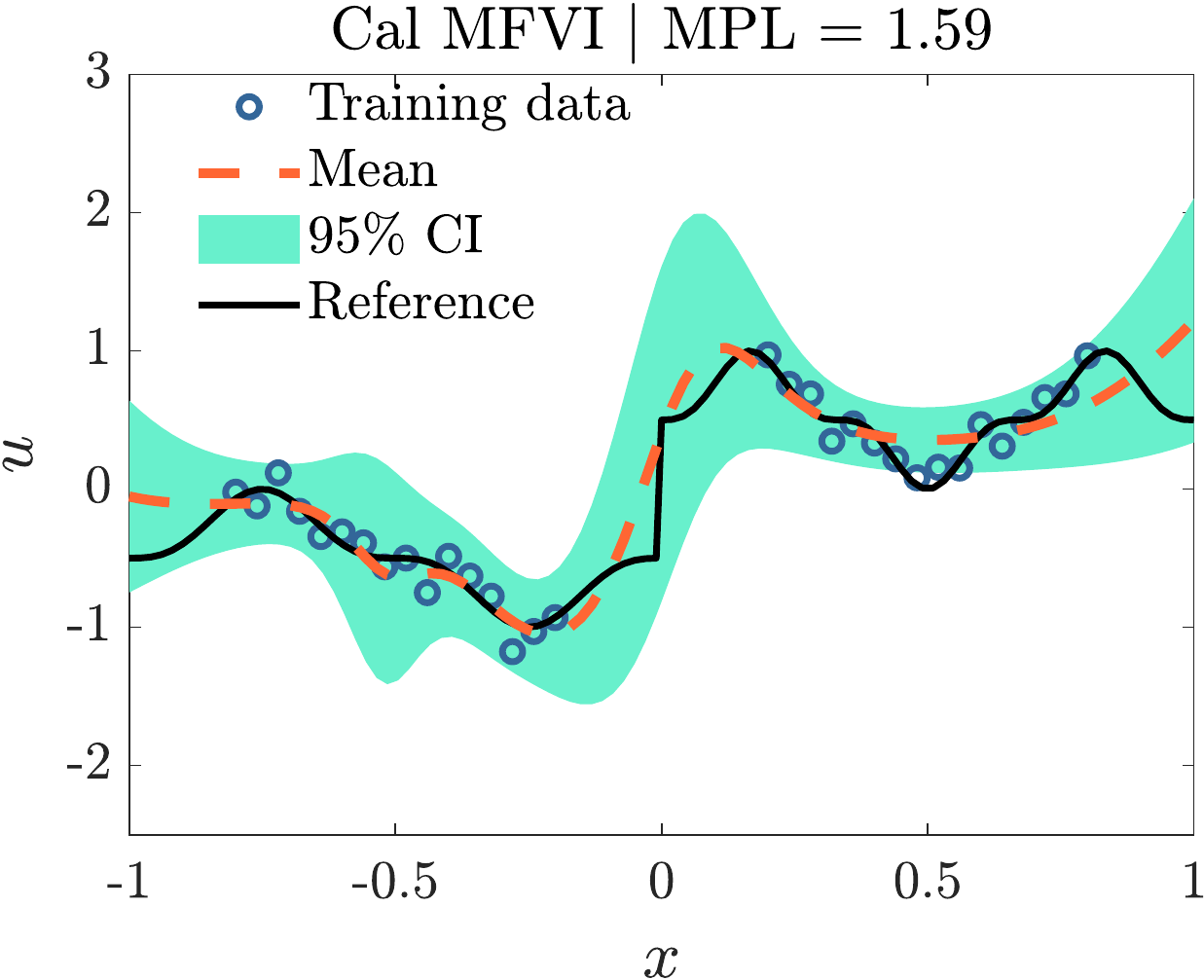}}
	\subcaptionbox{}{}{\includegraphics[width=0.32\textwidth]{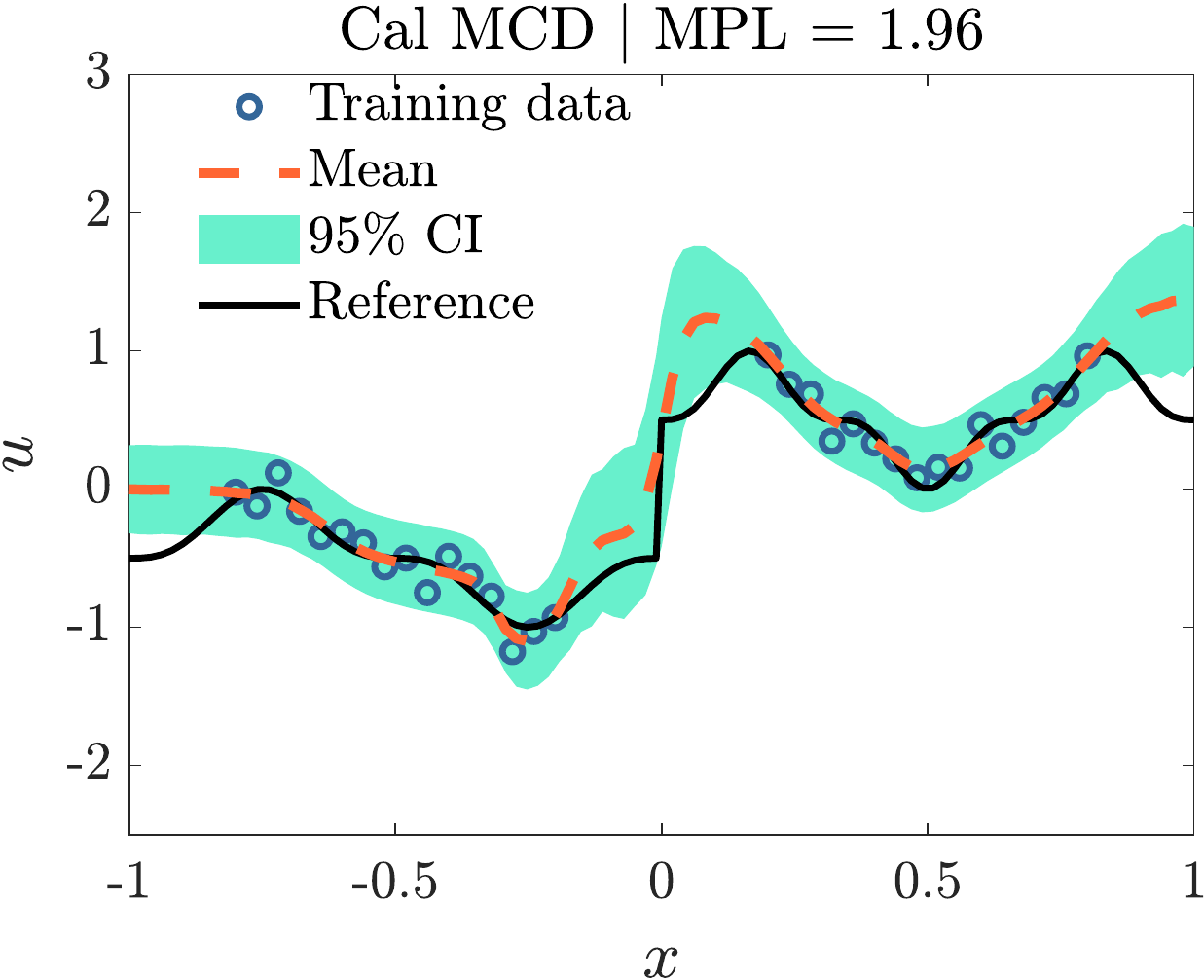}}
	\subcaptionbox{}{}{\includegraphics[width=0.32\textwidth]{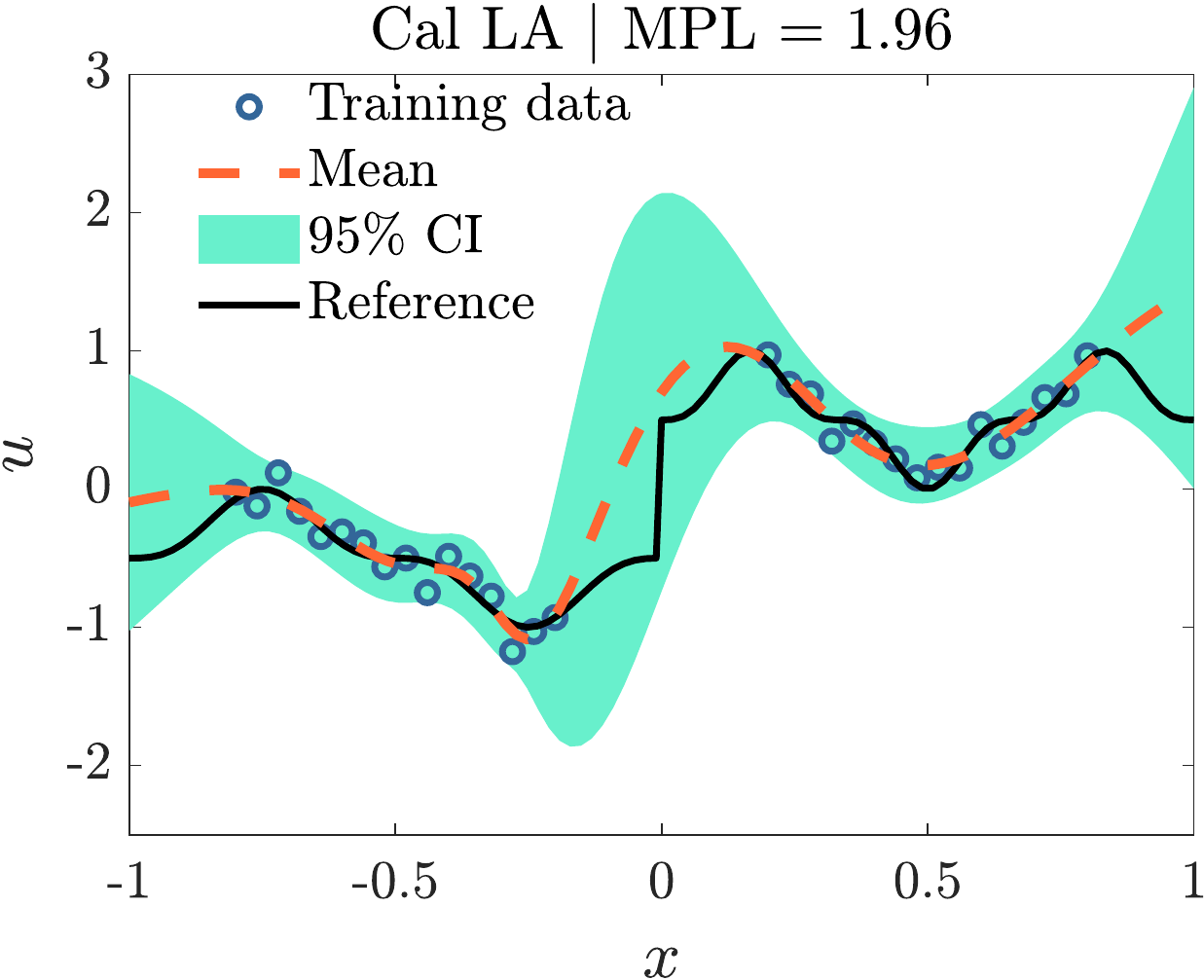}}
	\subcaptionbox{}{}{\includegraphics[width=0.32\textwidth]{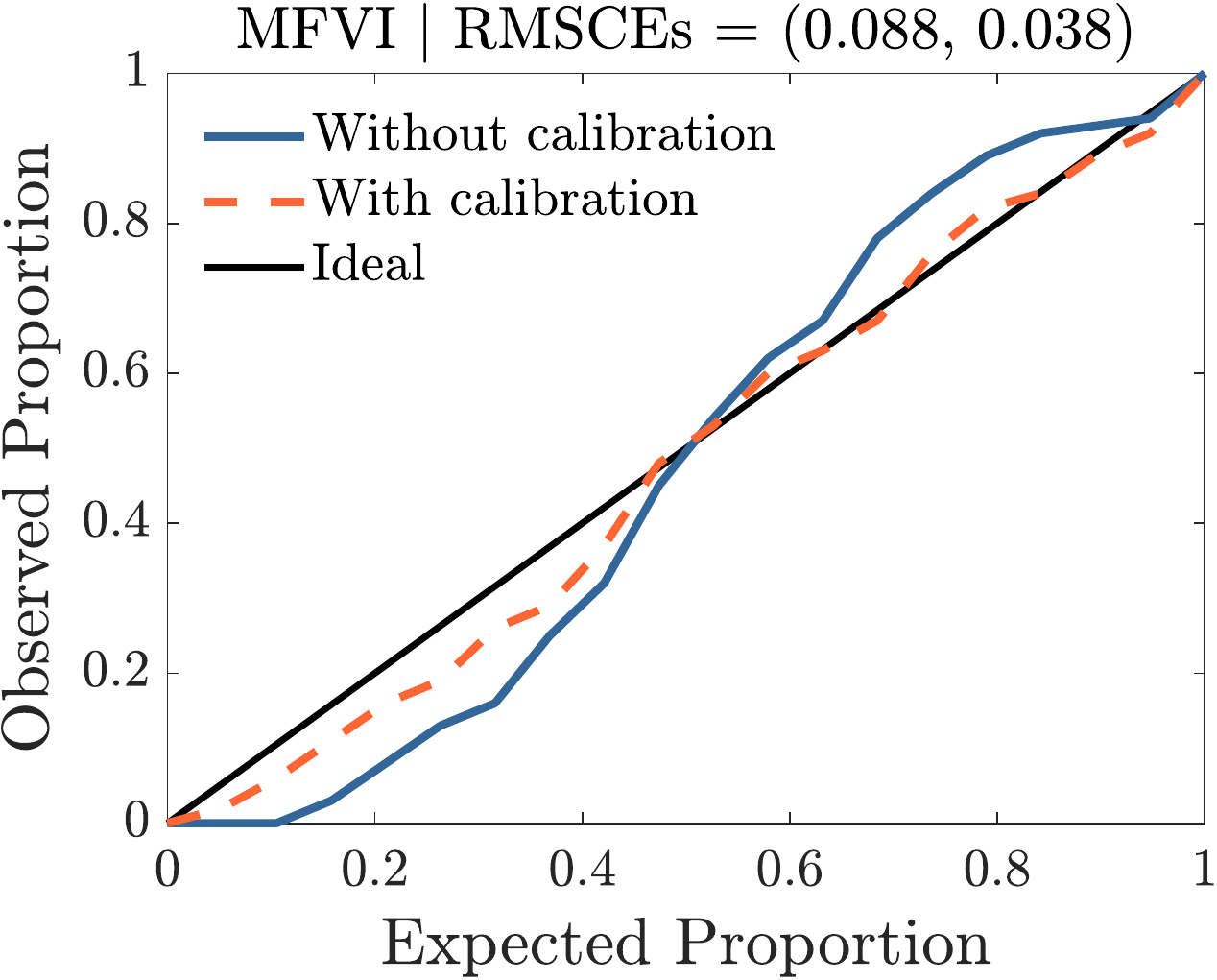}}
	\subcaptionbox{}{}{\includegraphics[width=0.32\textwidth]{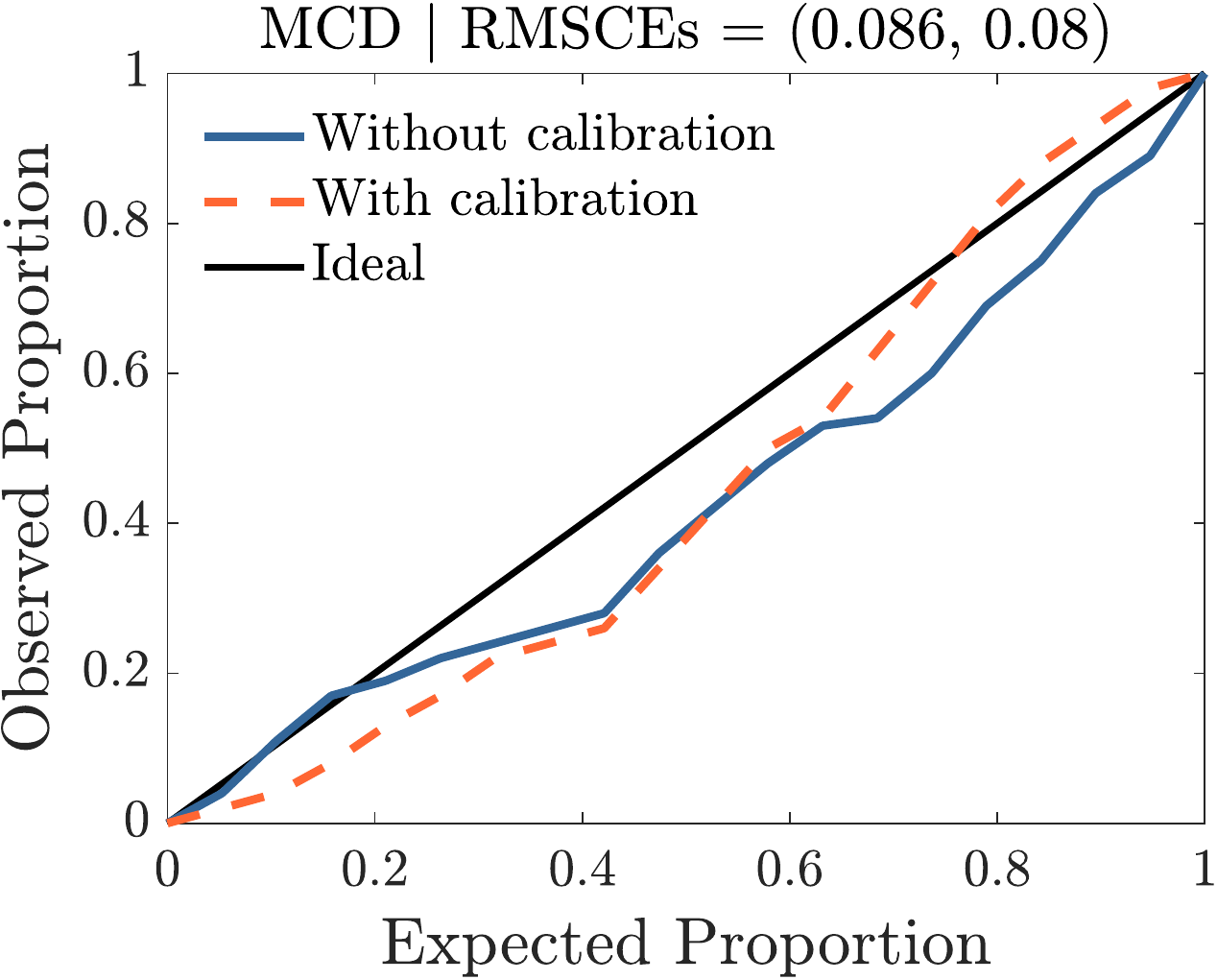}}
	\subcaptionbox{}{}{\includegraphics[width=0.32\textwidth]{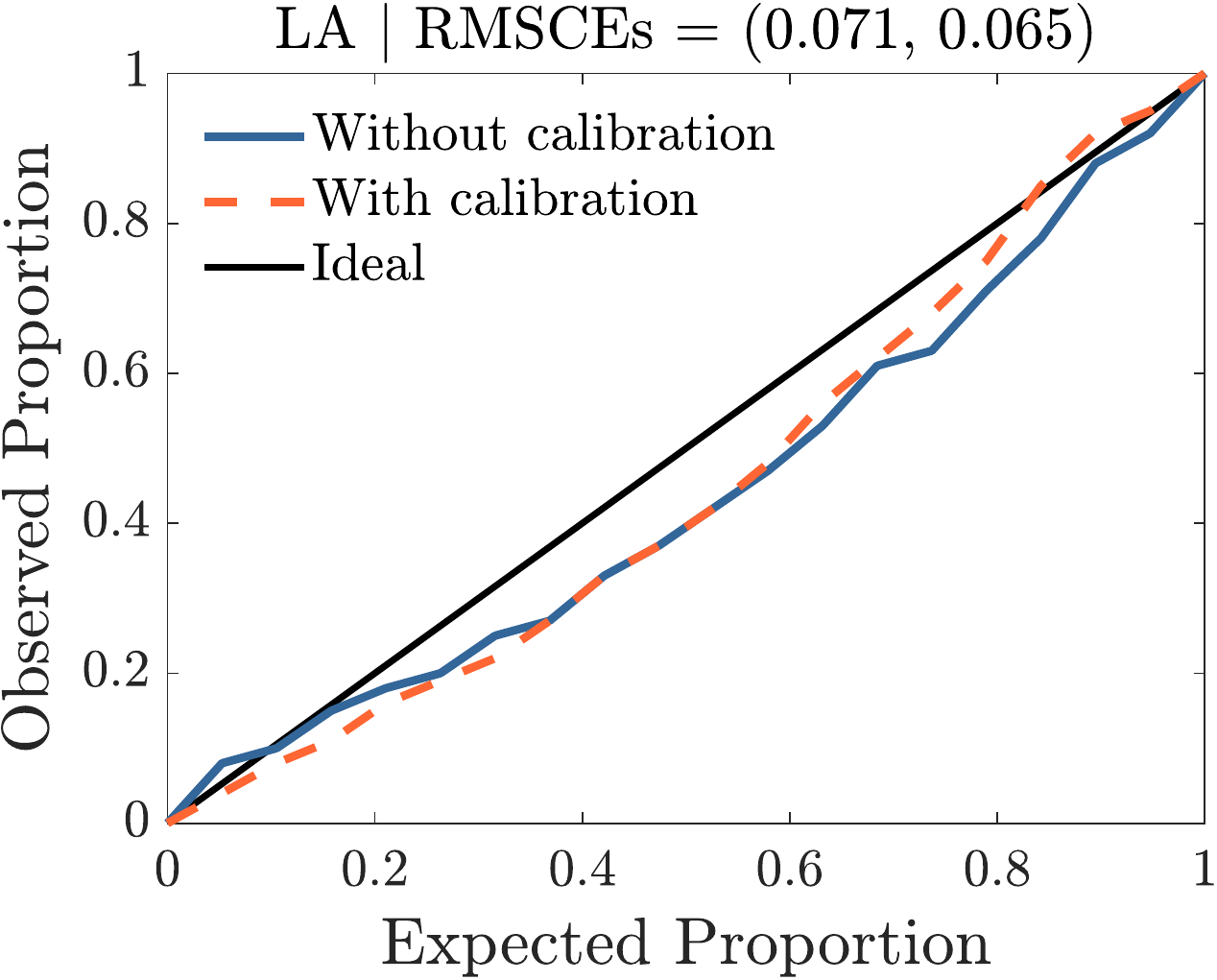}}
	\caption{
		Function approximation problem of Eq.~\eqref{eq:comp:func:func} | \textit{Known homoscedastic noise}: training data and exact function, as well as the mean and total uncertainty ($95\%$ CI) predictions of MFVI, MCD, and LA.
		\textbf{Top row:} uncalibrated predictions. 
		\textbf{Middle row:} calibrated predictions (Section~\ref{sec:eval:calib}). 
		\textbf{Bottom row:} calibration plots (see also Fig.~\ref{fig:eval:eval:misscal}) and RMSCEs before and after post-training calibration (in the parentheses). 
	}
	\label{fig:comp:func:homosc:kno:res:2}
\end{figure}

\begin{figure}[H]
	\centering
	\subcaptionbox{}{}{\includegraphics[width=0.32\textwidth]{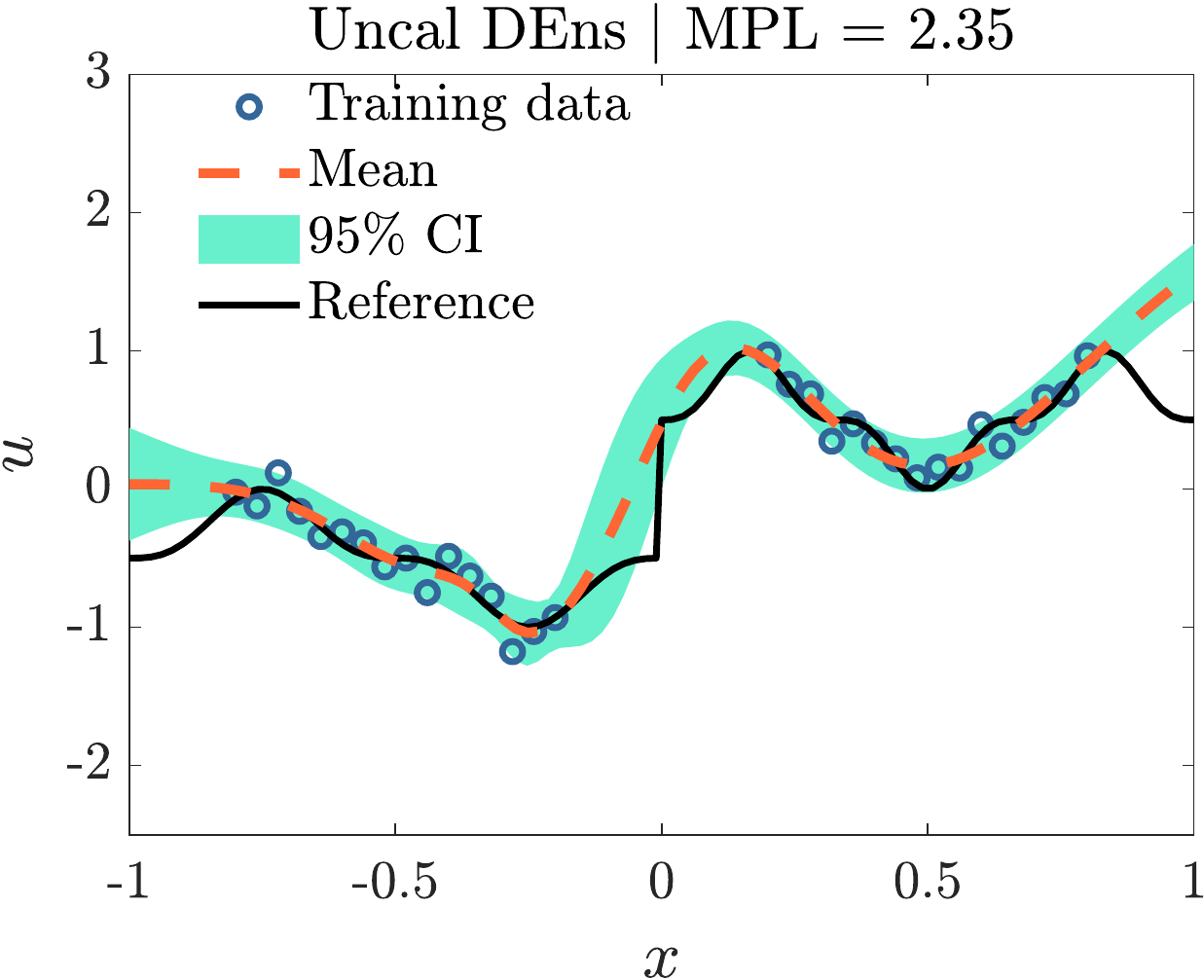}}
	\subcaptionbox{}{}{\includegraphics[width=0.32\textwidth]{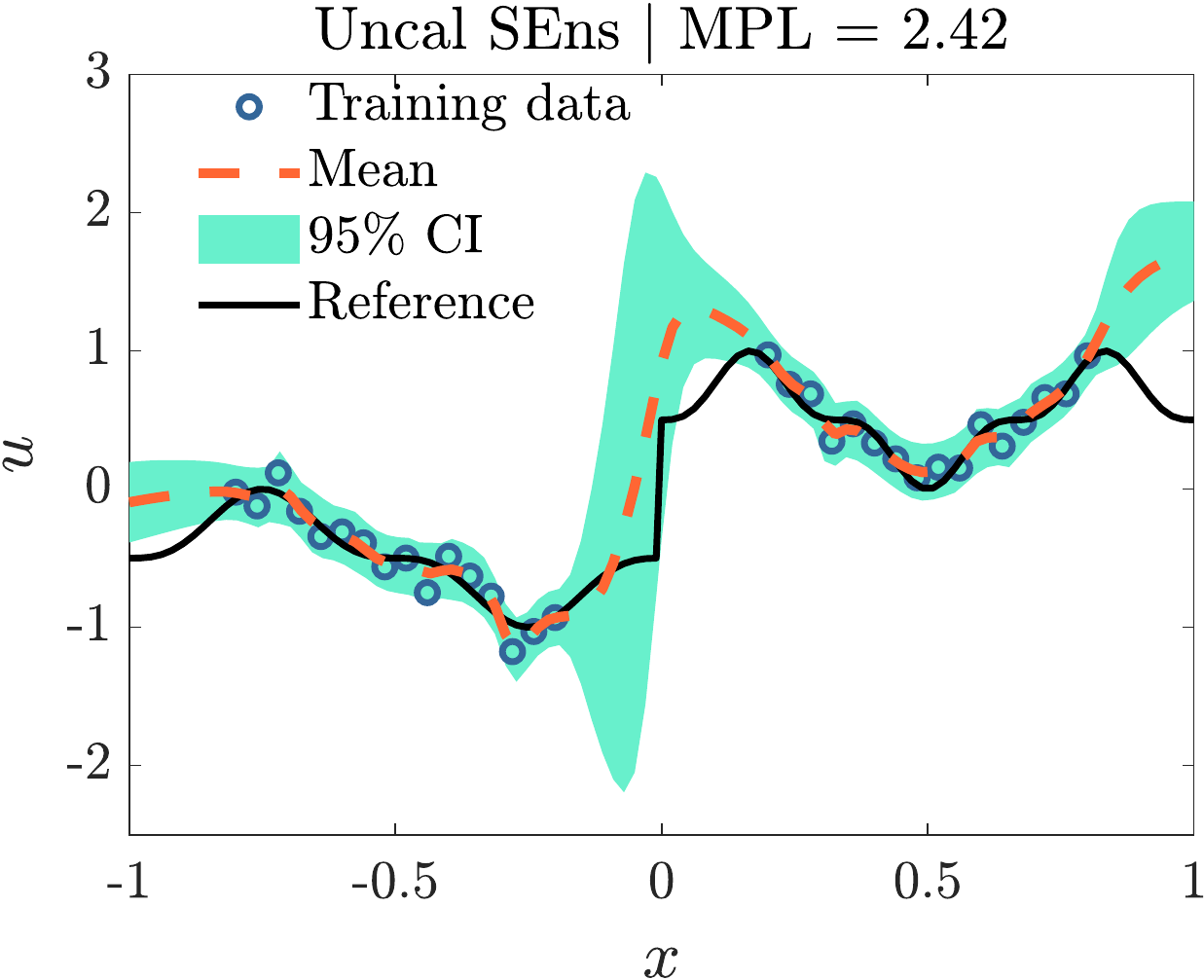}}
	\subcaptionbox{}{}{\includegraphics[width=0.32\textwidth]{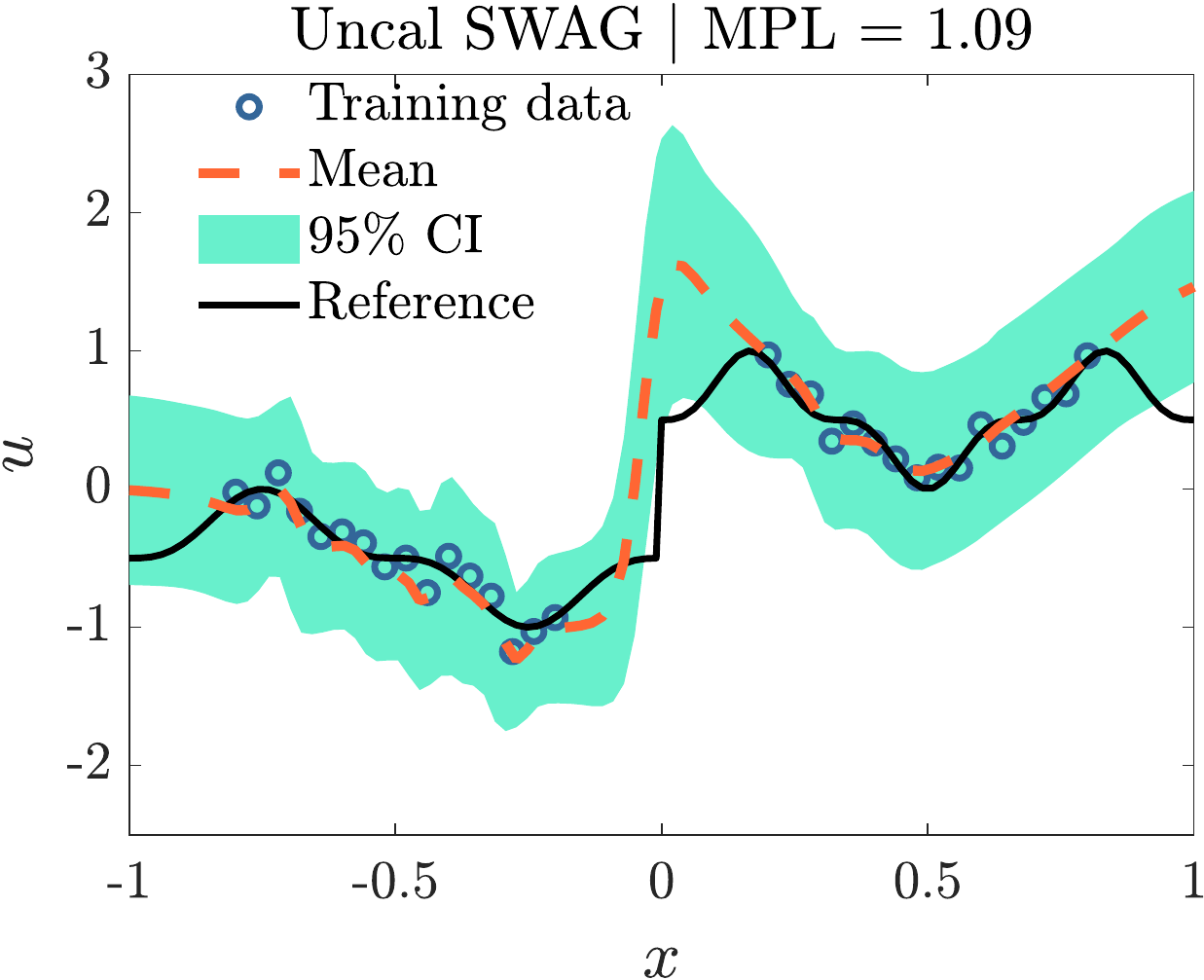}}
	\subcaptionbox{}{}{\includegraphics[width=0.32\textwidth]{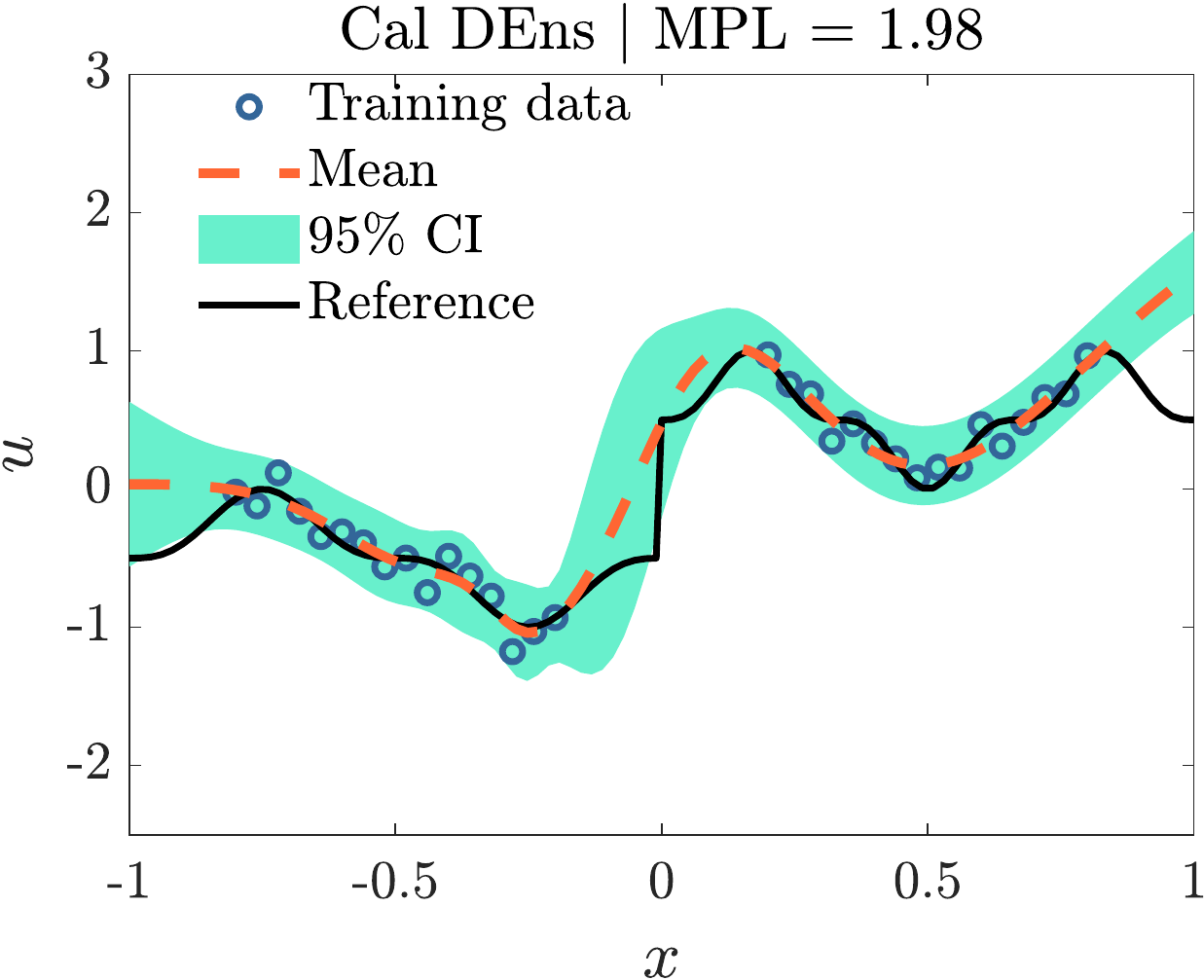}}
	\subcaptionbox{}{}{\includegraphics[width=0.32\textwidth]{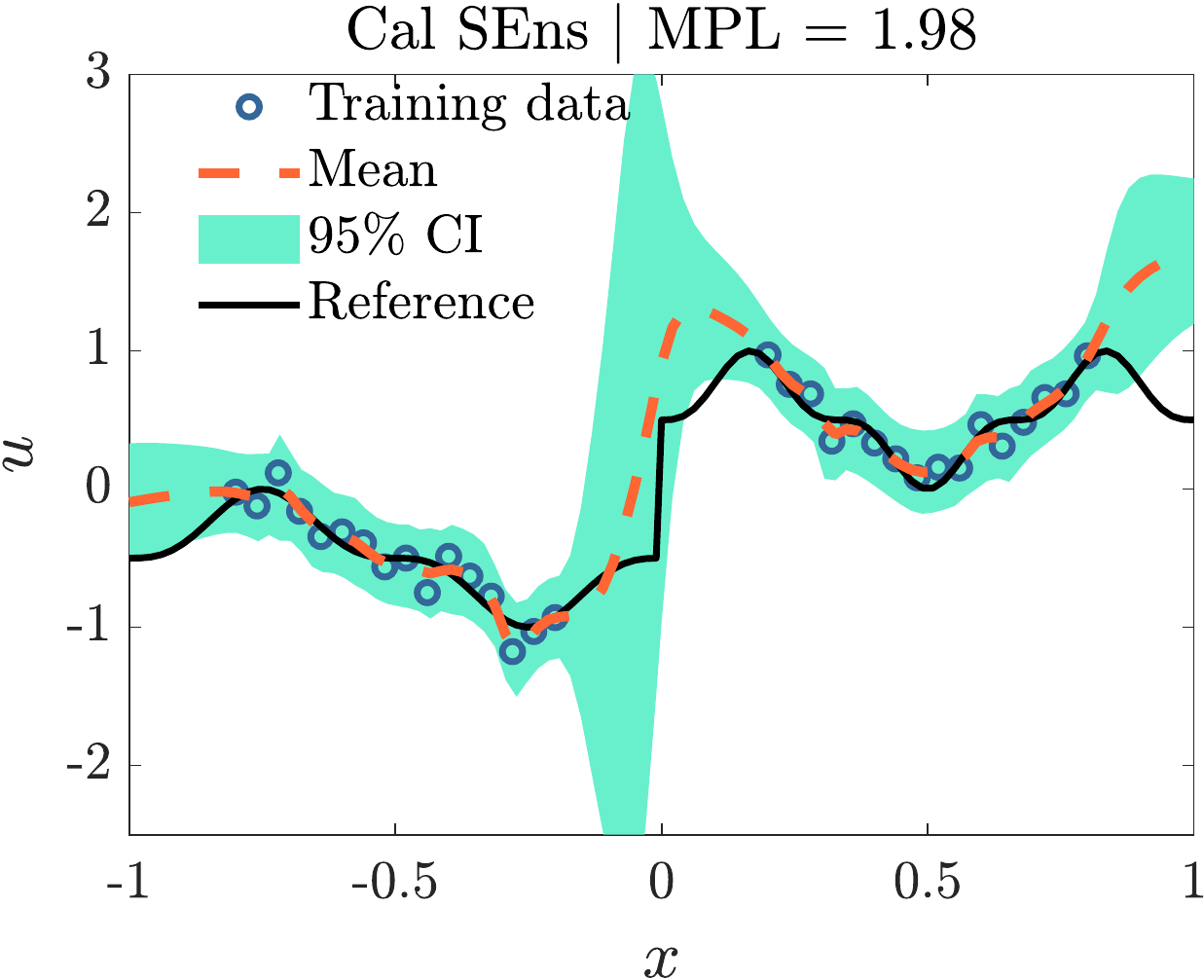}}
	\subcaptionbox{}{}{\includegraphics[width=0.32\textwidth]{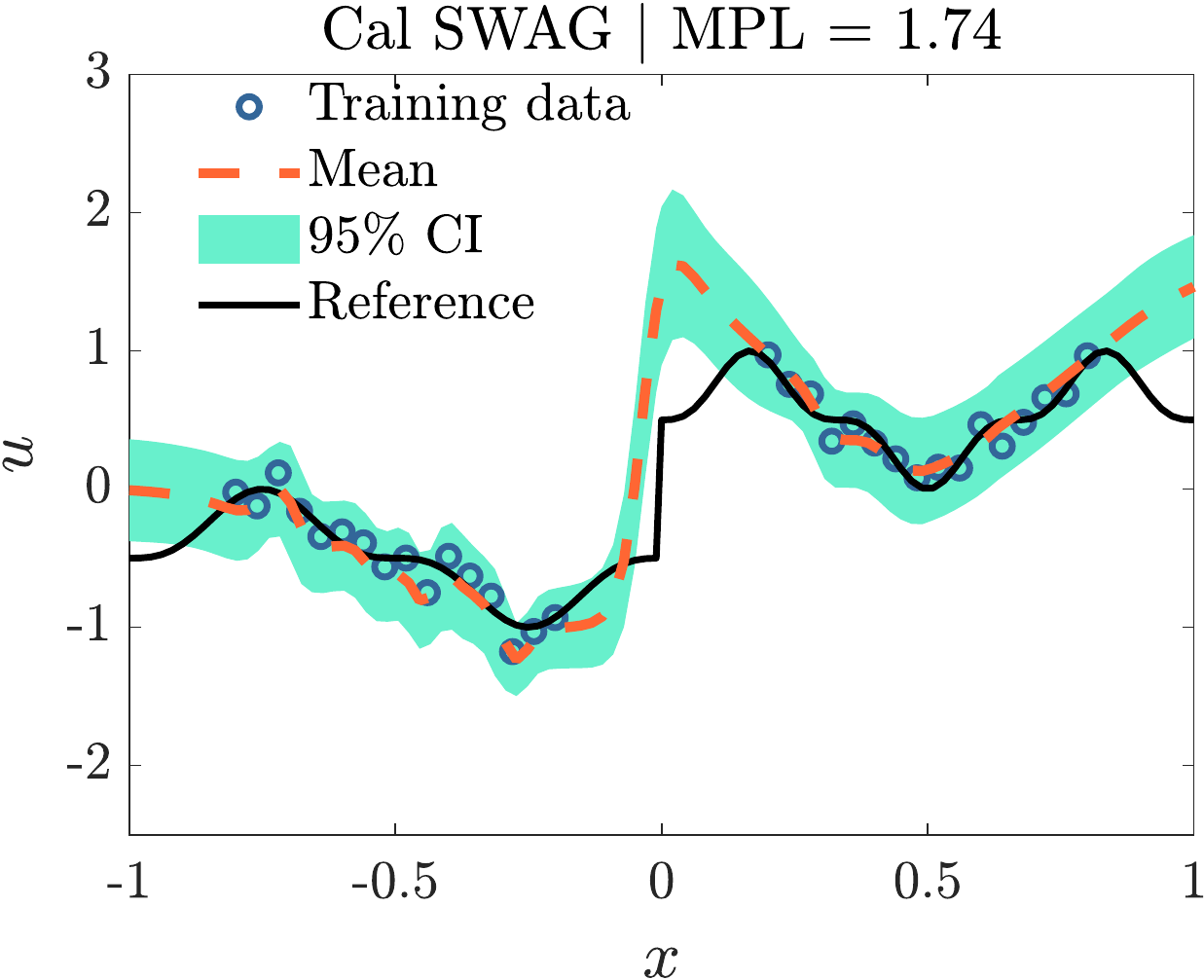}}
	\subcaptionbox{}{}{\includegraphics[width=0.32\textwidth]{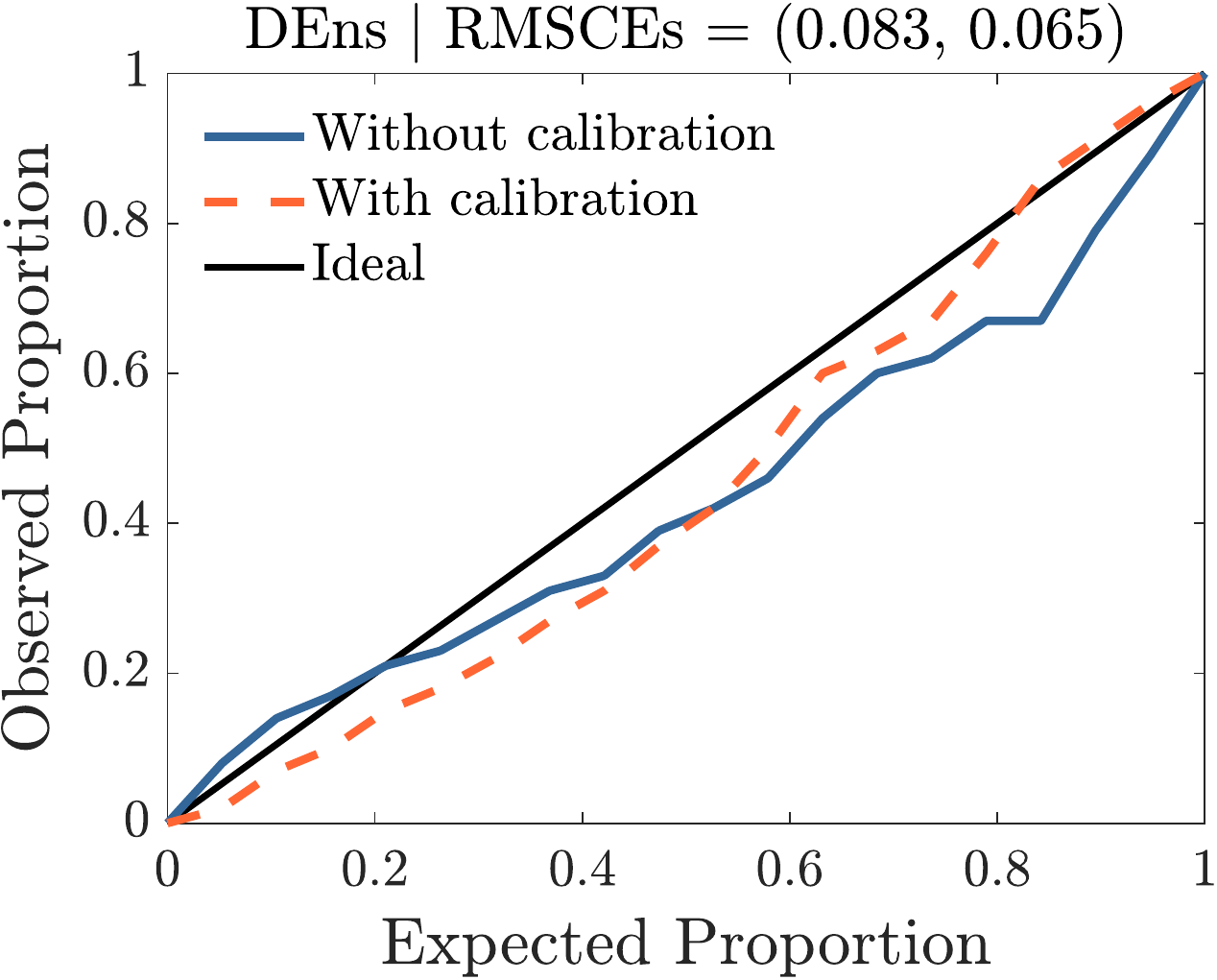}}
	\subcaptionbox{}{}{\includegraphics[width=0.32\textwidth]{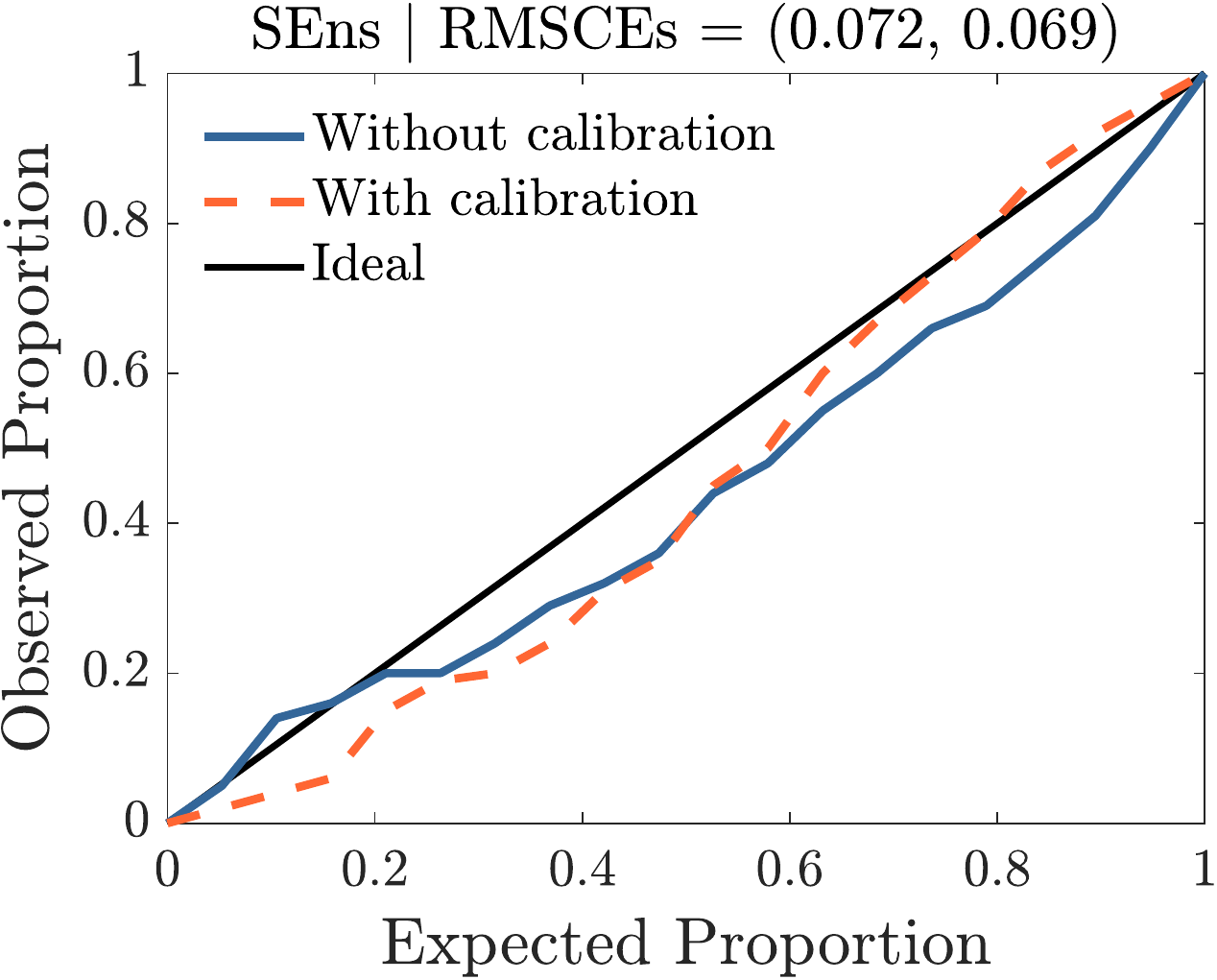}}
	\subcaptionbox{}{}{\includegraphics[width=0.32\textwidth]{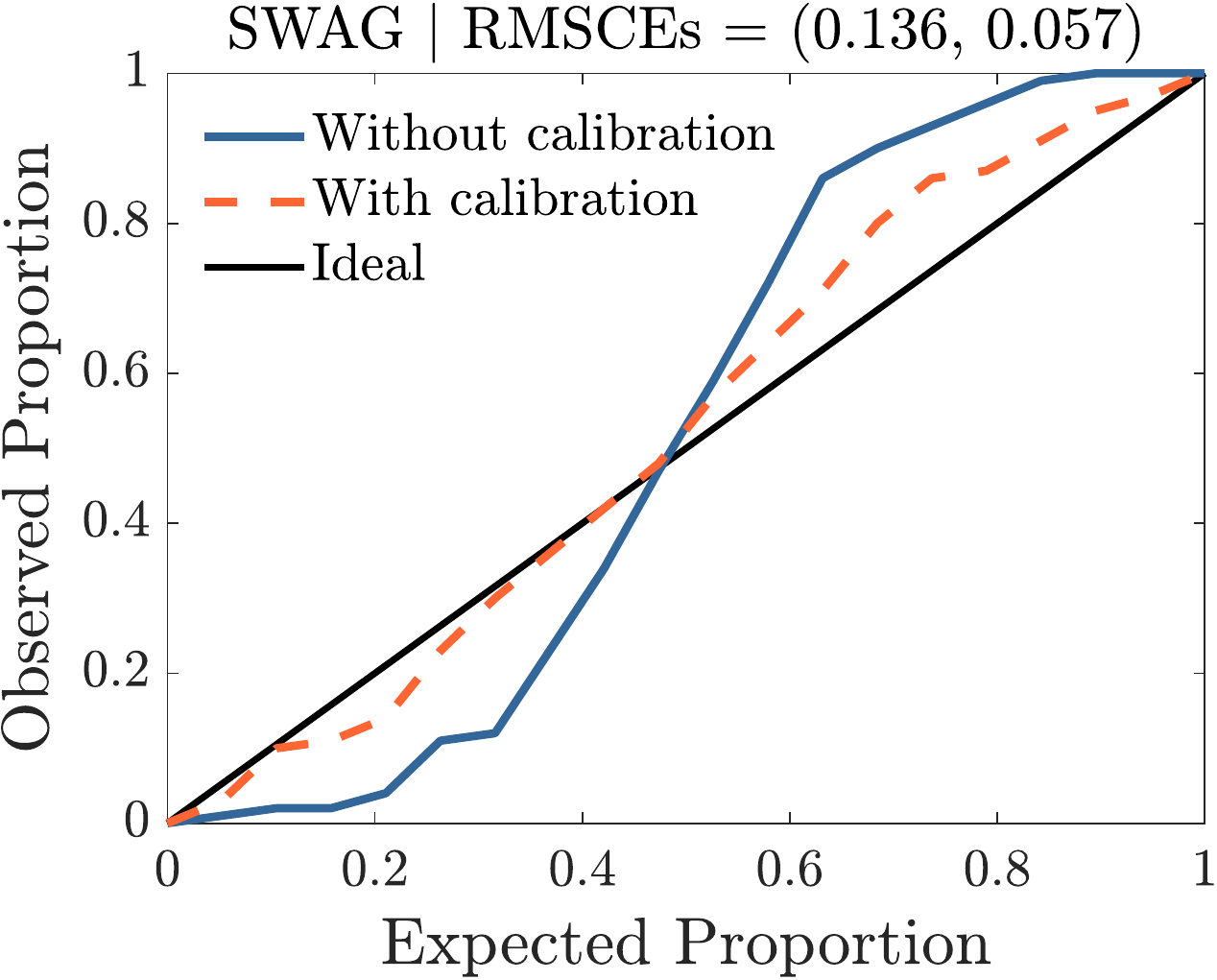}}
	\caption{
		Function approximation problem of Eq.~\eqref{eq:comp:func:func} | \textit{Known homoscedastic noise}: training data and exact function, as well as the mean and total uncertainty ($95\%$ CI) predictions of DEns, SEns, and SWAG.
		\textbf{Top row:} uncalibrated predictions. 
		\textbf{Middle row:} calibrated predictions (Section~\ref{sec:eval:calib}). 
		\textbf{Bottom row:} calibration plots (see also Fig.~\ref{fig:eval:eval:misscal}) and RMSCEs before and after post-training calibration (in the parentheses). 
	}
	\label{fig:comp:func:homosc:kno:res:3}
\end{figure}

\begin{figure}[H]
	\centering
	\subcaptionbox{}{}{\includegraphics[width=0.32\textwidth]{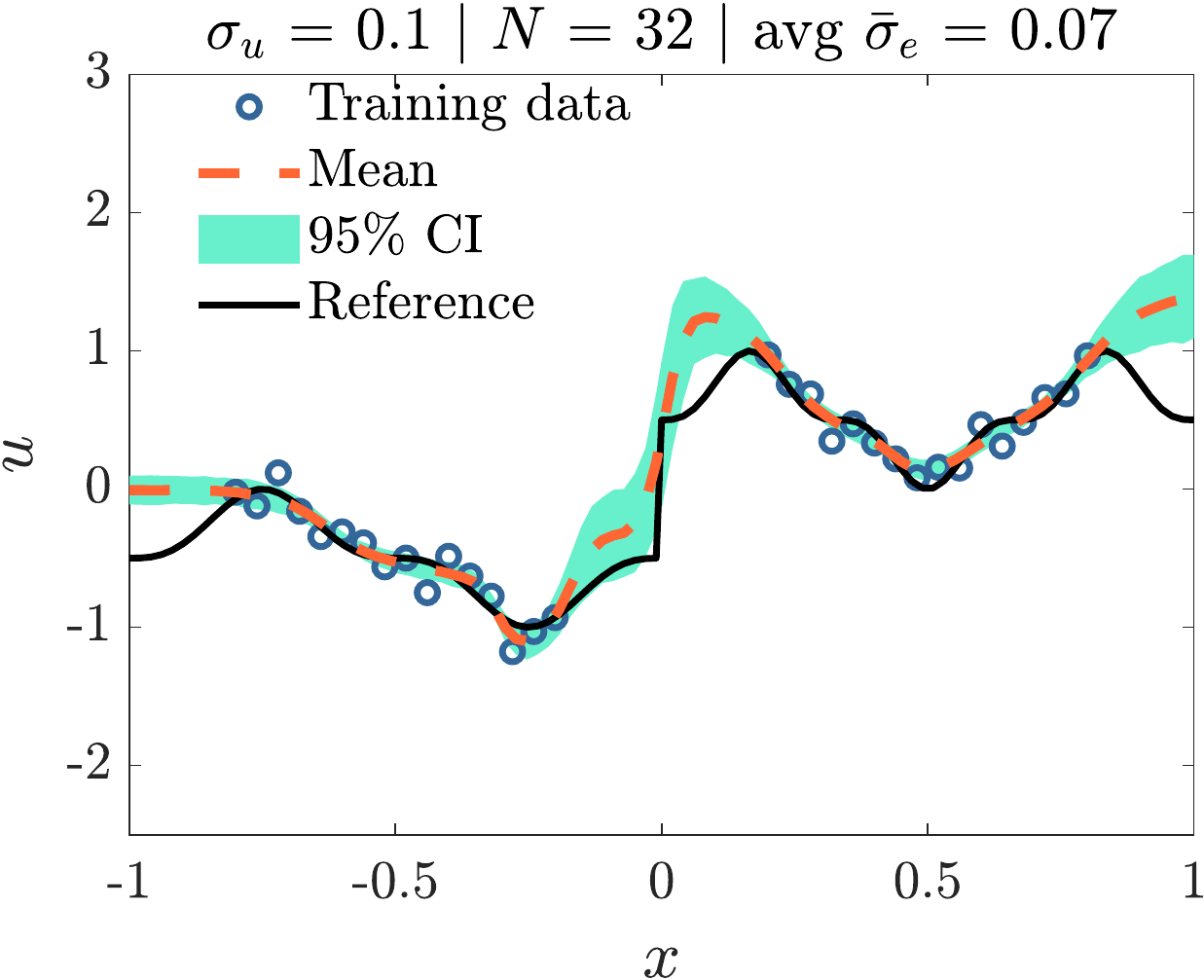}}
	\subcaptionbox{}{}{\includegraphics[width=0.32\textwidth]{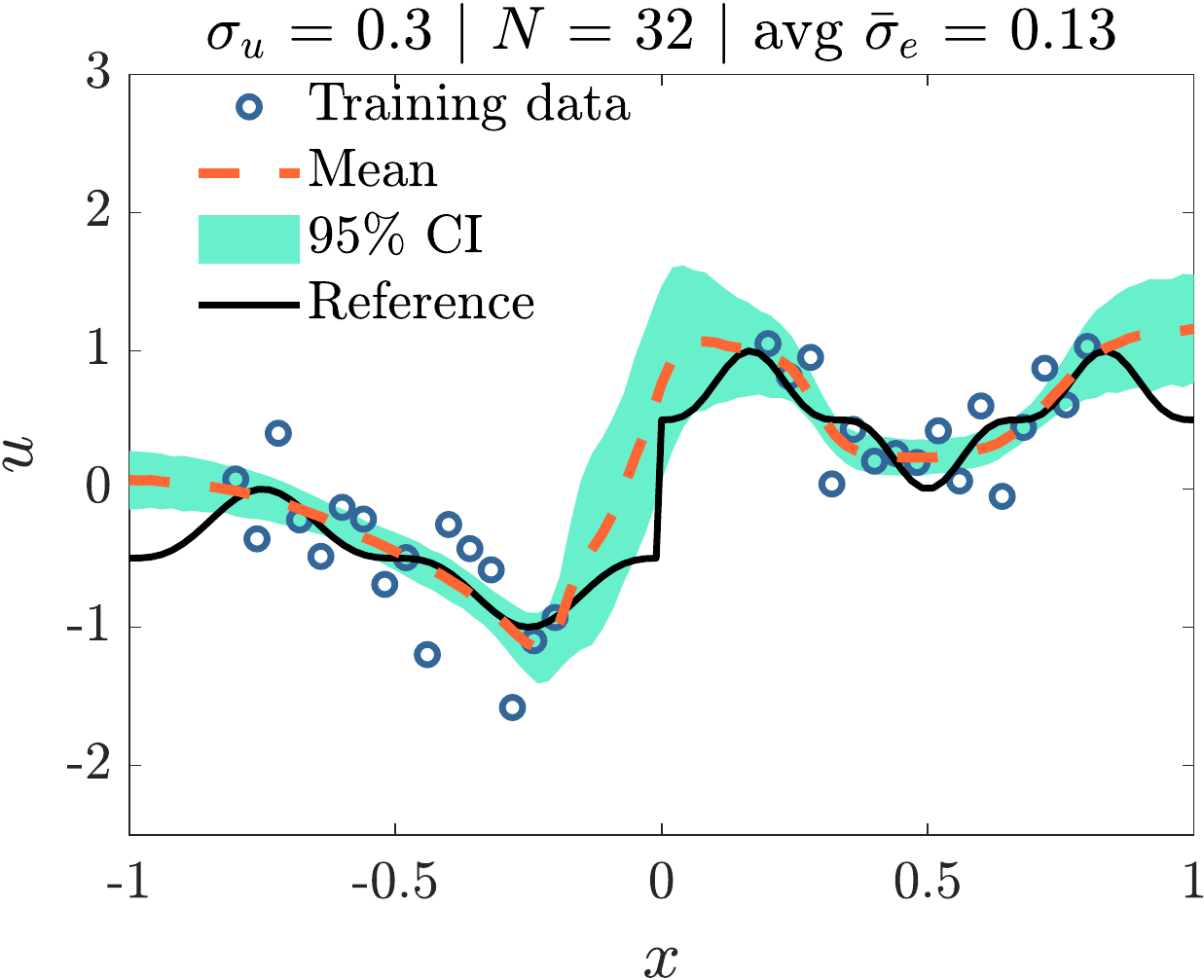}}
	\subcaptionbox{}{}{\includegraphics[width=0.32\textwidth]{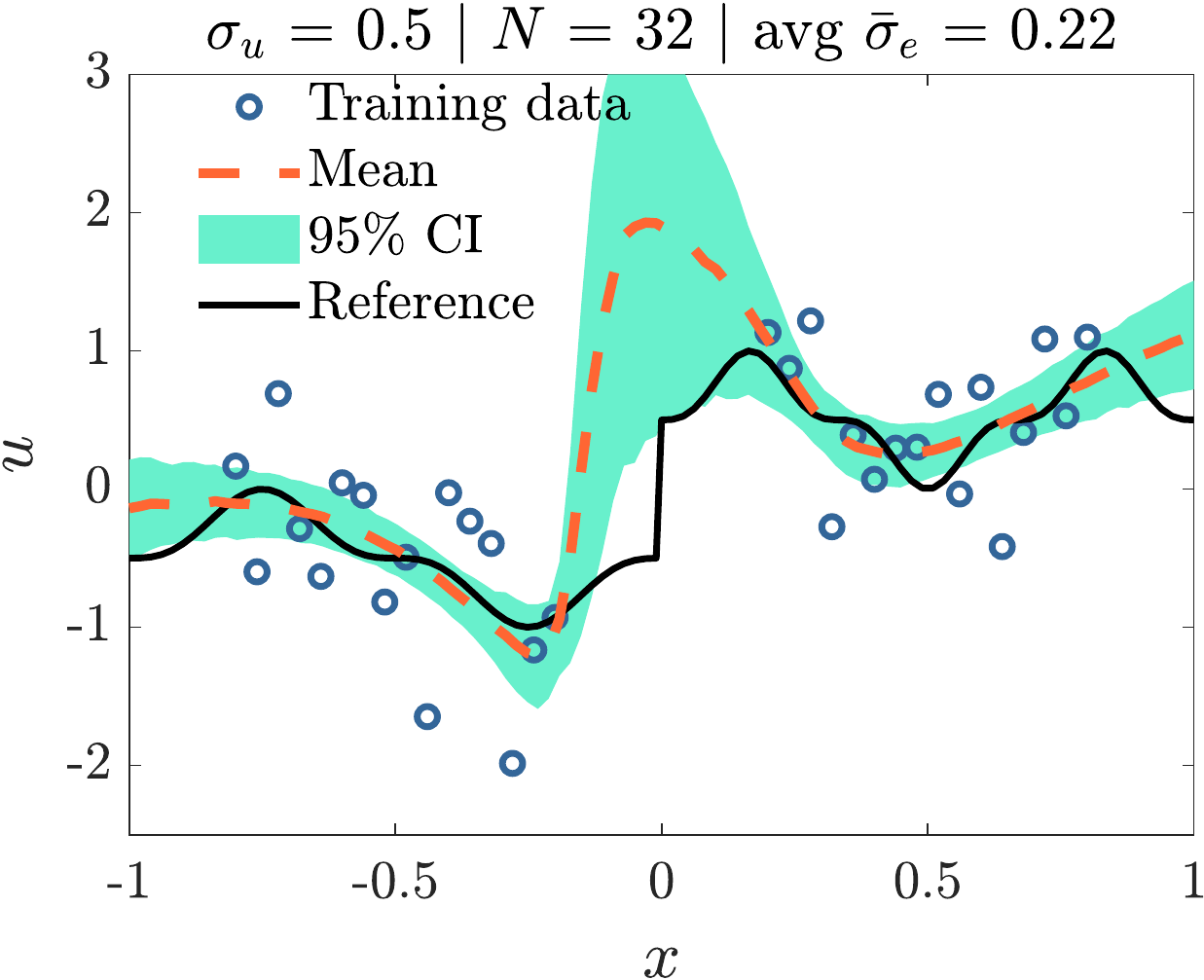}}
	\subcaptionbox{}{}{\includegraphics[width=0.32\textwidth]{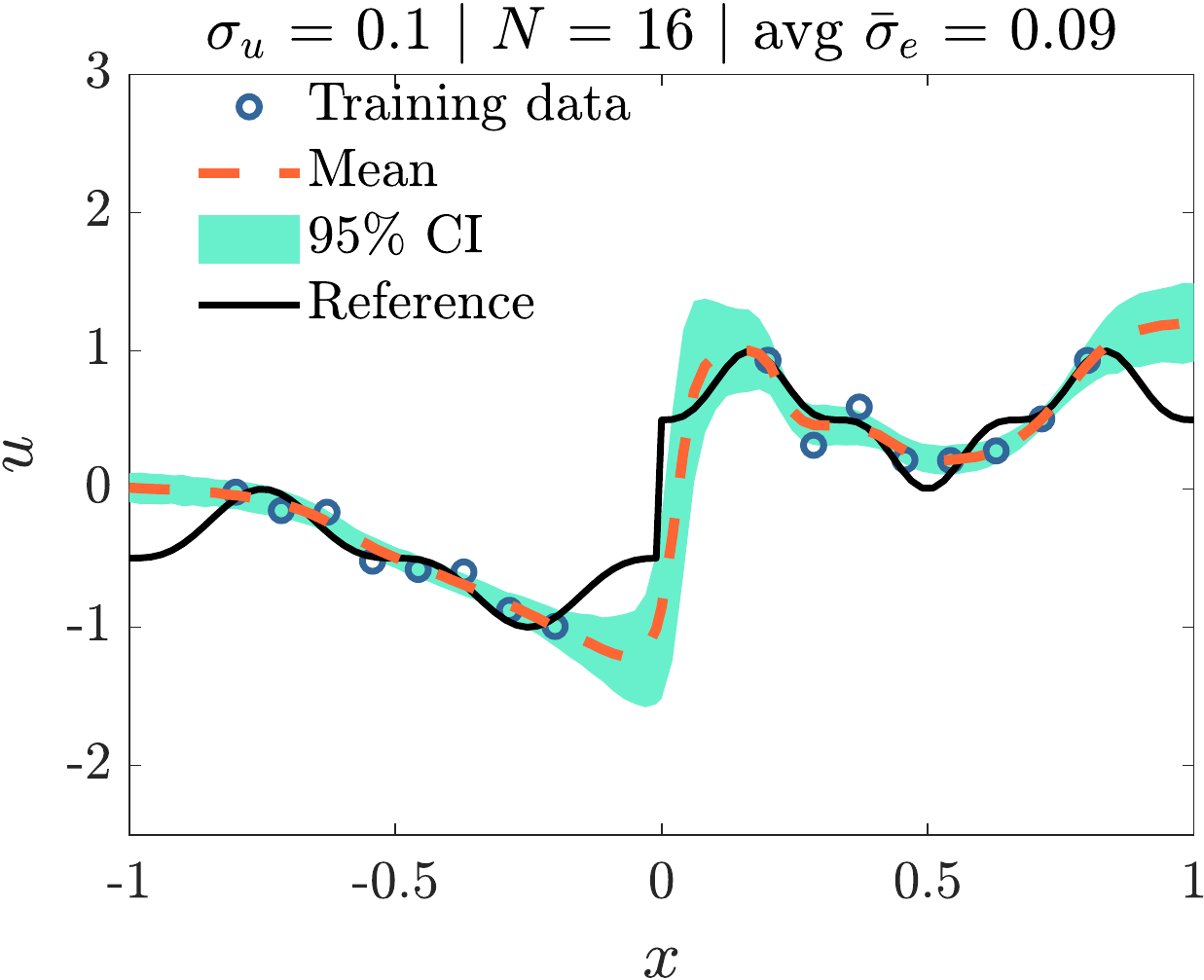}}
	\subcaptionbox{}{}{\includegraphics[width=0.32\textwidth]{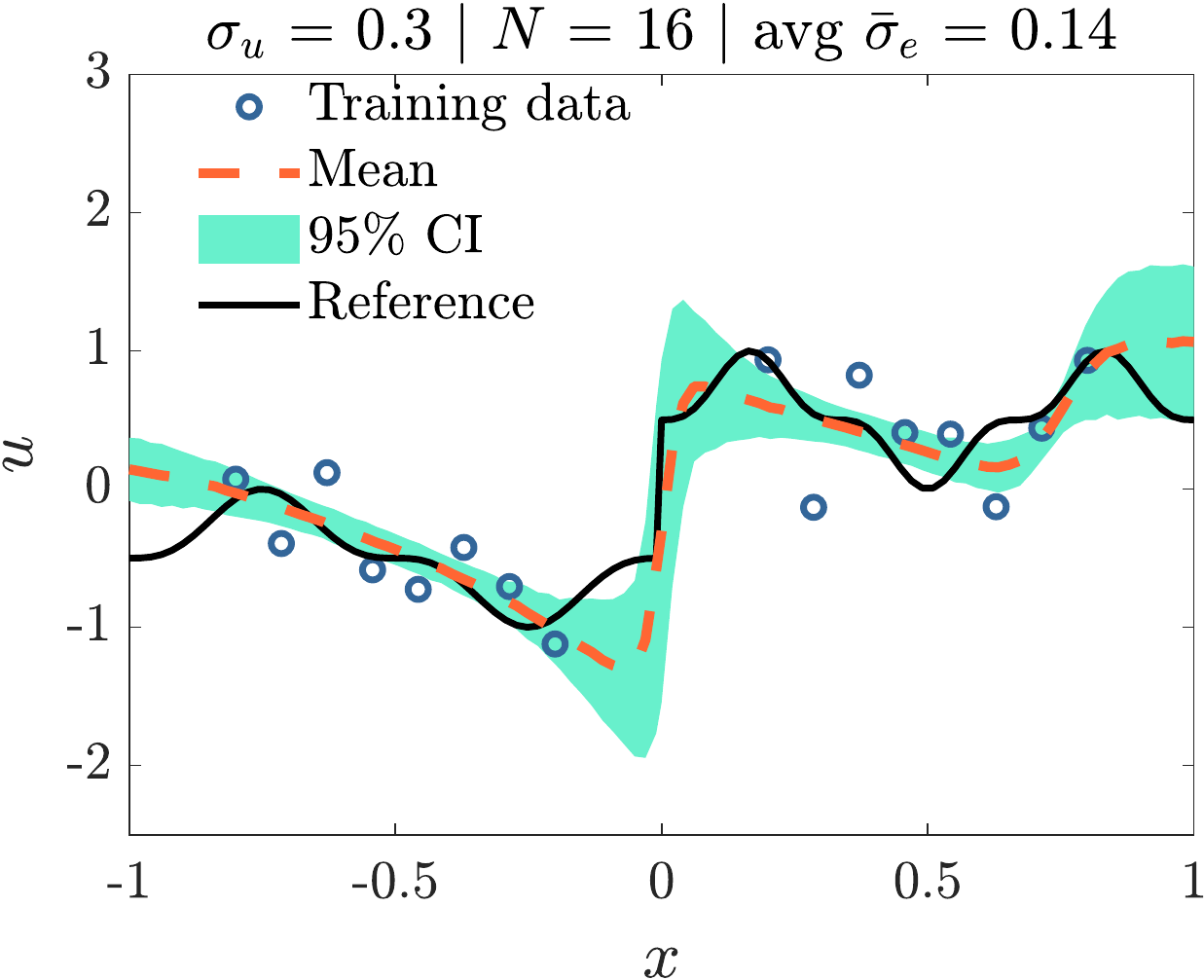}}
	\subcaptionbox{}{}{\includegraphics[width=0.32\textwidth]{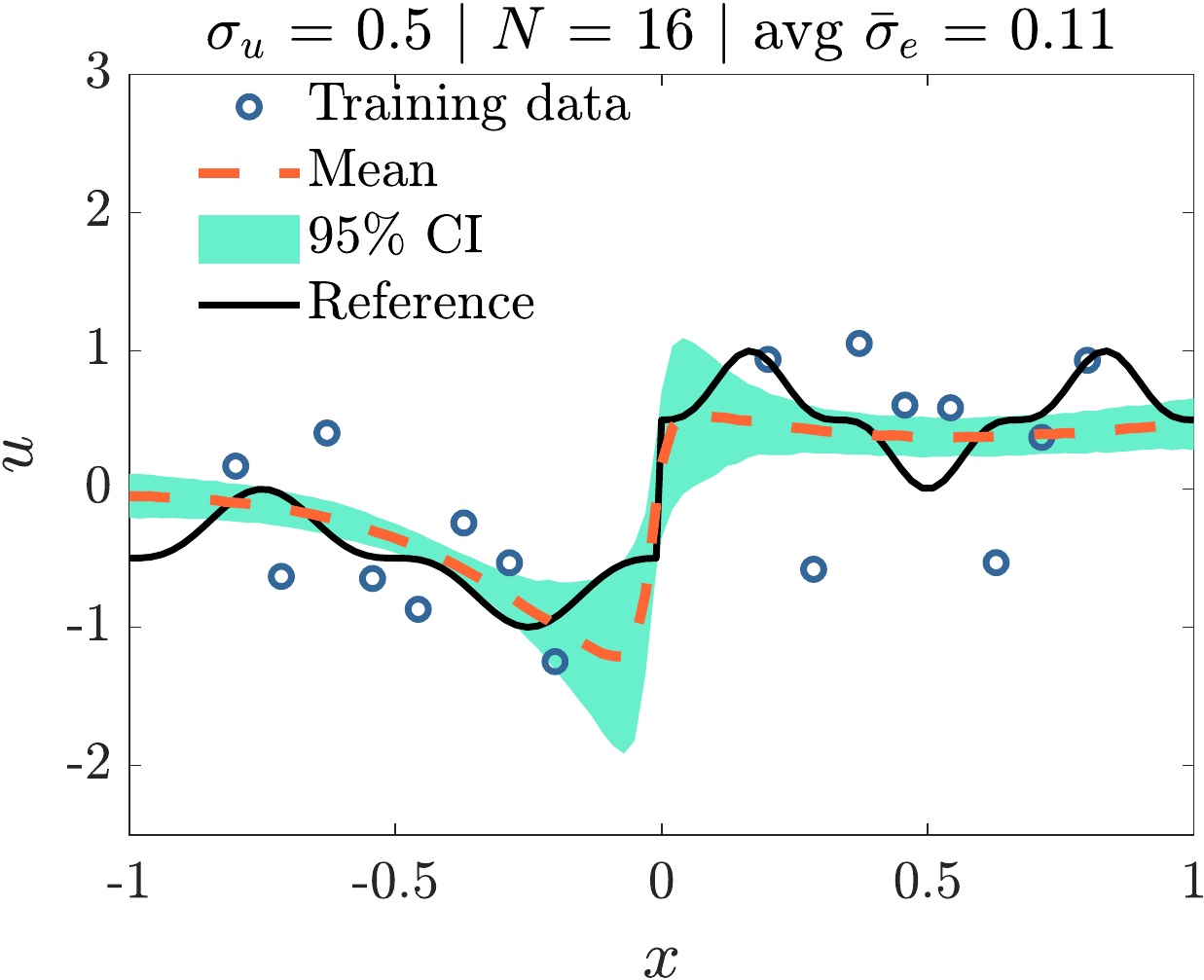}}
	\caption{
		Function approximation problem of Eq.~\eqref{eq:comp:func:func} | \textit{Known homoscedastic noise}:
		epistemic uncertainty of MCD (avg $\sigma_e^2$ along $x$) does not increase with increasing noise magnitude and with decreasing dataset size.
		Shown here are the training data and exact function, as well as the mean and epistemic uncertainty ($95\%$ CI) predictions, as obtained by MCD in conjunction with three noise scales ($\sigma_u=$ 0.1: left, 0.3: middle, 0.5: right) and two dataset sizes ($N=$ 32: top, 16: bottom).
	}
	\label{fig:comp:func:homosc:kno:epist:mcd}
\end{figure}

Here we evaluate the employed methods using OOD data (Table~\ref{tab:comp:func:homosc:kno:OOD}), and provide the results of post-training calibration (Figs.~\ref{fig:comp:func:homosc:kno:res:1}-\ref{fig:comp:func:homosc:kno:res:3}) and of MCD for various noise scales and dataset sizes (Fig.~\ref{fig:comp:func:homosc:kno:epist:mcd}).

\begin{table}[H]
	\centering
	\footnotesize
	\begin{tabular}{c|cc|ccccccc}
		\toprule
		Metric ($\times 10^2$)
		& GP & HMC & LD & MFVI & MCD & LA & DEns & SEns & SWAG \\
		\midrule
		RL2E ($\downarrow$) & 68.1 & 67.4 & 51.4 & \textbf{45.9} & 61.6 & 69.6 & 64.0 & 82.4 & 88.1 \\ 
		MPL ($\uparrow$) & 99.3 & 100.5 & 78.7 & 70.5 & 86.7 & 84.0 & \textbf{109.0} & 73.3 & 65.1 \\ 
		RMSCE ($\downarrow$) & 26.9 & 28.7 & 17.3 & \textbf{12.8} & 40.9 & 28.7 & 35.7 & 35.0 & 22.0 \\ 
		\midrule
		Cal RMSCE ($\downarrow$) & 24.3 & 25.8 & 24.3 & \textbf{11.4} & 36.1 & 27.2 & 31.7 & 31.1 & 27.0 \\ 
		PIW ($\downarrow$) & 165.1 & 142.3 & 150.6 & 218.0 & 63.5 & 165.5 & \textbf{61.9} & 113.3 & 151.2 \\ 
		SDCV ($\uparrow$) & 48.6 & 48.7 & 9.9 & 35.9 & 27.5 & 41.9 & 47.4 & \textbf{91.9} & 19.6 \\ 
		\bottomrule
	\end{tabular}
	\caption{
		Function approximation problem of Eq.~\eqref{eq:comp:func:func} | \textit{Known homoscedastic noise}: 
		although GP and HMC are expected to be the most accurate methods, final performance depends on the specific dataset considered and on tuning each technique individually.
		Further, techniques with large uncertainties for OOD data, although desirable, may perform worse in terms of MPL because of being conservative. See also \cite{yao2019quality,ashukha2021pitfalls} regarding the limitations of the evaluation metrics. Lastly, because calibration is performed using ID data, OOD performance may deteriorate. For example, compare the RMSCE and Cal RMSCE values corresponding to SWAG. 
		Here we evaluate OOD performance of nine UQ methods based on the metrics of Section~\ref{sec:eval}
		Noisy test data was used for all metrics.
		All values were obtained using noisy test data and they correspond to uncalibrated predictions, except for calibrated (Cal) RMSCE.
	}
	\label{tab:comp:func:homosc:kno:OOD}
\end{table}


\subsubsection{Unknown Student-t heteroscedastic noise}\label{app:comp:func:results:student}

In this section, each data noise sample for $x$ is given as $\epsilon_u = 0.5 |x| \epsilon_t$, where $\epsilon_t$ is a sample from a standard Student-t distribution with $\nu = 5$ degrees of freedom; see Fig.~\ref{fig:comp:func:hetero:datagen}(b). 
The variance of $\epsilon_u$ is given as $0.25 |x|^2 \frac{\nu}{\nu - 2}$. 
Similarly to Section~\ref{sec:comp:func:hetero}, the approximator NN has two outputs, one for the mean $u_{\theta}(x)$ and one for the variance $\sigma^2_u(x)$ of the Gaussian likelihood, and the results are obtained by h-HMC and h-MFVI.
This corresponds to a model misspecification case (Section~\ref{app:modeling:postemp}), because we employ a Gaussian likelihood, although the data distribution is Student-t.
For h-HMC and h-MFVI the results without posterior tempering for addressing this case are juxtaposed with the posterior tempering ones.

\begin{table}[H]
	\centering
	\footnotesize
	\begin{tabular}{c|cc|cc}
		\toprule
		Metric & \multicolumn{2}{c|}{Without PT} & \multicolumn{2}{c}{With PT ($\tau = 0.5$)}\\ \cline{2-5}
		($\times 10^2$) & h-HMC & h-MFVI & h-HMC & h-MFVI  \\
		\midrule
		RL2E ($\downarrow$)& \textbf{52.3} & 52.4 & 53.3 & \textbf{52.8} \\ 
		MPL ($\uparrow$)& \textbf{112.5} & 83.5 & \textbf{128.5} & 89.6 \\ 
		RMSCE ($\downarrow$)& \textbf{4.2} & 6.2 & 5.3 & \textbf{5.1} \\ 
		\bottomrule
	\end{tabular}
	\caption{
		Function approximation problem of Eq.~\eqref{eq:comp:func:func} | \textit{Unknown Student-t heteroscedastic noise}: ID performance evaluation of HMC and MFVI without and with posterior tempering (PT - Section~\ref{app:modeling:postemp}).
		All values were obtained using noisy test data and they correspond to uncalibrated predictions.
		Both techniques perform similarly based on RL2E, while HMC performs better based on MPL. Nevertheless, note that according to Figs.~\ref{fig:comp:func:student:hmc}-\ref{fig:comp:func:student:mfvi}, the uncertainty of HMC increases for OOD data, whereas the uncertainty of MFVI does not consistently. This important disadvantage of MFVI compared to more accurate techniques, such as HMC, may not be captured by the evaluation metrics.
	}
	\label{tab:comp:func:student}
\end{table}

\begin{figure}[!ht]
	\centering
	\subcaptionbox{}{}{\includegraphics[width=0.32\textwidth]{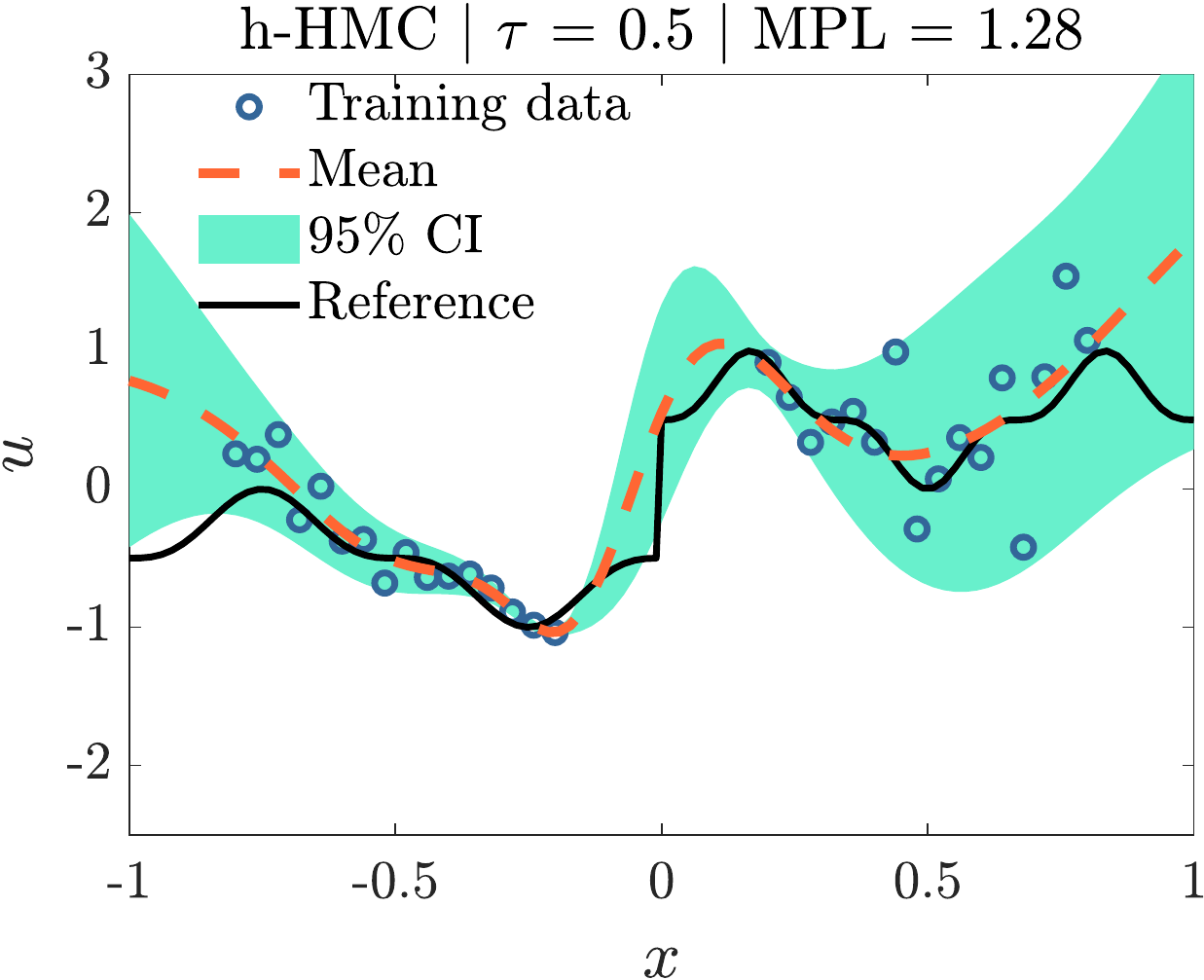}}
	\subcaptionbox{}{}{\includegraphics[width=0.32\textwidth]{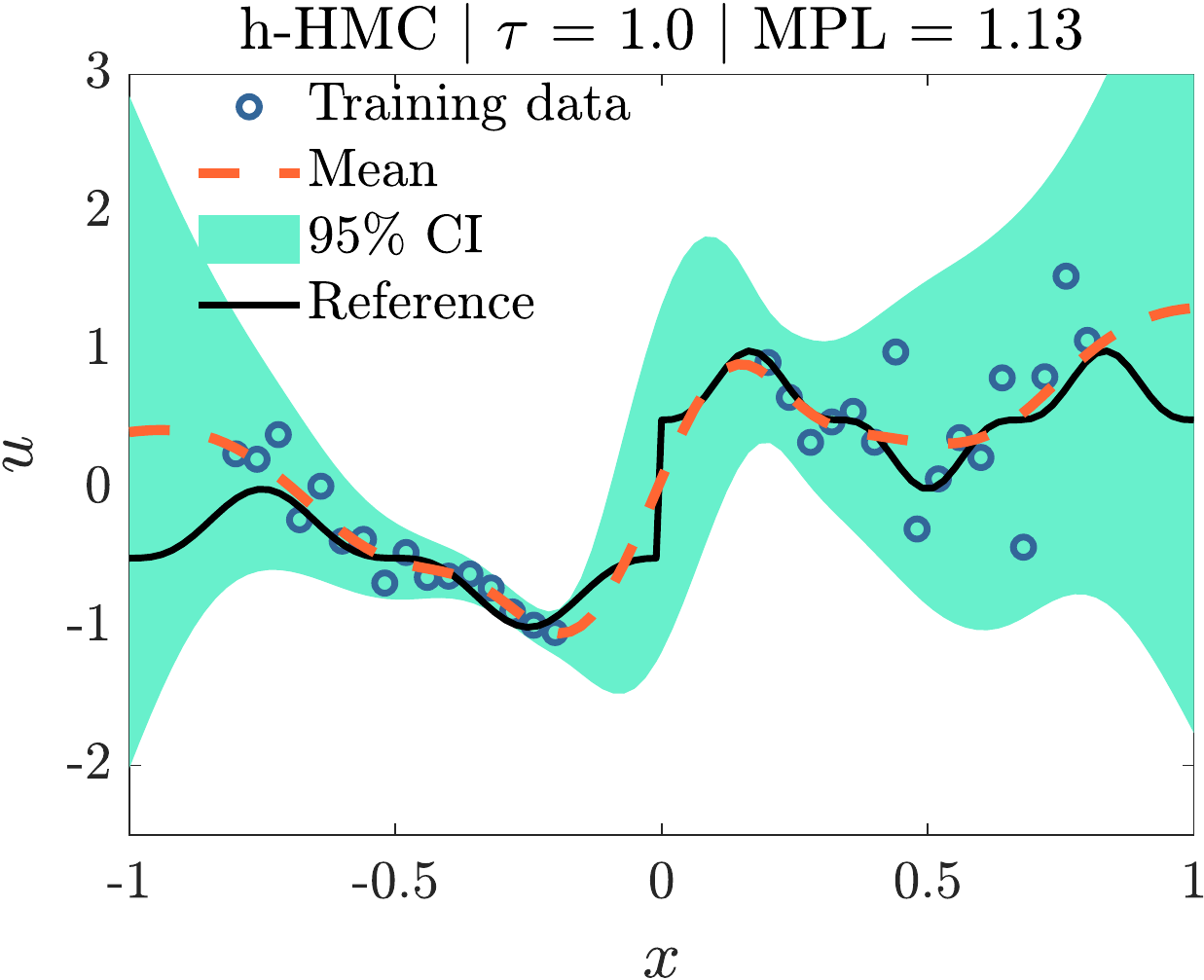}}
	\subcaptionbox{}{}{\includegraphics[width=0.32\textwidth]{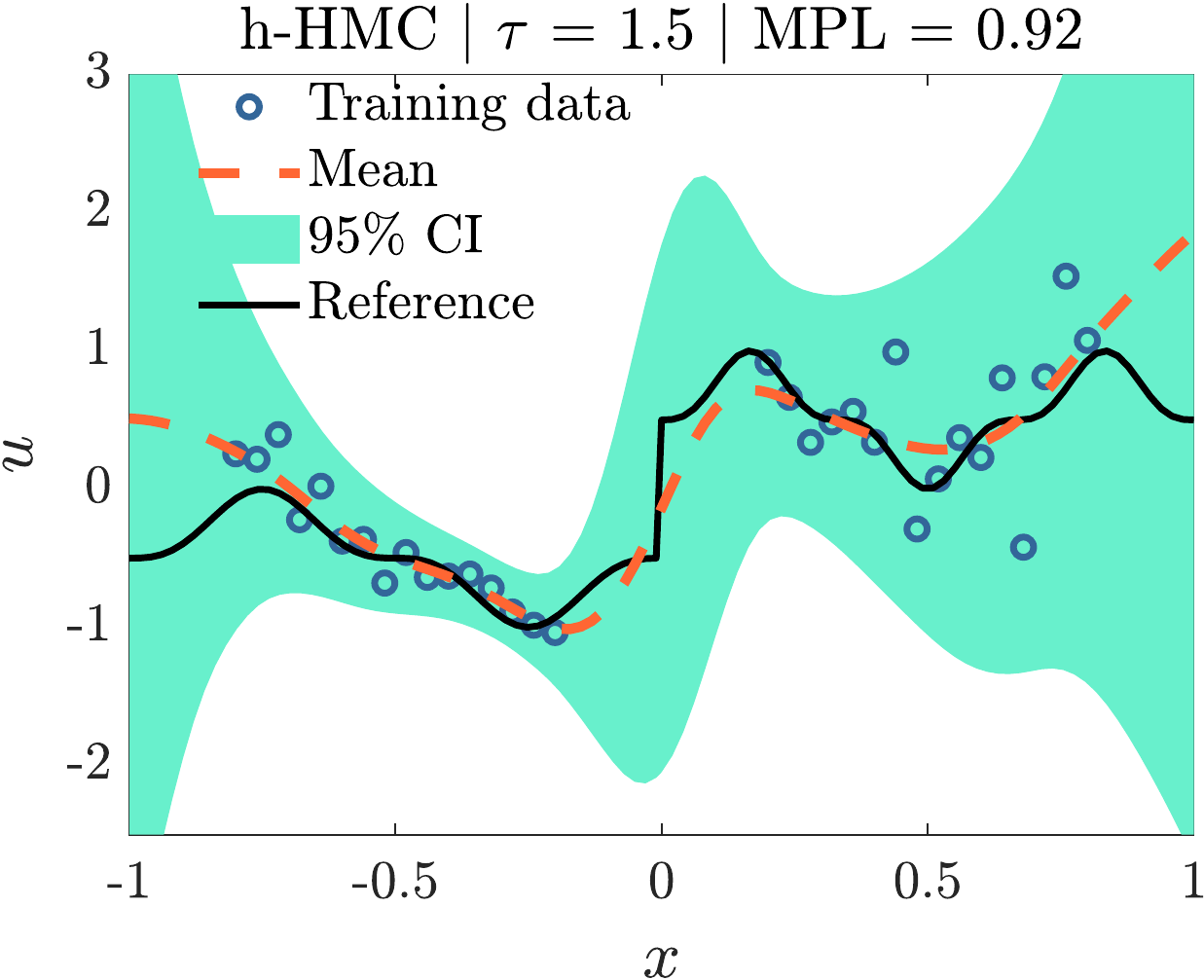}}
	\subcaptionbox{}{}{\includegraphics[width=0.32\textwidth]{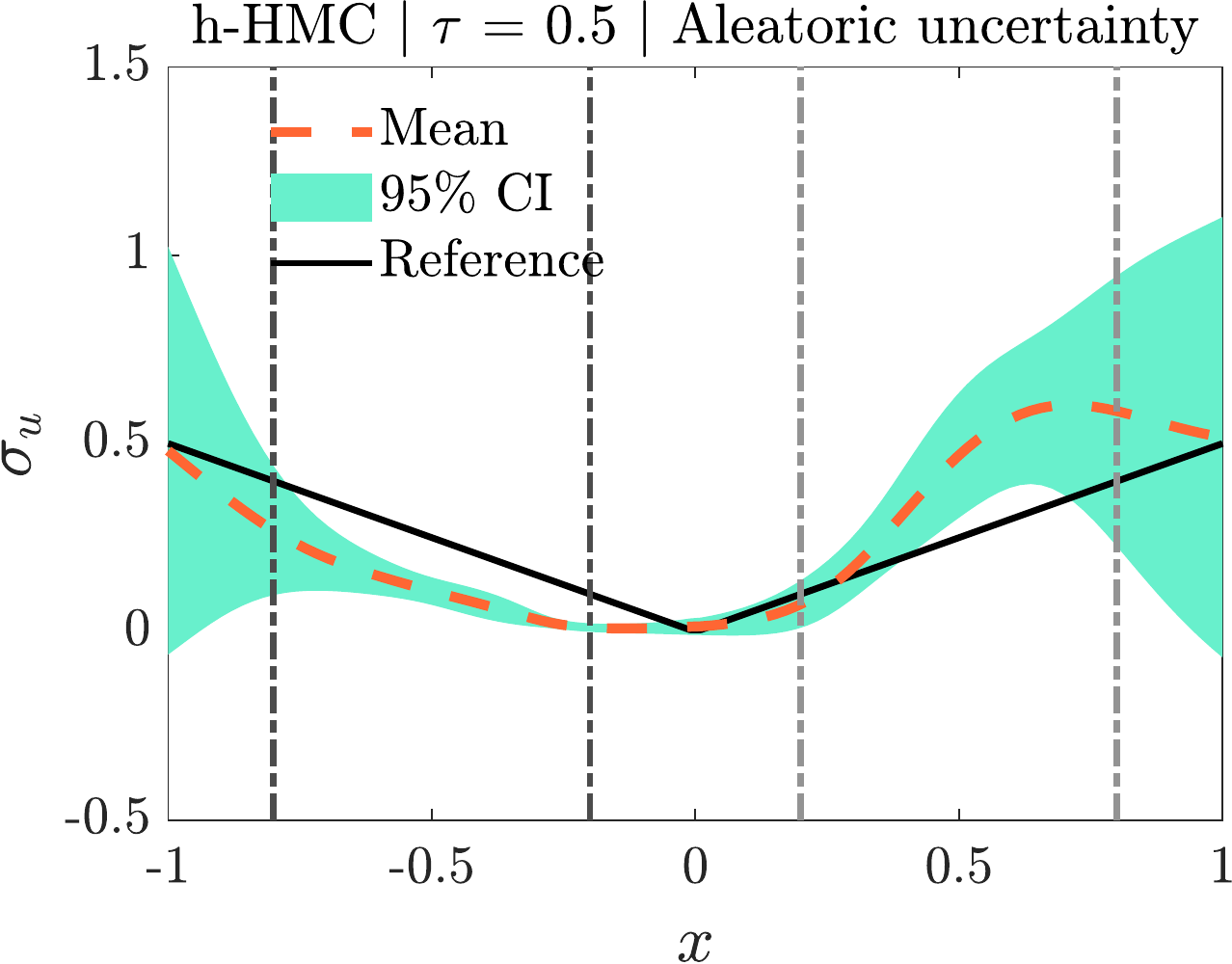}}
	\subcaptionbox{}{}{\includegraphics[width=0.32\textwidth]{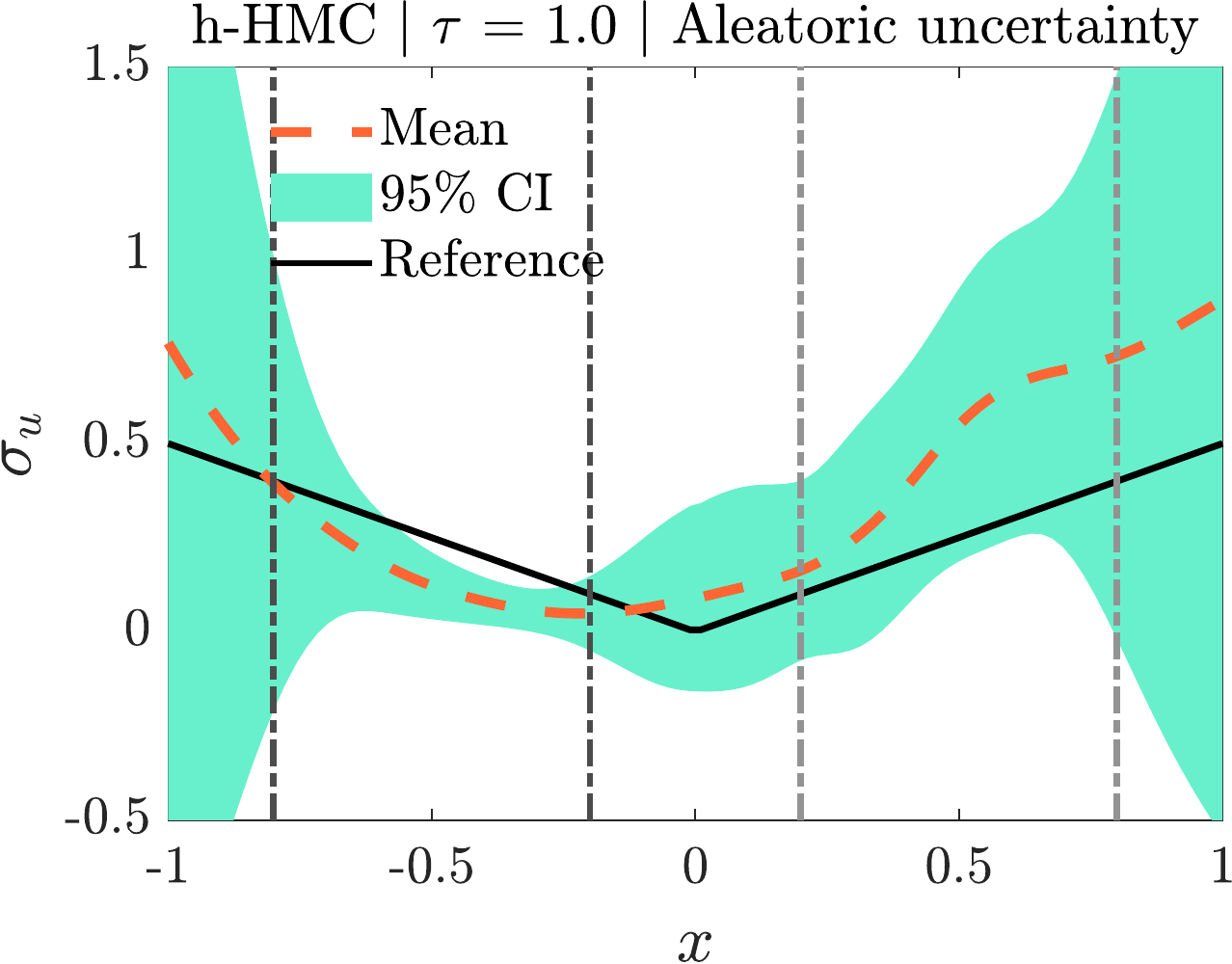}}
	\subcaptionbox{}{}{\includegraphics[width=0.32\textwidth]{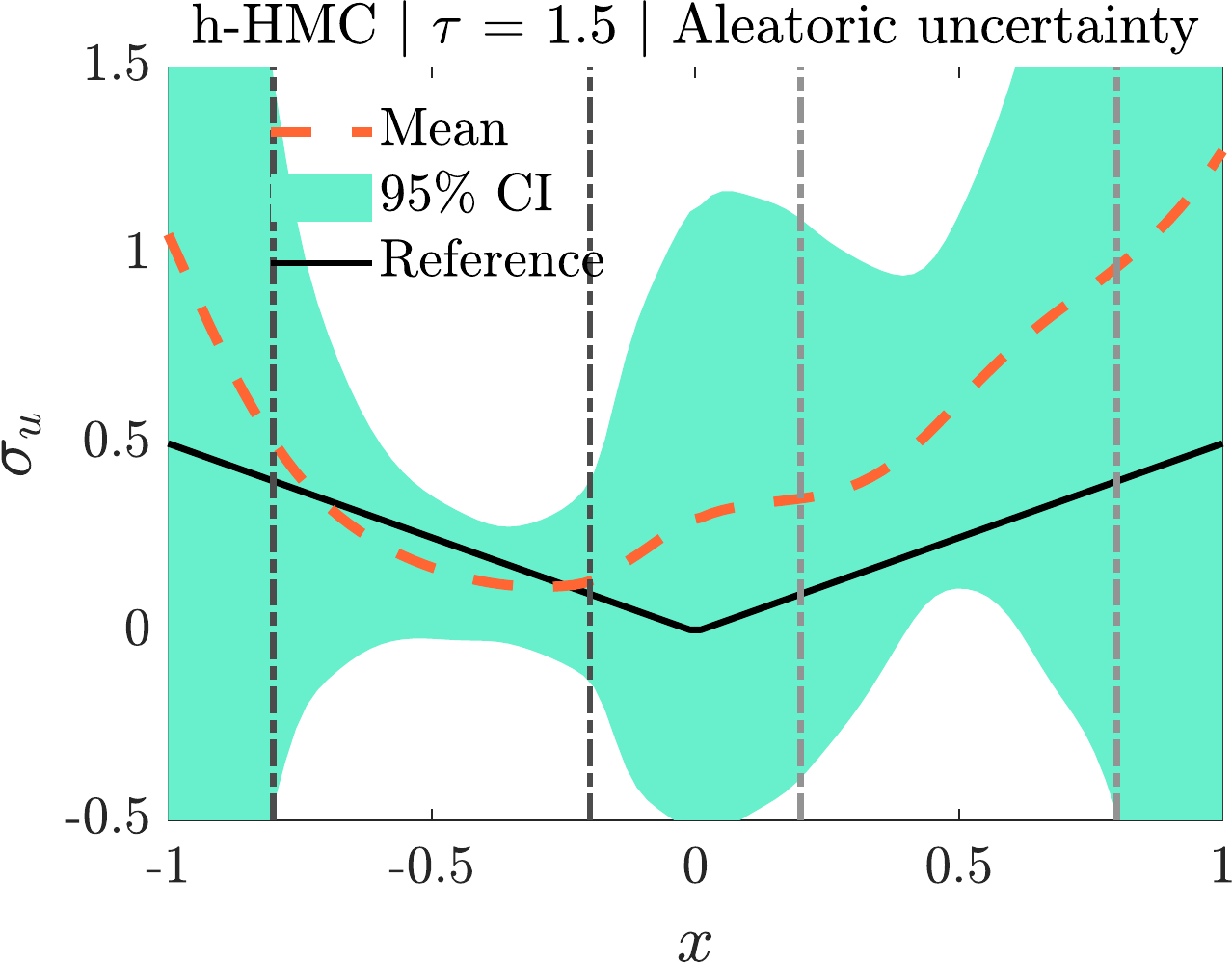}}
	\caption{
		Function approximation problem of Eq.~\eqref{eq:comp:func:func} | \textit{Unknown Student-t heteroscedastic noise}: by modifying the ``temperature'' $\tau$ we can control how conservative the predictions are.
		For this case of model misspecification, a cold posterior performs better in terms of predictive capacity (MPL); see also Table~\ref{tab:comp:func:student}.
		Note that data on the right side ``happen'' to be more noisy and this is reflected in the noise predictions.
		Here shown are h-HMC predictions for three values of $\tau$ (see Section~\ref{app:modeling:postemp}).
		\textbf{Top row:} mean and total uncertainty of $u(x)$. 
		\textbf{Bottom row:}
		mean and epistemic uncertainty of predicted aleatoric uncertainty $\bar{\sigma}_u(x)$.
		In the top row the training datapoints are also included, while in the bottom row the black and gray dot-dashed lines mark the training data intervals.
	}
	\label{fig:comp:func:student:hmc}
\end{figure}

\begin{figure}[!ht]
	\centering
	\subcaptionbox{}{}{\includegraphics[width=0.32\textwidth]{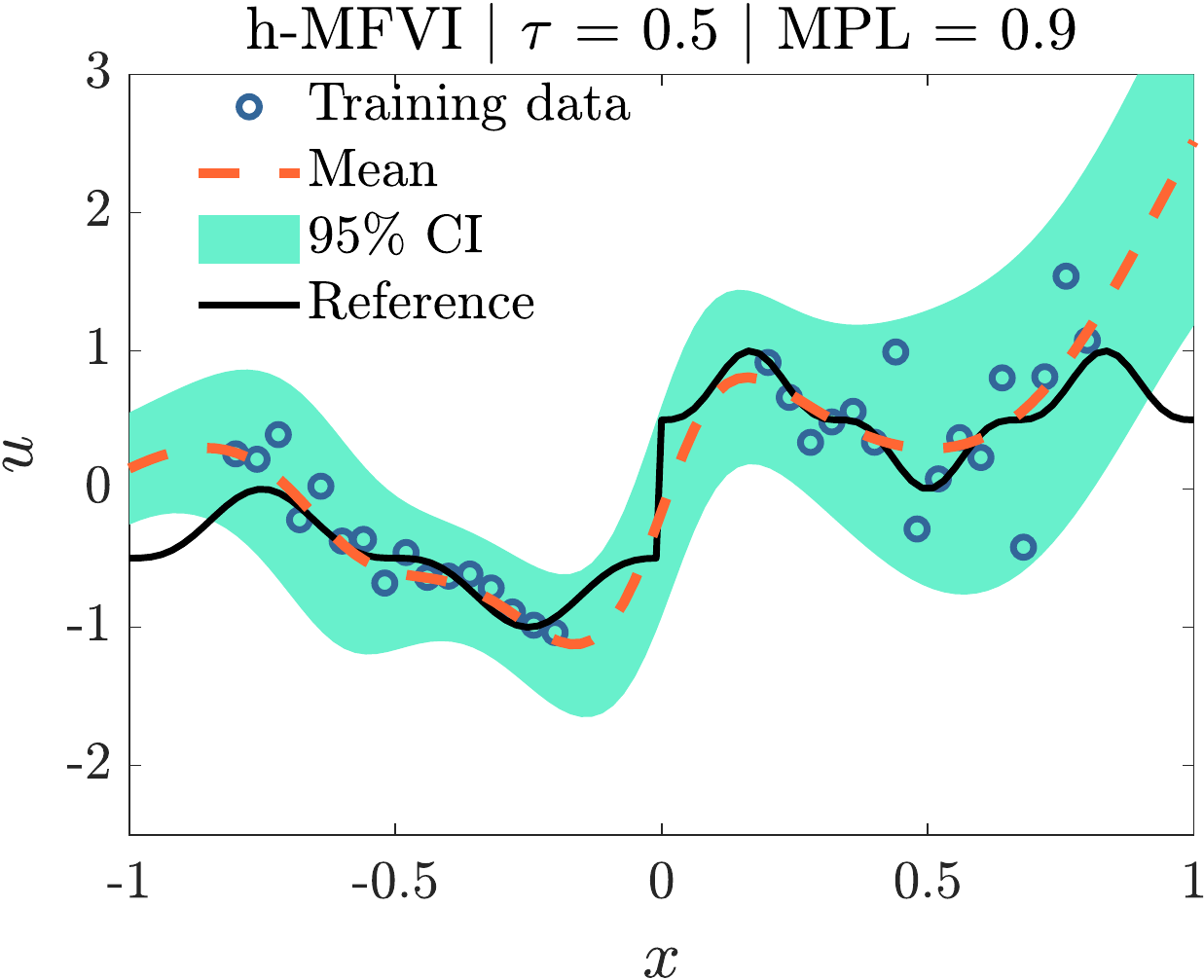}}
	\subcaptionbox{}{}{\includegraphics[width=0.32\textwidth]{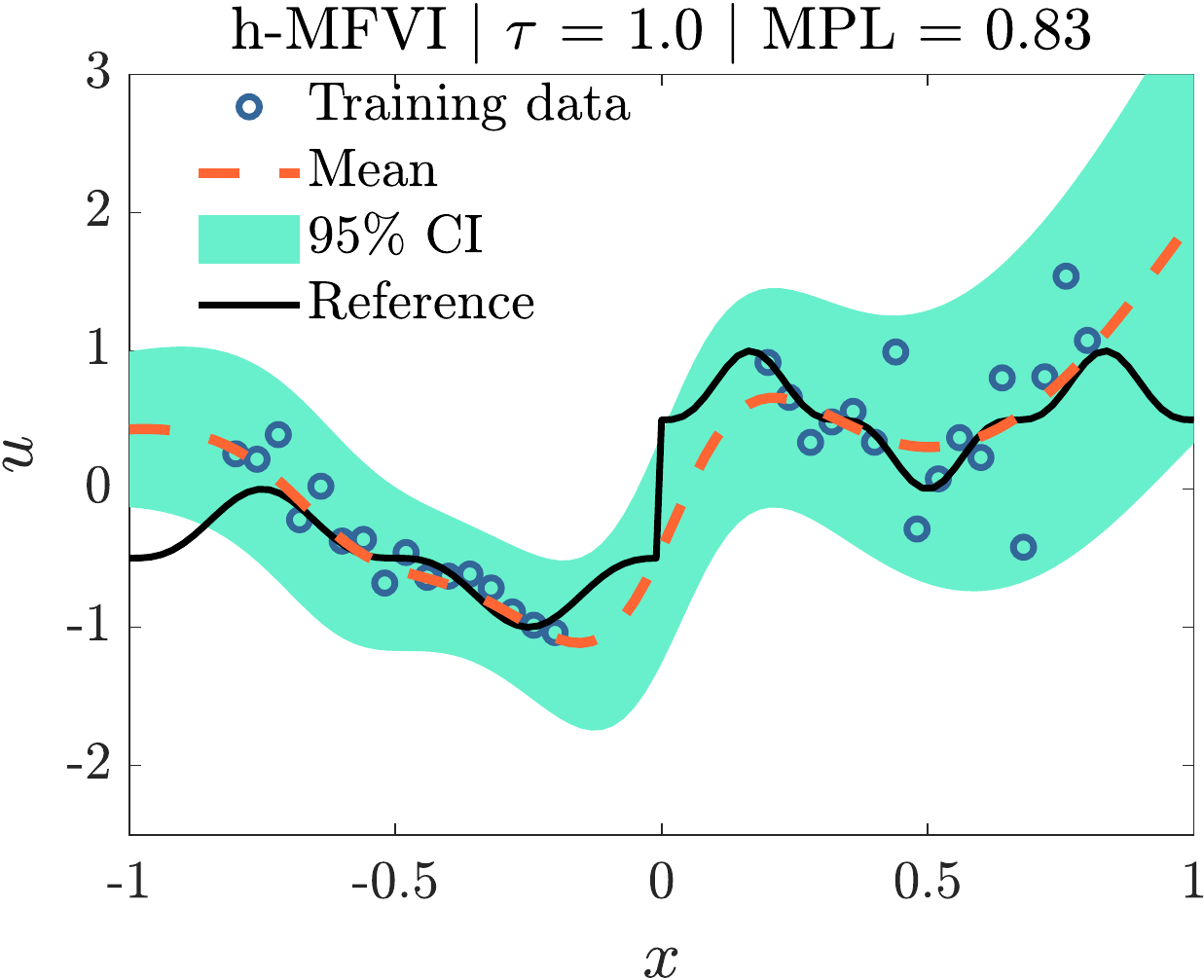}}
	\subcaptionbox{}{}{\includegraphics[width=0.32\textwidth]{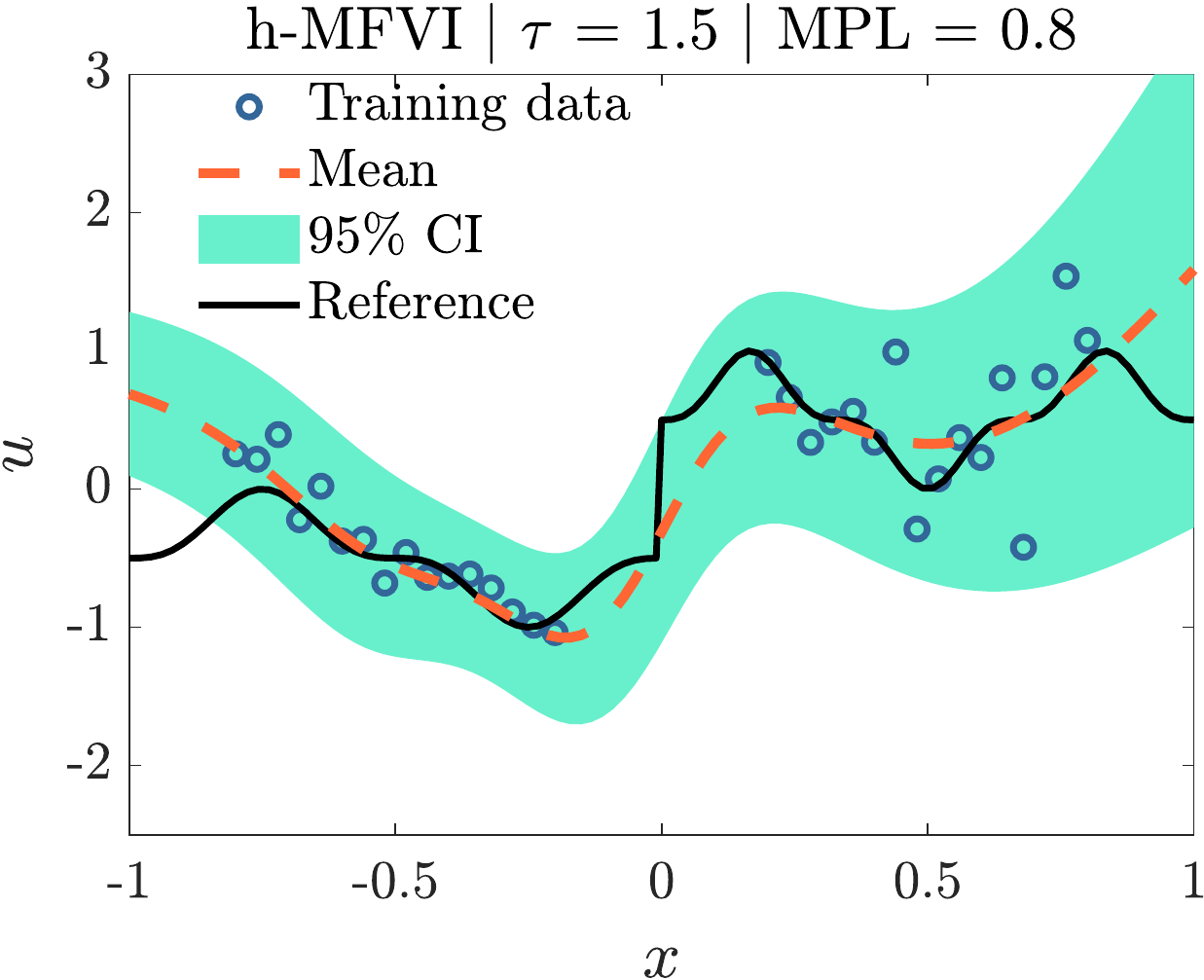}}
	\subcaptionbox{}{}{\includegraphics[width=0.32\textwidth]{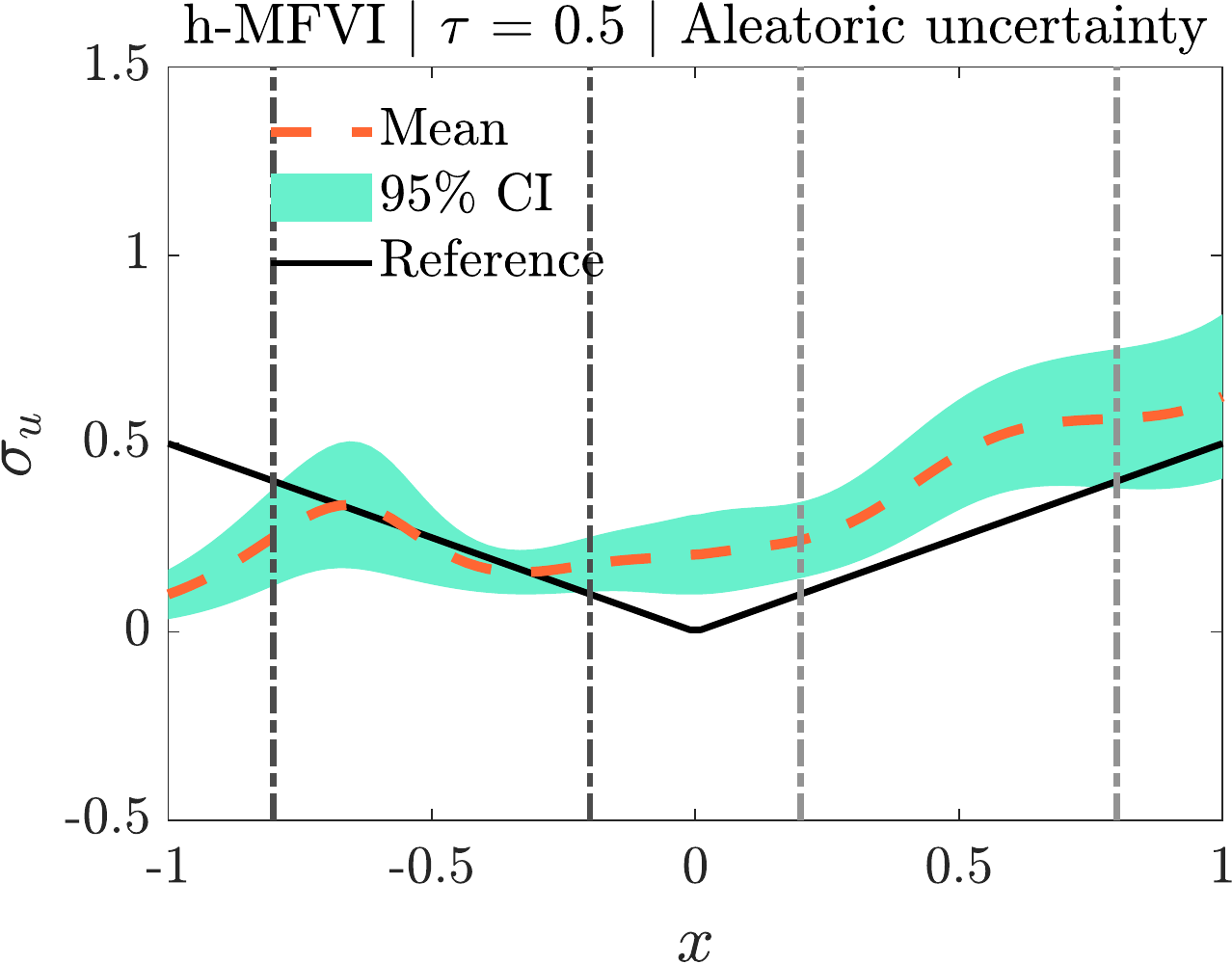}}
	\subcaptionbox{}{}{\includegraphics[width=0.32\textwidth]{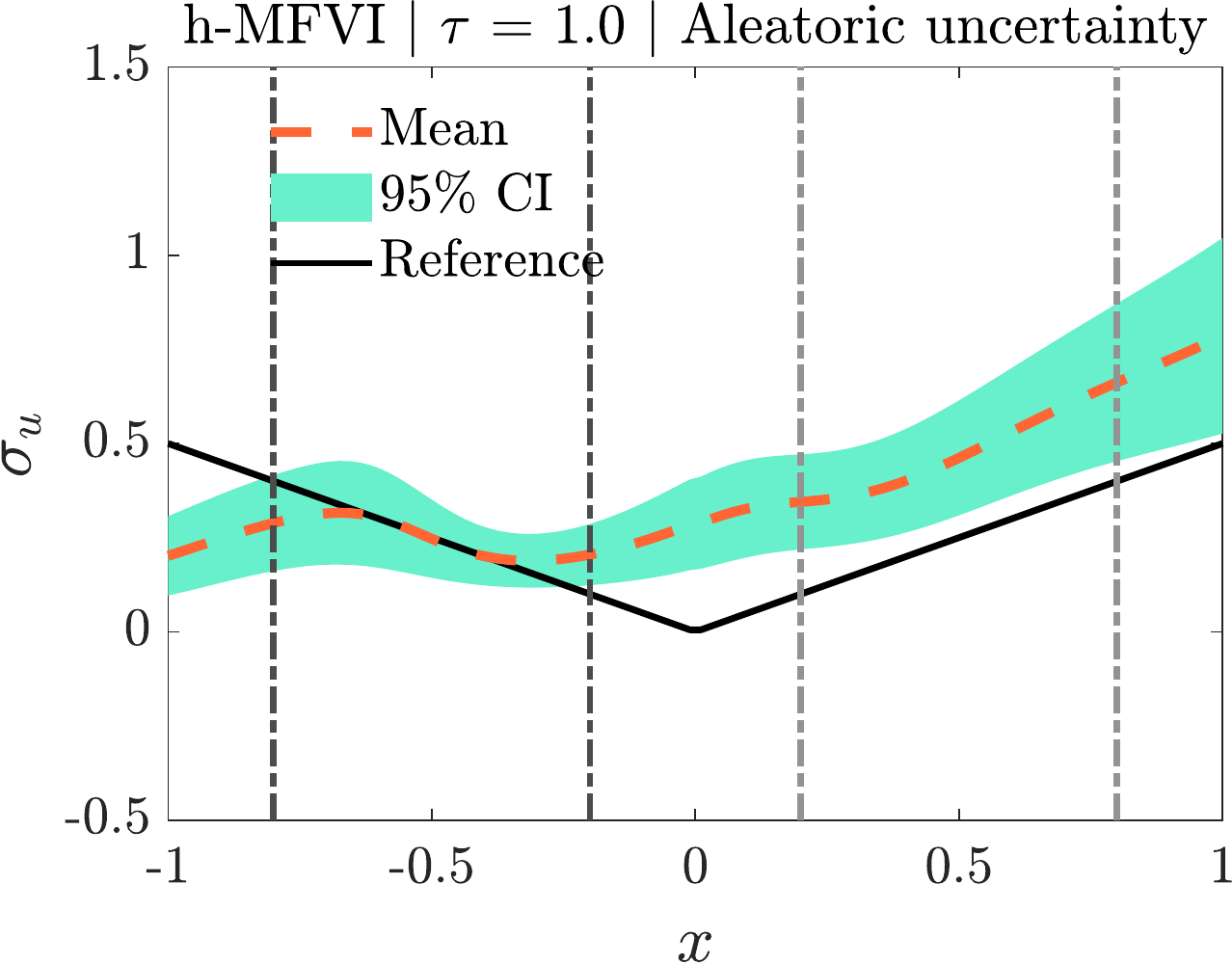}}
	\subcaptionbox{}{}{\includegraphics[width=0.32\textwidth]{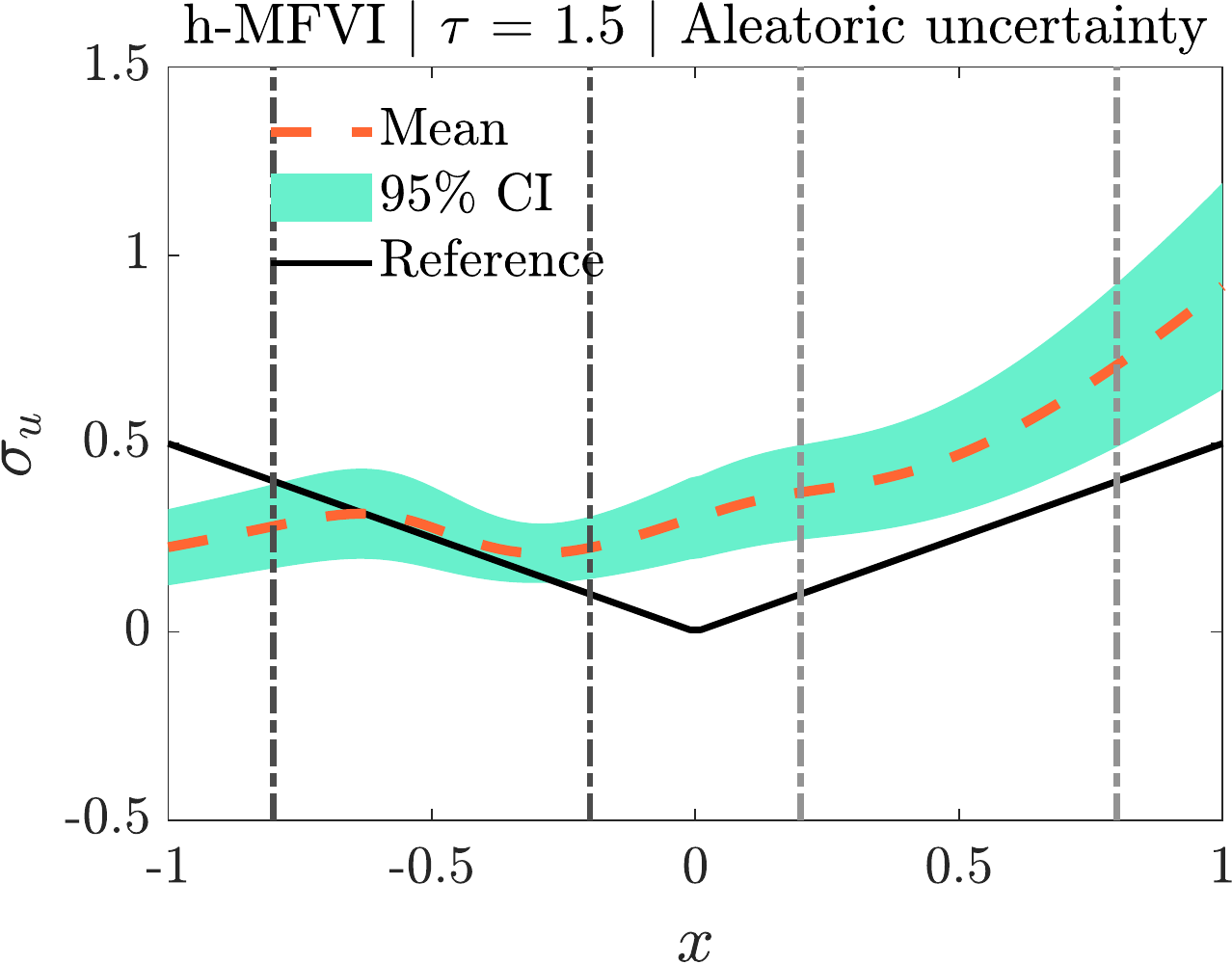}}
	\caption{
		Function approximation problem of Eq.~\eqref{eq:comp:func:func} | \textit{Unknown Student-t heteroscedastic noise}: by modifying the ``temperature'' $\tau$ we can control how conservative the predictions are.
		Here shown are h-MFVI predictions for three values of $\tau$ (see Section~\ref{app:modeling:postemp}).
		\textbf{Top row:} mean and total uncertainty of $u(x)$. 
		\textbf{Bottom row:}
		mean and epistemic uncertainty of predicted aleatoric uncertainty $\bar{\sigma}_u(x)$.
		In the top row the training datapoints are also included, while in the bottom row the black and gray dot-dashed lines mark the training data intervals.
	}
	\label{fig:comp:func:student:mfvi}
\end{figure}

\begin{figure}[!ht]
	\centering
	\subcaptionbox{}{}{\includegraphics[width=0.32\textwidth]{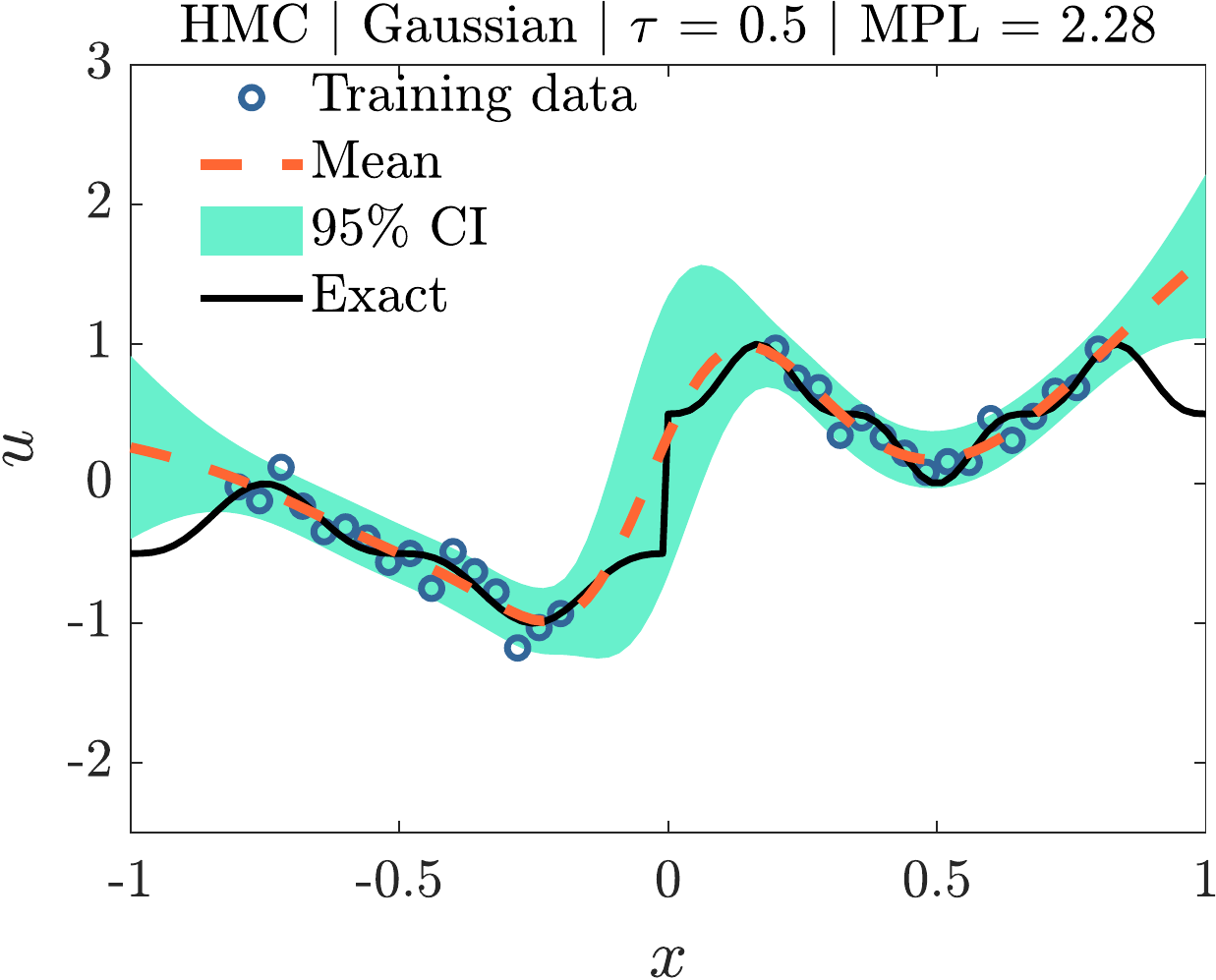}}
	\subcaptionbox{}{}{\includegraphics[width=0.32\textwidth]{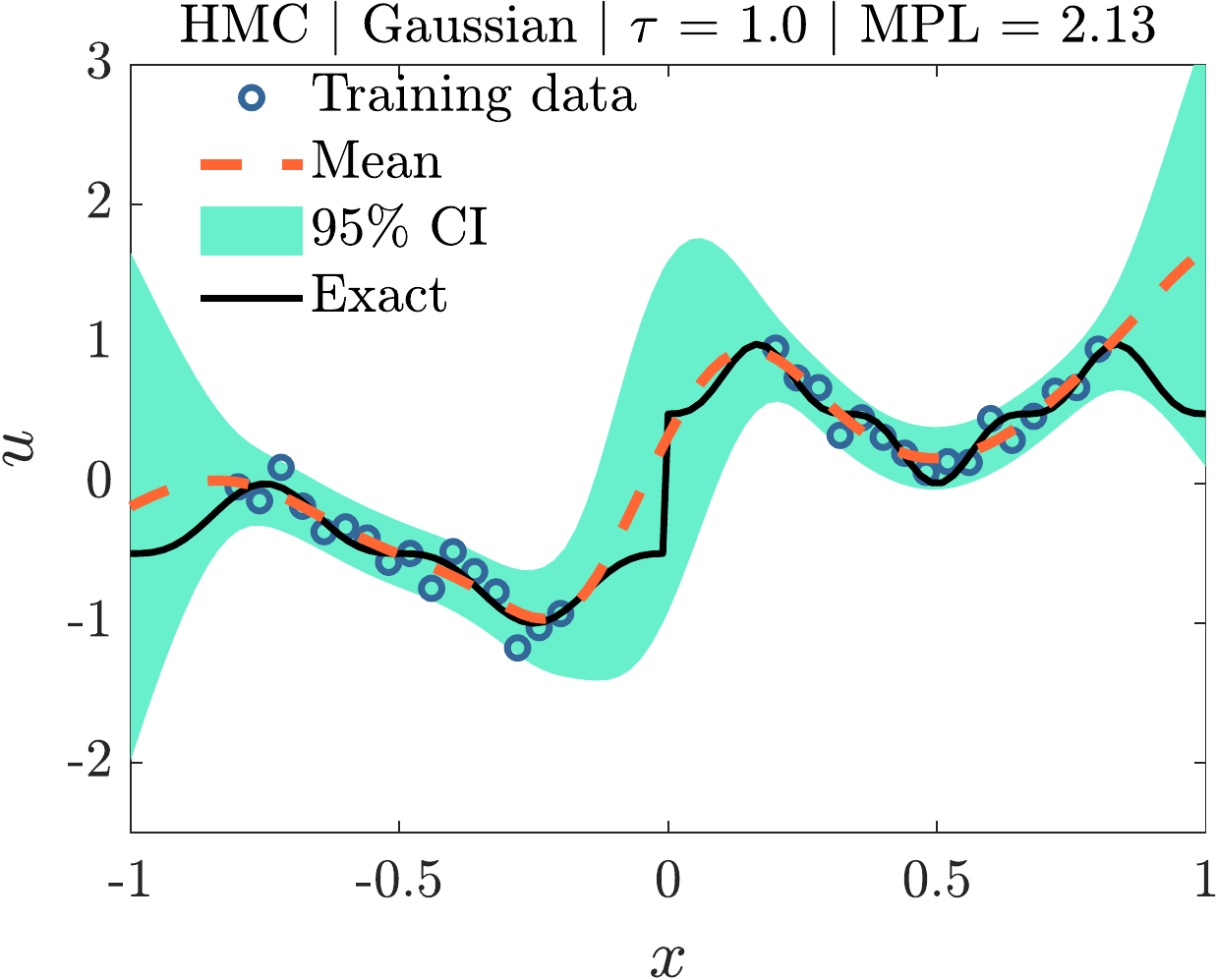}}
	\subcaptionbox{}{}{\includegraphics[width=0.32\textwidth]{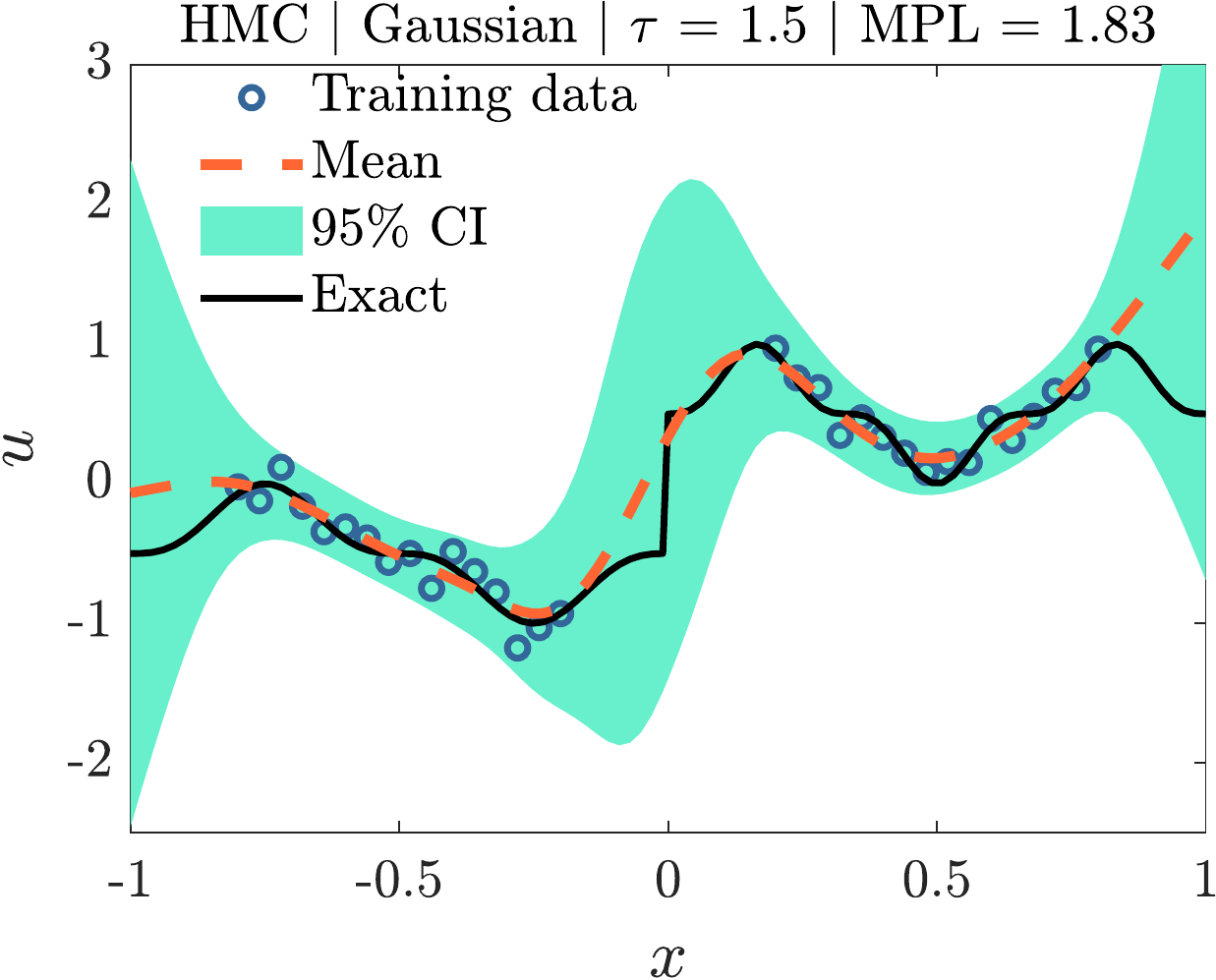}}
	\subcaptionbox{}{}{\includegraphics[width=0.32\textwidth]{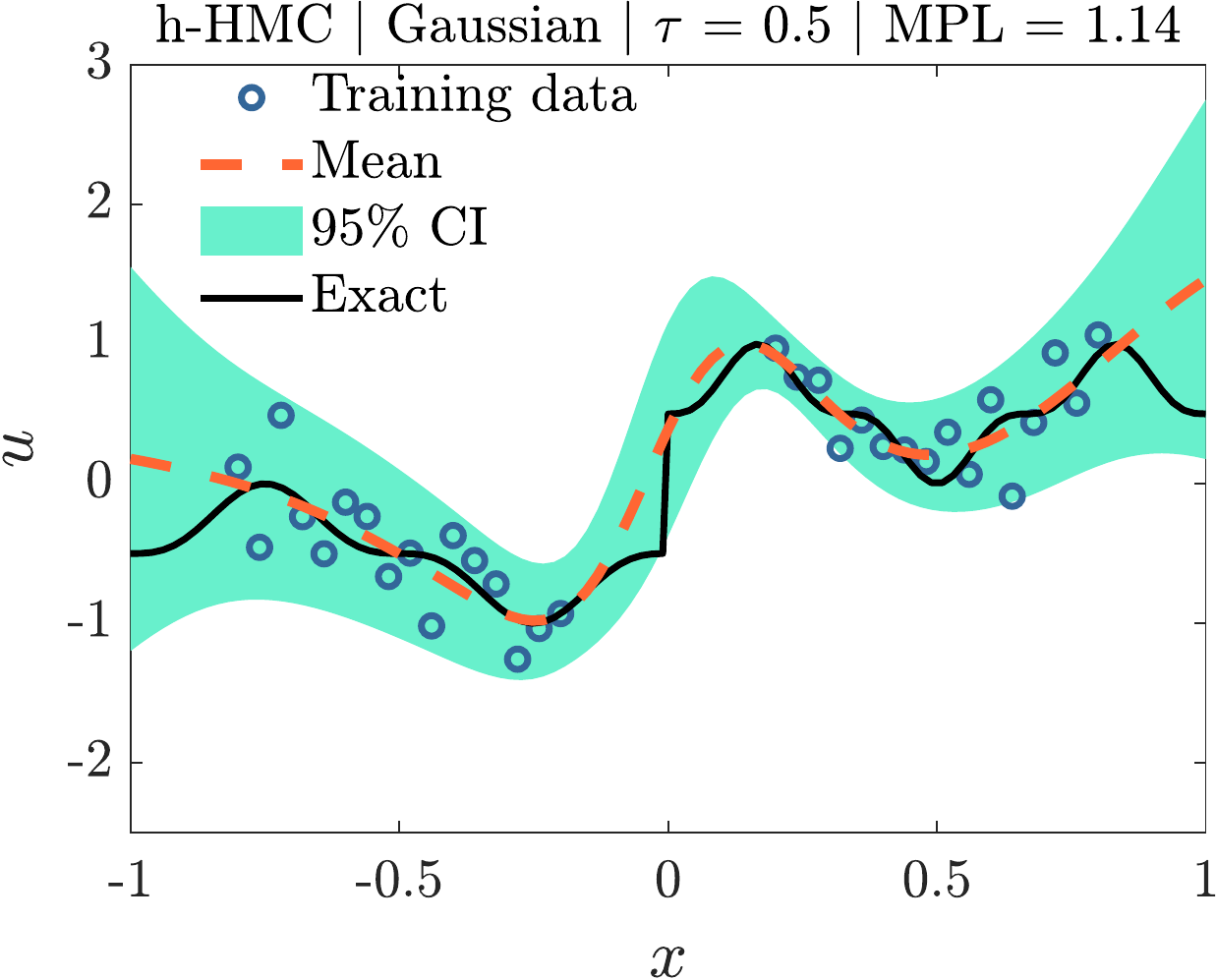}}
	\subcaptionbox{}{}{\includegraphics[width=0.32\textwidth]{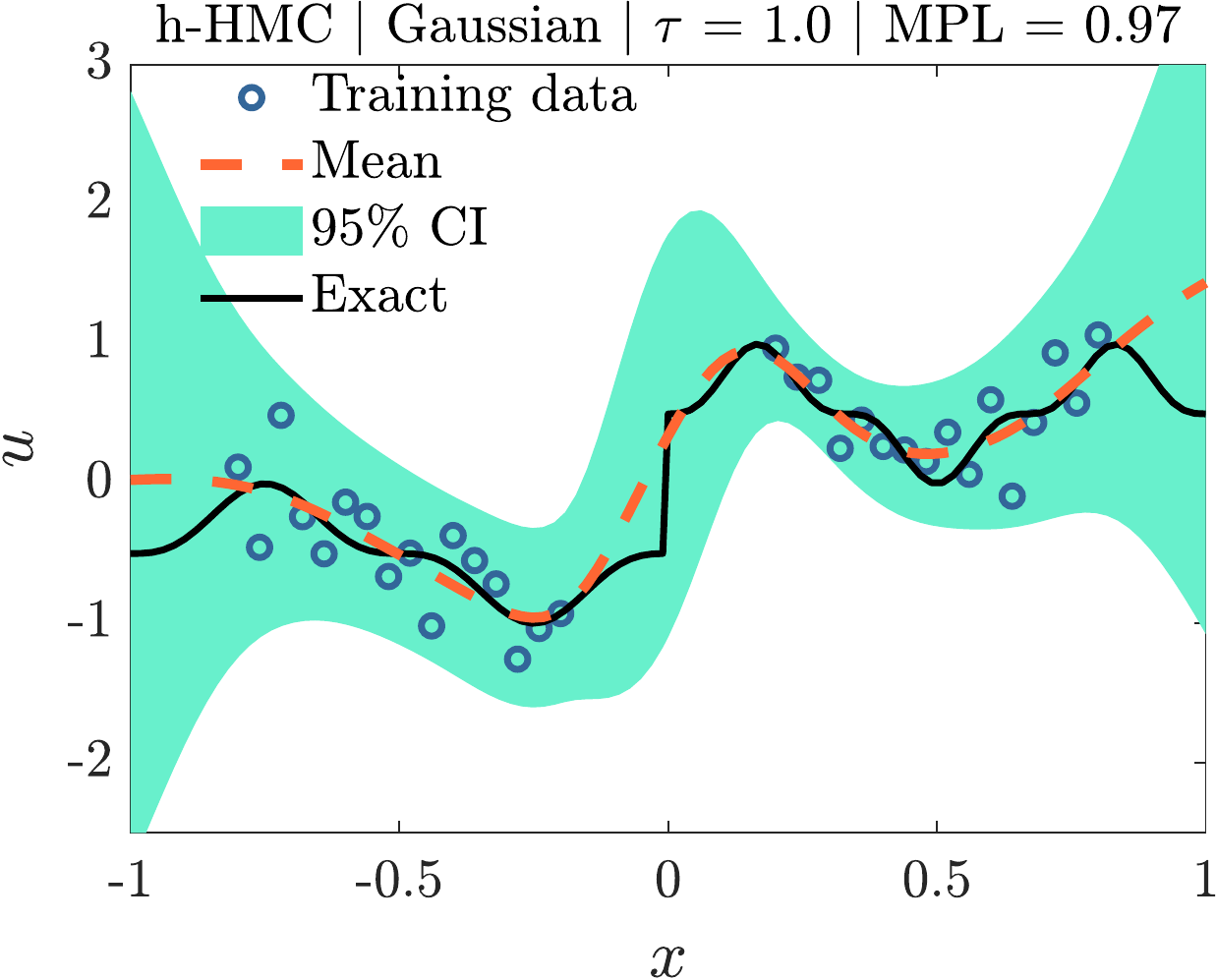}}
	\subcaptionbox{}{}{\includegraphics[width=0.32\textwidth]{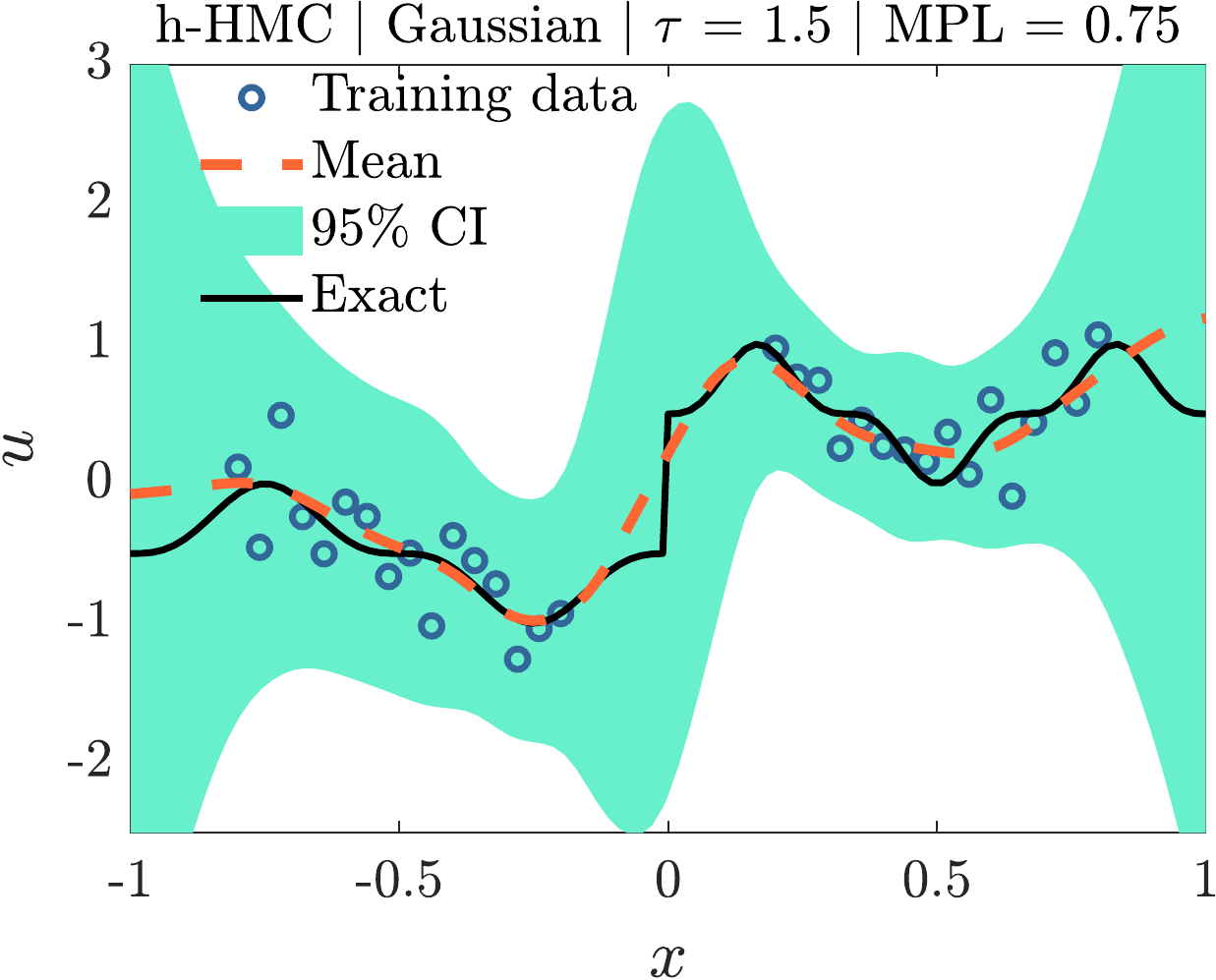}}
	\subcaptionbox{}{}{\includegraphics[width=0.32\textwidth]{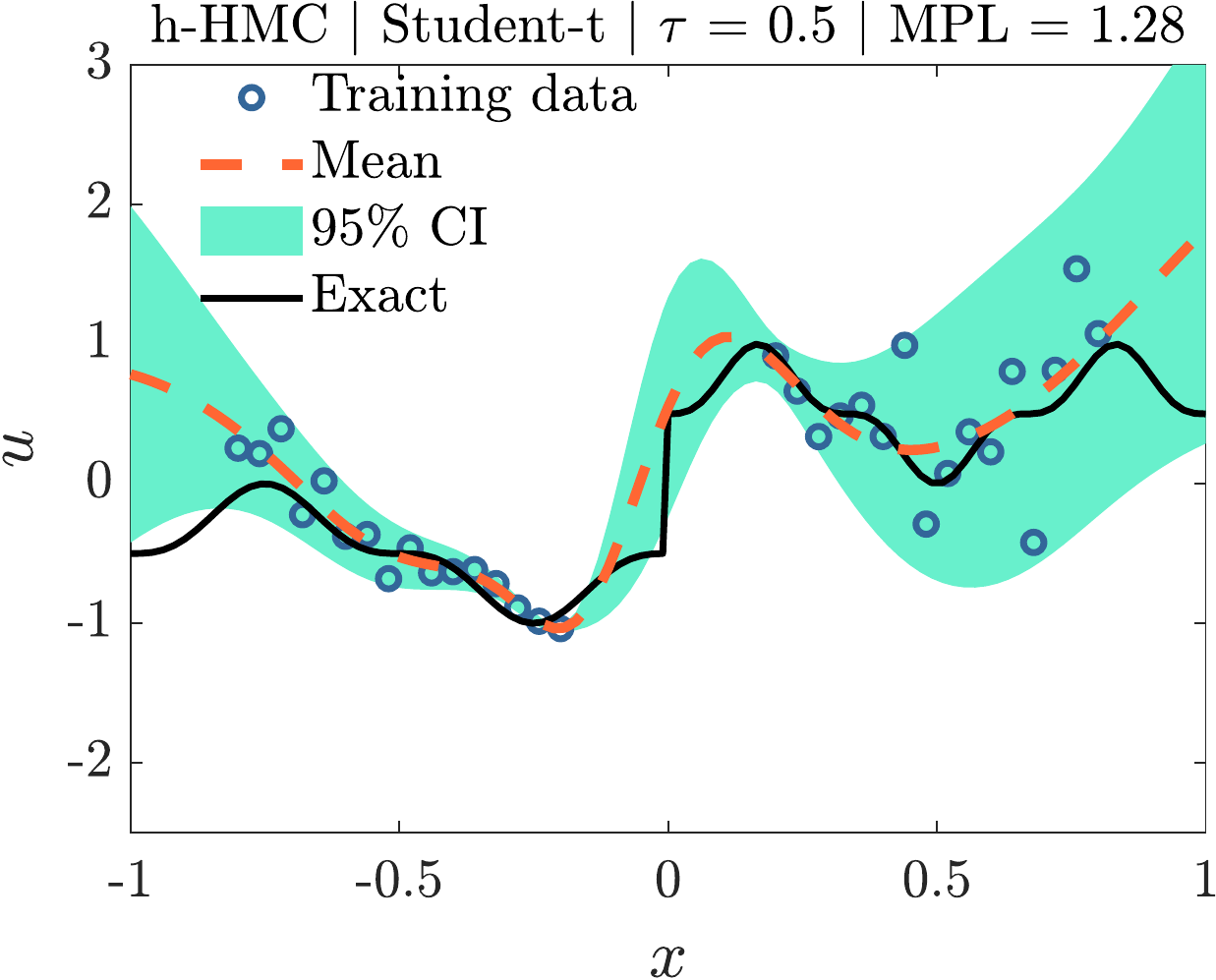}}
	\subcaptionbox{}{}{\includegraphics[width=0.32\textwidth]{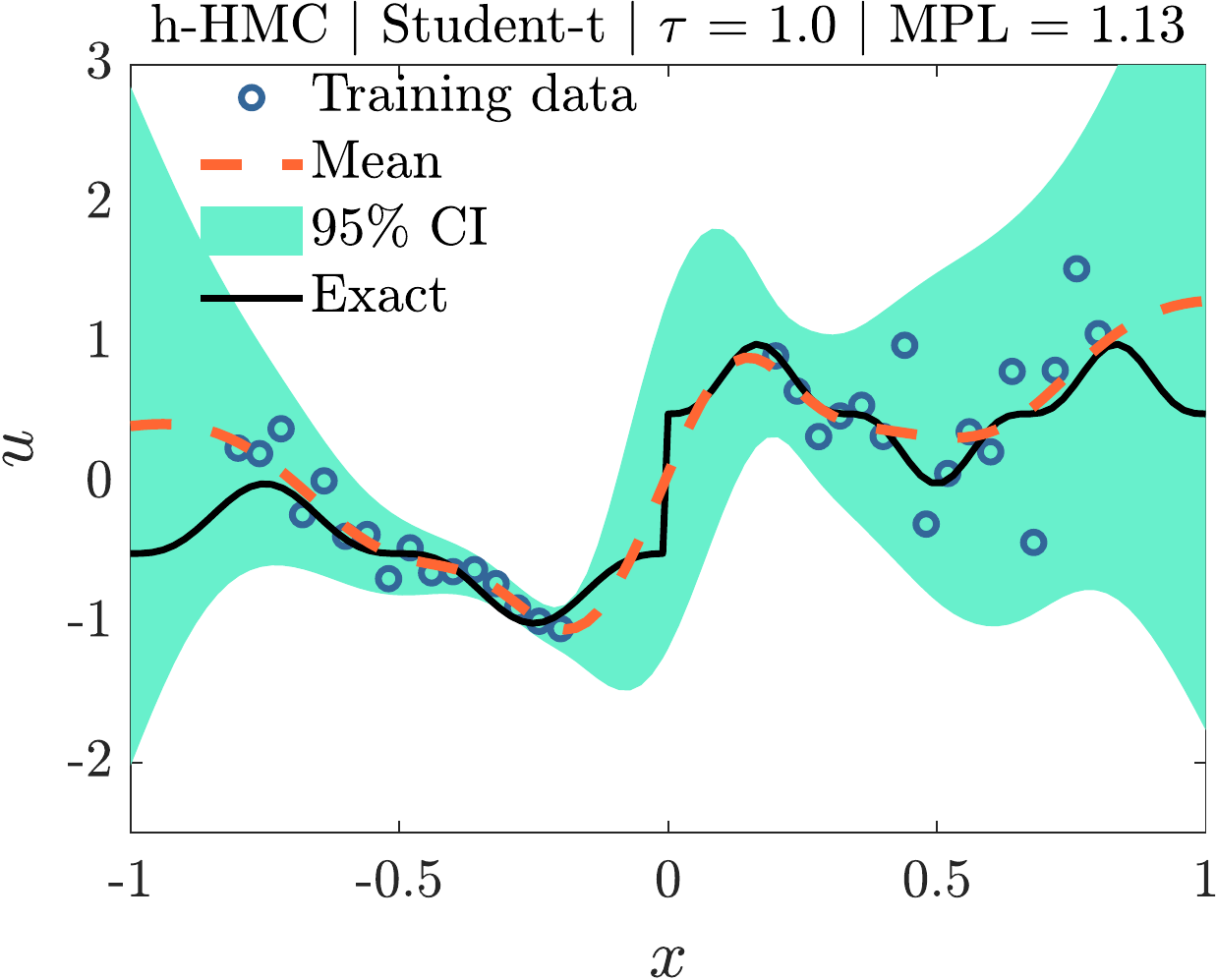}}
	\subcaptionbox{}{}{\includegraphics[width=0.32\textwidth]{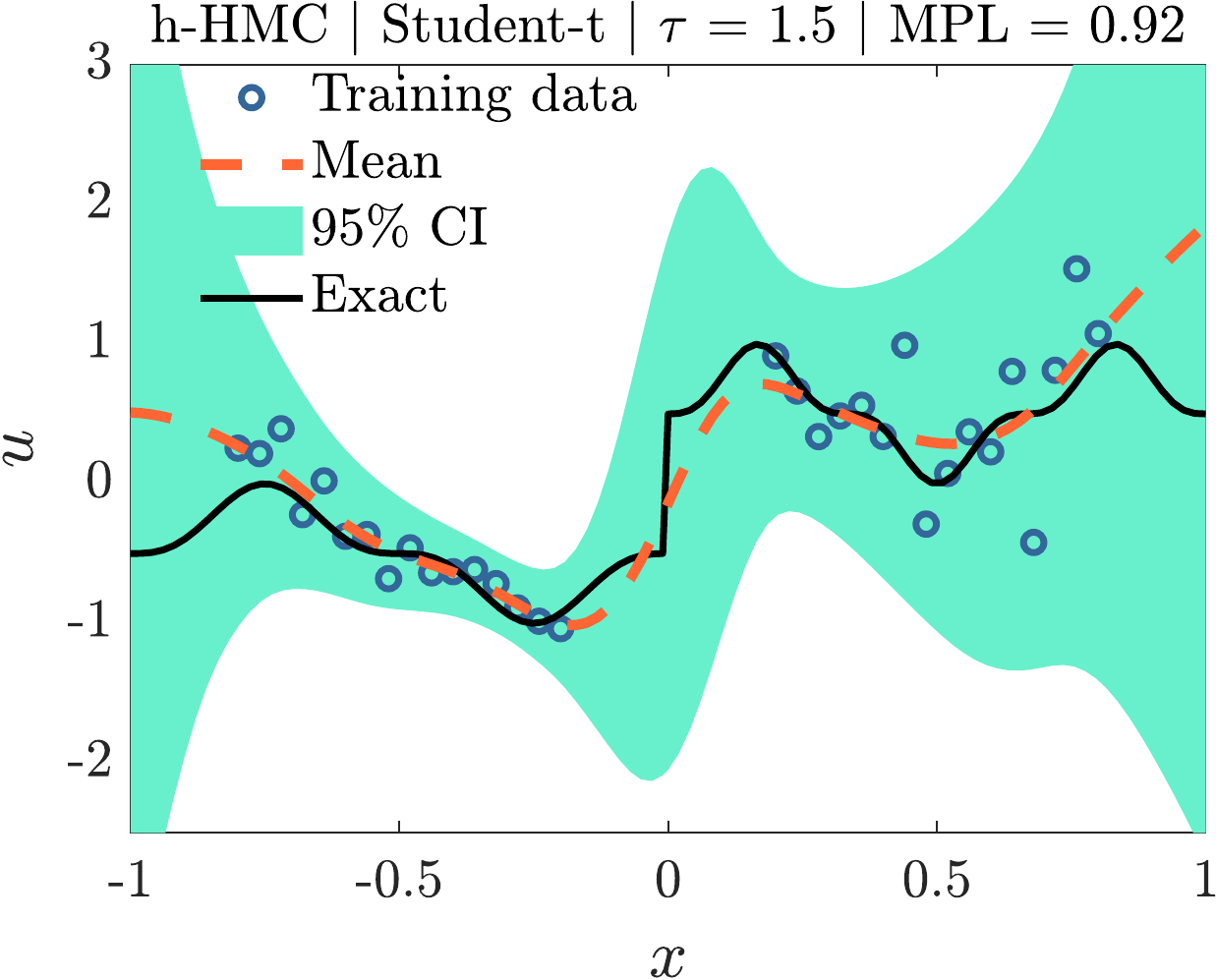}}
	\caption{
		Function approximation problem of Eq.~\eqref{eq:comp:func:func} | \textit{Gaussian or Student-t noise}: posterior tempering (Section~\ref{app:modeling:postemp}) can be construed as including one additional hyperparameter in the training procedure.
		Thus, it can be used both for cases where the assumed model is correct (a-c) and for model misspecification cases (d-i).
		Here shown are the mean and total uncertainty predictions of uncalibrated HMC for different posterior temperatures.
		\textbf{Top row:} known homoscedastic Gaussian noise.
		\textbf{Middle row:} unknown heteroscedastic Gaussian noise.
		\textbf{Bottom row:} unknown heteroscedastic Student-t noise.
	}
	\label{fig:comp:func:student:allcases}
\end{figure}

In Table~\ref{tab:comp:func:student} we evaluate h-HMC and h-MFVI with and without posterior tempering, based on the metrics of Section~\ref{sec:eval}.
In Figs.~\ref{fig:comp:func:student:hmc}-\ref{fig:comp:func:student:mfvi}, we provide the corresponding mean predictions and predicted aleatoric uncertainties.
By modifying the ``temperature'' $\tau$ we can control how conservative the predictions are.
For this case of model misspecification, a cold posterior performs better in terms of predictive capacity (MPL). 
Cold posteriors have been found to perform well also in other studies \cite{wenzel2020how,leimkuhler2020partitioned,zhang2019cyclical}.
In Fig.~\ref{fig:comp:func:student:allcases}, we employ posterior tempering also for Gaussian known homoscedastic and unknown heteroscedastic noise cases. 
In general, posterior tempering can be construed as including one additional hyperparameter in the training procedure.
Thus, it can be used both for cases where the assumed model is correct and for model misspecification cases.

%% file: appendix/IN_app_results_mixed_PINN.tex
\subsection{Additional results for the mixed PDE problem of Section~\ref{sec:comp:pinns}}\label{app:comp:pinns:results}

In this section, we present results pertaining to four more cases for the mixed PDE problem, i.e., (1) results from MFVI and MCD for the standard case of Section~\ref{sec:comp:pinns:stand}; (2) a ``large noise'' case in Section~\ref{app:comp:pinns:results:large} with a larger noise scale compared to the standard case; (3) an ``extrapolation'' case Section~\ref{app:comp:pinns:results:extra}, where the measurements of $\lambda(x)$ are concentrated in the interval $x \in [0, 1]$ and we extrapolate for $x \in [-1, 0]$; and (4) a ``steep boundary layers'' case in Section~\ref{app:comp:pinns:results:steep},  where the functions $u(t, x)$ and $\lambda(x)$ have large gradients for $x\approx -1$ and $x \approx 1$.

\subsubsection{Standard case: homoscedastic noise and uniform measurements in space and time}\label{app:comp:pinns:results:standard}

\begin{figure}[!ht]
	\centering
	\subcaptionbox{}{}{\includegraphics[width=0.32\textwidth]{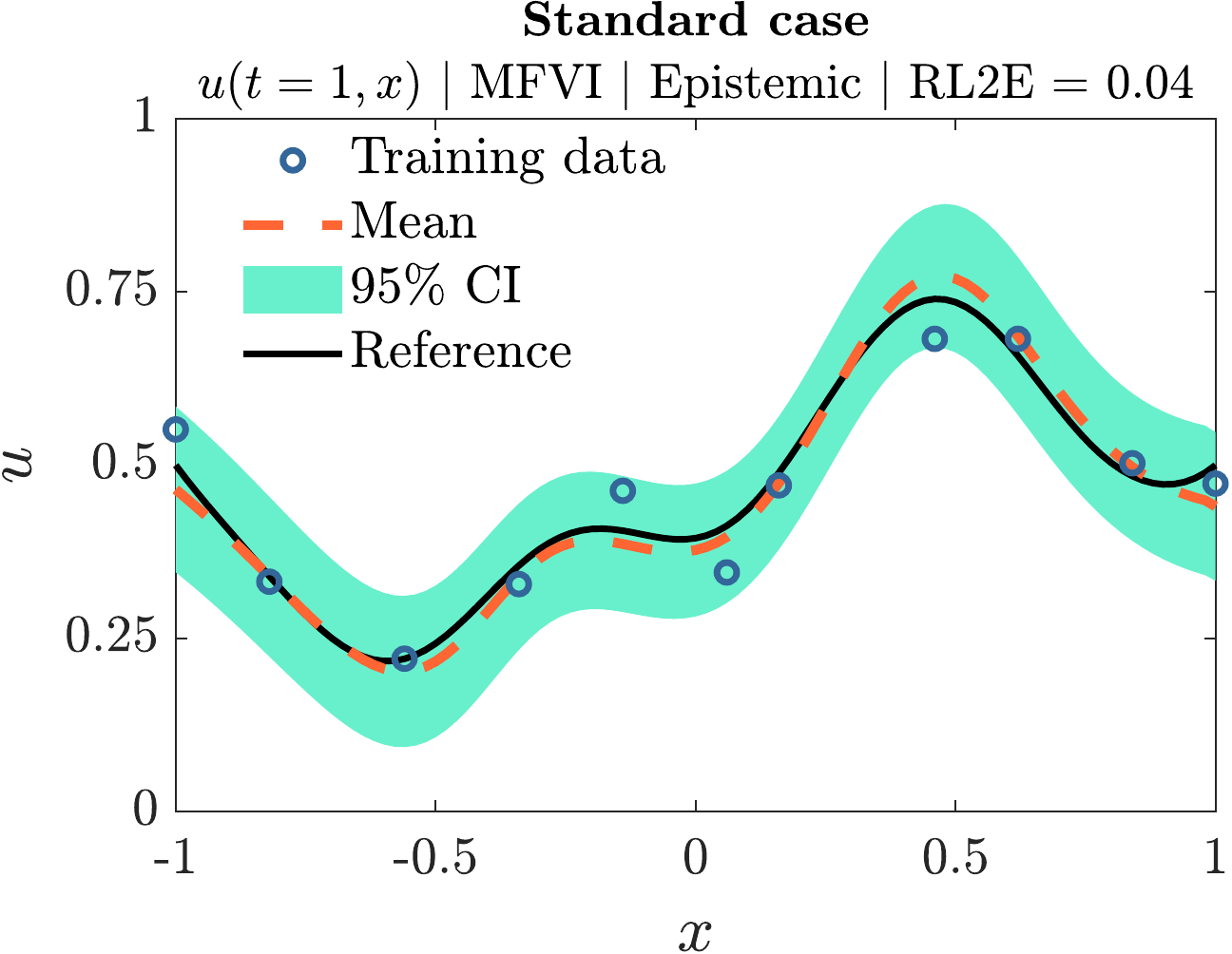}}
	\subcaptionbox{}{}{\includegraphics[width=0.32\textwidth]{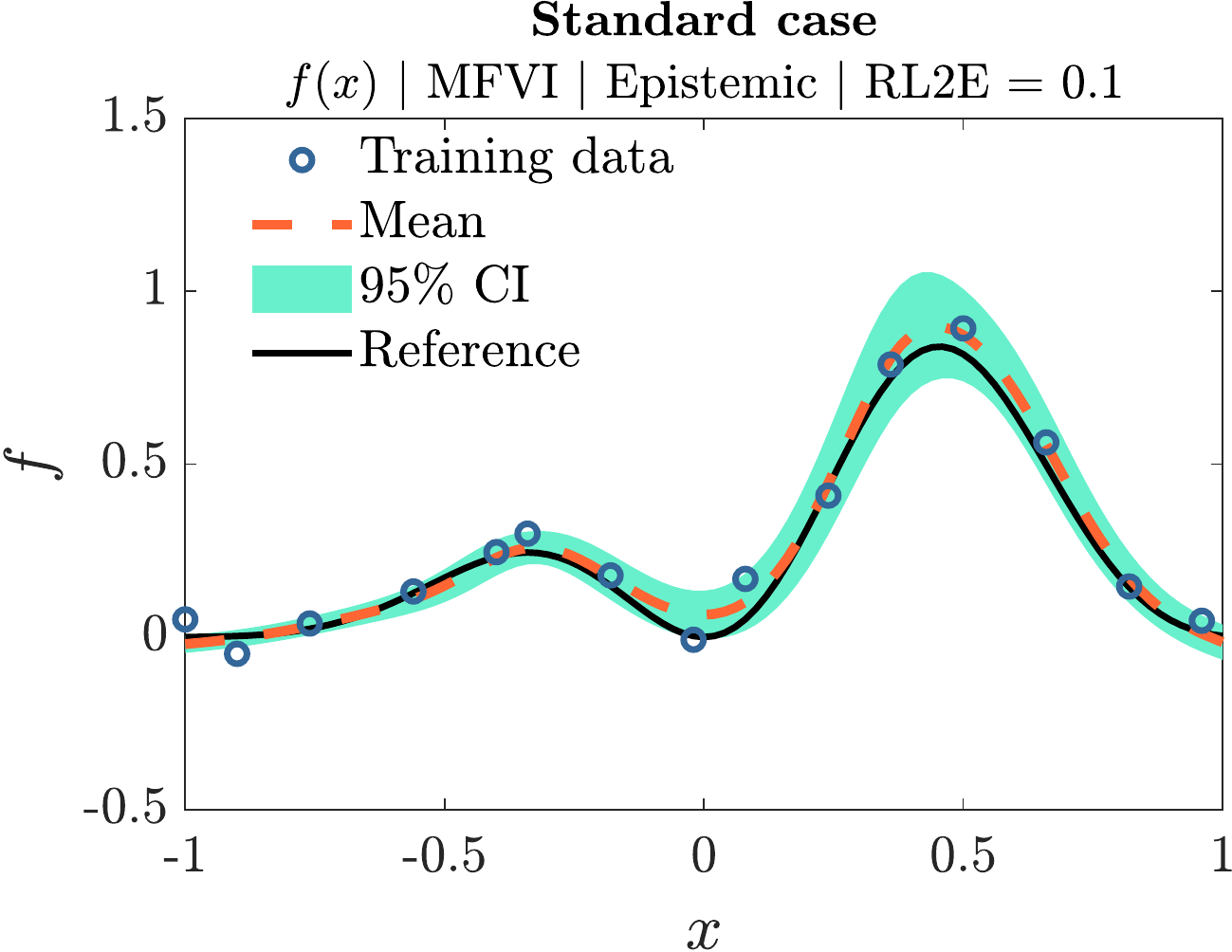}}
	\subcaptionbox{}{}{\includegraphics[width=0.32\textwidth]{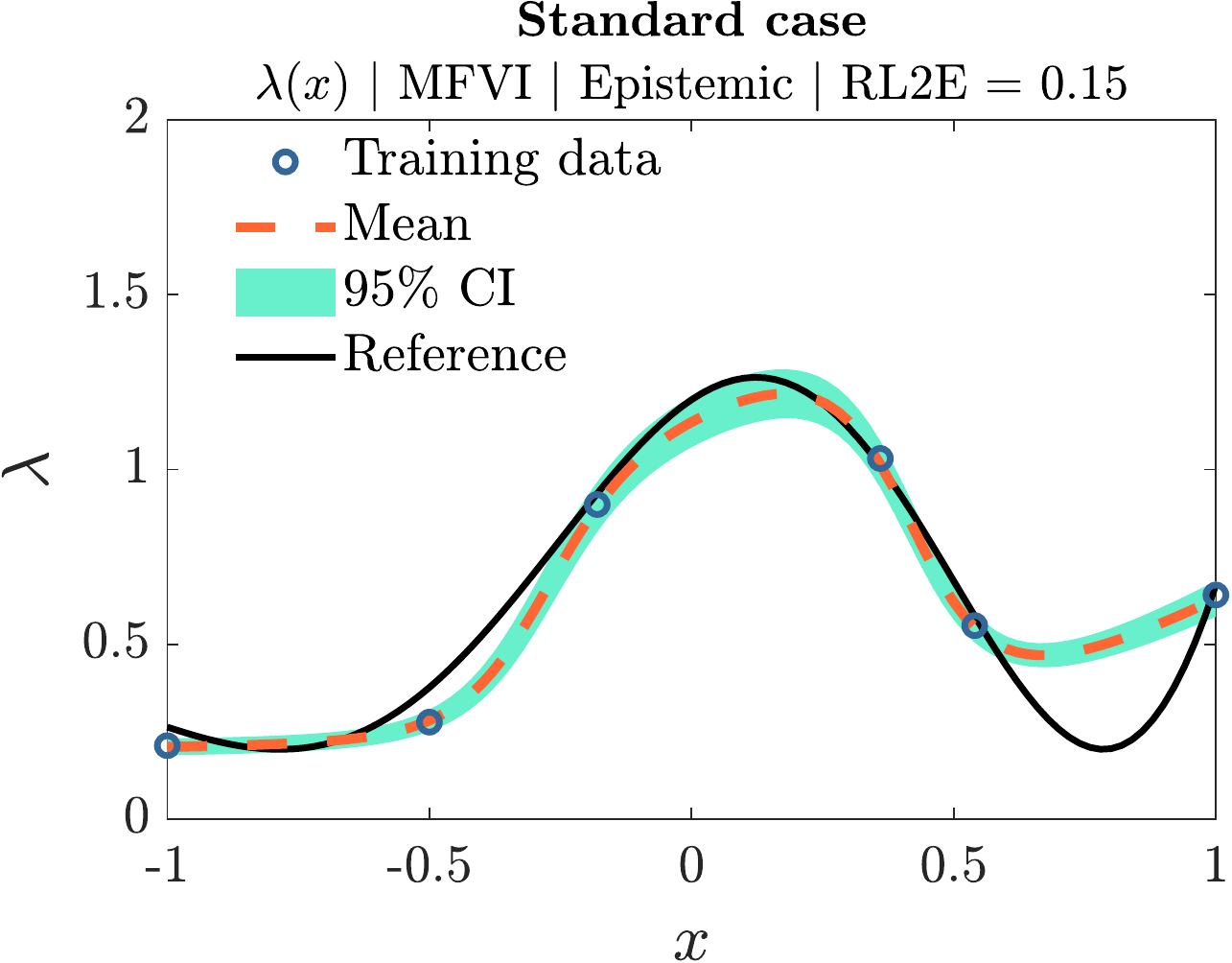}}
	\subcaptionbox{}{}{\includegraphics[width=0.32\textwidth]{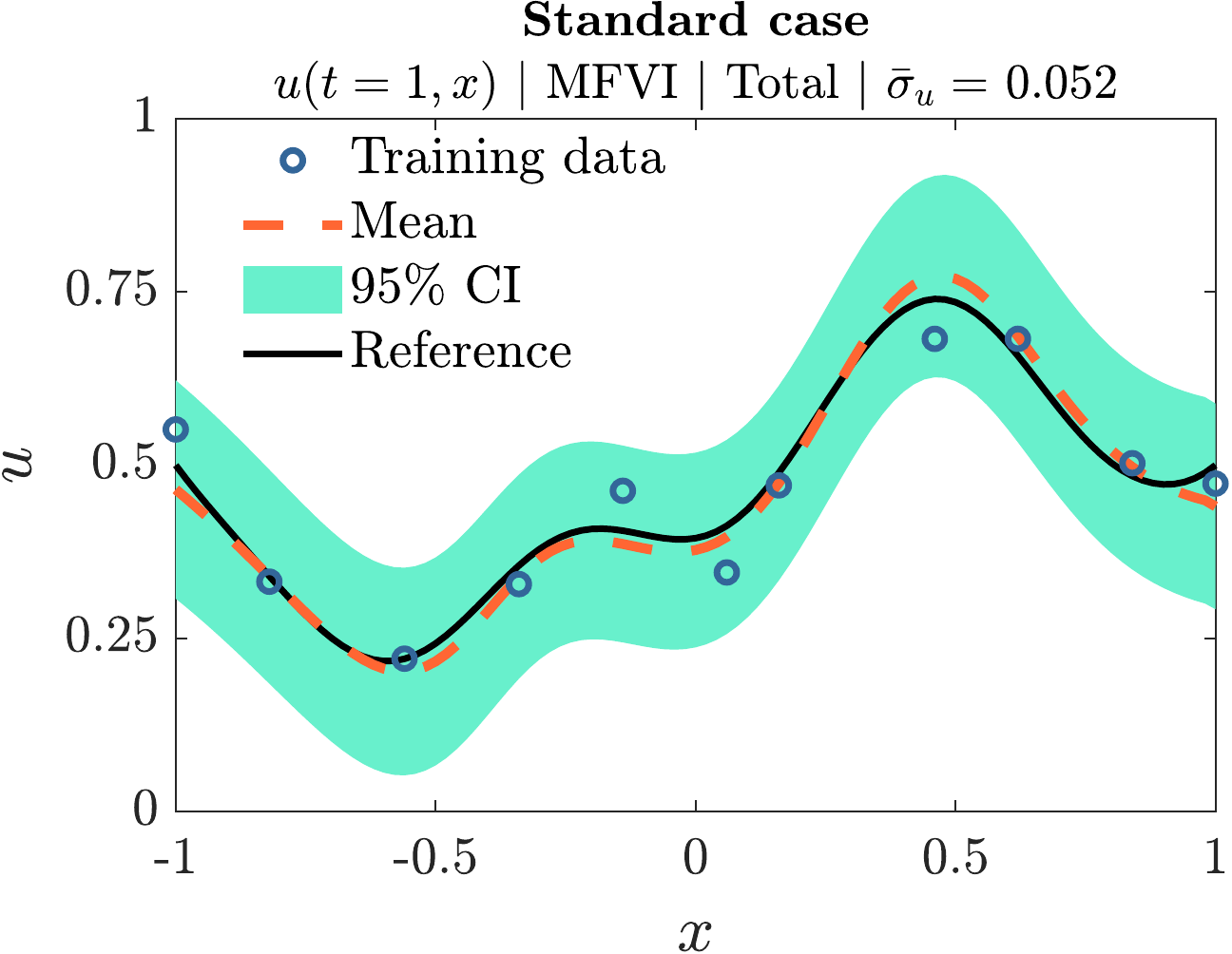}}
	\subcaptionbox{}{}{\includegraphics[width=0.32\textwidth]{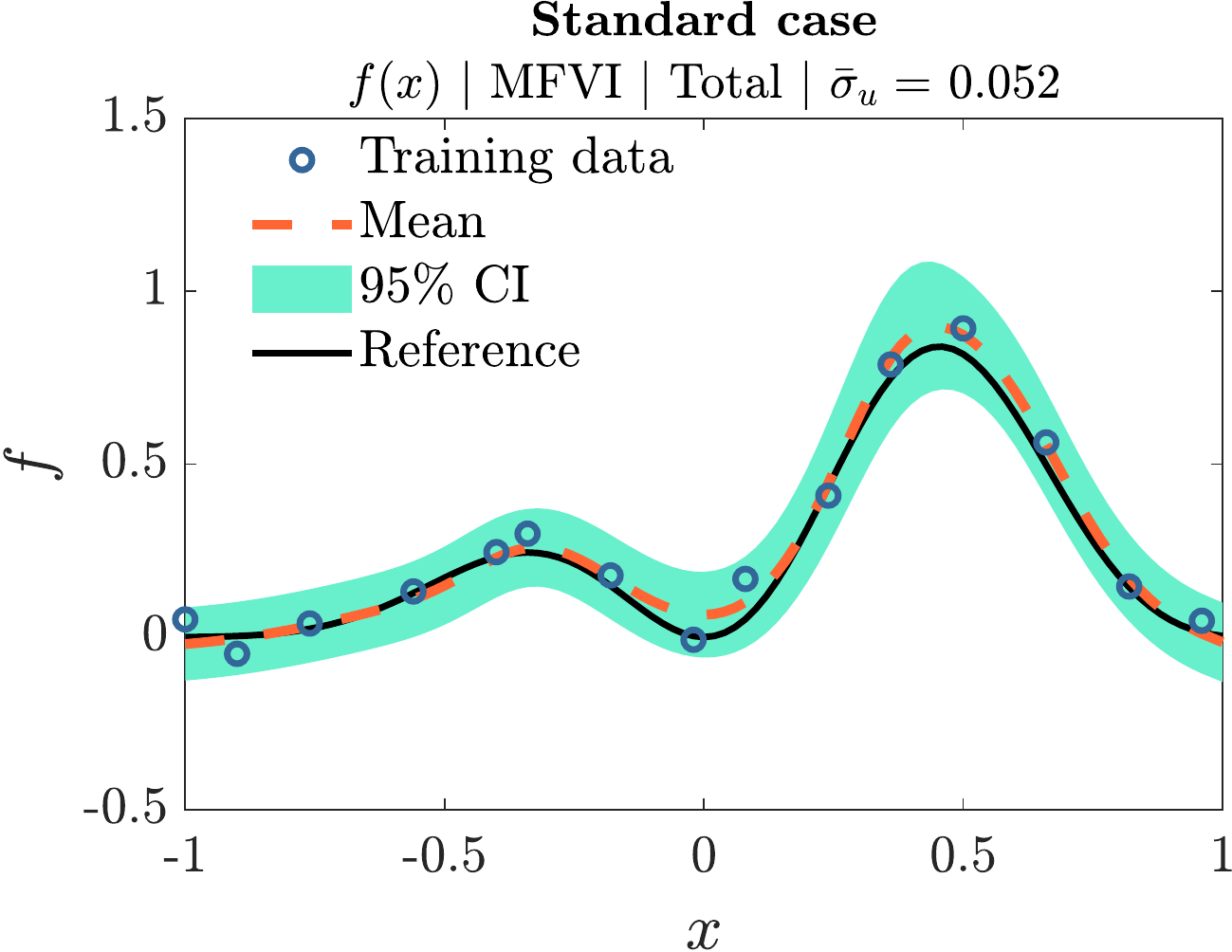}}
	\subcaptionbox{}{}{\includegraphics[width=0.32\textwidth]{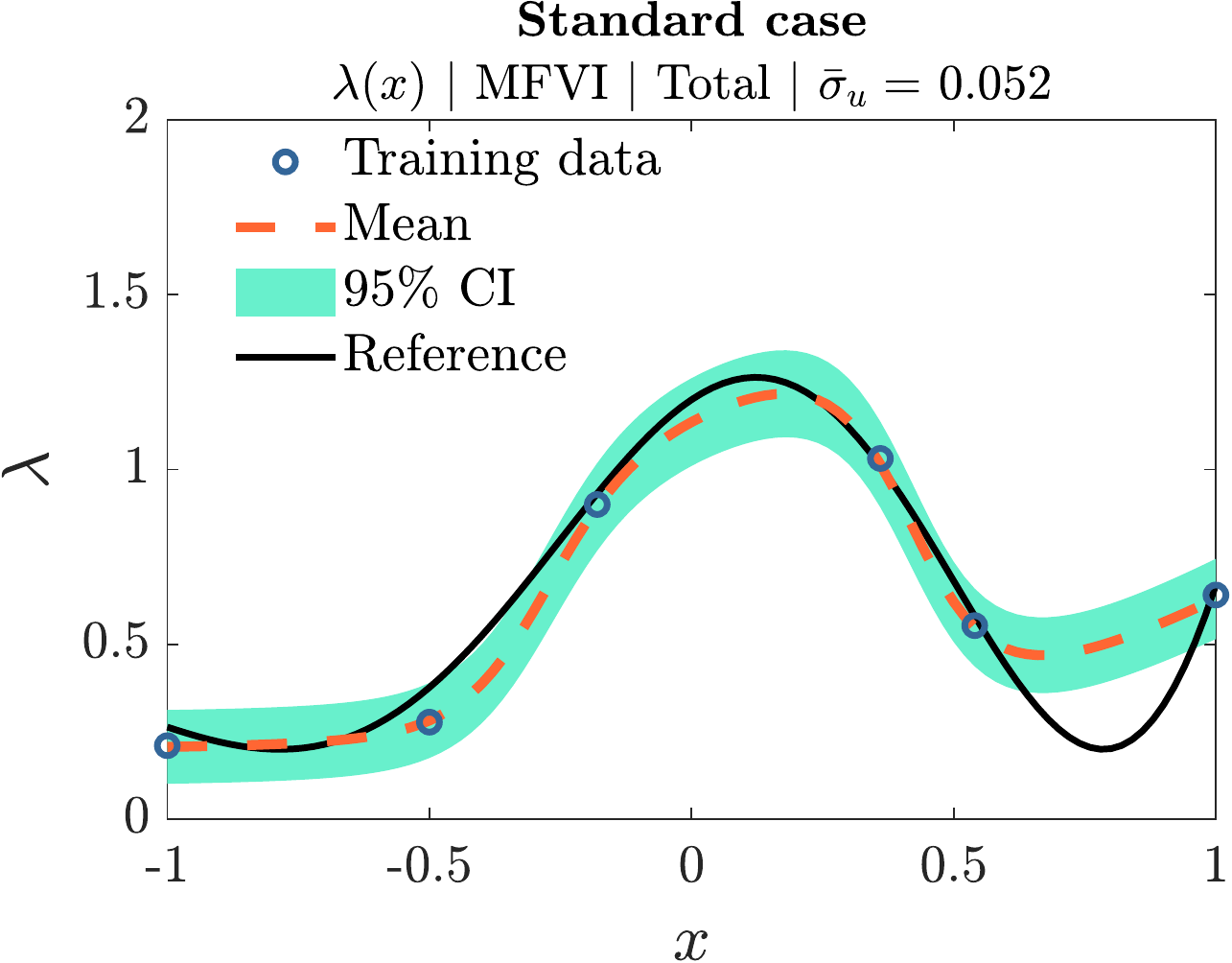}}
	\caption{
		Mixed PDE problem of Eq.~\eqref{eq:comp:pinns:stand} | \textit{Standard case}: training data, reference functions, as well as mean and uncertainty ($95\%$ CI) predictions of MFVI for $u$, $f$, and $\lambda$.
		\textbf{Top row:} epistemic uncertainty.
		\textbf{Bottom row:} total uncertainty including the \textit{learned} amount of aleatoric uncertainty, $\bar{\sigma}_u = 0.052$.
	}
	\label{fig:comp:pinns:stand:mfvi}
\end{figure}

\begin{figure}[!ht]
	\centering
	\subcaptionbox{}{}{\includegraphics[width=0.32\textwidth]{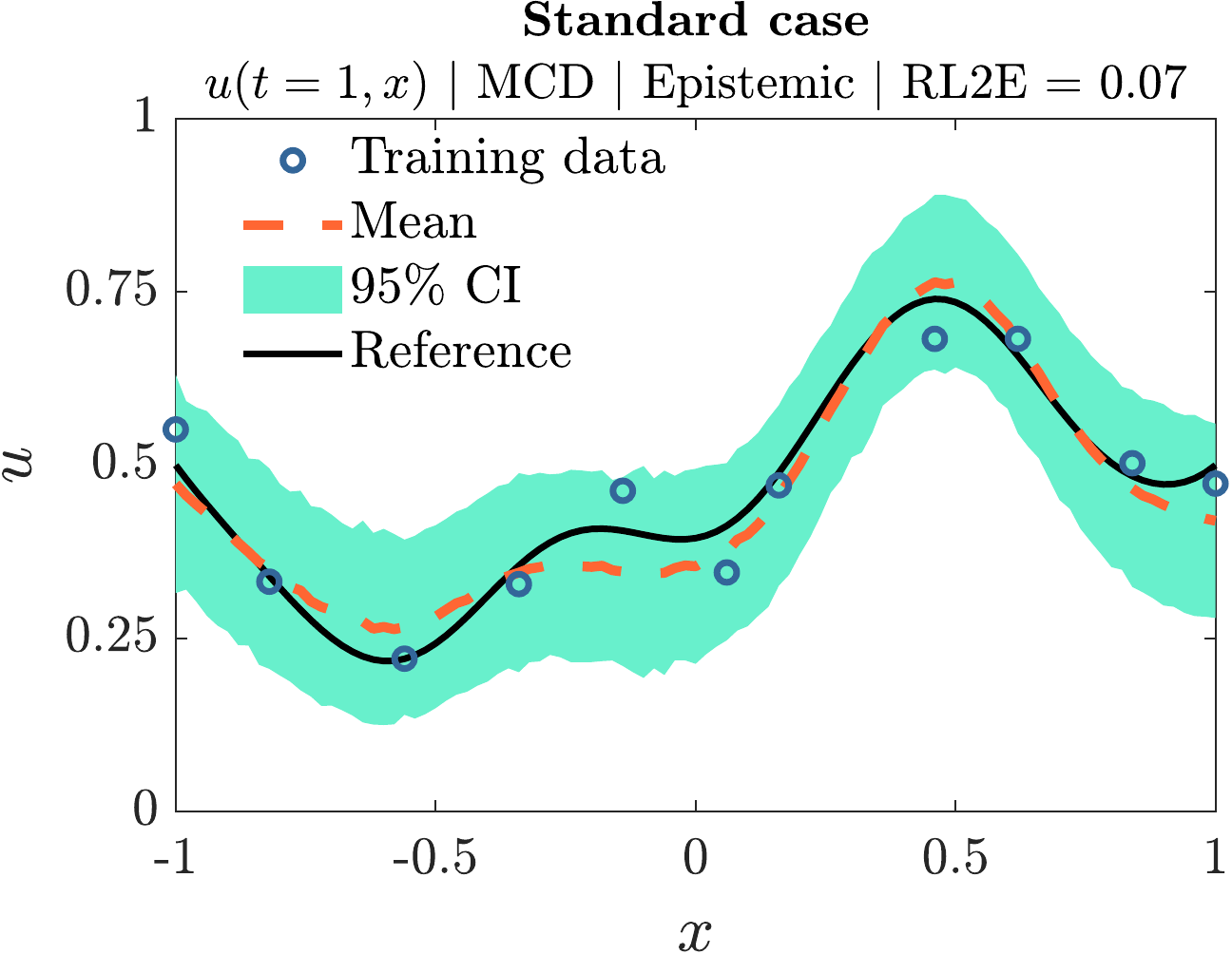}}
	\subcaptionbox{}{}{\includegraphics[width=0.32\textwidth]{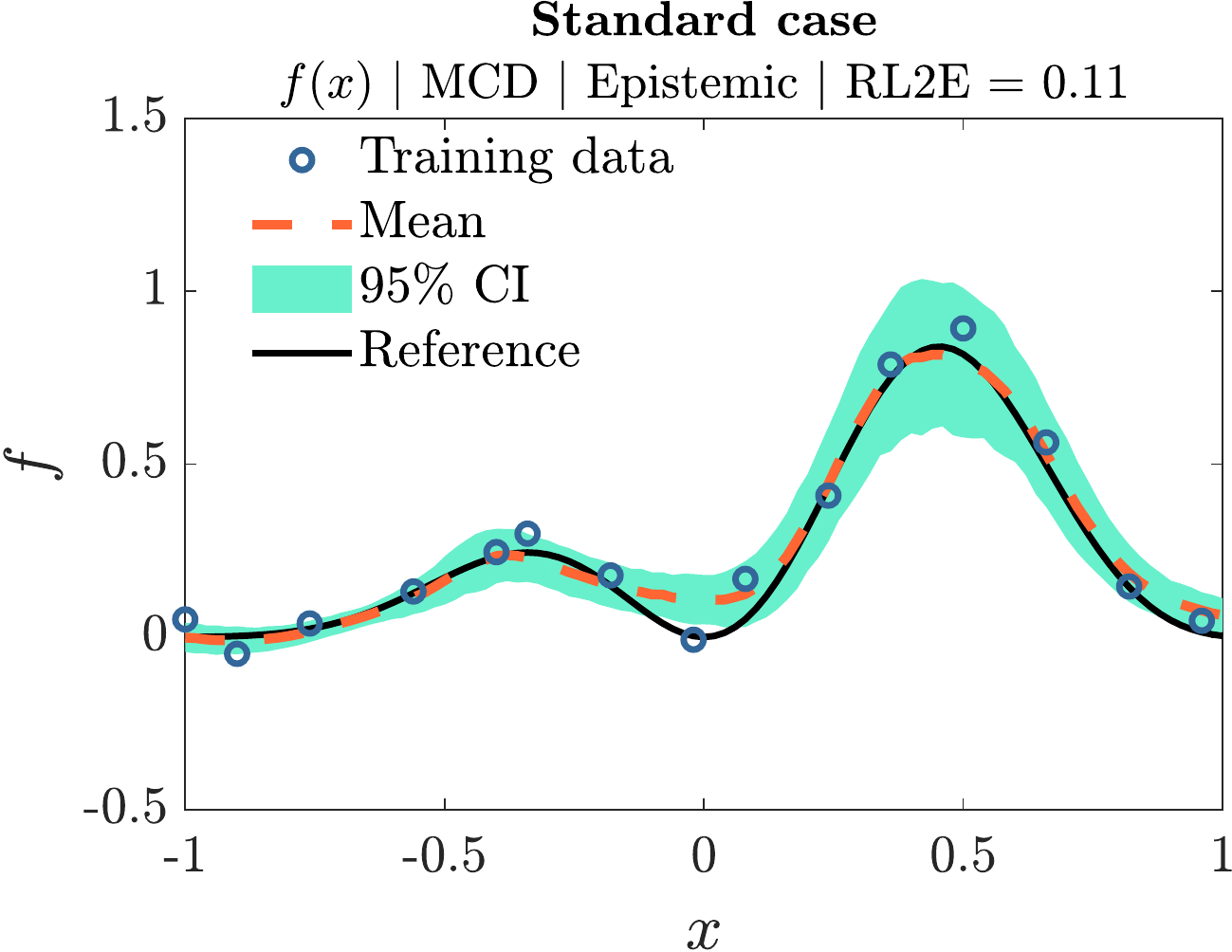}}
	\subcaptionbox{}{}{\includegraphics[width=0.32\textwidth]{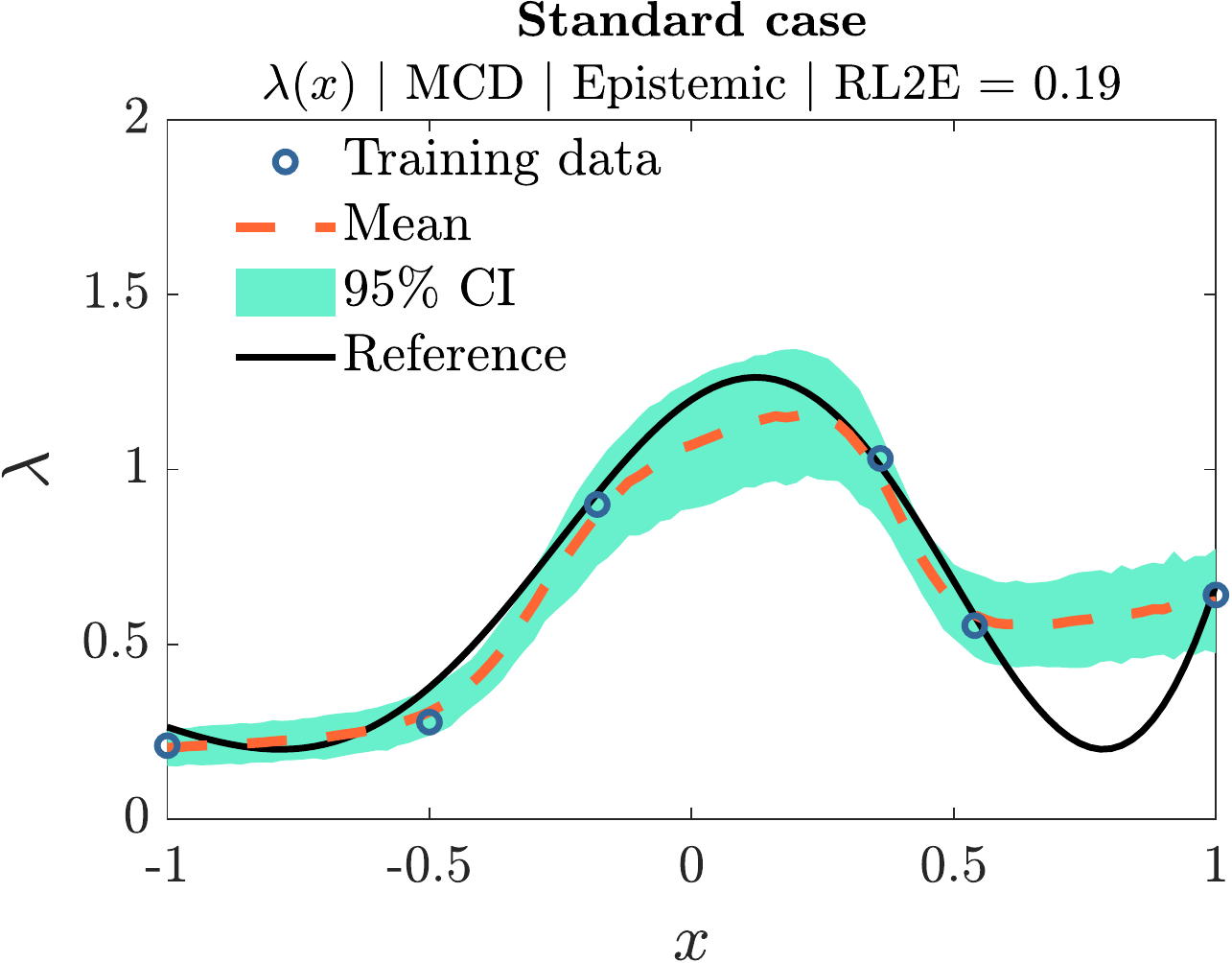}}
	\subcaptionbox{}{}{\includegraphics[width=0.32\textwidth]{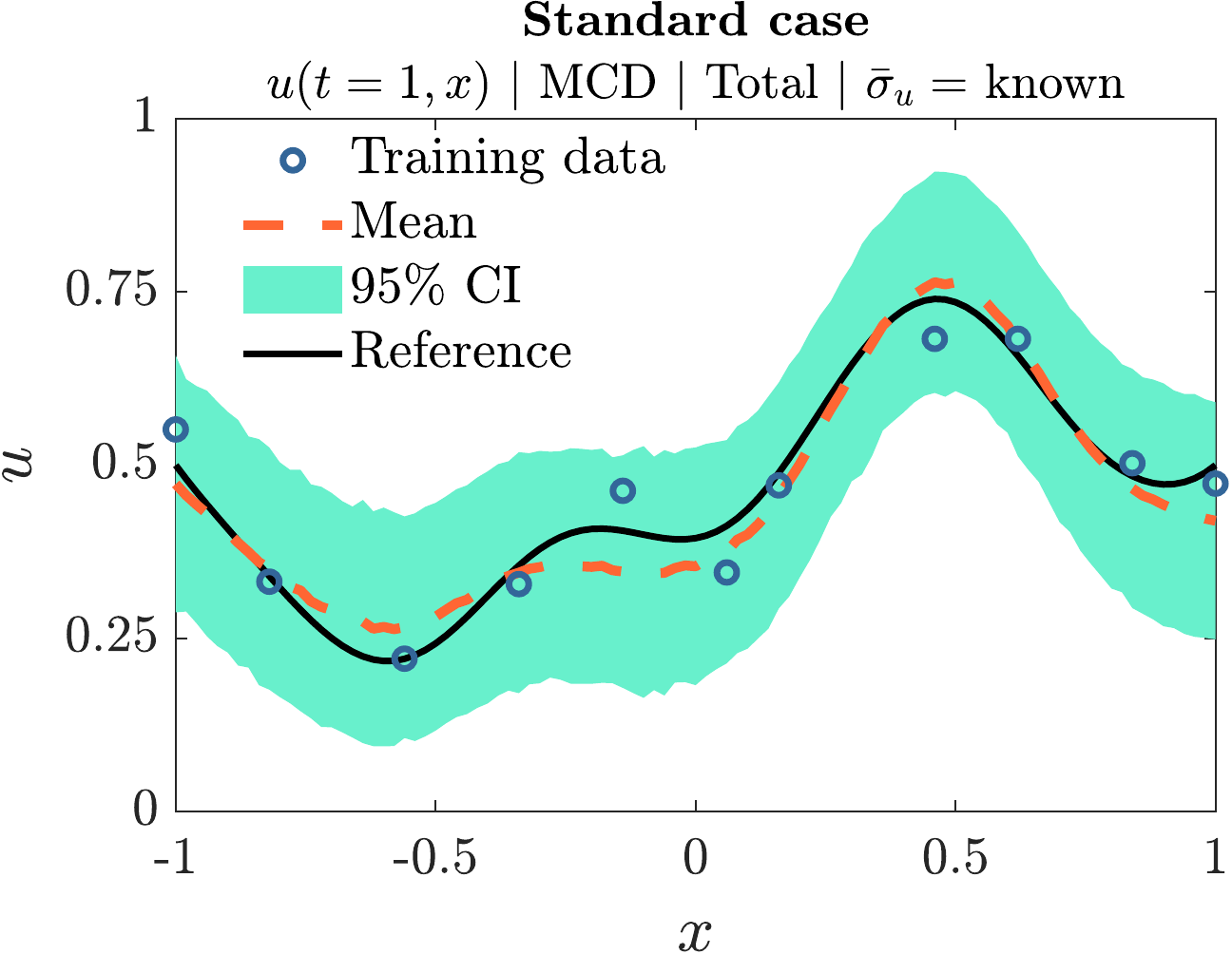}}
	\subcaptionbox{}{}{\includegraphics[width=0.32\textwidth]{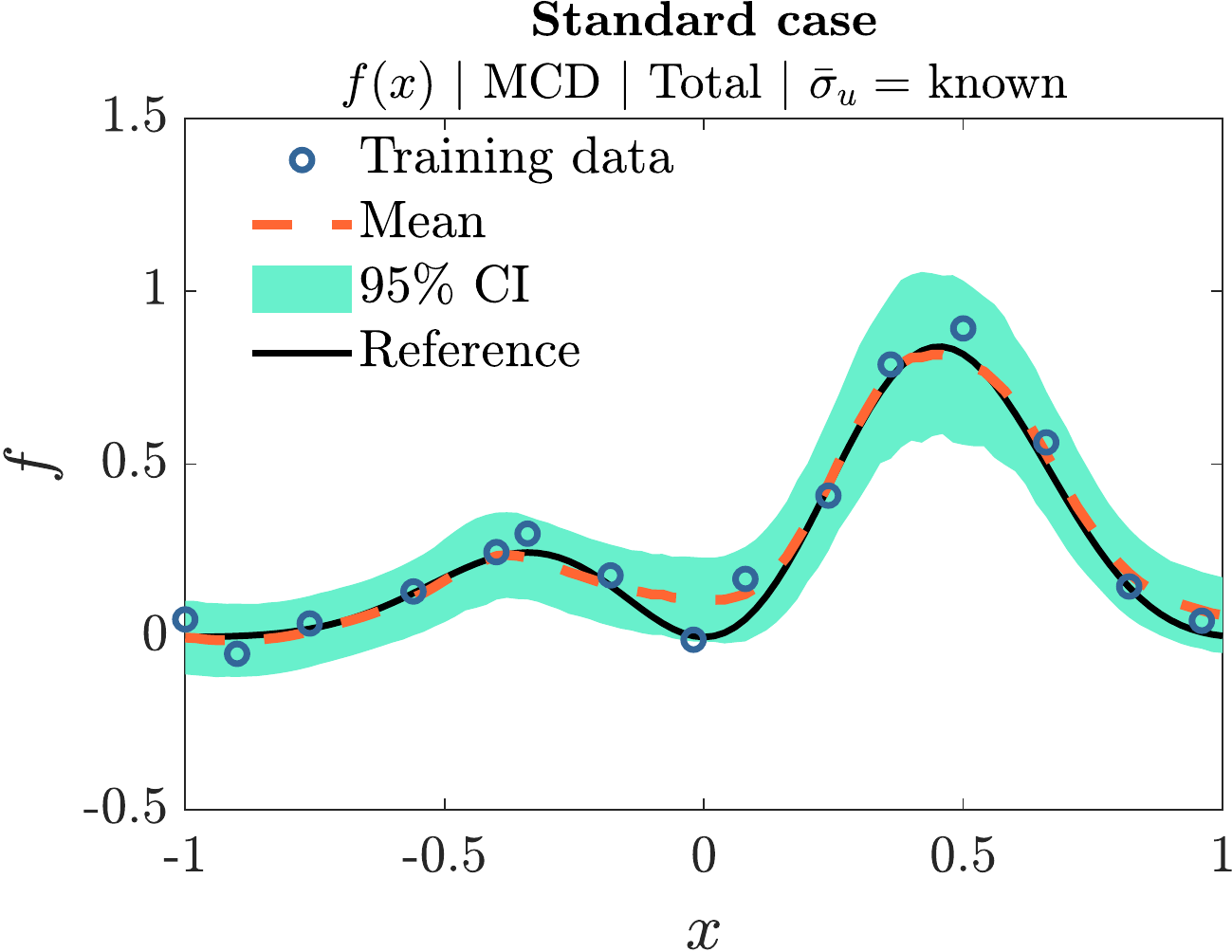}}
	\subcaptionbox{}{}{\includegraphics[width=0.32\textwidth]{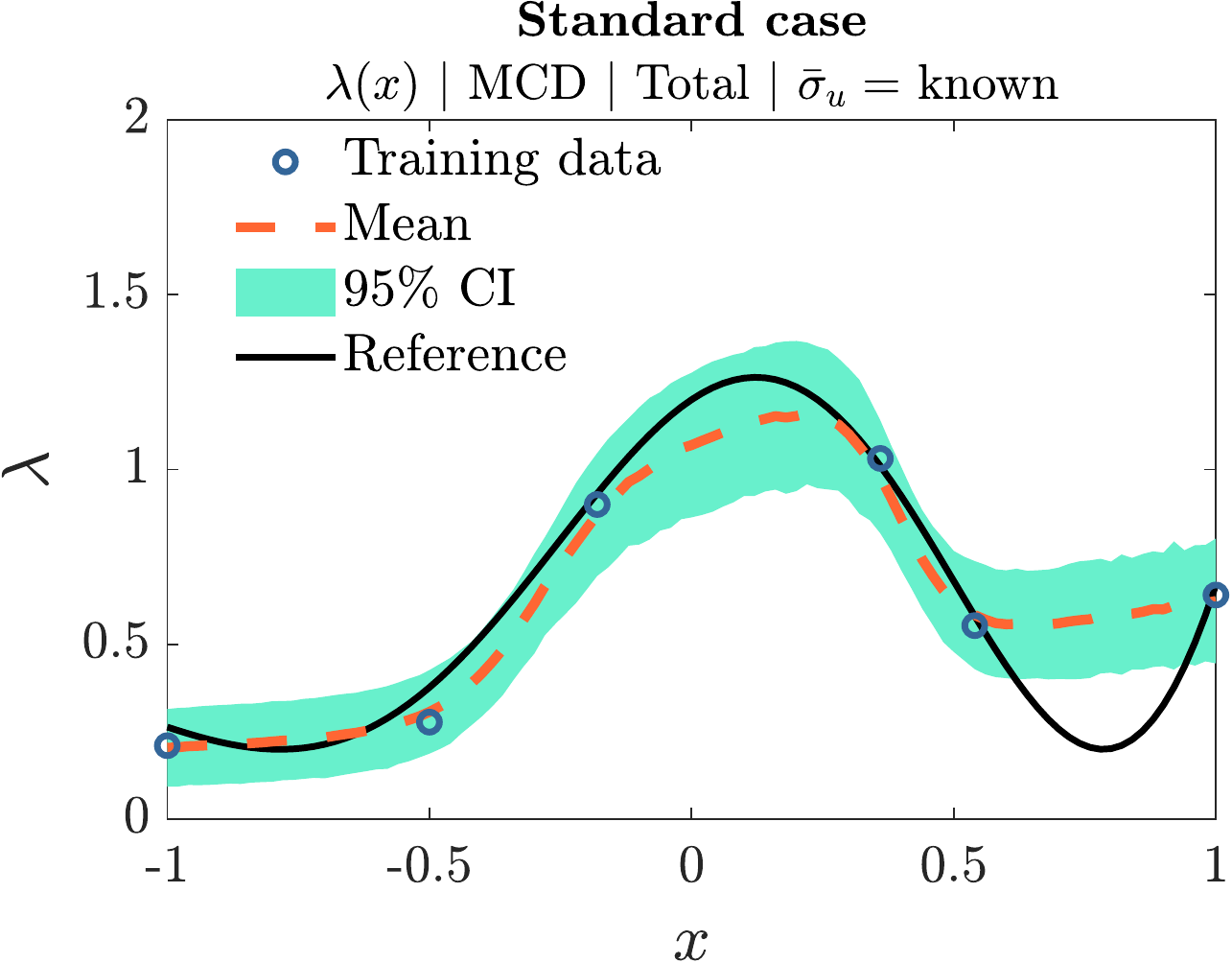}}
	\caption{
		Mixed PDE problem of Eq.~\eqref{eq:comp:pinns:stand} | \textit{Standard case}: training data, reference functions, as well as mean and uncertainty ($95\%$ CI) predictions of MCD for $u$, $f$, and $\lambda$.
		\textbf{Top row:} epistemic uncertainty.
		\textbf{Bottom row:} total uncertainty including the \textit{known} amount of aleatoric uncertainty, $\sigma_u = 0.05$.
	}
	\label{fig:comp:pinns:stand:mcd}
\end{figure}

The results obtained by MFVI (with learned noise) and MCD (with known noise) for the standard case of Section~\ref{sec:comp:pinns:stand} are provided in Figs. \ref{fig:comp:pinns:stand:mfvi} and \ref{fig:comp:pinns:stand:mcd}.
Although the predictions for $u$ and $f$ are satisfactory, the predictions for $\lambda$ are over-confident in some areas of the domain. 

\subsubsection{Large noise case: homoscedastic noise with large noise scale}\label{app:comp:pinns:results:large}

\begin{figure}[!ht]
	\centering
	\subcaptionbox{}{}{\includegraphics[width=0.32\textwidth]{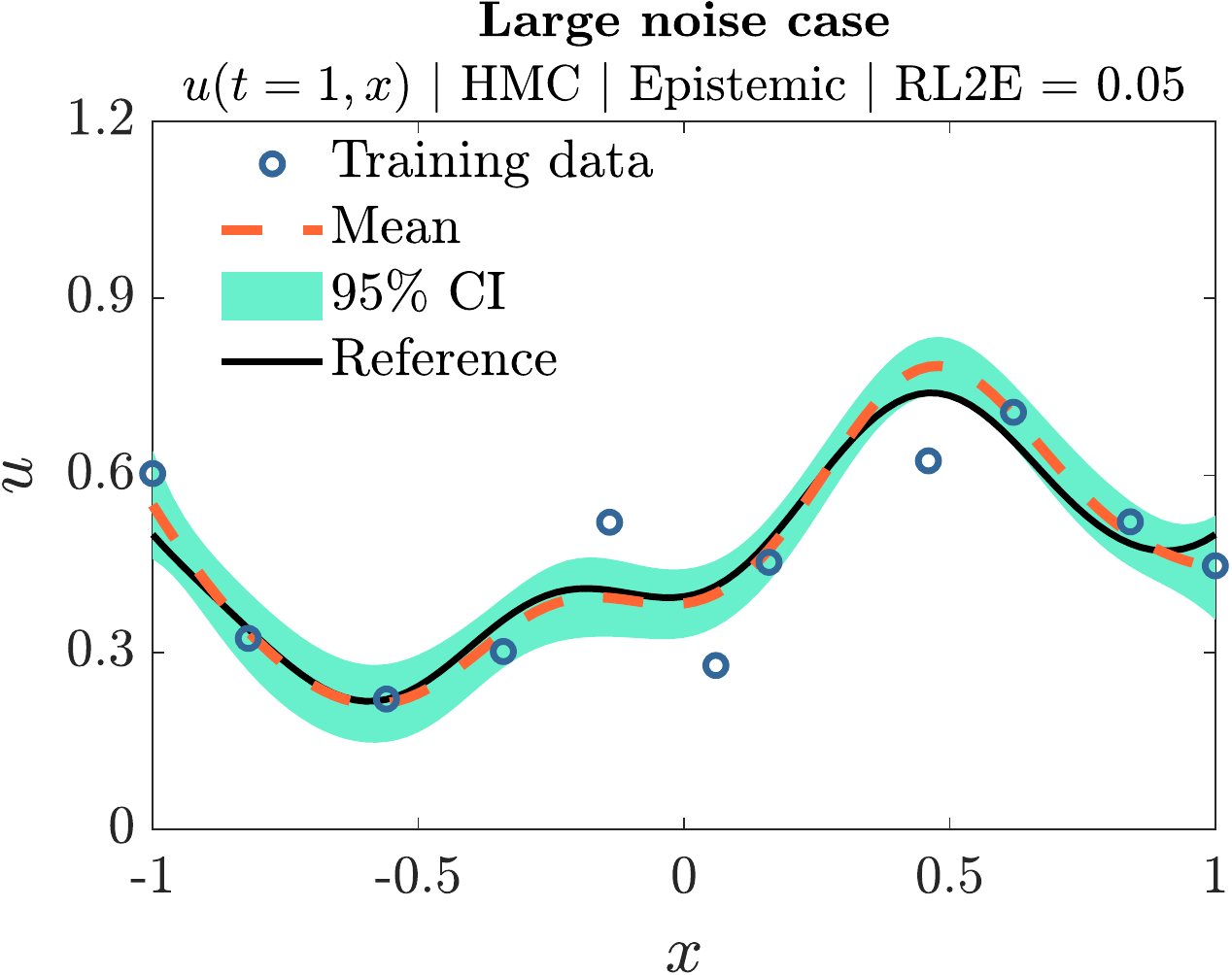}}
	\subcaptionbox{}{}{\includegraphics[width=0.32\textwidth]{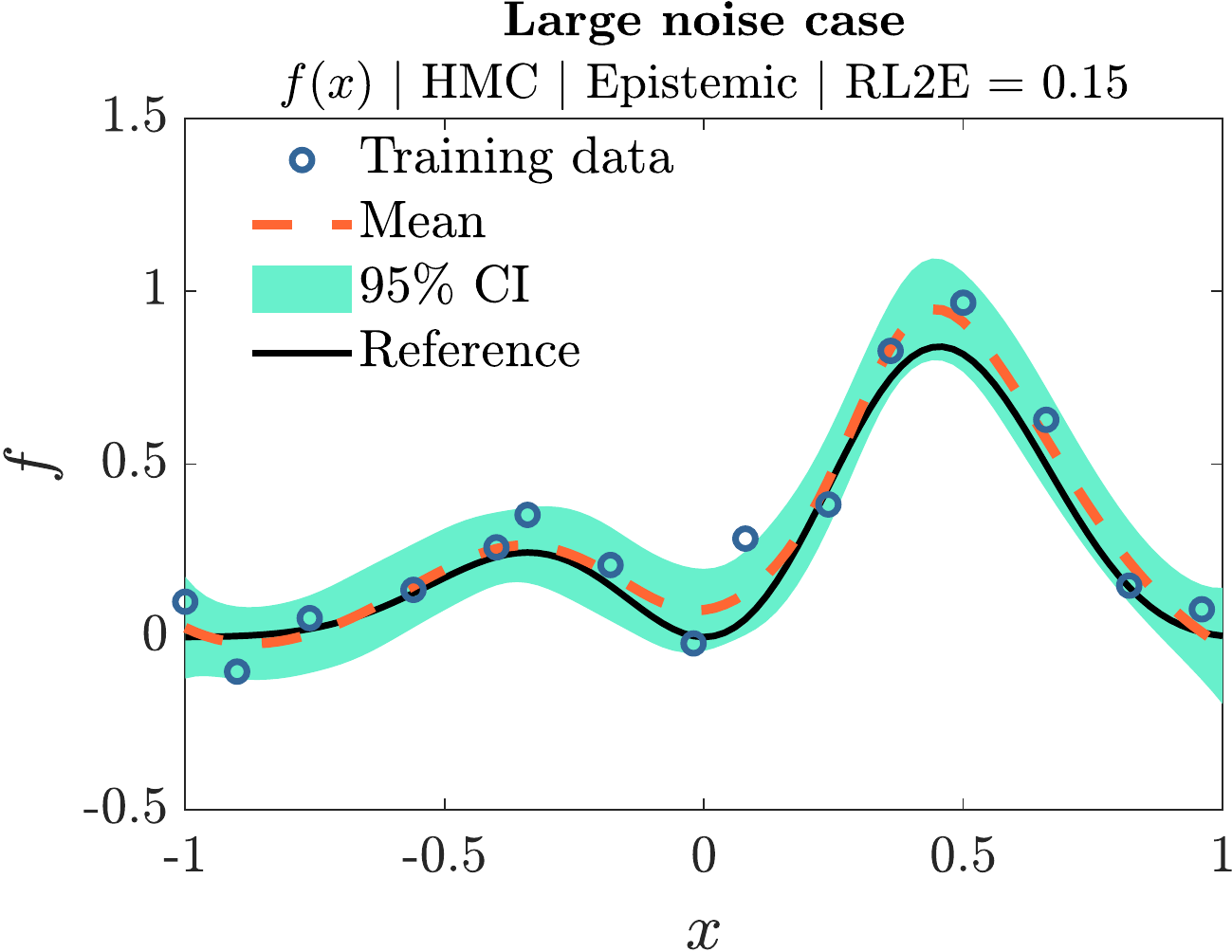}}
	\subcaptionbox{}{}{\includegraphics[width=0.32\textwidth]{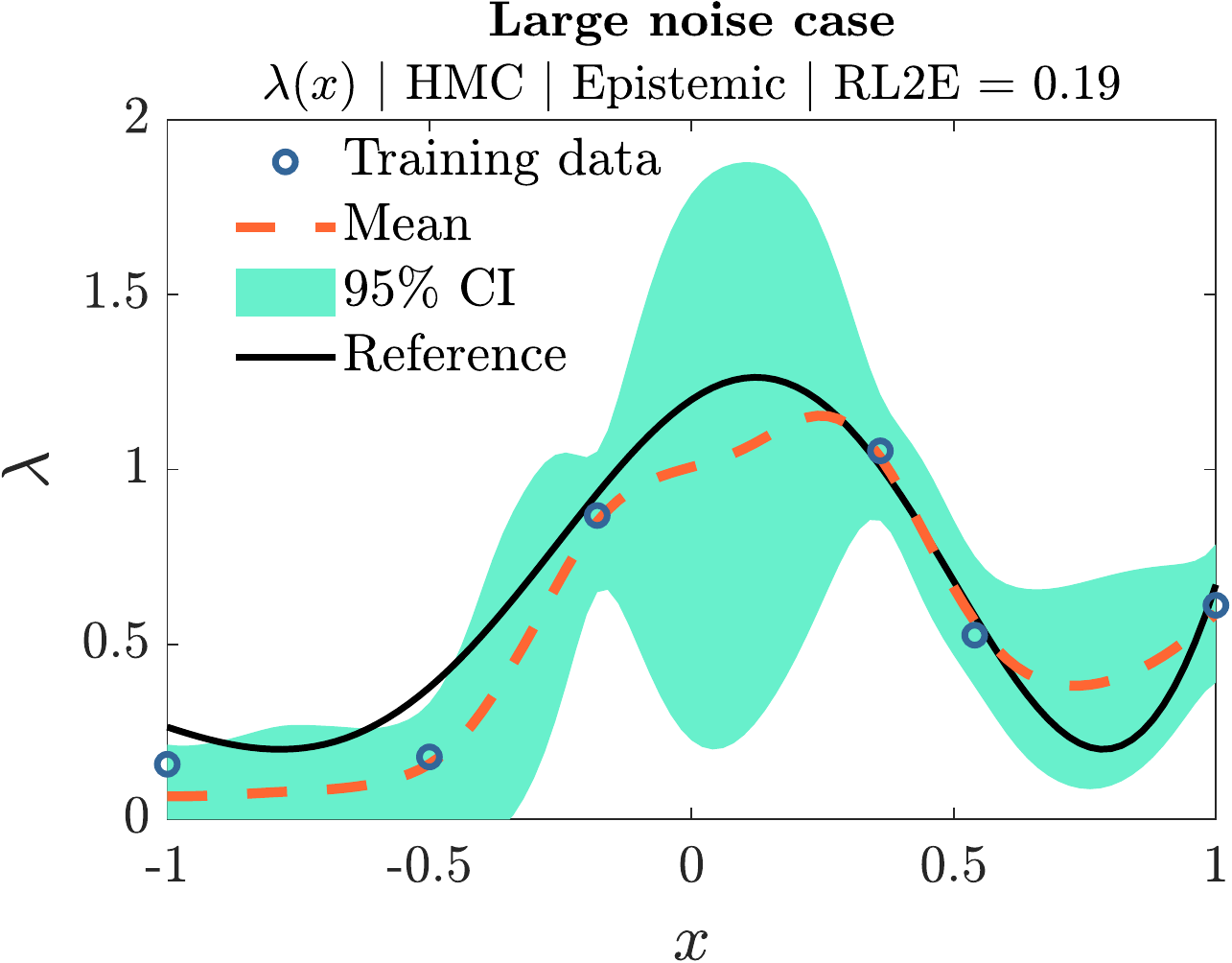}}
	\subcaptionbox{}{}{\includegraphics[width=0.32\textwidth]{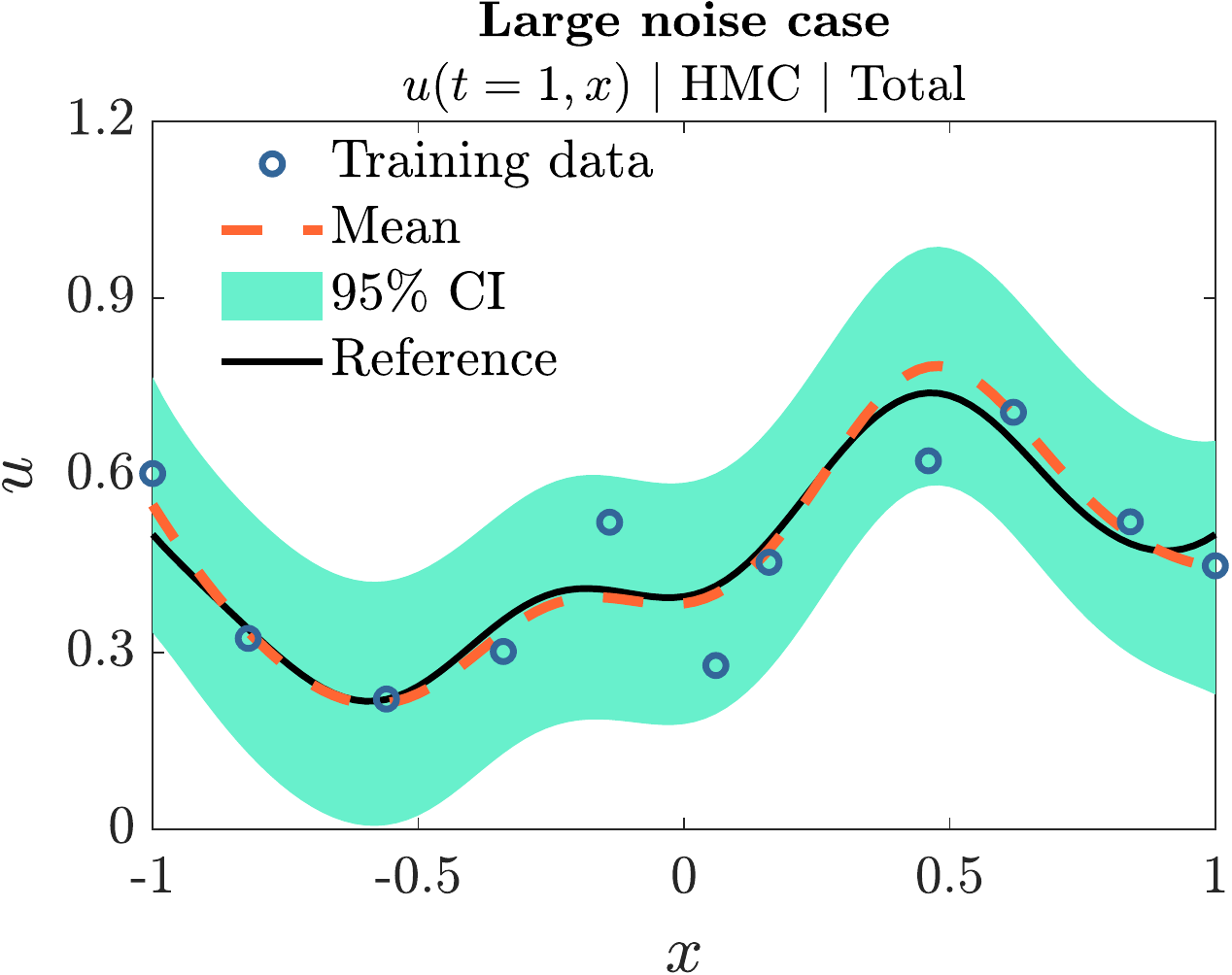}}
	\subcaptionbox{}{}{\includegraphics[width=0.32\textwidth]{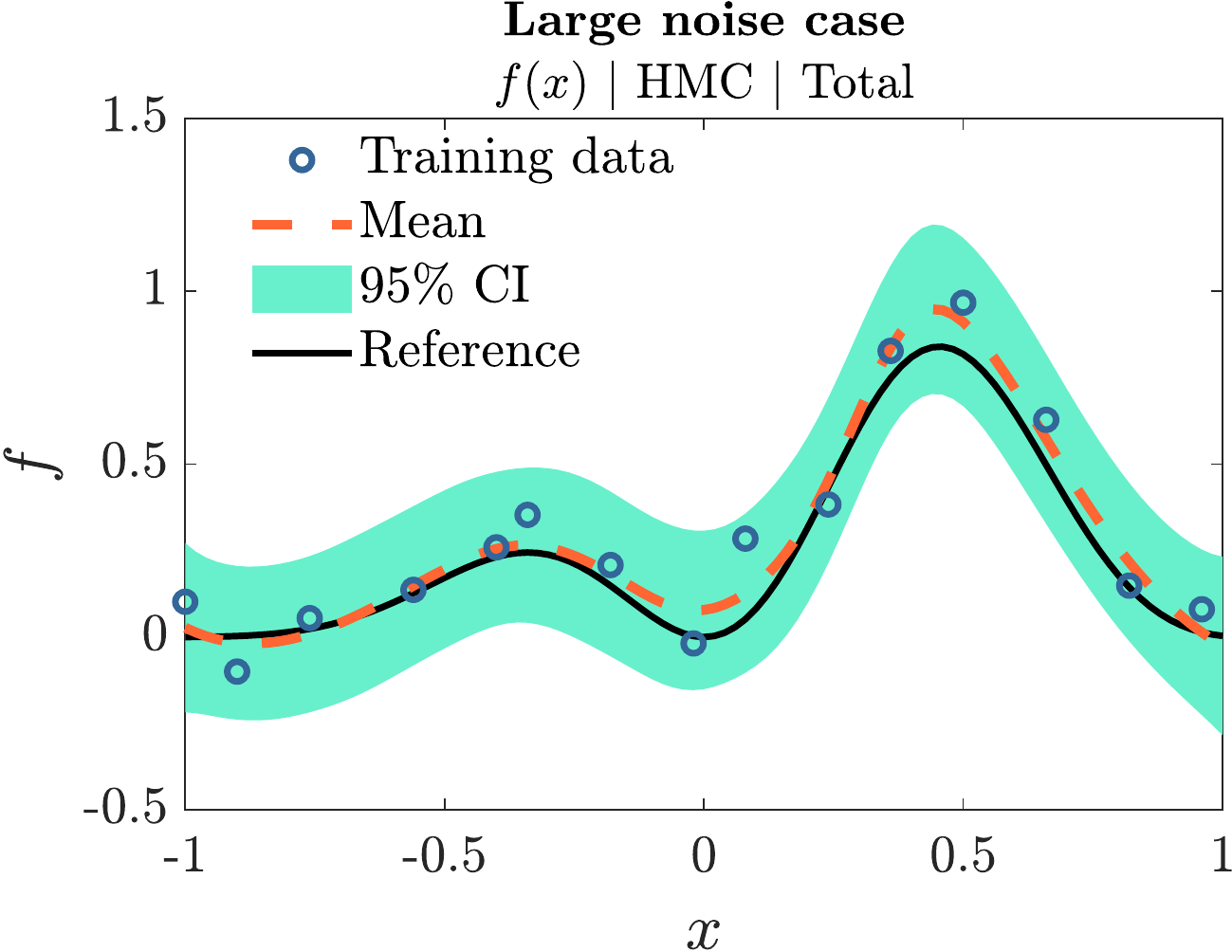}}
	\subcaptionbox{}{}{\includegraphics[width=0.32\textwidth]{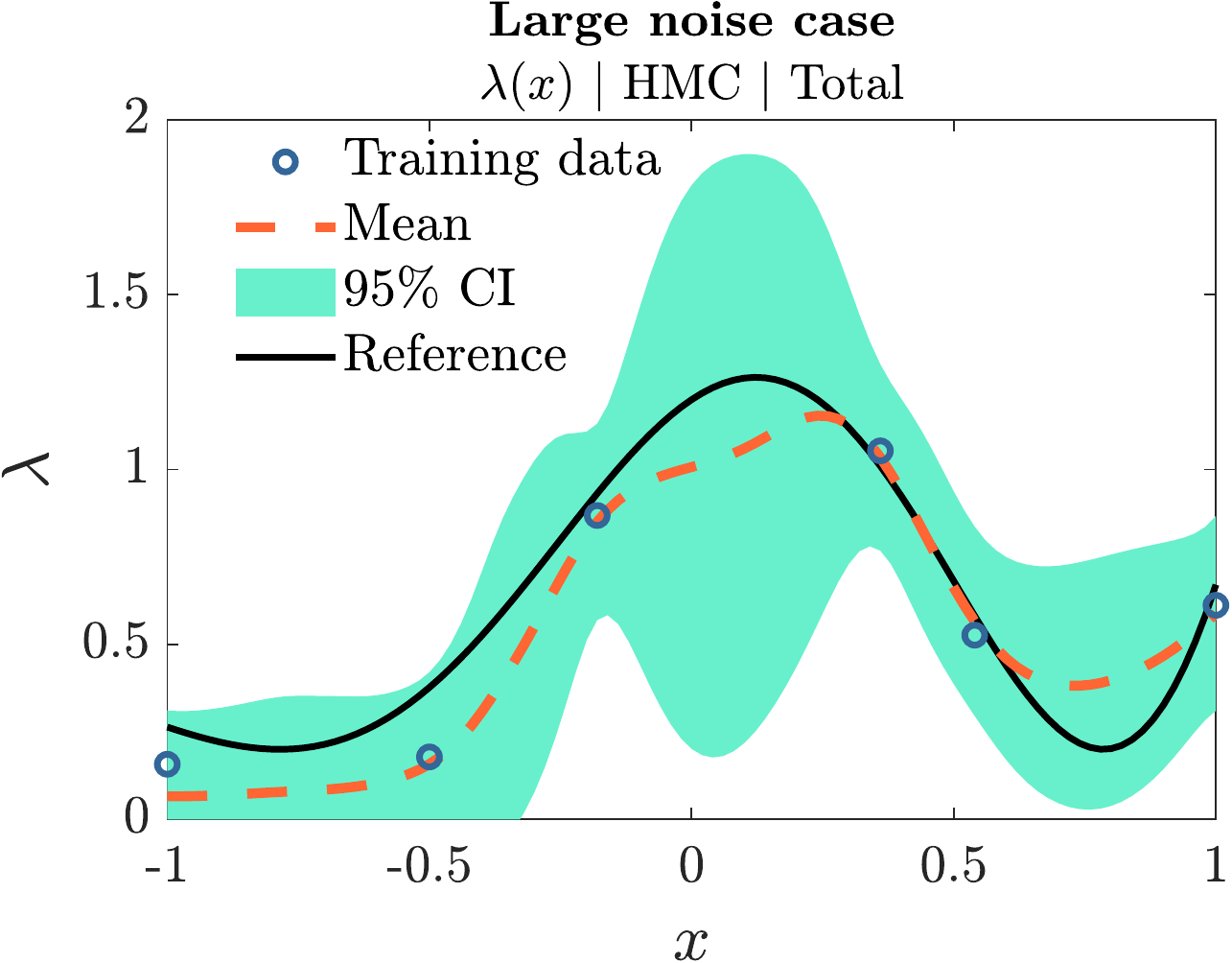}}
	\caption{
		Mixed PDE problem of Eq.~\eqref{eq:comp:pinns:stand} | \textit{Large noise case}: training data, reference functions, as well as mean and uncertainty ($95\%$ CI) predictions of HMC for $u$, $f$, and $\lambda$.
		\textbf{Top row:} epistemic uncertainty.
		\textbf{Bottom row:} total uncertainty including the \textit{known} amount of aleatoric uncertainty, $\sigma_u = 0.1$.
	}
	\label{fig:comp:pinns:large_noise}
\end{figure}

In this case, we keep the locations of all the training data the same as in Section \ref{sec:comp:pinns:stand} and only change the noise scale in the data from 0.05 to 0.1. The rest of the setup for this problem as well as the architecture of the BNN are the same as in Section \ref{sec:comp:pinns:stand}. The results from HMC are illustrated in Fig. \ref{fig:comp:pinns:large_noise}. As shown, (1) we obtain larger predicted total uncertainty, as expected, when comparing the results with the ones in the standard case in Section \ref{sec:comp:pinns:stand}, (2) the computational errors for $u$, $f$, and $\lambda$ are all within the $95 \%$ CIs. 

\subsubsection{Extrapolation case: data concentrated on one side of the space domain}\label{app:comp:pinns:results:extra}

\begin{figure}[!ht]
	\centering
	\subcaptionbox{}{}{\includegraphics[width=0.32\textwidth]{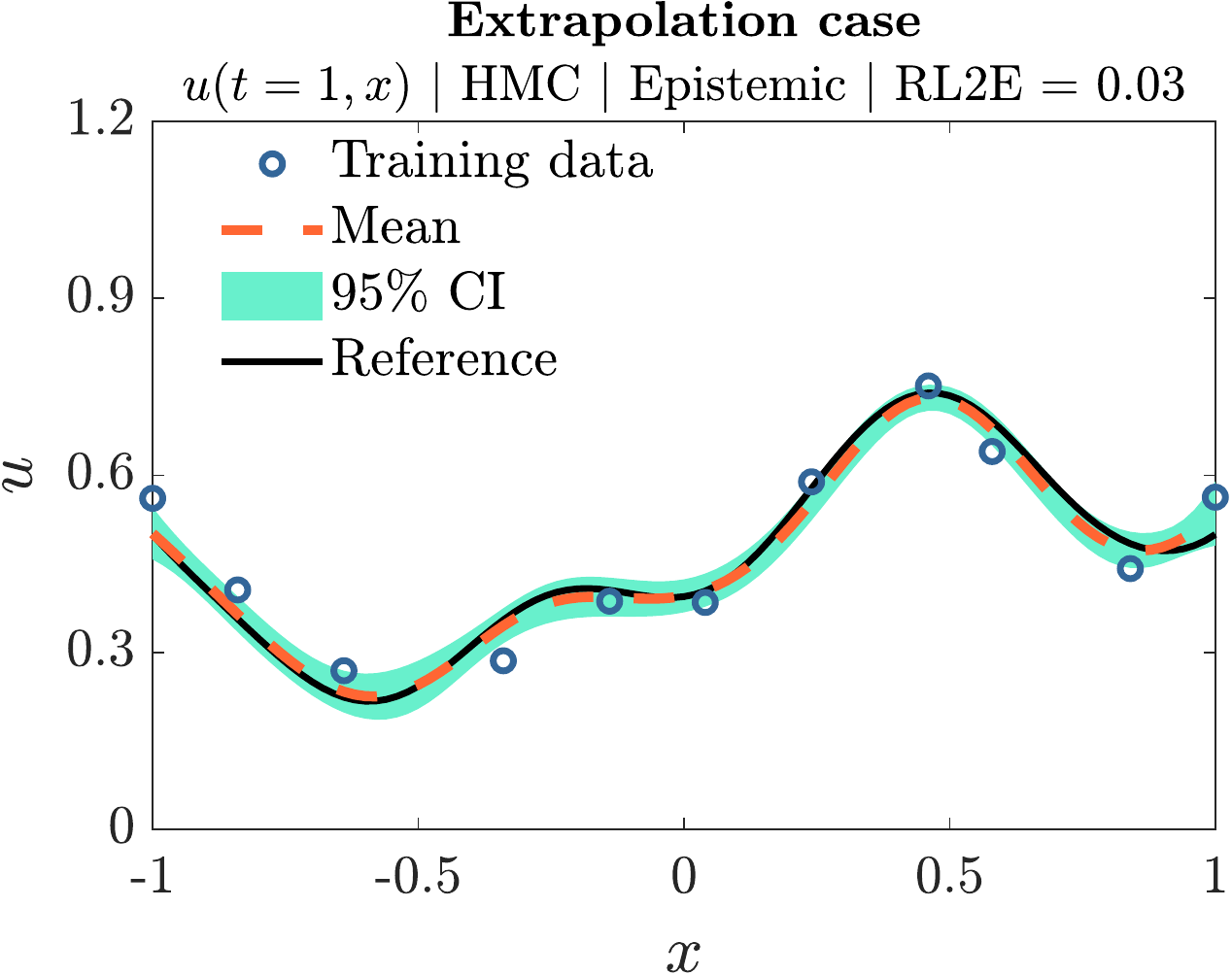}}
	\subcaptionbox{}{}{\includegraphics[width=0.32\textwidth]{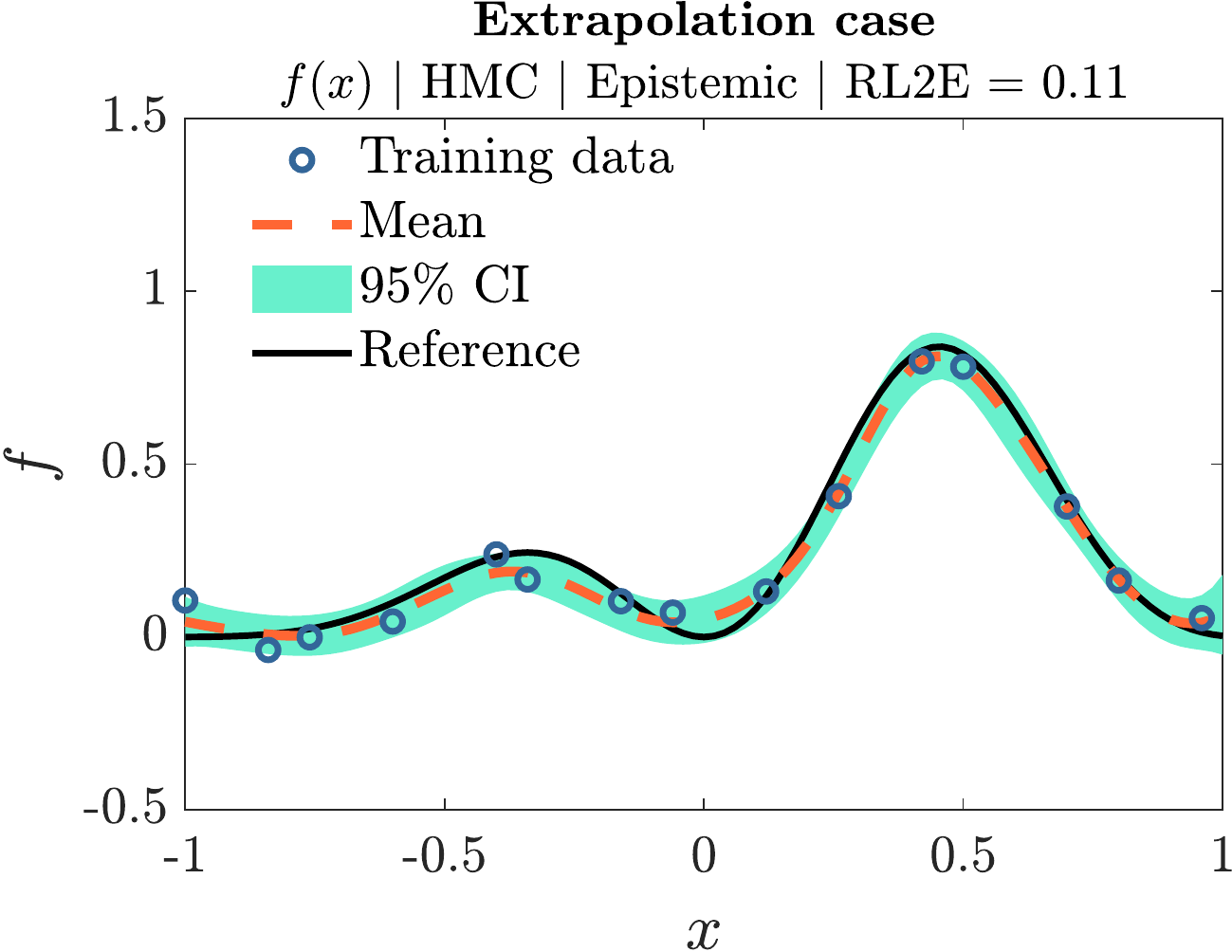}}
	\subcaptionbox{}{}{\includegraphics[width=0.32\textwidth]{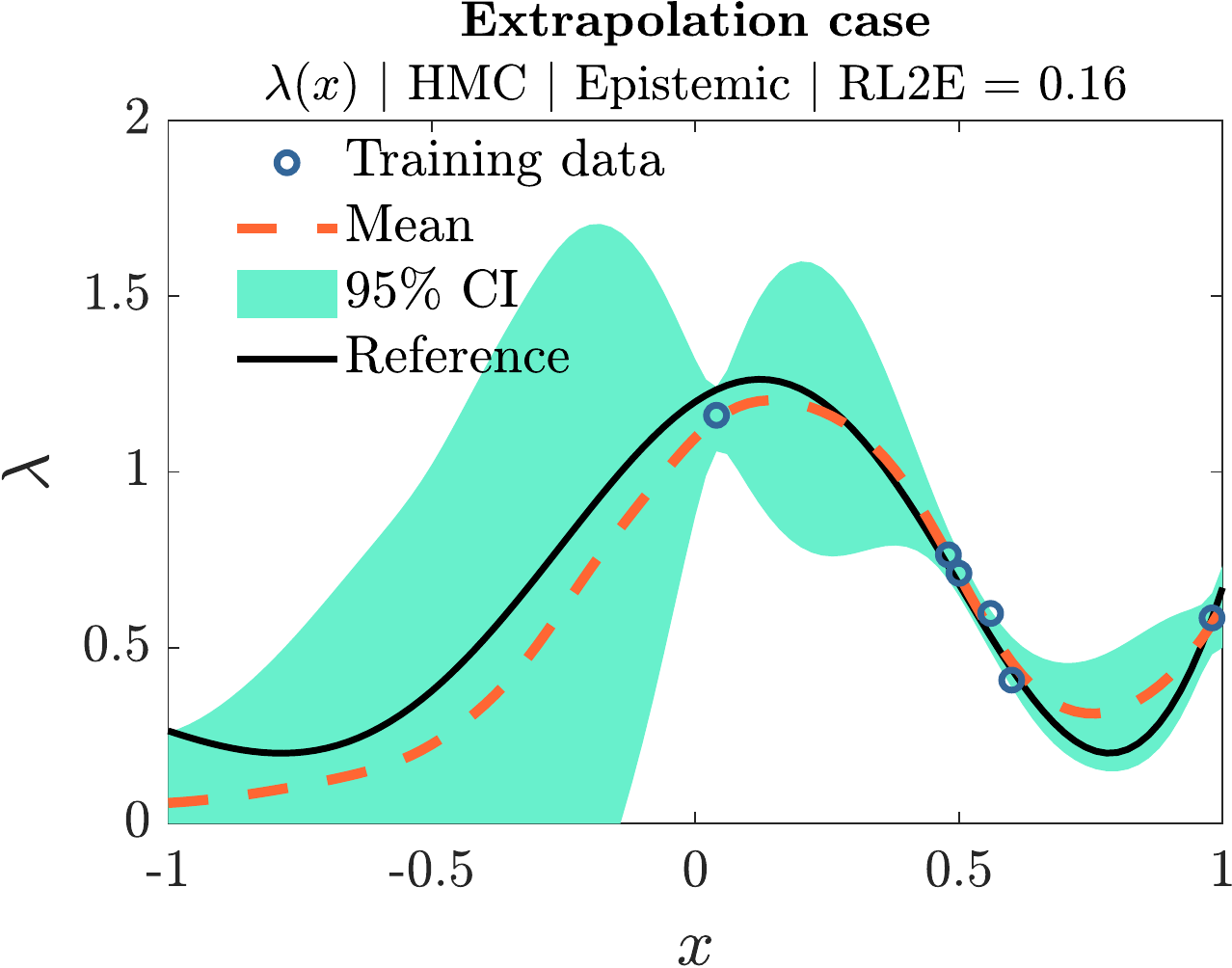}}
	\subcaptionbox{}{}{\includegraphics[width=0.32\textwidth]{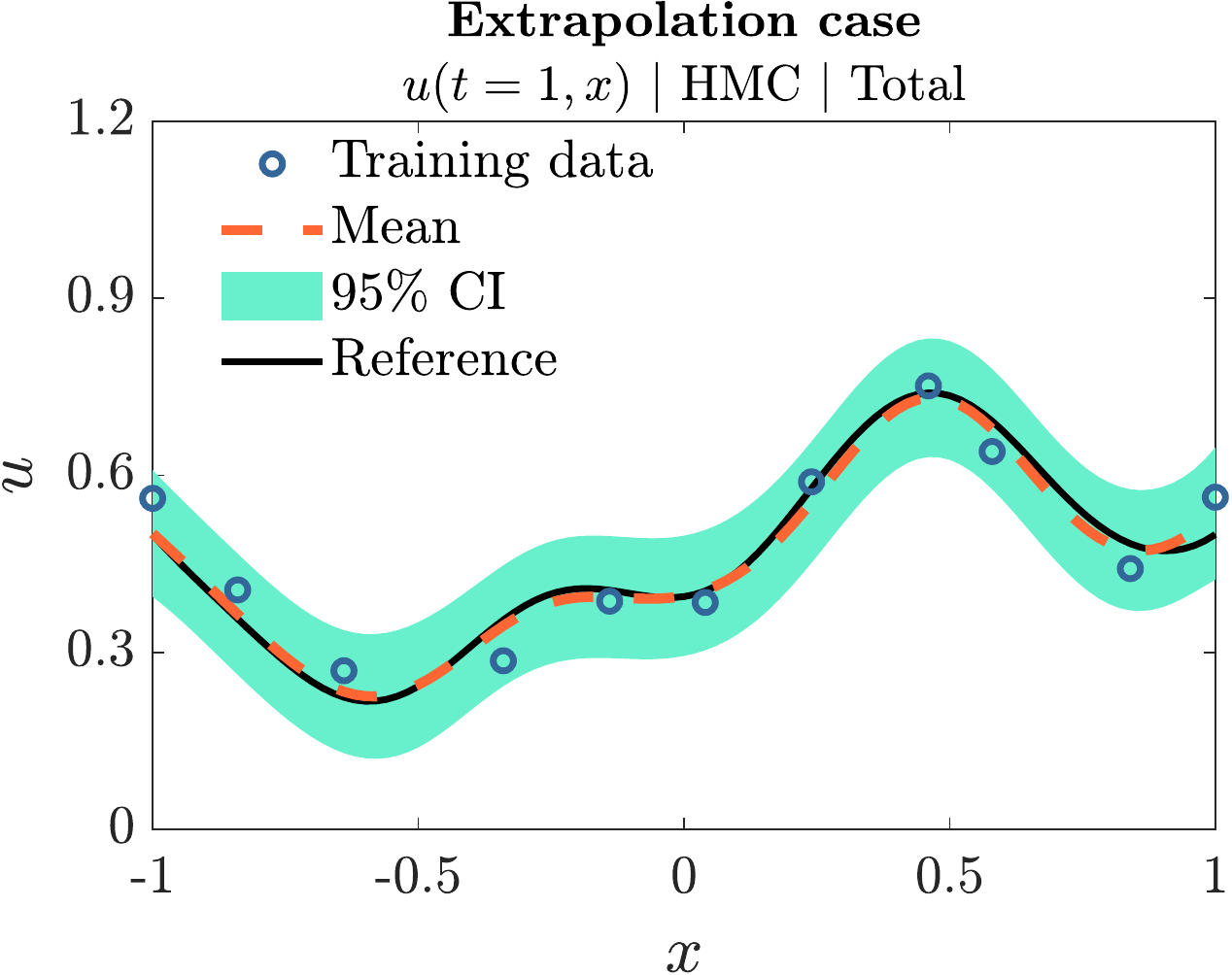}}
	\subcaptionbox{}{}{\includegraphics[width=0.32\textwidth]{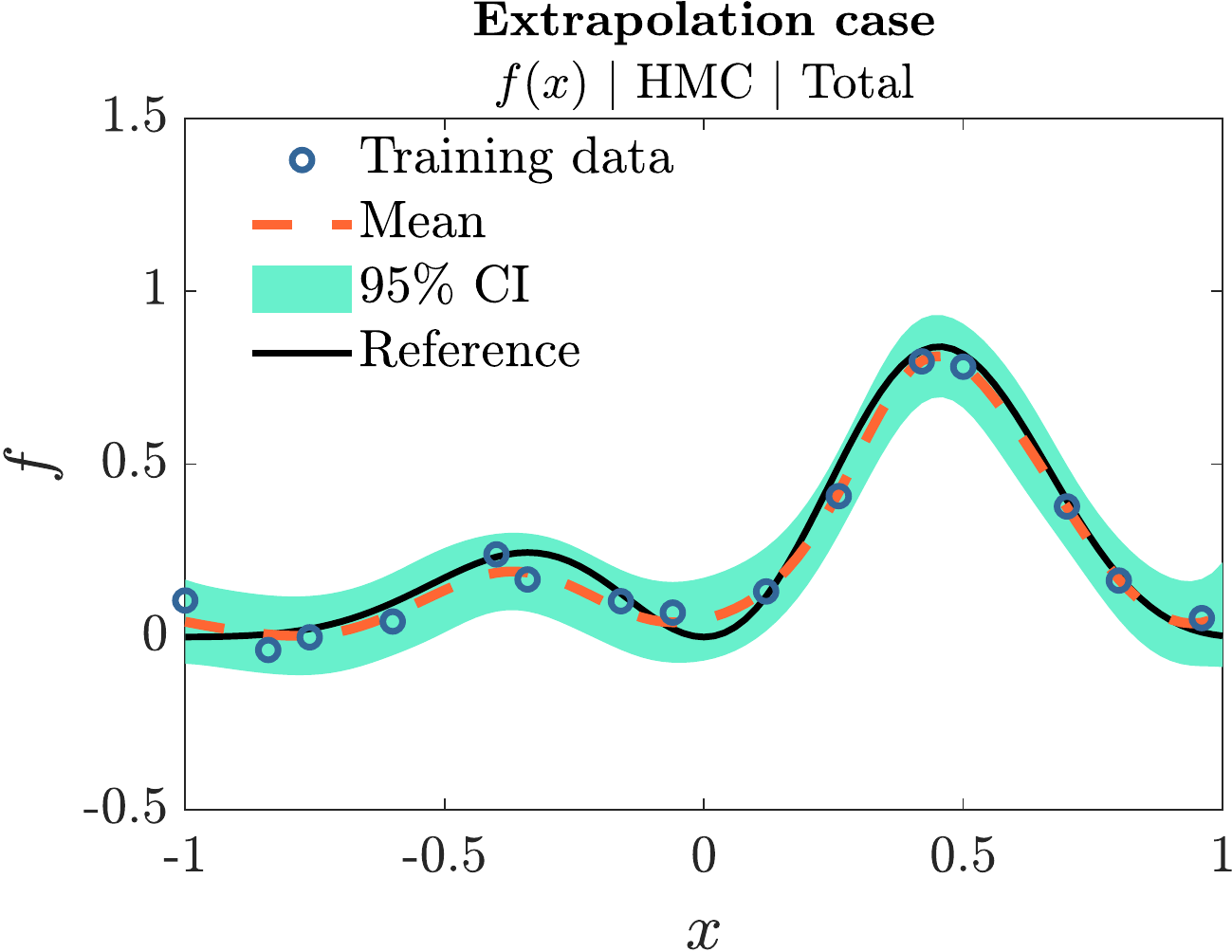}}
	\subcaptionbox{}{}{\includegraphics[width=0.32\textwidth]{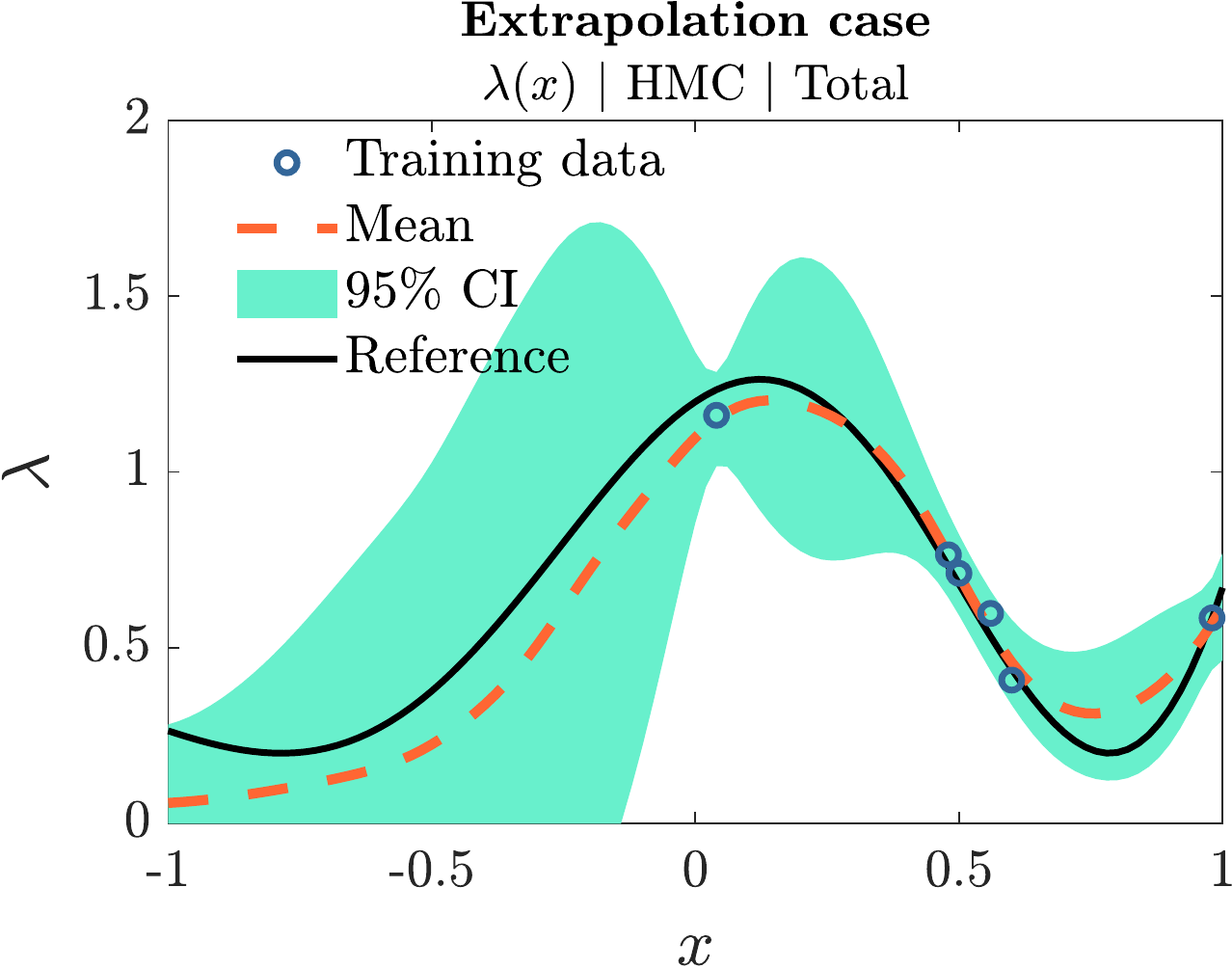}}
	\caption{
		Mixed PDE problem of Eq.~\eqref{eq:comp:pinns:stand} | \textit{Extrapolation case}: training data, reference functions, as well as mean and uncertainty ($95\%$ CI) predictions of HMC for $u$, $f$, and $\lambda$.
		\textbf{Top row:} epistemic uncertainty.
		\textbf{Bottom row:} total uncertainty including the \textit{known} amount of aleatoric uncertainty, $\sigma_u = 0.05$.
	}
	\label{fig:comp:pinns:extrapolation}
\end{figure}

Here we use the same training data for $u$ and $f$ as in Section \ref{sec:comp:pinns:stand}. In addition, we assume that we have the same number of observations on $\lambda$, but all the data are concentrated in $x \in [0, 1]$, as shown in Fig. \ref{fig:comp:pinns:extrapolation}. In Fig. \ref{fig:comp:pinns:extrapolation} we present the results obtained by HMC. As shown, (1) we can still obtain accurate predictions for $\lambda$ for $x \in [-1, 0]$, where we do not have any training data. This is because $\lambda$ is coupled with $u$ and $f$ through the PDE. Thus, information on $u$ and $f$ is leveraged to help the inference of $\lambda$, (2) the computational errors for $u$, $f$, and $\lambda$ are are all within the $95 \%$ CIs.

\subsubsection{Steep boundary layers case: large gradients close to the boundaries of the space domain}\label{app:comp:pinns:results:steep}

In this section, we consider a more challenging case, in which we change the reaction rate $\lambda$ to make $u$ more sharp at the two boundaries. The equations for the problem considered here are expressed as:
\begin{align}
	\partial_t u = D\partial^2_x u - k_r(x) u^3 + f(x), t \in [0, 1], ~ x \in [-1, 1], \label{eq:comp:pinns:pde:steep}\\
	u(-1, t) = u(1, t) = 1, ~ u(x, 0) = \cos^2(\pi x), D=0.01, \label{eq:comp:pinns:bcs}\\
	\lambda(x) = 0.2 + \exp(x^2)\cos^2(3x), \label{eq:comp:pinns:k}\\
	f(x) = \exp(-\frac{(x - 0.25)^2}{2l^2}) \sin^2(3 x), ~l = 0.4. \label{eq:comp:pinns:f}
\end{align}
We solve the above equations using the {\emph{Matlab PDE toolbox}} as in Section \ref{sec:comp:pinns:stand}, and use the numerical results as the reference solution as well as to generate the training data. 
Specifically, we keep the locations of measurements for $u$ and $\lambda$ the same as in Section \ref{sec:comp:pinns:stand}, and for $f$ we assign more measurements near the two boundaries and further delete some measurements away from boundaries. The total number of measurements of $f$ is $N_f=13$. The noise scales for the training data are $\sigma_u = \sigma_f = \sigma_\lambda = 0.05$.

\begin{figure}[!ht]
	\centering
	\subcaptionbox{}{}{\includegraphics[width=0.32\textwidth]{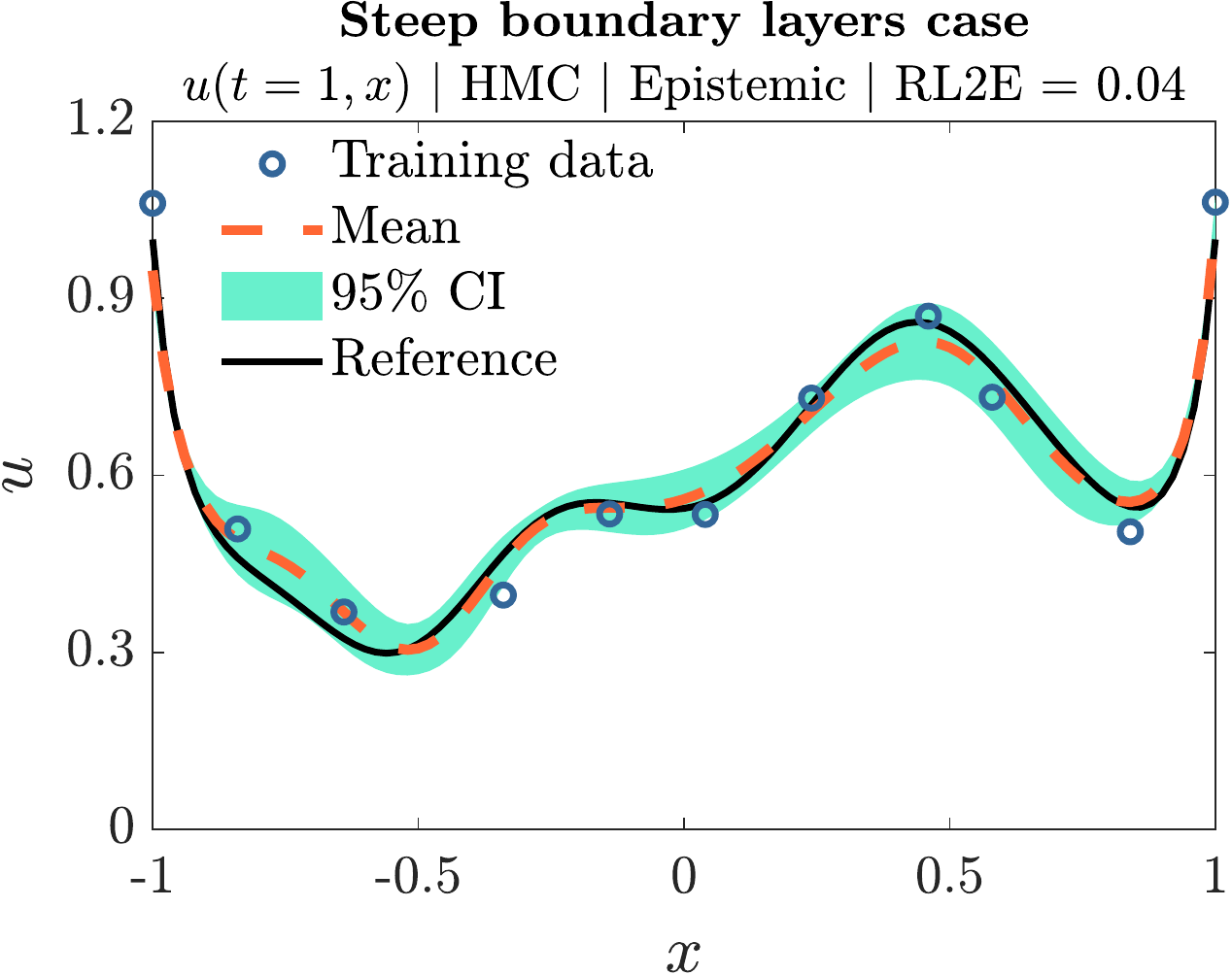}}
	\subcaptionbox{}{}{\includegraphics[width=0.32\textwidth]{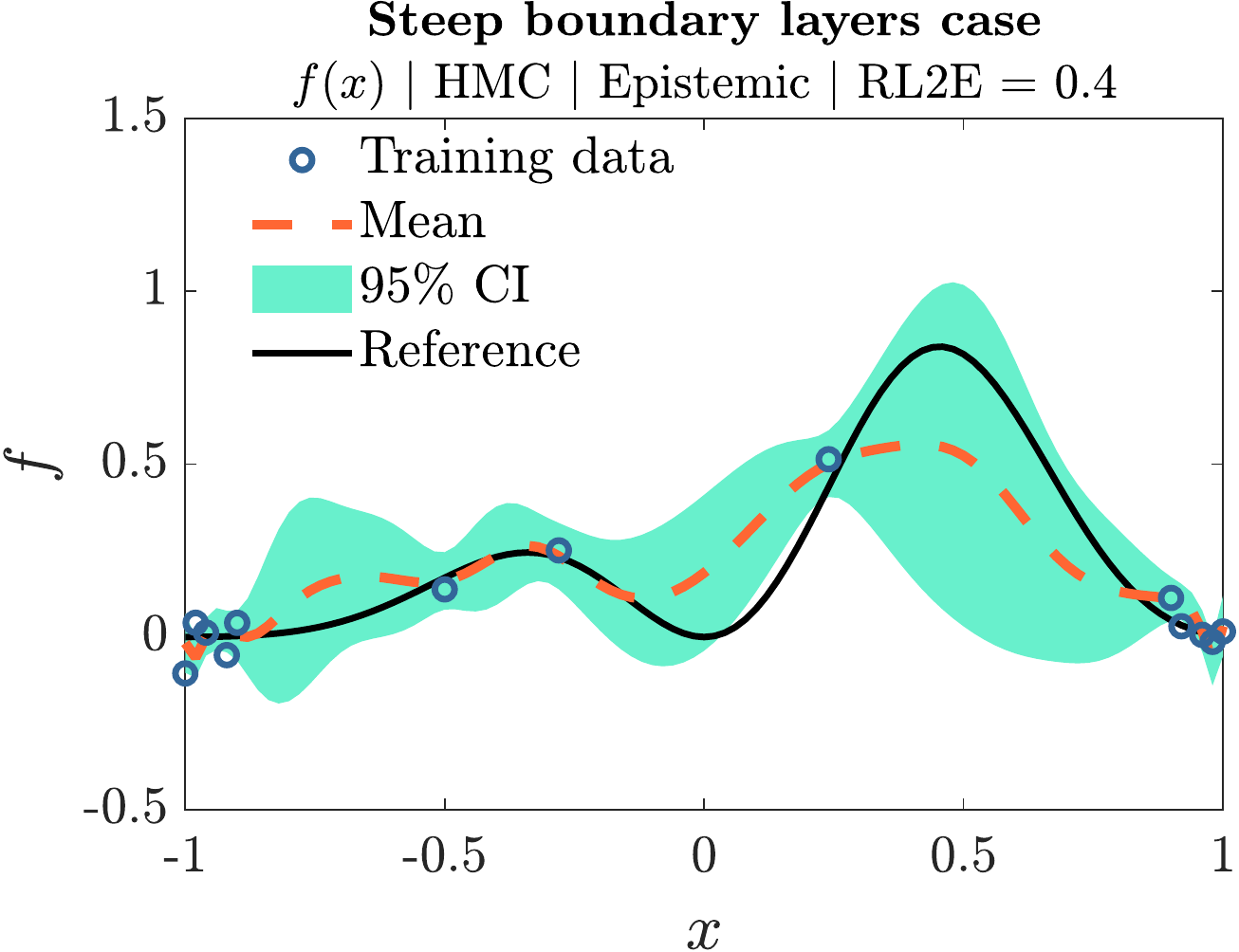}}
	\subcaptionbox{}{}{\includegraphics[width=0.32\textwidth]{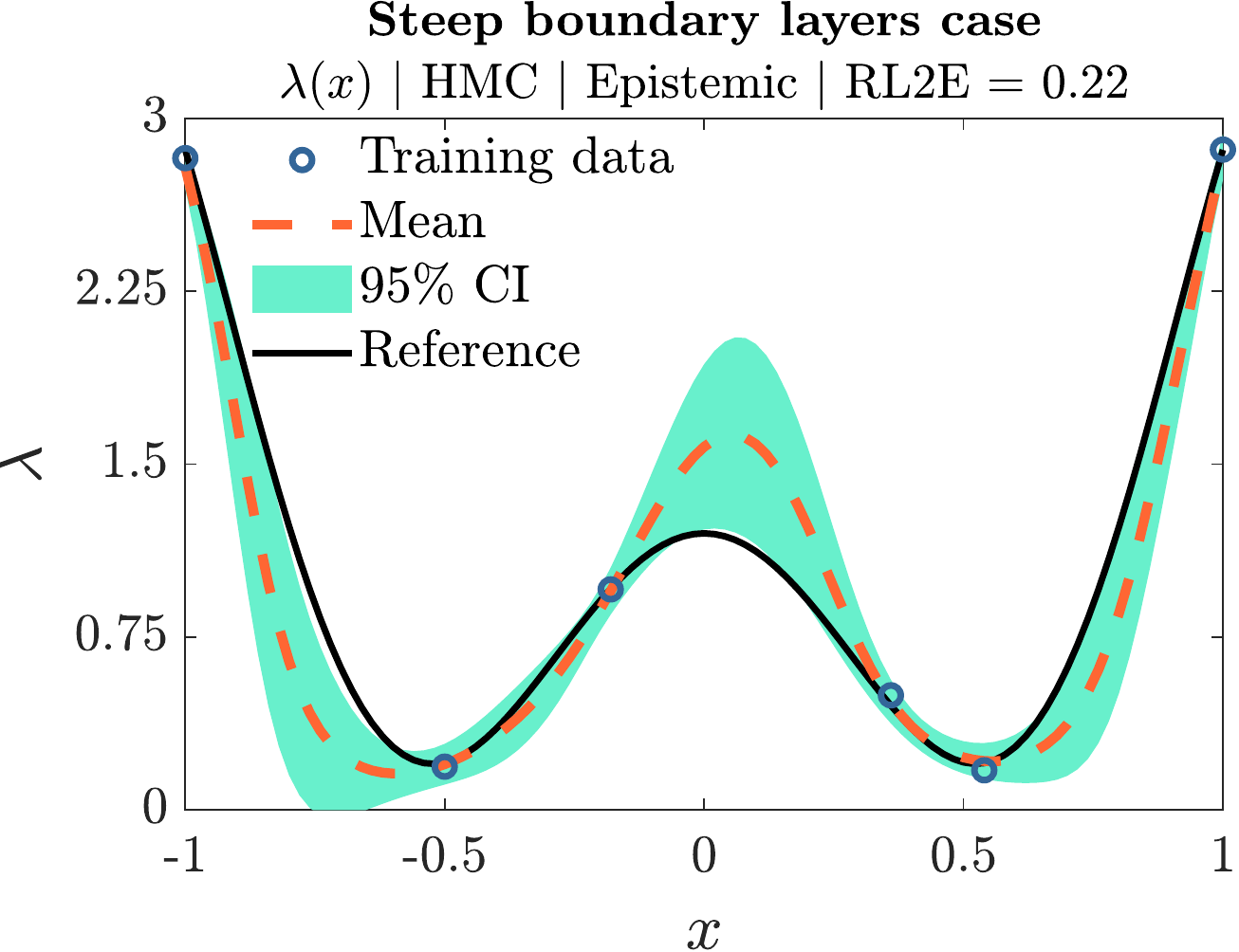}}
	\subcaptionbox{}{}{\includegraphics[width=0.32\textwidth]{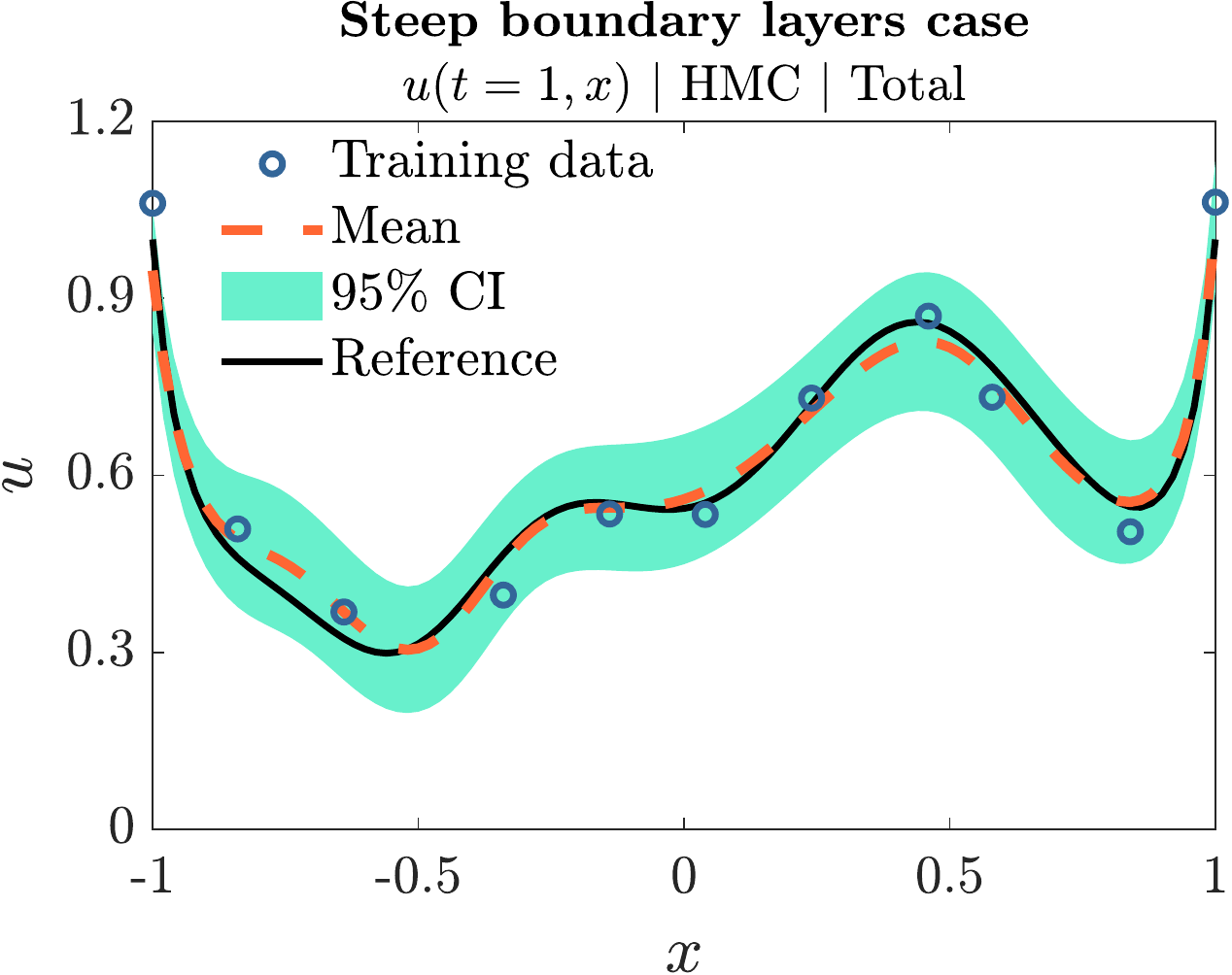}}
	\subcaptionbox{}{}{\includegraphics[width=0.32\textwidth]{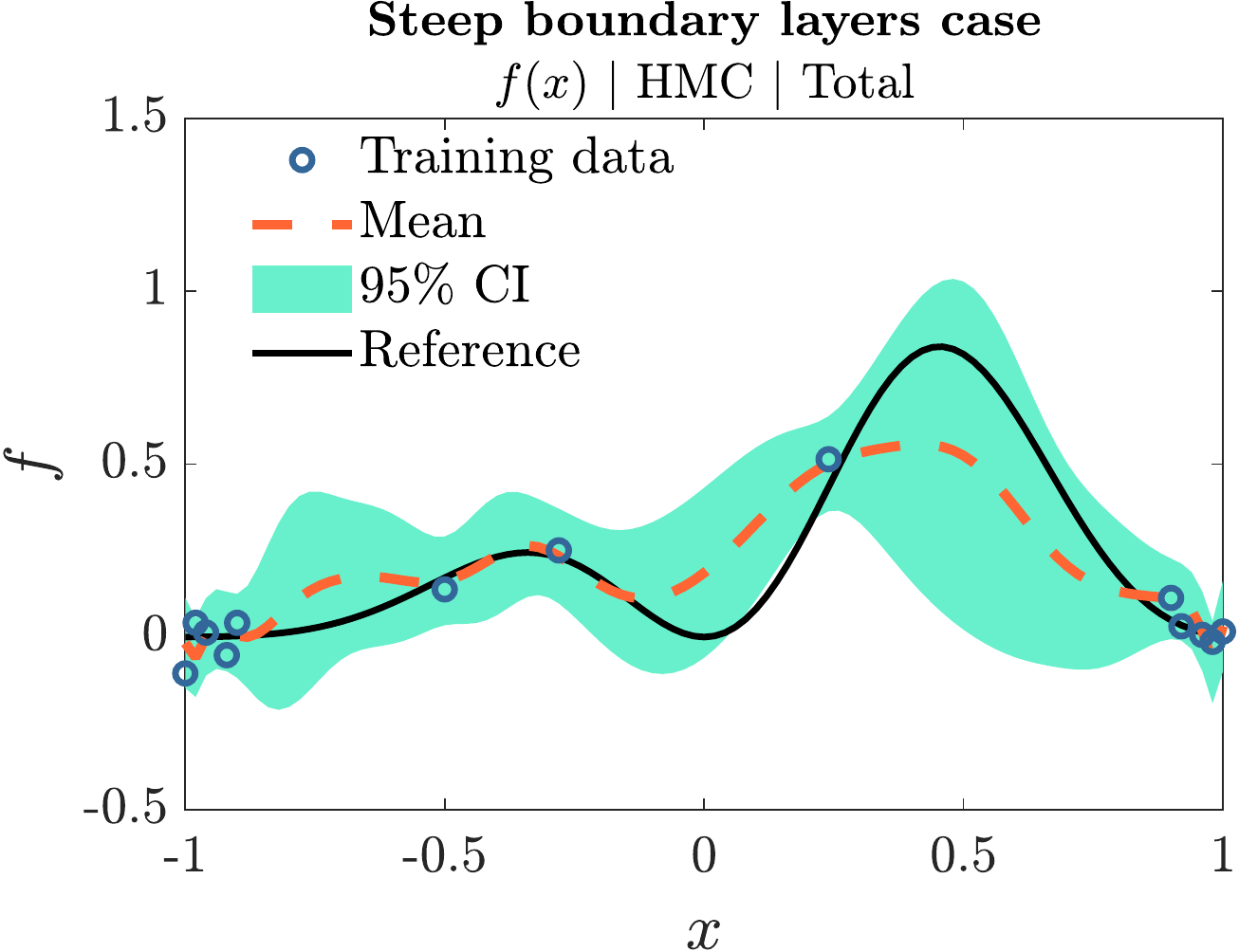}}
	\subcaptionbox{}{}{\includegraphics[width=0.32\textwidth]{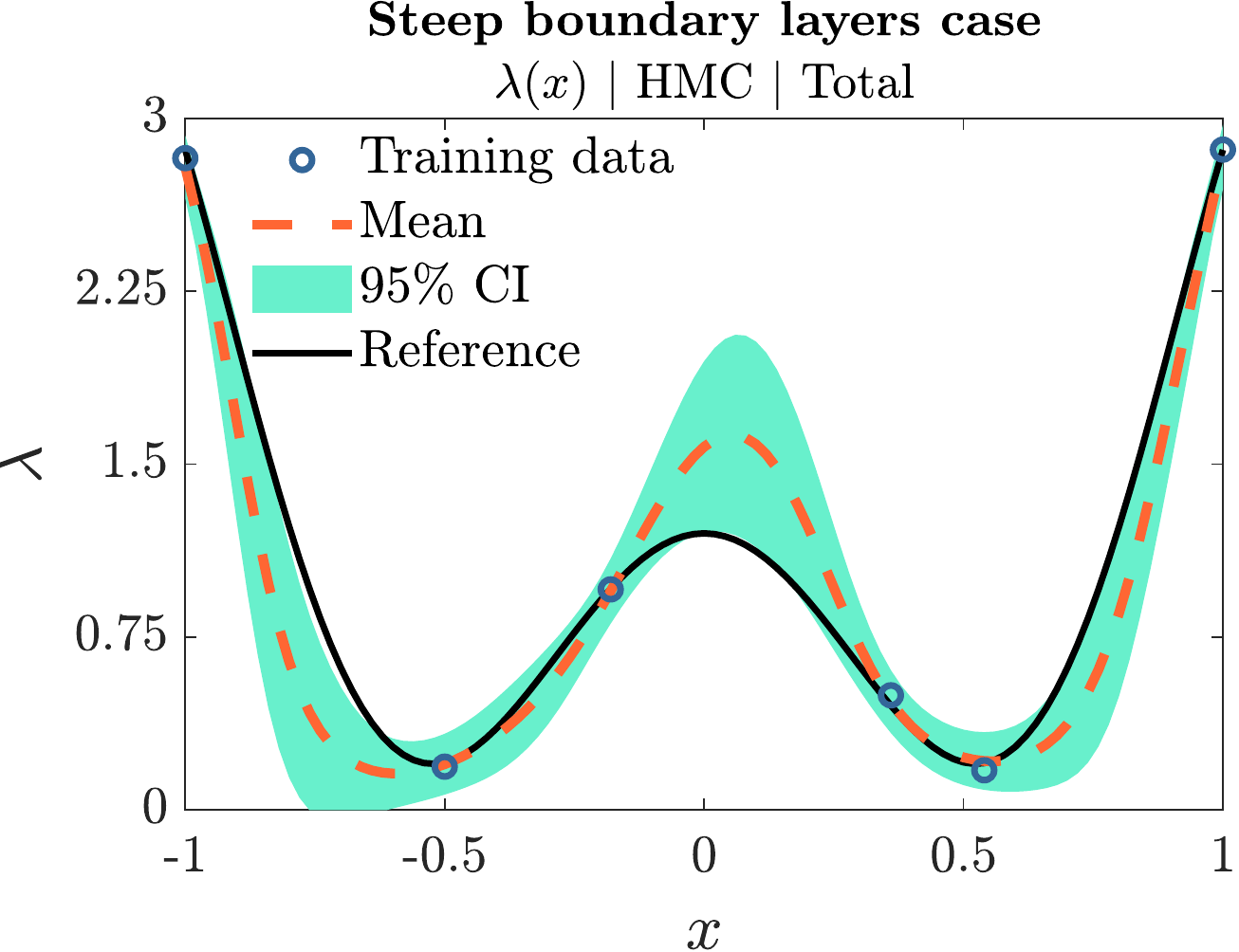}}
	\caption{
		Mixed PDE problem of Eq.~\eqref{eq:comp:pinns:stand} | \textit{Steep boundary layers case}: training data, reference functions, as well as mean and uncertainty ($95\%$ CI) predictions of HMC for $u$, $f$, and $\lambda$.
		\textbf{Top row:} epistemic uncertainty.
		\textbf{Bottom row:} total uncertainty including the \textit{known} amount of aleatoric uncertainty, $\sigma_u = 0.05$.
	}
	\label{fig:comp:pinns:steep}
\end{figure}

In Fig. \ref{fig:comp:pinns:steep} we present the results obtained by HMC. As shown, (1) the computational errors for $u$, $f$, and $\lambda$ are in most parts of the domains within the $95 \%$ CIs, (2) small fluctuations on the predictions of $f$ at the left part are observed, which may be caused by the steep boundary layer of $u$, since $f$ is related to the second-order derivative of $u$, and (3) no fluctuations on the predicted $f$ at the right part are observed, which is reasonable since we have a lot of data on $f$ near the right boundary, and the noise in the training data near the right boundary happens to be smaller than near the left boundary.

%% file: appendix/IN_app_results_stochastic.tex
\subsection{Additional results for the mixed stochastic problem of Section~\ref{sec:comp:stochastic}}\label{app:comp:stochastic:results}

In this section, we present the results corresponding to a post-training calibration experiment for the mixed SPDE problem of Section~\ref{sec:comp:stochastic}. 
First, we explain how to compute the calibration error (RMSCE) for this case. 
Consider one clean test realization containing values of $u$ sampled at $101$ points $x$.
Using one of U-PI-GAN, U-NNPC, or U-NNPC+, we produce 100 realizations of $u$, sampled at 101 points, with each one of the three ensemble samples. 
This results to 300 realizations. 
Equivalently, for each of the $1 \times 101 = 101$ test points of $u(x)$, there are 300 prediction samples and it is straightforward to compute the RMSCE according to Eq.~\eqref{eq:eval:eval:approx:rmsce}. If more than one test realizations are available, e.g., 100, we can use the same 300 predictions also for the additional realizations and compute the RMSCE based on $100 \times 101 = 10100$ test points. By doing so, each realization of $u(x)$ is using the same 300 predictions for computing the RMSCE. Similarly for $\lambda$.
Because in this case the test realizations are clean, we only consider epistemic uncertainty in the predictions (3 ensemble samples) and uncertainty due to stochasticity (100 realizations). We refer to this as total uncertainty in this case, although it does not include aleatoric uncertainty (see also Eq.~\eqref{eq:methods:sdes:var}).

\begin{figure}[!ht]
	\centering
	\subcaptionbox{}{}{\includegraphics[width=0.32\textwidth]{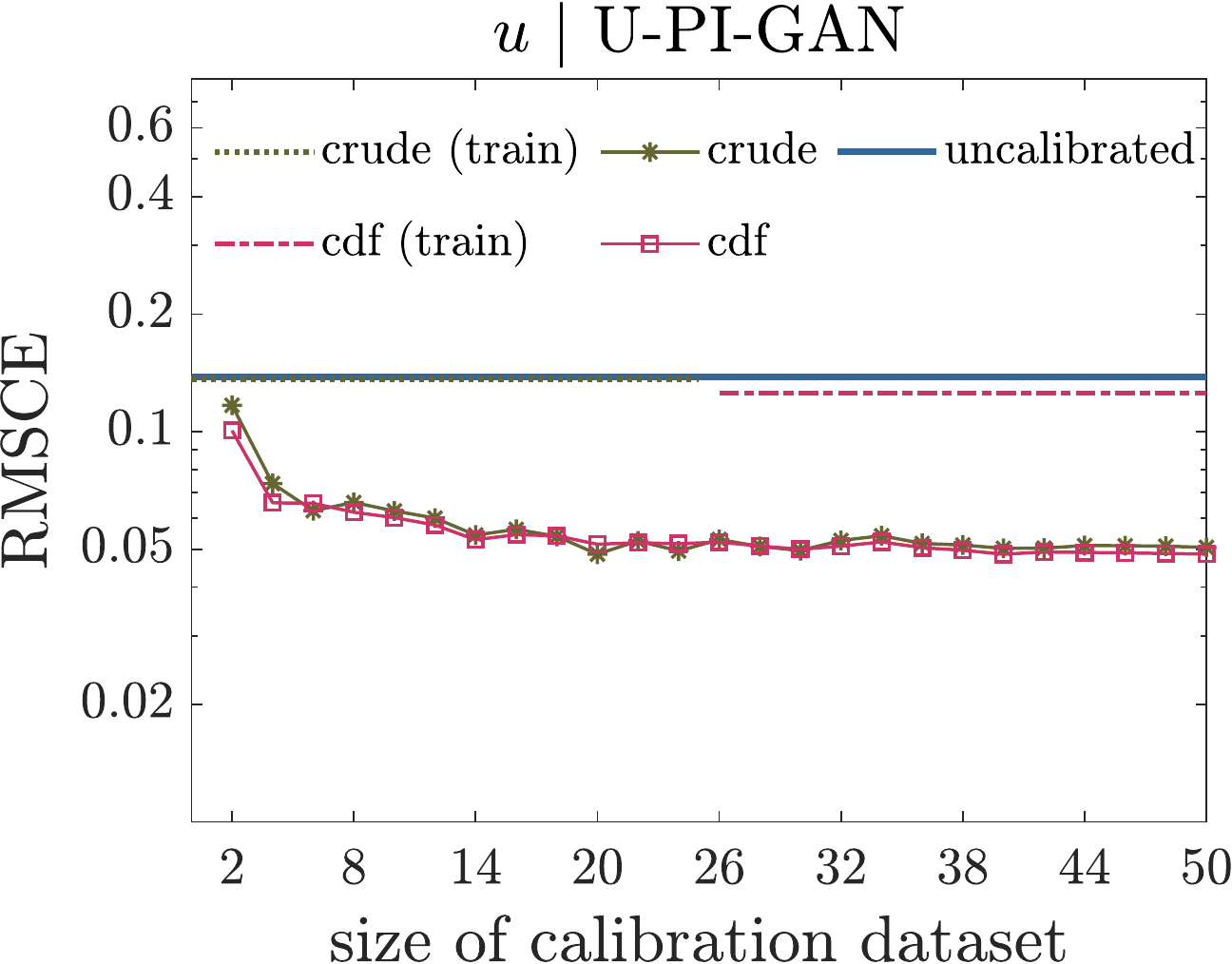}}
	\subcaptionbox{}{}{\includegraphics[width=0.32\textwidth]{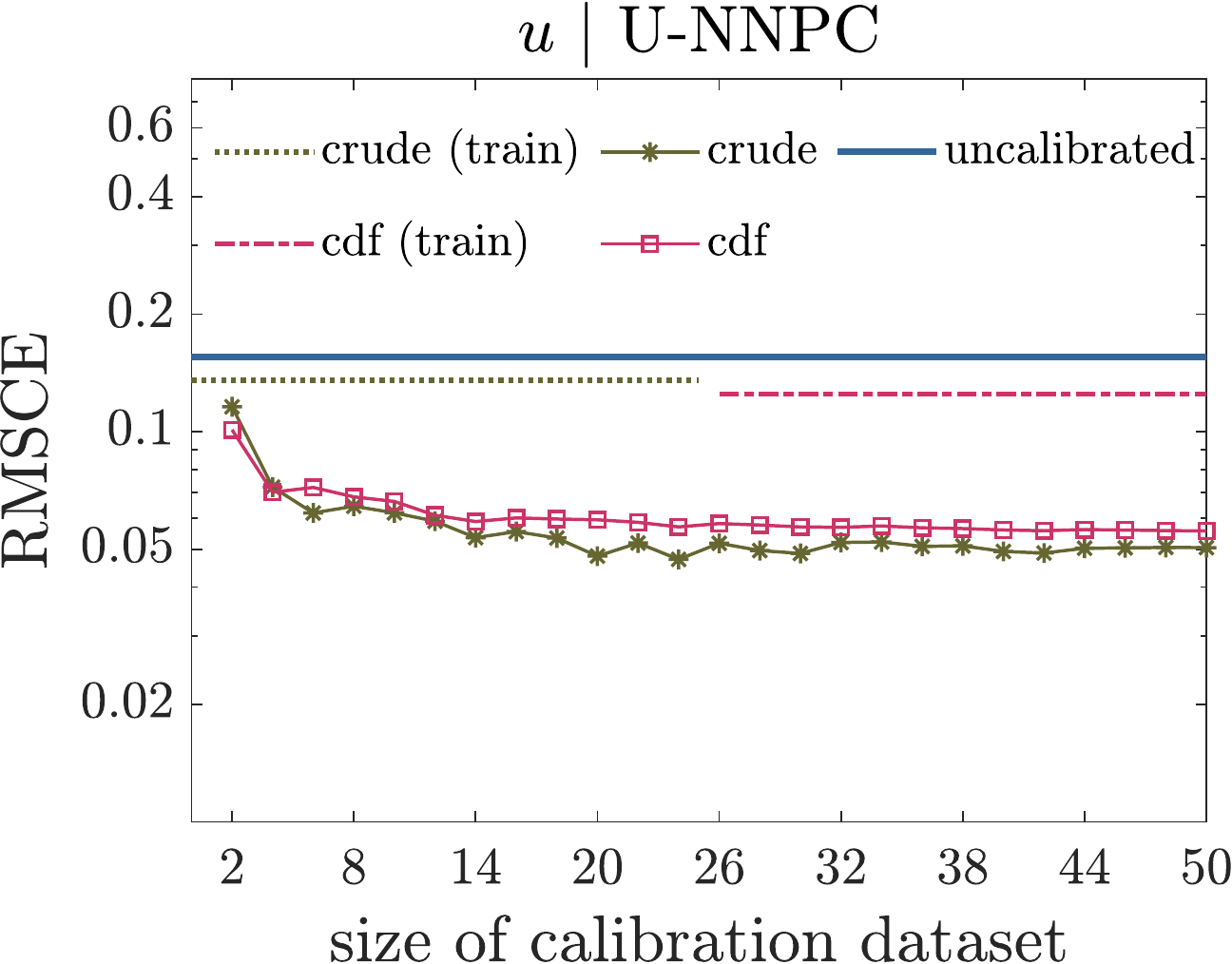}}
	\subcaptionbox{}{}{\includegraphics[width=0.32\textwidth]{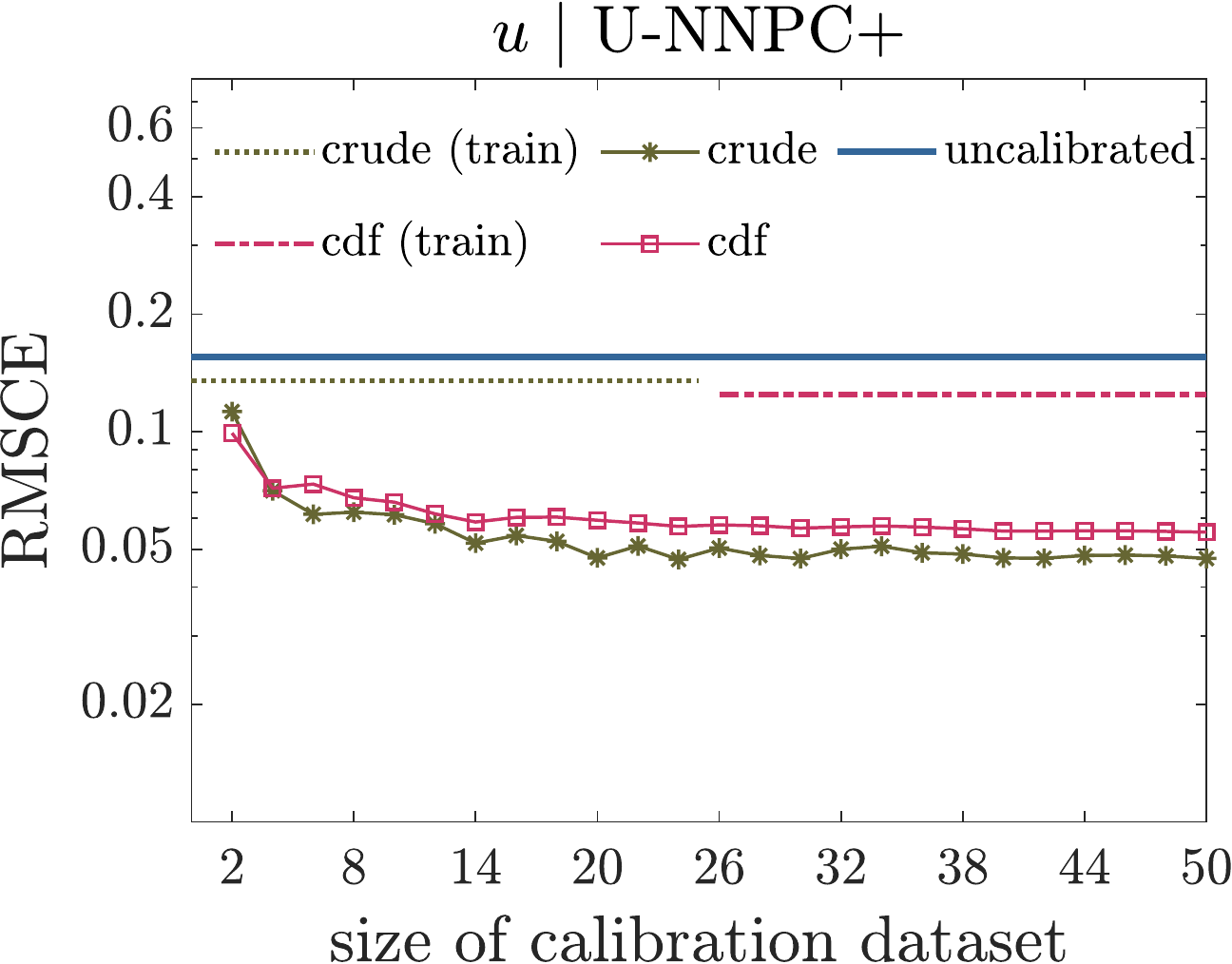}}
	\subcaptionbox{}{}{\includegraphics[width=0.32\textwidth]{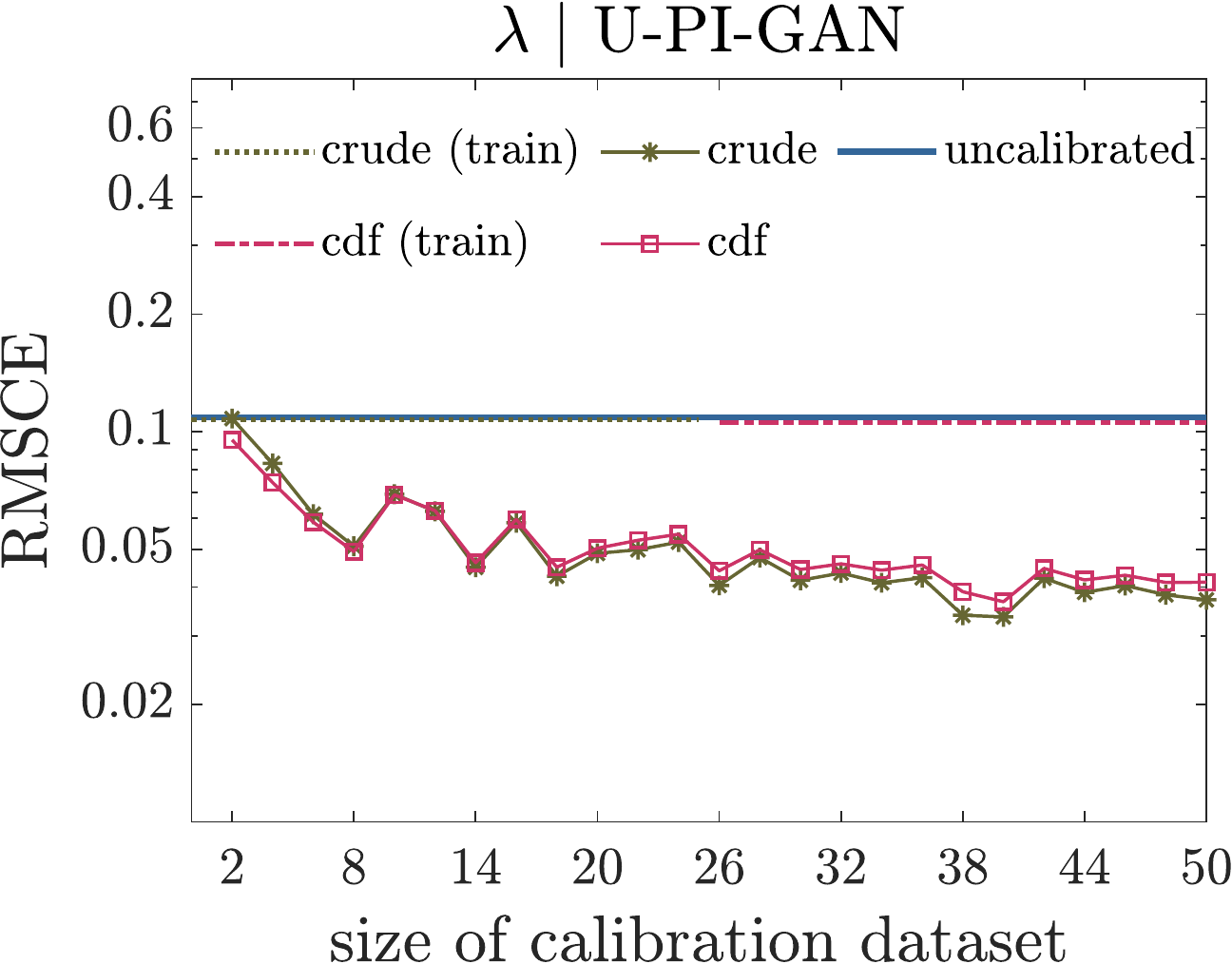}}
	\subcaptionbox{}{}{\includegraphics[width=0.32\textwidth]{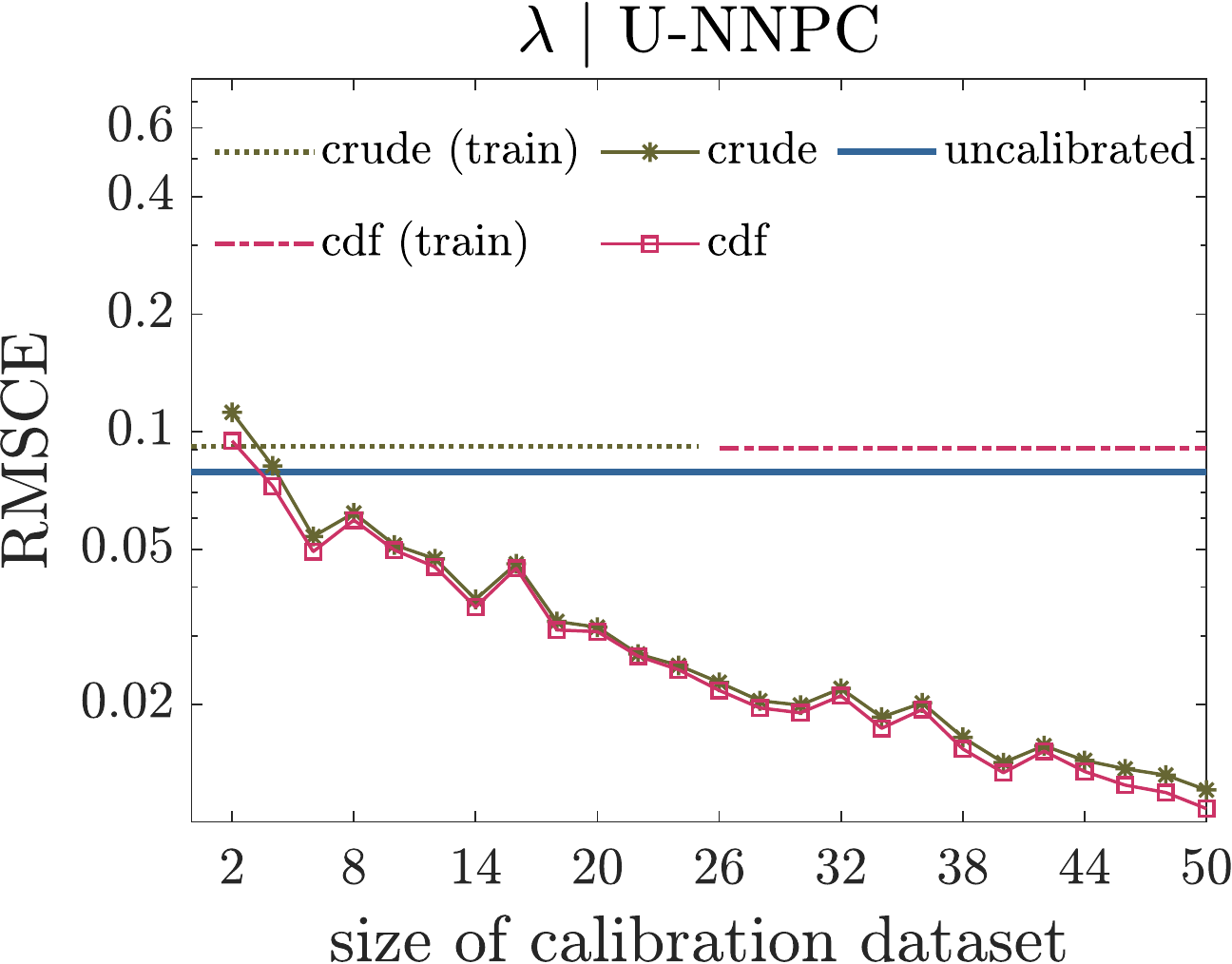}}
	\subcaptionbox{}{}{\includegraphics[width=0.32\textwidth]{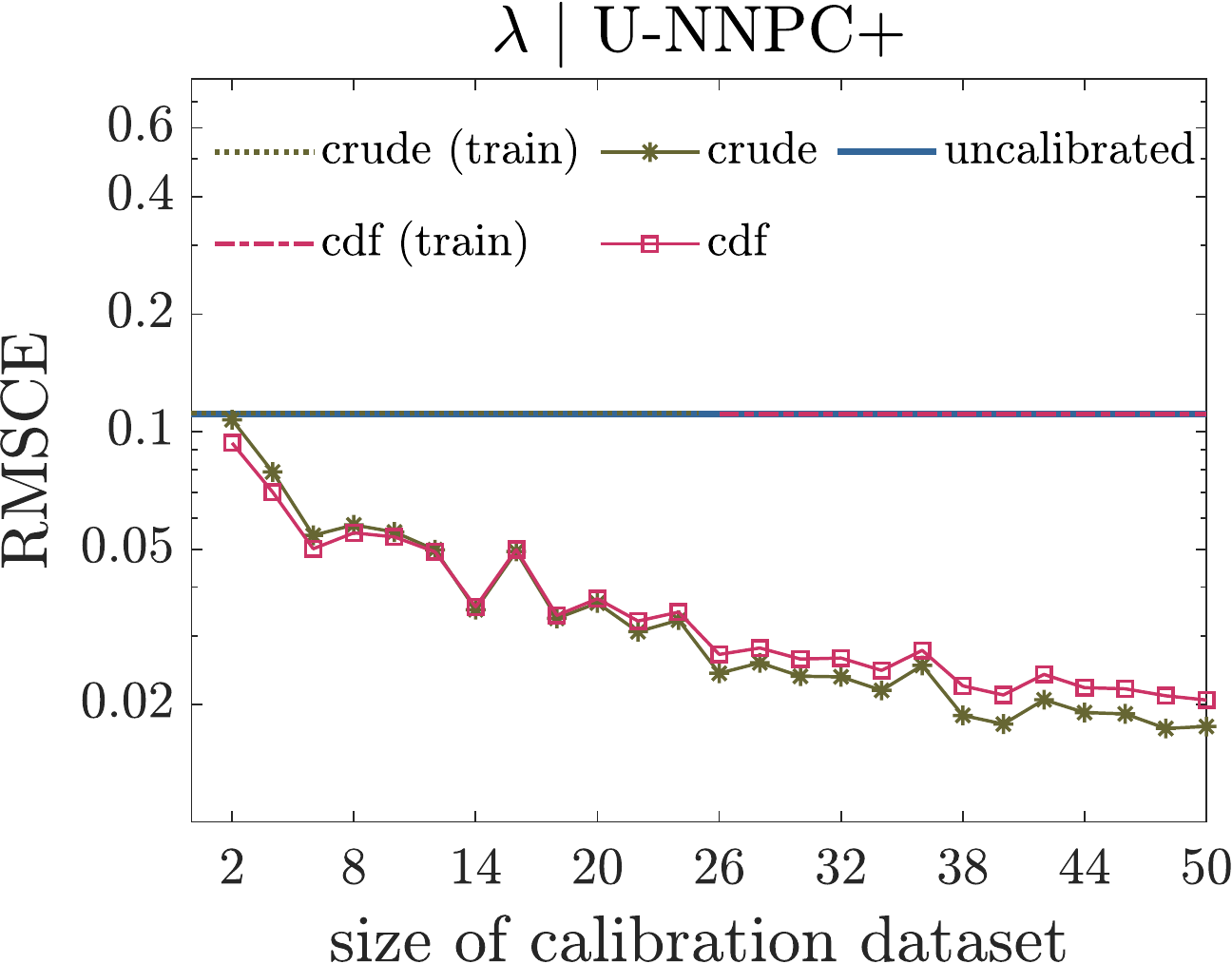}}
	\caption{
		Mixed stochastic problem of Eq.~\eqref{eq:comp:stochastic:eq}: even a few left-out noisy realizations (2-10) of $u$ and $\lambda$ can be used for calibrating the total uncertainty predictions of all considered UQ methods.
		Effect of calibration approach (Section~\ref{sec:eval:calib}) and calibration dataset (number of left-out realizations) on RMSCE for  U-PIGAN, U-NNPC, and U-NNPC+.
		Calibration is performed with total uncertainty, which in this case corresponds to epistemic and stochasticity uncertainties (see also Eq.~\eqref{eq:methods:sdes:var}).
	}
	\label{fig:comp:stochastic:calib}
\end{figure}

Next, for performing post-training calibration, we assume that we have 2-50 available realizations of $u$ and $\lambda$, not used during training. The data locations and noise are the same as in training. To perform calibration according to Section~\ref{sec:eval:calib}, we use the same 100 prediction realizations as we did for computing RMSCE, sampled at the training data locations.
Fig. \ref{fig:comp:stochastic:calib} shows the effect of calibration approach and calibration dataset (number of left-out realizations) on RMSCE for  U-PIGAN, U-NNPC, and U-NNPC+.
In some cases, even with a few (2-10) noisy left-out stochastic realizations, we can reduce RMSCE by a factor of 3; see, e.g., Fig.~\ref{fig:comp:stochastic:calib}c. 

%% file: appendix/IN_app_results_forward_PINN.tex
\subsection{Additional results for the forward PDE problem of Section~\ref{sec:comp:pinns:forw}}\label{app:comp:pinns:forw:results}

\begin{figure}[!ht]
	\centering
	\subcaptionbox{}{}{\includegraphics[width=0.32\textwidth]{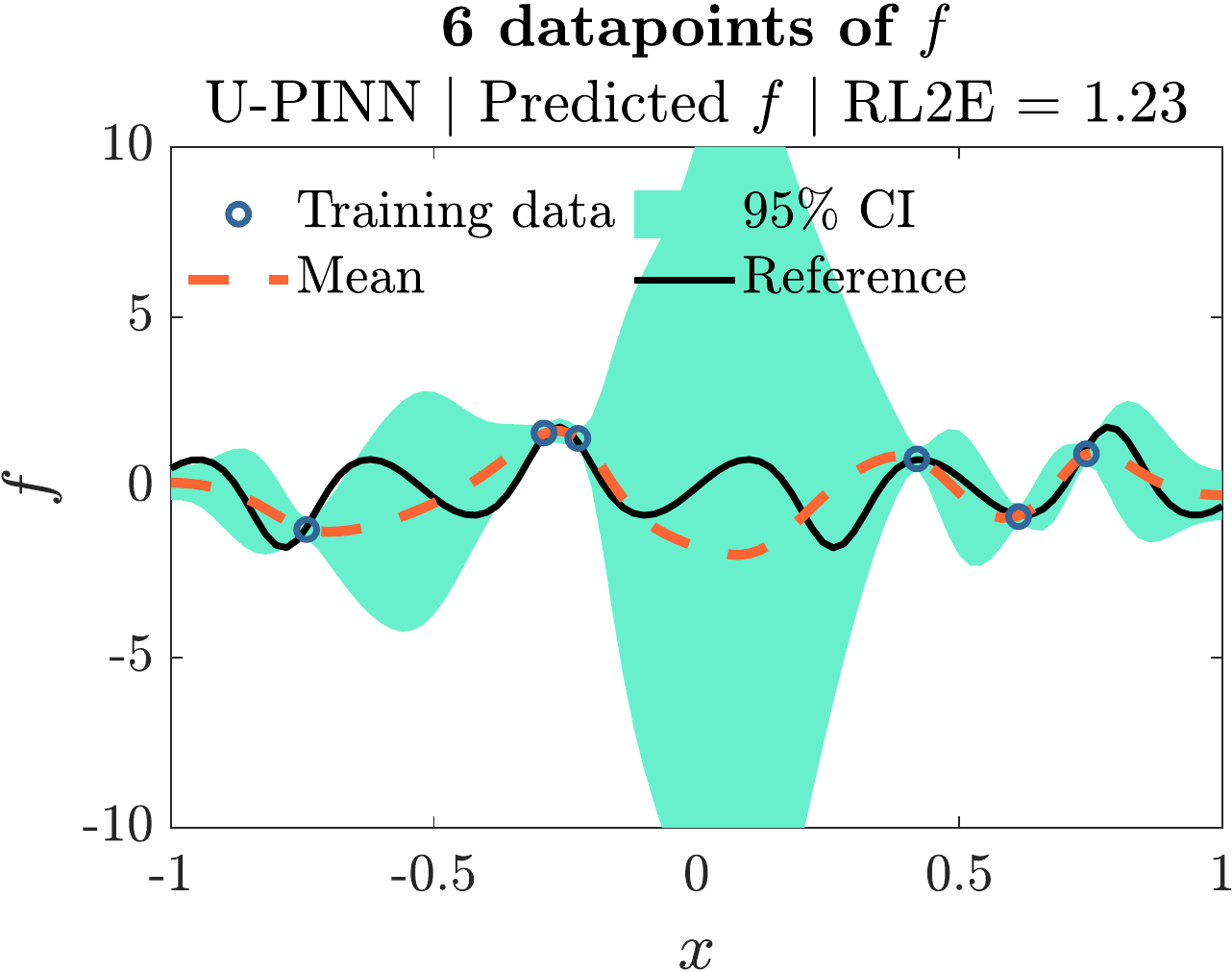}}
	\subcaptionbox{}{}{\includegraphics[width=0.32\textwidth]{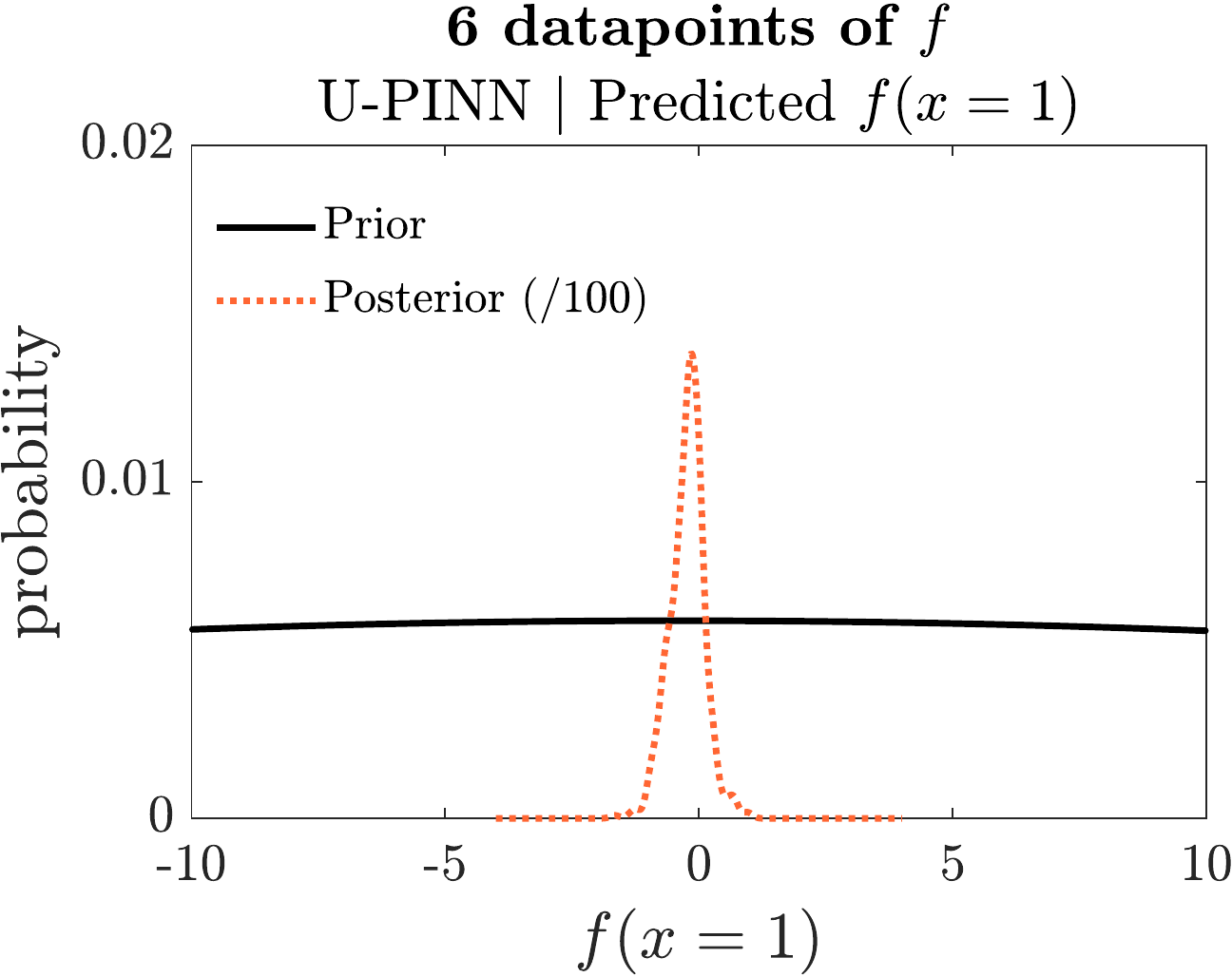}}
	\subcaptionbox{}{}{\includegraphics[width=0.32\textwidth]{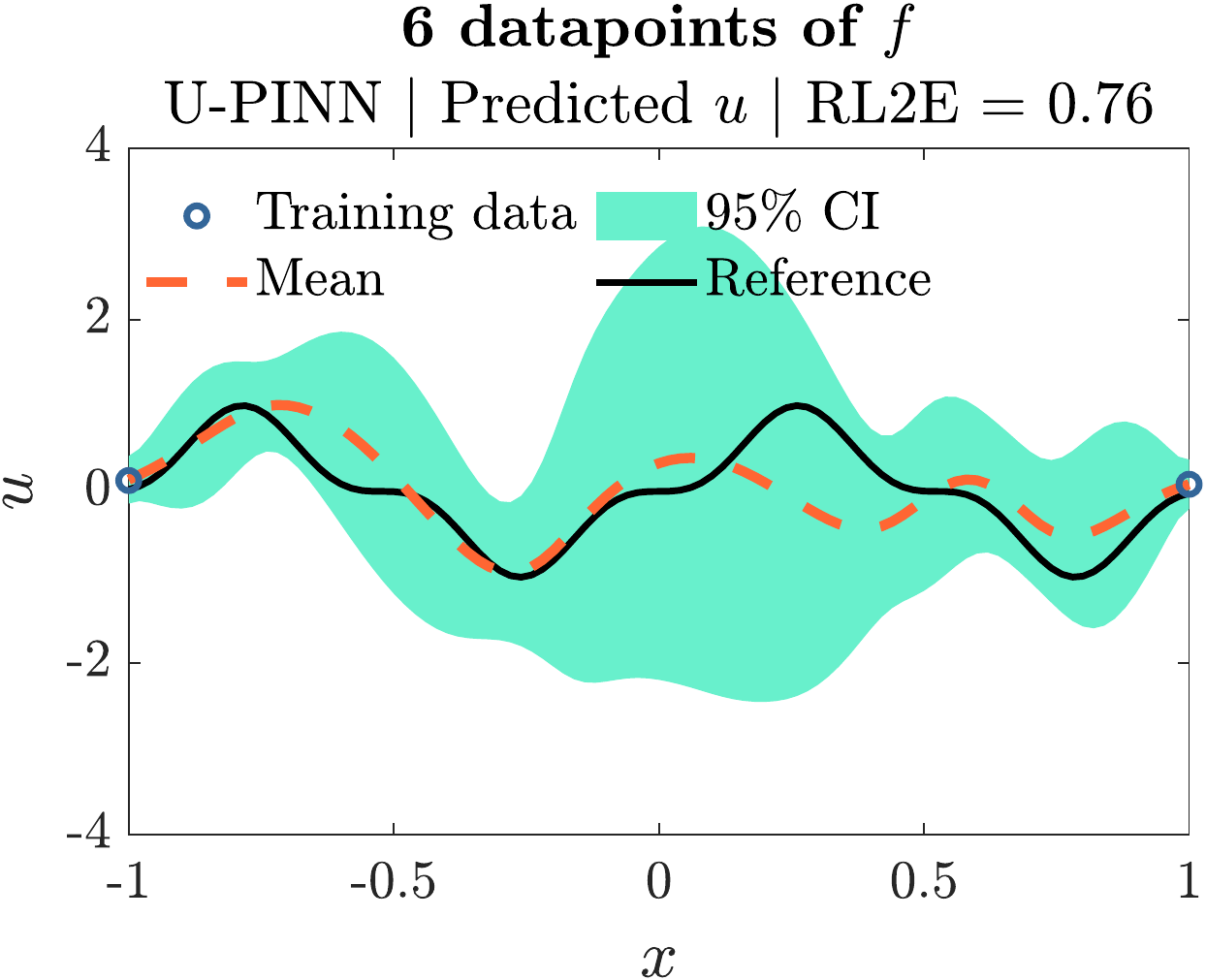}}
	\subcaptionbox{}{}{\includegraphics[width=0.32\textwidth]{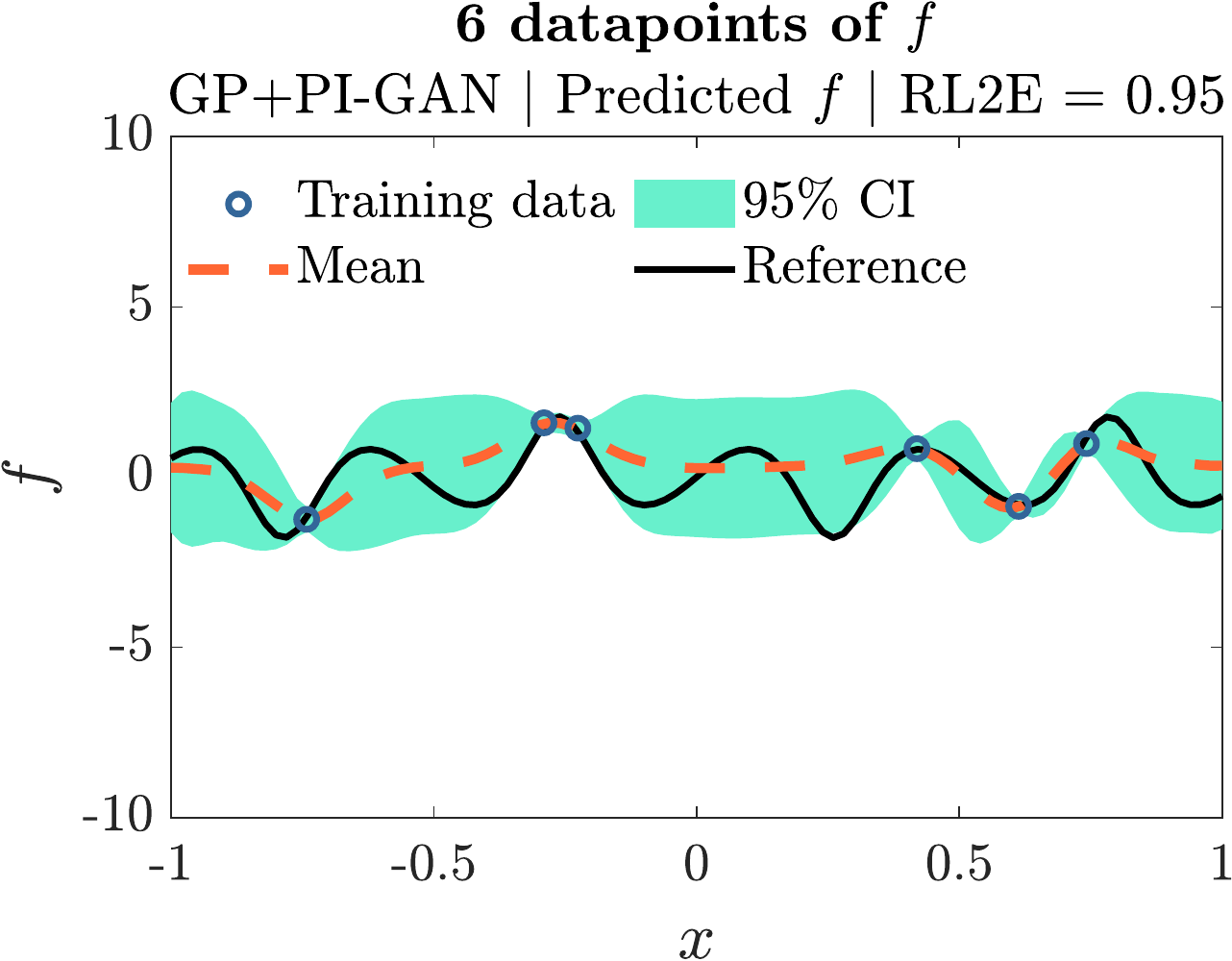}}
	\subcaptionbox{}{}{\includegraphics[width=0.32\textwidth]{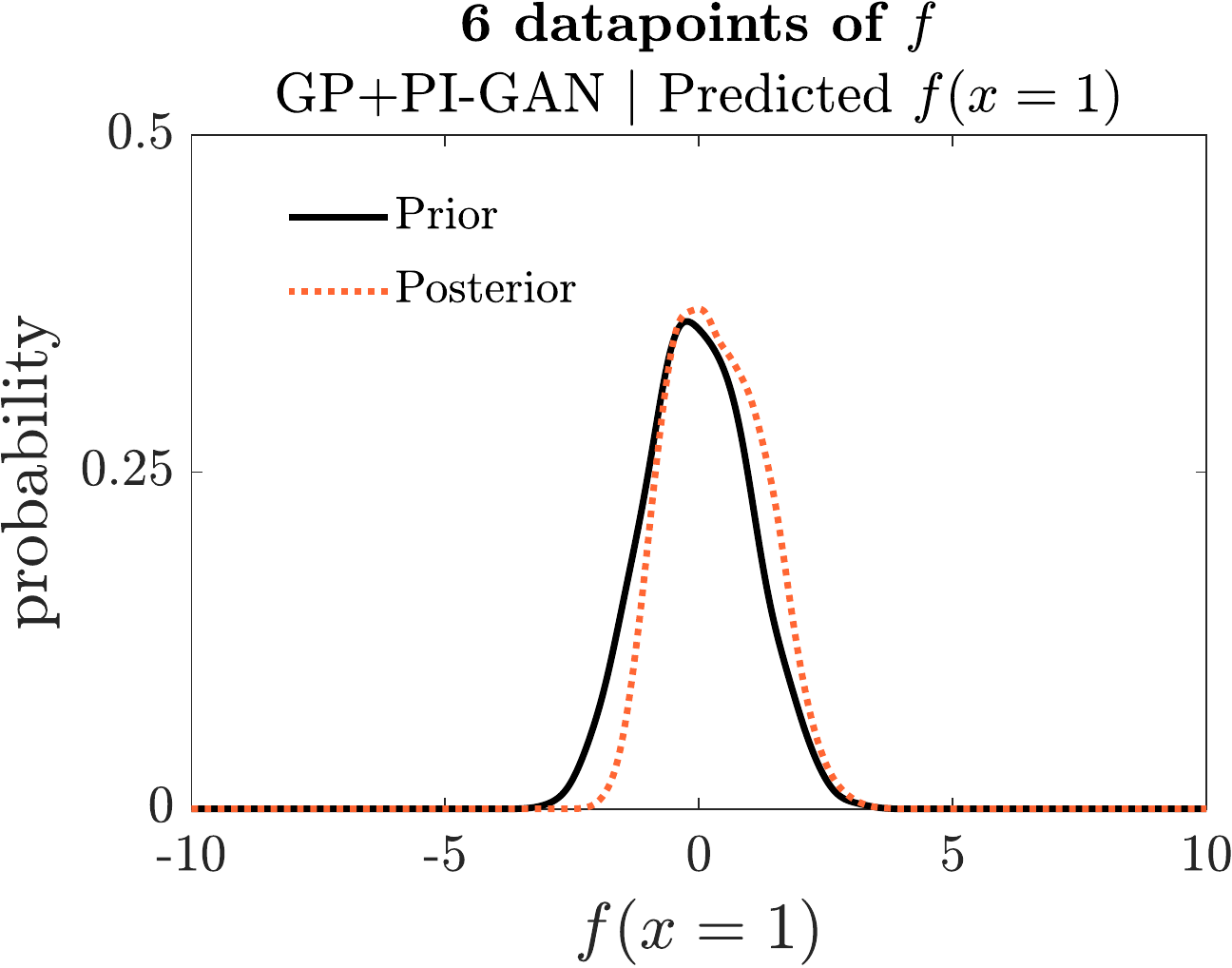}}
	\subcaptionbox{}{}{\includegraphics[width=0.32\textwidth]{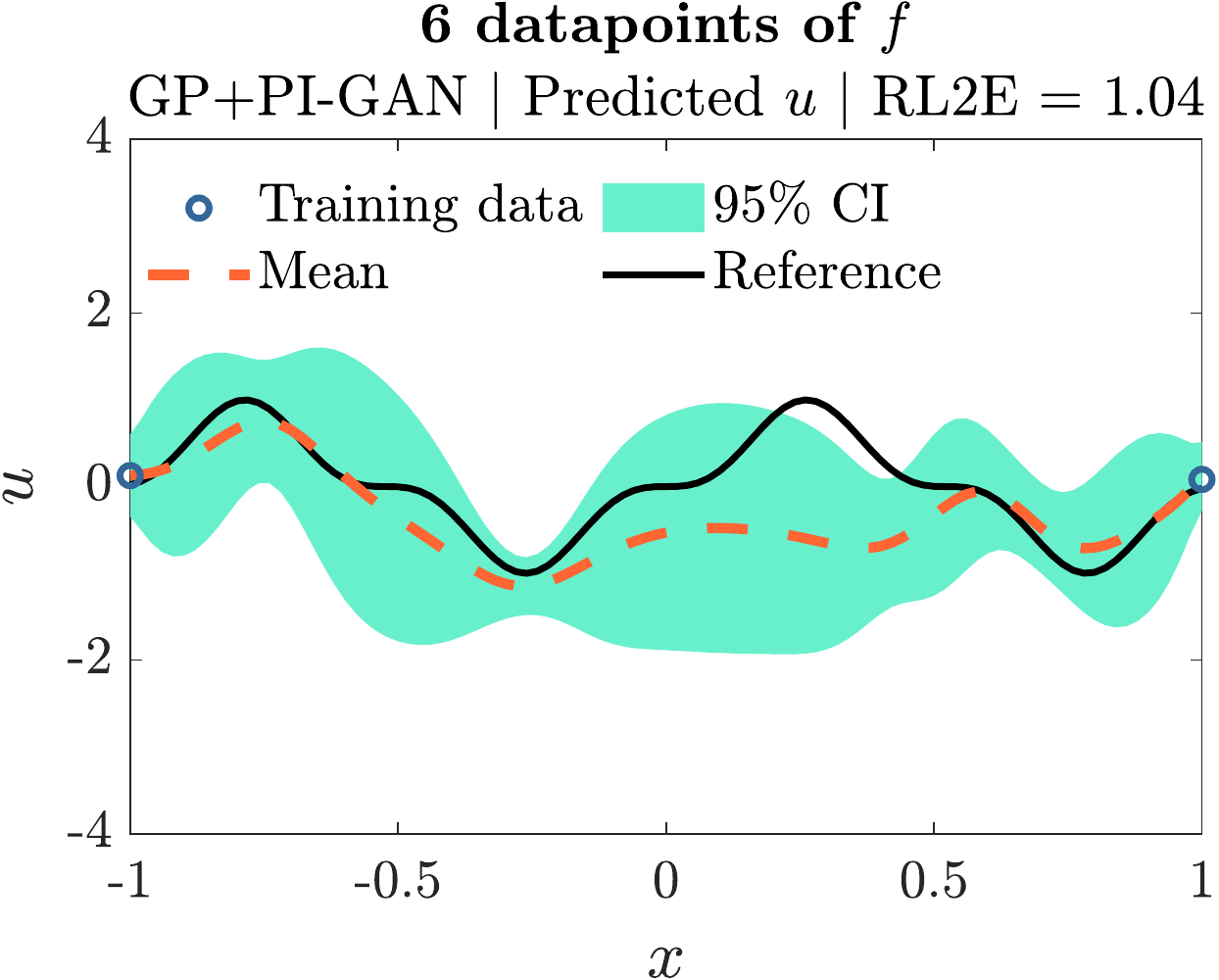}}
	\caption{
		Forward PDE problem of Eq.~\eqref{eq:comp:pinns:forw:pde} | $N_f=6$ \textit{datapoints of} $f$: we compare the U-PINN, which fits $f$ and $u$ simultaneously, with the herein proposed GP+PI-GAN, which first fits $f$ and then $u$.
		In contrast to U-PINN, the uncertainty of $u$ is underestimated by GP+PI-GAN, because the approximations of $u$ and $f$ are not coupled through the PDE of Eq.~\eqref{eq:comp:pinns:forw:pde}.
		Shown here are the training data, reference functions, as well as the mean and epistemic uncertainty ($95 \%$ CI) predictions of U-PINN (top) and GP+PI-GAN (bottom) for $u$ and $f$.
		The middle panels include the prior and posterior distributions of $f$ evaluated at $x=1$, i.e., at the right edge of the input domain where there is no available data for $f$. 
	}
	\label{fig:comp:forw:pinns:6}
\end{figure}

Here we present the results pertaining to the case of $N_f=6$ datapoints for $f$. Specifically, we do not have any measurements of $f$ at the two boundaries, as shown in Fig.~\ref{fig:comp:forw:pinns:6}. It is shown that the predicted uncertainties for $f$ by GP+PI-GAN at the two boundaries are larger than those by U-PINN, since the PDE is not used during training.  Further, the predicted uncertainty of $u$ is underestimated by GP+PI-GAN as compared to the results by U-PINN.

%% file: appendix/IN_app_results_DON.tex
\subsection{Details regarding the DeepONet input for the operator learning problem of Section~\ref{sec:comp:don}}\label{app:comp:don:lognormal}

Assume that $\lambda$ takes only positive values. 
For this reason, we consider that the noisy data of $\lambda$ at each $x, y$ follow a log-normal distribution with $\mathbb{E}[\lambda(x,y)] = \lambda_c(x,y)$ and $Var(\lambda(x,y)) = \alpha^2\lambda_c^2(x,y)$. 
That is, the mean of the data is equal to the true (``clean'') value $\lambda_c(x,y)$ and the data contains an amount of noise proportional to $\lambda_c(x,y)$ at each $(x,y)$.
In order to follow the modeling procedure of Section~\ref{sec:uqt:pre}, we consider $\tilde{\lambda} = \log \lambda$, which follows a Gaussian distribution with mean $\tilde{\lambda}_c = \log \lambda_c - \log(\sqrt{1+\alpha^2})$ and variance $\sigma^2 = \log(1+\alpha^2)$.
Thus, although the original dataset is expressed as $\cD_{\lambda} = \{x_i, y_i, \lambda_{i}\}_{i=1}^{N_{\lambda}}$, we utilize $\cD_{\tilde{\lambda}} = \{x_i, y_i, \tilde{\lambda}_{i}\}_{i=1}^{N_{\lambda}}$, which contains the logarithms of the data.
Suppose also that we standardize the new data using the sample means and standard deviations of the data used for training the DeepONet, on which the logarithm has been applied as well.
Specifically, we subtract from each $\tilde{\lambda}_{i}$ the sample mean $\mu_d(x,y)$ and divide by the sample standard deviation $\sigma_d(x,y)$ for each $(x,y)$.
Then the standardized datapoint $\hat{\lambda}(x,y)$ for each $(x,y)$ follows a Gaussian distribution with
\begin{equation}\label{eq:results:lognormal:stats}
	\begin{split}
		\mathbb{E}[\hat{\lambda}] & = \frac{\tilde{\lambda}_c-\mu_d}{\sigma_d} = \frac{\log \lambda_c - \log(\sqrt{1+\alpha^2}) -\mu_d}{\sigma_d},\\
		Var(\hat{\lambda}) & = \frac{\sigma^2}{\sigma_d^2} = \frac{\log(1+\alpha^2)}{\sigma_d^2}, 
	\end{split}
\end{equation}
where all quantities depend on $(x,y)$.
After fitting the approximator $\lambda_{\theta}$ using the standardized data, predictions for the original quantity $\lambda$ can be obtained via 
\begin{equation}
	\lambda_c(x,y) \approx \exp(\sigma_d(x,y) \lambda_{\theta}(x,y) + \log(\sqrt{1+\alpha^2}) + \mu_d(x,y) ).
\end{equation}
Finally, note that the $(x,y)$-dependent standardizations we applied above limit the approximator $\lambda_{\theta}$ in the sense that it can only be evaluated on the original $(x,y)$ grid used for training the DeepONet.
To avoid this limitation, either the data is not standardized or a shared standardization is used for all points.

\subsection{Additional results for the operator learning problem of Section~\ref{sec:comp:don}}\label{app:comp:don:results:fp}

In this section, we provide representative results pertaining to OOD inference data in operator learning, as well as to a post-training calibration experiment.

\begin{figure}[!ht]
	\centering
	\subcaptionbox{}{}{\includegraphics[width=0.32\textwidth]{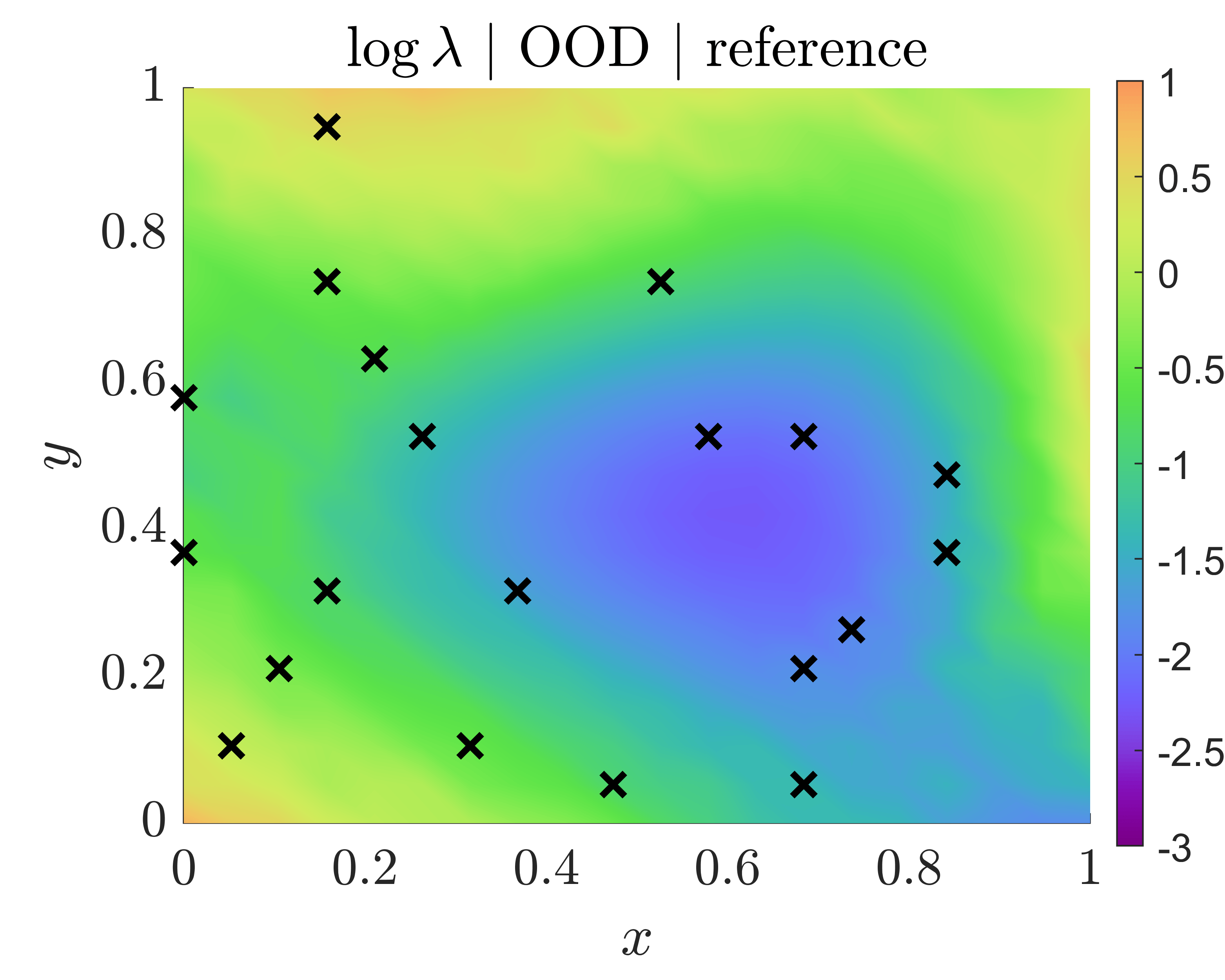}}
	\subcaptionbox{}{}{\includegraphics[width=0.32\textwidth]{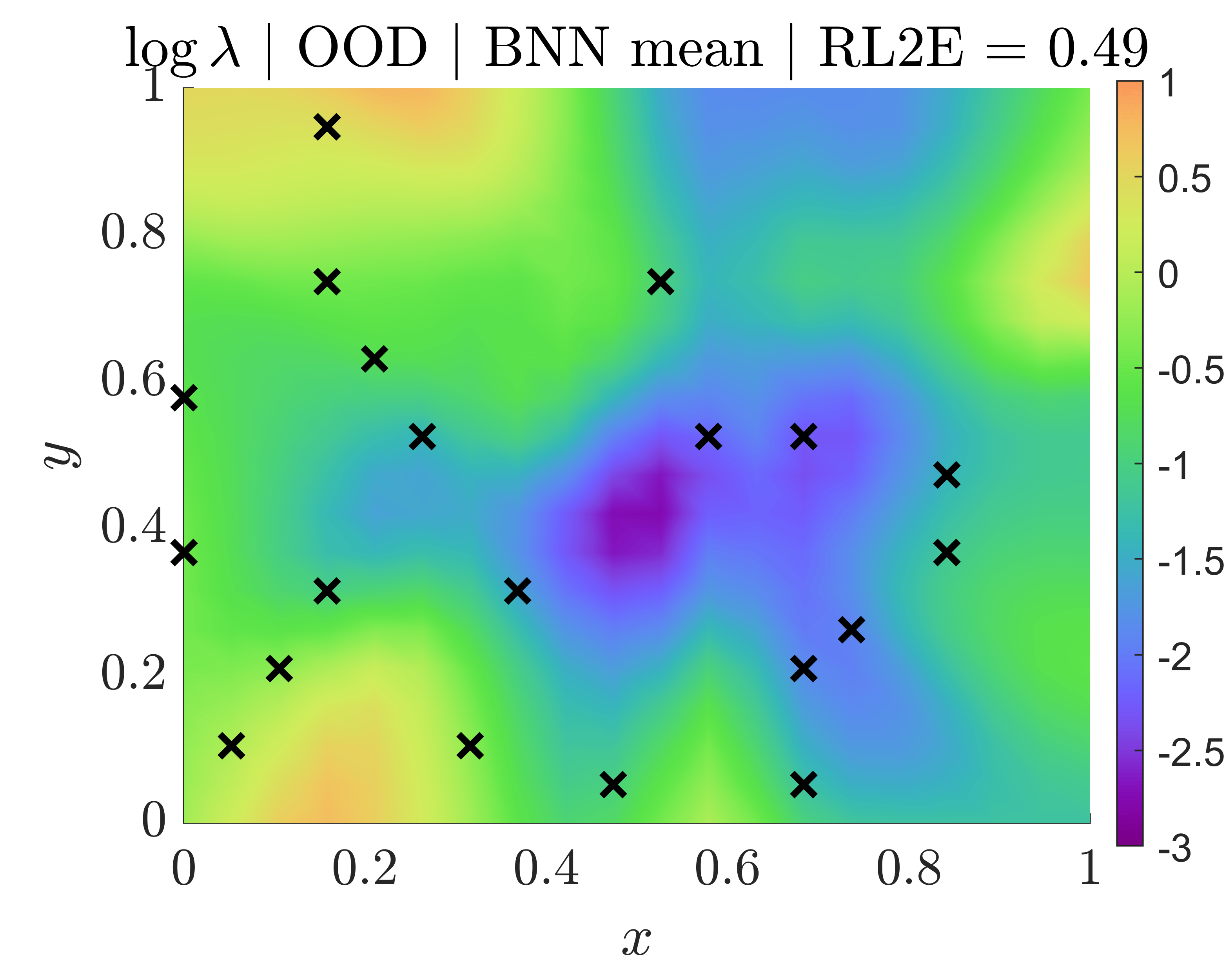}}
	\subcaptionbox{}{}{\includegraphics[width=0.32\textwidth]{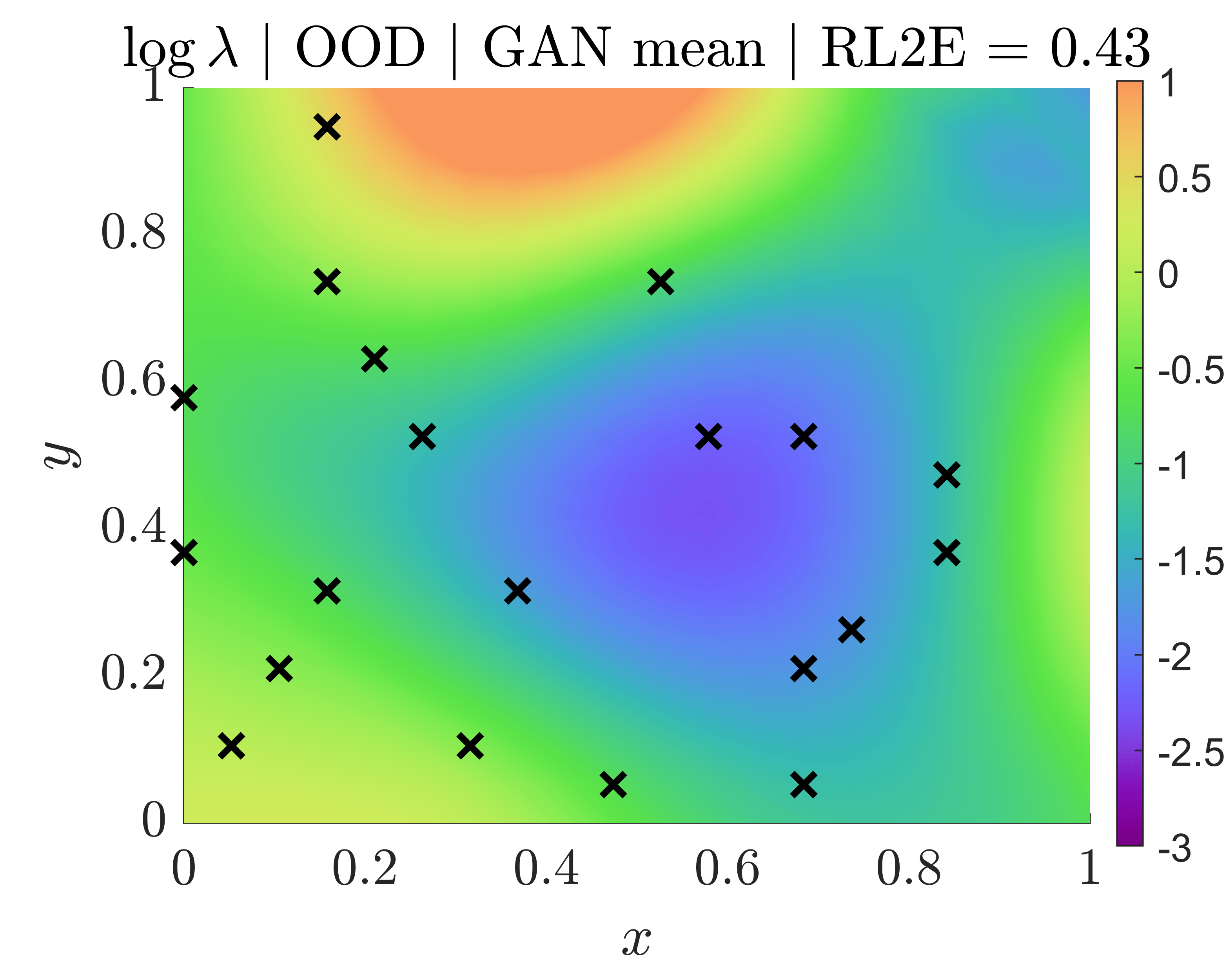}}
	\subcaptionbox{}{}{\includegraphics[width=0.32\textwidth]{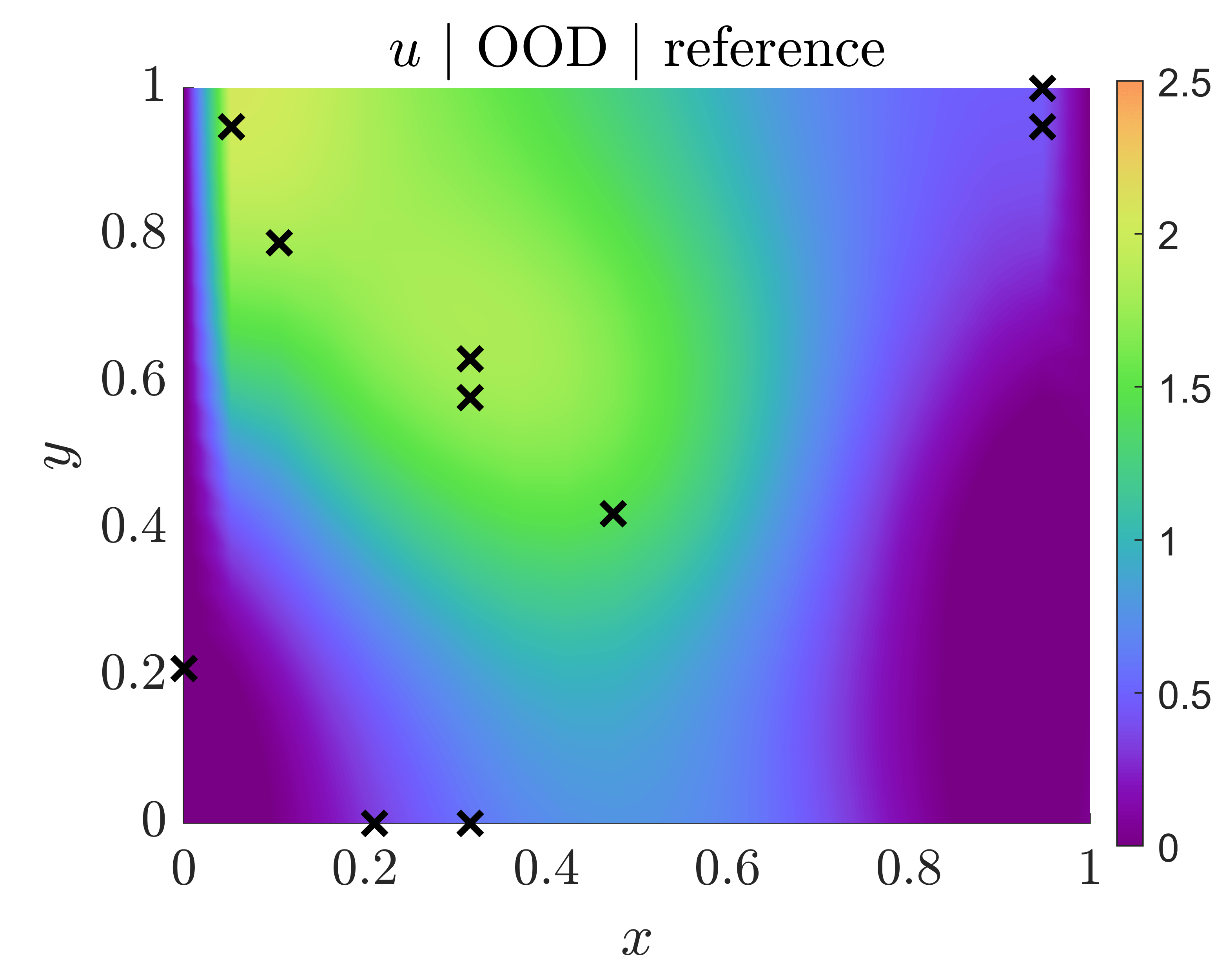}}
	\subcaptionbox{}{}{\includegraphics[width=0.32\textwidth]{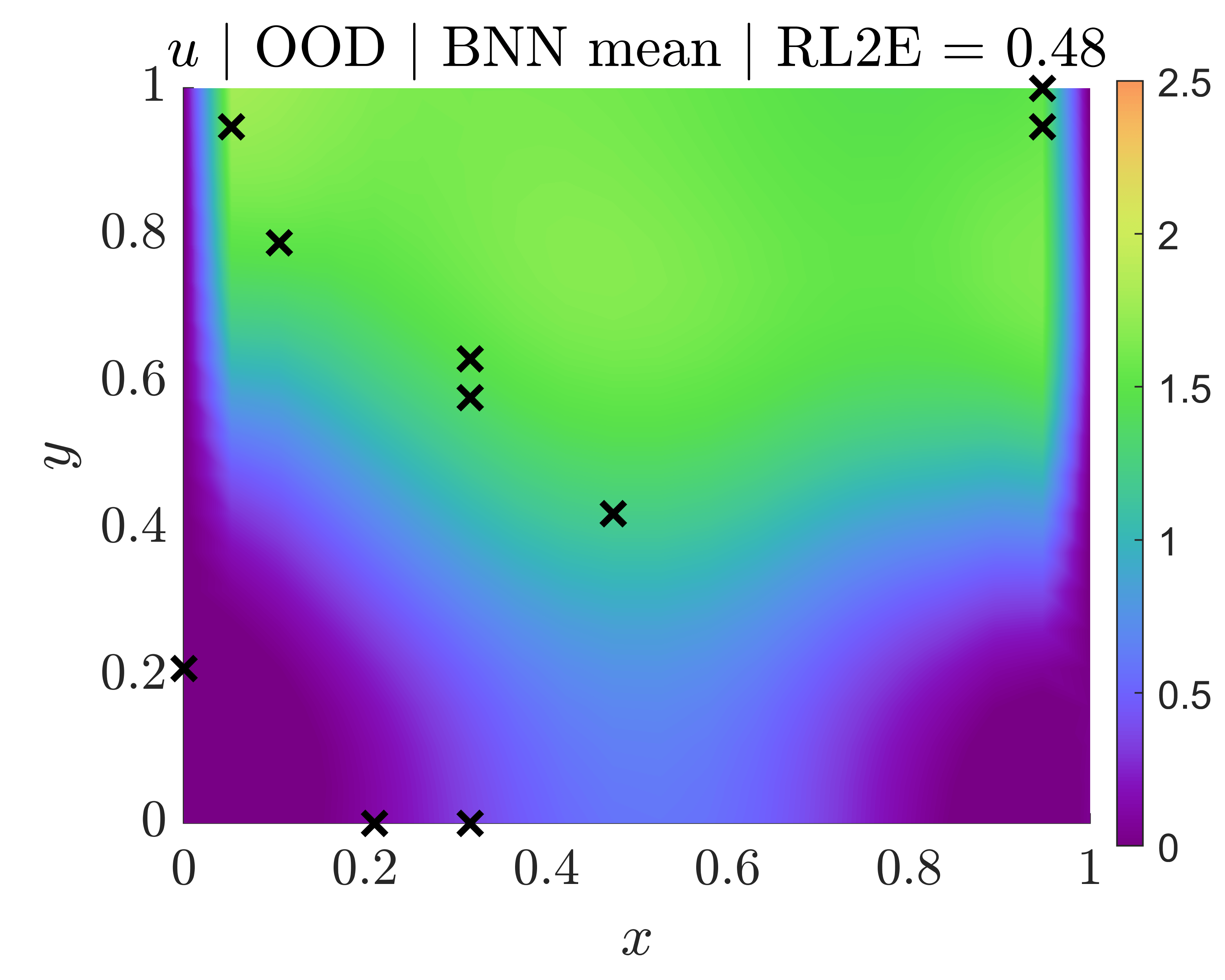}}
	\subcaptionbox{}{}{\includegraphics[width=0.32\textwidth]{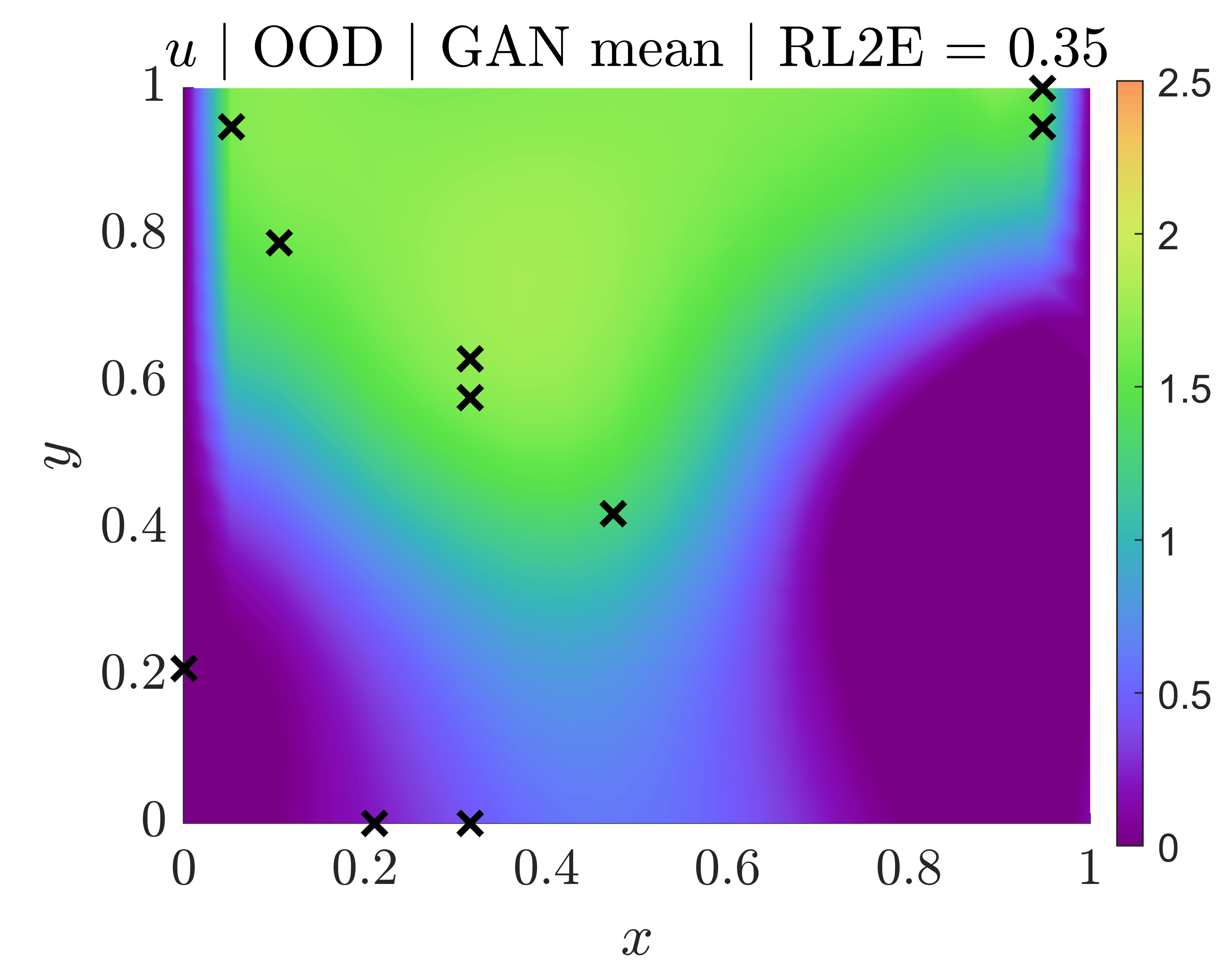}}
	\caption{
		Operator learning problem of Eqs.~\eqref{eq:comp:don:pde}-\eqref{eq:comp:don:bcs} | \textit{Limited and noisy inference data of $\lambda$ and $u$ (OOD)}:
		the performance of both PA-BNN-FP and PA-GAN-FP deteriorates significantly for OOD data with RL2E values close to 40-50$\%$.
		Nevertheless, PA-BNN-FP is more calibrated than PA-GAN-FP, as indicated by the RMSCE values of Table~\ref{tab:comp:don:fp}.
		This shows that PA-GAN-FP that is pre-trained with historical data of $\lambda$ produces confident predictions even for cases of unseen $\lambda$ during pre-training.
		Shown here are the reference input $\tilde{\lambda} = \log \lambda$ and solution $u$, the corresponding inference data locations (x markers), as well as the mean predictions by PA-BNN-FP and PA-GAN-FP.
		Top row results correspond to input $\log \lambda$, whereas bottom row to solution $u$.
	}
	\label{fig:comp:don:ood:gan_bnn_mean}
\end{figure}

\begin{figure}[!ht]
	\centering
	\subcaptionbox{}{}{\includegraphics[width=0.24\textwidth]{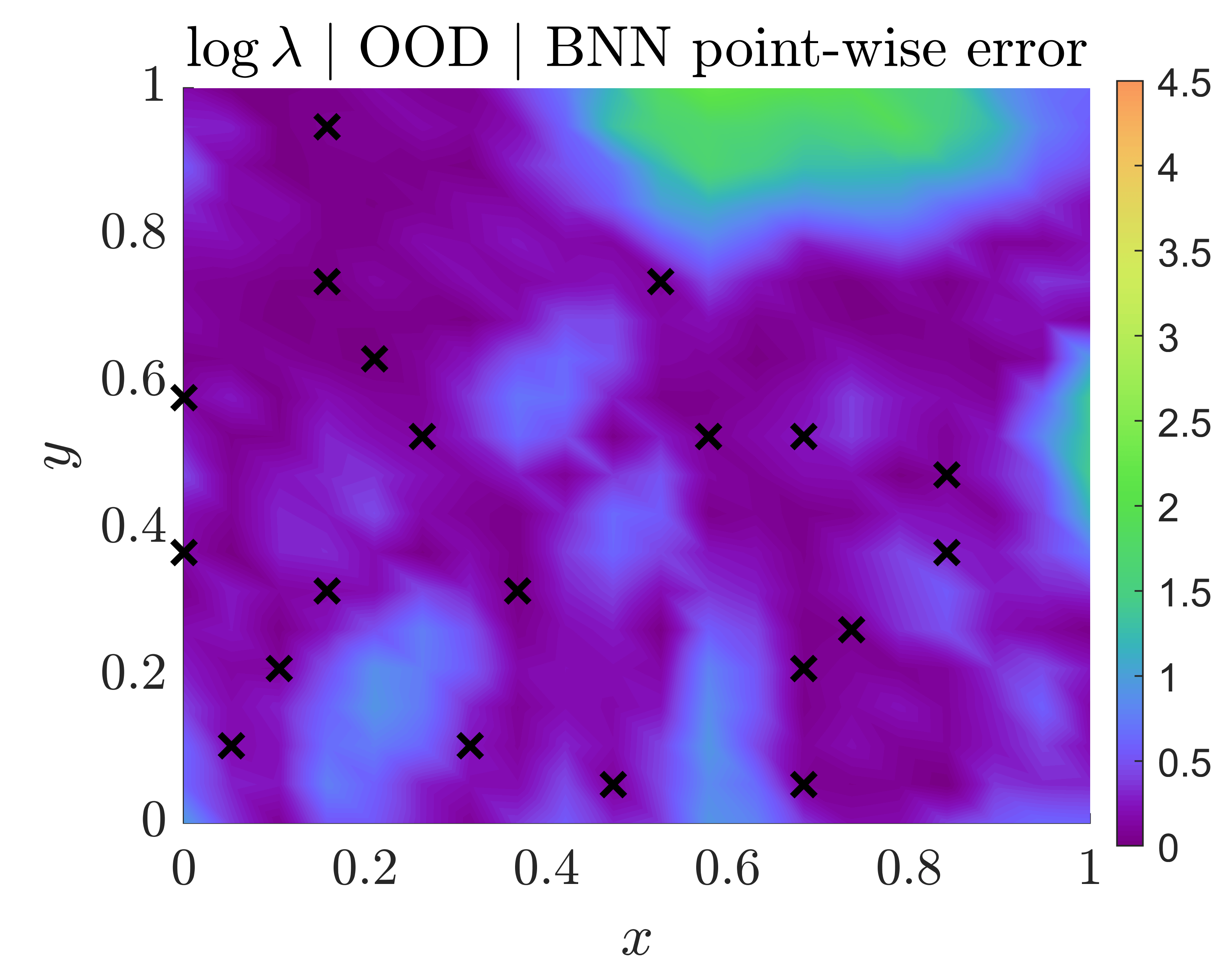}}
	\subcaptionbox{}{}{\includegraphics[width=0.24\textwidth]{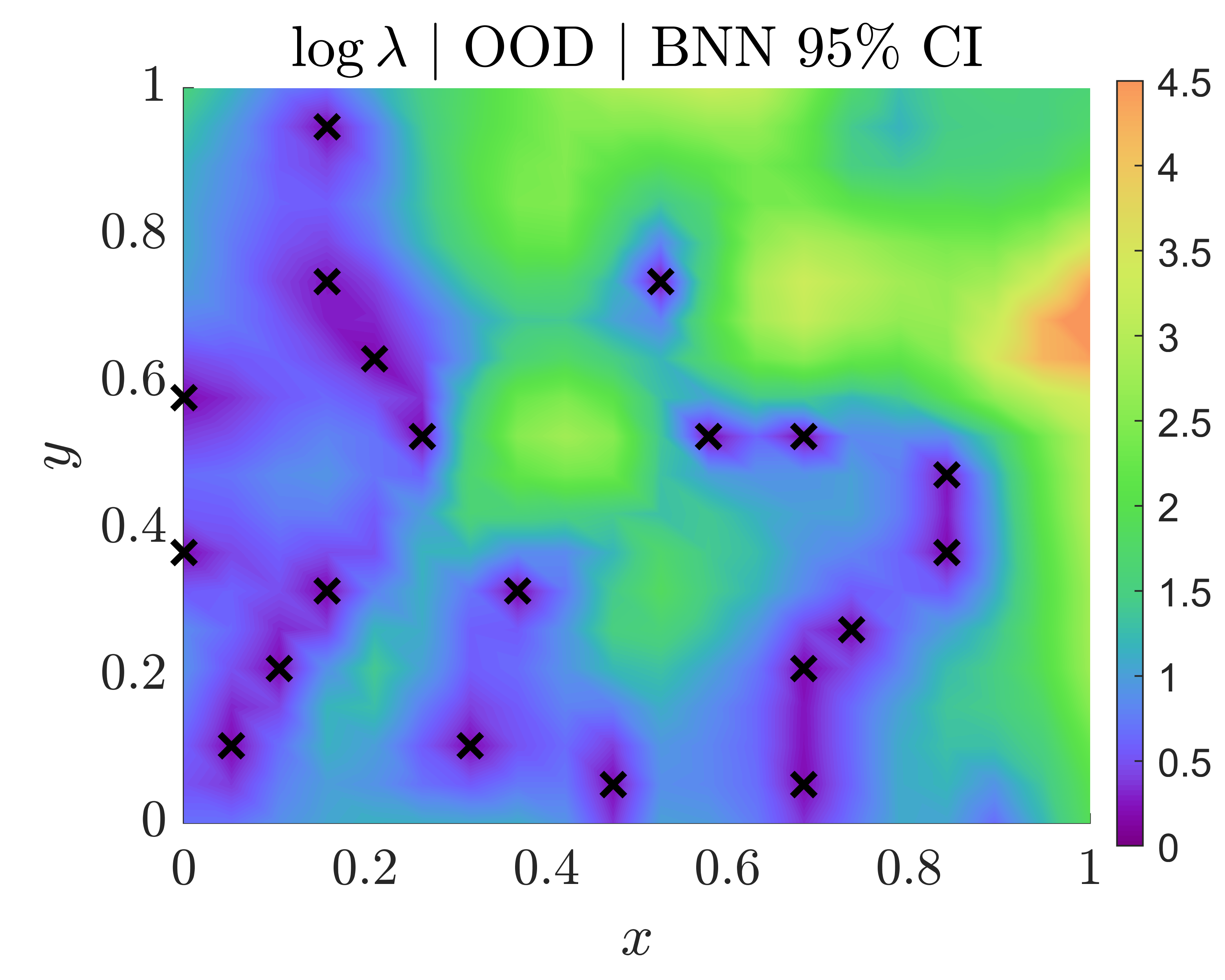}}
	\subcaptionbox{}{}{\includegraphics[width=0.24\textwidth]{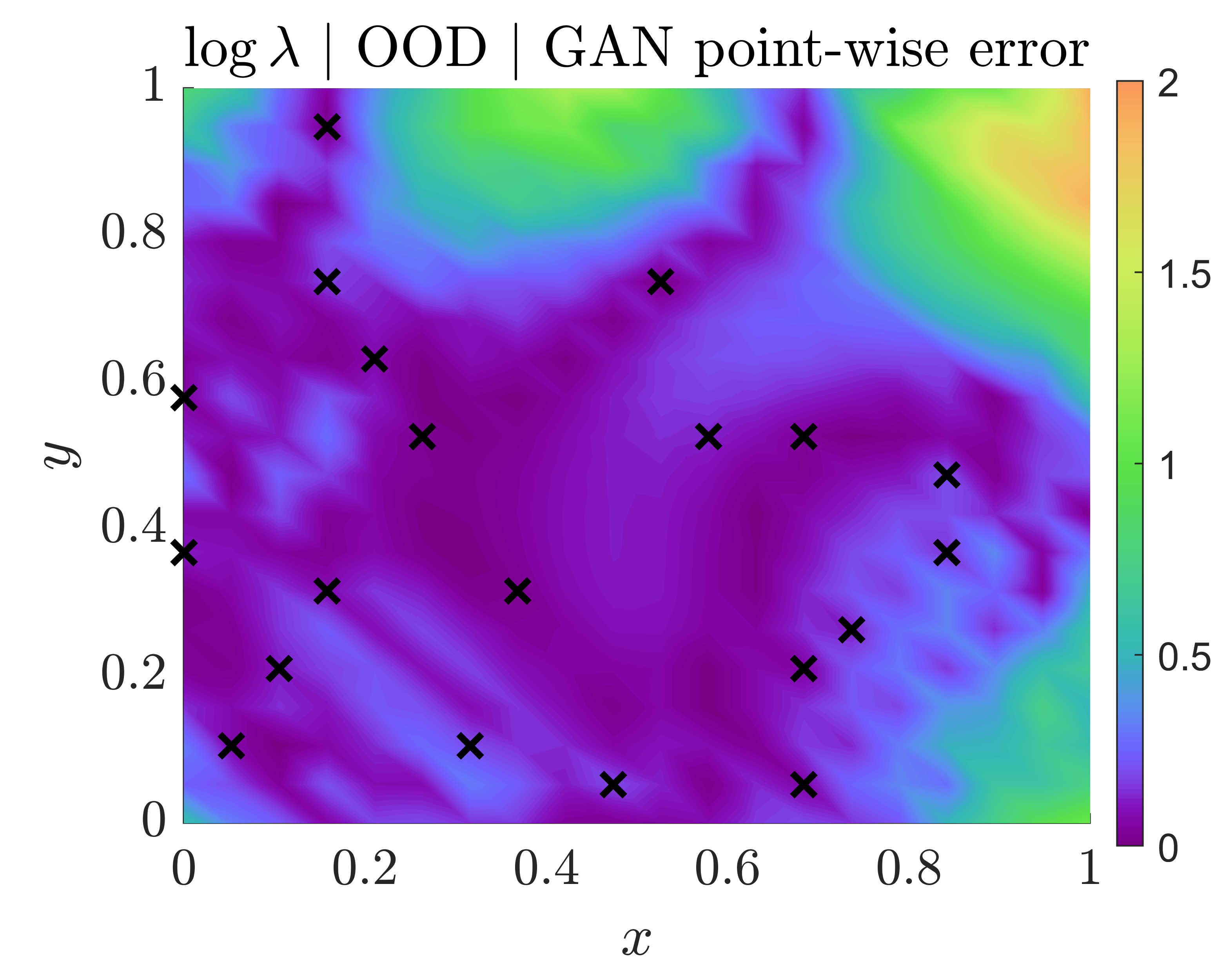}}
	\subcaptionbox{}{}{\includegraphics[width=0.24\textwidth]{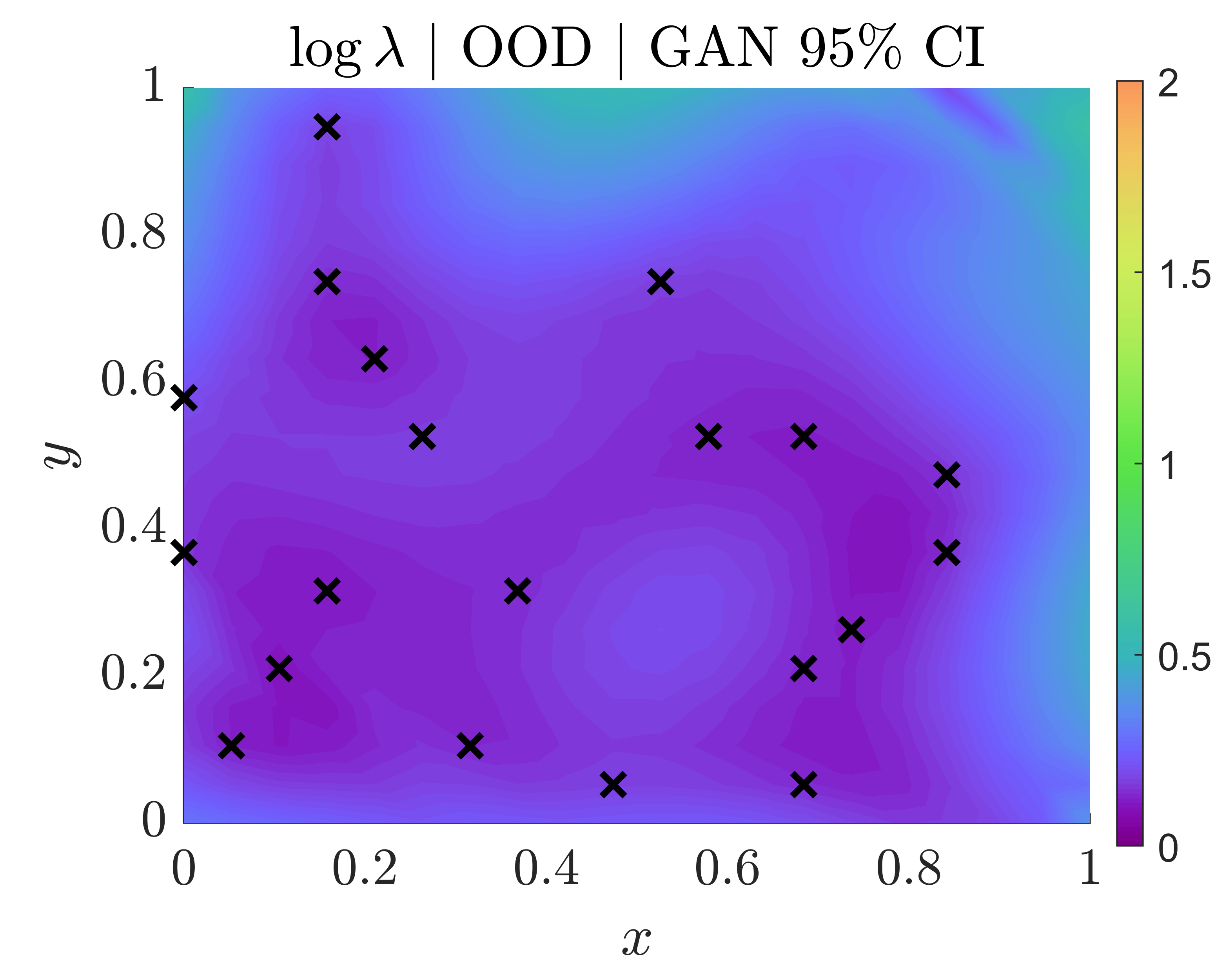}}
	\subcaptionbox{}{}{\includegraphics[width=0.24\textwidth]{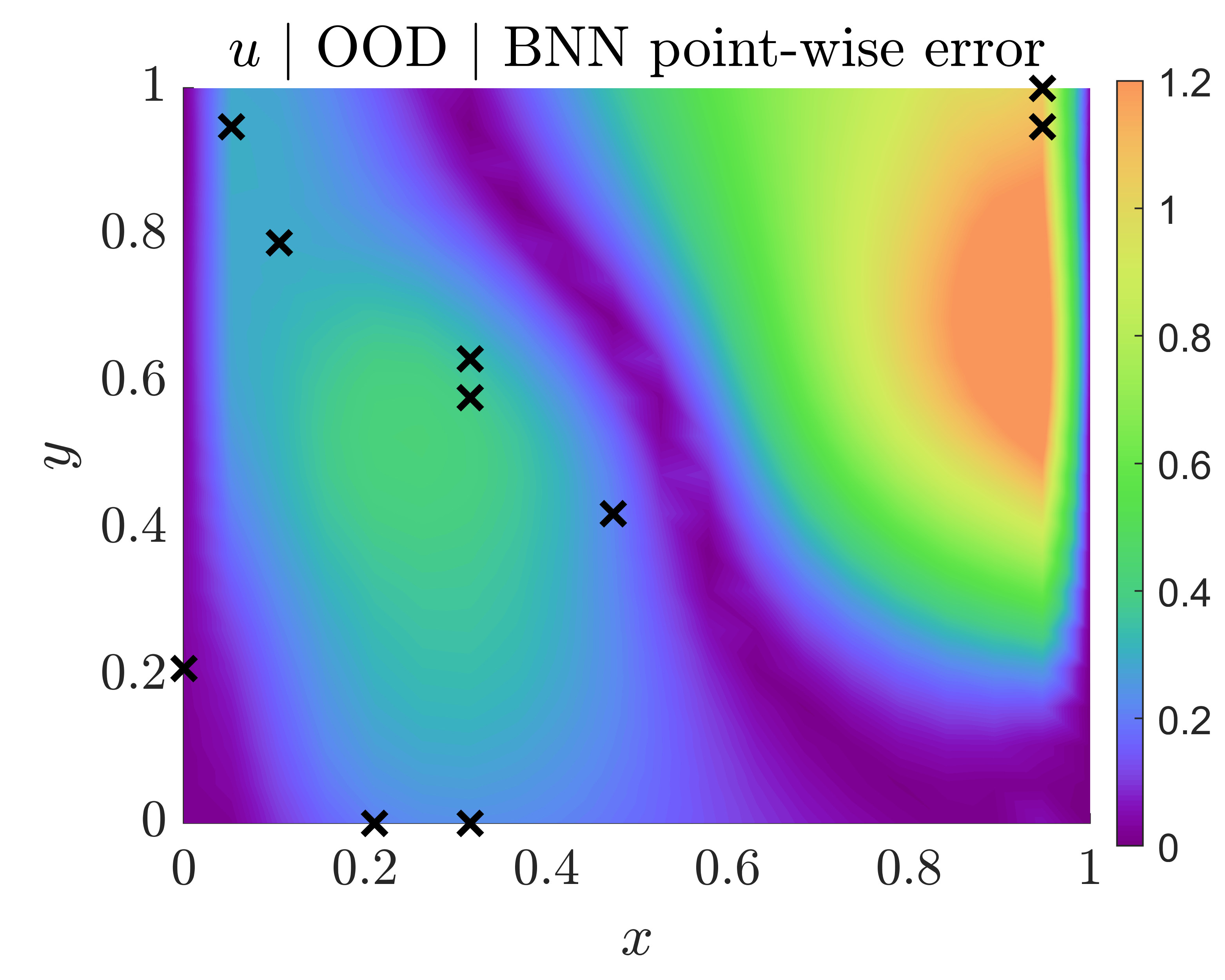}}
	\subcaptionbox{}{}{\includegraphics[width=0.24\textwidth]{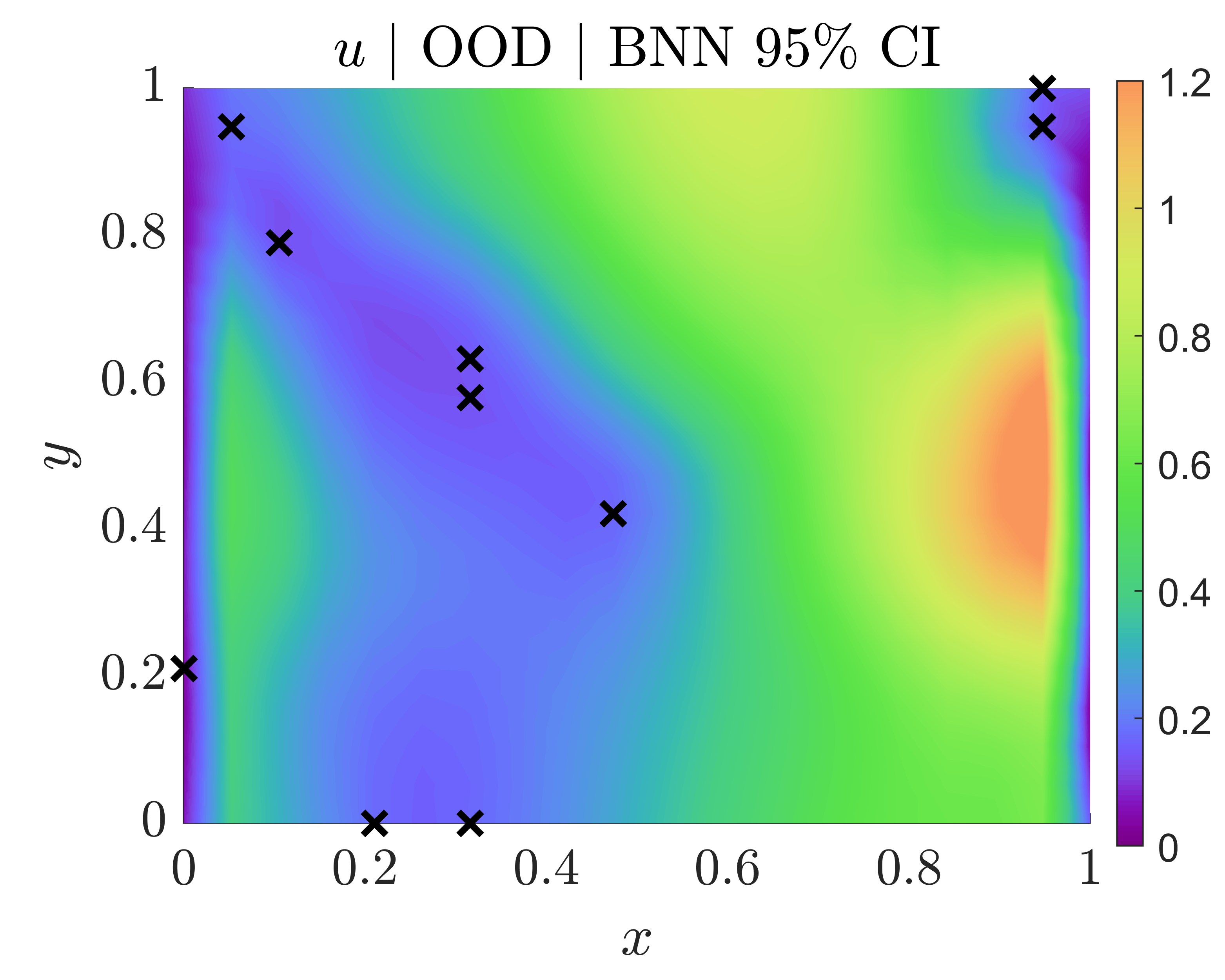}}
	\subcaptionbox{}{}{\includegraphics[width=0.24\textwidth]{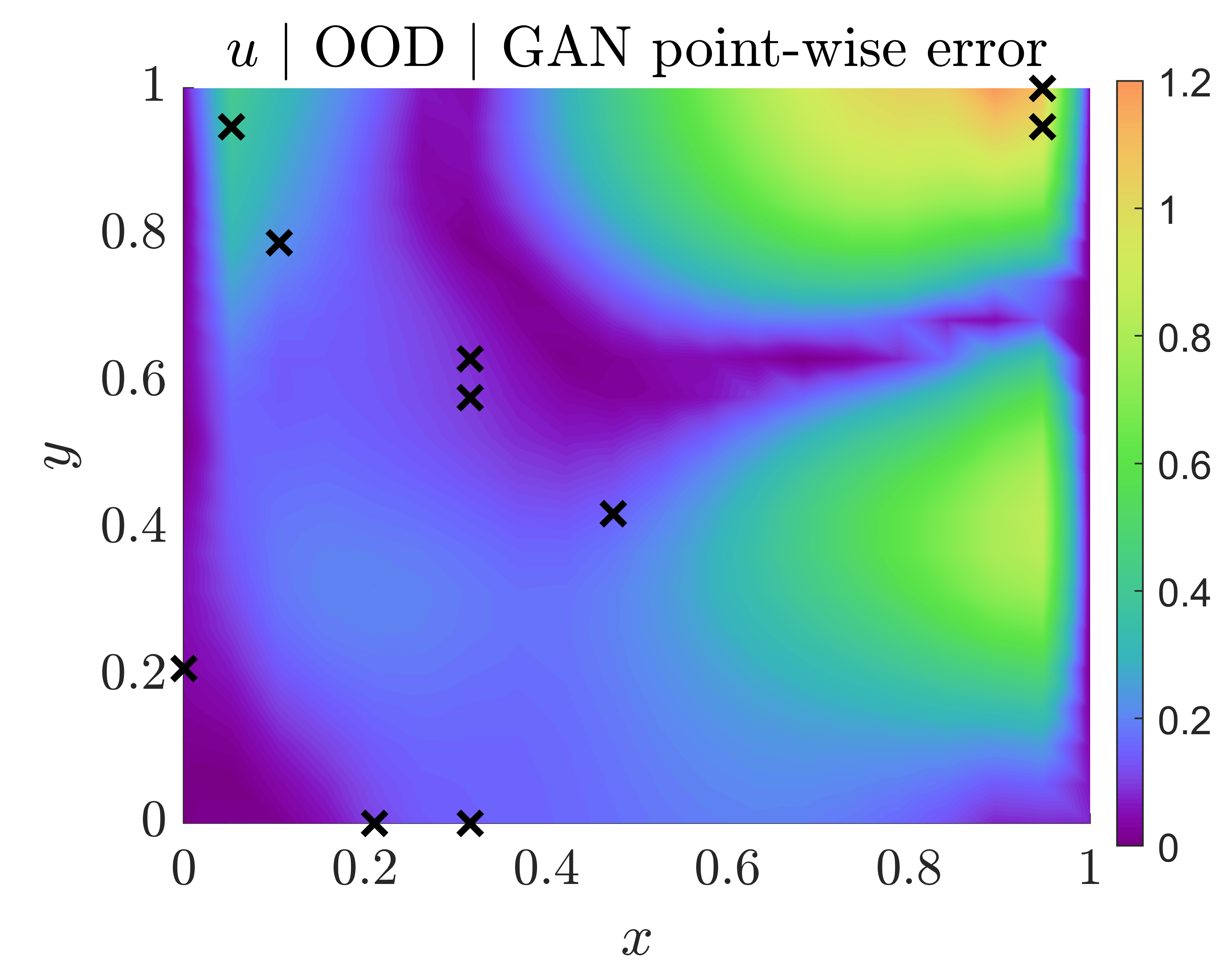}}
	\subcaptionbox{}{}{\includegraphics[width=0.24\textwidth]{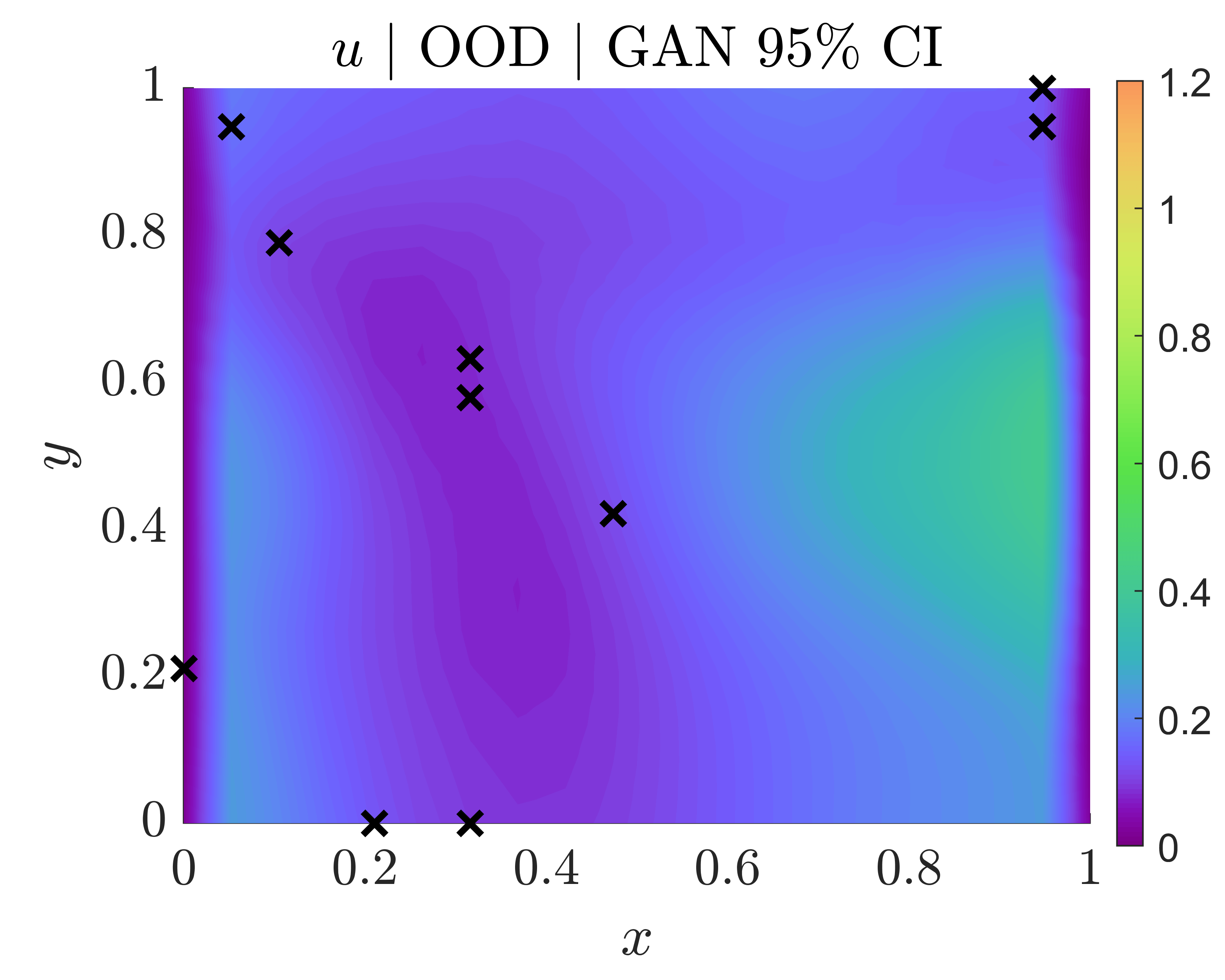}}
	\caption{
		Operator learning problem of Eqs.~\eqref{eq:comp:don:pde}-\eqref{eq:comp:don:bcs} | \textit{Limited and noisy inference data of $\lambda$ and $u$ (ID)}:
		the performance of both PA-BNN-FP and PA-GAN-FP deteriorates significantly for OOD data with RL2E values close to 40-50$\%$.
		Nevertheless, PA-BNN-FP is more calibrated than PA-GAN-FP, as indicated by the RMSCE values of Table~\ref{tab:comp:don:fp}.
		In this figure, we notice that PA-GAN-FP is over-confident and produces small epistemic uncertainties.
		This shows that PA-GAN-FP that is pre-trained with historical data of $\lambda$ produces confident predictions even for cases of unseen $\lambda$ during pre-training.
		Shown here are the absolute point-wise error and epistemic uncertainty predictions corresponding to input $\tilde{\lambda} = \log\lambda$ (a-d) and solution $u$ (e-f), as obtained by comparing PA-BNN-FP and PA-GAN-FP with reference solutions, as well as the corresponding inference data locations (x markers).
	}
	\label{fig:comp:don:ood:gan_bnn_std}
\end{figure}

In particular, we generate a sample for $\lambda$ from the Gaussian process with $l = 0.2$ in Eq. \eqref{eq:comp:don:gp}. The corresponding reference solution for $u$ is, subsequently, obtained by using the {\emph{Matlab PDE toolbox}}. 
Similarly to the ID evaluation performed in Section~\ref{sec:comp:don}, we assume that we have partial noisy measurements of $\lambda$ and $u$, and aim to reconstruct the fields for $\lambda$ and $u$. Specifically, we randomly select 20 and 10 data from the reference solutions of $\lambda$ and $u$, respectively, and then contaminate them with the same Gaussian noise as used in Section \ref{sec:comp:don:fp}. The results obtained by PA-BNN-FP and PA-GAN-FP are presented in Figs. \ref{fig:comp:don:ood:gan_bnn_mean}-\ref{fig:comp:don:ood:gan_bnn_std}. It is shown that (1) the RL2E values for the predicted means of $\lambda$ and $u$ by PA-BNN-FP and PA-GAN-FP are much larger than those in Fig. \ref{fig:comp:don:fp:mean:id} (ID evaluation), (2) the computational errors for $\lambda$ and $u$ are not covered within the $95\%$ CI in PA-GAN-FP, and (3) the computational errors are mostly covered within the $95\%$ CI in PA-BNN-FP, which may be attributed to the fact that the employed BNN here has a larger prior space than the GAN. More details on the detection of OOD data in operator learning are presented in Section \ref{sec:comp:don:ood}.

\begin{figure}[H]
	\centering
	\subcaptionbox{}{}{\includegraphics[width=0.24\textwidth]{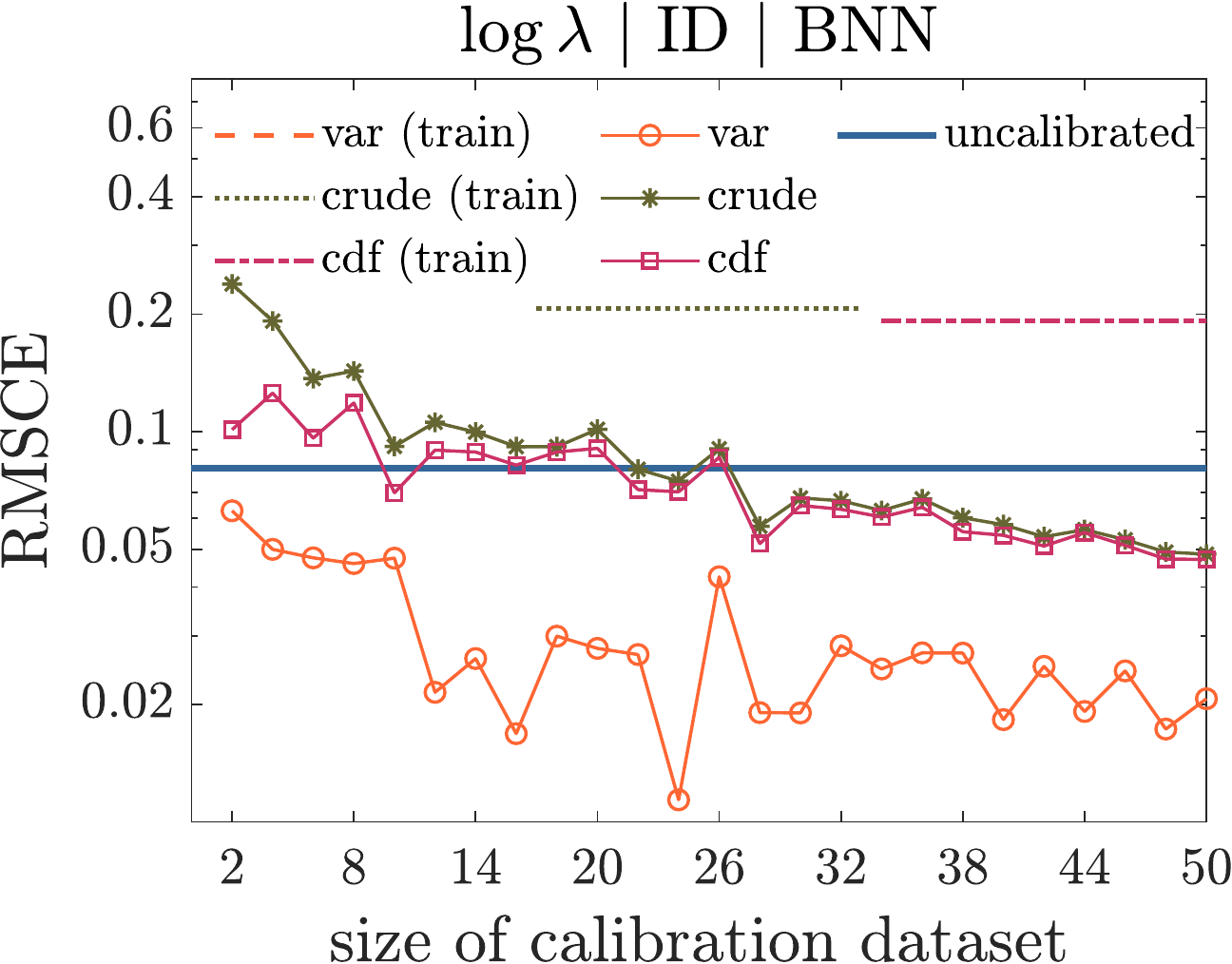}}
	\subcaptionbox{}{}{\includegraphics[width=0.24\textwidth]{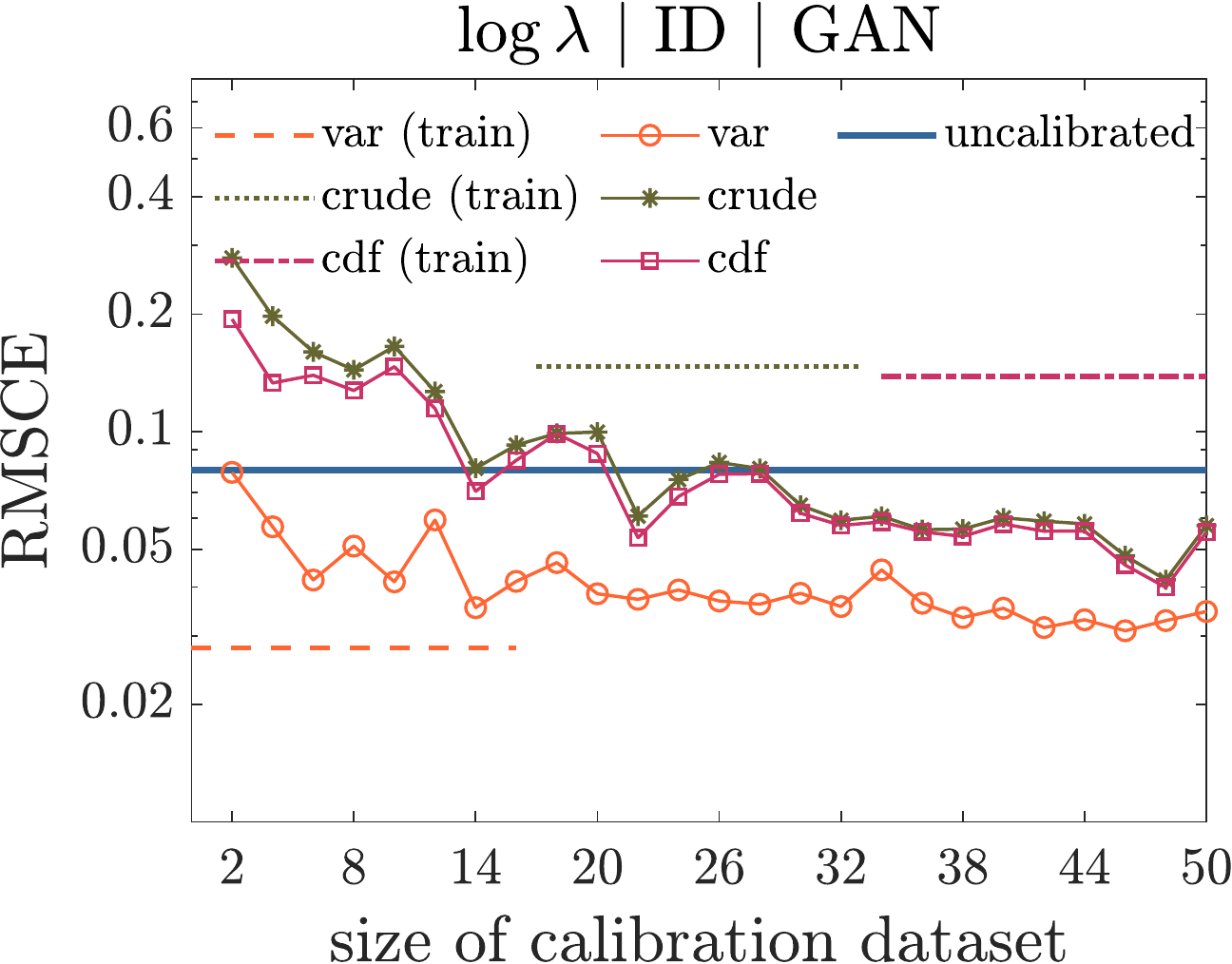}}
	\subcaptionbox{}{}{\includegraphics[width=0.24\textwidth]{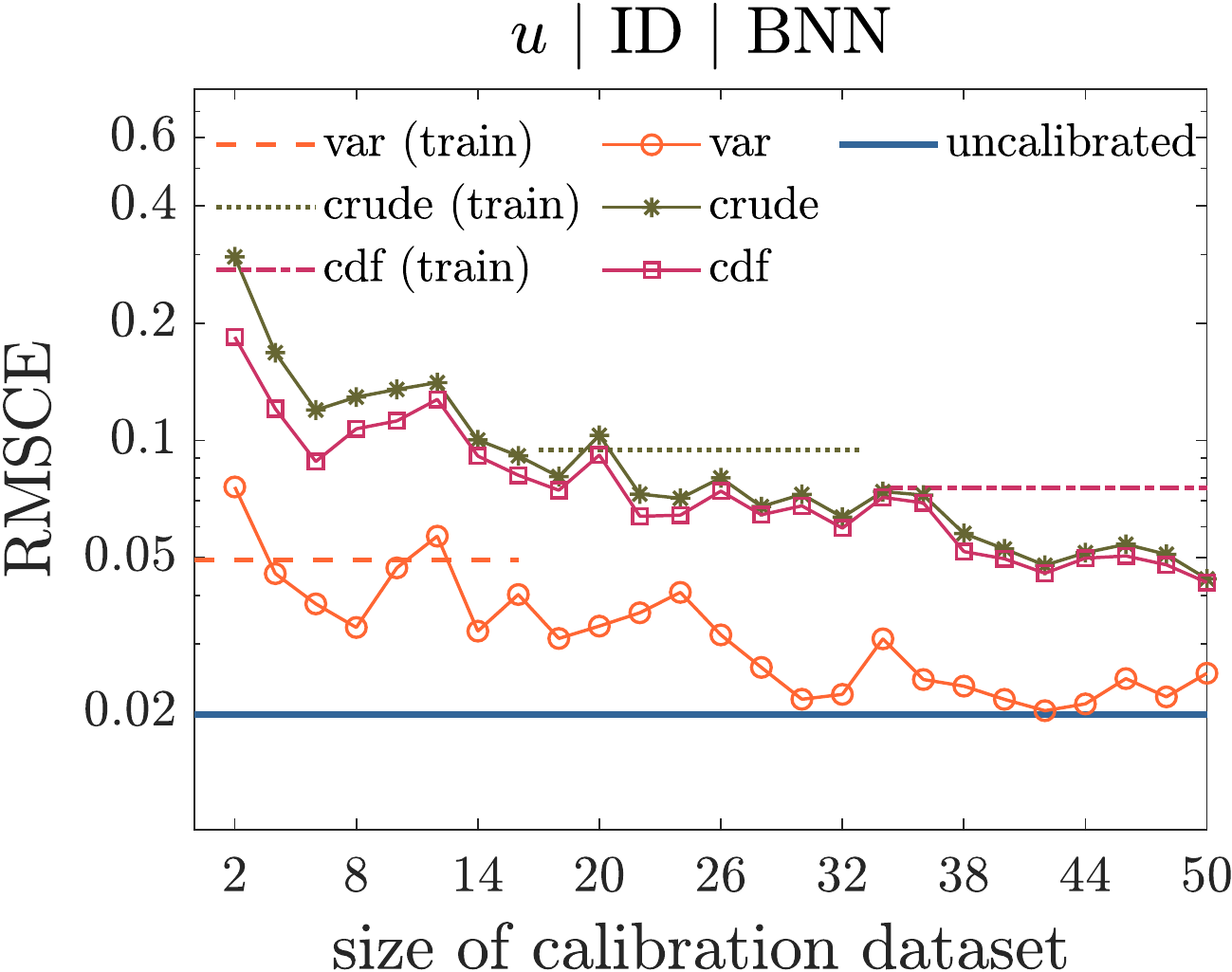}}
	\subcaptionbox{}{}{\includegraphics[width=0.24\textwidth]{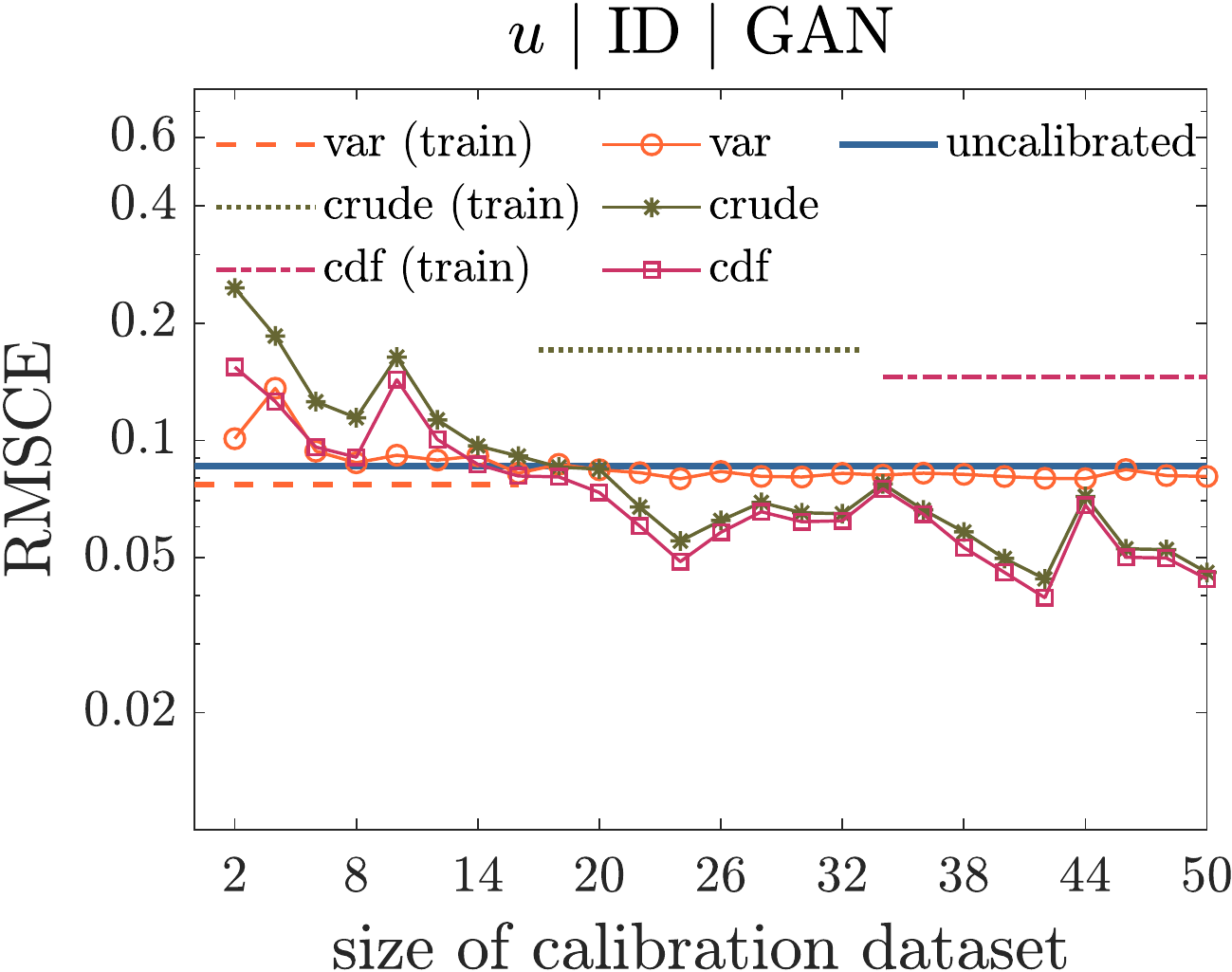}}
	\subcaptionbox{}{}{\includegraphics[width=0.24\textwidth]{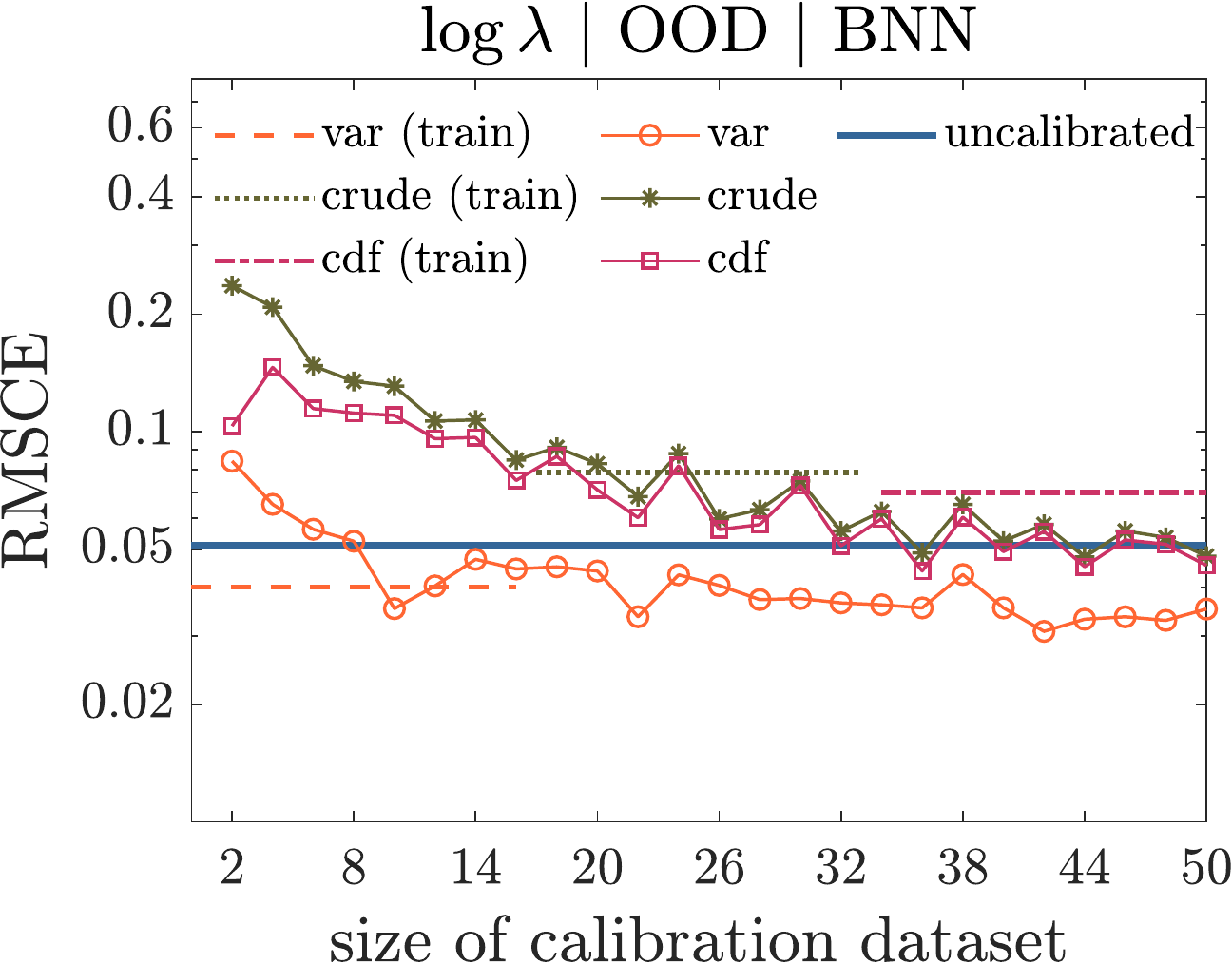}}
	\subcaptionbox{}{}{\includegraphics[width=0.24\textwidth]{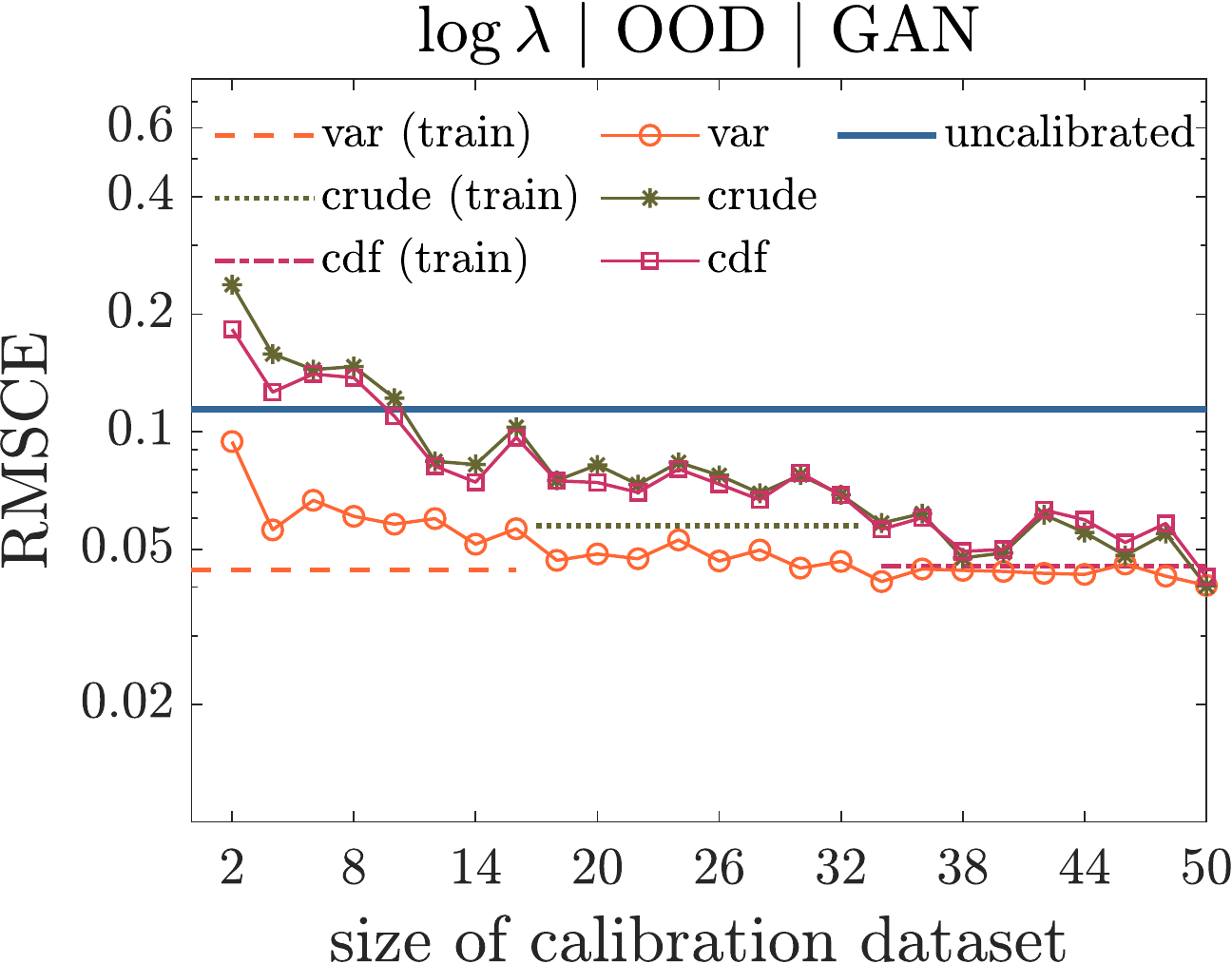}}
	\subcaptionbox{}{}{\includegraphics[width=0.24\textwidth]{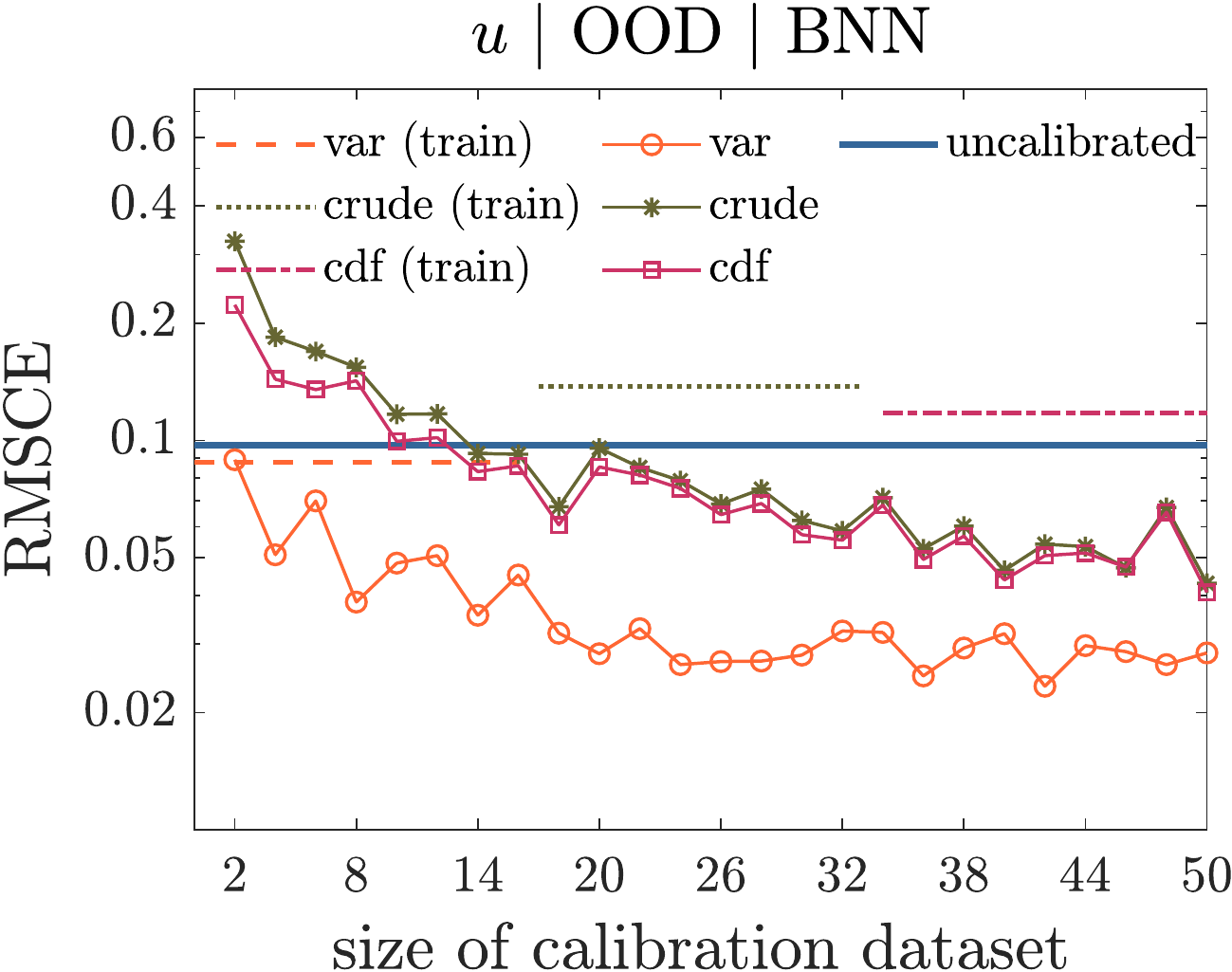}}
	\subcaptionbox{}{}{\includegraphics[width=0.24\textwidth]{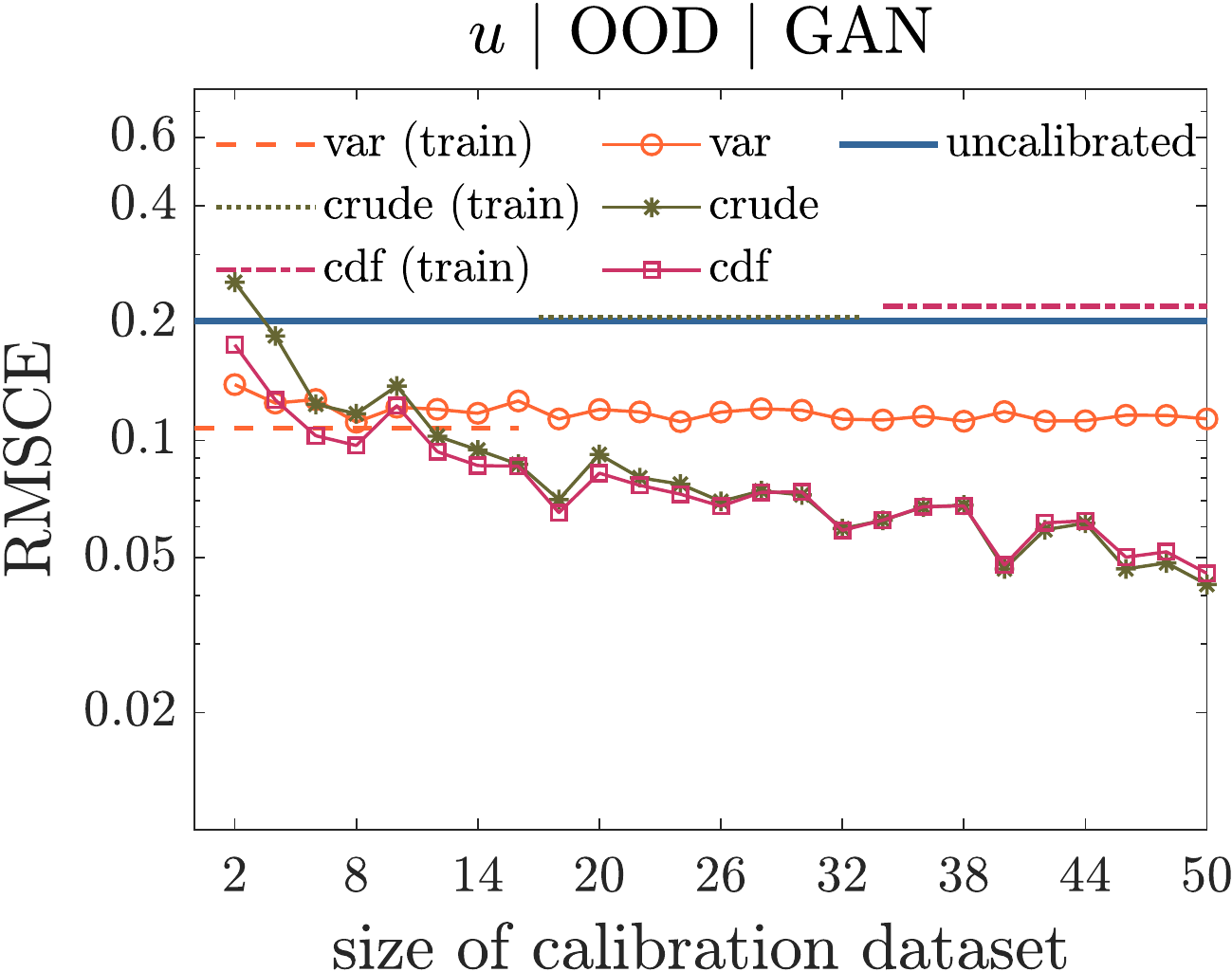}}
	\caption{
		Operator learning problem of Eqs.~\eqref{eq:comp:don:pde}-\eqref{eq:comp:don:bcs} | \textit{Limited and noisy inference data of $\lambda$ and $u$ (ID or OOD)}:
		performing post-training calibration even with a few left-out noisy datapoints (2-10) can reduce the calibration error by half in some cases.
		This is true especially for OOD data and for using the variance re-weighting technique.
		Shown here is the effect of calibration approach (Section~\ref{sec:eval:calib}) and calibration dataset on RMSCE for $\tilde{\lambda} = \log \lambda$ and $u$ predictions using PA-BNN-FP and PA-GAN-FP.
		Note, however, that there are cases in which calibration can make the predictions worse (e.g., part c of the figure), because it calibrates ``on average'', as explained in Section~\ref{sec:eval:calib}.
	}
	\label{fig:comp:don:fp:calib}
\end{figure}

Further, in Fig.~\ref{fig:comp:don:fp:calib} we perform post-training calibration using 2-50 left-out noisy data from $\lambda$ and $u$ during inference. Clearly, because the training data of $\lambda$ and $u$ were much less than 50, the objective of this experiment is only to illustrate the trend of decreasing RMSCE with increasing calibration dataset size and compare the three calibration methods. It is observed that post-training calibration even with a few datapoints (2-10) can reduce the calibration error by half in some cases. 
An interesting future research direction is to perform a cost-effectiveness study for comparing post-training calibration with active learning, i.e., with techniques that re-adjust the mean and uncertainty predictions in light of new data.

%% file: appendix/IN_app_hyperparameters.tex
\subsection{Hyperparameters used in the comparative study}
\label{app:comp:hyperparameters}

In this section, we provide all hyperparameters used in the comparative study. 
In Table \ref{tab:app:glossary_hyper} we provide the glossary for the hyperparameters. Note that the Adam optimizer in {\emph{Tensorflow}} is employed for optimization by default in the present study. In the Adam optimizer, we set the exponential decay rates for the first (i.e., $\beta_1$) and second (i.e., $\beta_2$) moment estimates in the training of GANs/PI-GANs as 0.5 and 0.9, respectively. They are set to the default values, i.e., $\beta_1 =0.9$ and $\beta_2 = 0.999$, in the training of standard NNs.
In Tables \ref{tab:comp:func:homosc:kno}-\ref{tab:comp:func:results:student}, we present the hyperparameters used for obtaining the results of Section \ref{sec:comp:func}. Specifically, we note that: (1) all results from HMC without Gibbs sampling in this study are implemented using the open-source package {\emph{tensorflow probability}} \cite{lao2020tfp}. In addition, we employ HMC with adaptive step size to make the computations more robust. For cases using HMC with Gibbs sampling, i.e., Figs.~\ref{fig:comp:func:homosc:kno:netsize} and \ref{fig:comp:func:homosc:unk:dists}, the step size is fixed throughout the whole HMC and Gibbs sampling process. We manually tune the step size to achieve the desired acceptance rate (around 0.6) in these two cases; and (2) for MFVI, we obtain the optimal training steps for early stopping based on the negative-log-likelihood of the validation data used in each case of Section \ref{sec:comp:func}. See more details regarding early stopping in Section \ref{app:methods:bnns:vi}. 
The hyperparameters used for implementing the methods employed in Sections~\ref{sec:comp:pinns:hetero} and \ref{app:comp:pinns:results:large}-\ref{app:comp:pinns:results:steep} are provided in Tables \ref{tab:app:comp:pinns:stand} and \ref{tab:app:comp:pinns:additional}.
The hyperparameters used for implementing the methods employed in Section \ref{sec:comp:stochastic} are provided in Table \ref{tab:app:comp:stochastic}.
The hyperparameters used for implementing the methods employed in Section \ref{sec:comp:pinns:forw} are provided in Table \ref{tab:app:comp:pinns:forw}.
The hyperparameters used for implementing the various methods employed in Section \ref{sec:comp:don} are provided in Tables \ref{tab:app:comp:don:fp} and \ref{tab:app:comp:don:dens}.

\begin{table}[H]
	\centering
	\footnotesize
	\begin{tabular}{r|l||r|l}
		\toprule
		\multicolumn{4}{c}{\textbf{Glossary of hyperparameters used in the comparative study}}\\
		\midrule
		$\epsilon$& learning rate or time step in leapfrog method &$T$& leapfrog steps\\
		$M$&number of posterior samples&wd& weight decay constant\\
		init&initialization method&train steps& number of training iterations \\	
		$\epsilon_{init}$&maximum learning rate (beginning of cycle)&dr& dropout rate \\
		$T_{cycles}$&number of cycles &$\epsilon_{final}$& minimum learning rate (end of cycle) \\
		$h_1$-$h_4$& conditional Gamma constants&$T_{used}$&number of last cycles used for sampling $\theta$\\
		$T_c$& number of discriminator iterations &$M'$&stochastic realizations\\
		&used per generator iteration&&\\
		\bottomrule
	\end{tabular}
	\caption{
	Glossary of hyperparameters used in the comparative study. The hyperparameter $\epsilon$ in HMC without Gibbs sampling is the step size at the initial time, which changes at every step for the first $0.8T$ steps, and then is fixed for the remaining steps. For HMC with Gibbs sampling, $\epsilon$ is fixed in the whole sampling process.
	Furthermore, $\epsilon$ in the Adam optimizer represents the learning rate.}
	\label{tab:app:glossary_hyper}
\end{table}

\begin{table}[H]
	\centering
	\footnotesize
	\begin{tabular}{c|c|c|c|c|c|c|c}
		\toprule
		Method & \multicolumn{7}{c}{Hyperparameter} \\
		\midrule
		\multirow{2}{*}{HMC} & $\epsilon$ & $T$ & burn-in steps  & $M$&&& \\ \cline{2-8}
		& 0.1 & 50 & $2 \times 10^3$ & $ 10^3$ &  && \\
		\midrule
		\multirow{2}{*}{LD} & $\epsilon$ & burn-in steps & M &&&& \\ \cline{2-8}
		& $ 10^{-4}$ & $5 \times 10^3$ & $ 10^3$ &  &  & & \\
		\midrule
		\multirow{2}{*}{MFVI} & $\epsilon$ & train steps & $M$ & init (mean) & init (std) & &\\ \cline{2-8}
		& $ 10^{-3}$ & 4,450 & $ 10^3$ & Xavier (normal) & $\log(1 + \exp(-6))$ &&  \\
		\midrule
		\multirow{2}{*}{MCD} & $\epsilon$&train steps& $M$& dr &init&&\\ \cline{2-8}
		& $10^{-3}$ & $2\times10^{4}$ & $10^{3}$ &$5\times10^{-2}$  &Xavier (normal)  &&  \\
		\midrule
		\multirow{2}{*}{LA} & $\epsilon_{MAP}$ & train steps (MAP) & $M$ &   init &  & &\\ \cline{2-8}
		& $ 10^{-2}$ & $3 \times 10^{4}$ & $ 10^{3}$ & Xavier   & &  &  \\
		\midrule
		\multirow{2}{*}{DEns} & $\epsilon$&train steps& $M$& wd& init&& \\ \cline{2-8}
		& $10^{-4}$ & $2\times10^{4}$ &10  & $5\times10^{-4}$ &Xavier (normal)  & & \\
		\midrule
		\multirow{2}{*}{SEns} & $\epsilon_{init}$ & train steps & $M$ & $\epsilon_{final}$&$T_{cycles}$&$T_{used}$&init \\ \cline{2-8}
		& $10^{-2}$& $2\times10^{4}$ & 20 & $10^{-4}$ &20  &20 & Xavier (normal) \\
		\midrule
		\multirow{2}{*}{SWAG} & $\epsilon_{init}$ & train steps & $M$ & $\epsilon_{final}$&$T_{cycles}$&$T_{used}$&init \\ \cline{2-8}
		& $10^{-2}$ & $2\times10^{4}$ & 50 & $10^{-4}$ & 10 & 5 & Xavier (normal) \\
		\bottomrule
	\end{tabular}
	\caption{
		 Section~\ref{sec:comp:func:homosc:kno}: hyperparameters used for implementing each technique. Note that (1) init (mean) and init (std) for the MFVI represent the initializations for the mean and standard deviation of the posterior distribution, respectively, which are different, and (2) the batch size in the MFVI is 32. The specified MFVI initializations and batch size are used throughout the whole study and are not repeated in the rest of the hyperparameter tables for simplicity. 
	}
	\label{tab:comp:func:homosc:kno}
\end{table}

\begin{table}[H]
	\centering
	\footnotesize
	\begin{tabular}{c|c|c|c|c|c|c}
		\toprule
		 $(N, \sigma_u)$&
		$(32, 0.1)$ & $(32, 0.3)$ & $(32, 0.5)$ & $(16, 0.1)$ & $(16, 0.3)$ & $(16, 0.5)$ \\
		\midrule
		 Dropout rate&
		0.05 & 0.2& 0.3& 0.1& 0.2& 0.3\\
		\bottomrule
	\end{tabular}
	\caption{
	Dropout rates used in the epistemic uncertainty experiment of Fig.~\ref{fig:comp:func:homosc:kno:epist:mcd}.
	}
	\label{tab:}
\end{table}

\begin{table}[H]
	\centering
	\footnotesize
	\begin{tabular}{c|c|c|c|c|c|c}
		\toprule
		Method & \multicolumn{6}{c}{Hyperparameter} \\
		\midrule
		\multirow{2}{*}{HMC} & $\epsilon$ & $T$ & burn-in steps & $M$ & $(h_1, h_2)$ & $(h_3, h_4)$  \\ \cline{2-7}
		& $2.5 \times 10^{-3}$ & 50 & $10^3$ & $10^3$ & (2, 0.25) & (2, 10)   \\
		\midrule
		\multirow{2}{*}{MFVI} & $\epsilon$ & train steps & $M$ & &&\\ \cline{2-7}
		& $ 10^{-3}$ & 22,170 & $ 10^{3}$ &  &  &\\
		\midrule
		\multirow{2}{*}{LA} & $\epsilon_{MAP}$ & train steps (MAP) & init & $\epsilon_{Prior}$ & train steps (Prior) & $M$ \\ \cline{2-7}
		& $ 10^{-2}$ & $3 \times 10^{4}$ & Xavier (normal) & $ 10^{-3}$ & $2 \times 10^3$  & $ 10^4$  \\
		\bottomrule
	\end{tabular}
	\caption{
		 Section~\ref{sec:comp:func:homosc:unk}: hyperparameters used for implementing each technique. $\epsilon_{MAP}$: learning rate for the MAP training; $\epsilon_{Prior}$: learning rate for the prior/noise scale optimization.
	}
	\label{tab:}
\end{table}

\begin{table}[H]
	\centering
	\footnotesize
	\begin{tabular}{c|c|c|c|c|c}
		\toprule
		Method & \multicolumn{5}{c}{Hyperparameter} \\
		\midrule
		\multirow{2}{*}{h-HMC} & $\epsilon$ & $T$ & burn-in steps & $M$ & \\ \cline{2-6}
		& 0.1 & 50 & $2 \times 10^{3}$ & $ 10^{3}$ &    \\
		\midrule
		\multirow{2}{*}{h-MFVI} & $\epsilon$ & train steps & $M$ &&\\ \cline{2-6}
		& $10^{-3}$ & 210 & $10^3$ &  &   \\
		\midrule
		\multirow{2}{*}{h-DEns} & $\epsilon$&train steps& $M$& wd&init \\ \cline{2-6}
		& $10^{-3}$ & $2 \times 10^4$ &10  & $10^{-2}$ & Xavier (normal) \\
		\bottomrule
	\end{tabular}
	\caption{
		 Section~\ref{sec:comp:func:hetero}: hyperparameters used for implementing each technique.
	}
	\label{tab:}
\end{table}

\begin{table}[H]
	\centering
	\footnotesize
	\begin{tabular}{c|c|c|c|c}
		\toprule
		Method & \multicolumn{4}{c}{Hyperparameter} \\
		\midrule
		\multirow{2}{*}{h-HMC} & $\epsilon$ & $T$ & burn-in steps & $M$  \\ \cline{2-5}
		& 0.1 & 50 & $2 \times 10^3$ & $ 10^3$   \\
		\midrule
		\multirow{2}{*}{h-MFVI} & $\epsilon$ & train steps & $M$ &\\ \cline{2-5}
		& $ 10^{-3}$ & $2,150$ & $ 10^3$ &  \\
		\bottomrule
	\end{tabular}
	\caption{\label{tab:comp:func:results:student}
		 Section~\ref{app:comp:func:results:student}: hyperparameters used for implementing each technique.
	}
	\label{tab:}
\end{table}

\begin{table}[H]
	\centering
	\footnotesize
	\begin{tabular}{c|c|c|c|c|c|c}
		\toprule
		Method & \multicolumn{6}{c}{Hyperparameter} \\
		\midrule
		U-PINN & $\epsilon$ & $T$ & burn-in steps & $M$ & $(h_1, h_2)$ & $(h_3, h_4)$  \\ \cline{2-7}
		with HMC& $10^{-2}$ & 50 & $5 \times 10^3$ & $10^3$ & $(25, 4\times 10^{-3})$ & $(0.5, 40)$   \\
		\midrule
		U-PINN & $\epsilon$  & train steps & $M$ &&&\\ \cline{2-7}
		with MFVI& $10^{-3}$  & $10^4$ & $10^3$ &&& \\
		\midrule
		U-PINN & $\epsilon$ & train steps & $M$ &dr&init \\ \cline{2-7}
		with MCD& $10^{-3}$ & $2 \times 10^4$ & $10^3$ & 0.05 & Xavier (normal)   \\
		\midrule
		U-PINN & $\epsilon$ & train steps & $M$ & wd & init \\ \cline{2-7}
		with DEns & $10^{-3}$ & $5 \times 10^3$ & 10 & 0.1 & Xavier (normal)  \\
		\midrule
		PI-GAN-FP & $\epsilon$ & $T$ & burn-in steps & $M$ &&\\ \cline{2-7}
		with HMC & 0.1 & 50 & $2 \times 10^3$ & $ 10^3$&&  \\
		\bottomrule
	\end{tabular}
	\caption{
		Section~\ref{sec:comp:pinns:stand}: hyperparameters used for implementing each technique.
	}
	\label{tab:app:comp:pinns:stand}
\end{table}

\begin{table}[H]
	\centering
	\footnotesize
	\begin{tabular}{c|c|c|c|c}
		\toprule
		Method & \multicolumn{4}{c}{Hyperparameter} \\
		\midrule
		U-PINN with h-HMC & $\epsilon$ &$T$ &burn-in steps & $M$  \\ \cline{2-5}
		in \S \ref{sec:comp:pinns:hetero} & $10^{-3}$ & 30 & $5 \times 10^{-3}$ & $10^3$   \\
		\midrule
		U-PINN with HMC & $\epsilon$ & $T$ & burn-in steps & $M$ \\ \cline{2-5}
		in \S \ref{app:comp:pinns:results:large} (large noise)& $10^-2$ & 50 & $5 \times 10^3$ & $10^3$   \\
		\midrule
		U-PINN with HMC & $\epsilon$ & $T$ & burn-in steps & $M$ \\ \cline{2-5}
		in \S \ref{app:comp:pinns:results:extra} (extrapolation)& $10^{-3}$ & 50 & $5 \times 10^3$ & $10^3$   \\
		\midrule
		U-PINN with HMC & $\epsilon$ & $T$ & burn-in steps & $M$ \\ \cline{2-5}
		in \S \ref{app:comp:pinns:results:steep} (steep boundaries) & $10^{-3}$ & 50 & $5 \times 10^3 $ & $10^3$   \\
		\bottomrule
	\end{tabular}
	\caption{
		Sections~\ref{sec:comp:pinns:hetero} and \ref{app:comp:pinns:results:large}-\ref{app:comp:pinns:results:steep}: hyperparameters used for implementing each technique.
	}
	\label{tab:app:comp:pinns:additional}
\end{table}

\begin{table}[H]
	\centering
	\footnotesize
	\begin{tabular}{c|c|c|c|c|c|c}
		\toprule
		Method & \multicolumn{6}{c}{Hyperparameter} \\
		\midrule
		U-PI-GAN & $\epsilon$ & $T_c$ & train steps &$M$ &$M'$ \\ \cline{2-7}
		with DEns& $10^{-3}$ & 5 & $10^5$ & $10^3$    \\
		\midrule
		U-NNPC & $\epsilon$ & train steps & $M$ &$M'$&wd& init\\ \cline{2-7}
		with DEns& $2\times 10^{-3}$ &  10,001 & 3 & 500 & $ 10^{-3}$ &Xavier (normal)  \\
		\midrule
		U-NNPC+ & $\epsilon$ &  train steps & $M$ &$M'$&wd& init \\ \cline{2-7}
		with DEns& $2\times 10^{-3}$ &  10,001 & 3 & 500 & $ 10^{-3}$ &Xavier (normal)  \\
		\bottomrule
	\end{tabular}
	\caption{
		Section~\ref{sec:comp:stochastic}: hyperparameters used for implementing each technique.
	}
	\label{tab:app:comp:stochastic}
\end{table}

\begin{table}[H]
	\centering
	\footnotesize
	\begin{tabular}{c|c|c|c|c}
		\toprule
		Method & \multicolumn{4}{c}{Hyperparameter} \\
		\midrule
		U-PINN & $\epsilon$ & $T$ & burn-in steps & $M$ \\ \cline{2-5}
		with HMC& 0.1 & 20 & $2 \times 10^{3}$ & $ 10^3$   \\
		\midrule
		\multirow{2}{*}{PI-GAN} & $\epsilon$ & train steps & $M$ &init \\ \cline{2-5}
		& $10^{-4}$ & $10^5$ & $10^3$ & Xavier (normal)  \\
		\bottomrule
	\end{tabular}
	\caption{
		Section~\ref{sec:comp:pinns:forw}: hyperparameters used for implementing each technique.
	}
	\label{tab:app:comp:pinns:forw}
\end{table}

\begin{table}[H]
	\centering
	\footnotesize
	\begin{tabular}{c|c|c|c|c}
		\toprule
		Method & \multicolumn{4}{c}{Hyperparameter} \\
		\midrule
		PA-BNN-FP & $\epsilon$ & $T$ & burn-in steps & $M$  \\ \cline{2-5}
		with HMC& 0.1 & 10 & $3 \times 10^3$ & $ 10^3$   \\
		\midrule
		PA-GAN-FP  & $\epsilon$ & $T$ & burn-in steps & $M$ \\ \cline{2-5}
		with HMC & 0.1 & 10 & $5 \times 10^3$ & $10^3$   \\
		\bottomrule
	\end{tabular}
	\caption{
		Section~\ref{sec:comp:don:fp}: hyperparameters used for implementing each technique.
	}
	\label{tab:app:comp:don:fp}
\end{table}

\begin{table}[H]
	\centering
	\footnotesize
	\begin{tabular}{c|c|c|c|c}
		\toprule
		Method & \multicolumn{4}{c}{Hyperparameter} \\
		\midrule
		U-DeepONet & $\epsilon$ & train steps & $M$ &init  \\ \cline{2-5}
		with DEns& $10^{-3}$ & $5 \times 10^4$ & 5 & Xavier (normal)   \\
		\bottomrule
	\end{tabular}
	\caption{
		Section~\ref{sec:comp:don:dens}: hyperparameters used for implementing each technique.
	}
	\label{tab:app:comp:don:dens}
\end{table}

%% file: appendix/IN_app_architecture.tex
\subsection{Neural network architectures used in the comparative study}
\label{app:comp:architecture}

In the function approximation example of Section \ref{sec:comp:func}, we have compared more than ten different UQ methods and all the employed NNs use the same architecture except for GANs for functional priors (h-HMC+FP - details in the end of the section).
That is, (1) 2 hidden layers with 50 neurons per layer, and (2) hyperbolic tangent function as the activation function. In addition, the prior distribution for each weight/bias of the BNNs is identical, which is a standard normal distribution. 

In the mixed PDE problem of Section~\ref{sec:comp:pinns}, we have compared six different UQ methods (including h-HMC). In each approach except for PI-GANs for functional priors (PI-GAN-FP - details in the end of the section), we employ two NNs as surrogates for the solution $u$ to Eq. \eqref{eq:comp:pinns:stand:pde}/\eqref{eq:comp:pinns:pde:steep} as well as the reaction rate $\lambda$. 
The NNs have the same architecture, i.e., (1) 2 hidden layers with 50 neurons per layer, and (2) hyperbolic tangent function as the activation function. Similarly to function approximation, the prior distribution for each weight/bias of the BNNs is identical, which is a standard normal distribution. The only difference in these two NNs lies in the input dimension, which for the former is 2 (because of space and time), whereas for the latter is 1 (space only).

In the forward PDE problem of Section \ref{sec:comp:pinns:forw}, we use a BNN as the surrogate for the solution $u$ to Eq. \eqref{eq:comp:pinns:forw:pde}, which has the same architecture as well as prior distribution as the BNNs employed in Section \ref{sec:comp:func}. In addition, we also employ the same architecture/priors for the BNNs used in the operator learning problem of Section \ref{sec:comp:don}. 
Finally, we present the architectures for the GANs/PI-GANs, NNPC, and neural operators in Tables \ref{tab:in_app_gan_arch}, \ref{tab:in_app_nnpc_arch}, and \ref{tab:in_app_deeponet_arch}, respectively.  Note that in Sections \ref{sec:comp:func:homosc:unk}, \ref{sec:comp:pinns:stand}, and \ref{sec:comp:don:fp}, we employ two NNs for the generator $G$, which take as inputs the Gaussian noise (the first architecture for $G$ in Table \ref{tab:in_app_gan_arch}) and coordinates (the second architecture for $G$ in Table \ref{tab:in_app_gan_arch}), respectively. 
This is done to enhance the computational efficiency in posterior estimation (see details in \cite{meng2021learning}). In Sections \ref{sec:comp:stochastic} and \ref{sec:comp:pinns:forw}, we use one NN for the generator $G$ (see details in \cite{meng2021learning,yang2020physicsinformed}), since we do not compute the posterior in these two cases. In addition, for the heteroscedastic noise case in Section \ref{sec:comp:func:homosc:unk} we have one more NN to approximate the varying noise scale, which takes $x$ as input, and outputs the noise scale at $x$. For this NN, we use 2 hidden layers with 20 neurons per layer, and the activation function is ReLu. 

\begin{table}[H]
\centering
{\footnotesize
\begin{tabular}{c|cc|cc|c|c}
\hline 
  & \multicolumn{2}{c|}{Generator $G$}  & \multicolumn{2}{c|}{Discriminator $D$} & \multirow{2}{*}{$K$} &
  \multirow{2}{*}{Training steps}  \\
  & width $\times$ depth  & Activation & width $\times$ depth & Activation &\\
  \hline
  
  {Section \ref{sec:comp:func:homosc:unk}} & $128 \times 2 / 128 \times 2$   & tanh/tanh & $128 \times 3$ & Leaky ReLu &   20  & 100,000 \\
  \hline
  {Section \ref{sec:comp:pinns:stand}} & $128 \times 2 / 128 \times 2$   & tanh/tanh & $256 \times 3$ & Leaky ReLu &  40 & 100,000 \\
  \hline
    {Section \ref{sec:comp:pinns:forw}} & $128 \times 2$   & tanh & $128 \times 3$ & Leaky ReLu &   40 &100,000 \\
 
    \hline
{Section \ref{sec:comp:stochastic}} & $128 \times 2$    & tanh & $128 \times 3$ & Leaky ReLu &   20 &100,000 \\
  \hline
  {Section \ref{sec:comp:don:fp}} & $128 \times 2 / 128 \times 2$   & tanh/tanh & $256 \times 3$ & Leaky ReLu &  100 &100,000 \\
  \hline 
\end{tabular}
}
\caption{
Architecture and training steps of GANs/PI-GANs in Section \ref{sec:comp}. The width and depth correspond to the hidden layers. $K$ is the dimension of $\theta$, which is the same as the output dimension of each NN in $G$ if we use two NNs for $G$. Note that (1) in Sections \ref{sec:comp:func:homosc:unk}, \ref{sec:comp:pinns:stand}, and \ref{sec:comp:don:fp}, we employ two NNs for $G$, which take as inputs the Gaussian noise (the first architecture for $G$ in this table) and coordinates (the second architecture for $G$ in this table), respectively.
This is done to enhance the computational efficiency in posterior estimation (see details in \cite{meng2021learning}).
Further, (2) in Section \ref{sec:comp:pinns:stand}, the architectures for $u$ and $\lambda$ are the same.}
\label{tab:in_app_gan_arch}
\end{table}

\begin{table}[H]
\centering
{\footnotesize
\begin{tabular}{c|cc|cc|cc|c}
\hline 
  & \multicolumn{2}{c|}{NN for $u$}  & \multicolumn{2}{c|}{NN for $\lambda$} & \multicolumn{2}{c|}{NN for $f'$ (NNPC+ only)} & Training  \\
  & width $\times$ depth  & Activation & width $\times$ depth & Activation  & width $\times$ depth  & Activation & steps\\
  \hline
  
    {Section \ref{sec:comp:stochastic}} & $4 \times 2 $ and $4\times 32$   & tanh &$4 \times 2 $ and $2\times 32$  & tanh & $4 \times 2 $ and $2\times 32$   & tanh &  10,001\\
  \hline 
\end{tabular}
}
\caption{
Architectures for NNPC and NNPC+ in Section \ref{sec:comp:stochastic}. 
For each of $u$ and $\lambda$, we use a NN for the mean and another one for the modes, e.g., $4 \times 2 $ and $4\times 32$, respectively; see \cite{zhang2019quantifying} for more implementation details.
}
\label{tab:in_app_nnpc_arch}
\end{table}

\begin{table}[H]
\centering
{\footnotesize
\begin{tabular}{c|cc|cc|c}
\hline 
  & \multicolumn{2}{c|}{Branch Net}  & \multicolumn{2}{c|}{Trunk Net} & \multirow{2}{*}{Training steps}  \\
  & width $\times$ depth  & Activation & width $\times$ depth & Activation &\\
  \hline
{Section \ref{sec:comp:don:fp}} & $128 \times 2 $   & tanh & $128 \times 3$ & tanh &  50,000 \\
  \hline 
  {Section \ref{sec:comp:don:dens}} & CNN   & ReLu  & $128 \times 3$ & tanh &  30,000 \\
  \hline 
\end{tabular}
}
\caption{
Architecture and training steps of DeepONets in Section \ref{sec:comp:don}. 
}
\label{tab:in_app_deeponet_arch}
\end{table}